\documentclass{article}
\usepackage[utf8]{inputenc}

\title{Feature Tracks are not Zero-Mean Gaussian}
\author{Stephanie Tsuei, Wenjie Mo, Stefano Soatto}
\date{October 2022}

\usepackage{graphicx}
\usepackage{amsmath}
\usepackage{amssymb}
\usepackage{booktabs}
\usepackage{xcolor}
\usepackage{multirow}
\usepackage{bbm}
\usepackage{subfigure}
\usepackage{float}

\usepackage[pagebackref,breaklinks,colorlinks]{hyperref}

\usepackage{fullpage}

\begin{document}

\maketitle

\begin{abstract}
In state estimation algorithms that use feature tracks as input, it is customary to assume that the errors in feature track positions are zero-mean Gaussian. Using a combination of calibrated camera intrinsics, ground-truth camera pose, and depth images, it is possible to compute ground-truth positions for feature tracks extracted using an image processing algorithm. We find that feature track errors are not zero-mean Gaussian and that the distribution of errors is conditional on the type of motion, the speed of motion, and the image processing algorithm used to extract the tracks.
\end{abstract}

\section{Introduction}

Many state estimation algorithms assume that measurements are zero-mean Gaussian. This is an explicit assumption in the Kalman Filter and its nonlinear variants \cite{thrun_probabilistic_2005, barrau_invariant_2018} and implicitly built-into the optimization problem of bundle adjustment algorithms \cite{mur-artal_orb-slam:_2015} and outlier-rejection algorithms \cite{civera_1-point_2009}. With extensive calibration with respect to temperature and mechanical alignment, the zero-mean Gaussian assumption is sufficient for the measurements of sensors such as inertial measurement units (IMUs) \cite{vectornav_imu_calibration, tedaldi_robust_2014}, even if it is still not completely true: Extended Kalman Filters (EKFs) that rely on these IMUs are deployed on safety-critical systems actively in use.

Even though several well-known algorithms for Simultaneous Localization and Mapping (SLAM) rely on the often-deployed EKF (e.g. \cite{jones_visual-inertial_2011,Geneva2020ICRA,bloesch_iterated_2017}), SLAM is still an active area of research. The existence of recently-released and actively used research benchmark datasets \cite{hilti_benchmark, tartanair2020iros} indicate that the robotics and computer vision communities still believe that performance of SLAM and an understanding of its failure cases are still insufficient, even after three decades of development \cite{early_slam_tutorial}. This motivates an examination into the fundamental assumptions of SLAM.

This manuscript visits the assumption that feature tracks, the ``measurements" of any indirect visual SLAM algorithm, contain only zero-mean Gaussian error. The covariance of the feature tracks is typically a tuning parameter to for all features at all times. We show that the feature track errors are not zero-mean Gaussian and furthermore, that the errors are conditional on the type of motion, the speed of motion, and the type of feature tracker used to extract the feature tracks. To our knowledge, this is the first study of the mean and covariance of feature tracks \emph{conditional} on the factors that affect them.

The organization of the paper is as follows. Section \ref{sec:feature_track_uq} details the methods. Section \ref{sec:feature_tracker_experiment_details} presents some key figures, and summarizes the error distribution of feature trackers. Section \ref{sec:discussion} ends with some concluding remarks. Additional figures from the experiment are given in the Appendix.

\subsection{Related Work}

\paragraph{Performance of feature detectors and descriptors conditional on nuisances.} The main metric used to evaluate feature detectors is \emph{repeatability} \cite{mikolajczyk_comparison_2005}, or the probability that a feature detector will detect the same feature across multiple images of the same scene under different illuminations and viewpoints. Other metrics are \emph{entropy} \cite{heinly_comparative_2012}, the spread of detected features over an image, and \emph{recall} \cite{aanaes_interesting_2012}, the number of features that are likely ``matchable'' to features in another image of the same scene. On the other hand, the primary metrics used to evaluate feature descriptors are \emph{precision} and \emph{recall}, calculated using pairs of ``matches'' that are found using the descriptor \cite{mikolajczyk_performance_2005}. The evaluation of feature detectors requires multiple images of the same scene. The evaluation of descriptors originally used the same datasets as the evaluation of detectors. To disentangle the problem of detecting features from the evaluation of feature description, two comprehensive datasets of image patches was released in 2017 \cite{balntas_hpatches_2017, maier_ground_2017}. At around the same time, \cite{schonberger_comparative_2017} evaluated both learned and handcrafted feature detectors and descriptors. Of most interest to us are \cite{heinly_comparative_2012}, which used a small dataset containing pure rotation, pure scaling, and illumination changes to evaluate the performance of various detector/descriptor combinations condition on each, \cite{zhao_image_2020}, which extended the datasets used in \cite{heinly_comparative_2012}, and \cite{aanaes_interesting_2012}, which evaluated the performance of feature detectors conditional on change in view angle and lighting condition. Tangentially interesting are \cite{hauagge_image_2012}, which released a dataset of image pairs that are geometrically consistent, but contain large changes in style (e.g. summer vs. winter) and lighting; and \cite{sattler_benchmarking_2018}, which contains groups of image sequences with similar motions, but large outdoor illumination changes.

\paragraph{Learning or Fitting a Covariance Matrix to Feature Tracks.} Early works sought to compute covariance of feature location using information in the RGB image. \cite{kanazawa_we_2001} approximated the covariance with the Hessian of the image centered at the feature point was the covariance of a detected feature -- the idea is that the sharper the curvature given by the Hessian, the more likely a convolutional filter will find the correct location of the feature. \cite{nickels_estimating_2002} contains a sum-of-squared-differences formula for computing feature track covariance. \cite{zeisl_estimation_2009} contains a formula for computing the covariance matrix of SIFT and SURF features. Later on, \cite{sheorey_uncertainty_2015} and \cite{wong_uncertainty_2017} present two methods to model the mean and covariance of Lucas-Kanade feature tracks. With the exception  of \cite{sheorey_uncertainty_2015}, which assumes that the location of a feature track could be a Gaussian Mixture Model, all other models assume that uncertainty is zero-mean Gaussian.

\section{Method}
\label{sec:feature_track_uq}

We wish to characterize the dependence of \textbf{mean error}, \textbf{mean absolute error}, \textbf{covariance}, \textbf{outlier ratio}, and \textbf{feature lifetime} on motion type, speed, tracker type, and when available, lighting. The types of motion investigated are:
\begin{itemize}
\item \textbf{Sideways motion} -- Linear movement with no rotation.
\item \textbf{Fixating motion} -- Moving in a constant radius around a central object. The camera is always pointed directly at the central object, creating some rotation.
\item \textbf{Forwards motion} -- Driving-like motion. The primary change frame-to-frame is scale. Points near the center of an image will stay near the center in subsequent frames.
\item \textbf{AR/VR motion} -- Movement that consists of mostly rotations around a persistent scene.
\end{itemize}
To vary speed, we skip frames at regular intervals from the image sequences. Nominal speed, or a speed of 1.00, means that all frames are used. A speed of 2.00 means that the feature tracker will only see every other frame, and a speed of 3.00 means that the feature tracker will only see one in every three frames. We do not test speeds below 1.00. The exact speeds tested depends on dataset. Finally, we also investigate the effect of two types of feature trackers:
\begin{itemize}
\item \textbf{Lucas-Kanade Sparse Optical Flow} \cite{lucas_iterative_1981}
\item \textbf{Correspondence Tracker} using the SIFT descriptor \cite{lowe_object_1999}. Although computationally expensive, the SIFT descriptor was chosen because of its availability and its performance when used in state estimation tasks \cite{schonberger_comparative_2017}. The descriptor of a feature track is set at the first frame it is detected and never updated.
\end{itemize}

We have chosen \emph{not} to study lens distortion, since this would require multiple similar datasets collected with different cameras. All images in all datasets either have been preprocessed to remove lens distortions, or simulated without lens distortions. Since the Lucas-Kanade tracker is differential, we also choose not to study a differential correspondence tracker that updates the descriptor of a feature track at every frame.

\subsection{Equations}

Consider a feature $i$ that was first detected at time $t^i_0$. If a depth image is available at time $t^i_0$ and $g_{sc}(t^i_0)$ is known, we may fix the feature's position in the spatial frame, $X_s^i$:
\begin{equation}
\begin{aligned}
    X^i_c(t^i_0) &= \pi^{-1}_K(x_p(t^i_0), Z^i_0) \\
    X^i_s &= g_{sc}(t^i_0) \circ X_c(t^i_0) \\
    \label{eq:fixing_Xs}
\end{aligned}
\end{equation}
In the above equation, $Z^i_c(t^i_0)$ is the third coordinate, or depth, of $X^i_c(t^i_0)$. Once, $X_s^i$ is fixed, we can then calculate the \textbf{``ground-truth feature track"} $\bar x_p^i(t)$:
\begin{equation}
    \bar x^i_p(t) = \pi_K(g_{sc}^{-1}(t) \circ X_s^i).
    \label{eq:gt_tracks}
\end{equation}
Some datasets provide a ground-truth point-cloud generated by a single lidar scan rather than a stream of depth images. A lidar scan is a point cloud with $M \sim 10^7$ points in the lidar frame $L$, which is defined as the camera frame at a particular time $t_L$: $\mathbf P_L = \{ P^0_L, P^1_L, \dots, P^M_L \}$. We can calculate the pixel coordinates of each point $j$ in $\mathbf P_L$: 
\begin{equation}
\pi_K(\mathbf P_L) = \{ \pi_K(P^0_L), \pi_K(P^1_L), \dots, \pi_K(P^M_L) \}
\label{eq:laser_scan_proj}
\end{equation}
Feature tracks visible at time $t_L$ can be associated with the nearest point in $\pi_K(\mathbf P_L)$. Suppose the nearest point in $\pi_K(\mathbf P_L)$ to feature $i$ is $P^j_L$. Then, the ground-truth track of feature $i$ is
\begin{equation}
\begin{aligned}
    X^i_s &= g_{sc}(t_L) \circ P^j_L \\
    \bar x^i_p(t) &= \pi_K(g_{sc}^{-1}(t) \circ X^i_s).
    \label{eq:dtu_px_groundtruth}
\end{aligned}
\end{equation}
Once we have a ground-truth feature track for feature $i$, we can calculate the error signal for that feature:
\begin{equation}
    e^i(t) = x_p^i(t) - \bar x_p^i(t)
    \label{eq:px_error_def}
\end{equation}
where $x_p^i(t)$ is the observed track.

For datasets that provide a ground-truth point cloud at a single frame, the \textbf{mean error at timestep $t$} is
\begin{equation}
    \mu(t) = \frac{1}{M(t)} \sum_{i=1}^{M(t)} e^i(t)
    \label{eq:mean_error_at_time}
\end{equation}
where $M(t)$ is the number of tracked features at time $t$. The \textbf{mean absolute error at timestep $t$}
\begin{equation}
    \kappa(t) = \frac{1}{M(t)} \sum_{i=1}^{M(t)} |e^i(t)|.
    \label{eq:mean_abs_error_at_time}
\end{equation}
Similarly, the \textbf{covariance at timestep $t$} is calculated by
\begin{equation}
    \Sigma(t) = \frac{1}{M(t)-1} \sum_{i=1}^{M(t)} e^i(t) e^i(t)^T.
    \label{eq:cov_at_time}
\end{equation}
It is only possible to compute $\mu(t)$, $\kappa(t)$, and $\Sigma(t)$ for features that are visible at time $t_L$, when the laser scan was acquired.

For datasets that provide a stream of depth images, we use different definitions of mean error, mean absolute error, and covariance. We can also use all features and not just those visible in a particular frame. The \textbf{mean error after $k$ timesteps} is
\begin{equation}
    \nu(k) = \frac{1}{\Psi(k)} \sum_{i=1}^{\Phi(k)} e^i(t^i_0+k\delta_t)
    \label{eq:mean_error_after_timesteps}
\end{equation}
where $\Psi(k)$ is the number of features in the entire dataset tracked for at least $k$ timesteps and $\delta_t$ is the length of each timestep. The \textbf{mean absolute error after $k$ timesteps is:}
\begin{equation}
    \eta(k) = \frac{1}{\Psi(k)} \sum_{i=1}^{\Psi(k)} |e^i(t^i_0+k\delta_t)|
    \label{eq:mean_abs_error_after_timesteps}
\end{equation}
where $\Phi(k)$ is the number of features tracked for at least $k$ timesteps and $\delta_t$ is the length of each timestep. Finally, the \textbf{covariance after $k$ timesteps} is given by
\begin{equation}
    \Phi(k) = \frac{1}{\Psi(k)-1} \sum_{i=1}^{\Psi(k)} e^i(t^i_0+k\delta_t) e^i(t^i_0+k\delta_t)^T.
    \label{eq:cov_after_timesteps}
\end{equation}
When depth data is available at all frames, we define the feature's 3D location at the frame it is first detected and use equations \eqref{eq:mean_error_after_timesteps}, \eqref{eq:mean_abs_error_after_timesteps}, \eqref{eq:cov_after_timesteps}. %

At each frame, a feature tracker will attribute some features in one frame to the features in the previous frame. Let $F(t)$ be the total number of features in the frame at time $t$. The features in each frame will consist of $f_0(t)$ correct attributions, $f_1(t)$ incorrect attributions, and $f_2(t)$ new features, where $f_0(t) + f_1(t) + f_2(t) = F(t)$ and $f_0(t) + f_1(t) \leq F(t-1)$. Outlier rejection algorithms are used to determine $f_0(t)$ and $f_1(t)$ in real-time. The \textbf{outlier ratio} is defined as:
\begin{equation}
\frac{f_1(t)}{F(t-1)}.
\end{equation}

Finally, the \textbf{feature lifetime} of a feature track is the total number of consecutive frames in which it found and successfully attributed. A feature is ``born" at the frame it is first detected and ``dies" if a feature is not found for a single frame.

\section{Experiment Details}
\label{sec:feature_tracker_experiment_details}

\subsection{Feature Tracker Configuration}
\label{sec:feature_tracker_configuration}

We used the feature tracker is the \texttt{Tracker} object integrated with XIVO, our in-house SLAM system. The tracker is configured to use the AGAST corner detector \cite{mair_adaptive_2010}, and to track between 1000 and 1200 features at a time. The AGAST corner detector was chosen for its speed and because it detects a large number of features in most scenes. The feature tracker was configured to track up to 1200 features per scene. We use RANSAC with $p=0.995$ and an error threshold of 3 pixels to reject outliers. More details on the \texttt{Tracker} object and XIVO can be found in Appendix \ref{chapter:about_xivo}.

Since the tracker software was programmed to be part of a larger system and not specifically for these experiments, the implementation of the Correspondence Tracker is not ideal. If a feature is visible in frames 0-5, but is not detected in frame 2, the tracker will drop the feature at frame 2 and initialize a new one in frame 3. This behavior is consistent with the definition of feature lifetime given in the previous section, but is not the ideal implementation for a Correspondence Tracker because there is always a possibility that a corner detector will not find the corner in one frame, or that a descriptor will be just a little too different in one particular frame because of lighting. A more ideal implementation of the Correspondence Tracker would drop frames after a $N_m$ missed frames, where $N_m > 1$ is an experimentally determined number. The definition of feature lifetime would also be changed to accommodate this more complex behavior. As a result of this choice, the distribution of feature lifetimes for the Correspondence Tracker are shorter than they otherwise would be. Furthermore, our experiments will fail to characterize trends that only appear at higher speeds.

\subsection{Dataset-Specific Details}

\paragraph{DTU Point Features Dataset.}

The DTU Point Features Dataset \cite{aanaes_interesting_2012} consists of sixty scenes of fixating motion. In the dataset, one or more objects is placed at the center of stage lit with up to 19 LEDs. A camera is mounted on a robot arm and moved in a precise manner at the stage. At each of 119 fixed locations, the camera acquires an image lit with one of the 19 LEDs, enabling lighting experiments using image-based relighting. The dataset contains a laser scan of the scene at a single frame, called the Key Frame. The original image size is 1600 $\times$ 1200. For speed, we use 800 $\times$ 600 px. grayscale versions of the images instead of the full resolution images.

\begin{figure}
    \centering
    \includegraphics[width=3.2in]{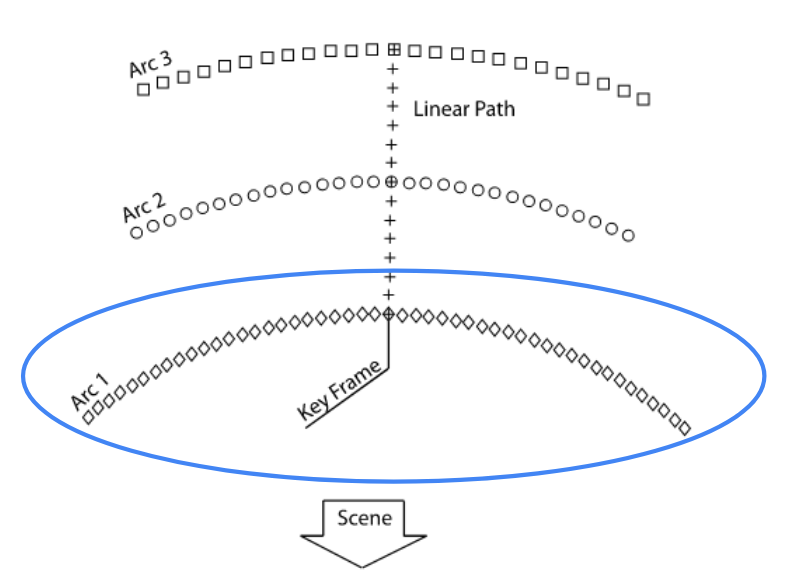}
    \includegraphics[width=3.2in]{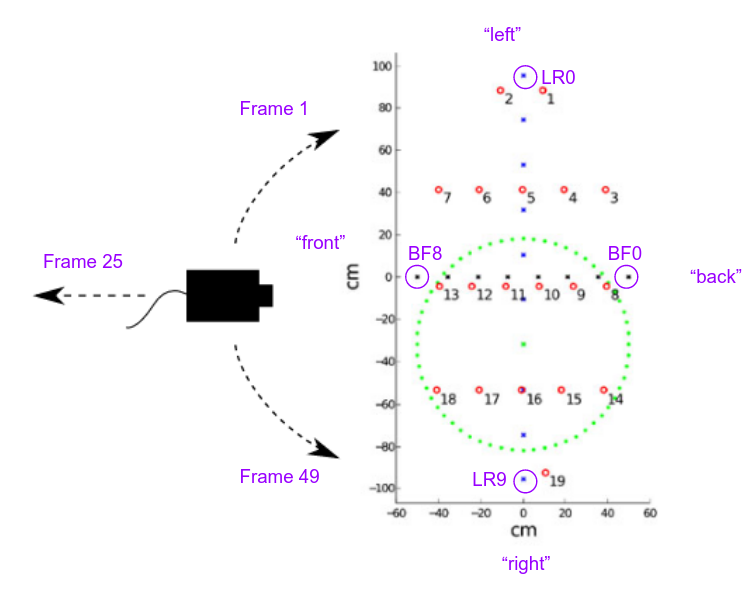}
    \caption{\textbf{An Illustration of the Light Stage Setup in the DTU Point Features Dataset.}  \textbf{Left:} The locations at which images were acquired in the DTU Point Features dataset form three arcs and a linear path. Laser scans of the scenes were collected at the Key Frame (front and center). Frames from Arc 1 (circled in blue) are used for this experiment. \textbf{Right:} Red circles depict the location of 19 physical LEDs used to light the scene, which are spaced out over the scene. At each camera position in the left figure, the authors of the DTU Point Features dataset acquired 19 images. In each image, exactly one of the 19 LEDs is switched on. Acquiring 19 images in each location this way facilitates experiments in lighting using image-based relighting. Diffuse lighting can be simulated by using all 19 photographs from each position equally. More intense directional lighting can be simulated by weighting some LEDs more than others. In our experiments, we vary lighting from back-to-front (BF0-BF7) and left-to-right (LR0-LR9) as the camera follows the motion of Arc 1. Lights LR0 - LR9 and BF0 - BF7 are calculated by using Gaussian-weights on the 19 lights with $\sigma=20$cm; Light LR6 is highlighted in green. Figures are reprinted and annotated with permission.}
    \label{fig:dtu_light_stage}
\end{figure}

We make use of the first 49 frames of each scene, or Arc 1 (see Figure \ref{fig:dtu_light_stage}). The Key Frame is Frame 25. We calculate mean error $\mu$, mean absolute error $\kappa$, and covariance $\Sigma$ using equations \eqref{eq:mean_error_at_time}, \eqref{eq:mean_abs_error_at_time}, and \eqref{eq:cov_at_time}. Since 3D data is only available at the Key Frame, calculation of errors and covariances only includes features that exist in Frame 25. Therefore, there is a bias towards longer tracks, as all short tracks that don't exist in Frame 25 are all tossed out. Since the ``ground-truth" position of each feature in 3D is defined by its position in Frame 25, all results will therefore show that Frame 25 has zero covariance and the lowest errors. Statistics on feature lifetime and outlier rejection, however, do include features that do not exist in Frame 25.

To compute the ground-truth location of a feature track, we must associate a feature track to a point in a laser scan point cloud (eq. \eqref{eq:dtu_px_groundtruth}). Since the point cloud does not cover every pixel in the image, associations between features and laser scan points are only made if the pixel value of the laser scan point (eq. \eqref{eq:laser_scan_proj}) is less than 0.25 pixels from the feature.  Associating a pixel to a laser scan point with the incorrect depth measurement will result in a very large calculated means in equation \eqref{eq:mean_error_at_time}. Even with the low 0.25 pixel threshold, this bad association can still happen around edges and corners of objects. So that our analyses do not include very many of these poor depth associations, we throw out feature tracks whose maximum error is greater than the 90th percentile.

Since the DTU Point Features dataset was designed to enable image-based relighting, we also investigated the effects of directional light in addition to speed and the tracker used. We tested the same directional lights as  \cite{aanaes_interesting_2012}. The position of each directional light is shown in Figure \ref{fig:dtu_light_stage}.

\paragraph{KITTI Vision Suite.} The raw data \cite{Geiger2012CVPR} in the KITTI Vision Suite consists RGB, GPS, IMU, and Lidar data captured from a moving vehicle. The motion captured in the images is predominantly forwards. The Lidar data was then processed into a separate benchmark dataset of depth images for single-image depth prediction and depth completion \cite{uhrig_sparsity_2017}. We make use stream \texttt{Image02}. Sequences containing ``still frames" (e.g. significant amount of waiting at a traffic light), are excluded. Excluding sequences containing still frames leaves 28 scenes for our experiments. Although this is fewer scenes than the DTU dataset, it is still more frames because most sequences are longer than 49 frames.

\begin{figure}[H]
    \centering
    \subfigure[Lucas-Kanade]{\includegraphics[width=0.48\textwidth]{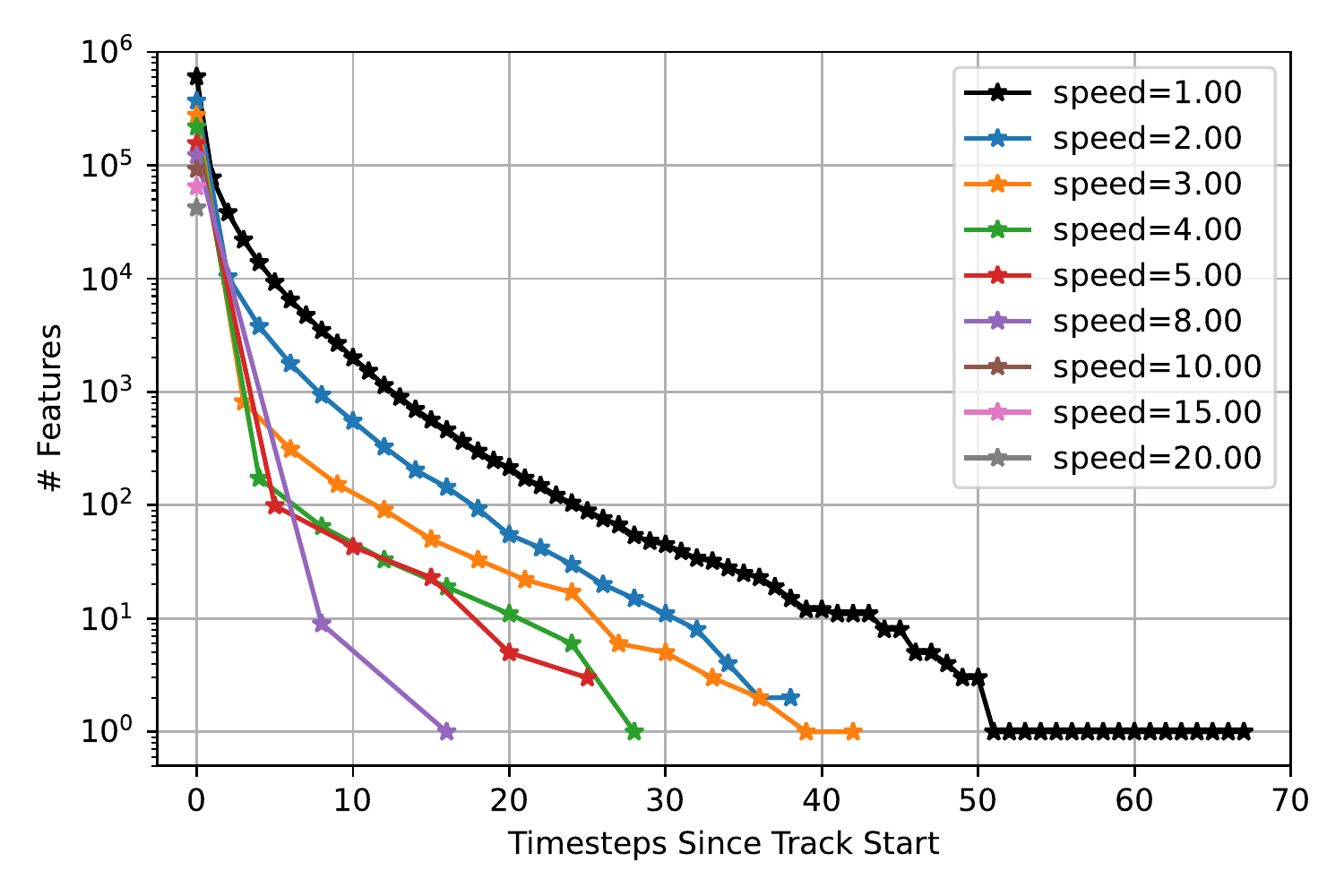}}
    \subfigure[Correspondence]{\includegraphics[width=0.48\textwidth]{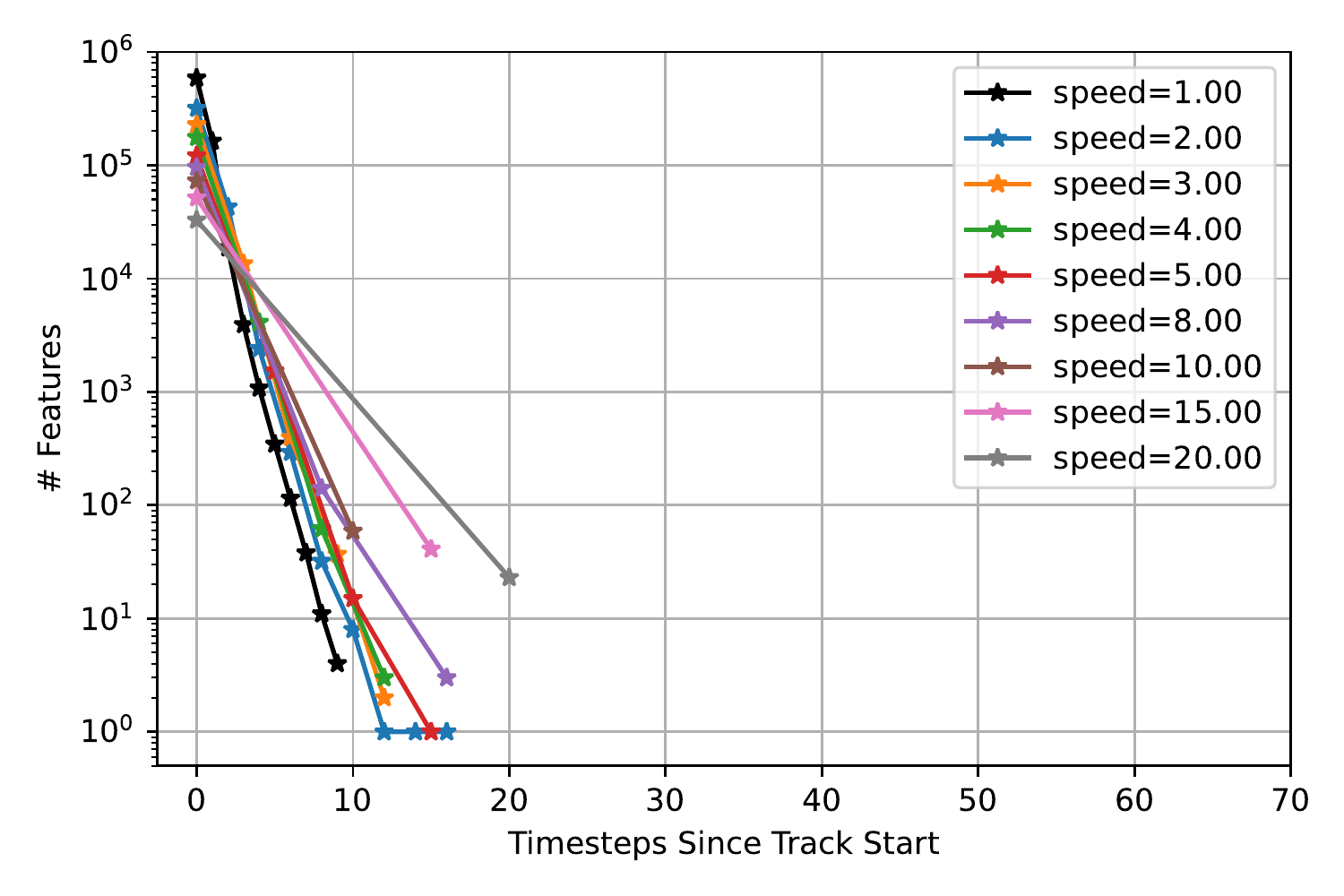}}
    \caption{Feature lifetime is plotted on the horizontal axis. The vertical axis, in log scale, shows the number of features in all 28 scenes that were tracked for at least that many frames. In both plots the number of features drops very fast. Note that for speeds greater than 8.00, the Lucas-Kanade tracker fails to match any features past one frame. \textbf{In subsequent analyses on the KITTI dataset, we only compute mean errors and covariances at timesteps with at least 100 features. We also only analyze speeds 1.00, 2.00, and 3.00 because higher speeds would otherwise be limited to $\le$ two timesteps.}}
    \label{fig:kitti_avg_feats}
\end{figure}

Since 3D data is available at every frame, we define a feature's 3D position using the depth image from the very first frame where it was detected. Therefore, we use mean error $\nu$ (eq. \eqref{eq:mean_error_after_timesteps}), absolute error $\eta$ (eq. \eqref{eq:mean_abs_error_after_timesteps}), and covariance $\Phi$ (eq. \eqref{eq:cov_after_timesteps}). To avoid errors due to bad depth measurements, we throw out the tracks whose maximum L2 error are above the 90th percentile and only calculate $\nu$, $\eta$, and $\Phi$ at timesteps where there are at least 100 features (see Fig. \ref{fig:kitti_avg_feats}).

\begin{figure}
\centering
\includegraphics[width=0.48\textwidth]{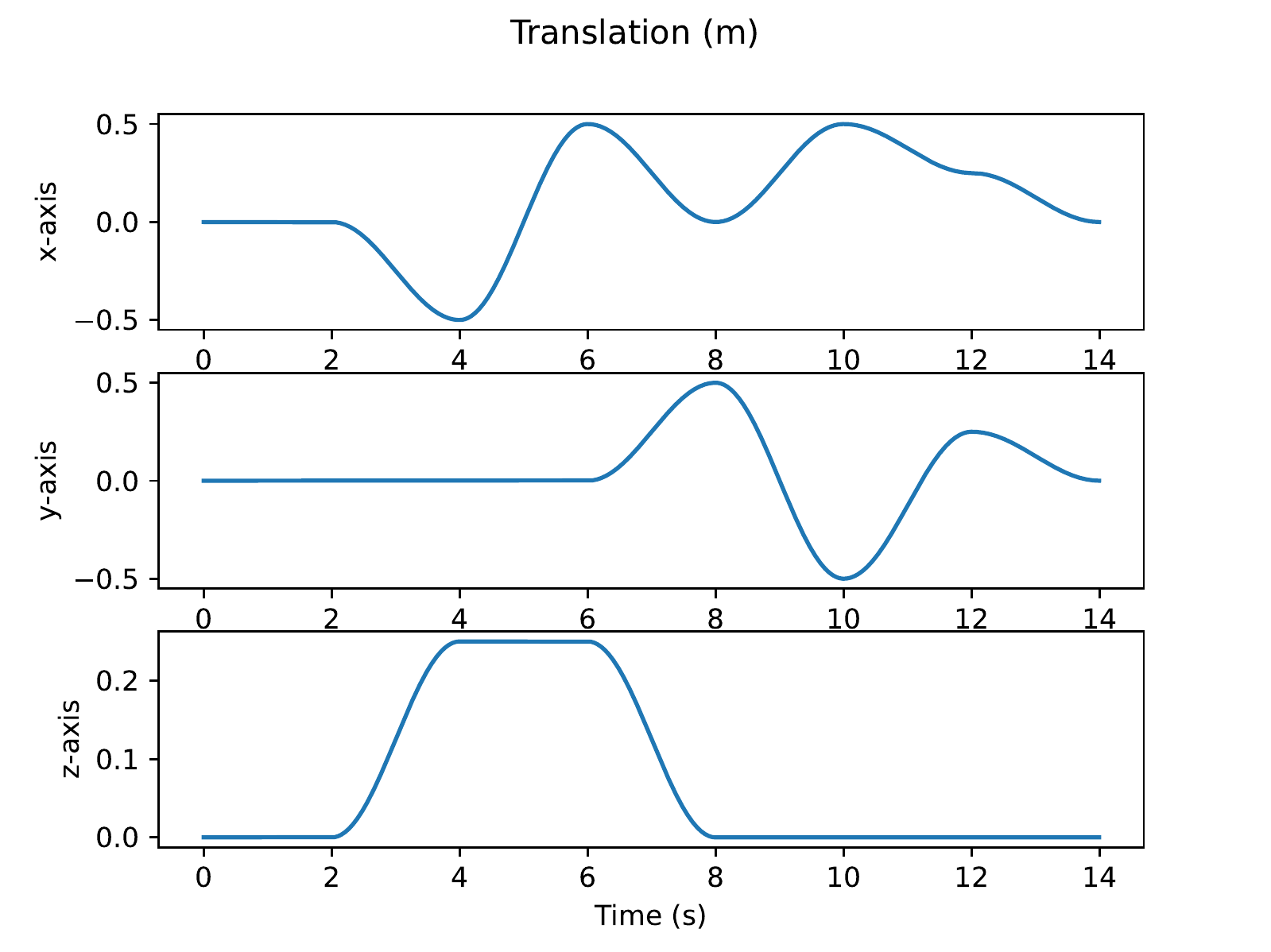}
\includegraphics[width=0.48\textwidth]{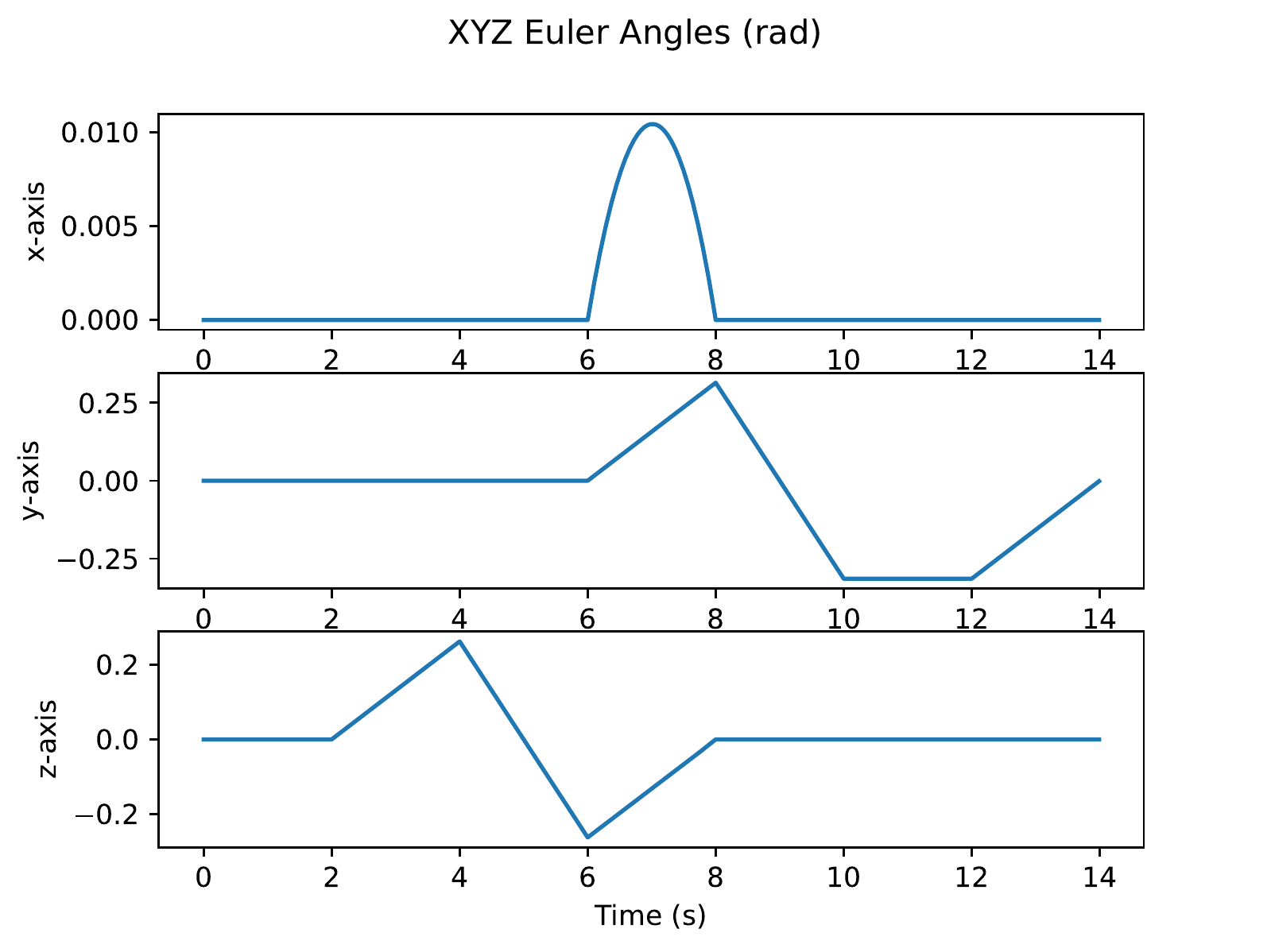}
\caption{\textbf{The trajectory generated for the AR/VR scenario.} The commanded trajectory used to collected the AR/VR data was generated from the translation (left) and rotation (right) plotted above. Translation is generated point-to-point using haversines and rotation is generated from slerping.}
\label{fig:gazebo_arvr_traj}
\end{figure}

\paragraph{Simulated Supplementary Data.} For AR/VR motions and sideways motions, we collected simulated RGB-D data in Gazebo. The simulation consisted of a Microsoft Kinect, modified so that RGB and depth data would be co-located, mounted on a Hector quadrotor \cite{hector_quadrotor} in ROS Melodic. The scene consisted of large objects from the Open Source Robotics Foundation's Gazebo Model Library. Images have a resolution of 800 $\times$ 600 pixels. In the subsequent sections, we refer to these datasets as ``Gazebo Linear" and ``Gazebo AR/VR". The AR/VR trajectory used to collect data is shown in Figure \ref{fig:gazebo_arvr_traj}.

\begin{figure}[H]
\centering
\includegraphics[width=0.48\textwidth]{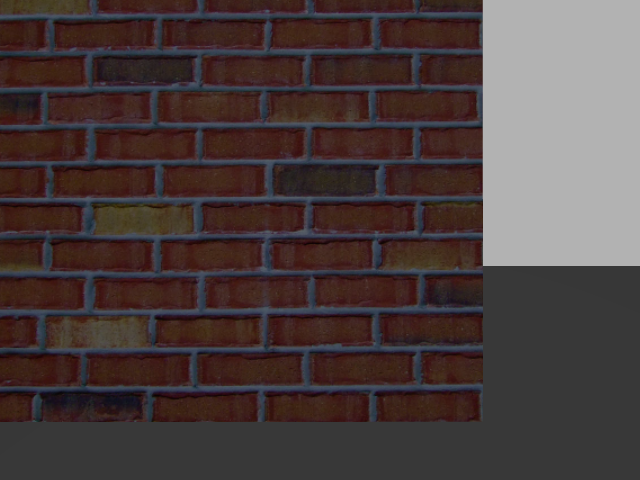}
\includegraphics[width=0.48\textwidth]{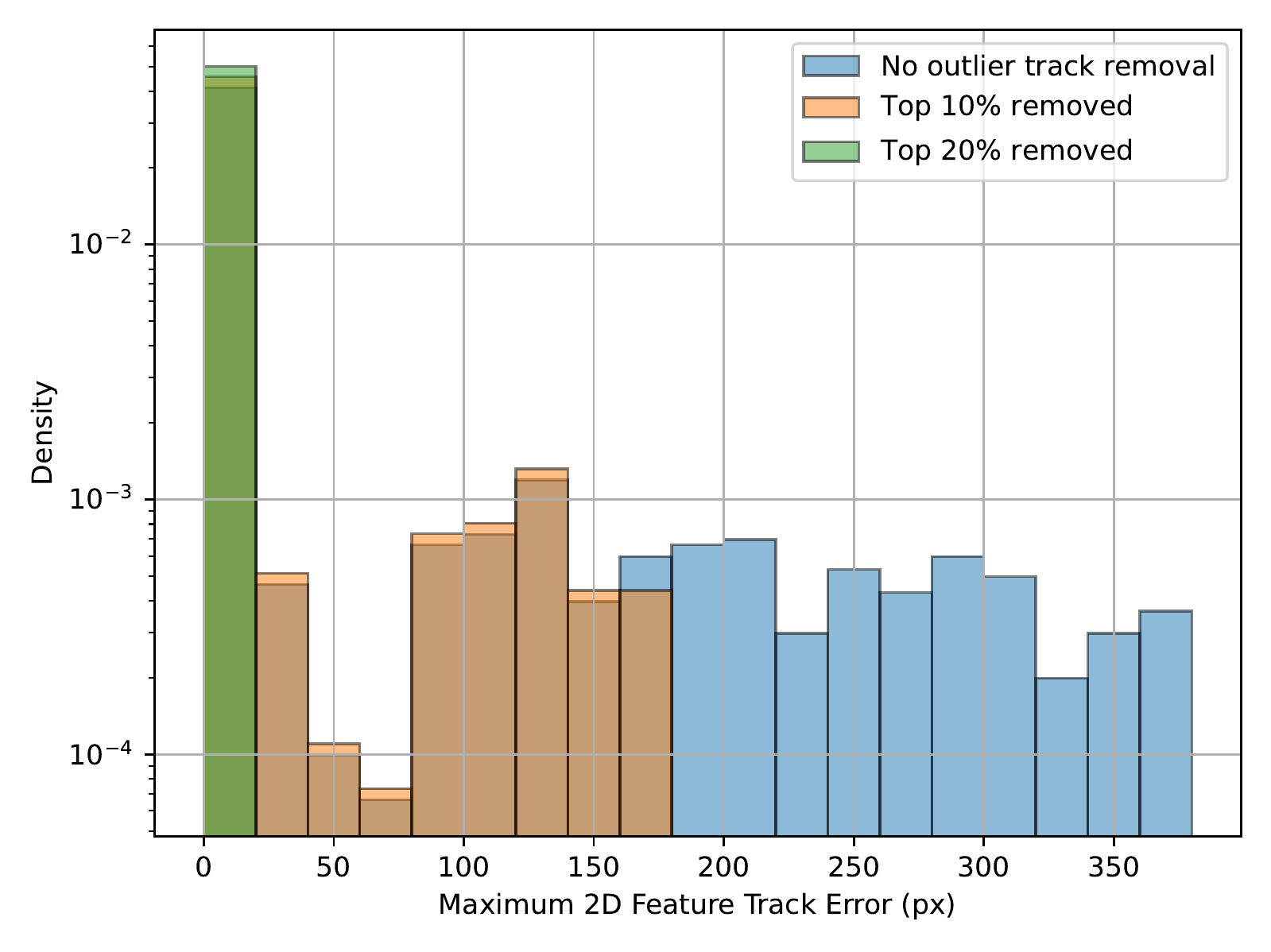}
\caption{\textbf{Gazebo Linear Dataset: We will throw out the 20\% of tracks with the most error instead of the 10\% of tracks.} The right figure plots the histogram density of the maximum L2 error of all feature tracks of one scene in log scale. The corresponding scene is pictured on the left. The large errors  that still remain after removing the 10\% of tracks with the most errors are caused by track propagation along smooth edges when the AGAST feature detector does not select perfect corners, as well as the asynchronous collection of RGB and depth images in the Gazebo simulator. The errors caused by track propagation along smooth edges are unlikely to occur in real world data, where backgrounds and textures are less ideal.}
\label{fig:gazebo_linear_error_throwout}
\end{figure}

In the Gazebo Linear dataset, we throw out tracks whose errors are above the 80-th percentile due to drift that naturally occurs when using the Lucas-Kanade Tracker in an environment containing straight and crisp edges parallel to the direction of motion. More details are given in Figure \ref{fig:gazebo_linear_error_throwout}. In the Gazebo AR/VR dataset, the we throw out tracks whose errors are above the 90-th percentile, as motions are no longer parallel to the straight edges.

\subsection{Results}

Overall, we find that mean error, mean absolute error, covariance, feature lifetime, and outlier ratio are all dependent on the type of motion, the tracker used, and the speed. For the DTU Point Features dataset, we found no dependence on the existence of directional light unless the directional light happened to cause tracking failure at high speeds. In Tables \ref{tab:dtu_summary_table} - \ref{tab:gazebo_arvr_summary_table}, we list the exact dependence of mean error, mean absolute error, feature lifetime, covariance, and outlier ratio on each independent variable. Differences in Tables \ref{tab:dtu_summary_table} - \ref{tab:gazebo_arvr_summary_table} lead us to conclude that feature tracks are dependent on motion, tracker, and speed, but not the existence of directional light.

\begin{figure}[H]
    \centering
    \includegraphics[width=4.00in]{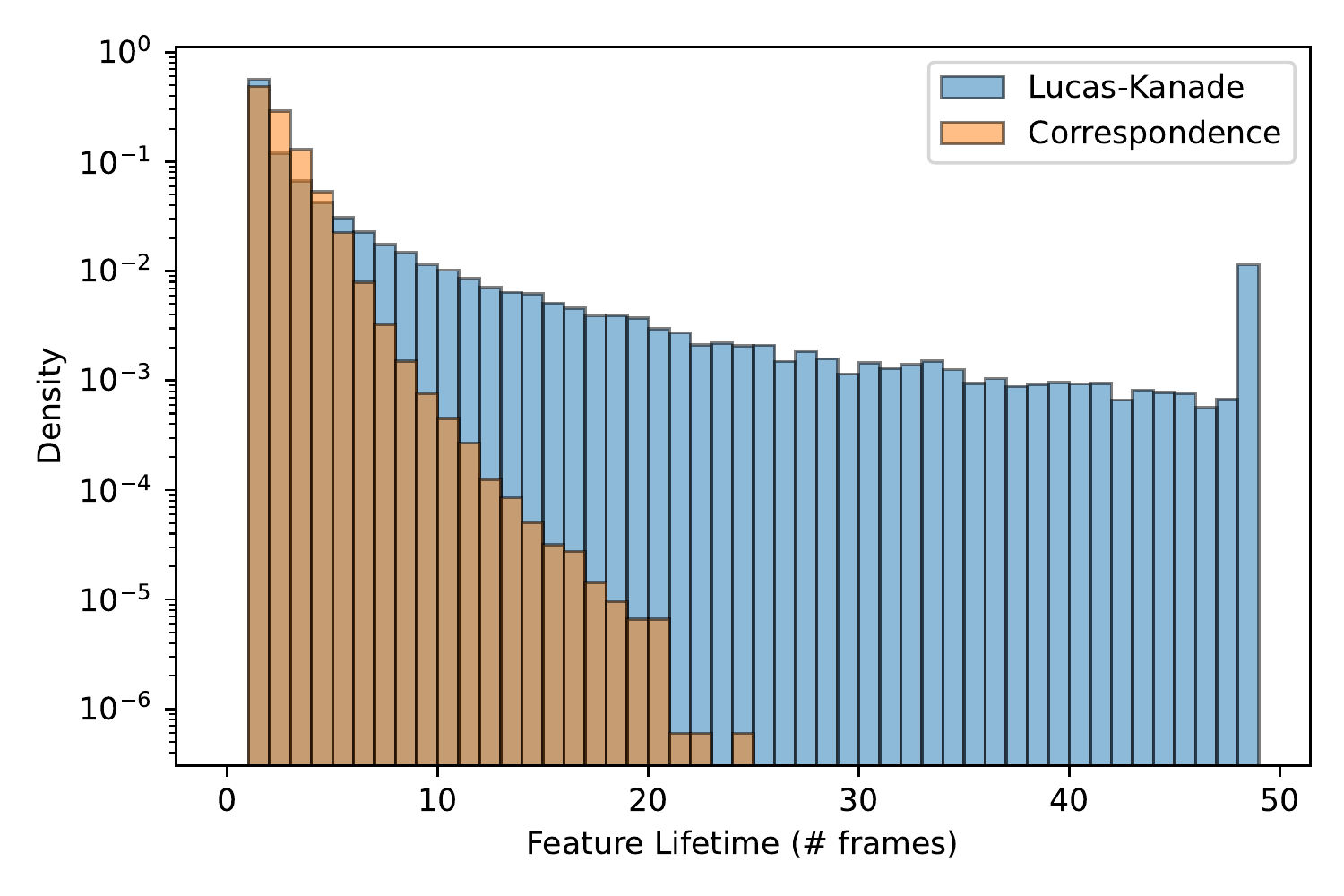}
    \caption{\textbf{DTU Point Features Dataset: Feature lifetimes generated by the Lucas-Kande Tracker is a long tailed distribution.} The histograms above plot feature lifetime density in log scale for scenes with diffuse lighting and no skipped frames (speed=1.00) for the Lucas-Kanade (blue) and Correspondence (orange) Trackers. There is a long tail of tracks with longer lifetimes when we use a sparse optical flow rather than correspondences.}
    \label{fig:dtu_track_lifetime}
\end{figure}

\begin{figure}[H]
    \centering
    \includegraphics[width=4in]{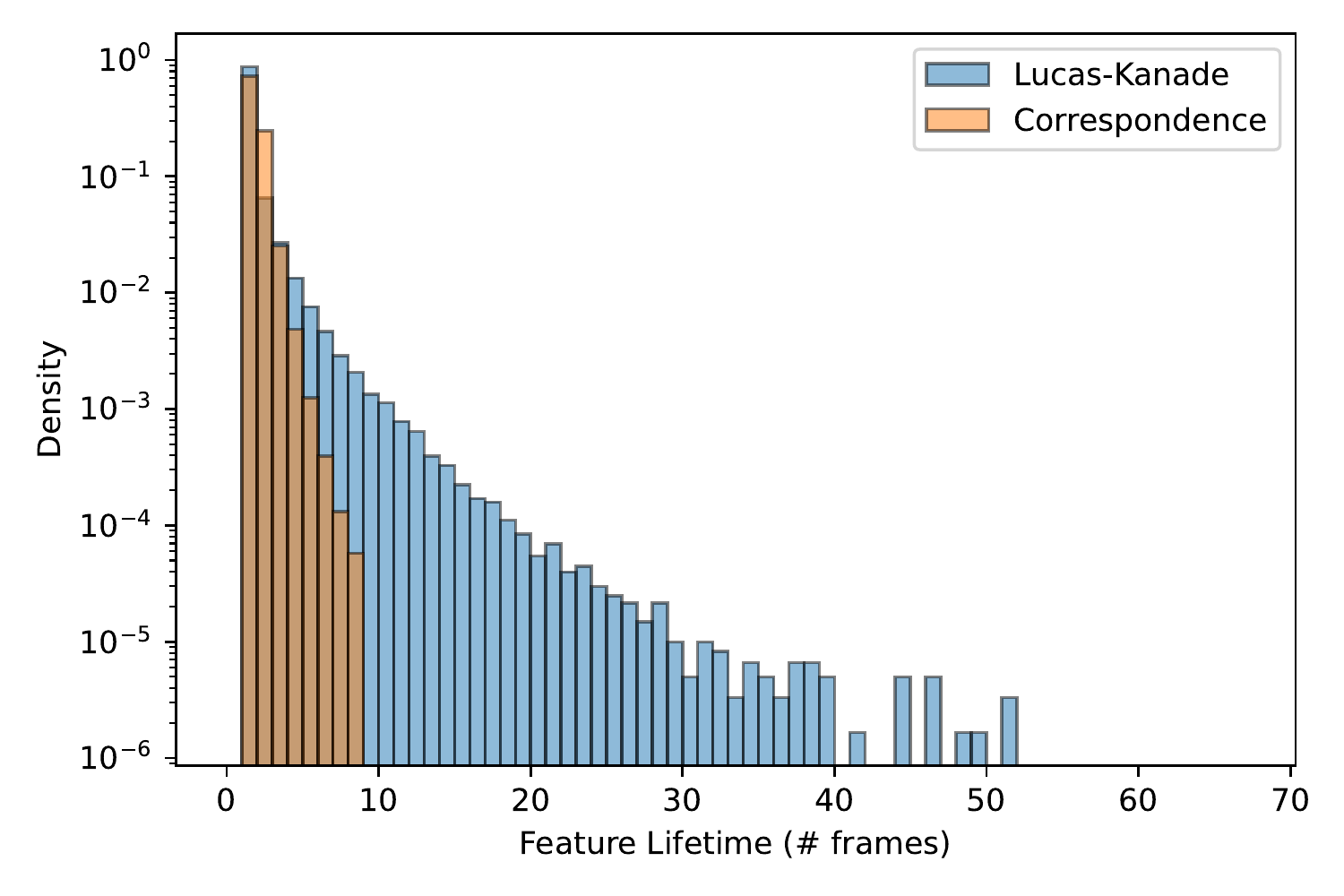}
    \caption{\textbf{KITTI Dataset: Most features live for less than five frames.} The distribution of feature lifetimes is plotted as a log-scale histogram for both the Lucas-Kanade and Correspondence-Based Tracker at nominal speed. The Lucas-Kanade Tracker produces a long tail of features with longer lifetimes. Features with long-lifetimes are those far away from the car's camera, in the center of the image.}
    \label{fig:kitti_feature_lifetime}
\end{figure}

\begin{figure}[H]
    \centering
    \includegraphics[width=4in]{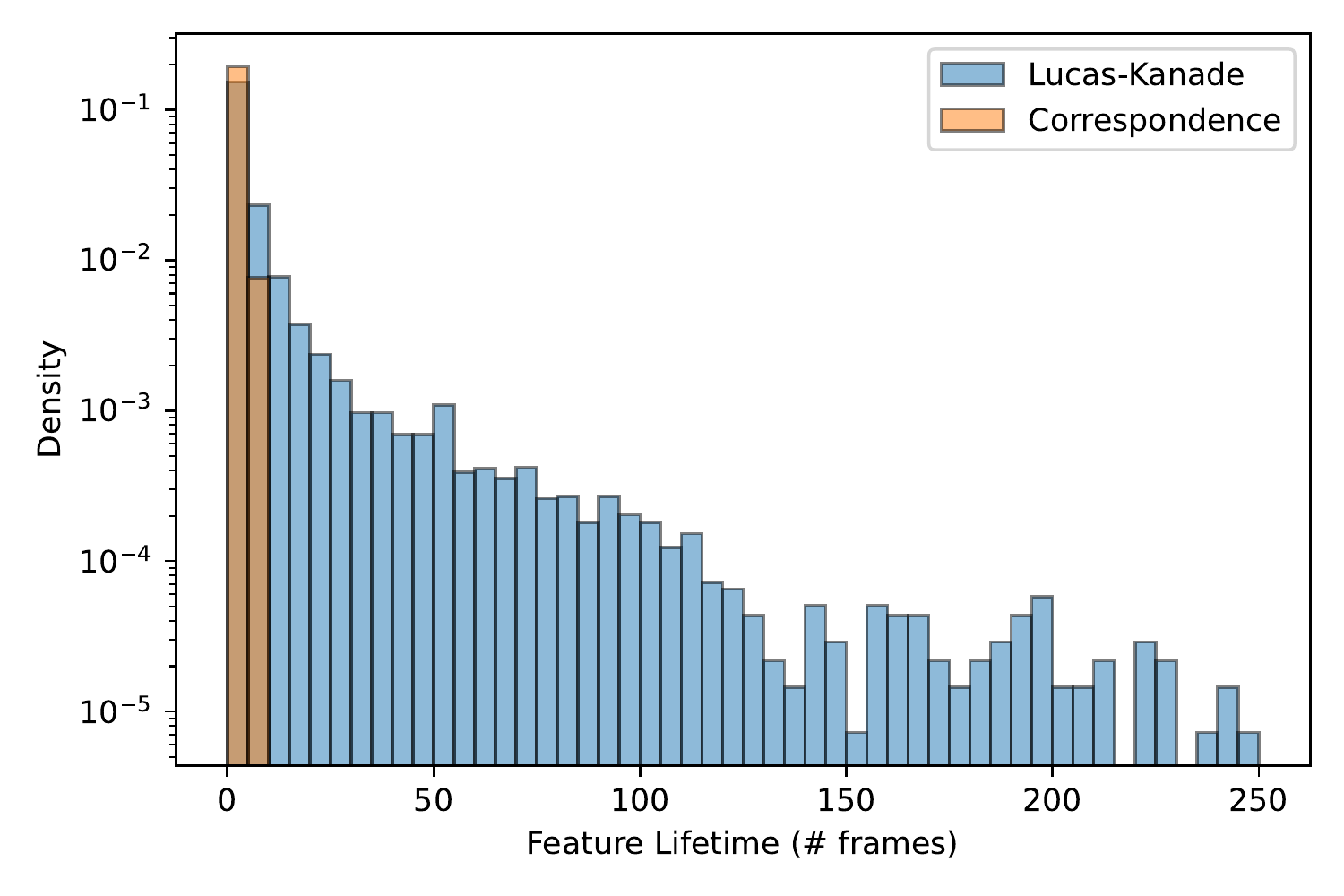}
    \caption{\textbf{Gazebo Linear Dataset: Feature Lifetime is usually $\leq$ five frames.} The distribution of feature lifetimes is plotted as a log-scale histogram for both the Lucas-Kanade and Correspondence Tracker at nominal speed. Many features live for less $\leq$ five frames, especially when the Correspondence Tracker is used. However, Lucas-Kanade produces a long tail of features with longer lifetimes.}
    \label{fig:gazebo_linear_feature_lifetime}
\end{figure}

\begin{figure}[H]
    \centering
    \subfigure[$\nu(t)$, Horizontal Coordinate]{
        \includegraphics[width=0.48\textwidth]{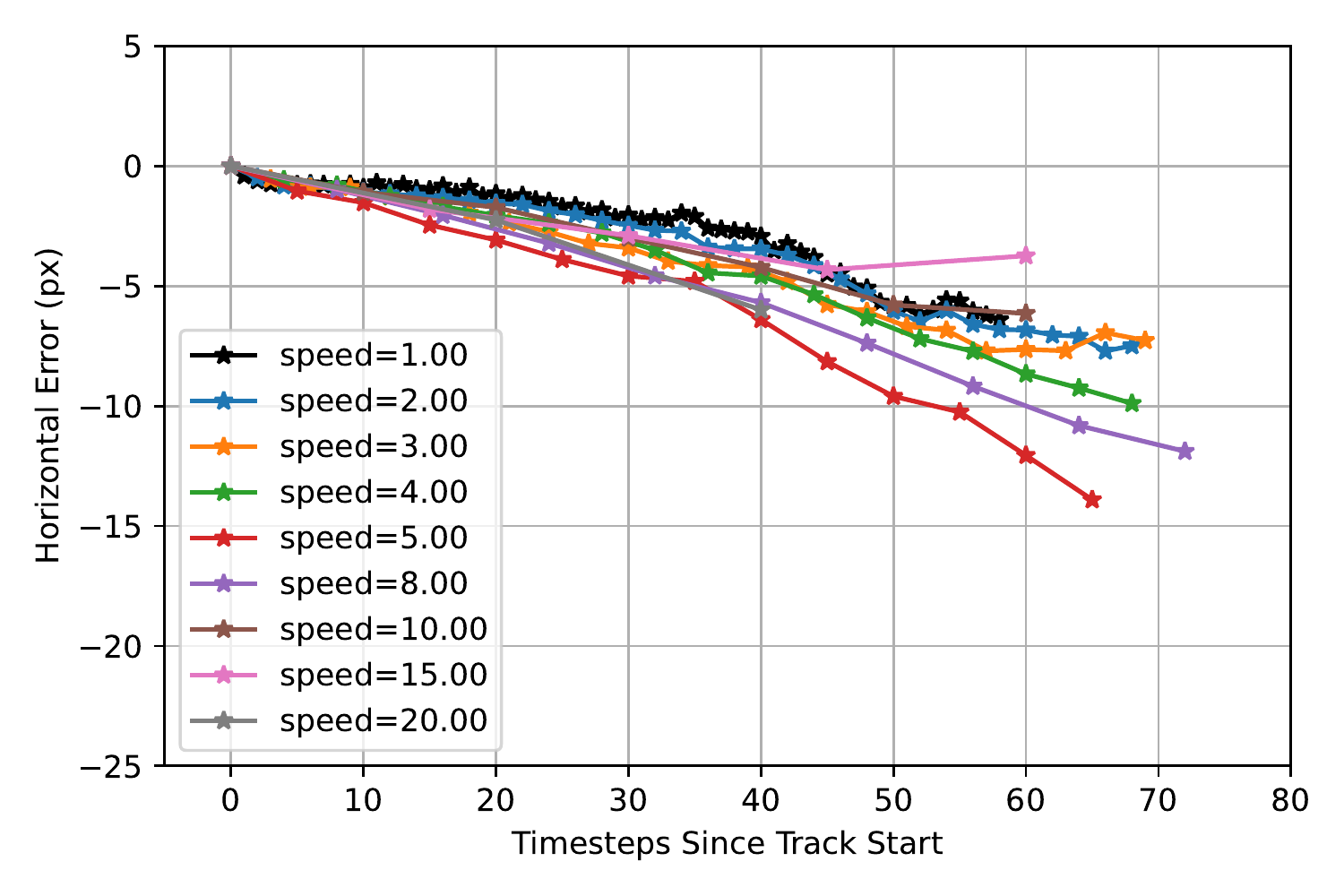}
        \includegraphics[width=0.48\textwidth]{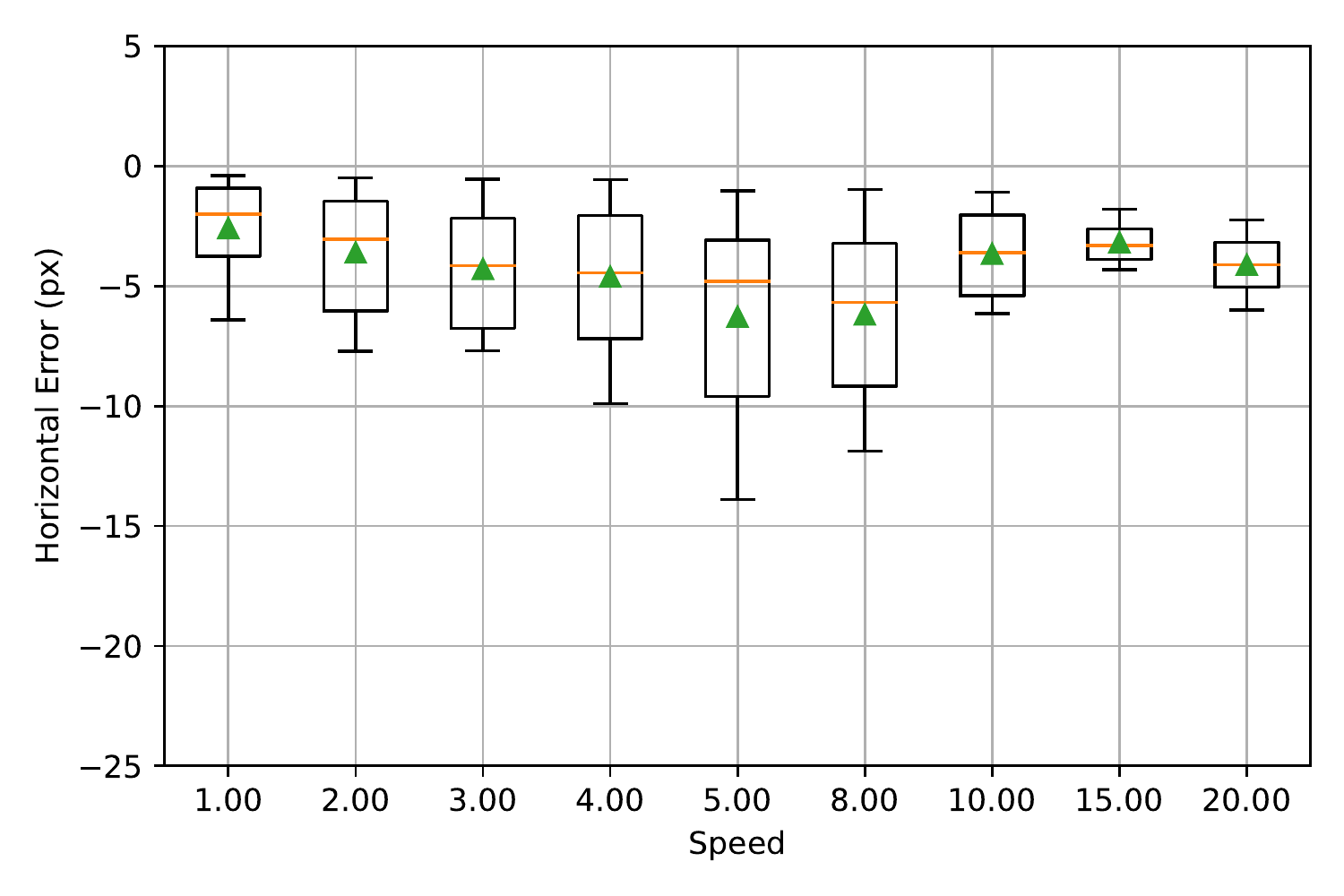}
    }
    \subfigure[$\nu(t)$, Vertical Coordinate]{
        \includegraphics[width=0.48\textwidth]{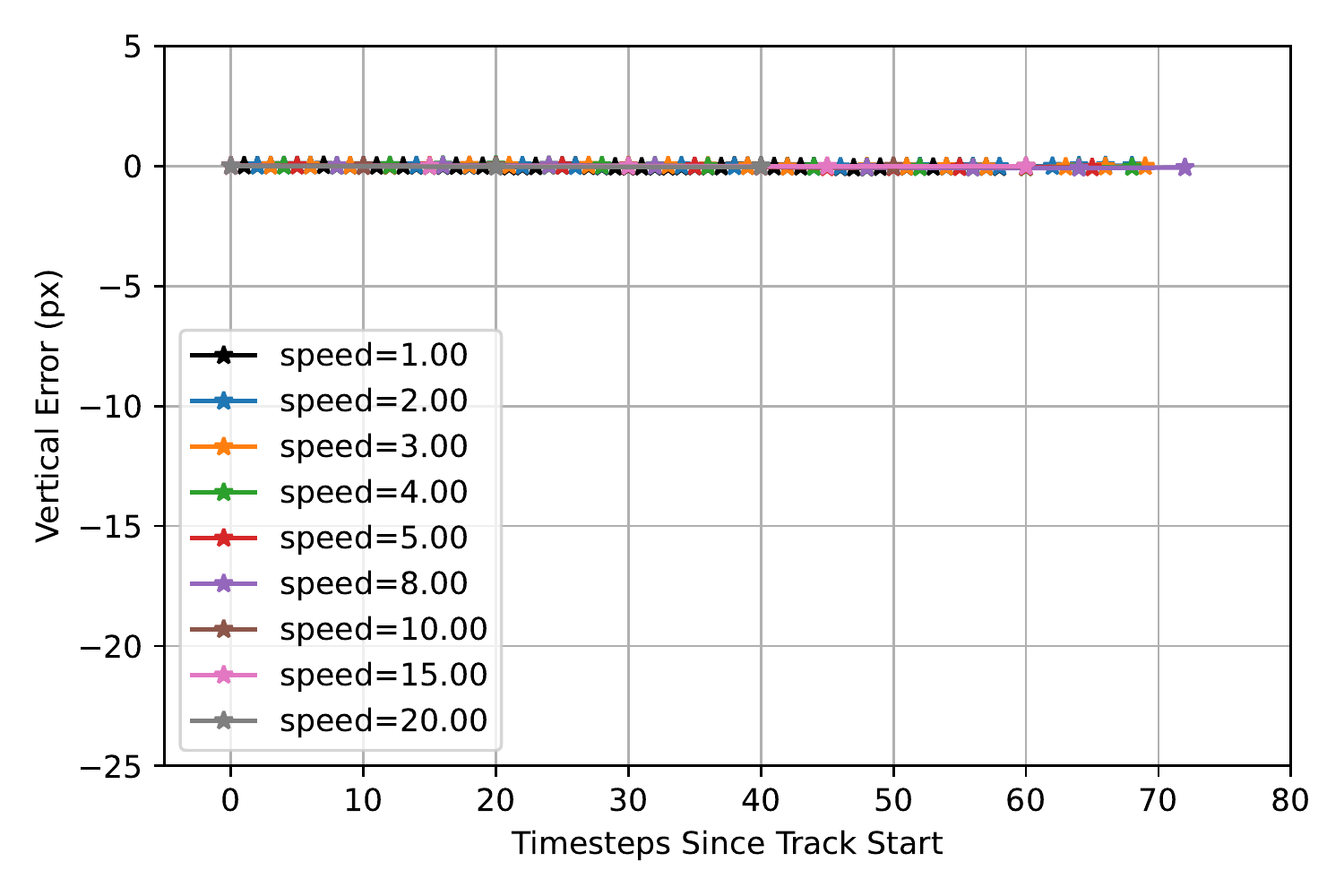}
        \includegraphics[width=0.48\textwidth]{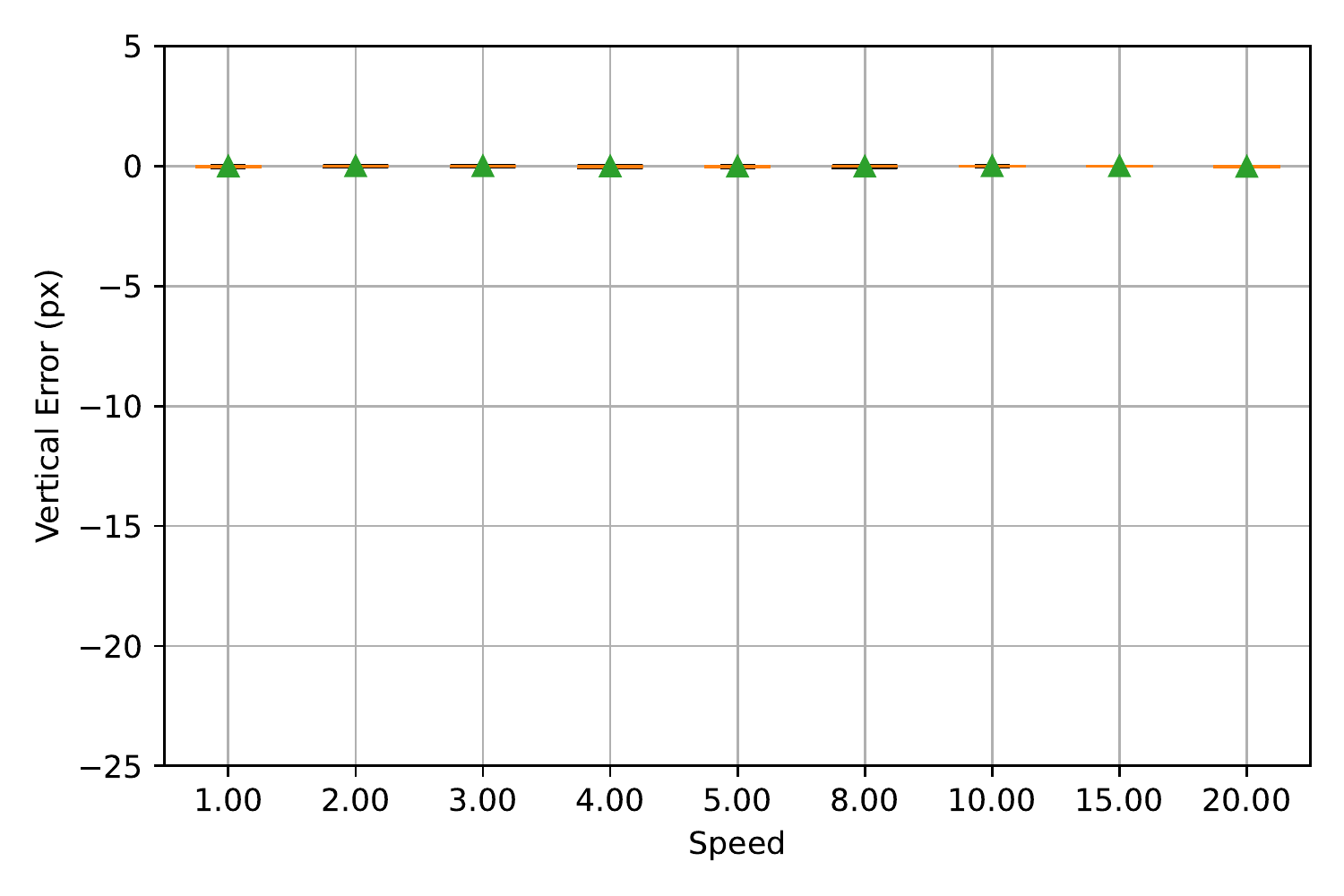}
    }
    \caption{\textbf{Gazebo Linear Dataset: Mean errors increase with speed when using the Lucas-Kanade Tracker.}
    The left column contains plots of the horizontal (top row) and vertical (bottom row) components of the mean tracking error $\nu(t)$ at each timestep $t$ after initial feature detection at multiple speeds. Each dot corresponds to a processed frame; lines for higher speeds contain data from fewer frames and therefore show fewer dots. The right column plots the ordinate values of each line for $t>0$ in the left figures as a box plot: means are shown as green triangles and medians are shown as orange lines.
    The top-right shows that mean errors in the horizontal coordinate become more negative as speed is increased from 1.00 to 8.00. The mean error then decreases for speeds=10.00 (brown line), 15.00 (pink line), and 20.00 (gray line), showing that both the number of elapsed frames, and the speed are both factors that affect $\nu(t)$. For all speeds, mean error is close to zero in the vertical coordinate.
    }
    \label{fig:gazebo_linear_LK_meanerror}
\end{figure}

\begin{figure}[H]
    \centering
    \includegraphics[width=4in]{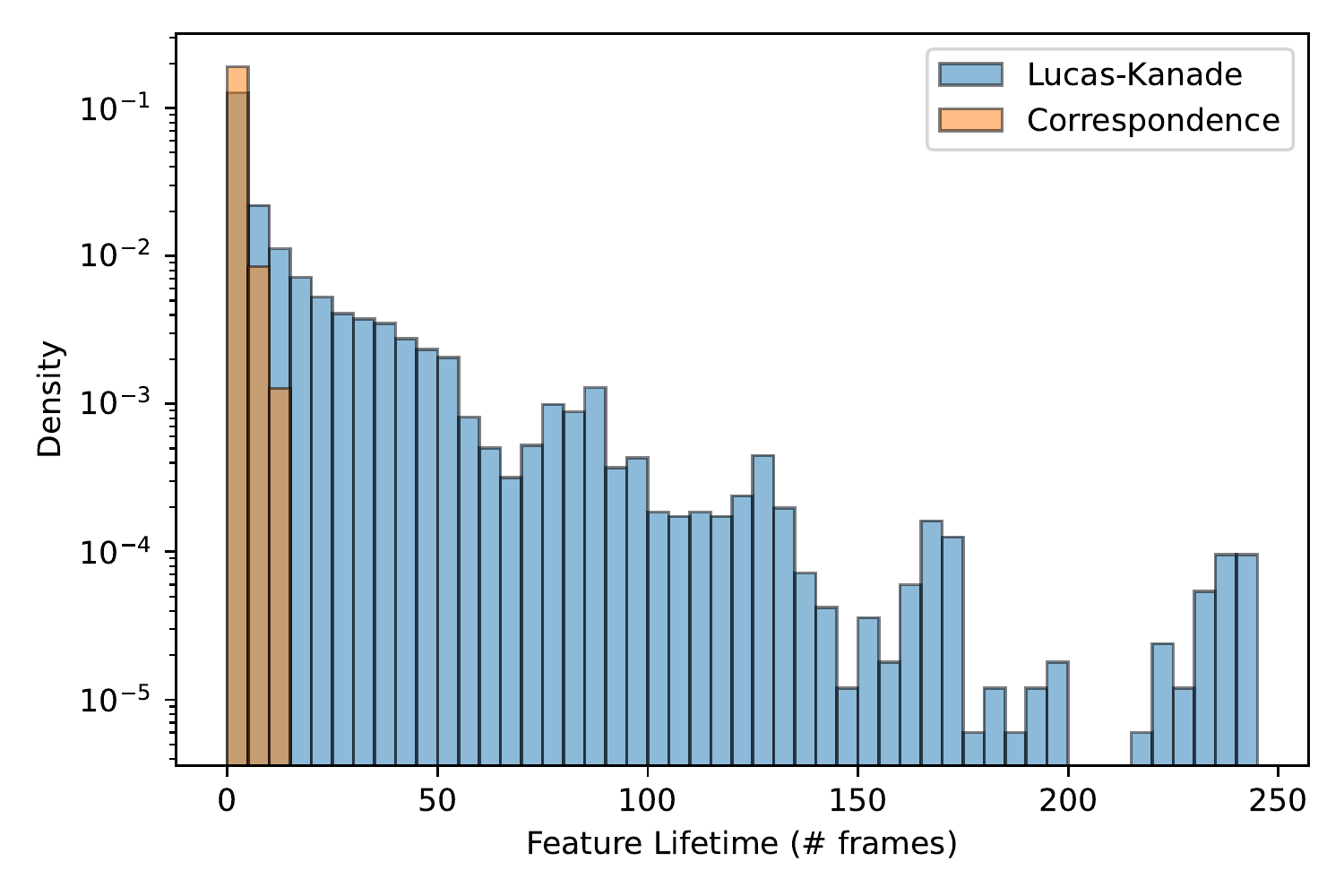}
    \caption{\textbf{Gazebo AR/VR Dataset: Most features live for less than 10 frames.} The distribution of feature lifetimes is plotted as a log-scale histogram for both the Lucas-Kanade and Correspondence Tracker at nominal speed. Many features live for less than 10 frames, especially when the Correspondence Tracker is used. However, Lucas-Kanade produces a long tail of features with longer lifetimes.}
    \label{fig:gazebo_arvr_feature_lifetime}
\end{figure}

\begin{figure}[H]
    \centering
    \subfigure[$\nu(t)$, Horizontal Coordinate]{
        \includegraphics[width=0.48\textwidth]{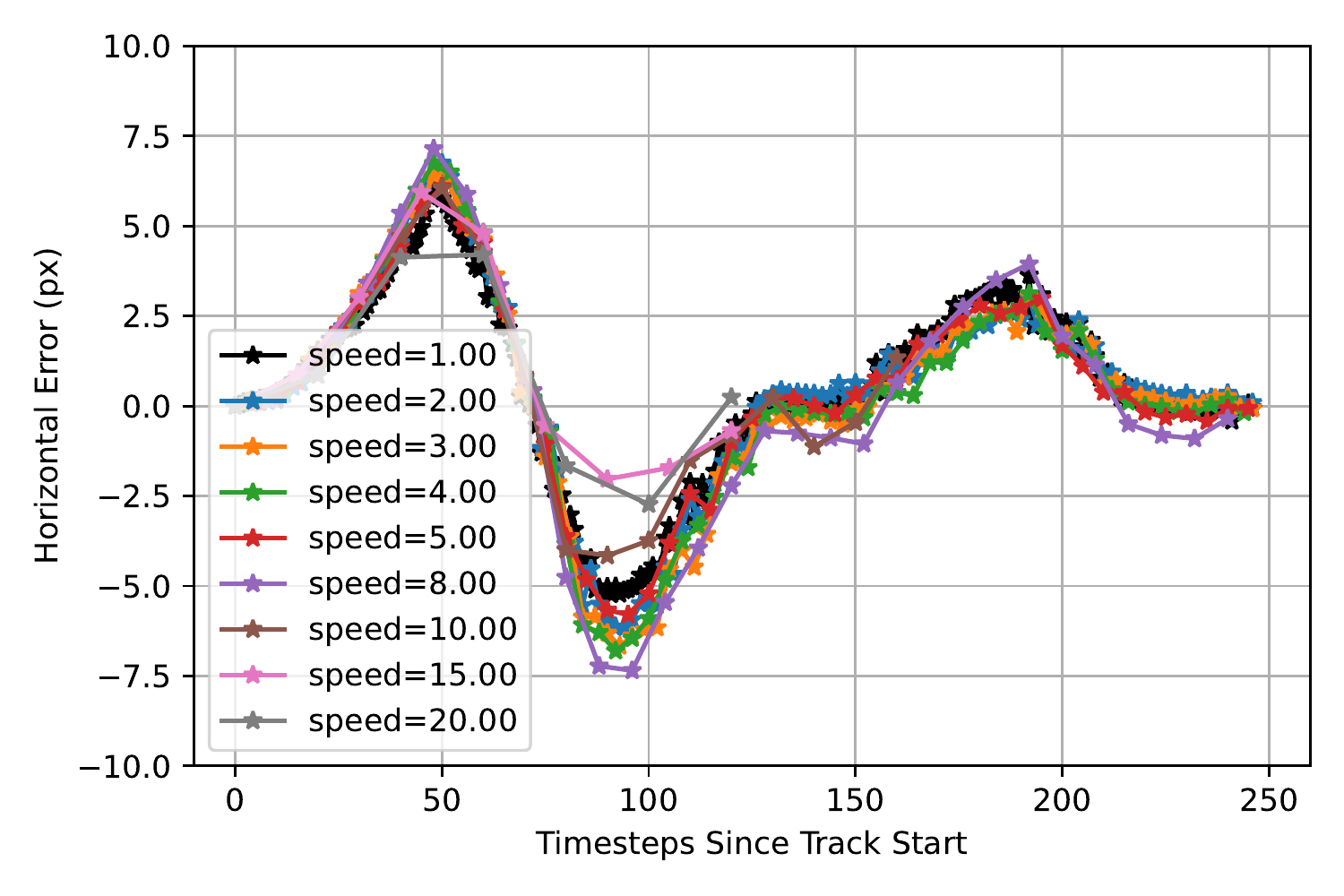}
        \includegraphics[width=0.48\textwidth]{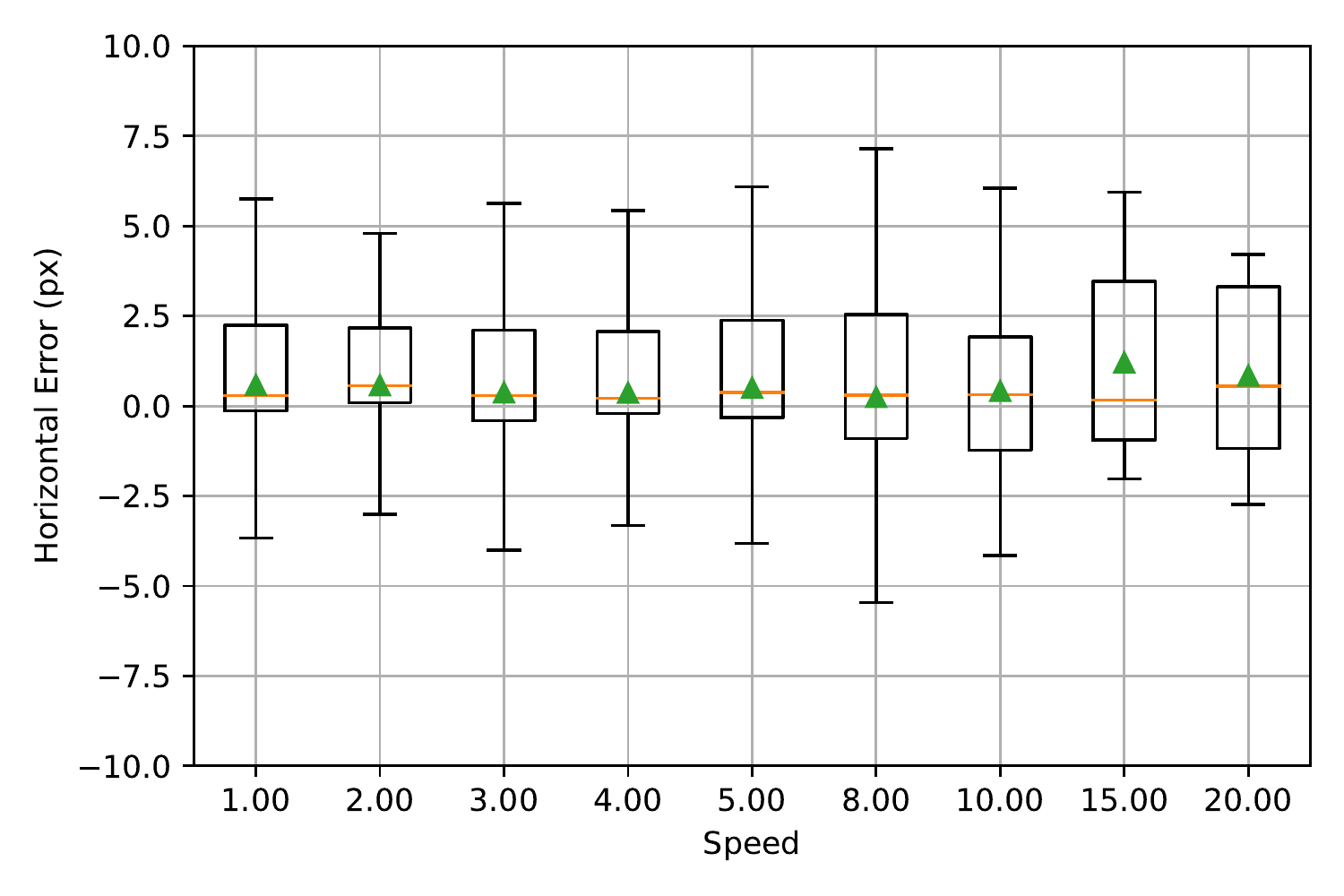}
    }
    \subfigure[$\nu(t)$, Vertical Coordinate]{
        \includegraphics[width=0.48\textwidth]{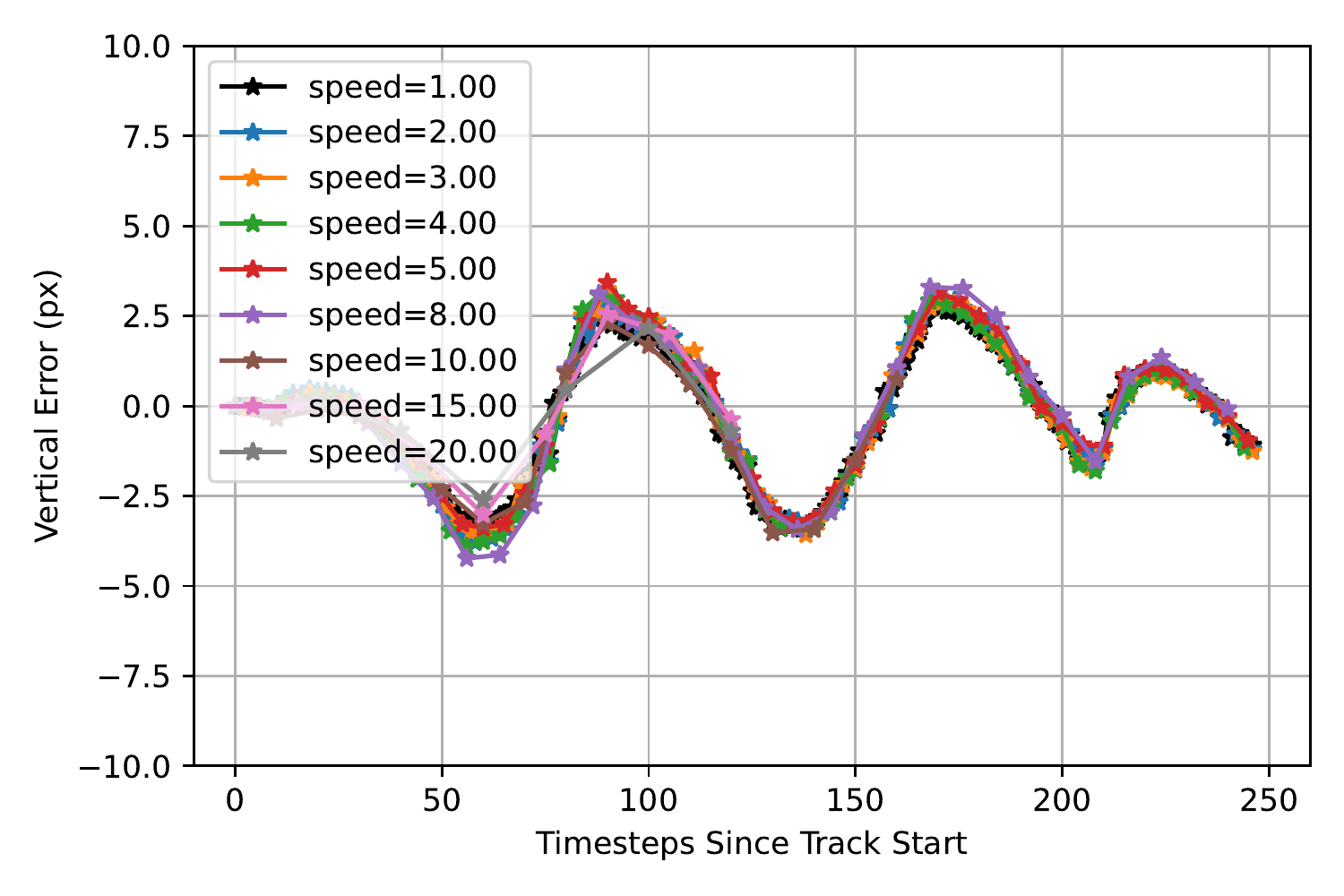}
        \includegraphics[width=0.48\textwidth]{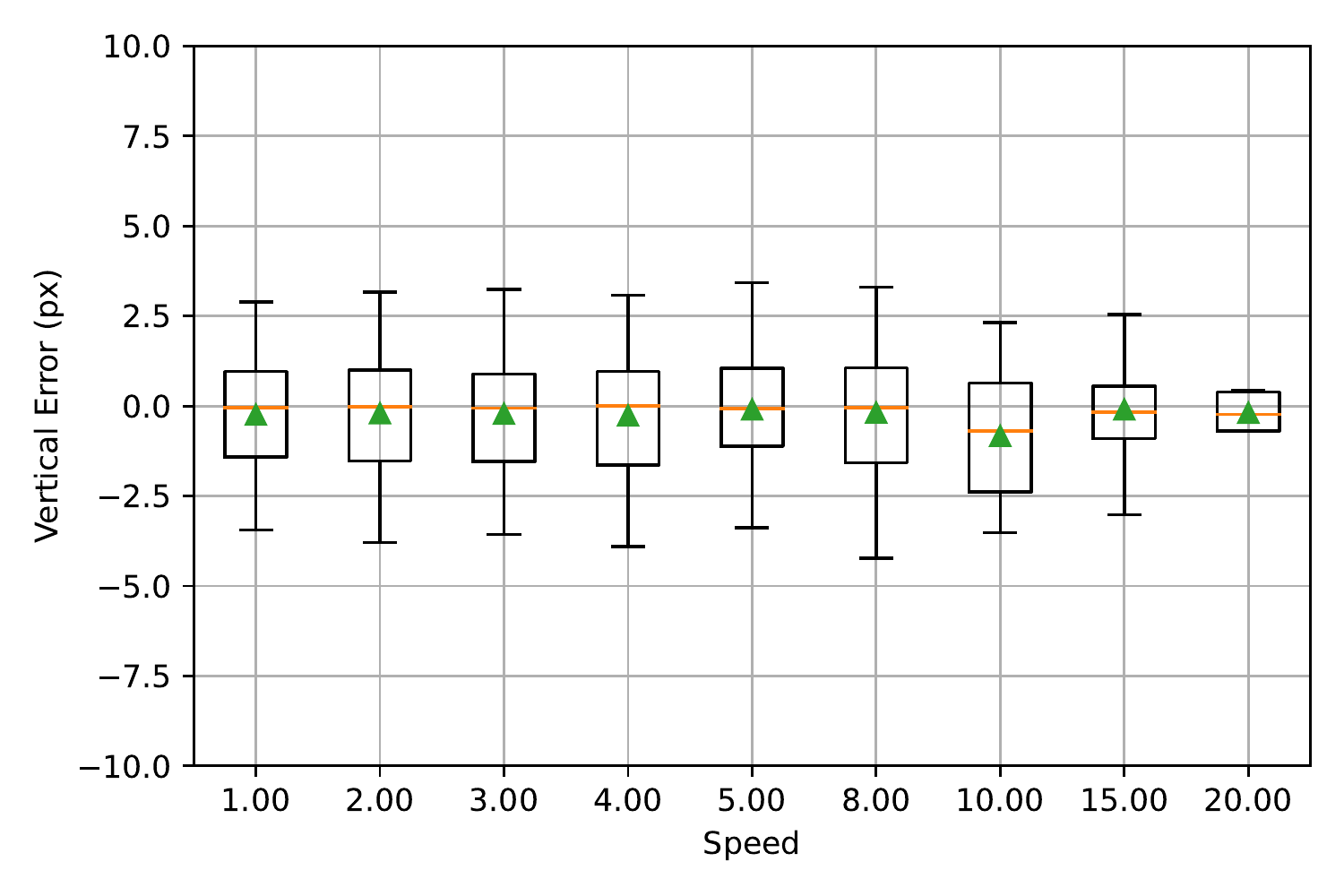}
    }
    \caption{\textbf{Gazebo AR/VR Dataset: Mean tracking errors are not affected by speed when using the Lucas-Kanade Tracker.}
    The left column contains plots of the horizontal (top row) and vertical (bottom row) components of the mean tracking error $\nu(t)$ at each timestep $t$ after initial feature detection at multiple speeds. Each dot corresponds to a processed frame; lines for higher speeds contain data from fewer frames and therefore show fewer dots. The right column plots the ordinate values of each line for $t>0$ in the left figures as a box plot: means are shown as green triangles and medians are shown as orange lines.
    The box plots show that over time, mean and median errors are not much affected by speed in both the horizontal and vertical coordinates. This is also illustrated by the fact that in the left column, all lines are on top of one another. In the left plots, some of the lines at higher speeds (brown, pink, and gray) show less mean error than the lines at lower speeds, indicating that the number of frames, and not just absolute distance, also affects the total drift.
    }
    \label{fig:gazebo_arvr_LK_meanerror}
\end{figure}

One notable difference between the Lucas-Kanade and Correspondence Trackers is that feature tracks produced by the Lucas-Kanade Tracker drift steadily while the Correspondence Tracker does not. This is because the Lucas-Kanade Tracker is differential, i.e. the characterization of a feature will slightly change frame to frame. For the Correspondence Tracker, this is not true. Therefore, the location of the feature track will drift, and the direction and magnitude of drift is dependent on the direction of motion. With left-to-right fixating motion, drift is positive (see Figure \ref{fig:dtu_diffuse_1.00_meanerror}). With left-to-right linear motion, drift is negative, and also larger (see Figure \ref{fig:gazebo_linear_LK_meanerror}). In AR/VR motion, the direction of drift changes with motion (see Figure \ref{fig:gazebo_arvr_LK_meanerror}). The flipside is that the Lucas-Kanade tracker generates features with a longer lifetime (see Figures \ref{fig:dtu_track_lifetime}, \ref{fig:kitti_feature_lifetime}, \ref{fig:gazebo_linear_feature_lifetime}, \ref{fig:gazebo_arvr_feature_lifetime}). When motion is fixating, the Correspondence Tracker also drifts about the direction of motion (see Figure \ref{fig:dtu_mean_error_sideways}).

Finally, we note that the zero-mean Gaussian assumption holds when motion is predominantly forwards and we are using the Correspondence Tracker (see Figures \ref{fig:kitti_match_meanerror} and \ref{fig:kitti_match_cov}). All figures supporting the assertions in this section are given in the Appendix.

\begin{table}[htp]
    \centering
    \begin{tabular}{p{1in}|p{1.0in}|p{1.0in}|p{2.5in}}
                & \textbf{Tracker} & \textbf{Lighting} & \textbf{Speed} \\
    \hline
    $\mu(t)$ & No (fig. \ref{fig:dtu_diffuse_1.00_meanerror}) & No (figs. \ref{fig:dtu_lighting_mu_LK}, \ref{fig:dtu_lighting_mu_match}) & No (figs. \ref{fig:dtu_match_diffuse_mean_error_varyspeed}, \ref{fig:dtu_LK_mean_varyspeed}) \\
    \hline
    $\kappa(t)$ & Yes (fig. \ref{fig:dtu_diffuse_1.00_MAE_cov}) & No (fig. \ref{fig:dtu_diffuse_1.00_MAE_cov}) & Yes for Correspondence Tracker (fig. \ref{fig:dtu_match_diffuse_MAE_varyspeed}), No for Lucas-Kanade Tracker (fig. \ref{fig:dtu_LK_MAE_varyspeed})\\
    \hline
    $\Sigma(t)$ & Yes (fig. \ref{fig:dtu_diffuse_1.00_MAE_cov}) & No (fig. \ref{fig:dtu_lighting_sigma_LK}, \ref{fig:dtu_lighting_sigma_match}) & Yes for Correspondence Tracker (fig. \ref{fig:dtu_match_diffuse_cov_varyspeed}), No for Lucas-Kanade Tracker (fig. \ref{fig:dtu_LK_cov_varyspeed})\\
    \hline
    Feature Lifetime & Yes (fig. \ref{fig:dtu_track_lifetime}) & No  (fig. \ref{fig:dtu_lighting_feature_lifetimes}) & Yes (fig.  \ref{fig:dtu_active_features}) \\
    \hline
    Outlier Ratio & Yes (figs. \ref{fig:dtu_track_outliers_lights}, \ref{fig:dtu_track_outliers_speed}) & No (fig.  \ref{fig:dtu_track_outliers_lights}) & Yes  (fig. \ref{fig:dtu_track_outliers_speed}) \\
    \end{tabular}
    \caption{\textbf{DTU Point Features Results Summary.} Cells contain whether or not the dependent variables in the left column are affected by the independent variables listed in the top row. Entries also contain figure numbers containing justification. The ``Tracker" and ``Lighting" columns contain references to figures containing plots at nominal speed. Although not indicated in the table, Figures \ref{fig:dtu_speed2.00_percent_outlier} - \ref{fig:dtu_LK_cov_speed12.00} in the Appendix show that the existence of directional lighting continues to not affect outlier ratio, mean error, mean absolute error, and covariance at higher speeds for both the Lucas-Kanade and Correspondence Trackers.}
    \label{tab:dtu_summary_table}
\end{table}

\begin{table}[htp]
    \centering
    \begin{tabular}{p{1in}|p{1.5in}|p{2.50in}}
                & \textbf{Tracker}  & \textbf{Speed} \\
    \hline
    $\nu(t)$ & No (fig. \ref{fig:kitti_1.00_meanerror}) & Yes (figs. \ref{fig:kitti_LK_meanerror}, \ref{fig:kitti_match_meanerror}) \\
    \hline
    $\eta(t)$ & Yes (fig. \ref{fig:kitti_1.00_error_cov}) & No for Correspondence Tracker (fig. \ref{fig:kitti_match_MAE}), Yes for Lucas-Kanade Tracker (figs. \ref{fig:kitti_LK_MAE}) \\
    \hline
    $\Phi(t)$ & Yes (fig. \ref{fig:kitti_1.00_error_cov}) & No for Correspondence Tracker (fig. \ref{fig:kitti_match_cov}), Yes for Lucas-Kanade Tracker (fig. \ref{fig:kitti_LK_cov}) \\ 
    \hline
    Feature Lifetime & Yes (fig. \ref{fig:kitti_feature_lifetime}) & Yes (fig. \ref{fig:kitti_avg_feats}) \\
    \hline
    Outlier Ratio & Yes (fig. \ref{fig:kitti_outlier_ratio}) & Yes (fig. \ref{fig:kitti_outlier_ratio})\\
    \end{tabular}
    \caption{\textbf{KITTI Results Summary.} Cells contain whether or not the dependent variables in the left column are affected by the independent variables listed in the top row. Entries also contain figure numbers containing justification.}
    \label{tab:kitti_summary_table}
\end{table}

\begin{table}[htp]
    \centering
    \begin{tabular}{p{1in}|p{1.5in}|p{2.5in}}
                & \textbf{Tracker}  & \textbf{Speed} \\
    \hline
    $\nu(t)$ & Yes (fig. \ref{fig:gazebo_linear_1.00_meanerror}) & No for Correspondence Tracker (fig. \ref{fig:gazebo_linear_match_meanerror}), Yes for Lucas-Kanade Tracker   (fig. \ref{fig:gazebo_linear_LK_meanerror}) \\
    \hline
    $\eta(t)$ & Yes (fig. \ref{fig:gazebo_linear_1.00_error_cov}) & Yes (figs. \ref{fig:gazebo_linear_LK_MAE}, \ref{fig:gazebo_linear_match_MAE}) \\
    \hline
    $\Phi(t)$ & Yes (fig. \ref{fig:gazebo_linear_1.00_error_cov}) & Yes (figs.  \ref{fig:gazebo_linear_match_cov}, \ref{fig:gazebo_linear_LK_cov}) \\ 
    \hline
    Feature Lifetime & Yes (fig. \ref{fig:gazebo_linear_feature_lifetime}) & Yes (fig. \ref{fig:gazebo_linear_avg_feats}) \\
    \hline
    Outlier Ratio & Yes (fig. \ref{fig:gazebo_linear_outlier_ratio}) &  No for Correspondence Tracker, Yes for Lucas-Kanade Tracker (fig. \ref{fig:gazebo_linear_outlier_ratio})\\
    \end{tabular}
    \caption{\textbf{Gazebo Linear Results Summary.} Cells contain whether or not the dependent variables in the left column are affected by the independent variables listed in the top row. Entries also contain figure numbers containing justification.}
    \label{tab:gazebo_linear_summary_table}
\end{table}

\begin{table}[htp]
    \centering
    \begin{tabular}{p{1in}|p{1.5in}|p{2.5in}}
                & \textbf{Tracker}  & \textbf{Speed} \\
    \hline
    $\nu(t)$ & Yes (fig. \ref{fig:gazebo_arvr_1.00_meanerror}) & No (fig. \ref{fig:gazebo_arvr_LK_meanerror}, \ref{fig:gazebo_arvr_match_meanerror}) \\
    \hline
    $\eta(t)$ & Yes (fig. \ref{fig:gazebo_arvr_1.00_error_cov}) & Yes for Correspondence   Tracker (fig. \ref{fig:gazebo_arvr_match_MAE}), No for Lucas-Kanade  Tracker (fig. \ref{fig:gazebo_arvr_LK_MAE})  \\
    \hline
    $\Phi(t)$ & Yes (fig. \ref{fig:gazebo_arvr_1.00_error_cov}) &  Yes for Correspondence   Tracker (fig. \ref{fig:gazebo_arvr_match_cov}), No for Lucas-Kanade Tracker (fig. \ref{fig:gazebo_arvr_LK_cov})  \\ 
    \hline
    Feature Lifetime & Yes (fig. \ref{fig:gazebo_arvr_feature_lifetime}) & Yes (fig. \ref{fig:gazebo_arvr_avg_feats}) \\
    \hline
    Outlier Ratio & Yes (fig. \ref{fig:gazebo_arvr_outlier_ratio}) & No for Correspondence Tracker, Yes for Lucas-Kanade Tracker (fig. \ref{fig:gazebo_arvr_outlier_ratio})\\
    \end{tabular}
    \caption{\textbf{Gazebo AR/VR Results Summary.} Cells contain whether or not the dependent variables in the left column are affected by the independent variables listed in the top row. Entries also contain figure numbers containing justification.}
    \label{tab:gazebo_arvr_summary_table}
\end{table}

\section{Discussion}
\label{sec:discussion}

Other than the caveat about the Correspondence Tracker noted in Section \ref{sec:feature_tracker_configuration}, the main limitation of this work is that there are more variables we could have tested, but chose not to. Examples of variables we chose not to test are the choice of feature detector and descriptor, and characteristics in the scene. For example, would the Correspondence Tracker have as little drift when moving forwards in an indoor environment and comparing BRIEF descriptors? Testing for conditionality on more variables inevitably leads to an unmanageable experiment, so we chose to lock in the feature detector and descriptor to well-performing available options and let the dataset dictate available scenes. Nevertheless, our work is a first step in characterizing the dependence of mean error, mean absolute error, covariance, feature lifetime, and outlier ratio on motion, tracker, speed, and the existence of directional lighting. The main conclusion is that the common zero-mean Gaussian assumption is rarely true. This conclusion motivates a few areas of future work.

The most immediate direction of future work is to continue to use the Extended Kalman Filter and dynamically adapt filter parameters, such as covariance estimates and the number of tracked features, to the scene. Since feature tracks are not zero-mean, covariance estimates will have to be enlarged so that feature tracks containing the extra bias are not outliers. Machine learning approaches to adapting the covariance already exist \cite{vega-brown_cello_2013, liu_deep_2018}. Since statistical methods are not often desirable in safety critical systems, it is of interest to compare performance when covariance is adjusted by a learned model to when covariance is adjusted by a finite state machine. While this approach is the most immediate, it does not address the fact that it brings no convergence guarantees in a downstream state estimation process and will therefore require extensive testing for each application.

The second area of future work is to adapt existing state estimation algorithms to accommodate feature tracks that are not zero-mean Gaussian. It may not be possible, however, to design a filter that is both computationally tractable, guaranteed to converge, and simple enough to implement on a complex, realistic system. This motivates the study in the next chapter, and the third area of future work.

The third direction of future work is to adjust individual feature tracks \emph{before} they are used by a state estimation algorithm that assumes that measurements are zero-mean Gaussian. This is the approach used for IMUs: errors in IMUs measurements are primarily dependent on temperature and mechanical alignment errors, so IMU measurements are adjusted for temperature and known mechanical misalignments before they are passed to a downstream computer. For feature tracks, the calibration table would be more complex, as it is dependent on speed, motion type, and the type of tracker used.

\section{Supporting Figures}

\subsection{Supporting Figures for DTU Point Features Dataset}
\label{sec:all_dtu_figs}

\begin{figure}[H]
\centering
\includegraphics[width=0.48\textwidth]{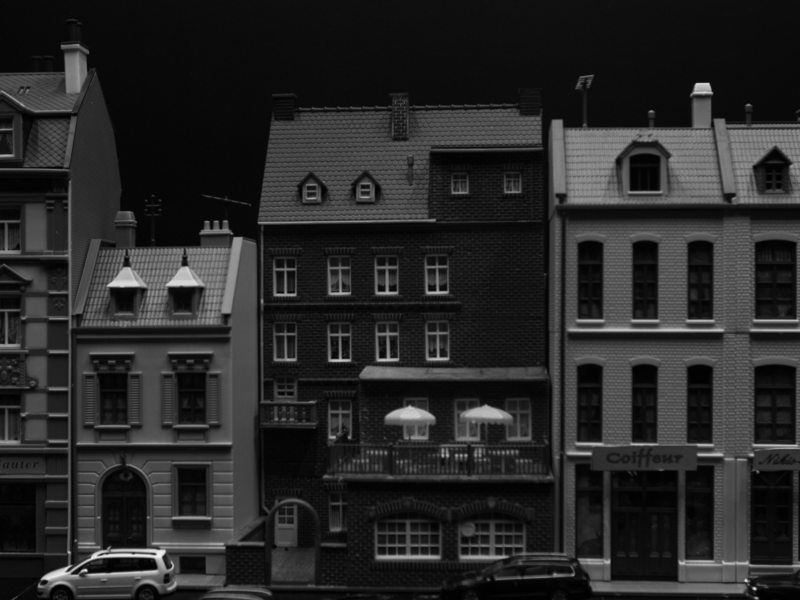}
\includegraphics[width=0.48\textwidth]{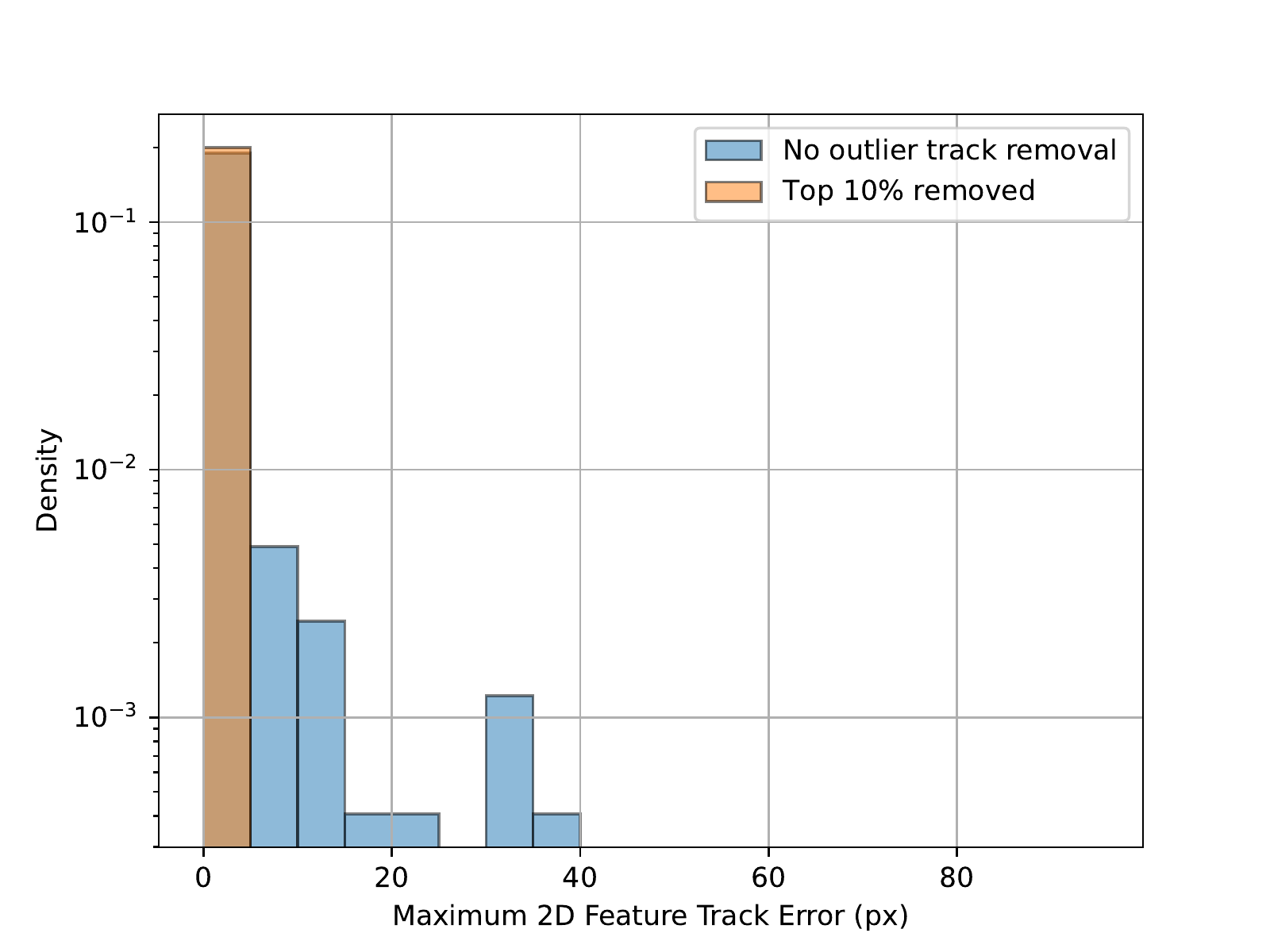}
\caption{\textbf{DTU Point Features Dataset: We will throw out the 10\% of tracks with the most error from each scene.} The right figure plots the histogram density of all feature tracks' maximum L2 error in log scale. The corresponding scene is pictured on the left. Outliers in the blue histogram are caused by noisy depth measurements and the imperfect association of features with laser scan points.}
\label{fig:dtu_error_throwout}
\end{figure}

\begin{figure}[H]
    \centering
    \subfigure[Lucas-Kanade]{\includegraphics[width=0.48\textwidth]{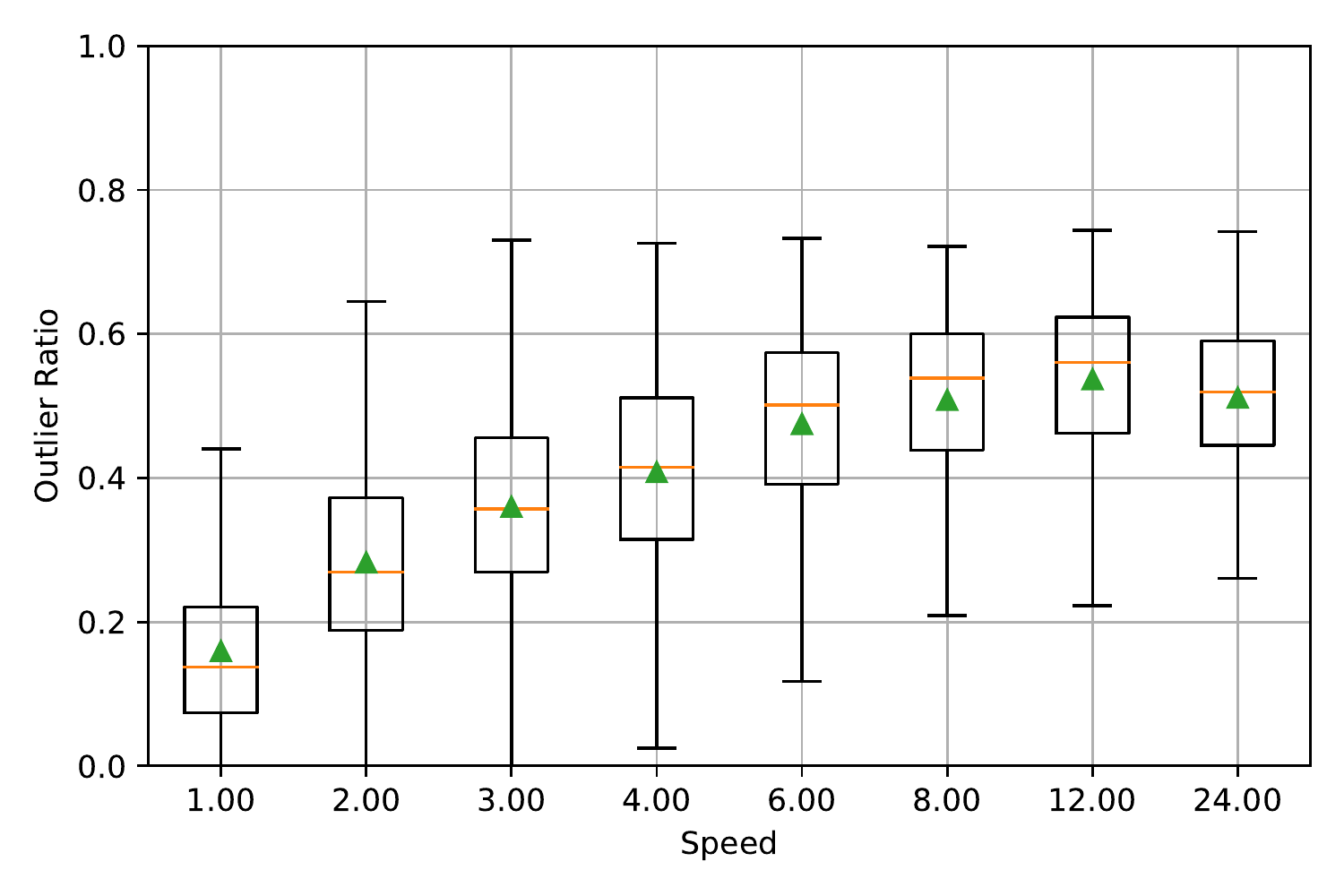}}
    \subfigure[Correspondence]{\includegraphics[width=0.48\textwidth]{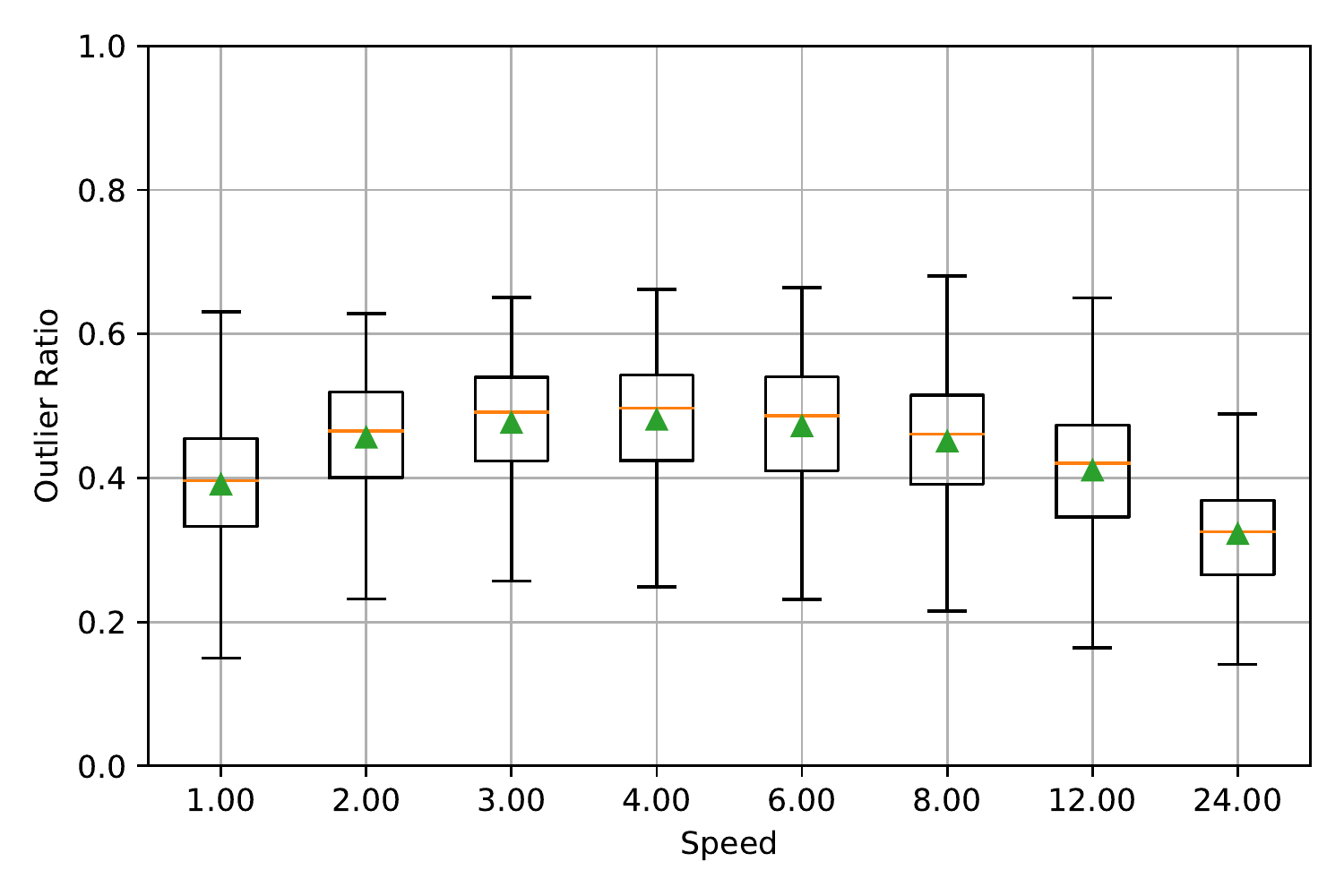}}
    \caption{\textbf{DTU Point Features Dataset: Outlier Ratio Depends on Speed.} In the box-and-whisker plots above, the orange line is the median, the green triangle is the mean, and the box extends from the first to the third quartiles. The whiskers extend up to 1.5x the length of the boxes. Outlier ratios increase with speed for all tested feature trackers to a point, and then falls slightly. Each box-and-whisker is computed using features from all 60 scenes, one tracker, and one speed. Outlier ratios then decrease at higher speeds not because the tracker is more accurate, but because the percentage of features that fail to be tracked from frame to frame increases.}
    \label{fig:dtu_track_outliers_speed}
\end{figure}

\begin{figure}[H]
    \centering
    \subfigure[Lucas-Kanade]{
        \includegraphics[width=0.48\textwidth]{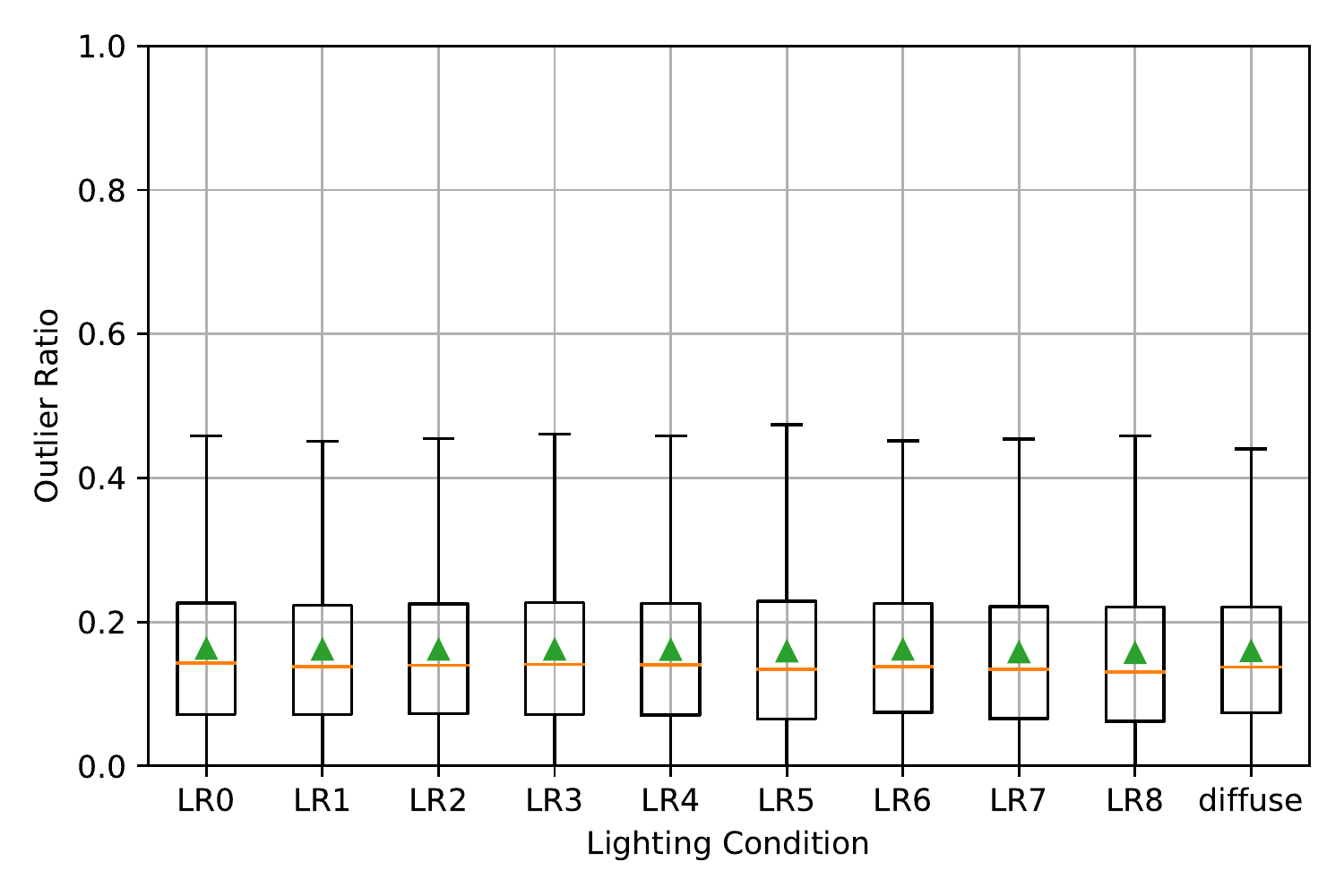}
        \includegraphics[width=0.48\textwidth]{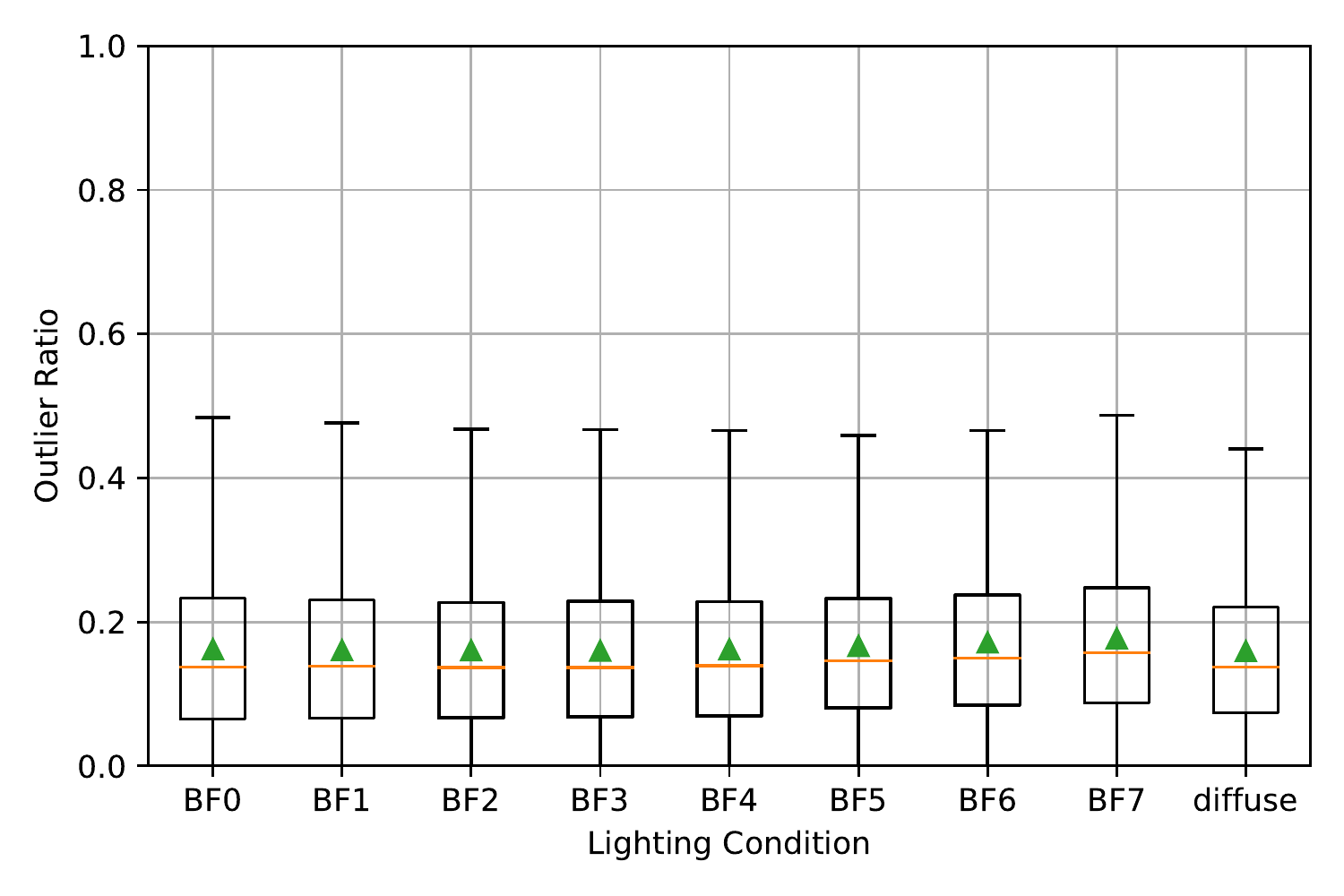}
    }
    \subfigure[Correspondence]{
        \includegraphics[width=0.48\textwidth]{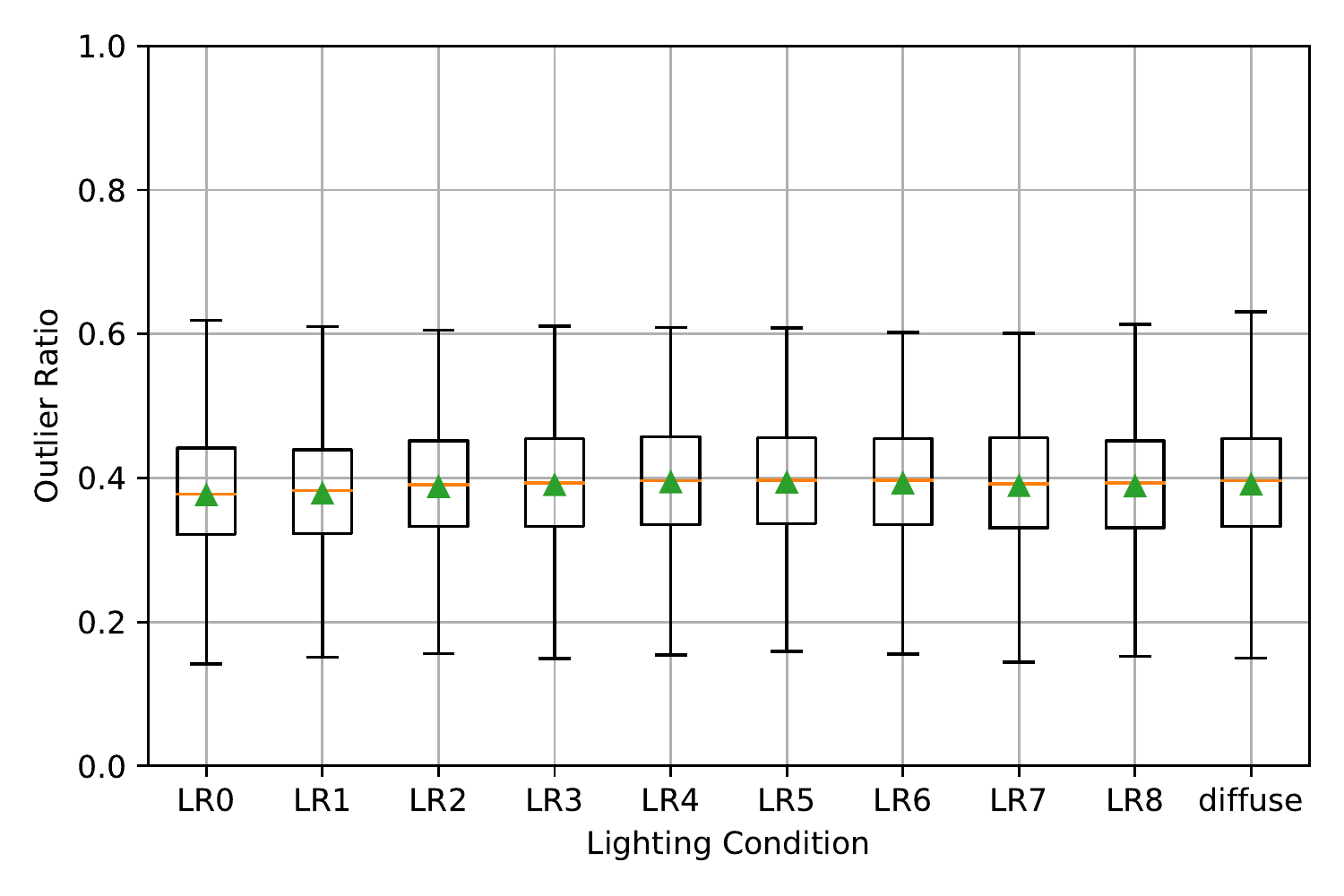}
        \includegraphics[width=0.48\textwidth]{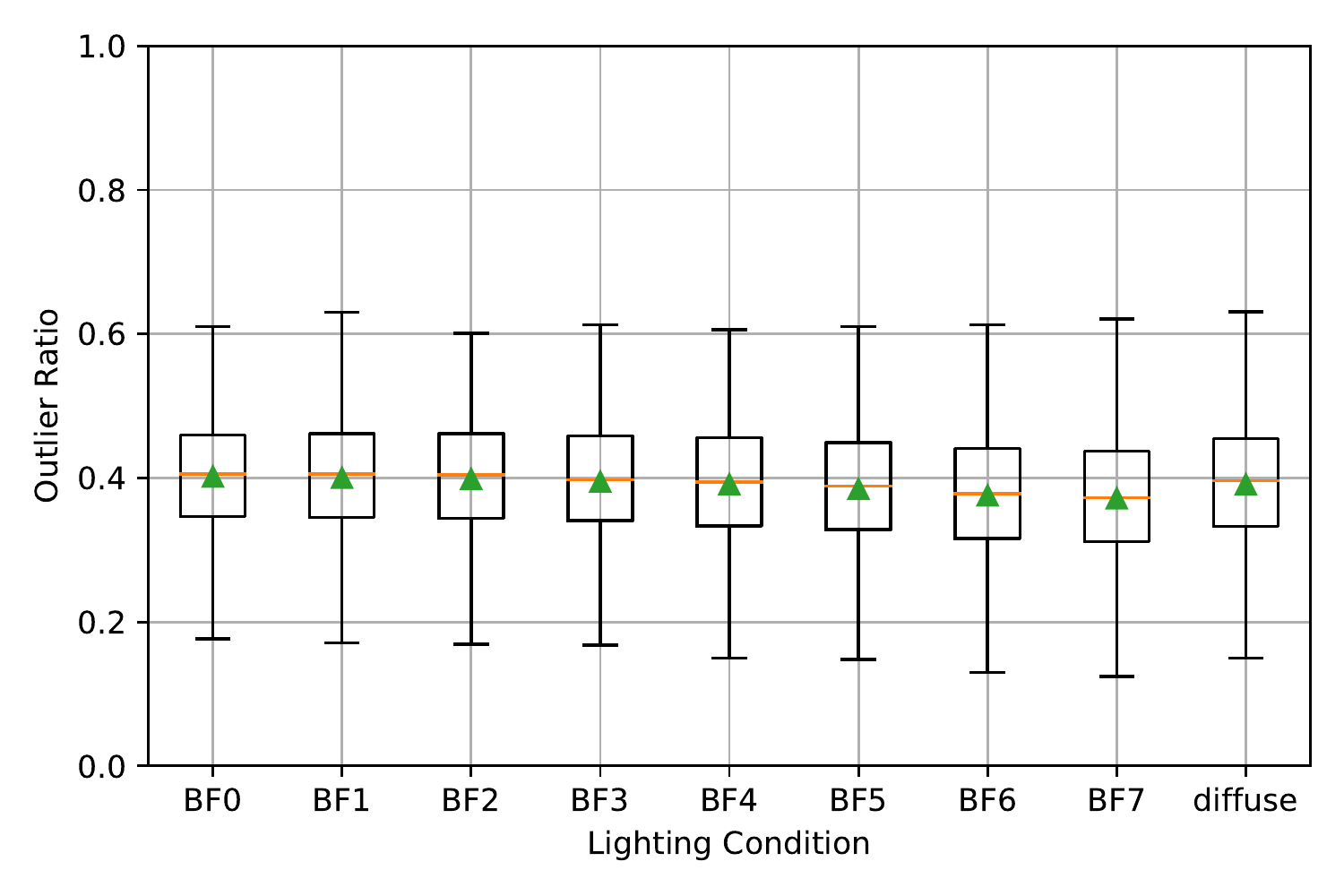}
    }   
    \caption{\textbf{DTU Point Features Dataset: Outlier ratio does not depend on the existence of directional lighting.} In the box-and-whisker plots above, the orange line is the median, the green triangle is the mean, and the box extends from the first to the third quartiles. The whiskers extend up to 1.5x the length of the boxes. Each box-and-whisker plot is computed using features from all 60 scenes, one tracker, speed=1.00, and one of the lighting conditions in Figure \ref{fig:dtu_light_stage}. The distribution of outlier ratio is approximately the same for all lighting conditions.}
    \label{fig:dtu_track_outliers_lights}
\end{figure}

\begin{figure}[H]
    \centering
    \subfigure[Lucas-Kanade]{
        \includegraphics[width=0.48\textwidth]{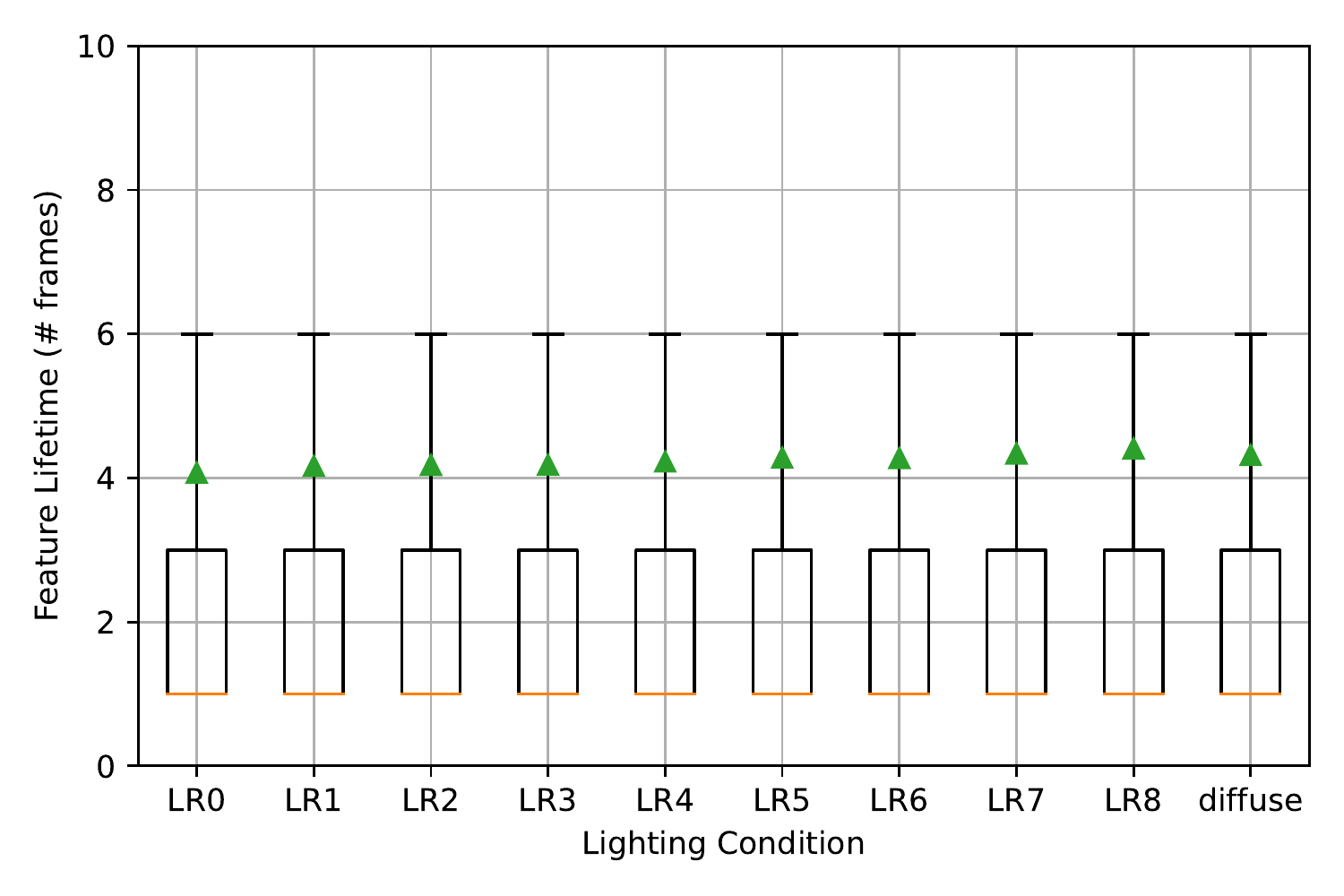}
        \includegraphics[width=0.48\textwidth]{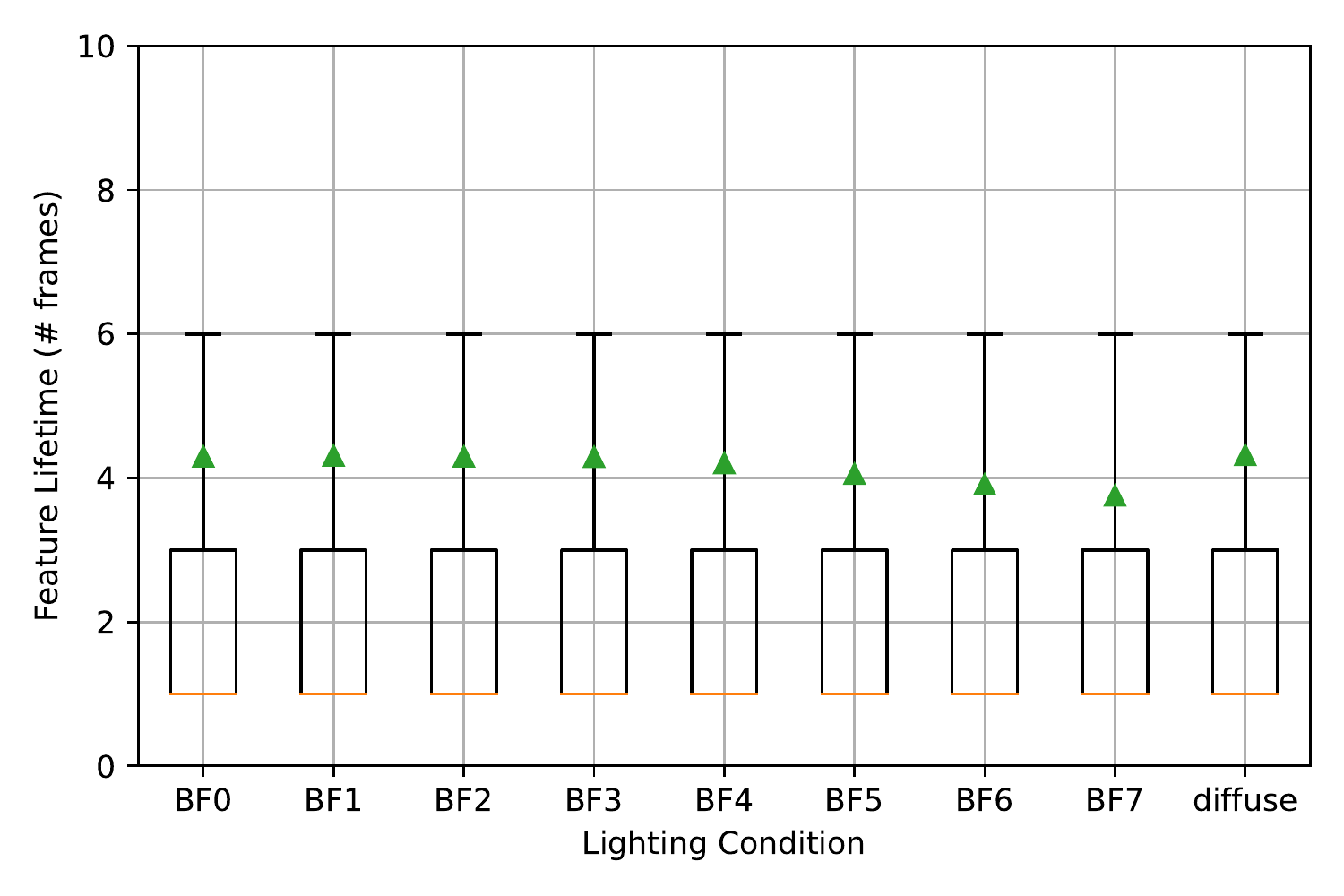}
    }
    \subfigure[Correspondence]{
        \includegraphics[width=0.48\textwidth]{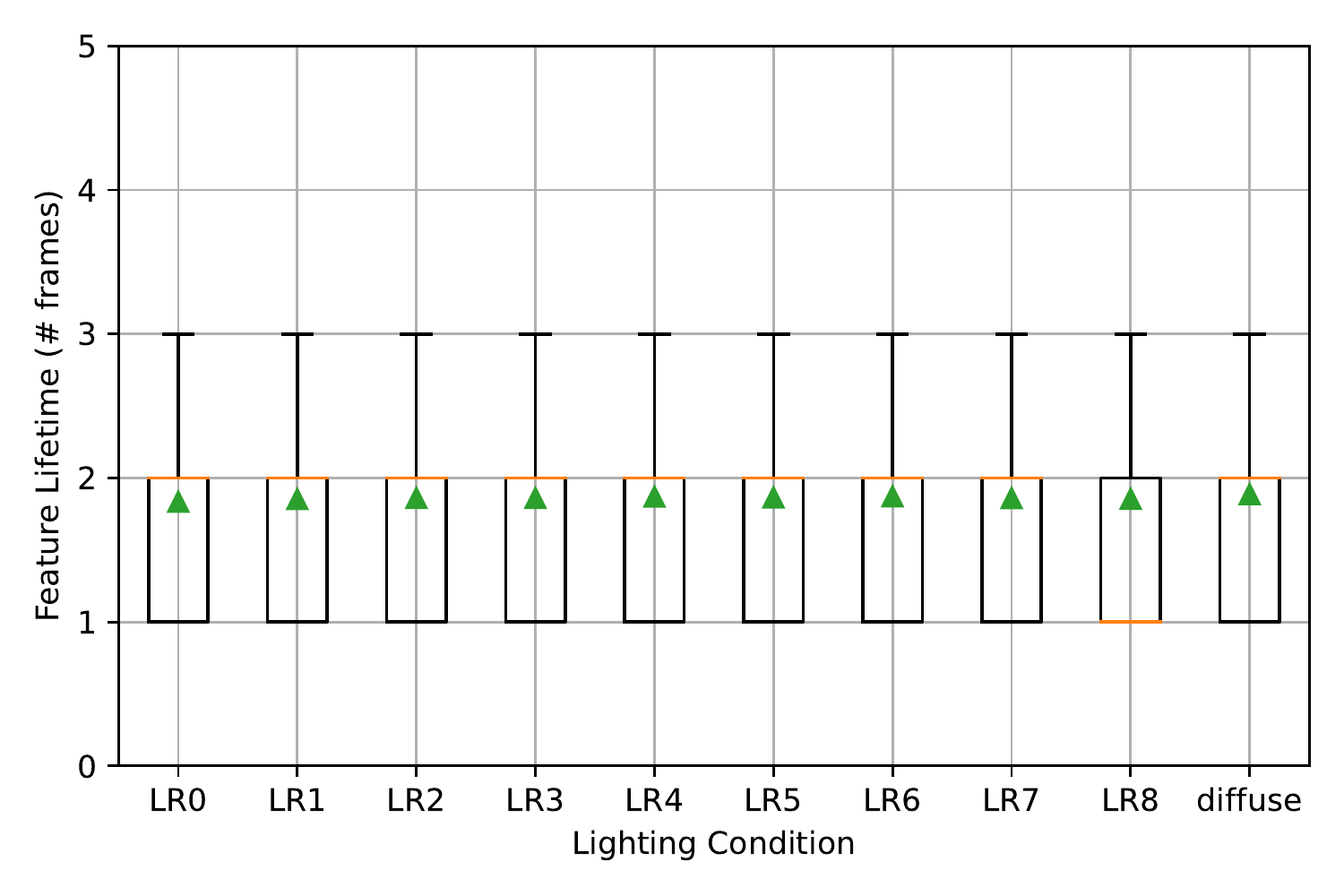}
        \includegraphics[width=0.48\textwidth]{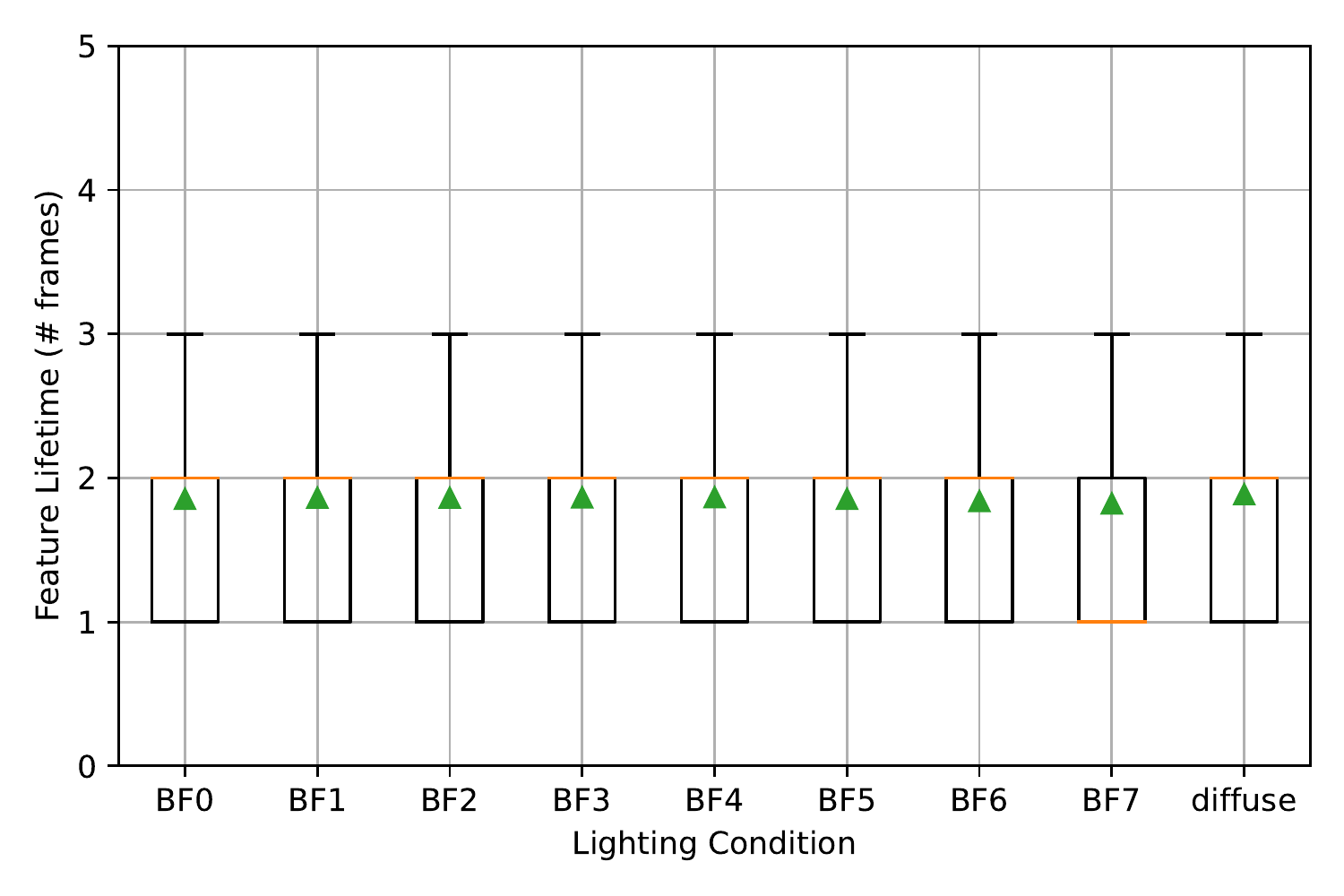}
    }
    \caption{\textbf{DTU Point Features Dataset: Feature lifetime does not depend on the existence of directional lighting.}  In the box-and-whisker plots above, the orange line is the median, the green triangle is the mean, and the box extends from the first to the third quartiles. The whiskers extend up to 1.5x the length of the boxes. Outlier ratios increase with speed for all tested feature trackers to a point, and then falls slightly. Each box-and-whisker is computed using features from all 60 scenes, one tracker, and one speed. The distribution of feature lifetime is approximately the same for all lighting conditions.}
    \label{fig:dtu_lighting_feature_lifetimes}
\end{figure}

\begin{figure}[H]
    \centering
    \subfigure[Lucas-Kanade]{\includegraphics[width=0.48\textwidth]{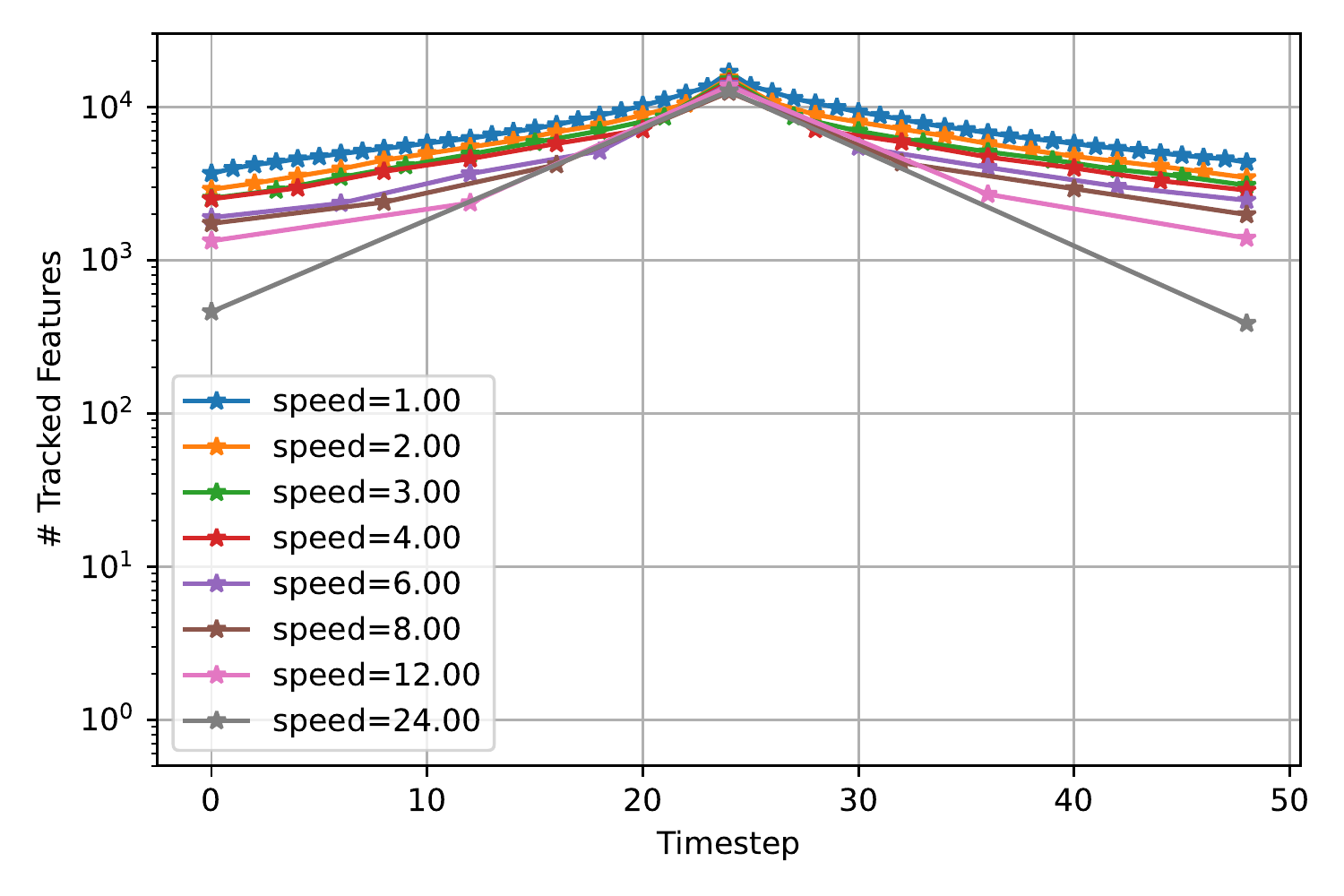}}
    \subfigure[Correspondence]{\includegraphics[width=0.48\textwidth]{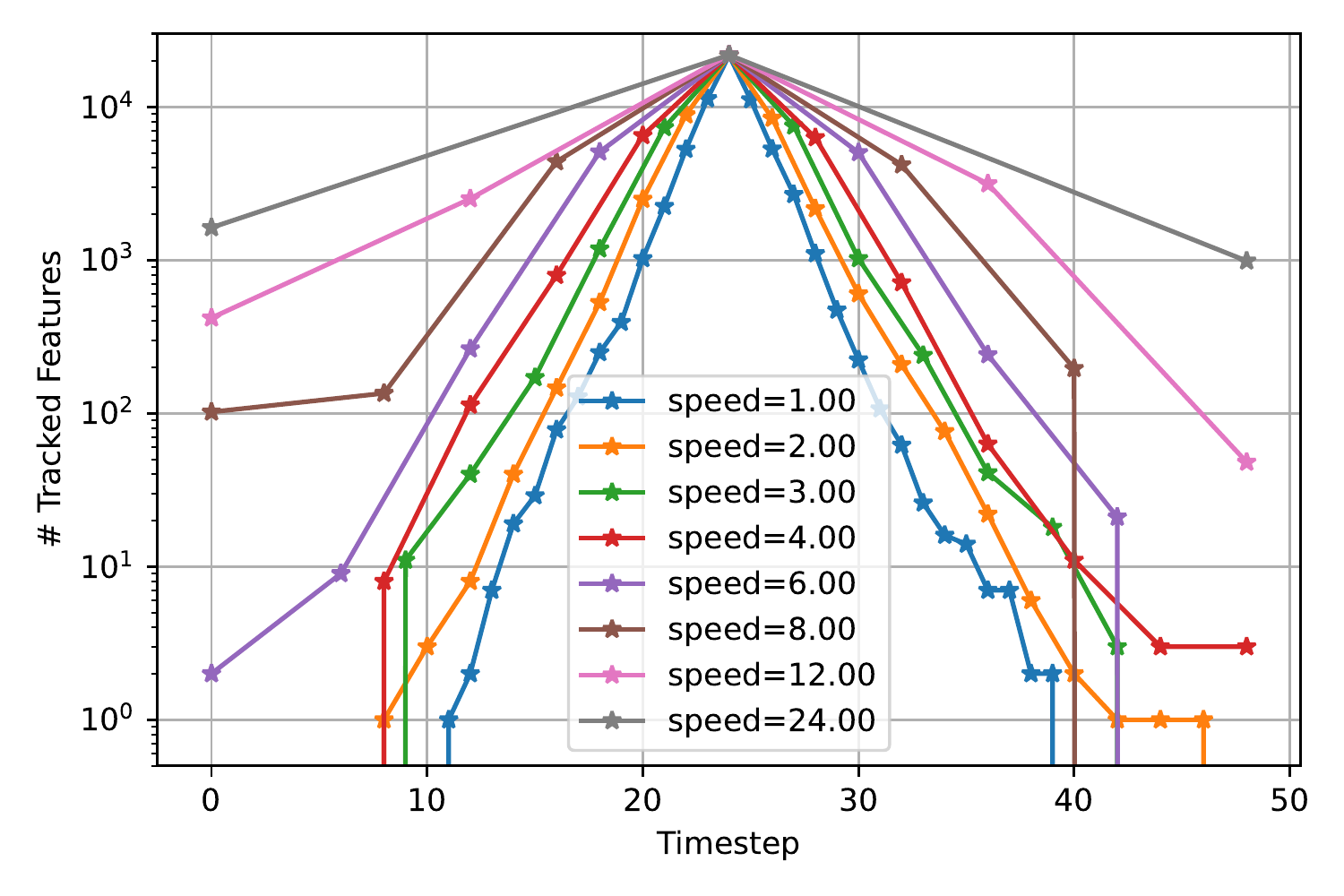}}
    \caption{Each curve shows the total number of tracked features at each timestep for the Lucas-Kanade Tracker (left) and the Correspondence Tracker (right) in log scale. Each dot on a curve is a frame in the sequence and each curve is computed using all features visible in the Key Frame under diffuse lighting and one speed. The number of features that can be used to compute mean $\mu(t)$ and covariance $\Sigma(t)$ declines quickly away from the Key Frame when using the Correspondence Tracker. When using the Lucas-Kanade Tracker, a slower speed means that more features are tracked for more frames. When using the Correspondence Tracker, the number of features tracked is dependent on the number of frames as well as the speed for the reasons noted in Section \ref{sec:feature_tracker_configuration}. The closer two frames are (i.e., the slower the speed), the fewer features are dropped between them. This is consistent with previously known results about the precision and recall of feature descriptors \cite{mikolajczyk_performance_2005, schonberger_comparative_eval_2017, WU2017150}. \textbf{We limit calculations of mean error $\mu(t)$, mean absolute error, $\kappa(t)$, and covariance $\Sigma(t)$ to timesteps that contain at least 100 features.}}
    \label{fig:dtu_active_features}
\end{figure}

\begin{figure}[H]
    \centering
    \subfigure[$\mu(t)$, Horizontal Coordinate]{\includegraphics[width=0.48\textwidth]{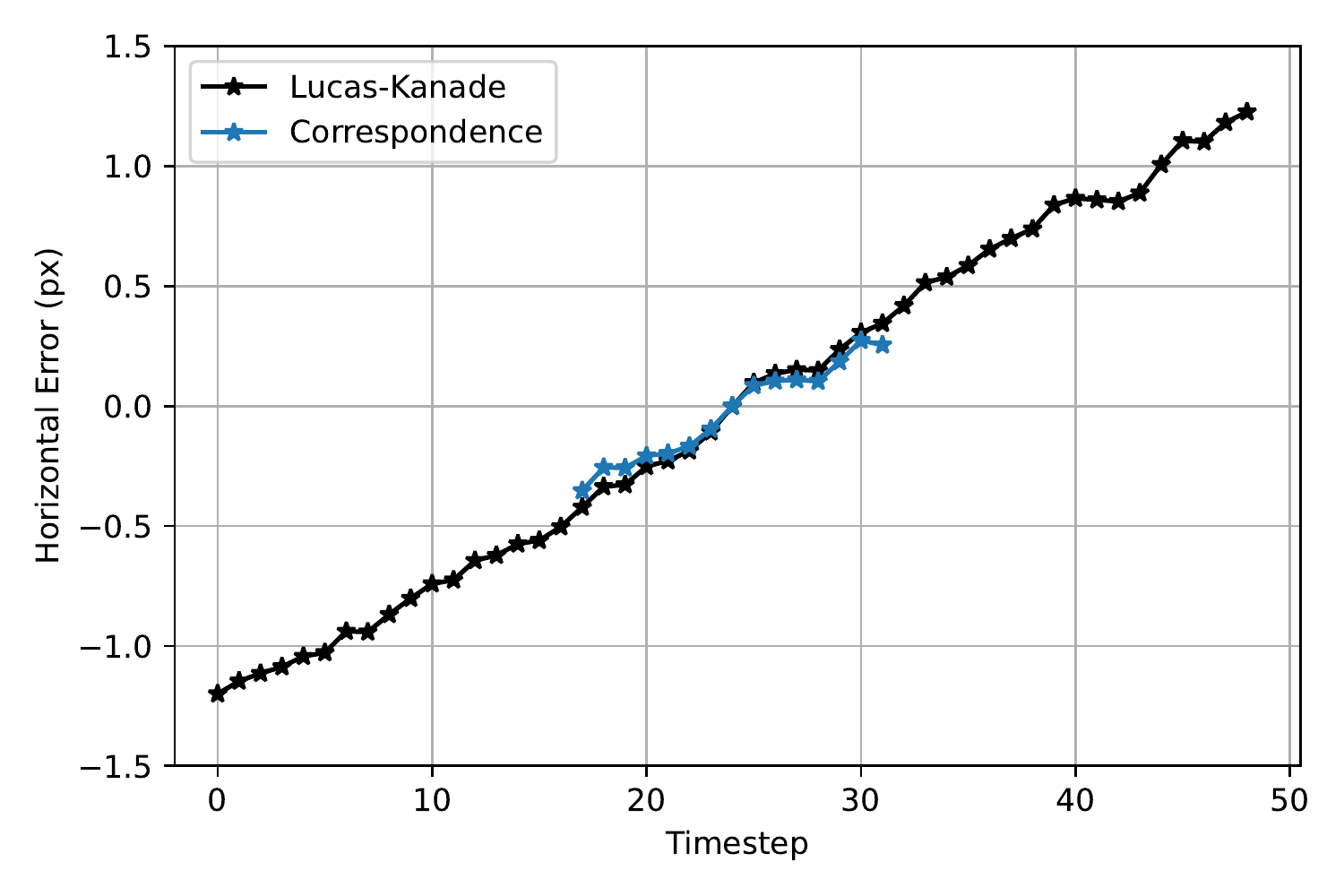}}
    \subfigure[$\mu(t)$, Vertical Coordinate]{\includegraphics[width=0.48\textwidth]{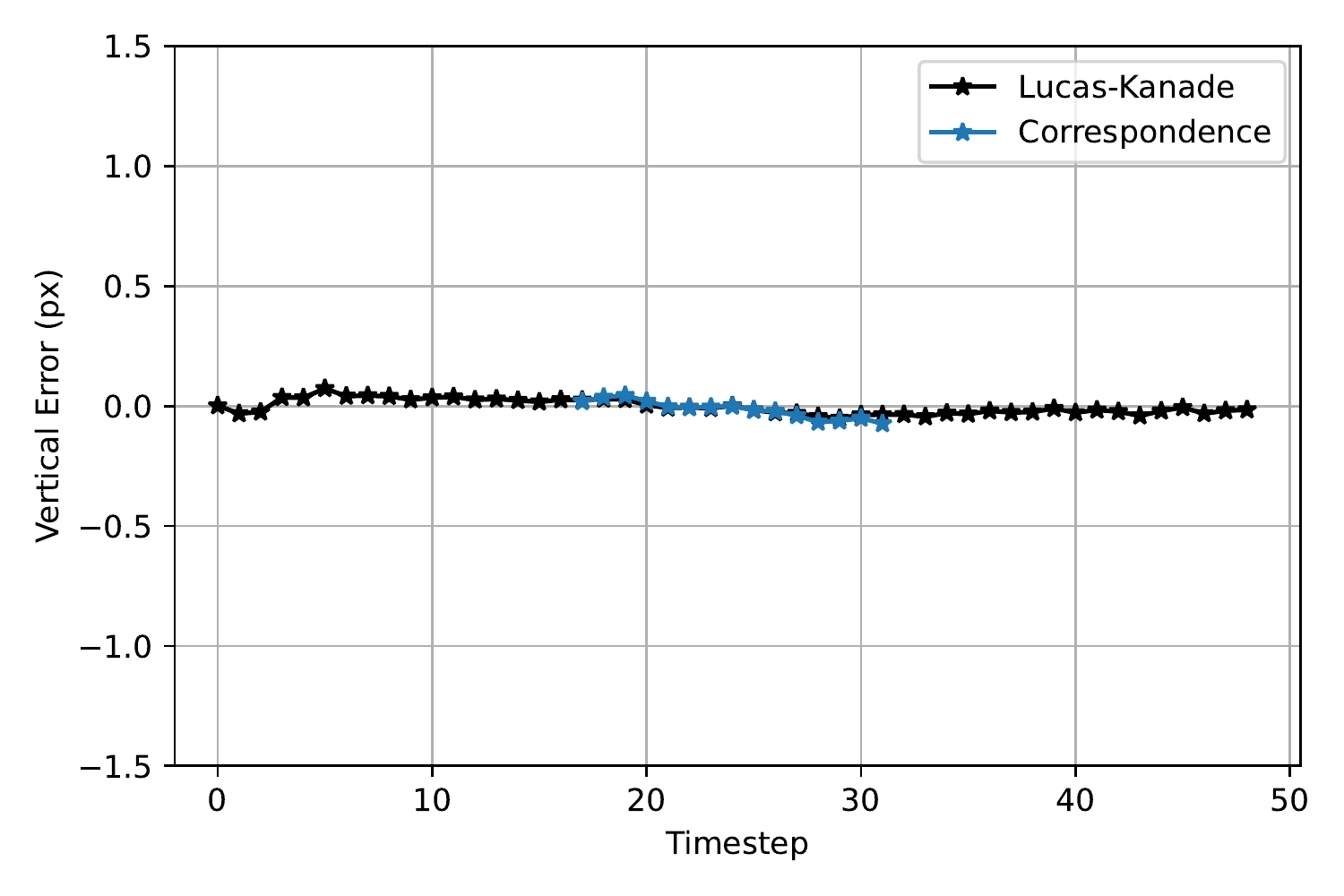}}
    \caption{\textbf{DTU Point Features Dataset: At nominal speed and with diffuse lighting, the tracker used has little effect on $\mu(t)$.} Lines shown are mean feature track errors $\mu(t)$ at each timestep $t$ calculated over all scenes. The blue lines are feature track errors calculated using the Lucas-Kanade Tracker and the orange lines are feature track errors calculated using the Correspondence Tracker. Lines are cut-off to timesteps where at least 100 features with 3D data are available (see Fig. \ref{fig:dtu_active_features}). The orange lines are on top of the blue lines, therefore the tracker used does not affect mean error.}
    \label{fig:dtu_diffuse_1.00_meanerror}
\end{figure}

\begin{figure}[H]
    \centering
   \subfigure[$\kappa(t)$, Horizontal Coordinate]{\includegraphics[width=0.48\textwidth]{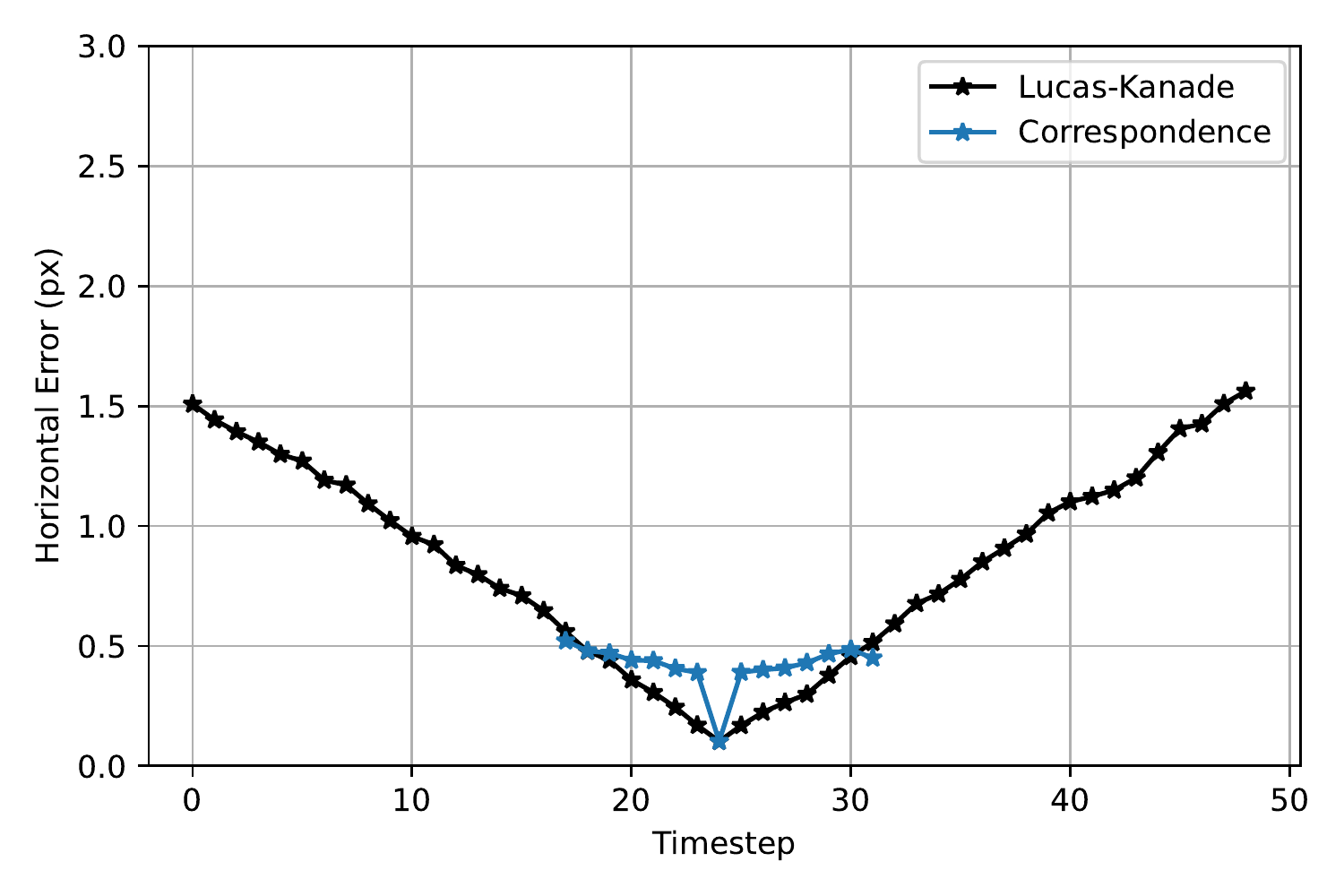}}
    \subfigure[$\kappa(t)$, Vertical Coordinate]{\includegraphics[width=0.48\textwidth]{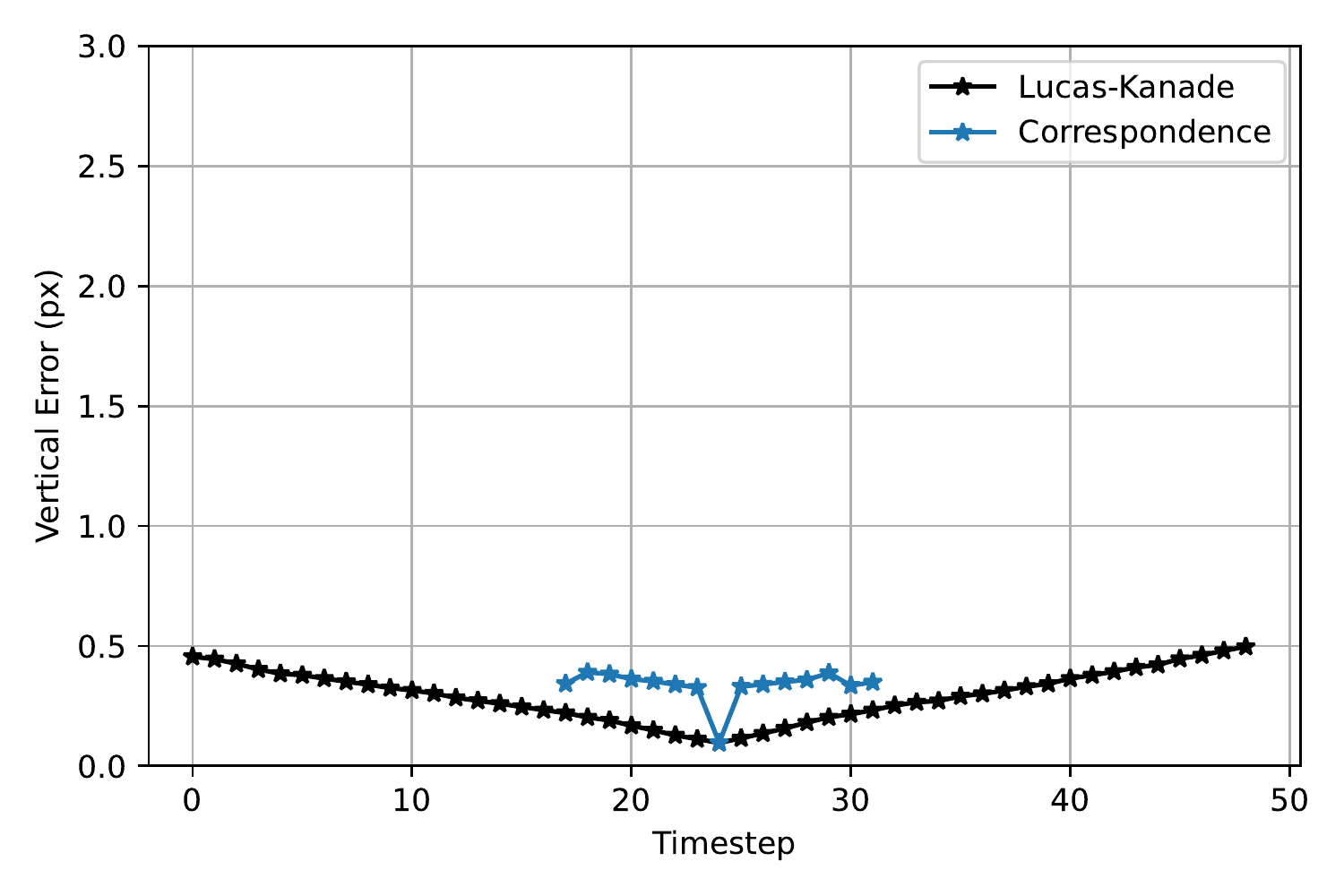}}
    \subfigure[$\Sigma(t)$, Horizontal Covariance]{\includegraphics[width=0.48\textwidth]{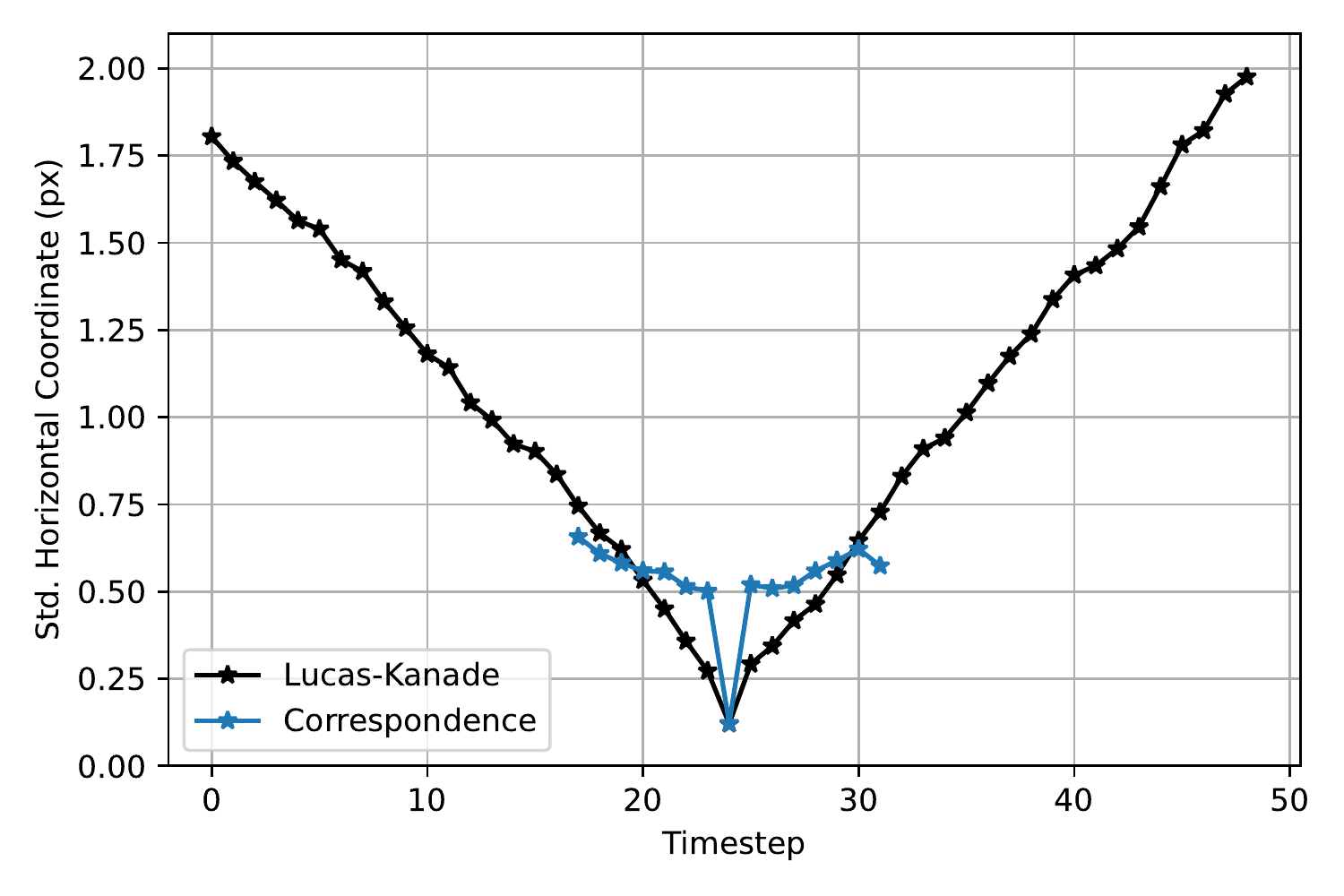}}
    \subfigure[$\Sigma(t)$, Vertical Coordinate]{\includegraphics[width=0.48\textwidth]{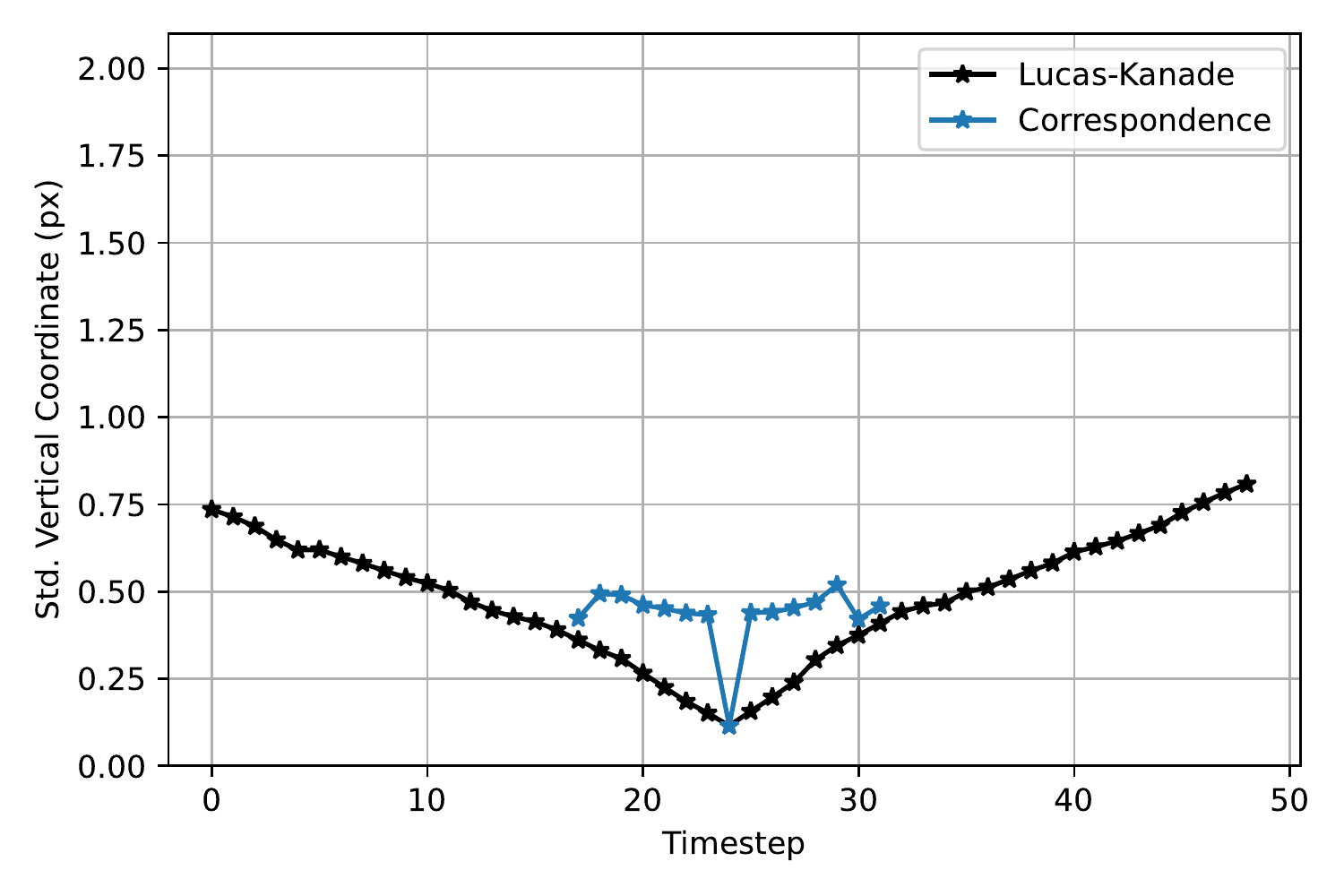}}
    \caption{\textbf{DTU Point Features Dataset: At nominal speed and under diffuse lighting, the tracker used does affect mean absolute error $\kappa(t)$ and covariance $\Sigma(t)$.} Lines shown are horizontal and vertical coordinates of $\kappa(t)$ (top row), and $\Sigma(t)$ (bottom row) calculated using all tracks from all scenes. Each dot corresponds to a single frame. Mean absolute error and covariance for the Correspondence Tracker are roughly constant with respect to time, while the same values for the Lucas-Kanade Tracker increases steadily with time away from the Key Frame.}
    \label{fig:dtu_diffuse_1.00_MAE_cov}
\end{figure}

\begin{figure}[H]
    \centering
    \subfigure[$\mu(t)$, Horizontal Coordinate]{
        \includegraphics[width=0.48\textwidth]{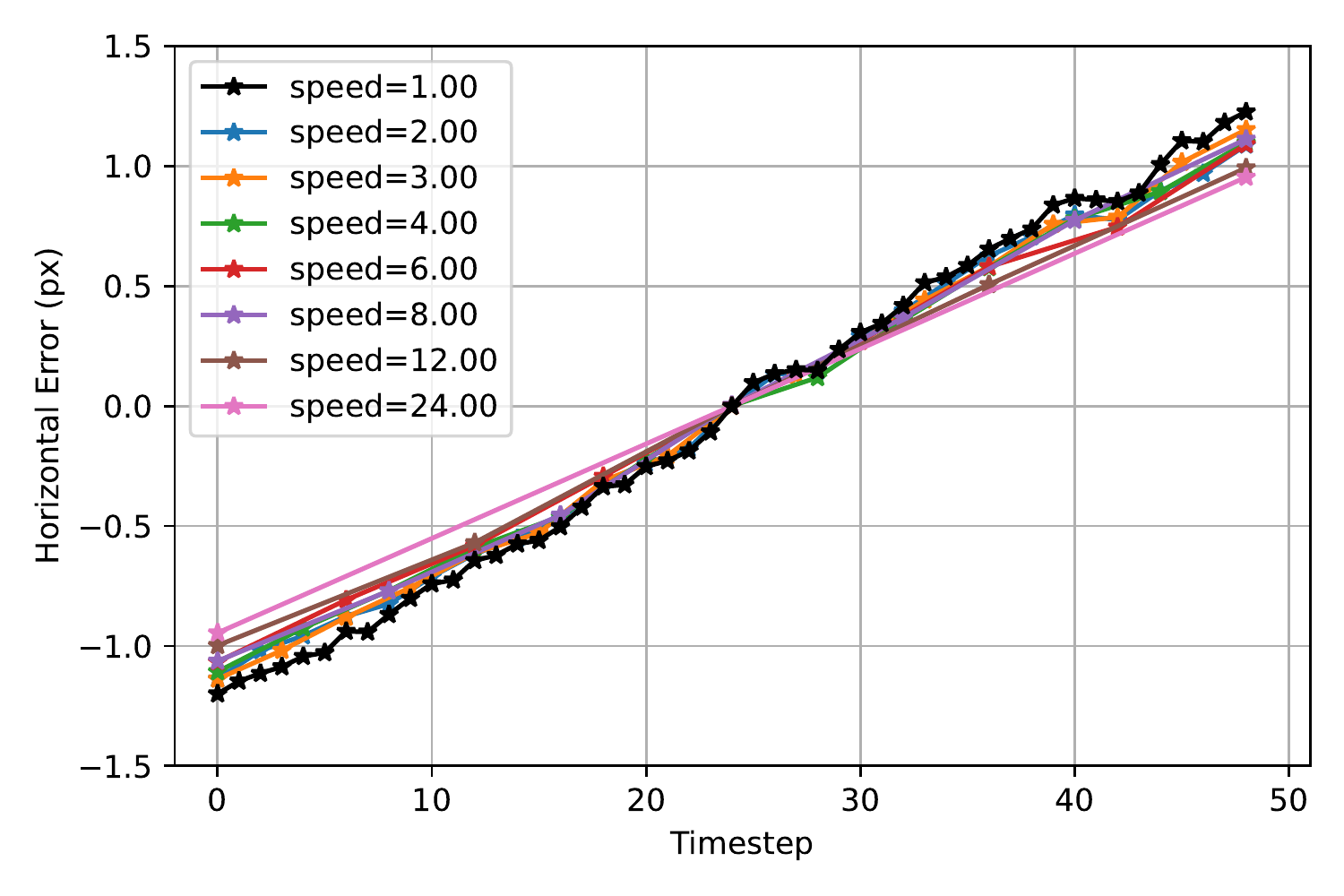}
        \includegraphics[width=0.48\textwidth]{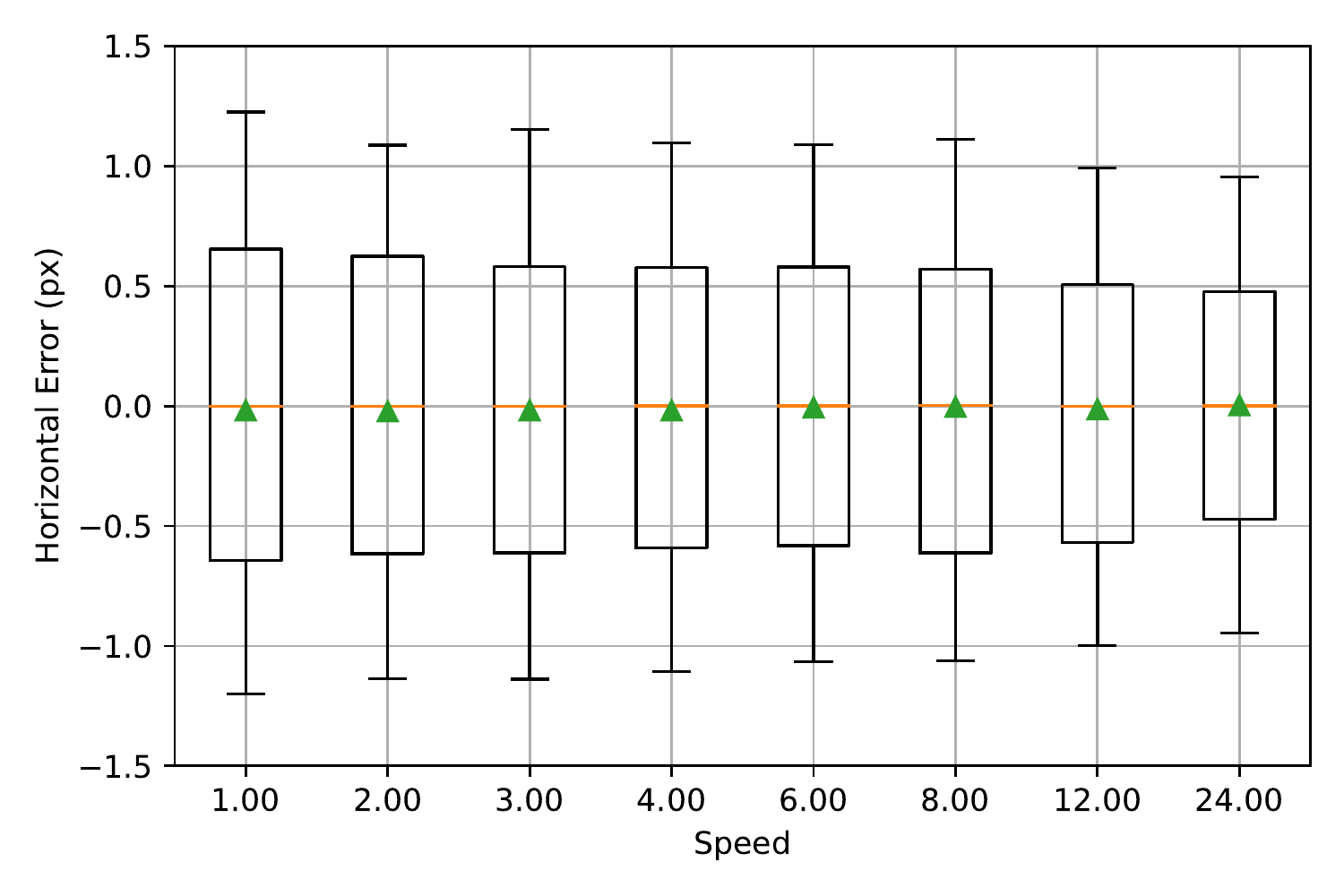}
    }
    \subfigure[$\mu(t)$, Vertical Coordinate]{
        \includegraphics[width=0.48\textwidth]{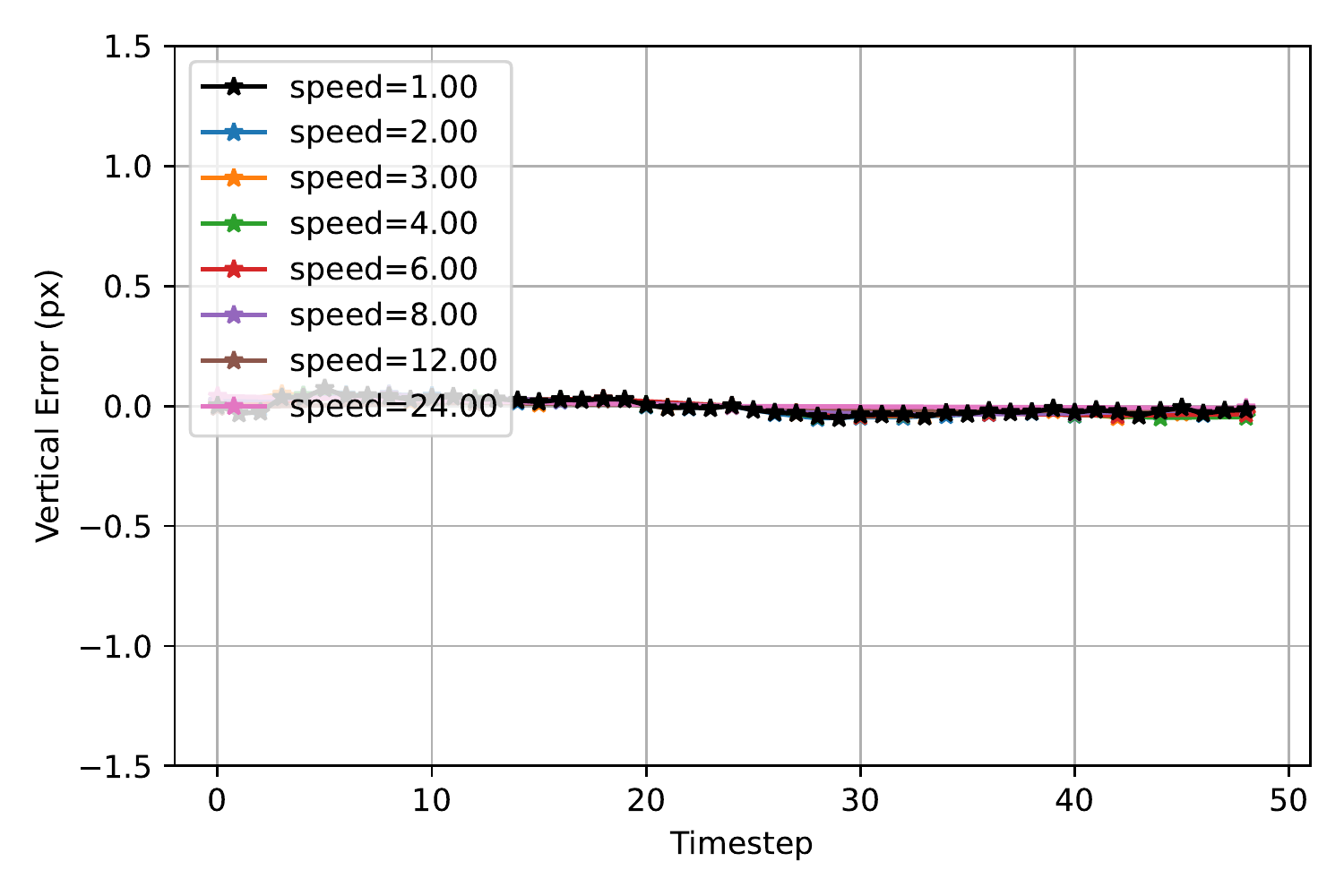}
        \includegraphics[width=0.48\textwidth]{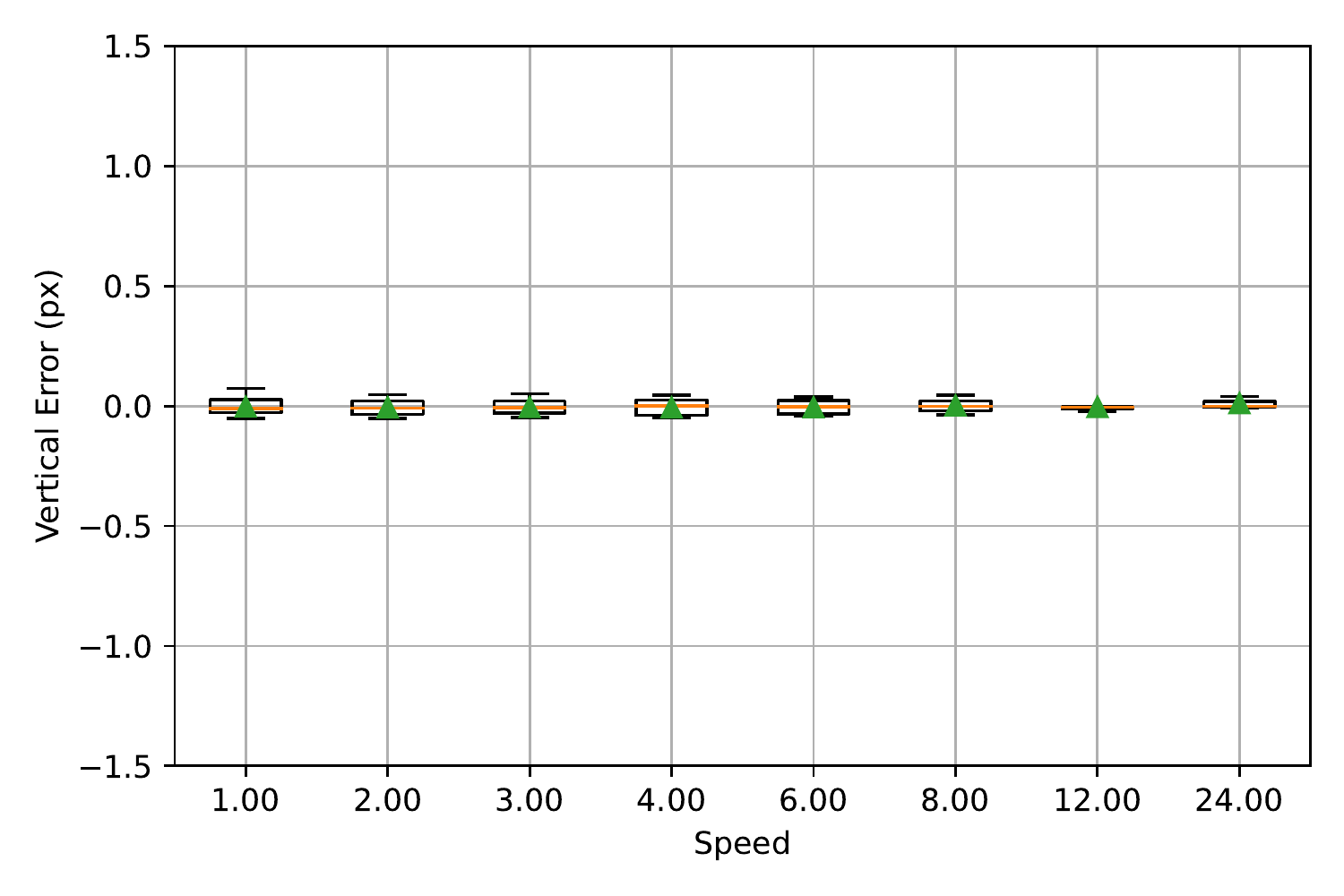}
    }
    \caption{\textbf{DTU Point Features Dataset: When using the Lucas-Kanade Tracker with diffuse lighting, speed affects mean error.} The left column contains plots of the horizontal (top row) and vertical (bottom row) coordinates of mean error $\mu(t)$ at each timestep and multiple speeds. Each dot corresponds to a processed frame; lines for higher speeds contain data from fewer frames and therefore show fewer dots. The right column plots the ordinate value of each line in the left figures as a box plot: means are shown as green triangles and medians are shown as orange lines. As speed is increased, the slope of the horizontal components of $\mu(t)$ in the left plots (eq. \eqref{eq:mean_error_at_time}) decreases and the height of each box in the right plot decreases, i.e. the absolute magnitude of $\mu(t)$ slighty decreases.  This trend indicates the existence of two speed-related components that affect $\mu(t)$: the difference between frames and the number of frames that have elapsed; the former has a much larger effect than the latter. The latter occurs because the exact point that the Lucas-Kanade Tracker tracks drifts with each frame. Fewer frames means that the tracked point has fewer opportunities to drift.}
    \label{fig:dtu_LK_mean_varyspeed}
\end{figure}

\begin{figure}[H]
    \centering
    \subfigure[$\kappa(t)$, Horizontal Coordinate]{
        \includegraphics[width=0.48\textwidth]{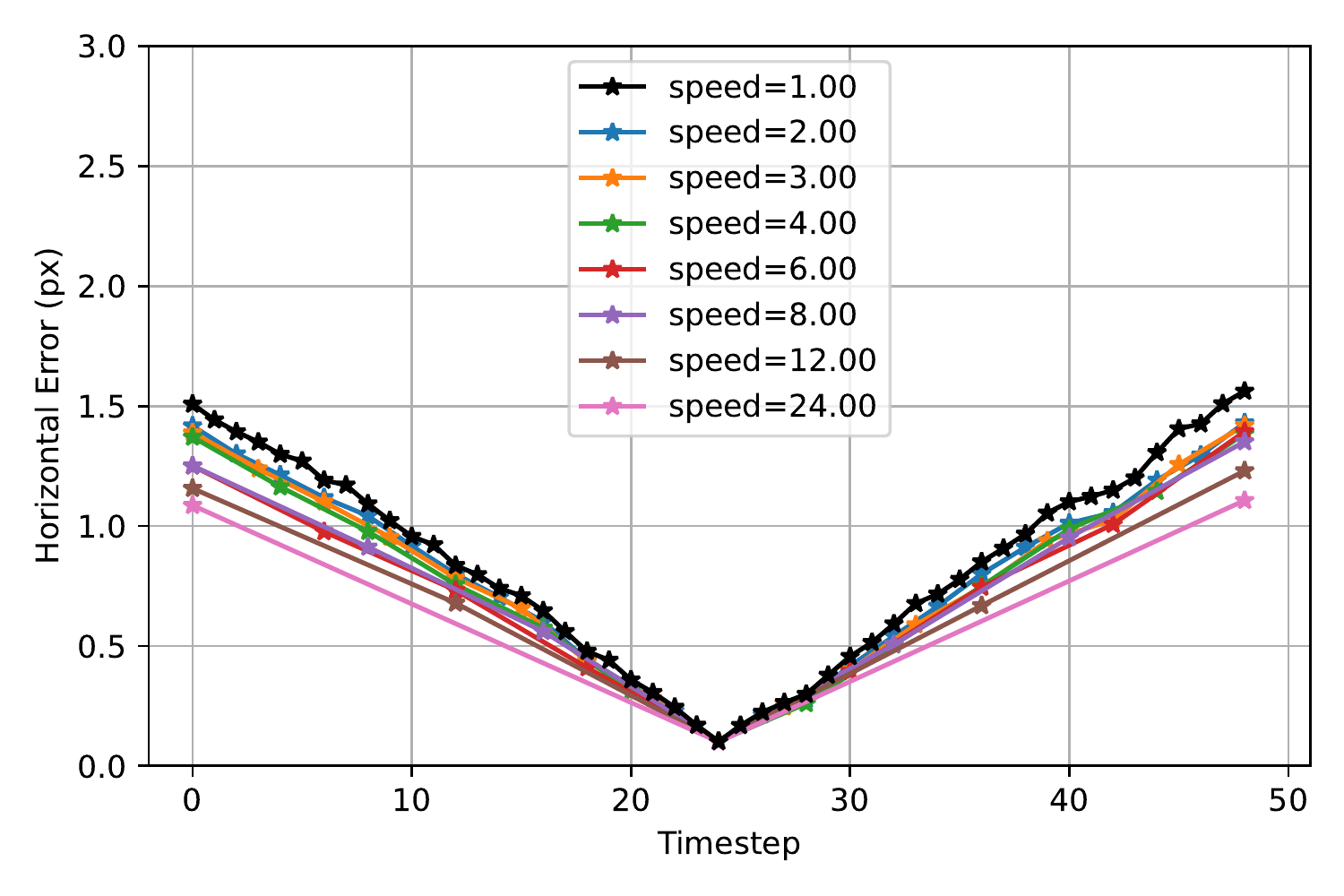}
        \includegraphics[width=0.48\textwidth]{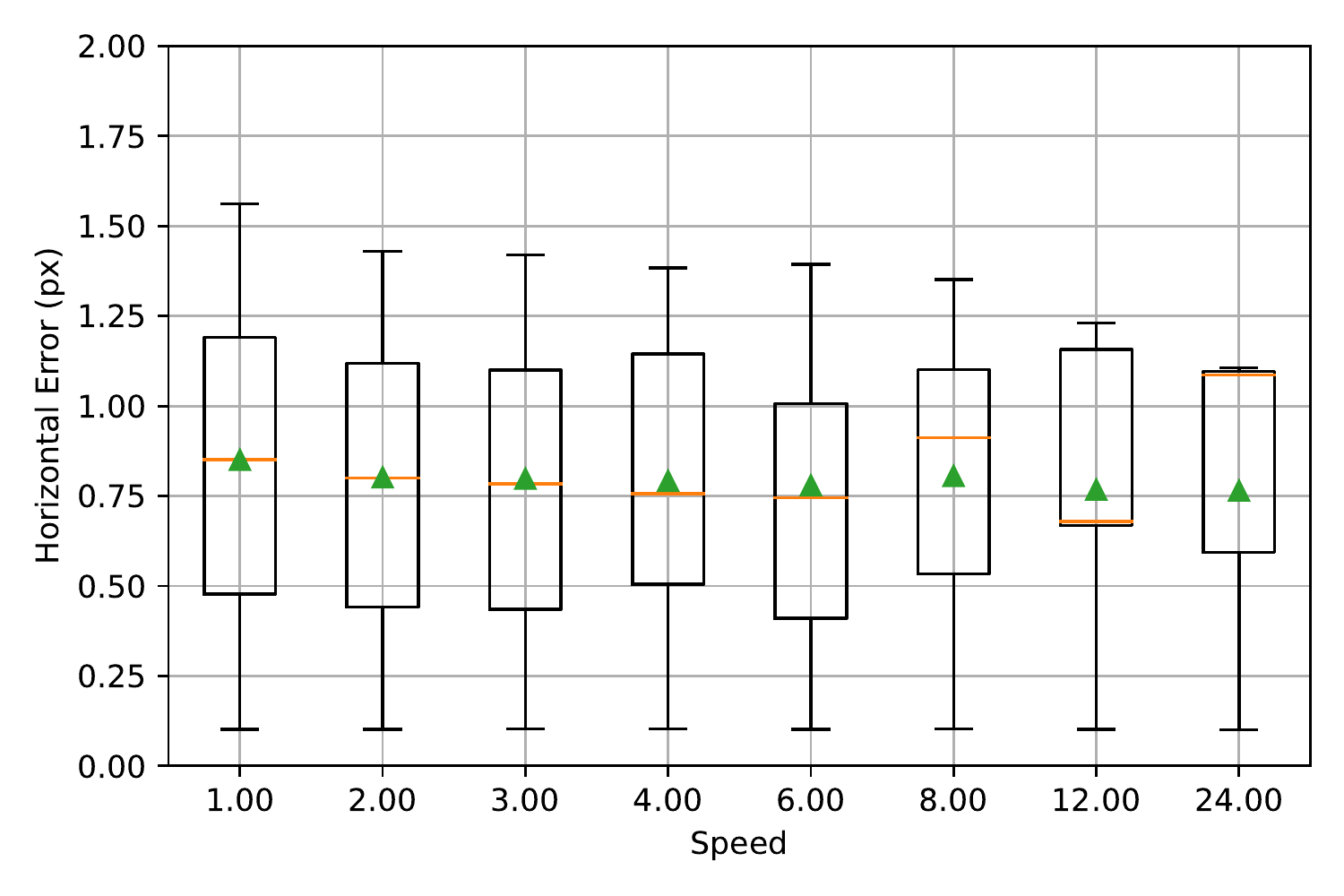}
    }
    \subfigure[$\kappa(t)$, Vertical Coordinate]{
        \includegraphics[width=0.48\textwidth]{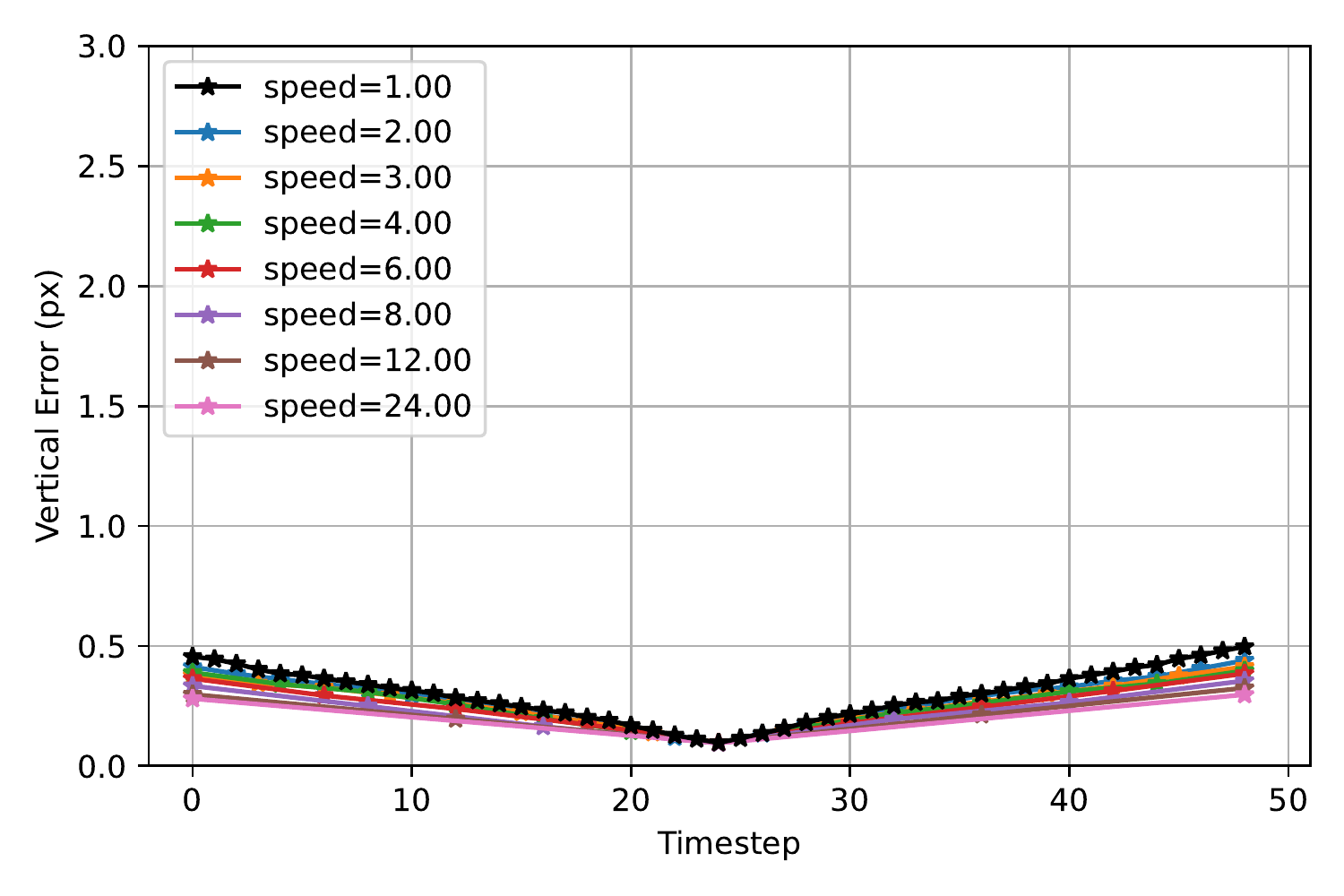}
        \includegraphics[width=0.48\textwidth]{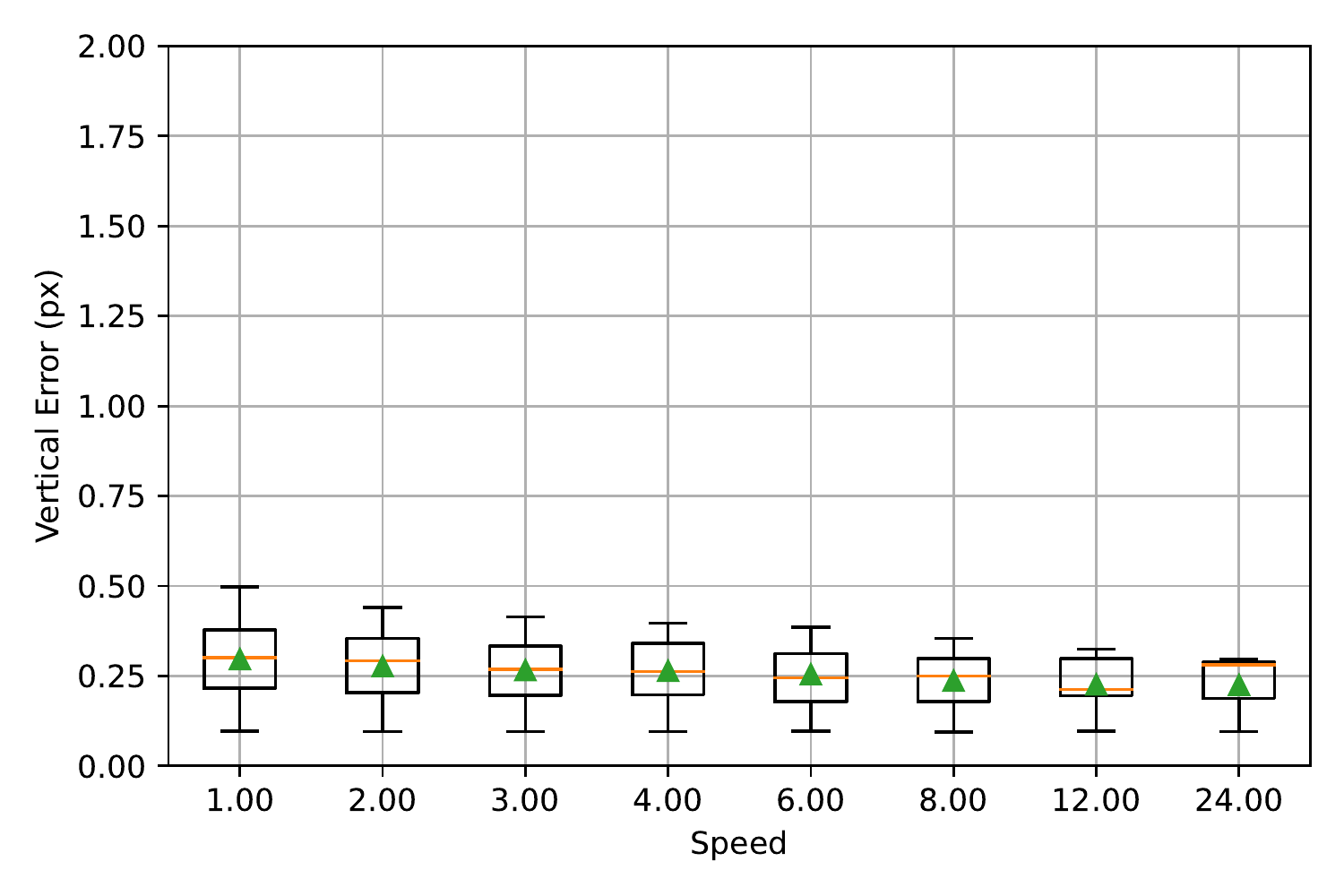}
    } 
    \caption{\textbf{DTU Point Features Dataset: When using the Lucas-Kanade Tracker with diffuse lighting, speed affects mean absolute error.} The left column contains plots of the horizontal (top row) and vertical (bottom row) coordinates of mean absolute error $\kappa(t)$ at each timestep and multiple speeds. Each dot corresponds to a processed frame; lines for higher speeds contain data from fewer frames and therefore show fewer dots. The right column plots the ordinate value of each line in the left figures as a box plot: means are shown as green triangles and medians are shown as orange lines. As speed is increased, the mean absolute error at each timestep slightly decreases. This indicates the existence of two speed-related components that affect $\kappa(t)$: the difference between frames and the number of frames that have elapsed; the former has a much larger effect than the latter. The latter occurs because the exact point that the Lucas-Kanade Tracker tracks drifts with each frame. Fewer frames means that the tracked point has fewer opportunities to drift.}
    \label{fig:dtu_LK_MAE_varyspeed}
\end{figure}

\begin{figure}[H]
    \centering
   \subfigure[$\Sigma(t)$, Horizontal Covariance]{
        \includegraphics[width=0.48\textwidth]{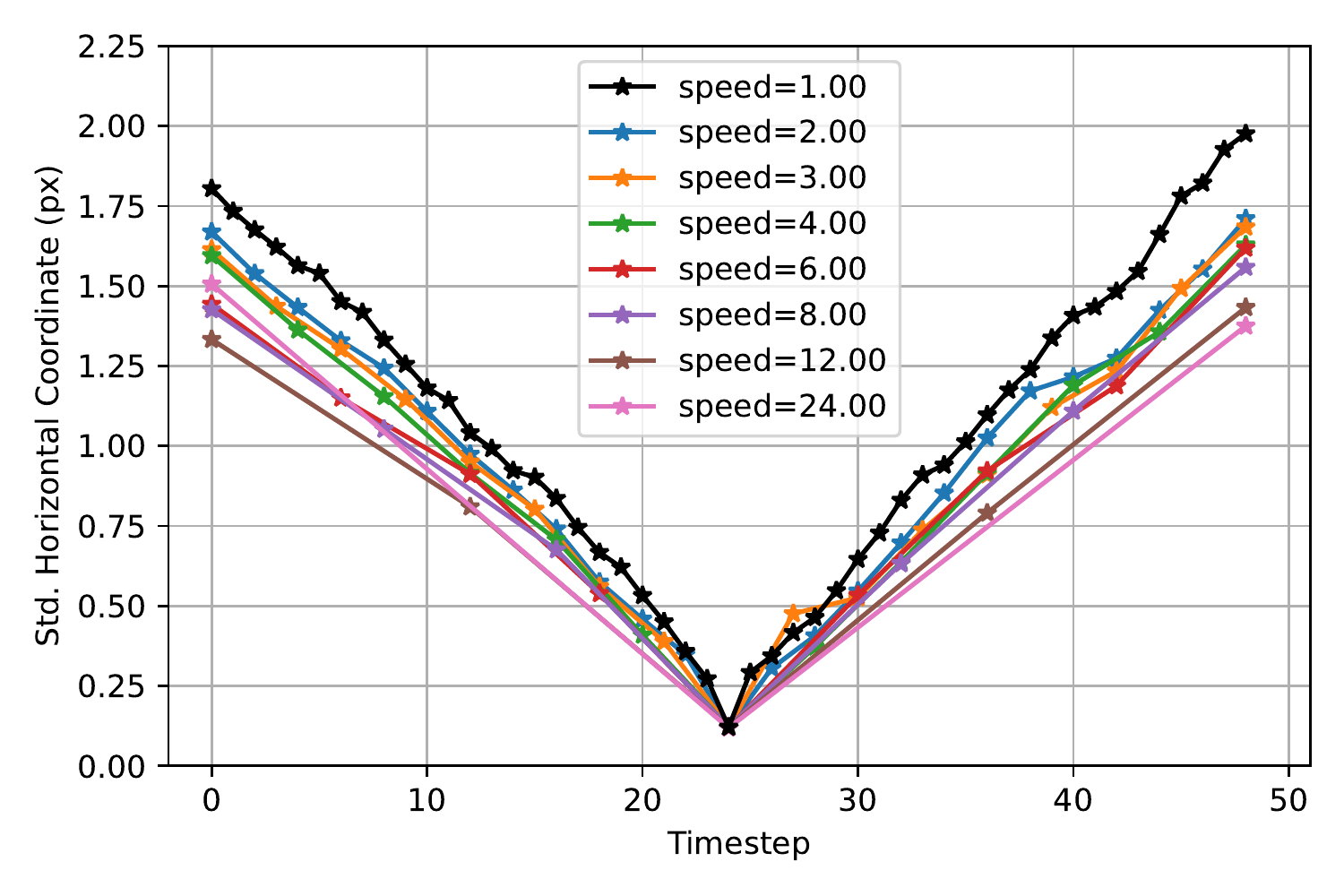}
        \includegraphics[width=0.48\textwidth]{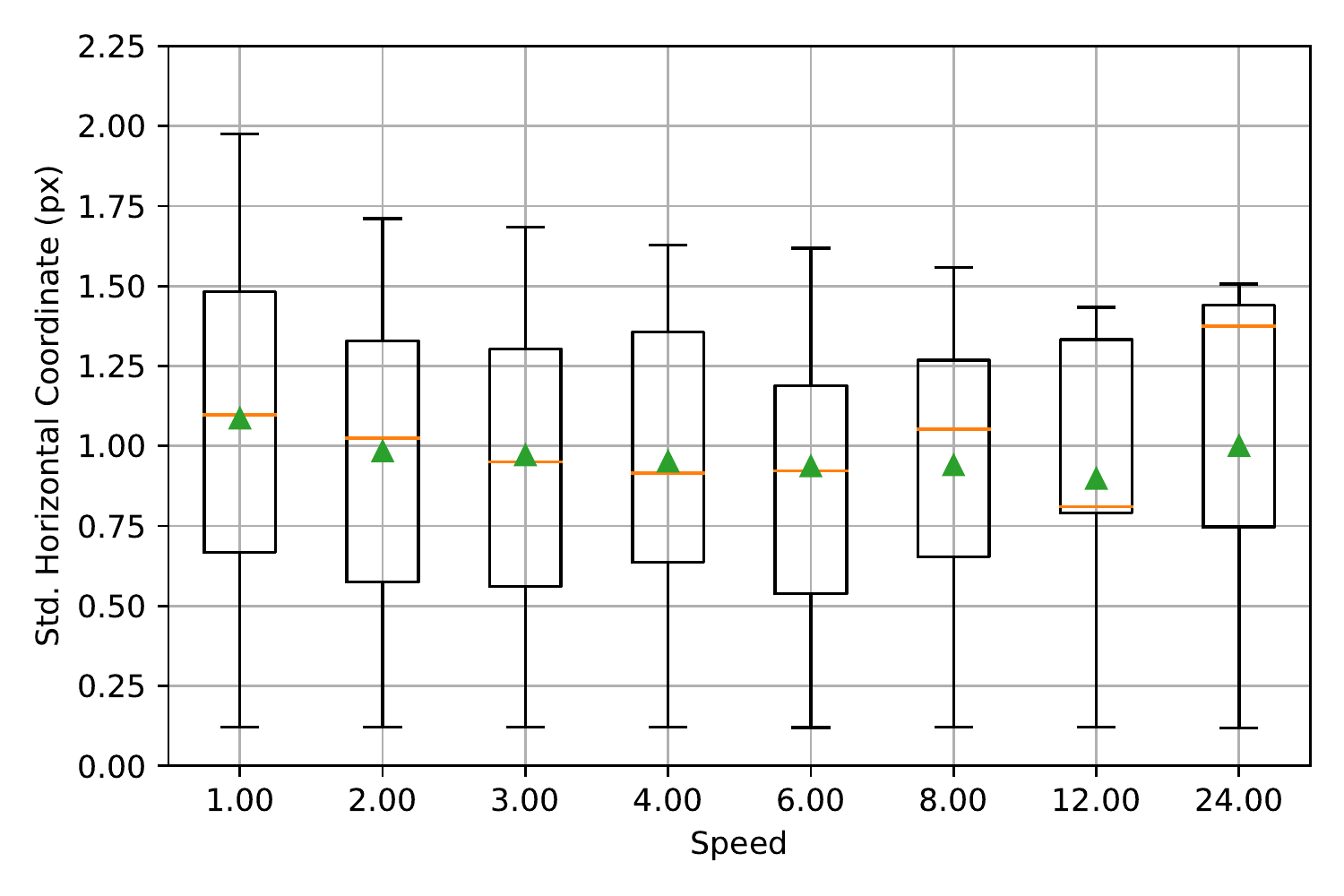}
    }
    \subfigure[$\Sigma(t)$, Vertical Coordinate]{
        \includegraphics[width=0.48\textwidth]{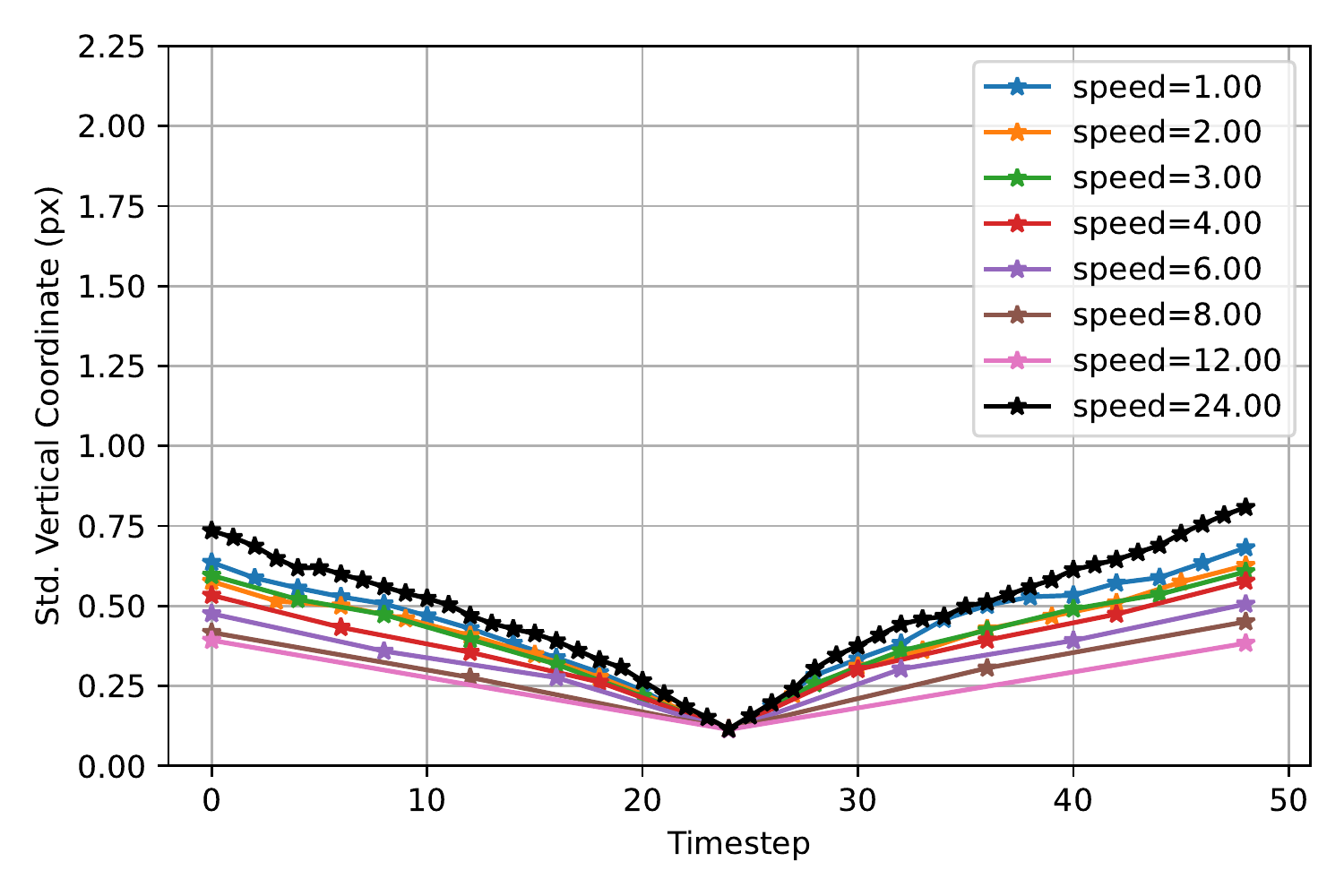}
        \includegraphics[width=0.48\textwidth]{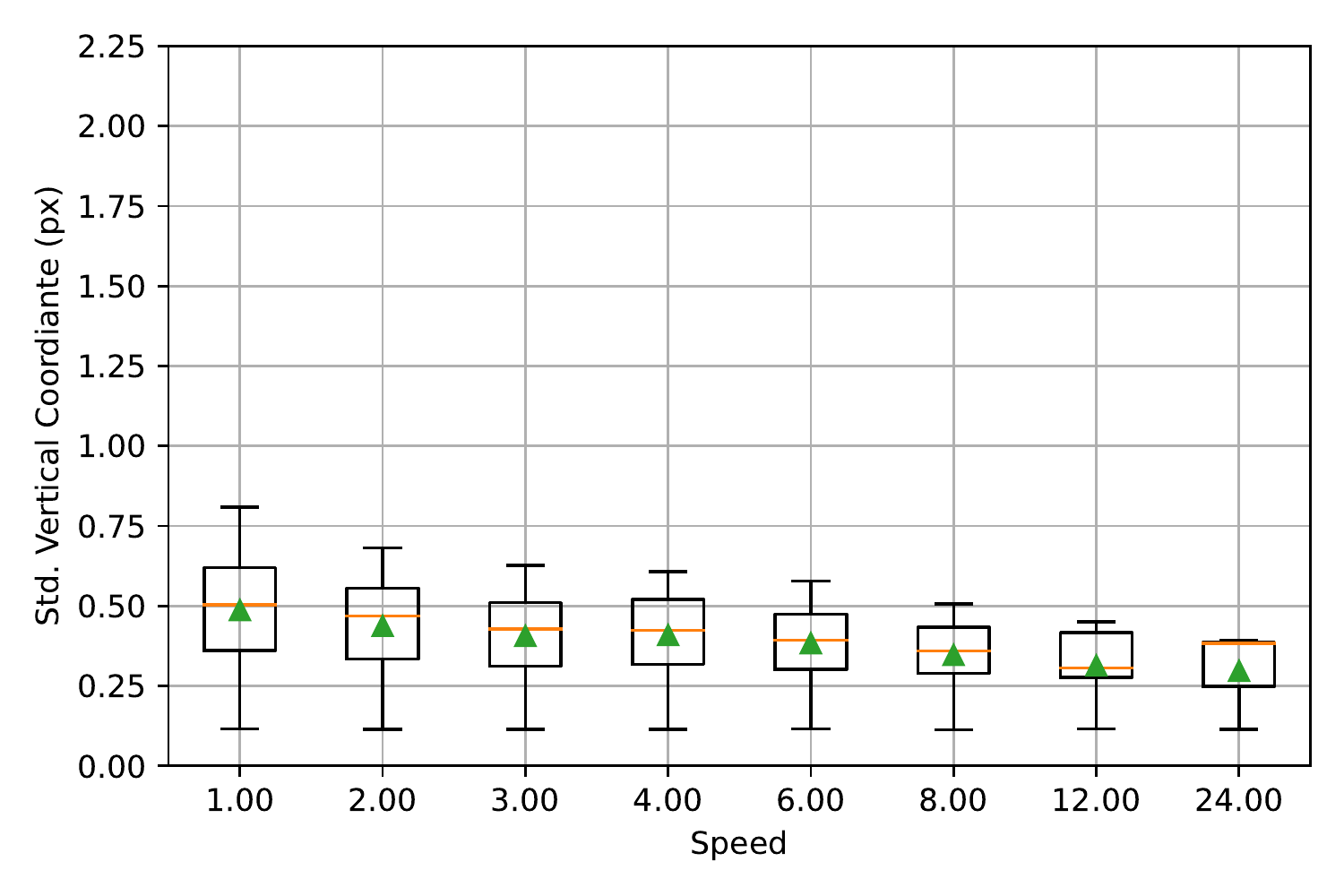}
    }
    \caption{\textbf{DTU Point Features Dataset: When using the Lucas-Kanade Tracker with diffuse lighting, speed affects covariance.} The left column contains plots of the square root of the horizontal (top row) and vertical (bottom row) coordiantes of $\Sigma(t)$ at each timestep and multiple speeds. Each dot cooresponds to a processed frame; lines for higher speeds contain data from fewer frames and therefore show fewer dots. The right column plots the ordinate value of each line in the left figures as a box plot: means are shown as green triangles and medians are shown as orange lines. As speed is increased, the covariance of both the horizontal and vertical coordiantes slightly decreases; the lines in the left plot become slightly less steep and mean values of covariance get slightly smaller. This indicates the existence of two speed-related components to these statistics: the difference between frames and the number of frames that have elapsed; the former has a much larger effect than the latter. The latter occurs because the exact point that the Lucas-Kanade Tracker tracks drifts with each frame. Fewer frames means that the tracked point has fewer opportunities to drift.}
    \label{fig:dtu_LK_cov_varyspeed}
\end{figure}

\begin{figure}[H]
    \centering
    \subfigure[$\mu(t)$, Horizontal Coordinate]{
        \includegraphics[width=0.48\textwidth]{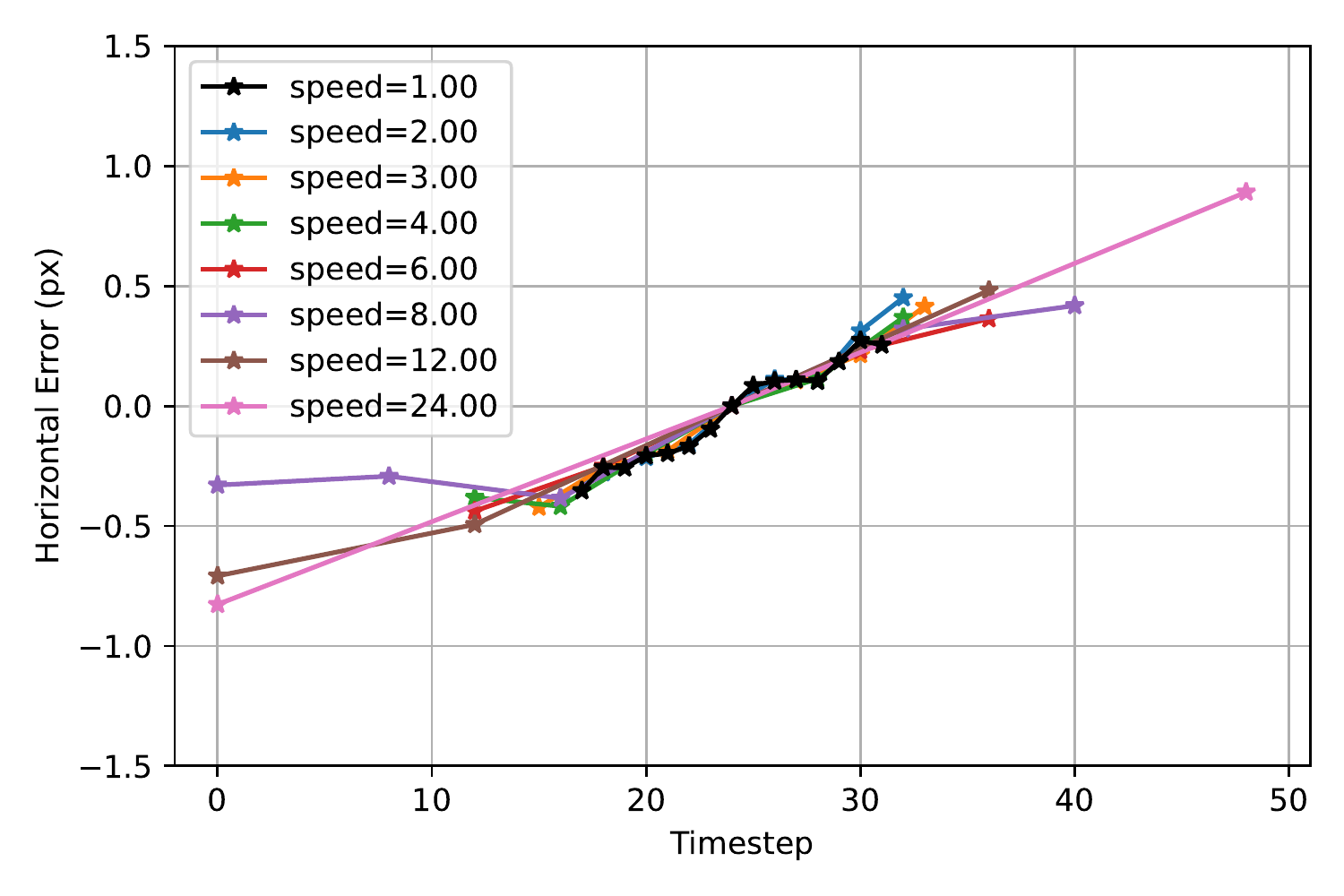}
        \includegraphics[width=0.48\textwidth]{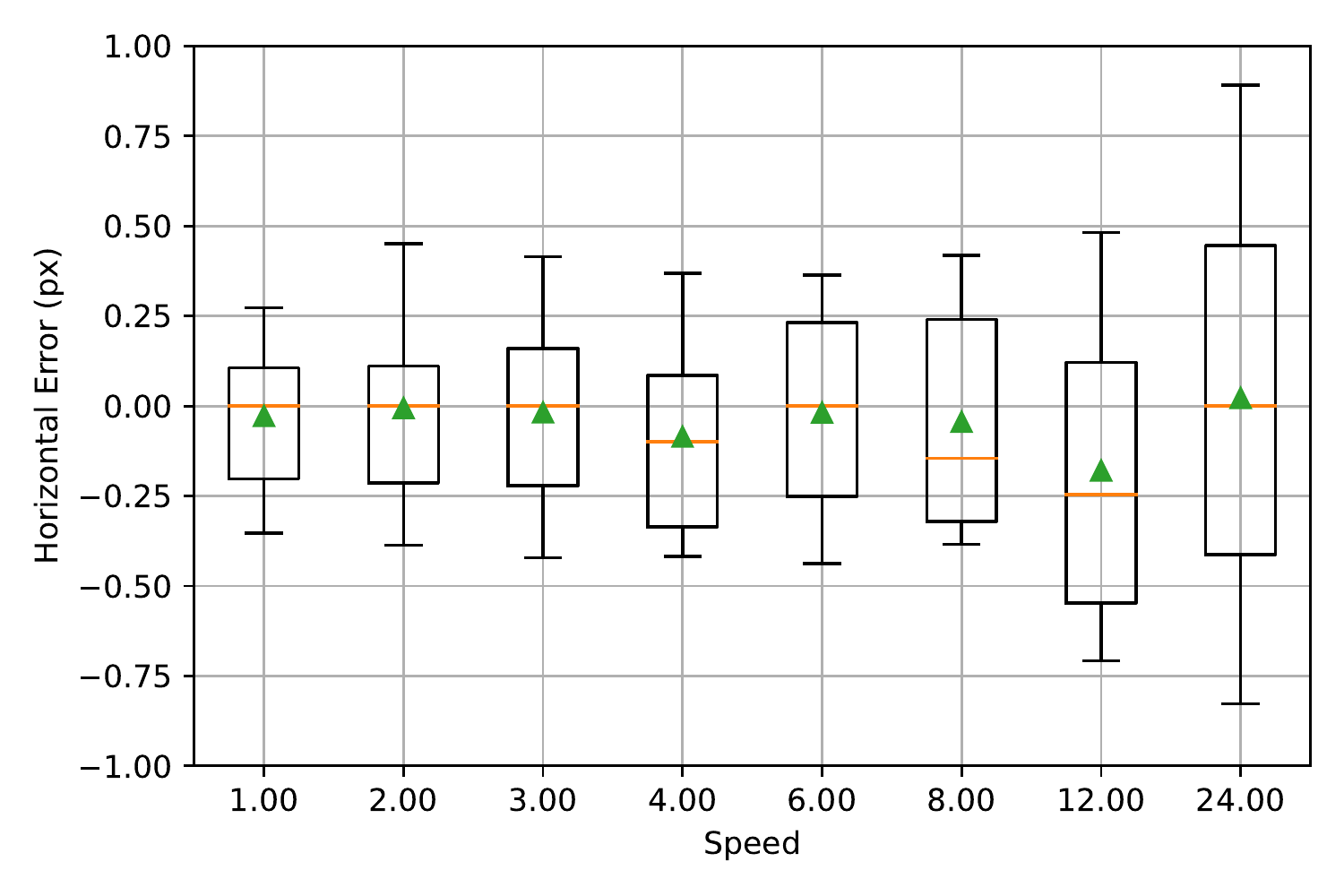}
    }
    \subfigure[$\mu(t)$, Vertical Coordinate]{
        \includegraphics[width=0.48\textwidth]{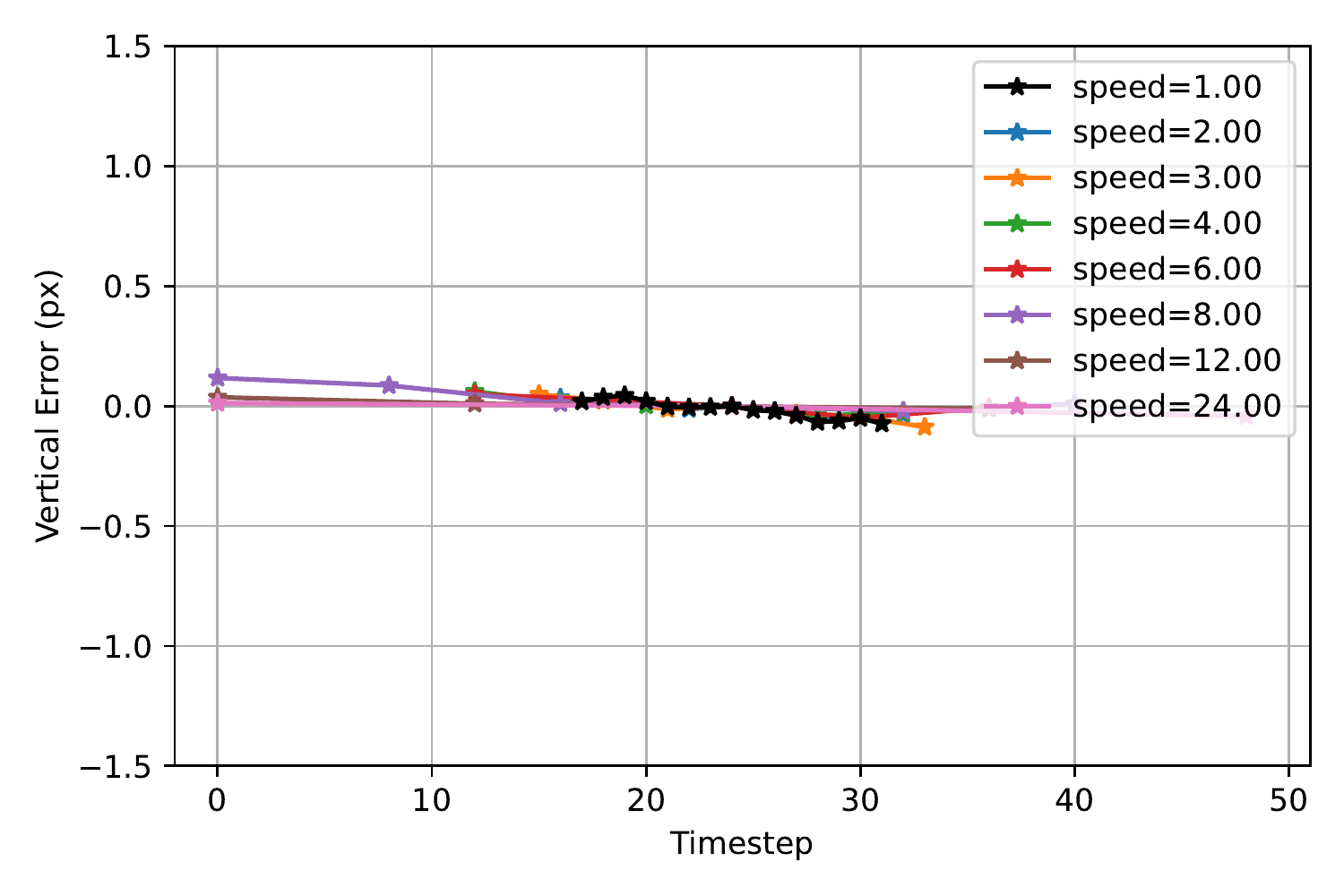}
        \includegraphics[width=0.48\textwidth]{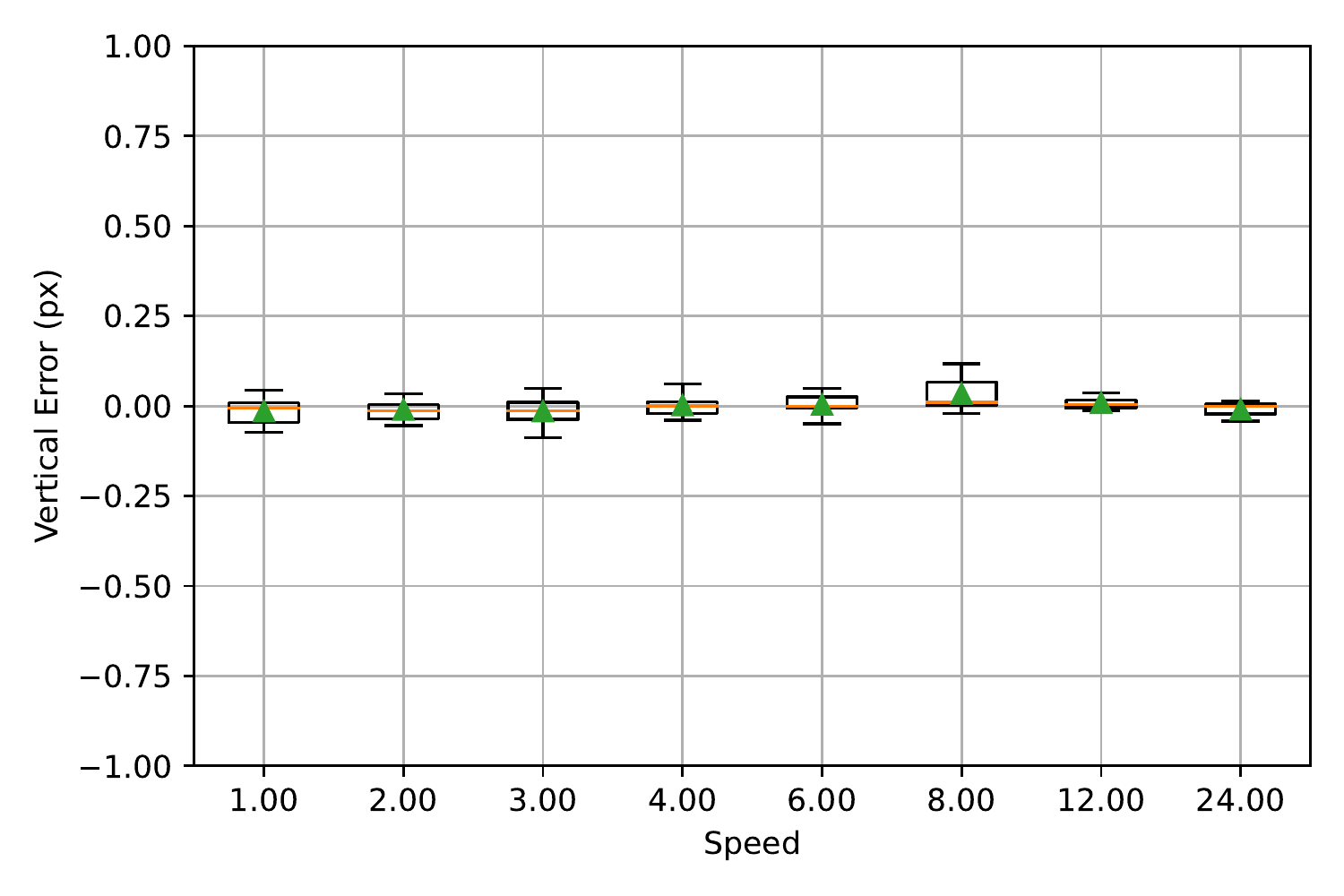}
    }
    \caption{\textbf{DTU Point Features Dataset: When using the Correspondence Tracker with diffuse lighting, mean error $\mu(t)$ is not affected by speed.}  The left column contains plots of the horizontal coordinate (top row) and vertical coordinate (bottom row) of $\mu(t)$ at each timestep and multiple speeds. Each dot corresponds to a processed frame; lines for higher speeds contain data from fewer frames and therefore show fewer dots. The right column plots the ordinate value of each line in the left figures as a box plot: means are shown as green triangles and medians are shown as orange lines. Both the line and box plots only contain timesteps that contain at least 100 tracked features (see Fig. \ref{fig:dtu_active_features}), leading to some asymmetry of the lines about the Key Frame, i.e. the line plot of the horizontal coordinate of $\mu(t)$ at speed=8.00 starts at timestep 0, but ends at timestep 40 even though the Key Frame is at timestep 24. As speed is increased, there is no change in both the horizontal and vertical coordinates, as all lines in the left column plots are on top of one another. The boxes in the box plots of the horizontal coordinate are taller for higher speeds because the time cutoff for those speeds is longer than for the lower speeds, allowing more error to appear in the tracked features.}
    \label{fig:dtu_match_diffuse_mean_error_varyspeed}
\end{figure}

\begin{figure}[H]
    \centering
    \subfigure[$\kappa(t)$, Horizontal Coordinate]{
        \includegraphics[width=0.48\textwidth]{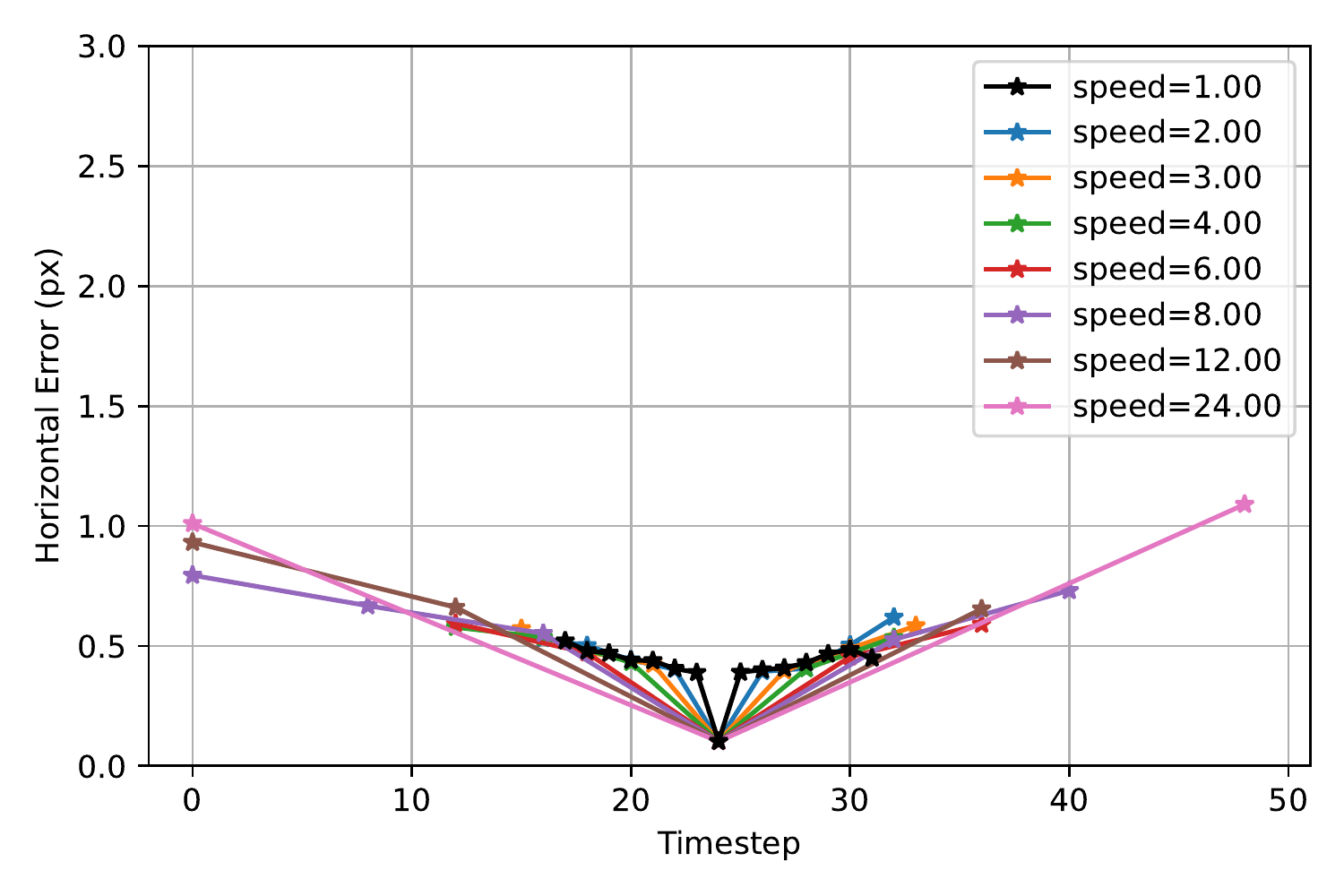}
        \includegraphics[width=0.48\textwidth]{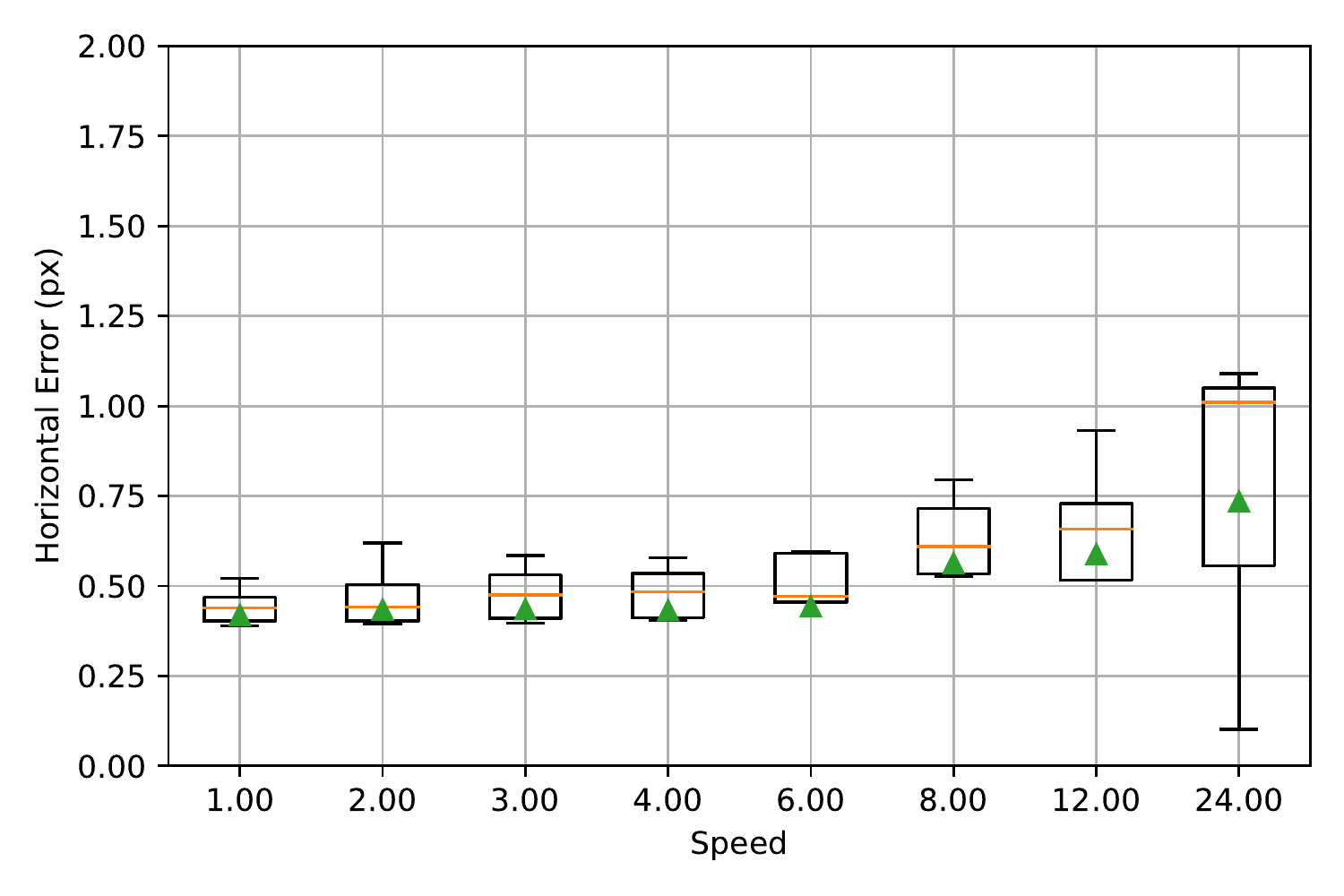}
    }
    \subfigure[$\kappa(t)$, Vertical Coordinate]{
        \includegraphics[width=0.48\textwidth]{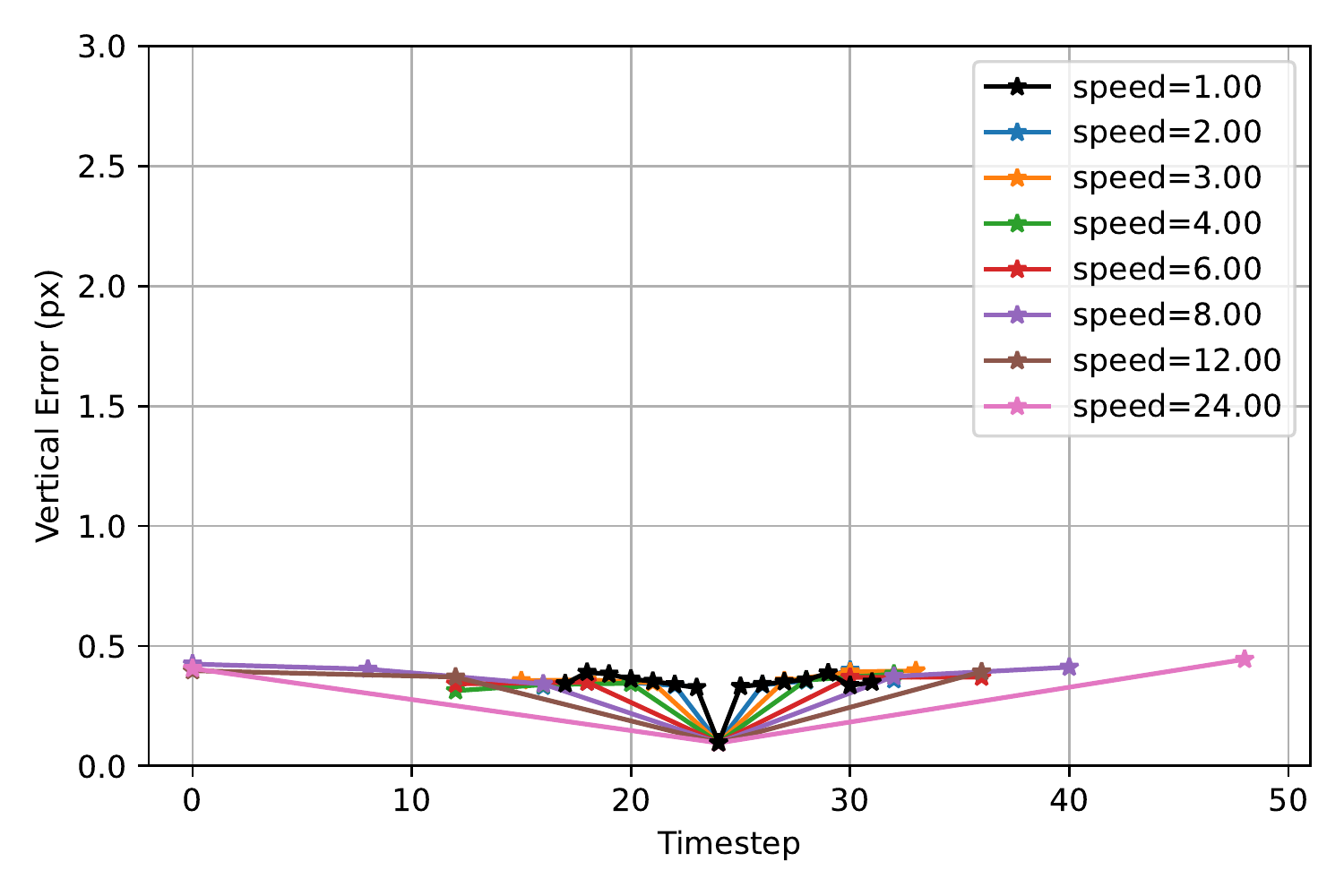}
        \includegraphics[width=0.48\textwidth]{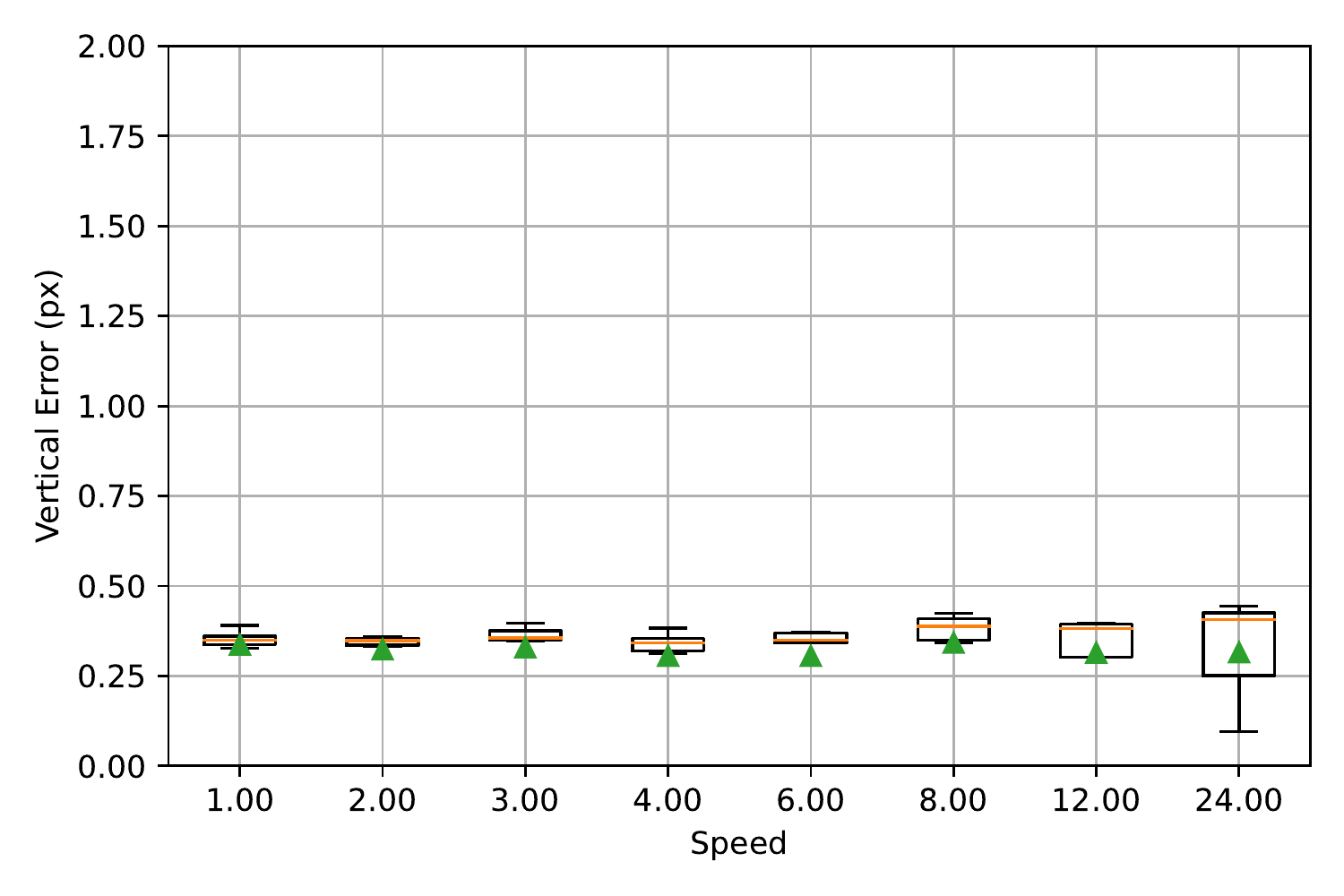}
    }
    \caption{\textbf{DTU Point Features Dataset: When using the Correspondence Tracker with diffuse lighting, mean absolute error $\kappa(t)$ increases in the horizontal direction, but not the vertical direction, as speed is increased.}
    The left column contains plots of the horizontal coordinate (top row) and vertical coordinate (bottom row) of $\kappa(t)$ at each timestep and multiple speeds. Each dot corresponds to a processed frame; lines for higher speeds contain data from fewer frames and therefore show fewer dots. The right column plots the ordinate value of each line in the left figures as a box plot: means are shown as green triangles and medians are shown as orange lines. Both the line and box plots only contain timesteps that contain at least 100 tracked features (see Fig. \ref{fig:dtu_active_features}). The mean and median values of the horizontal coordinate of $\kappa(t)$ increases as speed is increased.
    }
    \label{fig:dtu_match_diffuse_MAE_varyspeed}
\end{figure}

\begin{figure}[H]
    \centering
    \subfigure[$\Sigma(t)$, Horizontal Covariance]{
      \includegraphics[width=0.48\textwidth]{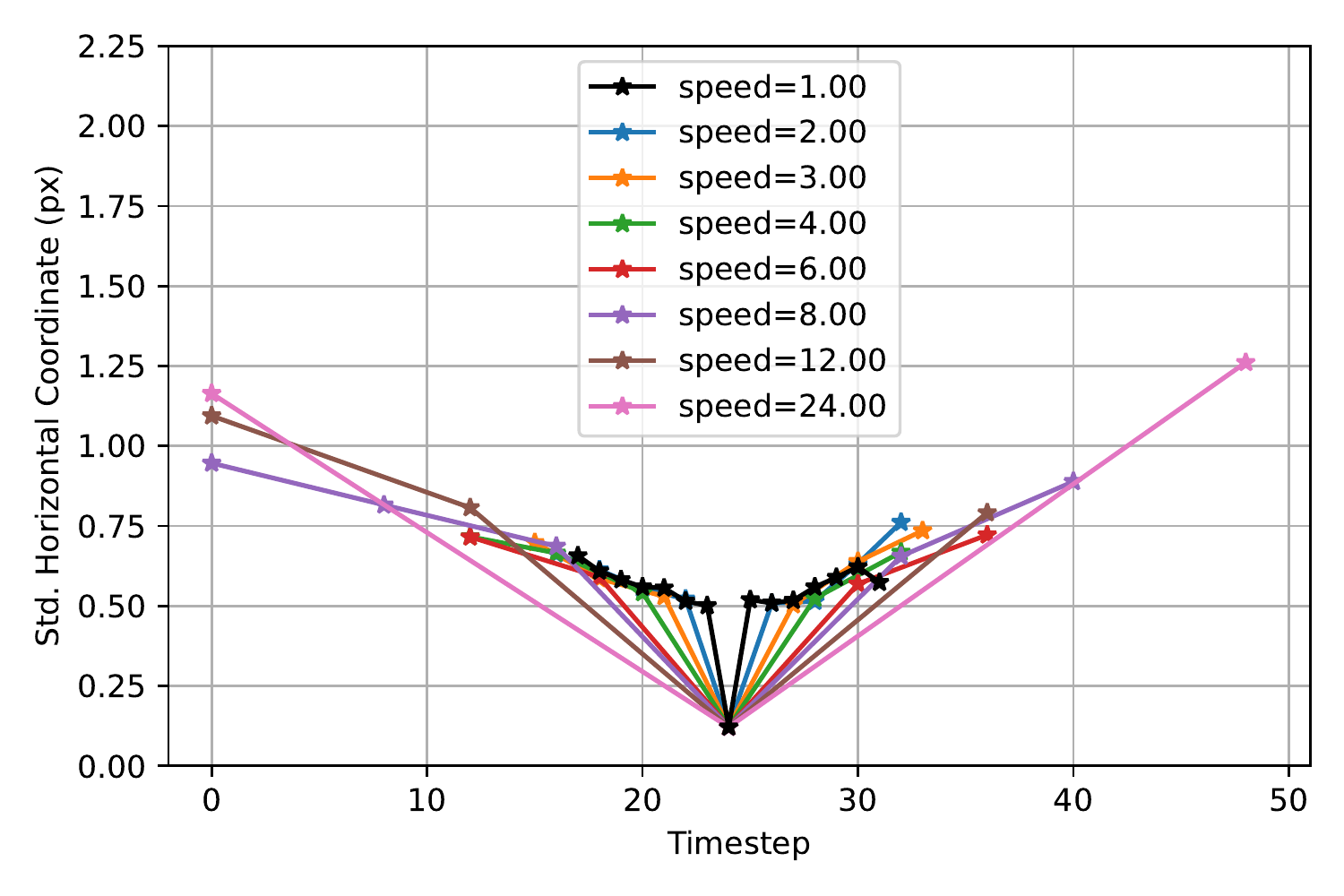}
      \includegraphics[width=0.48\textwidth]{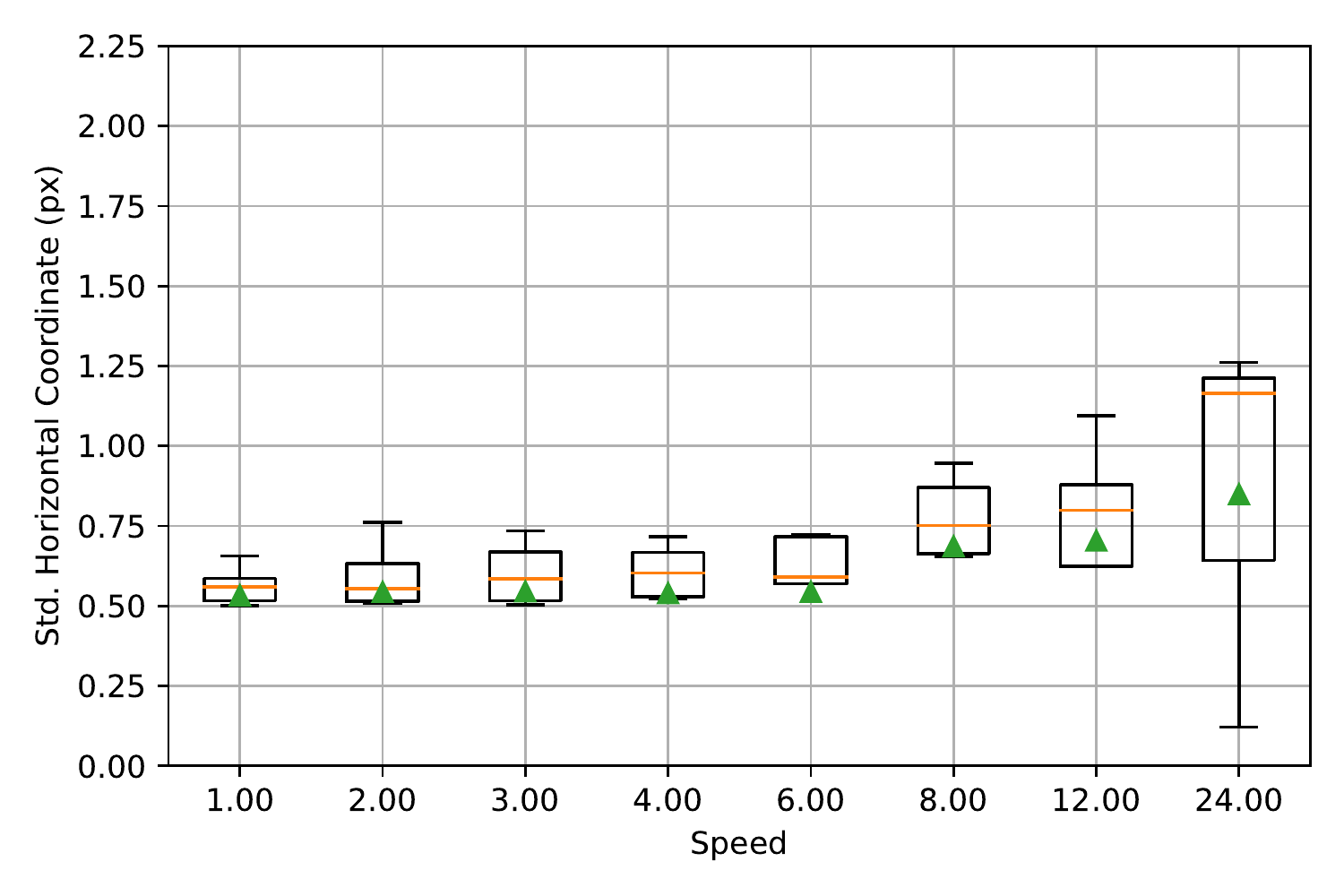}
    }
    \subfigure[$\Sigma(t)$, Vertical Coordinate]{
      \includegraphics[width=0.48\textwidth]{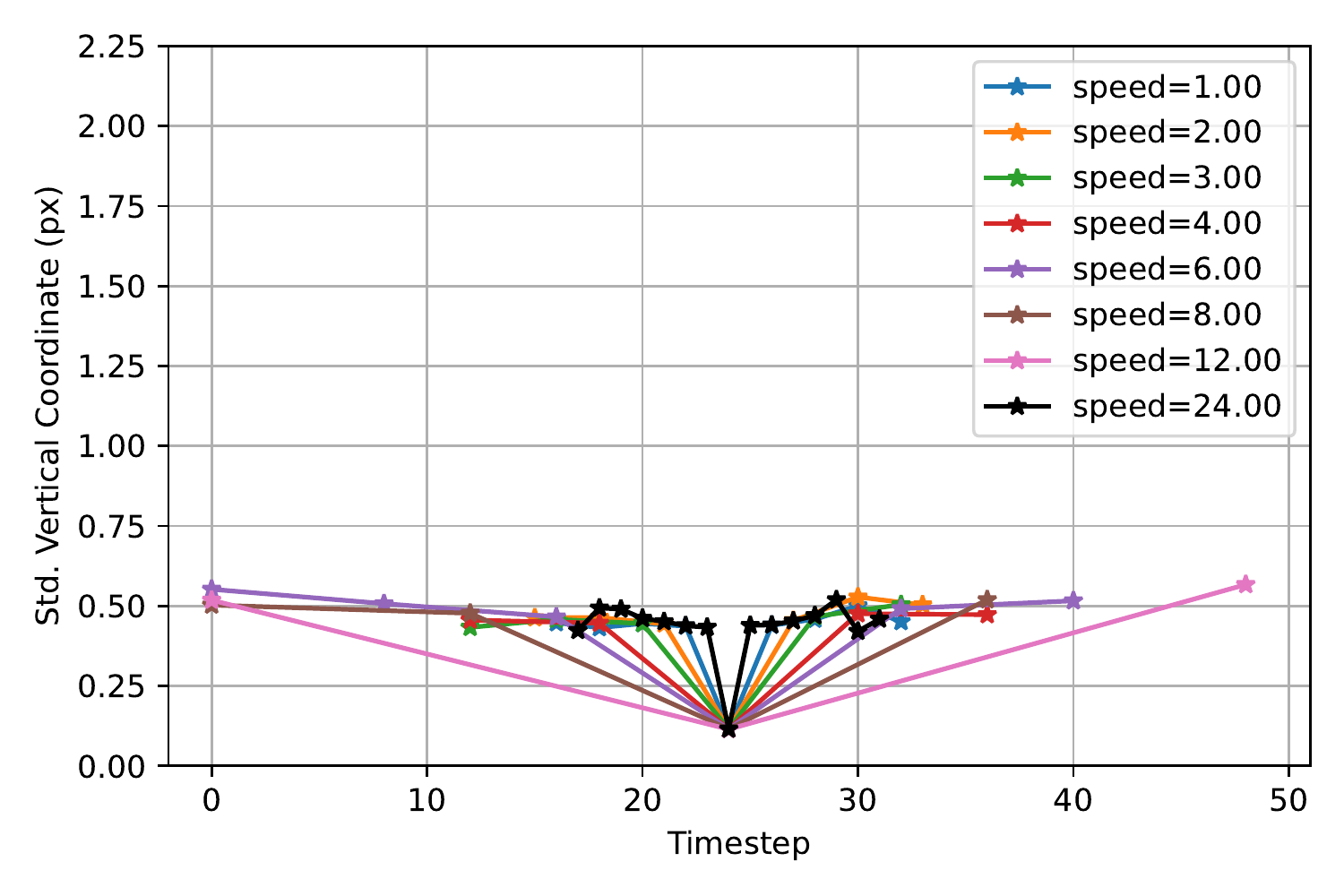}
      \includegraphics[width=0.48\textwidth]{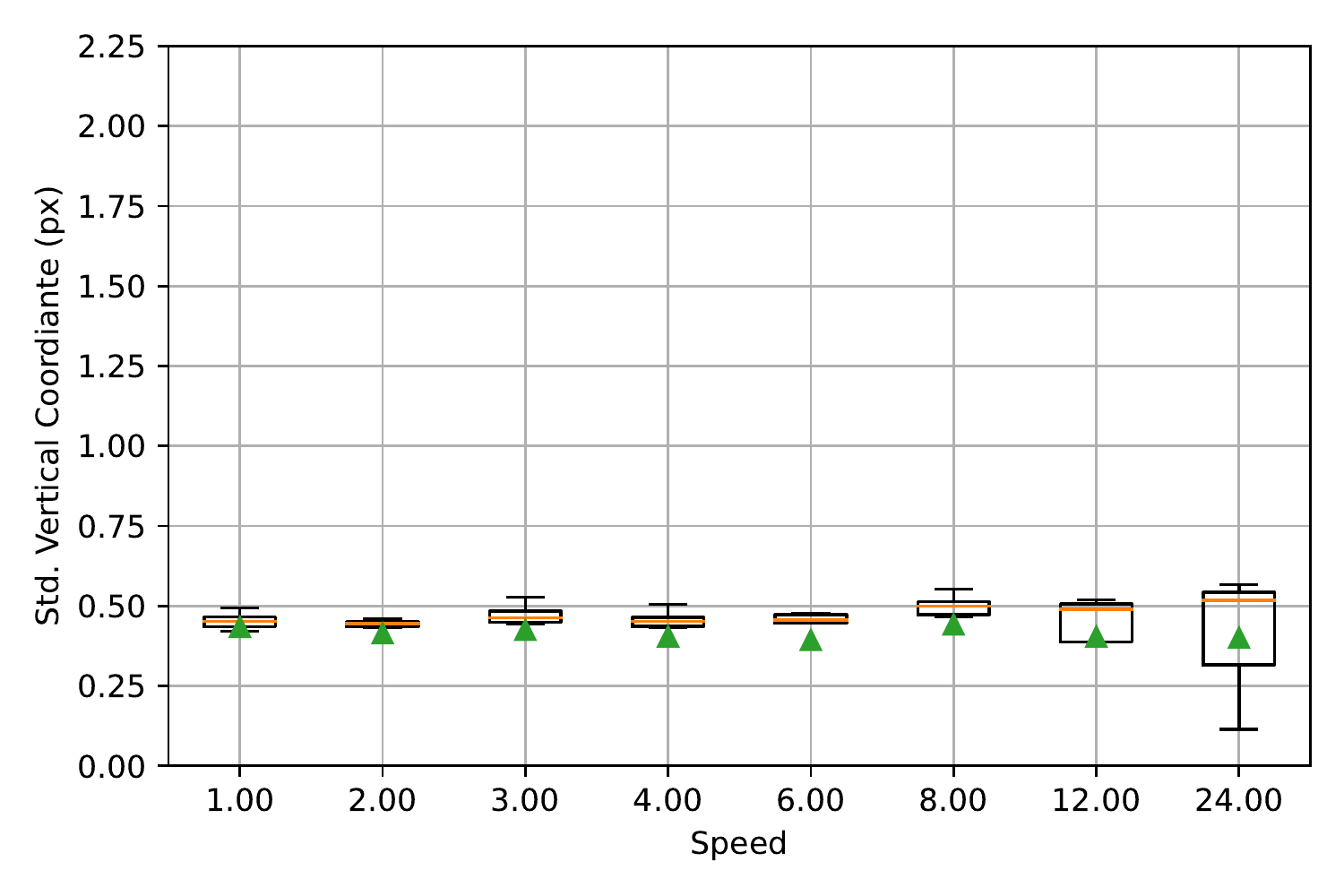}
    }
    \caption{\textbf{DTU Point Features Dataset: When using the Correspondence Tracker with diffuse lighting, covariance $\Sigma(t)$ increases in the horizontal direction, but not the vertical direction, as speed is increased.}  
    The left column contains plots of the square root of the horizontal coordinate (top row) and vertical coordinate (bottom row) of $\Sigma(t)$ at each timestep and multiple speeds. Each dot corresponds to a processed frame; lines for higher speeds contain data from fewer frames and therefore show fewer dots. The right column plots the ordinate value of each line in the left figures as a box plot: means are shown as green triangles and medians are shown as orange lines. Both the line and box plots only contain timesteps that contain at least 100 tracked features (see Fig. \ref{fig:dtu_active_features}). The mean and median values of the horizontal value of $\Sigma(t)$ as speed is increased.
    }
    \label{fig:dtu_match_diffuse_cov_varyspeed}
\end{figure}

\begin{figure}[H]
    \centering
    \subfigure[$\mu(t)$, Horizontal Coordinate]{
        \includegraphics[width=0.48\textwidth]{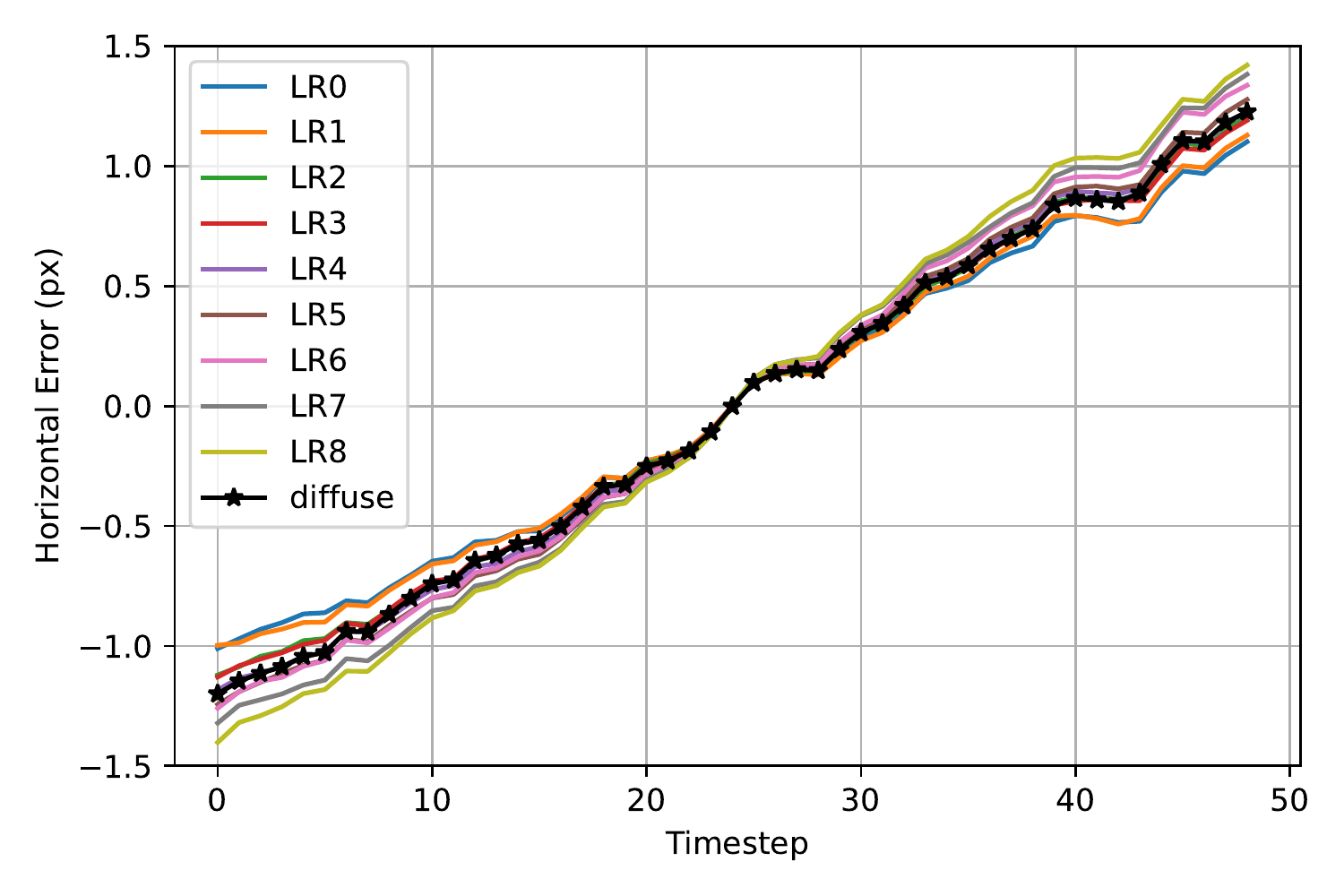}
        \includegraphics[width=0.48\textwidth]{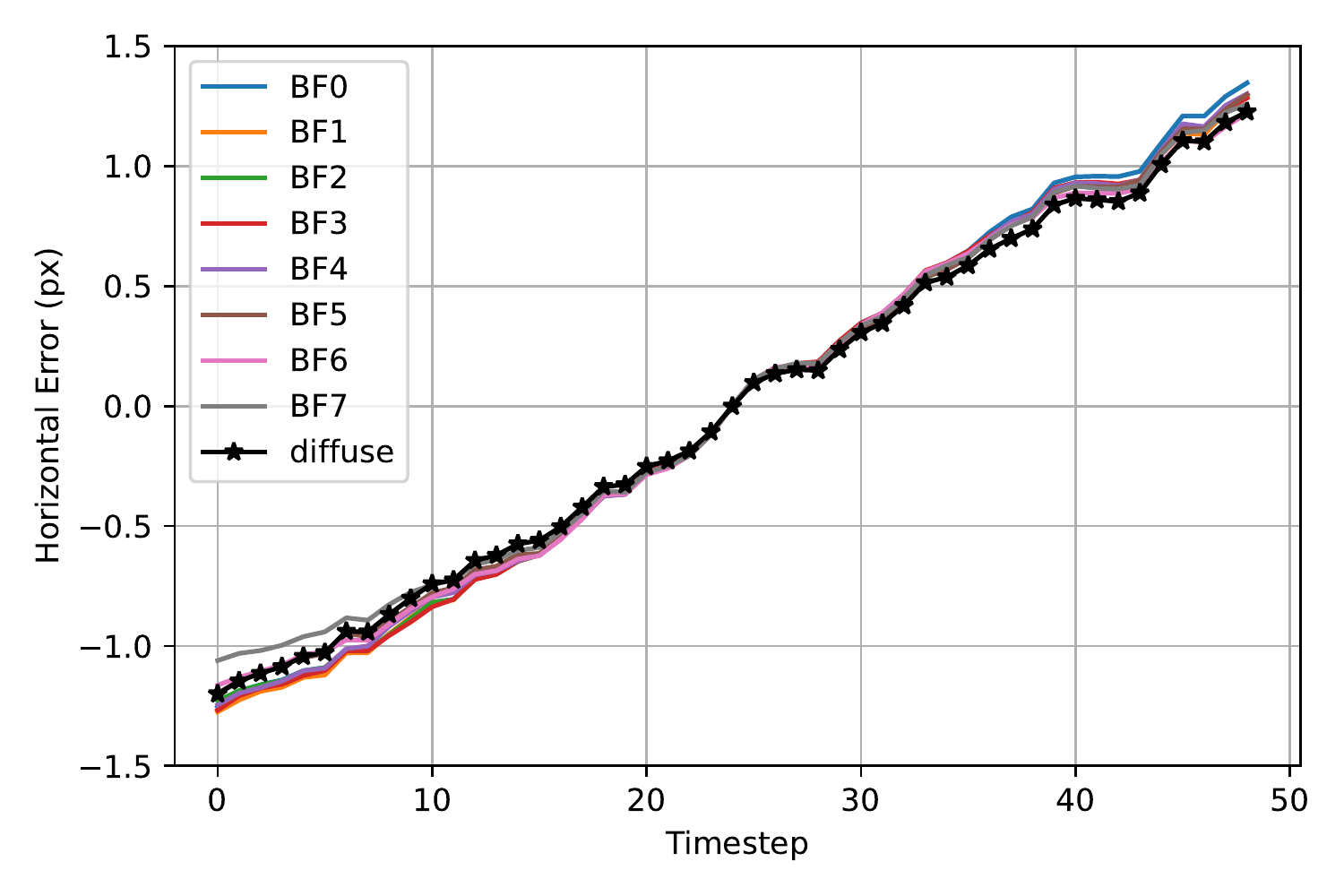}
    }
    \subfigure[$\mu(t)$, Vertical Coordinate]{
        \includegraphics[width=0.48\textwidth]{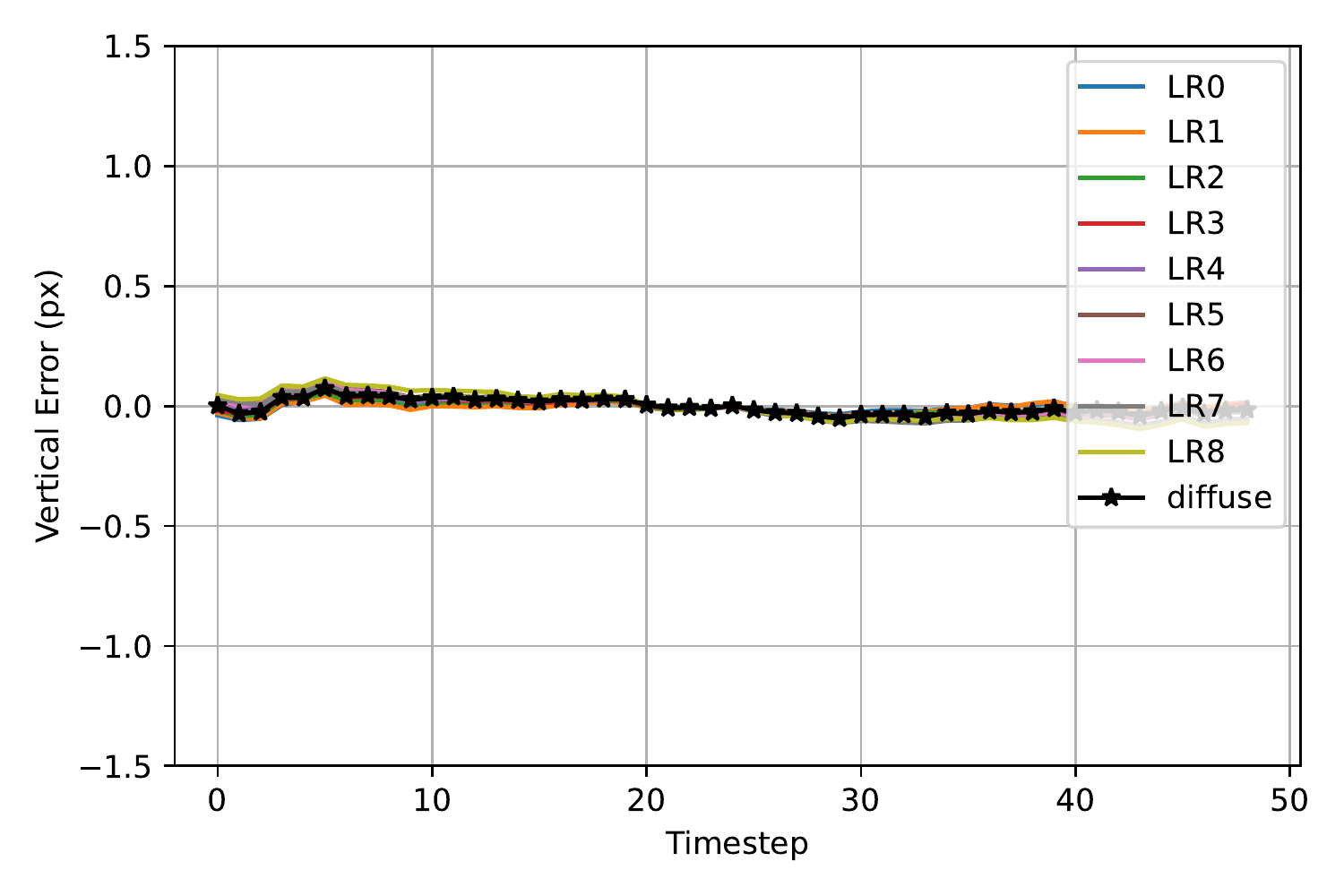}
        \includegraphics[width=0.48\textwidth]{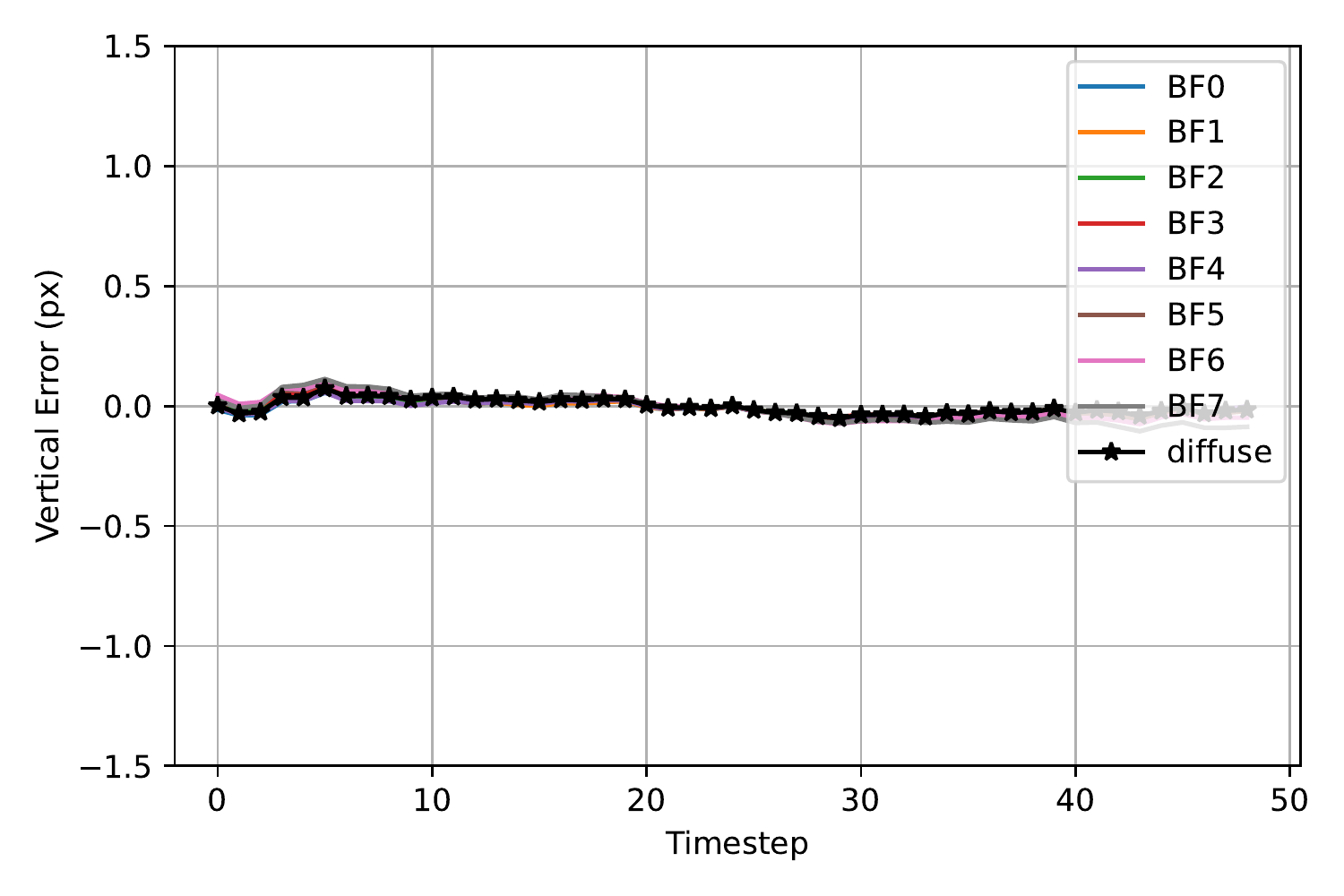}
    }
    \caption{\textbf{DTU Point Features Dataset: The existence of directional lighting does not change trends in mean error $\mu(t)$ when using the Lucas-Kande Tracker at nominal speed.} We compute $\mu(t)$ using diffuse lighting (black lines) and each of the directional lighting conditions listed in Figure \ref{fig:dtu_light_stage} using all tracks from all 60 scenes. Results for the horizontal coordinate are in the top row and results for the  vertical coordinate are in the bottom row. Timesteps are limited to those that contain at least 100 features. The variation of $\mu(t)$ due to the existence of directional lighting is at most 10 percent the size of the variation common to all plotted lines. The effect of directional lighting is relatively small because changes between adjacent frames are small whether or not the scene contains directional lighting.}
    \label{fig:dtu_lighting_mu_LK}
\end{figure}

\begin{figure}[H]
    \centering
    \subfigure[$\mu(t)$, Horizontal Coordinate]{
        \includegraphics[width=0.48\textwidth]{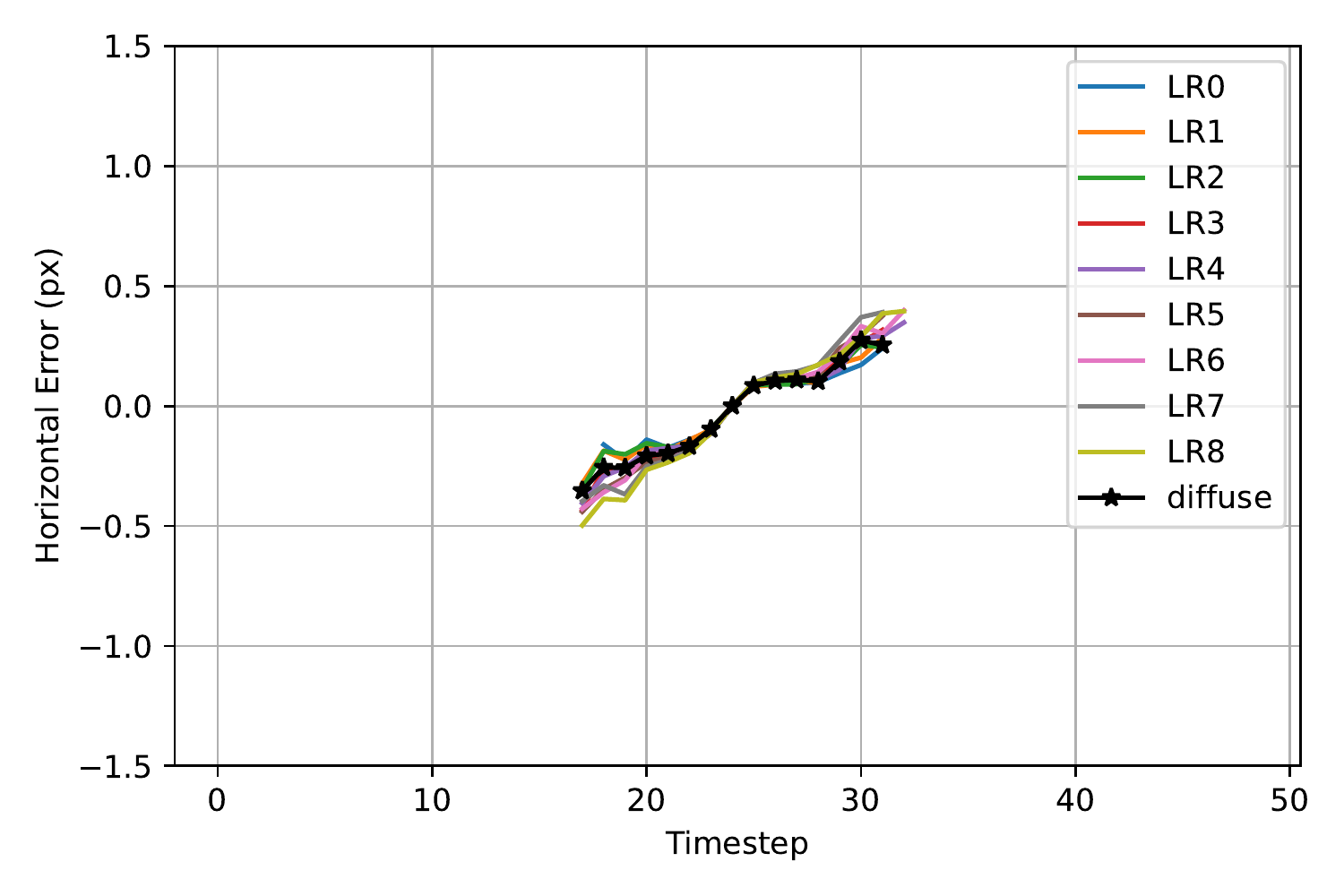}
        \includegraphics[width=0.48\textwidth]{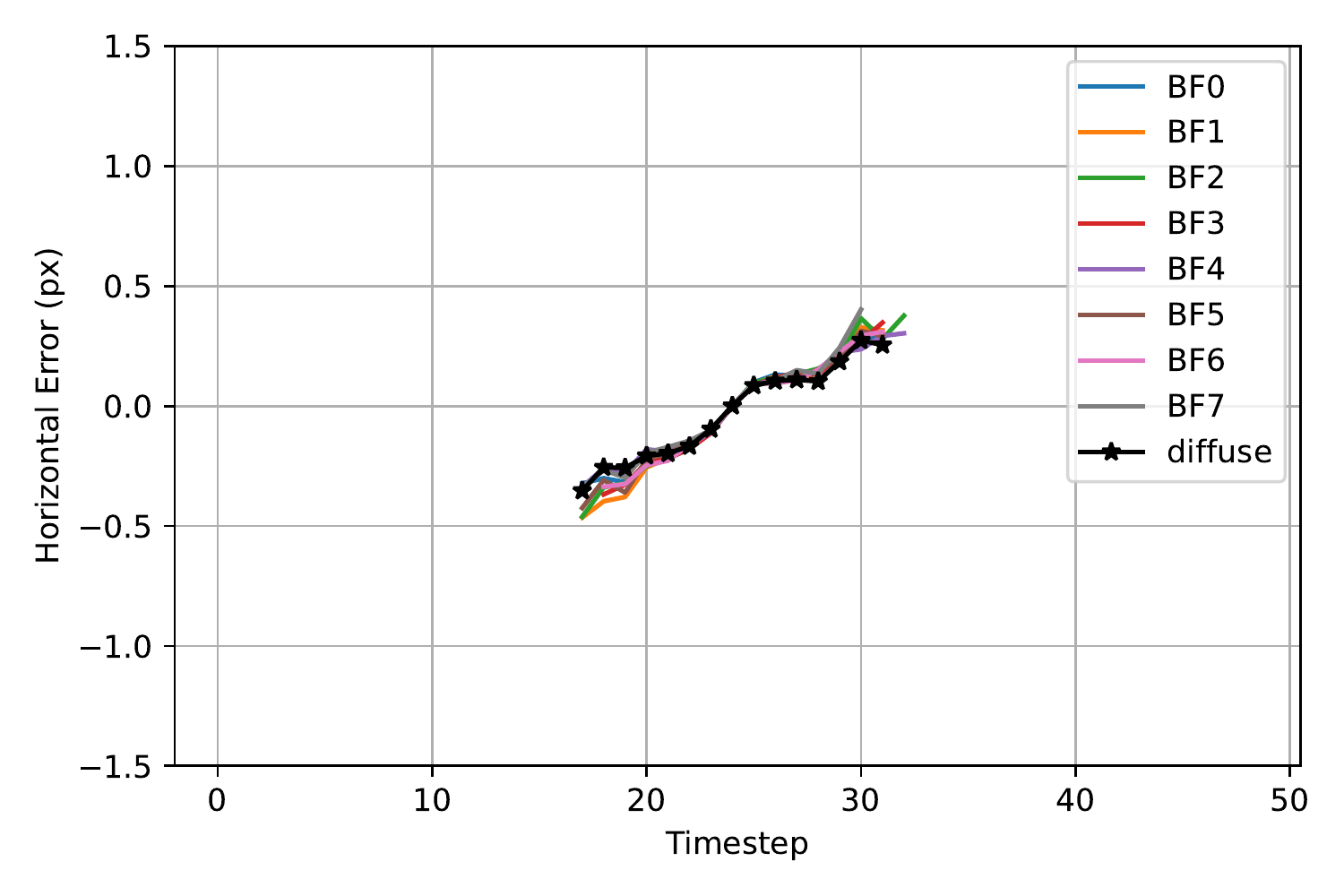}
    }
    \subfigure[$\mu(t)$, Vertical Coordinate]{
        \includegraphics[width=0.48\textwidth]{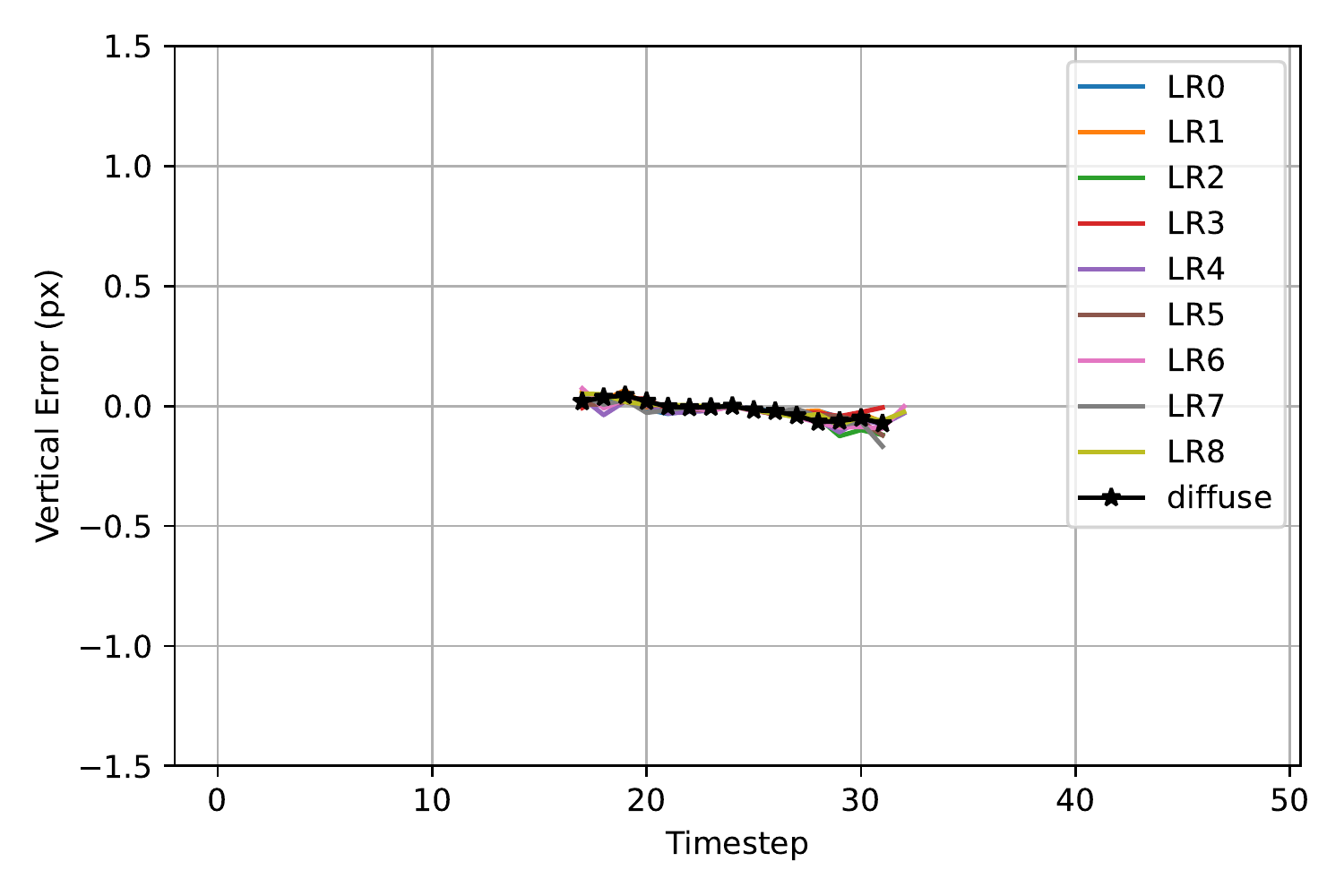}
        \includegraphics[width=0.48\textwidth]{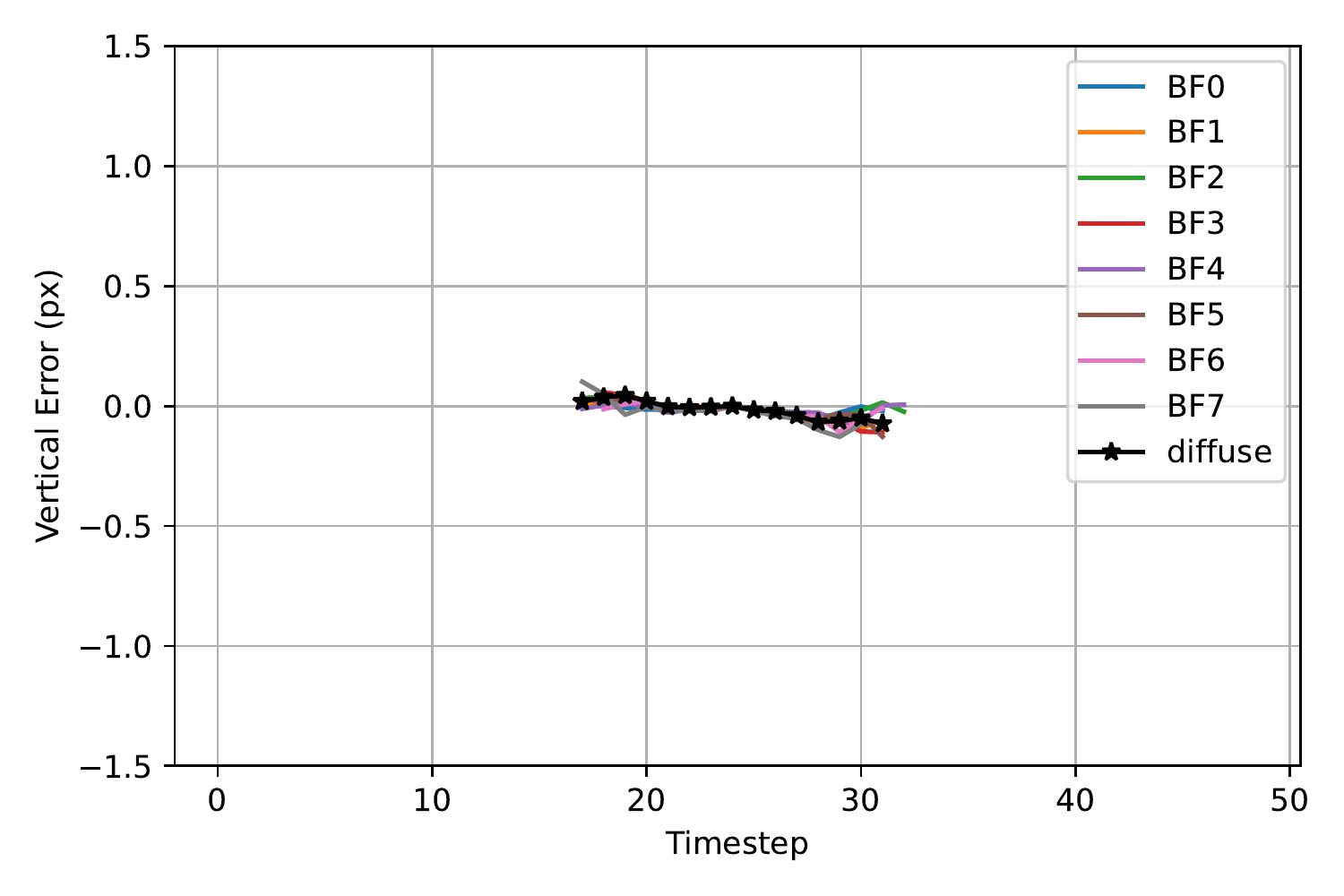}
    }
    \caption{\textbf{DTU Point Features Dataset: The existence of directional lighting does not change trends in mean error $\mu(t)$ when using the Correspondence Tracker at nominal speed.} We compute $\mu(t)$ using diffuse lighting (black lines) and each of the directional lighting conditions listed in Figure \ref{fig:dtu_light_stage} using all tracks from all 60 scenes. Results of the horizontal coordinate are shown in the top row and results for the vertical coordinate are shown in the bottom row. Timesteps are limited to those that contain at least 100 features. The variation of $\mu(t)$ due to the existence of directional lighting is at most 10 percent of the variation common to all plotted lines. The effects of directional lighting is relatively small because changes between adjacent frames are small whether or not the scene contains directional lighting.}
    \label{fig:dtu_lighting_mu_match}
\end{figure}

\begin{figure}[H]
    \centering
    \subfigure[$\kappa(t)$, Horizontal Coordinate]{
        \includegraphics[width=0.48\textwidth]{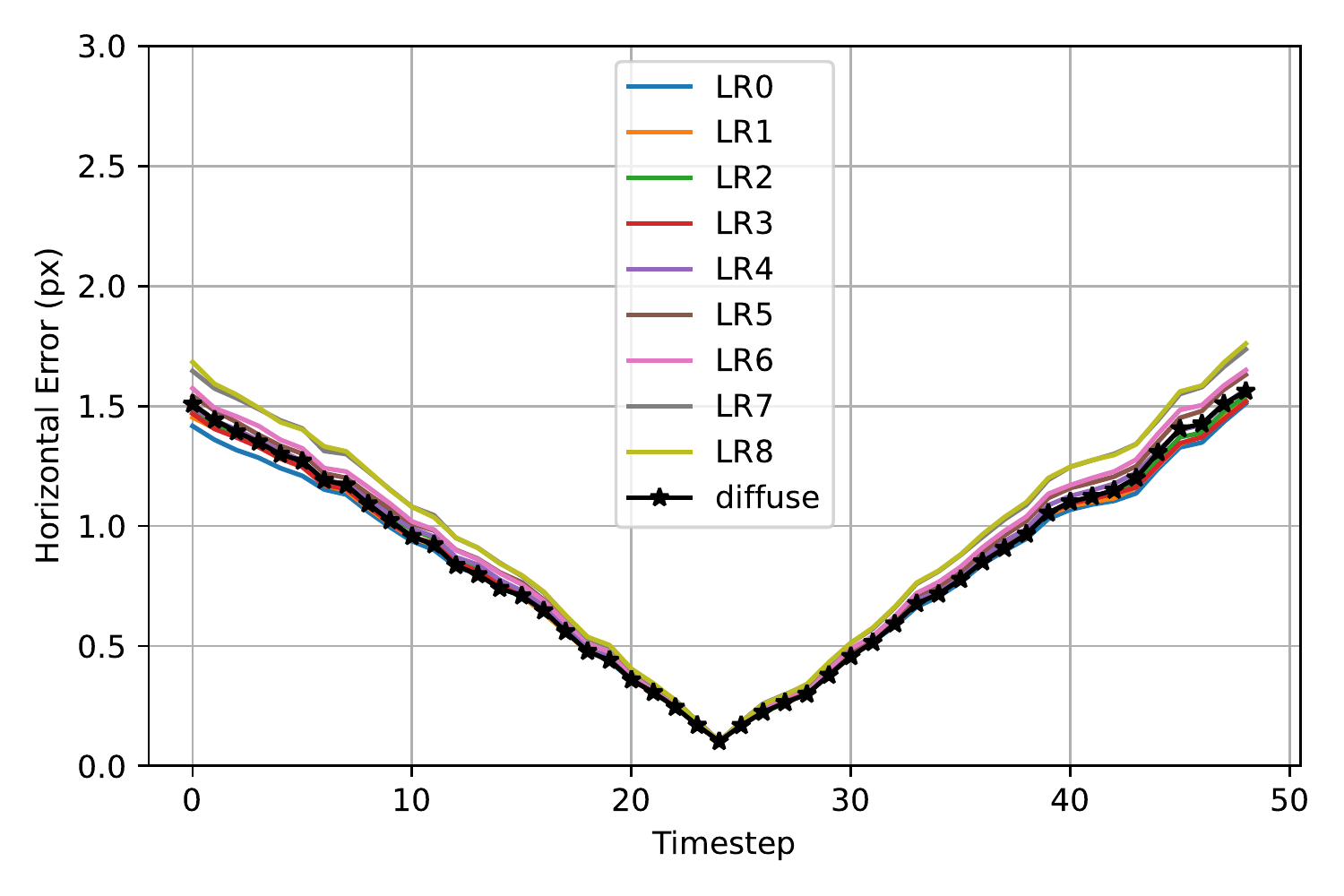}
        \includegraphics[width=0.48\textwidth]{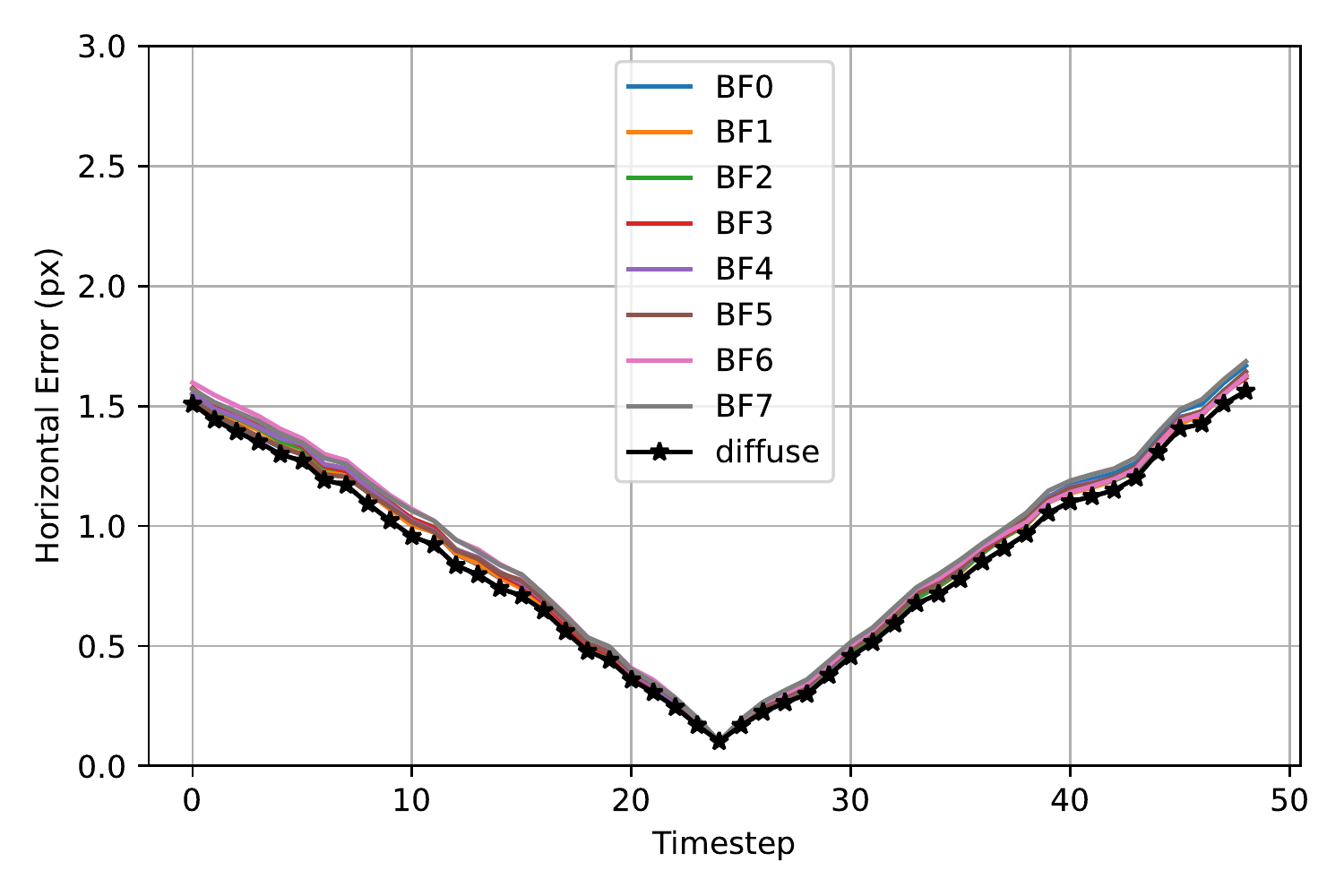}
    }
    \subfigure[$\kappa(t)$, Vertical Coordinate]{
        \includegraphics[width=0.48\textwidth]{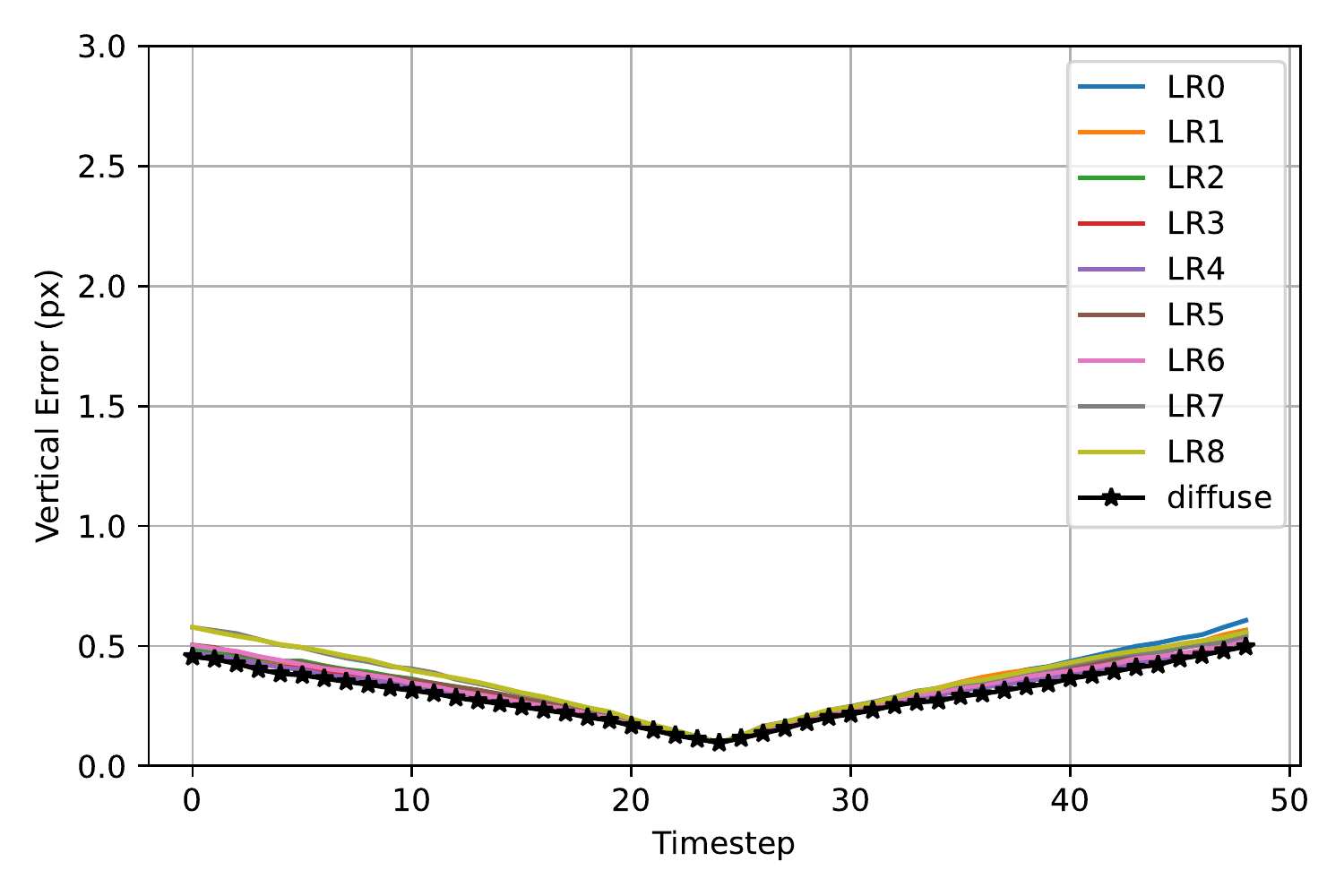}
        \includegraphics[width=0.48\textwidth]{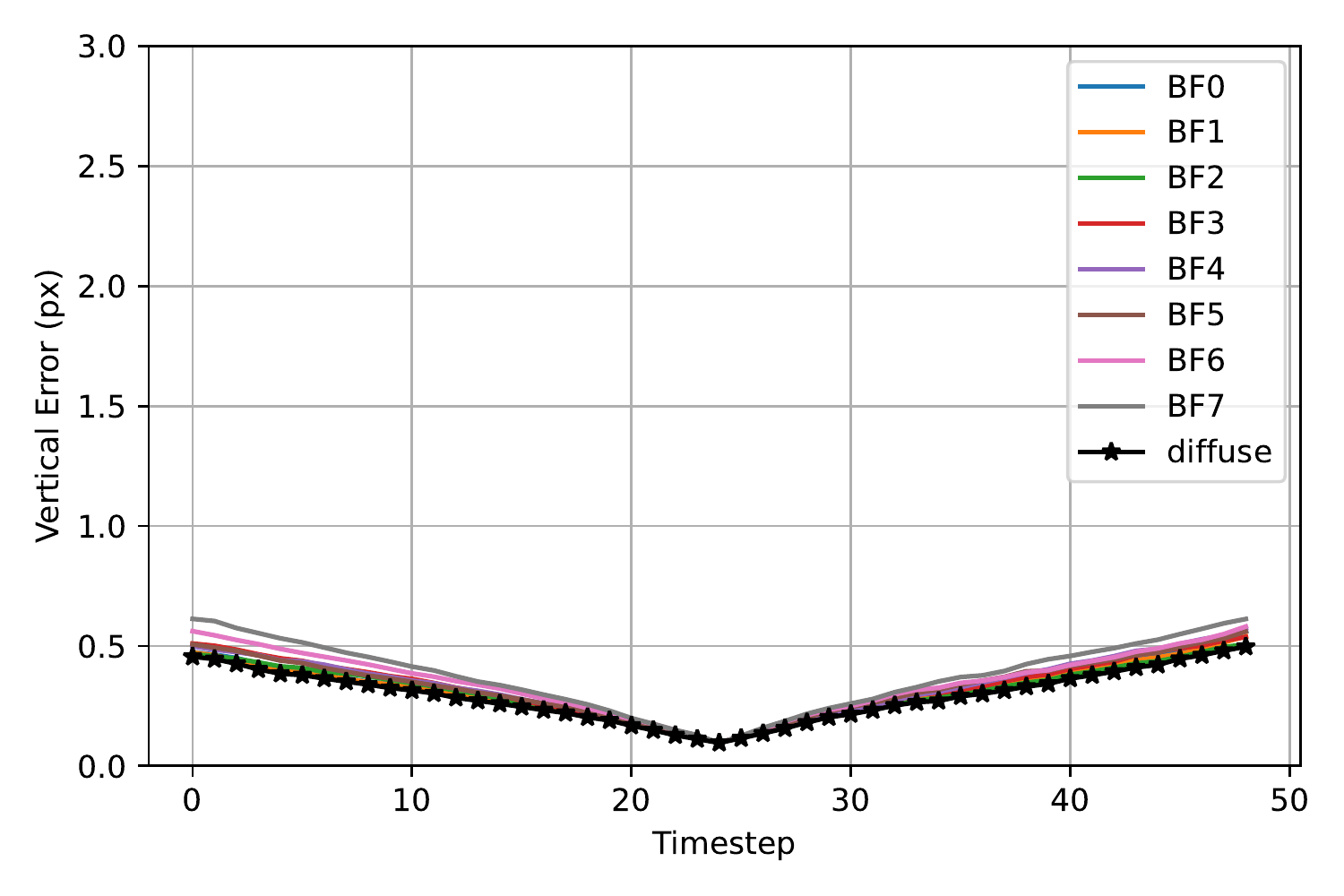}
    }
    \caption{\textbf{DTU Point Features Dataset: The existence of directional lighting does not change trends in mean absolute error $\kappa(t)$ when using the Lucas-Kanade Tracker at nominal speed.} We compute $\mu(t)$ at each timestep using diffuse lighting (black lines) and each of the directional lighting conditions listed in Figure \ref{fig:dtu_light_stage} using all tracks from all 60 scenes. Timesteps are limited to those that contain at least 100 features. The variation of $\kappa(t)$ due to the existence of directional lighting is at most 10 percent of the variation common to all plotted lines. The effect of directional lighting is relatively small because changes between adjacent frames are small whether or not the scene contains directional lighting.}
    \label{fig:dtu_lighting_omega_LK}
\end{figure}

\begin{figure}[H]
    \centering
    \subfigure[$\kappa(t)$, Horizontal Coordinate]{
        \includegraphics[width=0.48\textwidth]{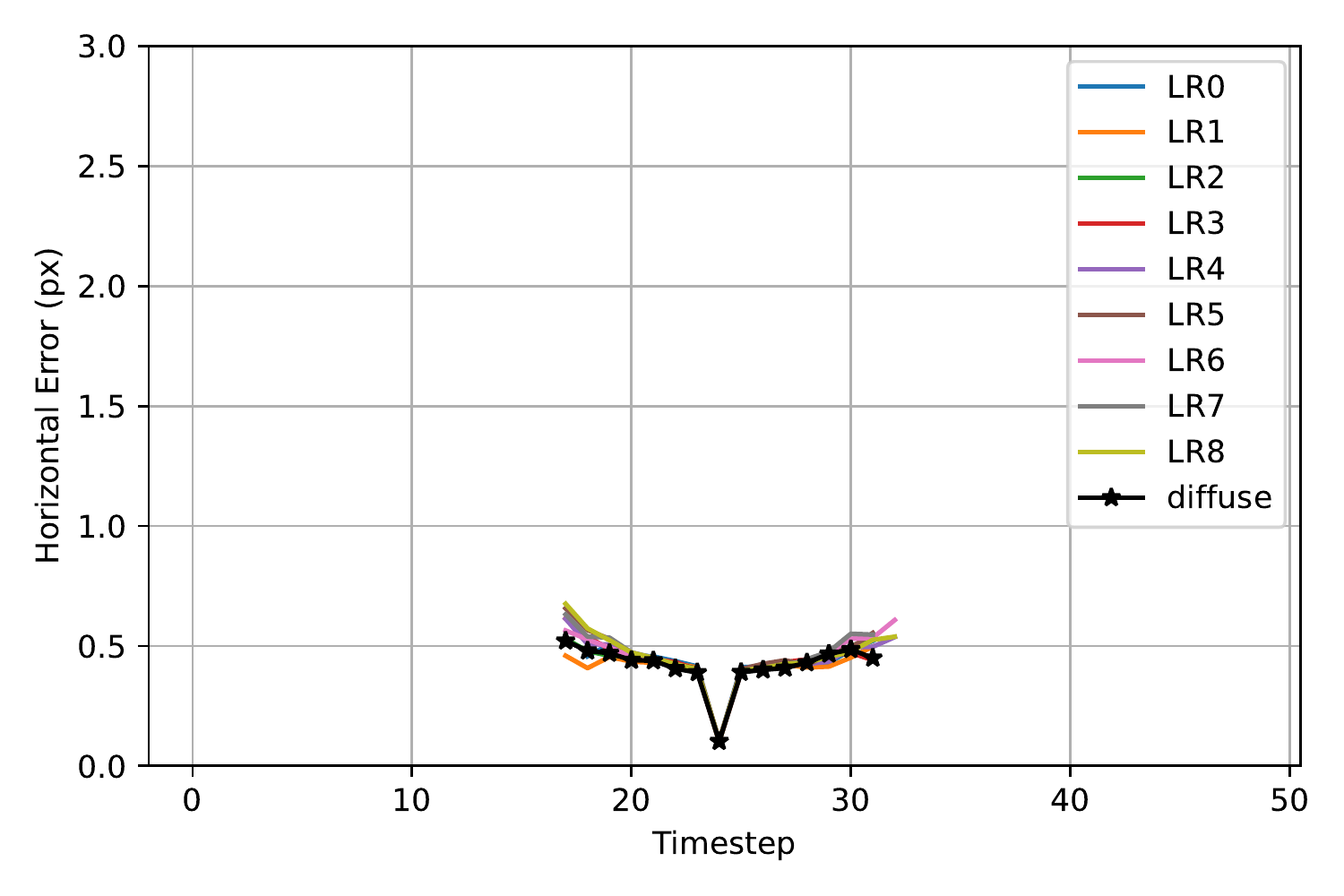}
        \includegraphics[width=0.48\textwidth]{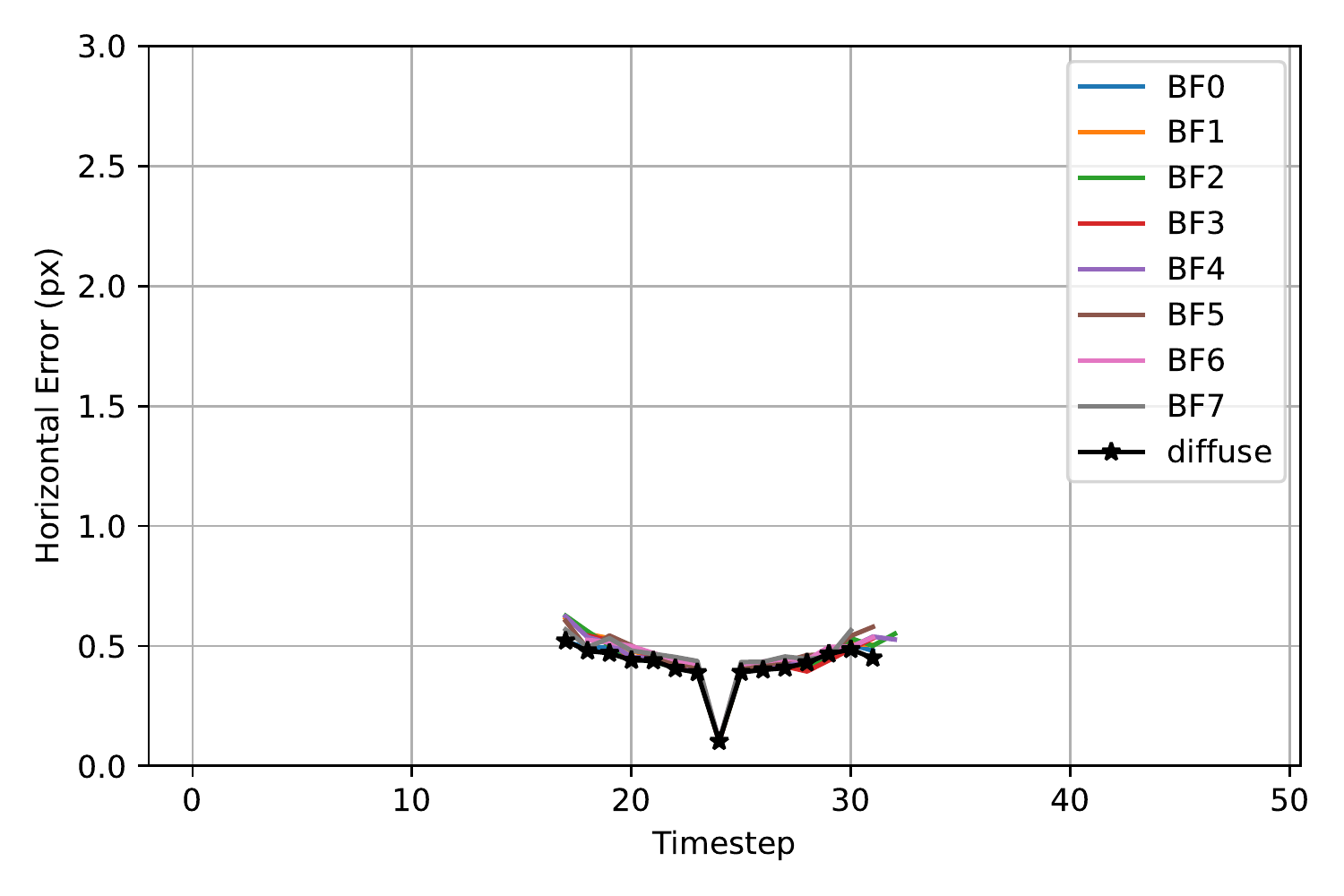} 
    }
    \subfigure[$\kappa(t)$, Vertical Coordinate]{
        \includegraphics[width=0.48\textwidth]{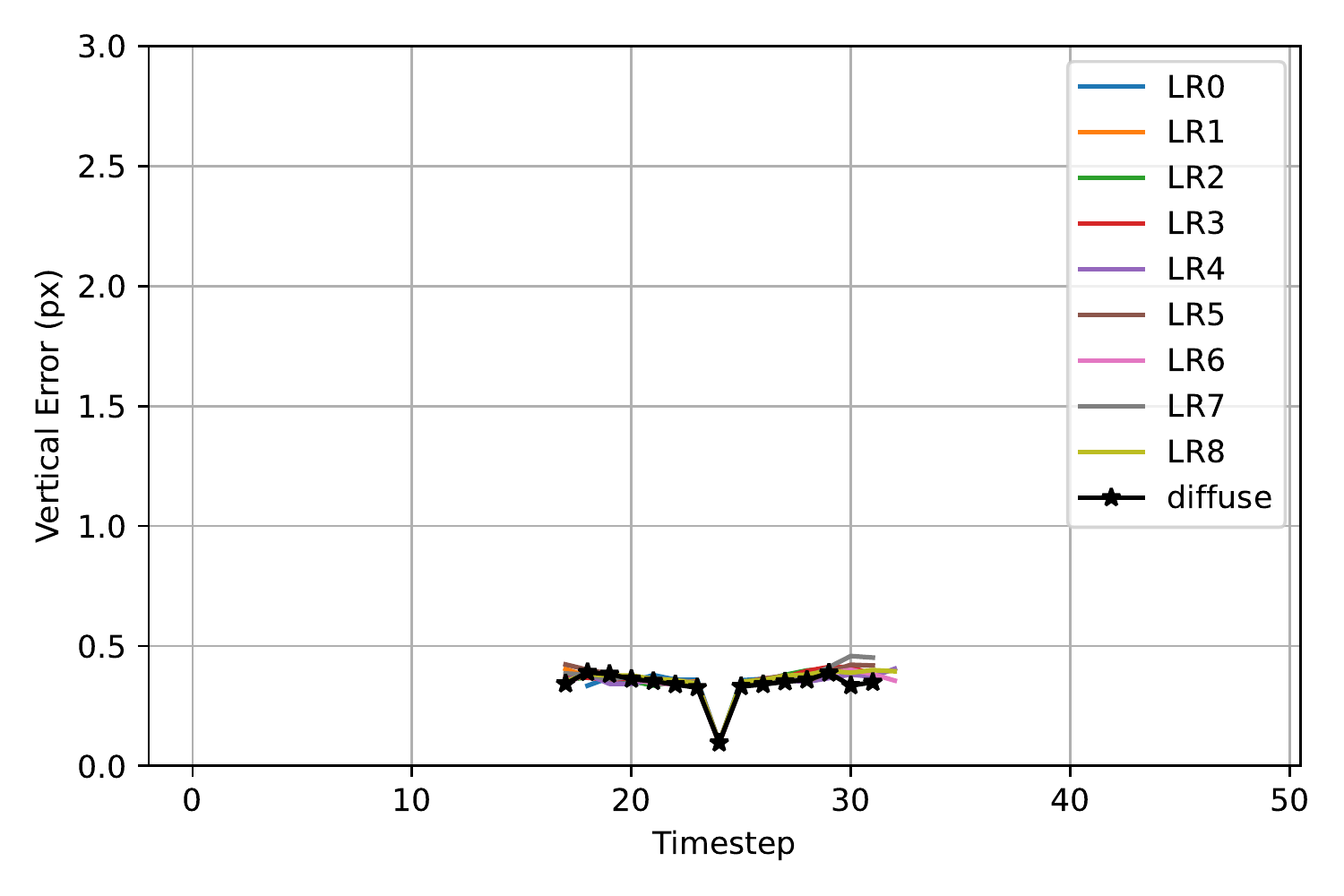} 
        \includegraphics[width=0.48\textwidth]{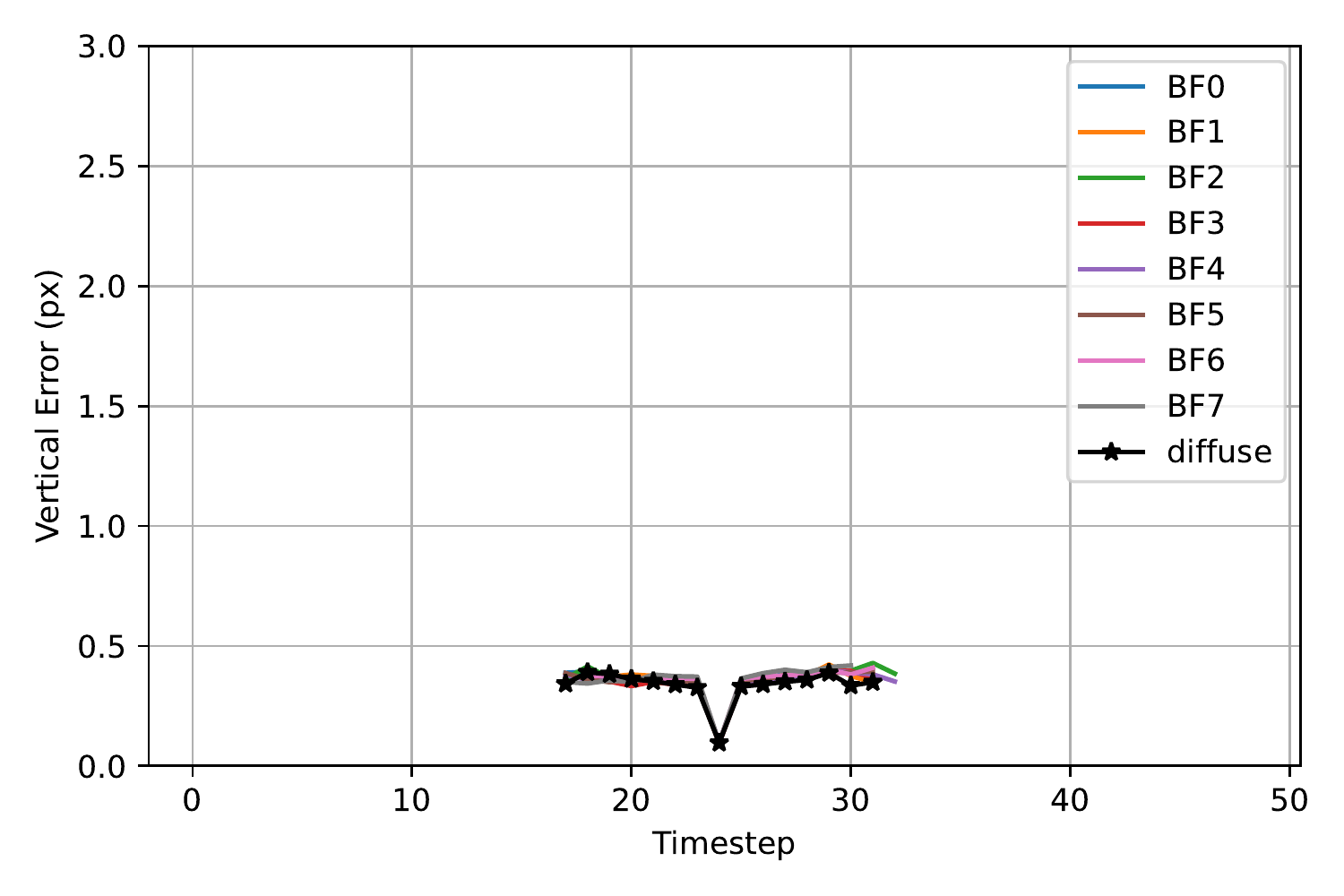} 
    }
    \caption{\textbf{DTU Point Features Dataset: The existence of directional lighting does not change trends in mean absolute error $\kappa(t)$ when using the Correspondence Tracker at nominal speed.} We compute $\kappa(t)$ using diffuse lighting (black lines) and each of the directional lighting conditions listed in Figure \ref{fig:dtu_light_stage} using all tracks from all 60 scenes. Timesteps are limited to those that contain at least 100 features. The variation of $\kappa(t)$ due to the existence of directional lighting is at most 10 percent of the variation common to all plotted lines. The effect of directional lighting is relatively small because changes between adjacent frames are small whether or not the scene contains directional lighting.}
    \label{fig:dtu_lighting_omega_match}
\end{figure}

\begin{figure}[H]
    \centering
    \subfigure[$\Sigma(t)$, Horizontal Coordinate]{
        \includegraphics[width=0.48\textwidth]{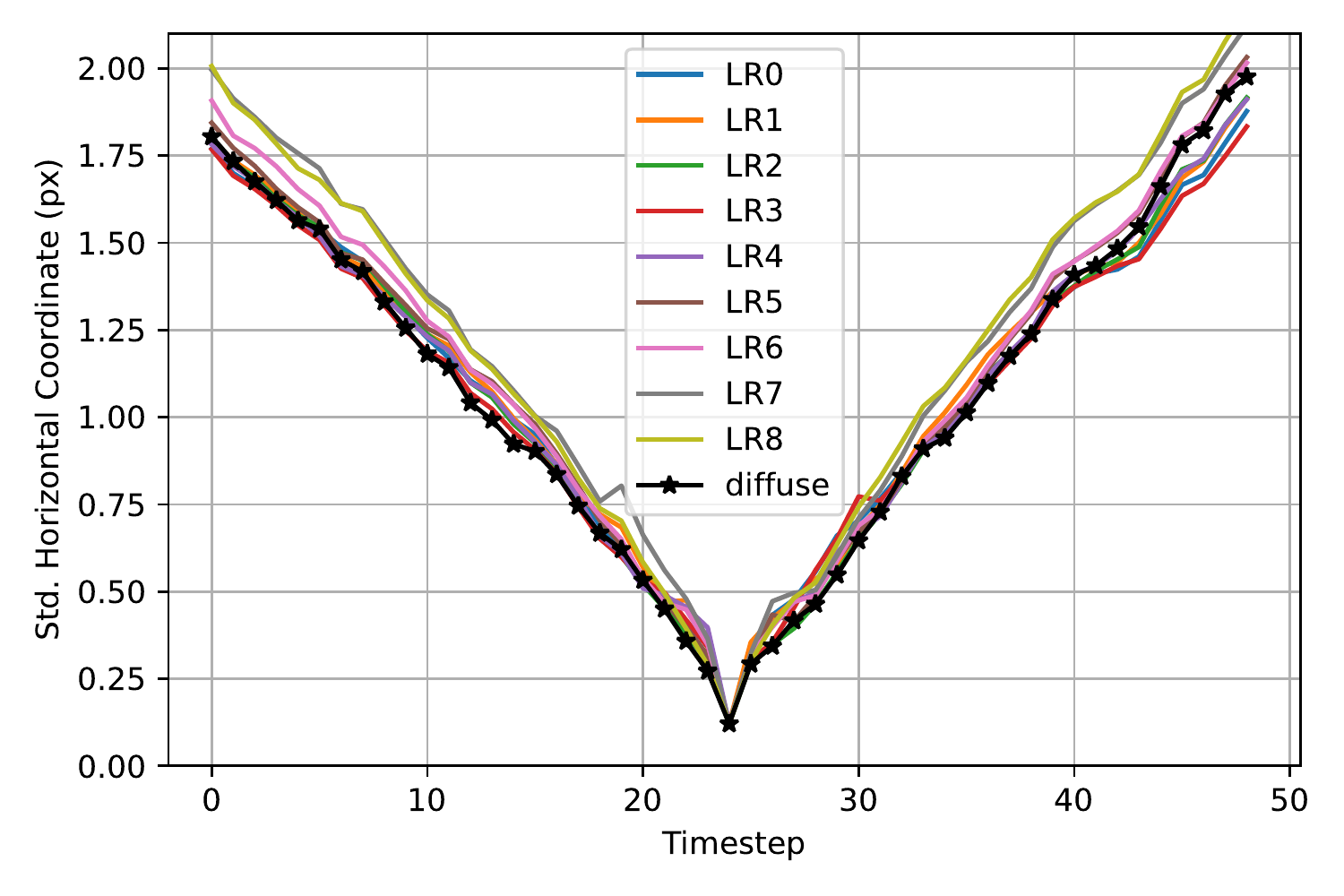}
        \includegraphics[width=0.48\textwidth]{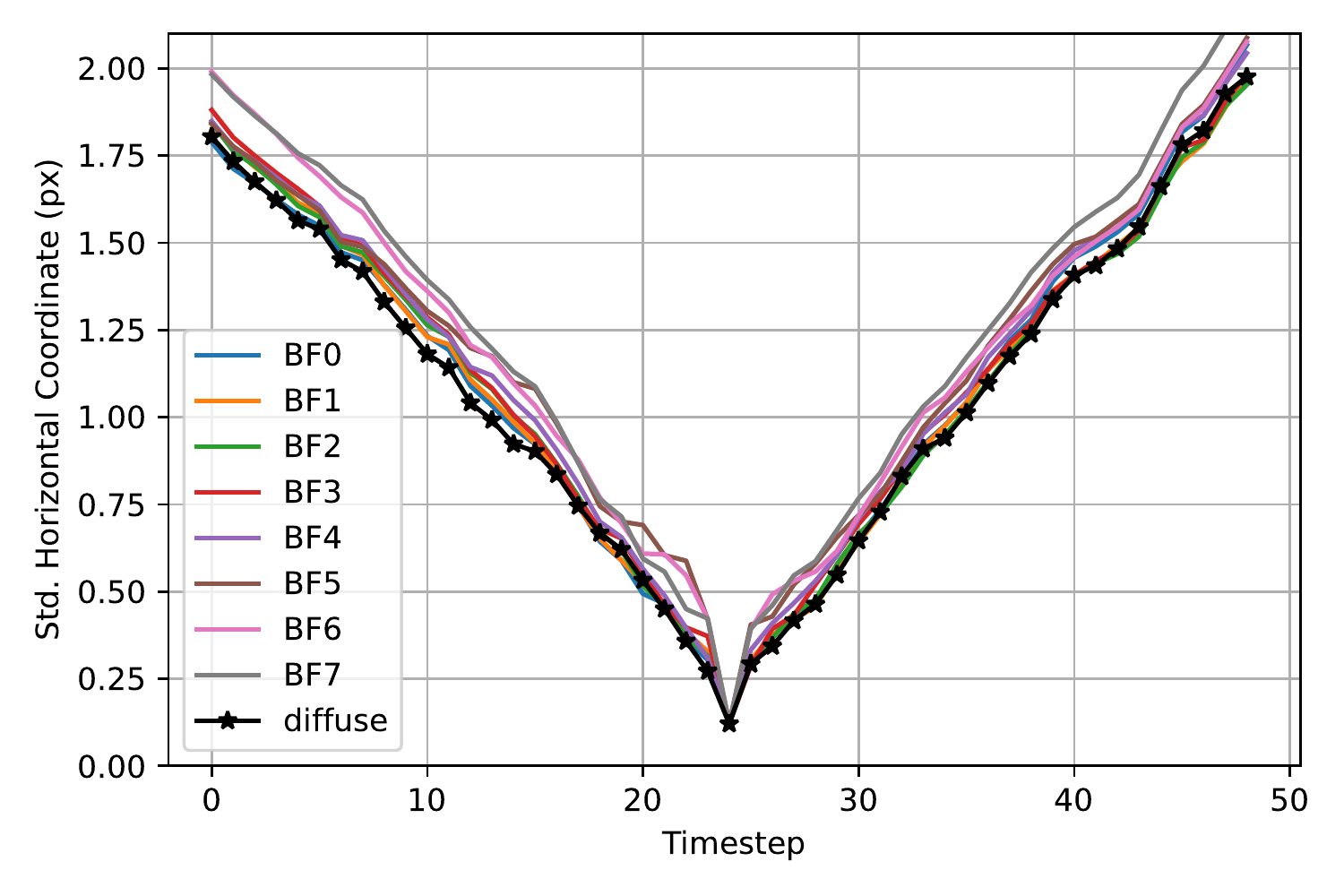}
    }
    \subfigure[$\Sigma(t)$, Vertical Coordinate]{
        \includegraphics[width=0.48\textwidth]{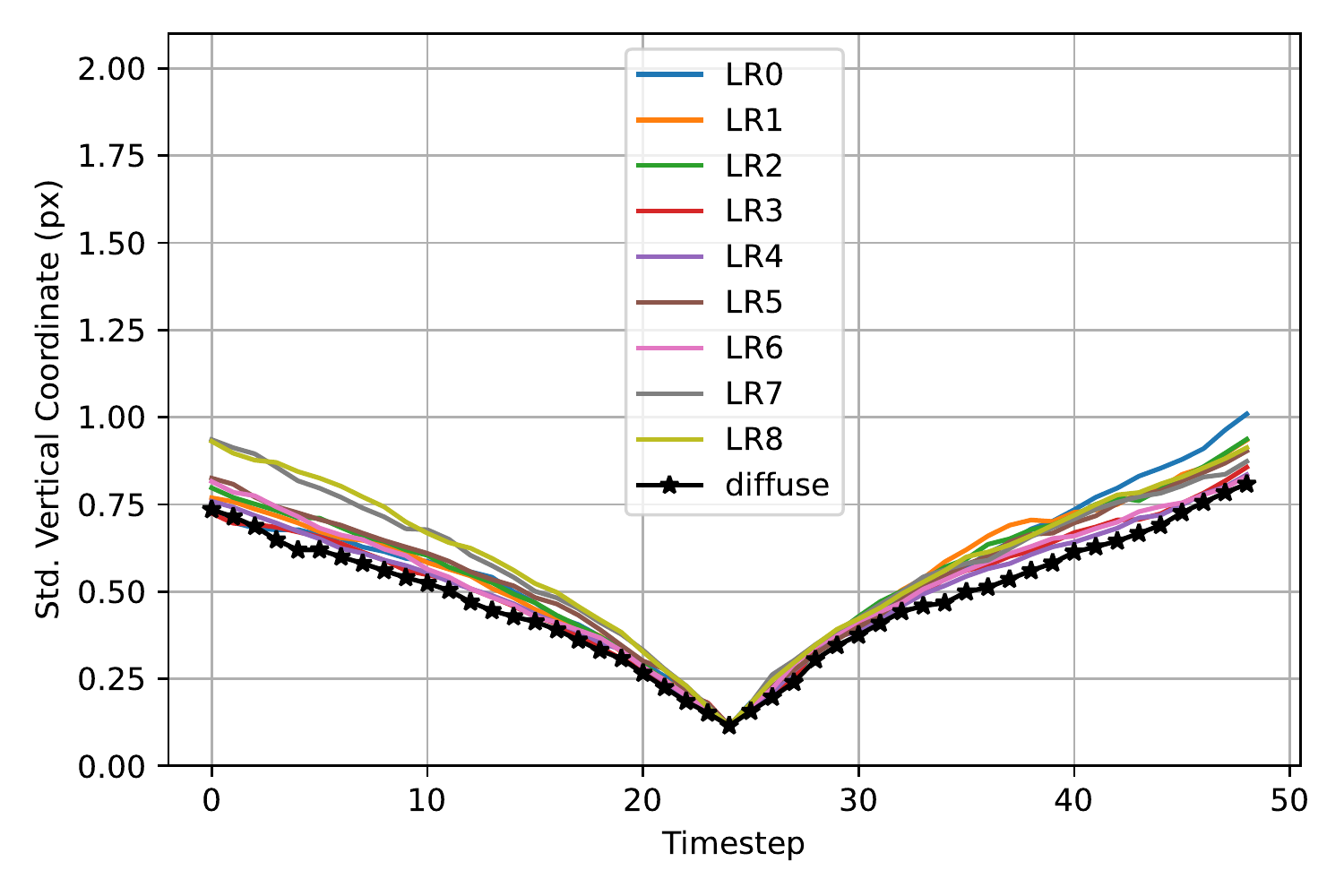}
        \includegraphics[width=0.48\textwidth]{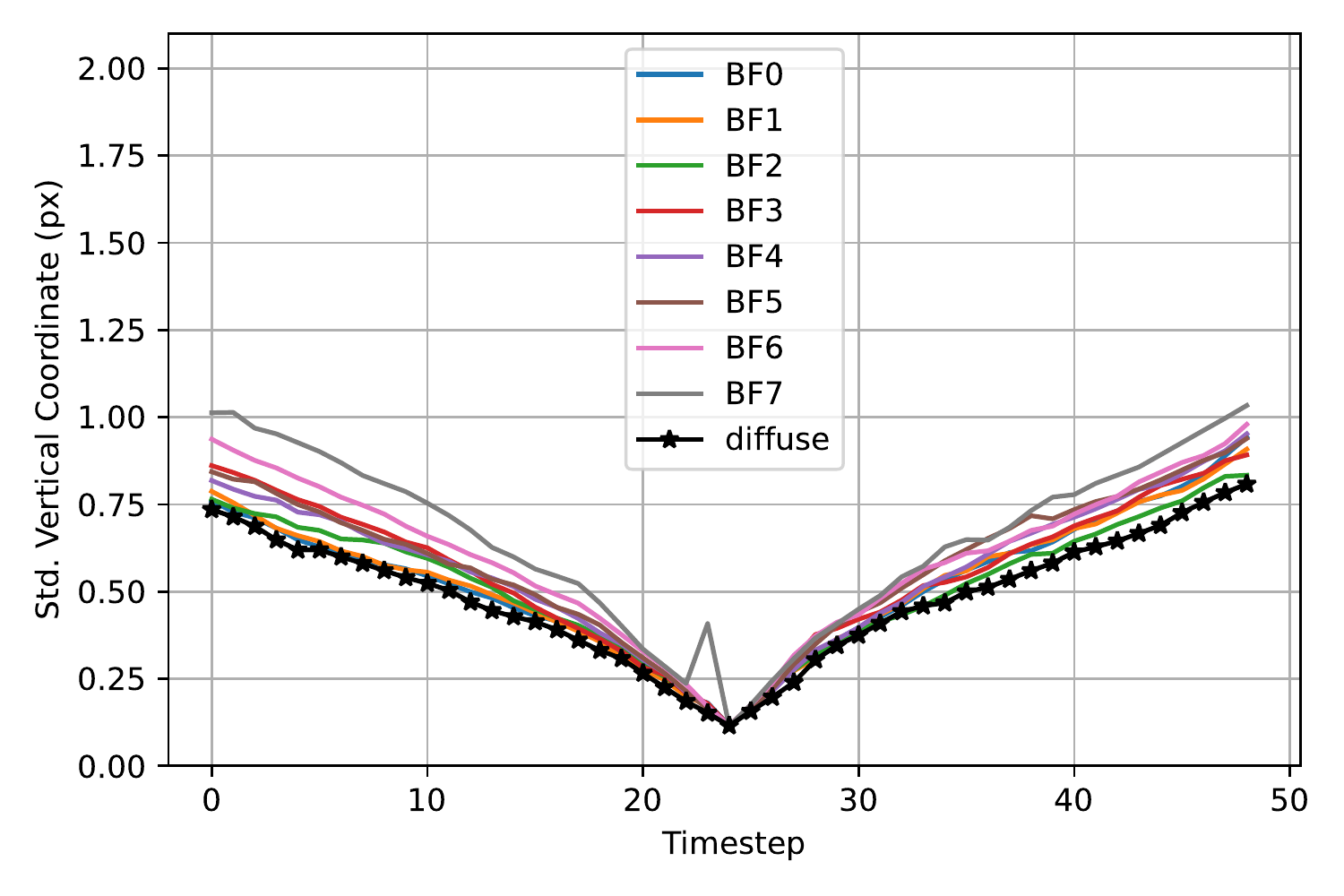}
    }
    \caption{\textbf{DTU Point Features Dataset: The existence of directional lighting does not change trends in covariance $\Sigma(t)$ when using the Lucas-Kanade Tracker at nominal speed.} We compute $\Sigma(t)$ using diffuse lighting (black lines) and each of the directional lighting conditions listed in Figure \ref{fig:dtu_light_stage} using all tracks from all 60 scenes. Timesteps are limited to those that contain at least 100 features. 
    The variation of $\Sigma(t)$ due to the existence of directional lighting is at most 10 percent of the variation common to all plotted lines. The effect of directional lighting is relatively small because changes between adjacent frames are small whether or not the scene contains directional lighting. The blip in the bottom-right figure is due to one specific scene where the AGAST tracker finds very few features, causing a failure in tracking and outlier rejection, and then calculation of $\Sigma(t)$ downstream. }
    \label{fig:dtu_lighting_sigma_LK}
\end{figure}

\begin{figure}[H]
    \centering
    \subfigure[$\Sigma(t)$, Horizontal Coordinate]{
        \includegraphics[width=0.48\textwidth]{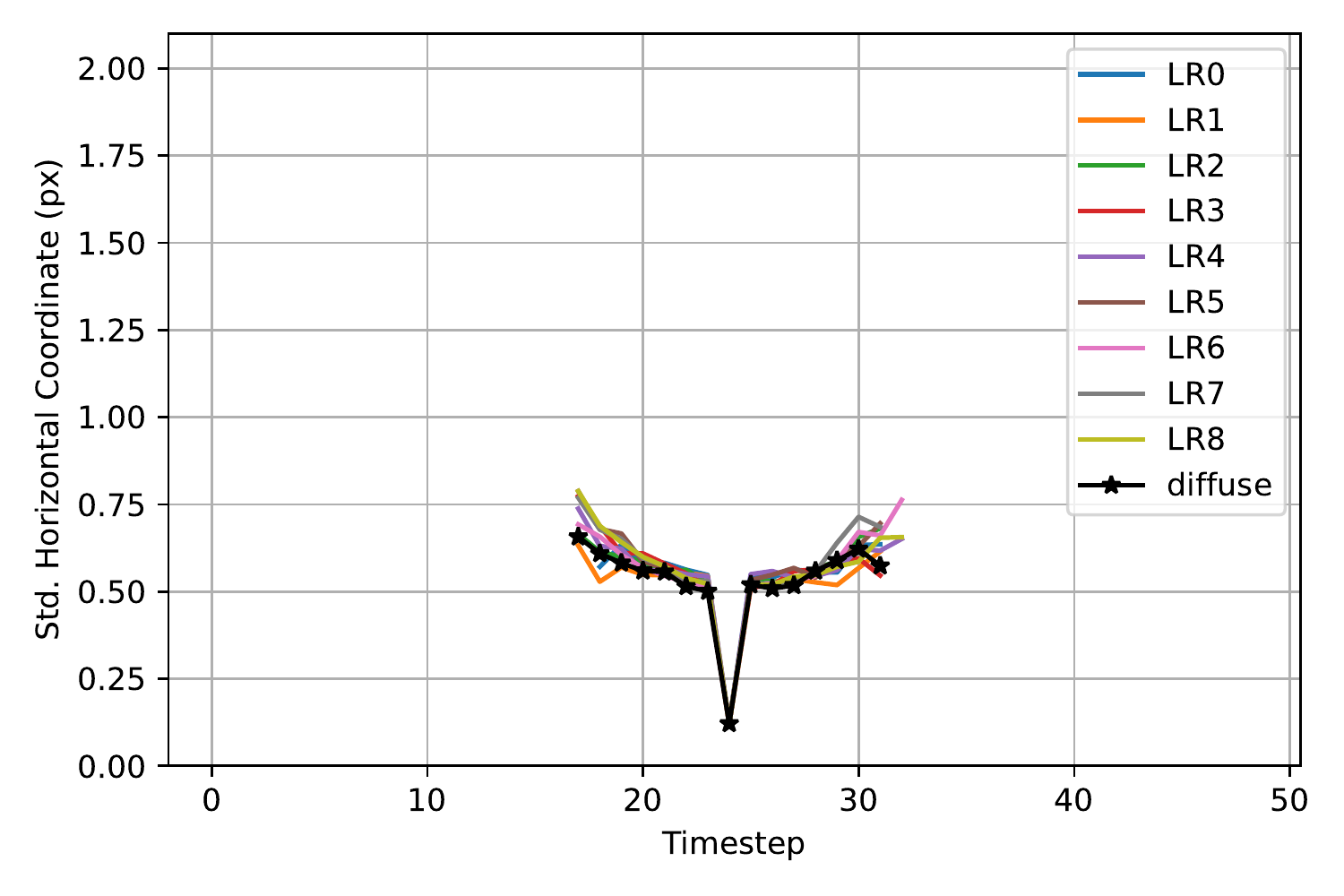}
        \includegraphics[width=0.48\textwidth]{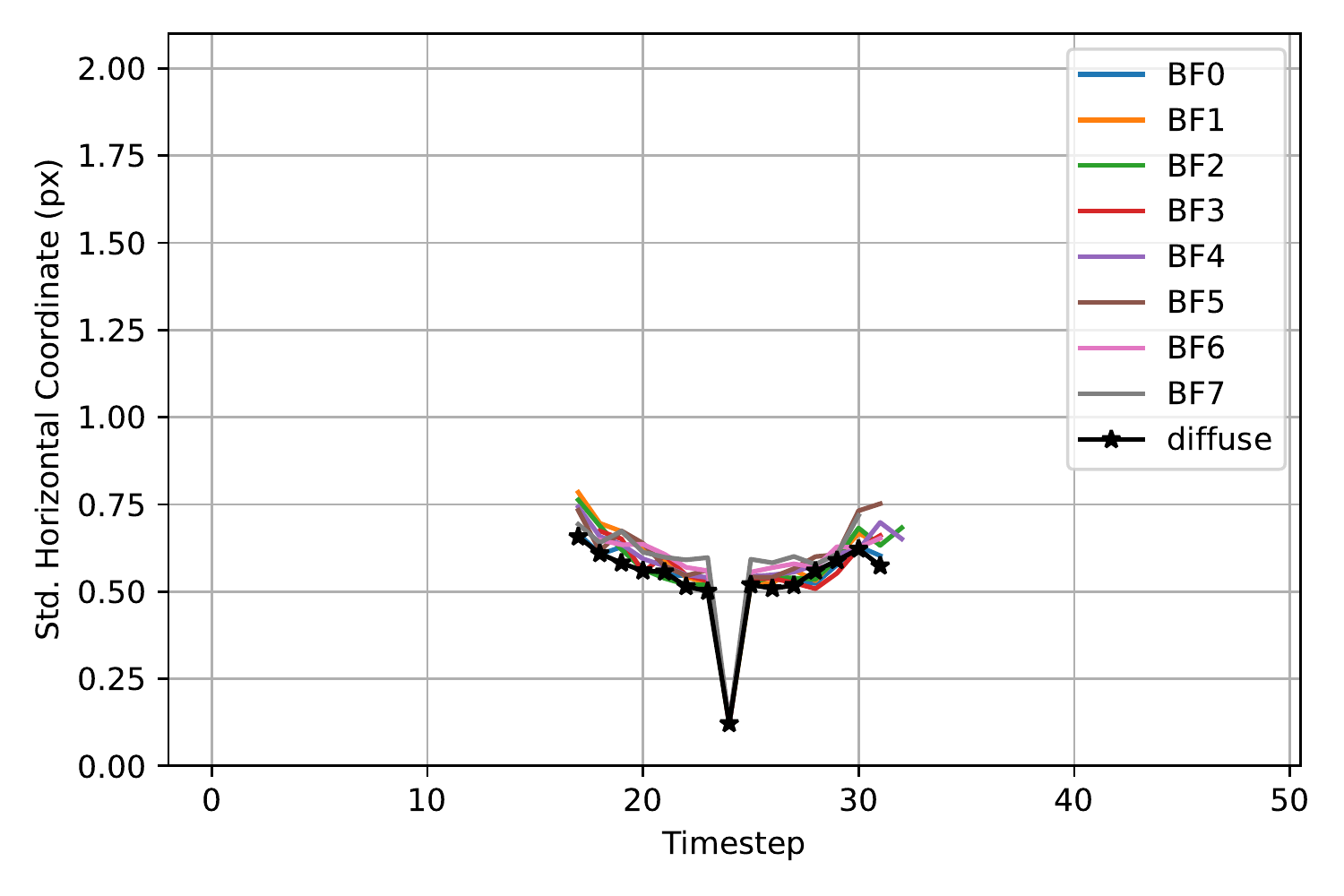} 
    }
    \subfigure[$\Sigma(t)$, Vertical Coordinate]{
        \includegraphics[width=0.48\textwidth]{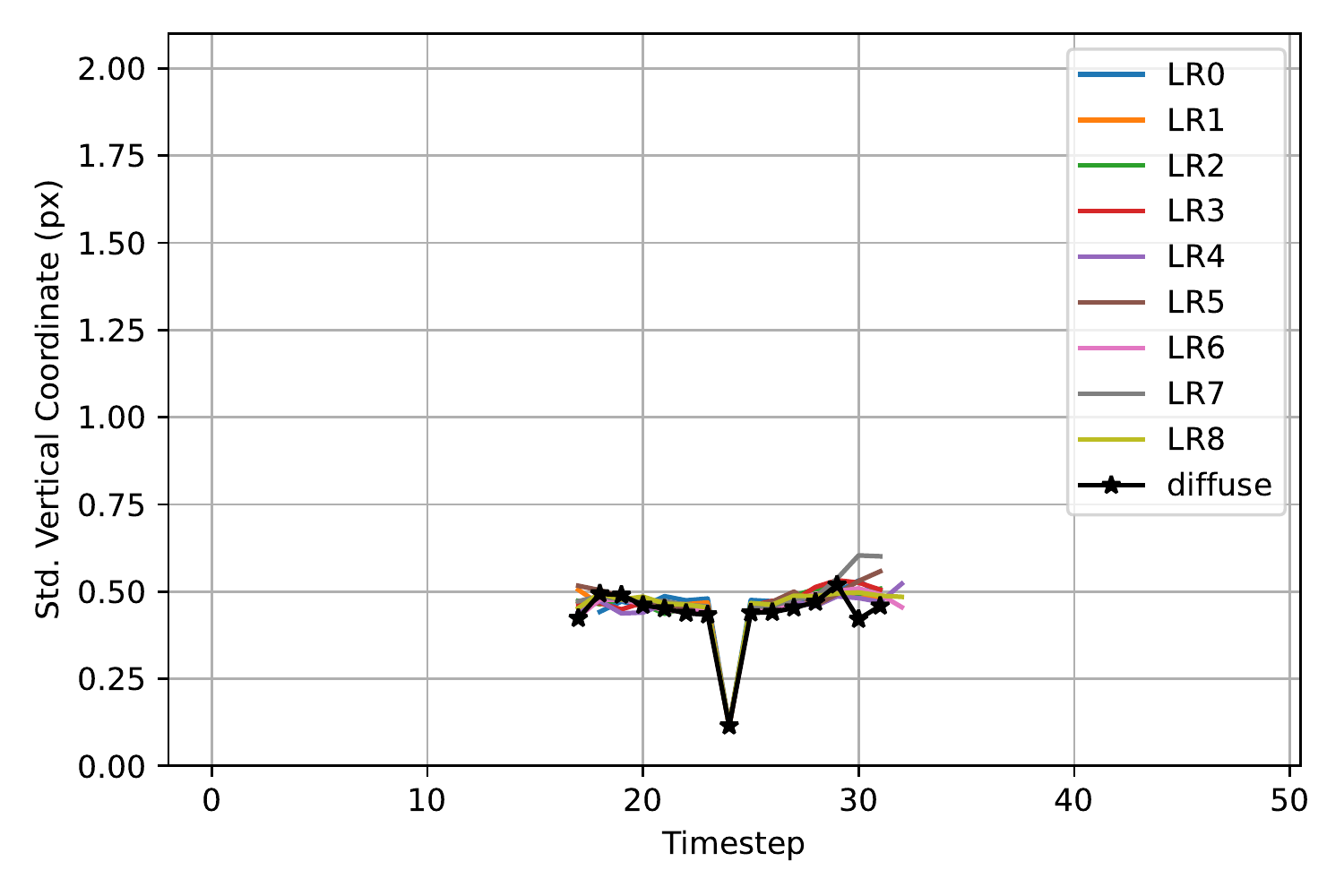} 
        \includegraphics[width=0.48\textwidth]{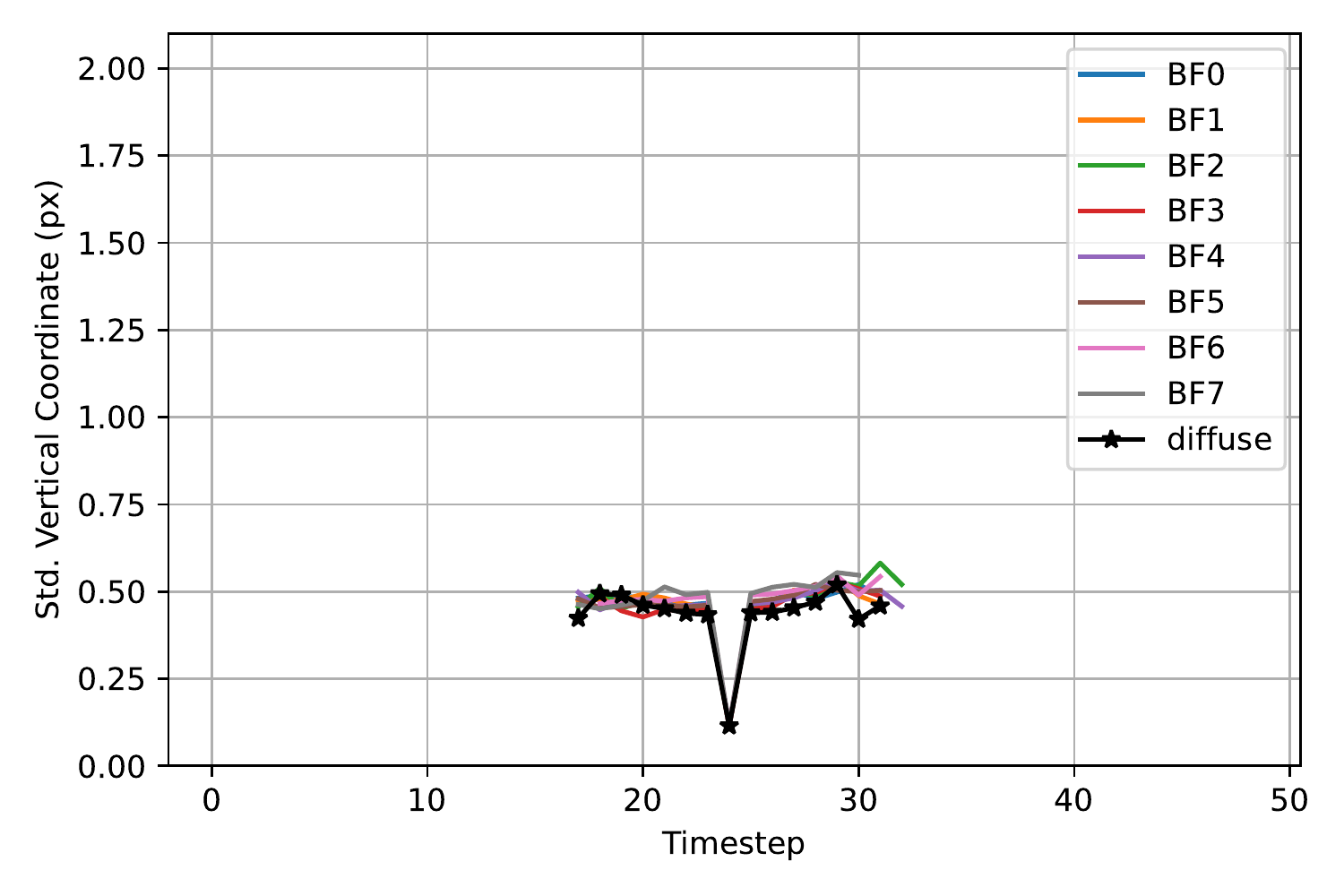} 
    }
    \caption{\textbf{DTU Point Features Dataset: The existence of directional lighting does not change trends in covariance $\Sigma(t)$ when using the Correspondence Tracker at nominal speed.} We compute $\Sigma(t)$ using diffuse lighting (black lines) and each of the directional lighting conditions listed in Figure \ref{fig:dtu_light_stage} using all tracks from all 60 scenes. Timesteps are limited to those that contain at least 100 features. The variation of $\Sigma(t)$ due to the existence of directional lighting is at most 10 percent of the variation common to all plotted lines. The effect of directional lighting is relatively small because changes between adjacent frames are small whether or not the scene contains directional lighting.}
    \label{fig:dtu_lighting_sigma_match}
\end{figure}

\begin{figure}[H]
    \centering
    \subfigure[Lucas-Kanade]{
        \includegraphics[width=0.48\textwidth]{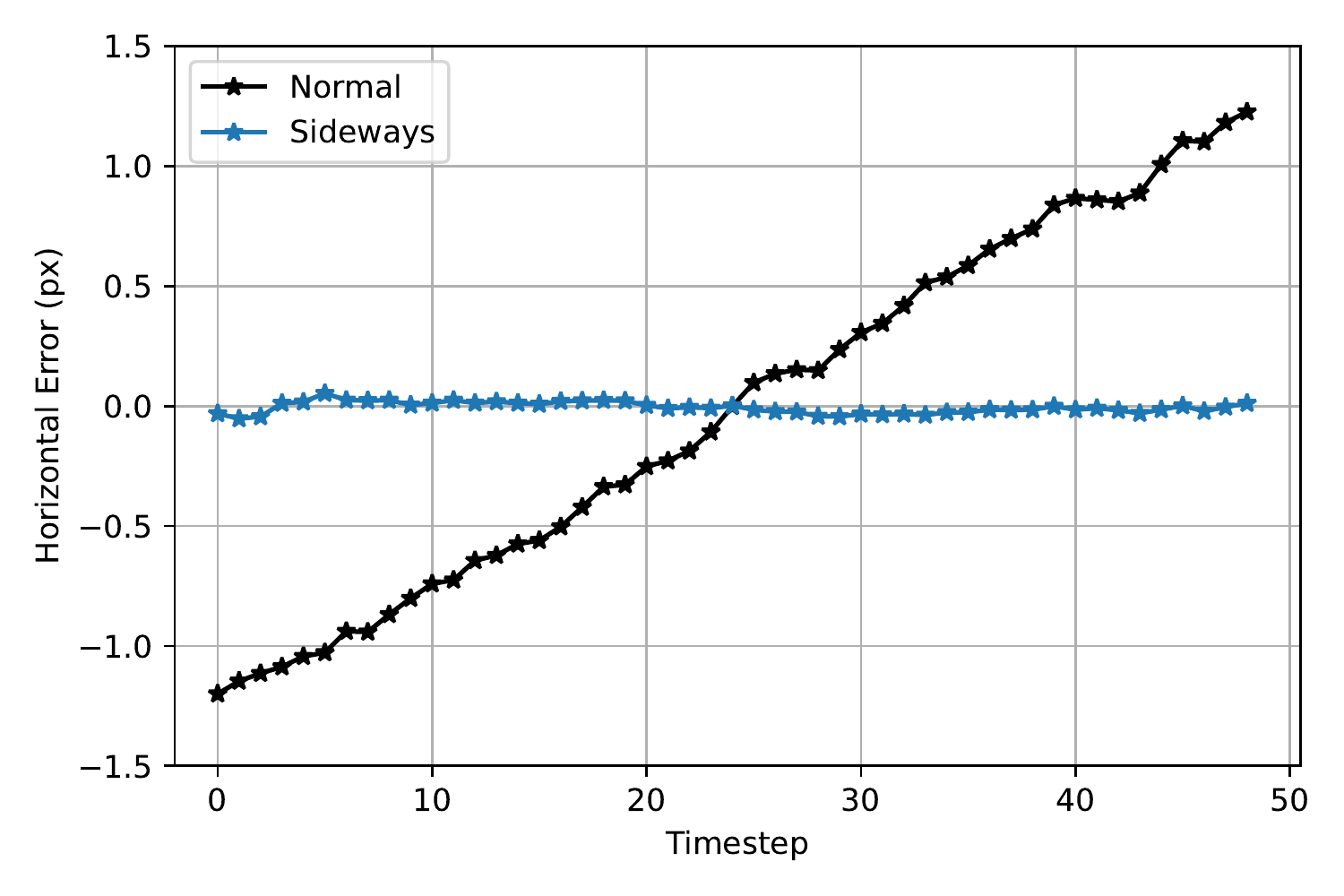}
        \includegraphics[width=0.48\textwidth]{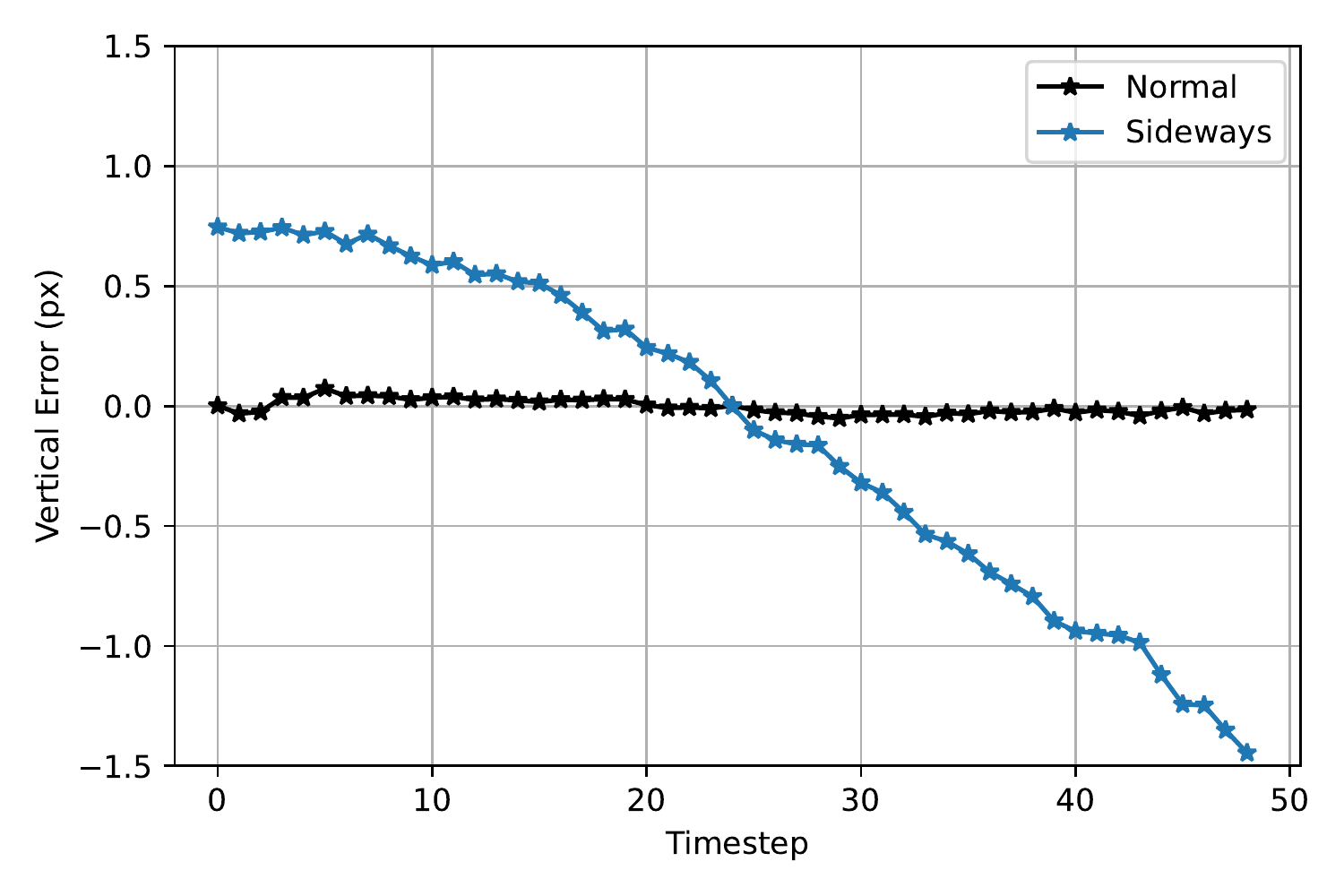}
    }
    \subfigure[Correspondence]{
        \includegraphics[width=0.48\textwidth]{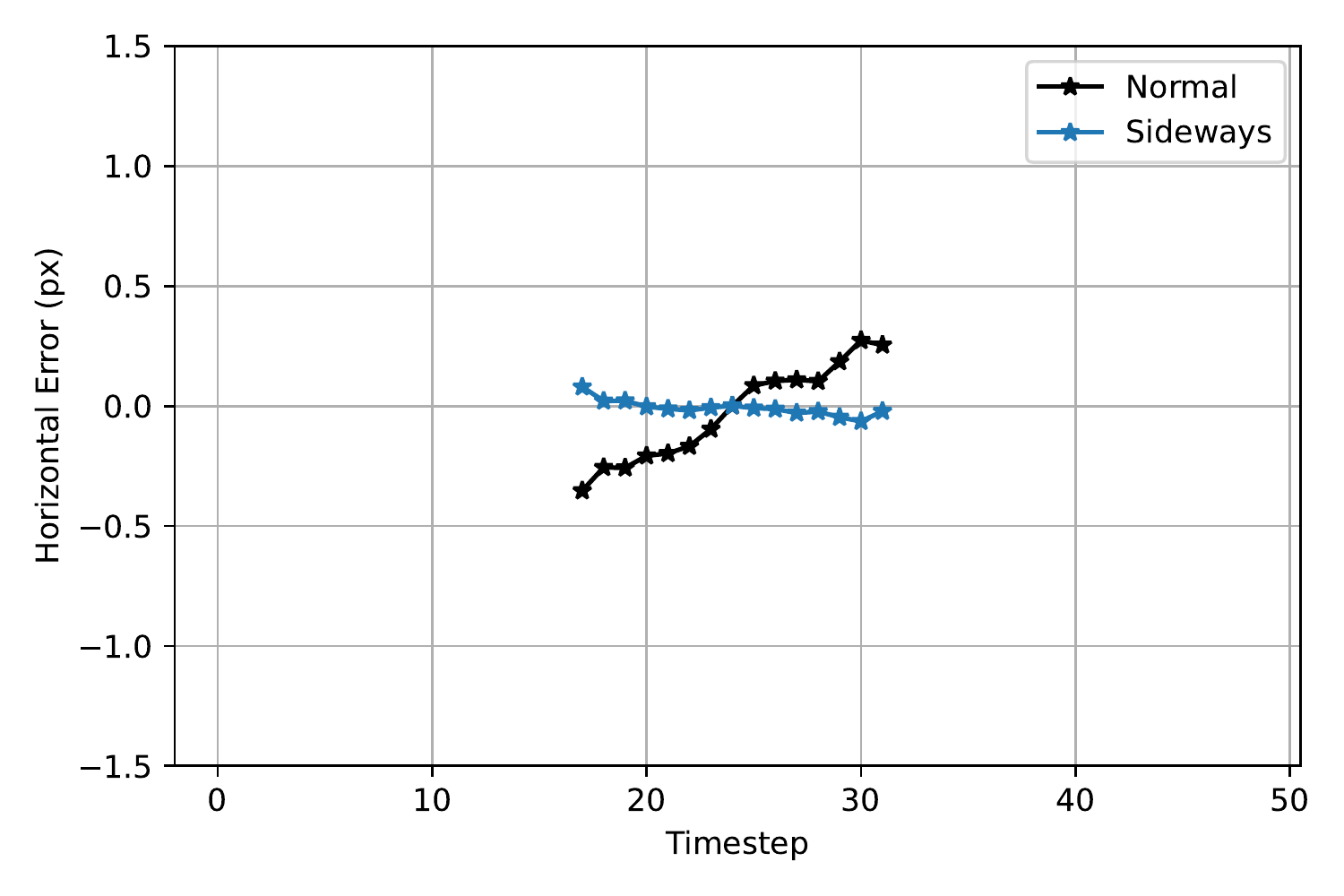} 
        \includegraphics[width=0.48\textwidth]{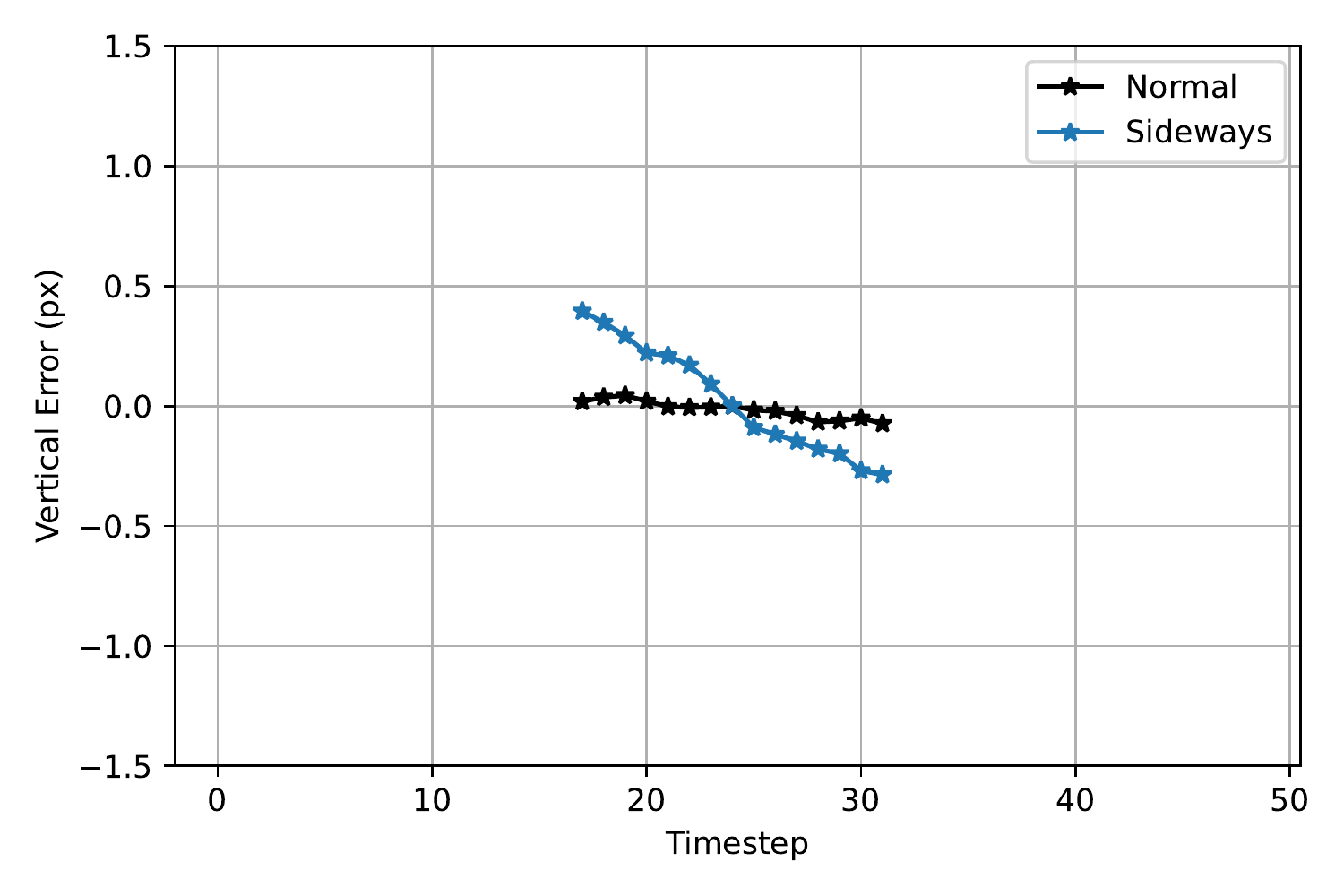} 
    }
    \caption{\textbf{DTU Point Features Dataset: Mean errors are larger about the direction of motion for both the Lucas-Kanade and Correspondence Trackers.} In Figures \ref{fig:dtu_LK_mean_varyspeed}, \ref{fig:dtu_match_diffuse_mean_error_varyspeed}, \ref{fig:dtu_lighting_mu_LK}, and \ref{fig:dtu_lighting_mu_match}, the horizontal component (left column) of $\mu(t)$ was always larger than the vertical component (right column). When images are rotated 90 degrees counterclockwise (``sideways''), the trend is reversed. Errors shown above are computed for the Lucas-Kanade Tracker at nominal speed and in diffuse lighting.}
    \label{fig:dtu_mean_error_sideways}
\end{figure}

\begin{figure}[H]
    \centering
    \subfigure[$\kappa(t)$, Horizontal Coordinate]{\includegraphics[width=0.48\textwidth]{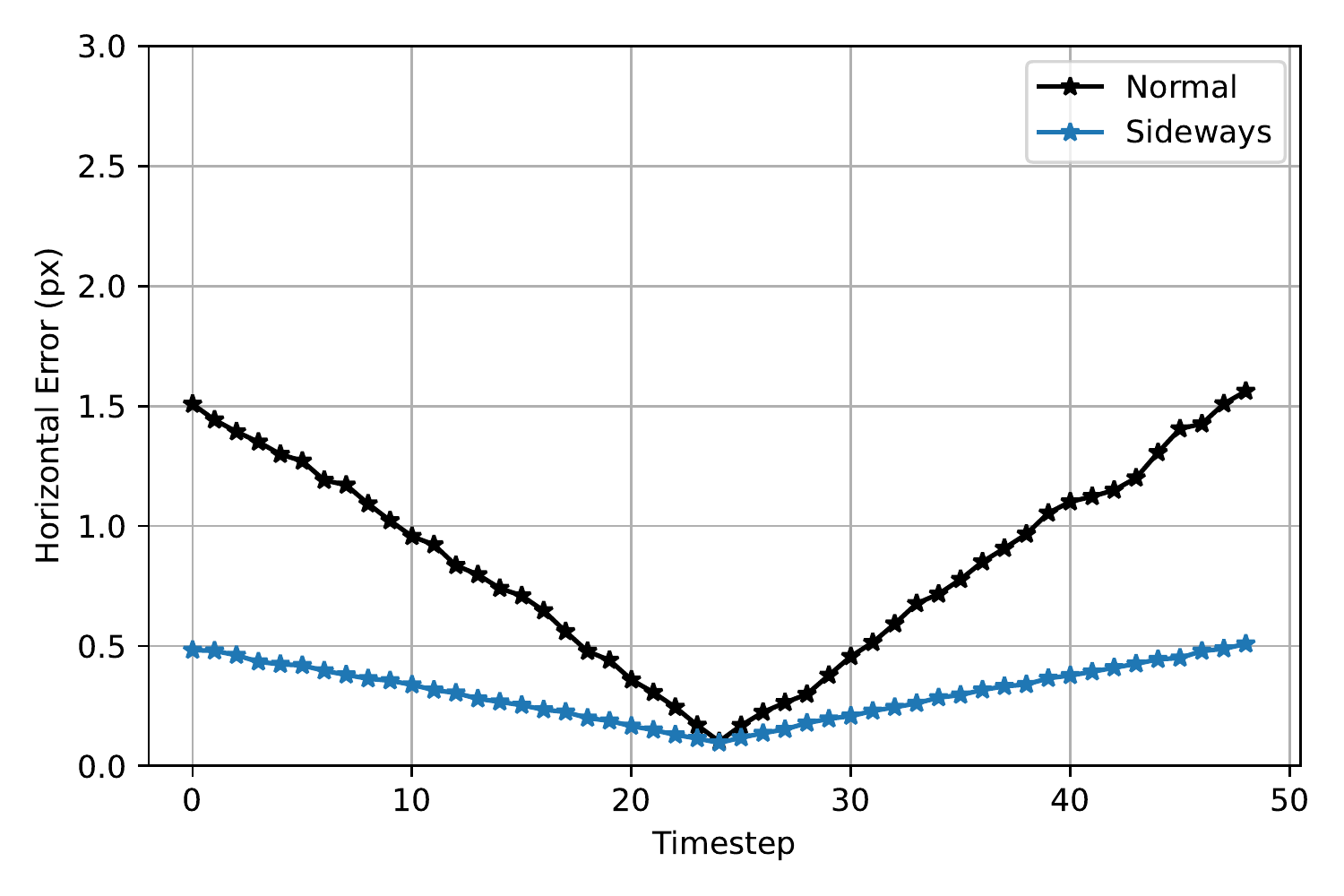}}
    \subfigure[$\kappa(t)$, Vertical Coordinate]{\includegraphics[width=0.48\textwidth]{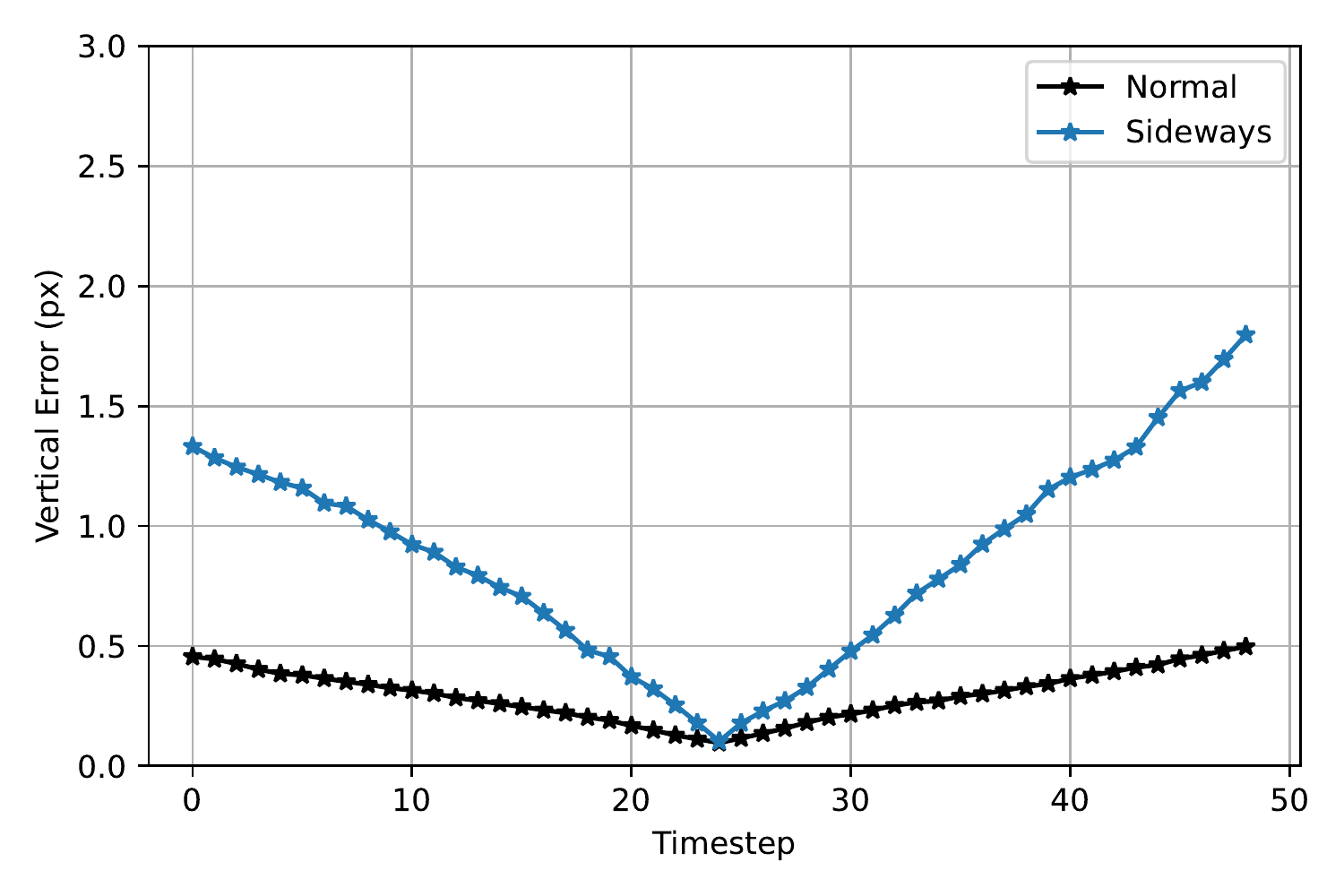}}
    \caption{\textbf{DTU Point Features Dataset: Mean absolute errors are larger about the direction of motion when using the Lucas-Kanade Tracker.} In Figures \ref{fig:dtu_LK_mean_varyspeed}, \ref{fig:dtu_match_diffuse_mean_error_varyspeed}, \ref{fig:dtu_lighting_mu_LK}, and \ref{fig:dtu_lighting_mu_match}, the horizontal component of $\kappa(t)$ was always larger than the vertical component. When images are rotated 90 degress counterclockwise (``sideways''), the trend is reversed. Mean absolute errors shown above are computed for the Lucas-Kanade Tracker at nominal speed and in diffuse lighting. }
    \label{fig:dtu_abs_error_sideways}
\end{figure}

\begin{figure}[H]
    \centering
    \subfigure[$\kappa(t)$, Horizontal Coordinate]{\includegraphics[width=0.48\textwidth]{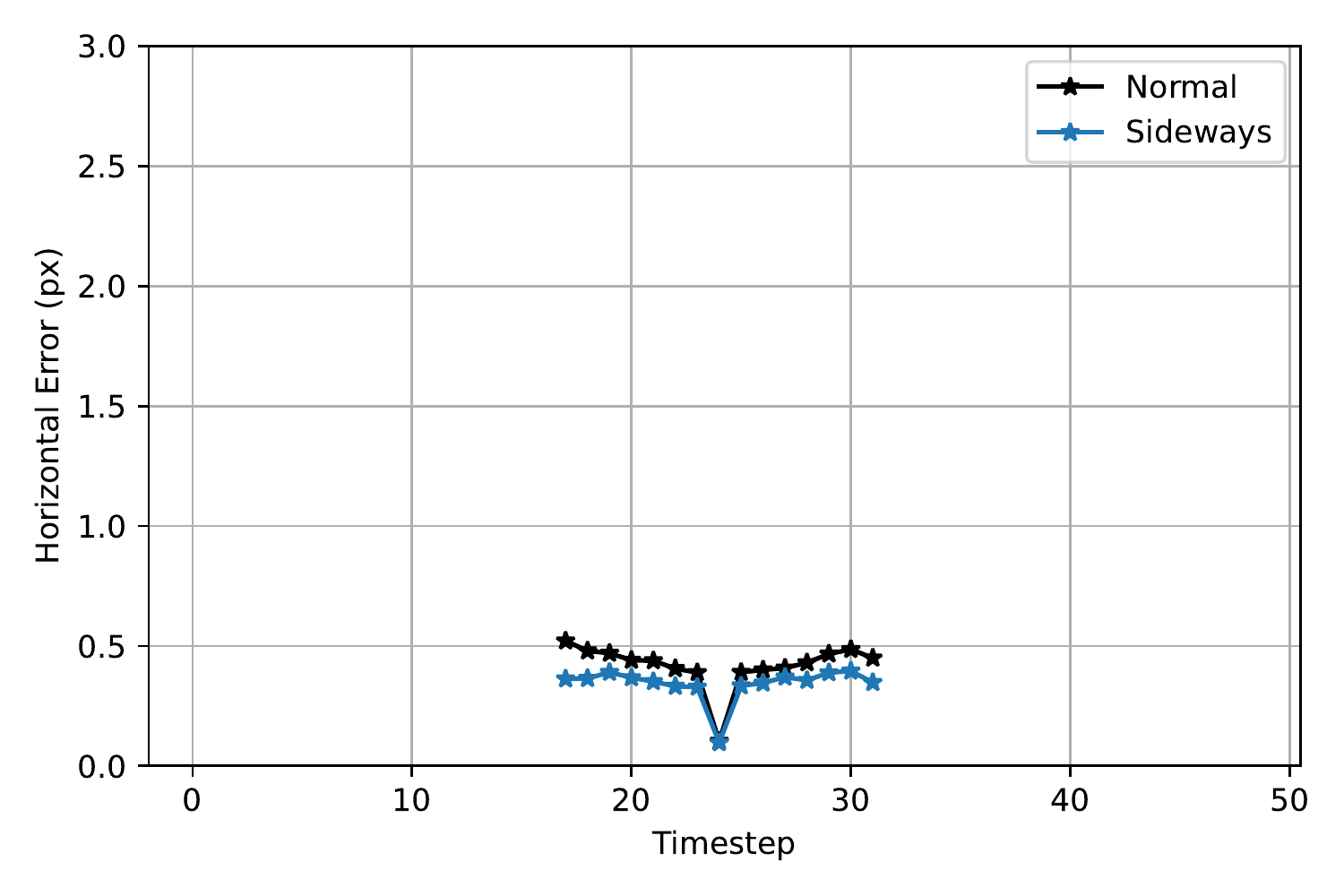}}
    \subfigure[$\kappa(t)$, Vertical Coordinate]{\includegraphics[width=0.48\textwidth]{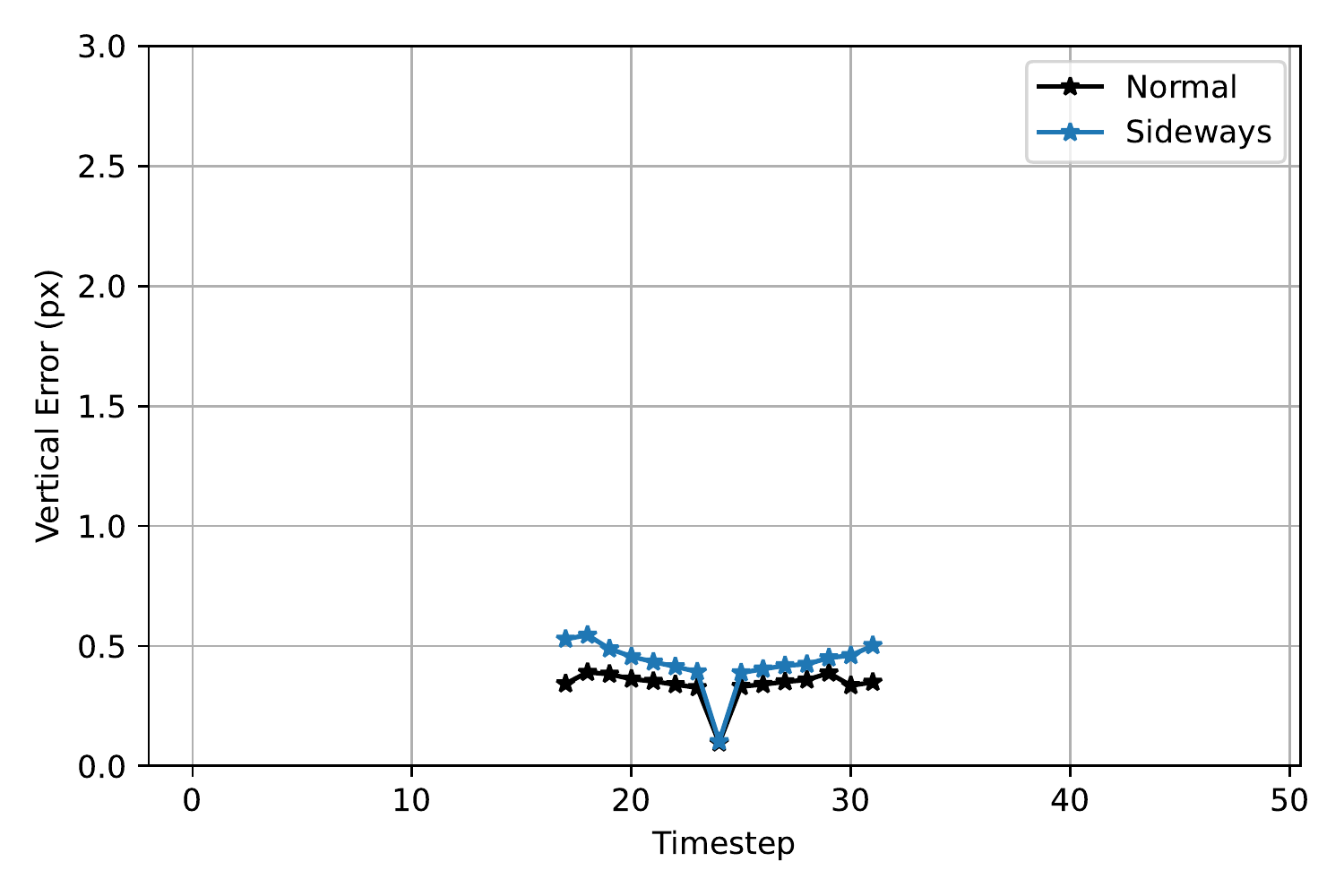}}
    \caption{\textbf{DTU Point Features Dataset: The direction of motion does not affect mean absolute error when using the Correspondence Tracker.} In Figures \ref{fig:dtu_match_diffuse_MAE_varyspeed}, and \ref{fig:dtu_lighting_omega_match}, the difference between the horizontal and vertical components of $\kappa(t)$ was a fraction of the size of $\kappa(t)$ in both components. When images are rotated 90 degrees counterclockwise (``sideways''), the trend is the same. Errors shown above are computed for the Correspondence Tracker at nominal speed and in diffuse lighting. }
    \label{fig:dtu_match_abs_error_sideways}
\end{figure}

\begin{figure}[H]
    \centering
    \subfigure[$\Sigma(t)$, Horizontal Coordinate]{\includegraphics[width=0.48\textwidth]{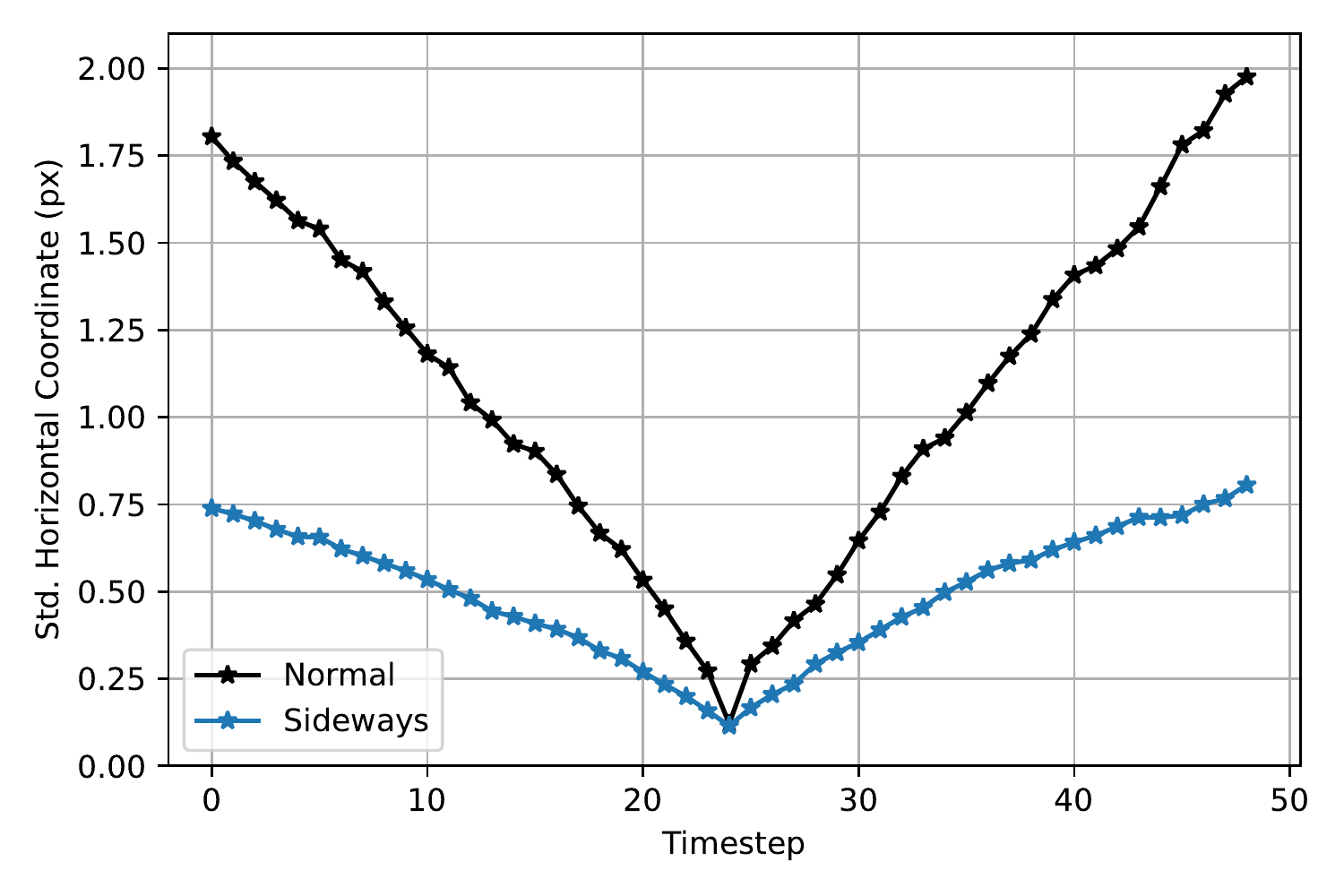}}
    \subfigure[$\Sigma(t)$, Vertical Coordinate]{\includegraphics[width=0.48\textwidth]{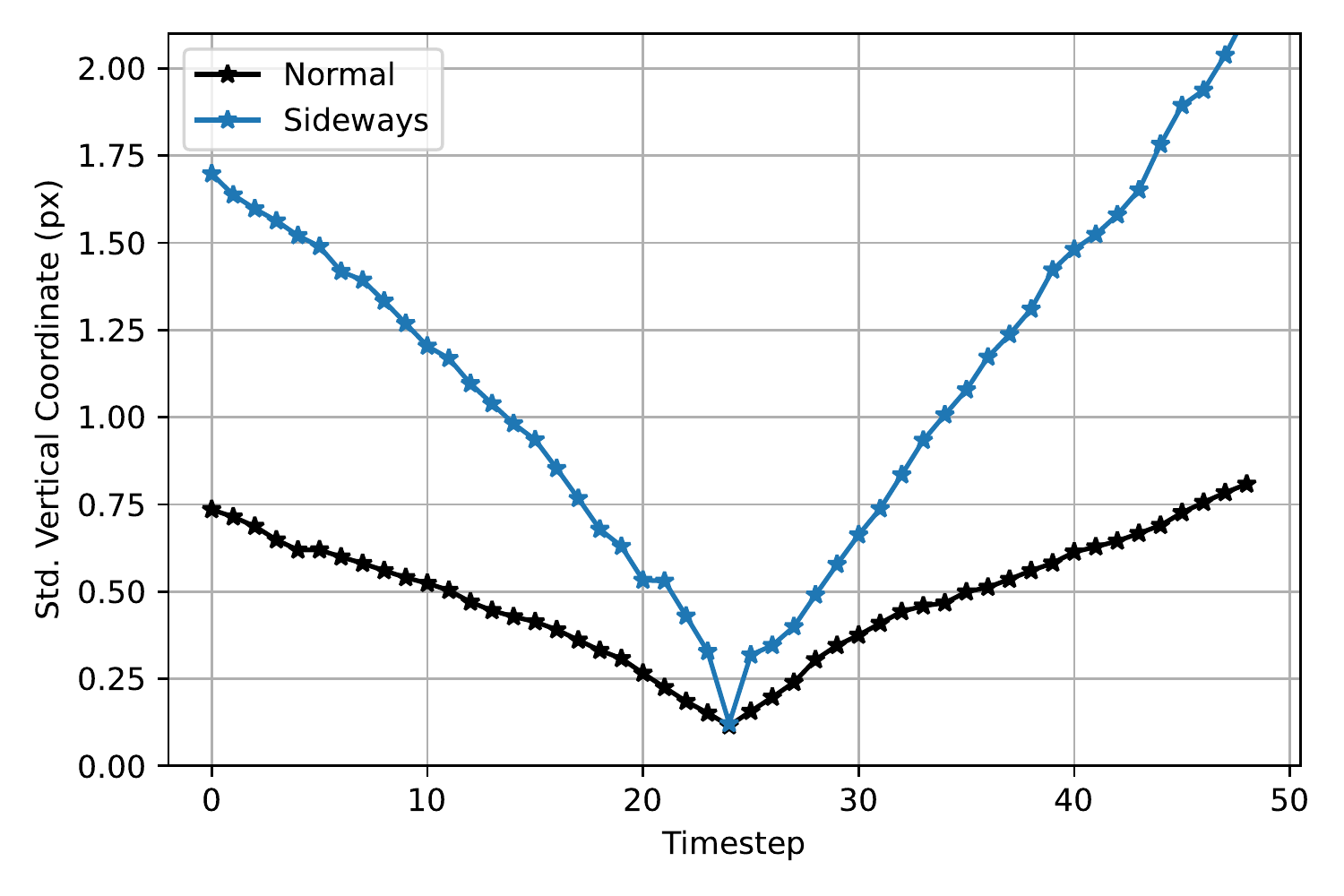}}
    \caption{\textbf{DTU Point Features Dataset: Covariances are larger about the direction of motion when using the Lucas-Kanade Tracker.} In Figures \ref{fig:dtu_LK_cov_varyspeed}, \ref{fig:dtu_match_diffuse_cov_varyspeed}, \ref{fig:dtu_lighting_sigma_LK}, and \ref{fig:dtu_lighting_sigma_match}, the horizontal component of $\Sigma(t)$ was always larger than the vertical component. When images are rotated 90 degrees counterclockwise (``sideways"), the trend is reversed for both errors (top row) and covariance (bottom row). Errors above are computed for the Lucas-Kanade Tracker at nominal speed and in diffuse lighting.}
    \label{fig:dtu_error_sideways}
\end{figure}

\begin{figure}[H]
    \centering
    \subfigure[$\Sigma(t)$, Horizontal Coordinate]{\includegraphics[width=0.48\textwidth]{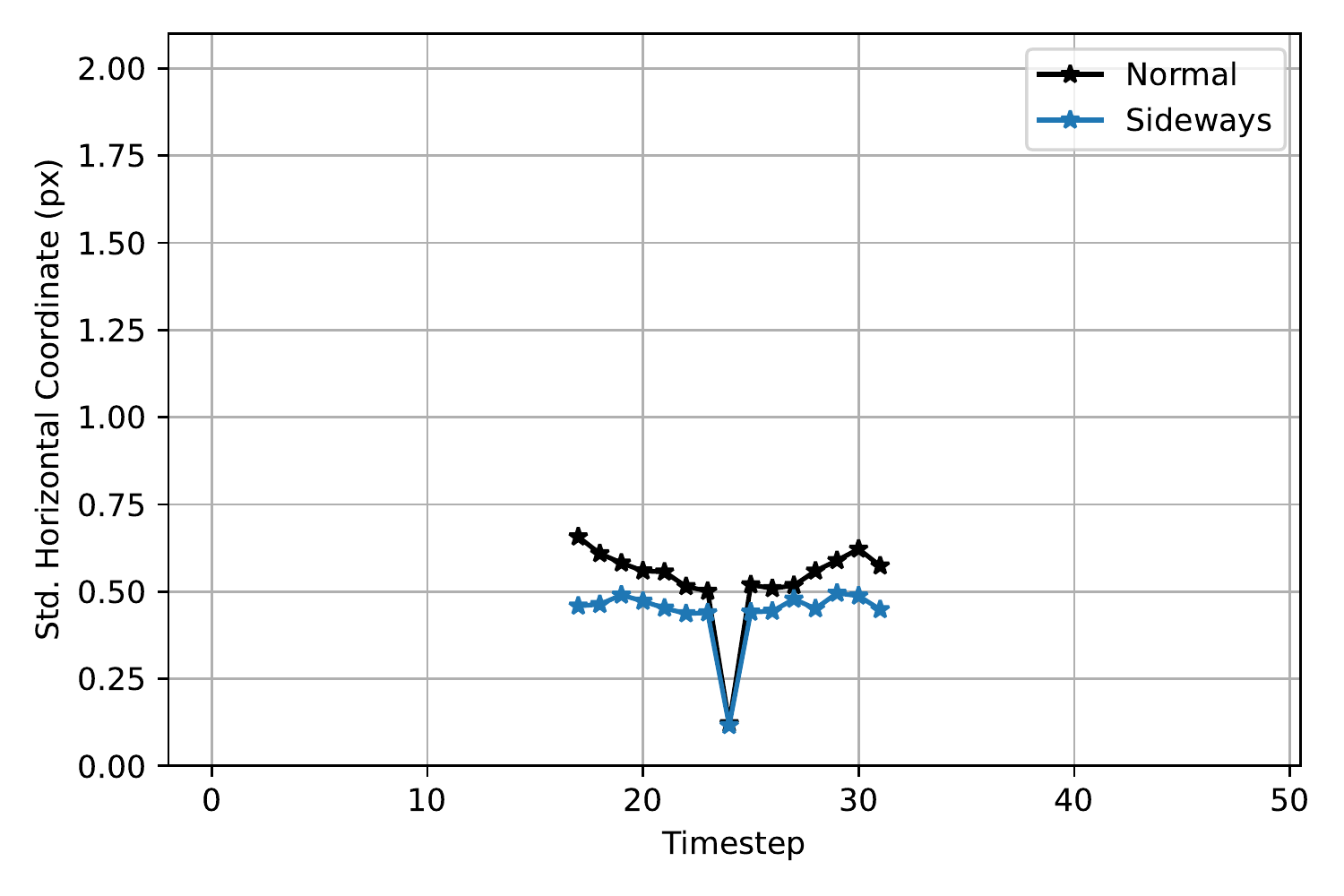}}
    \subfigure[$\Sigma(t)$, Vertical Coordinate]{\includegraphics[width=0.48\textwidth]{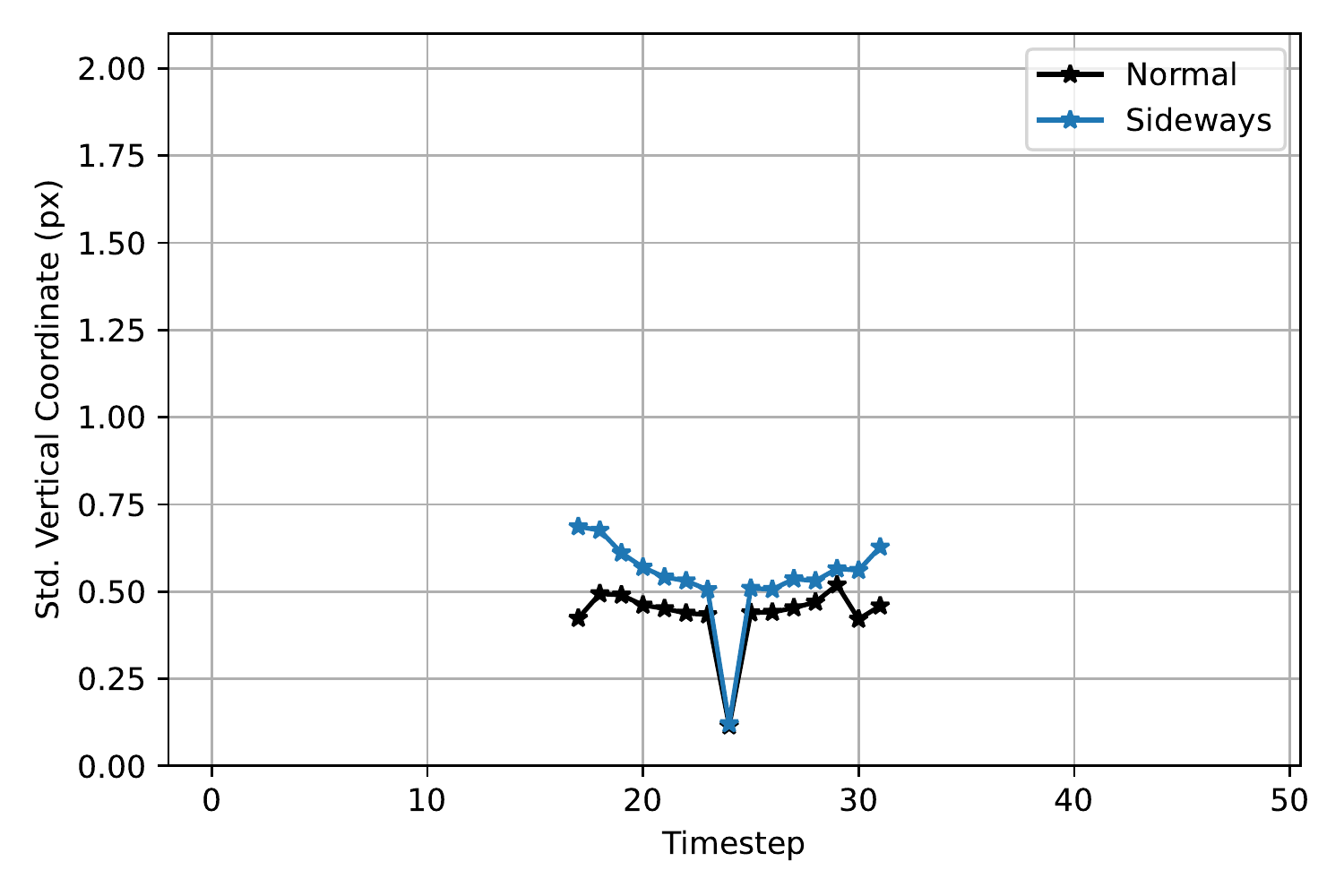}}
    \caption{\textbf{DTU Point Features Dataset: The direction of motion does not affect covariance when using the Correspondence Tracker.} In Figures \ref{fig:dtu_match_diffuse_cov_varyspeed}, and \ref{fig:dtu_lighting_sigma_match}, the difference between the horizontal and vertical components of $\Sigma(t)$ was a fraction of the size of $\Sigma(t)$ in both components. When images are rotated 90 degrees counterclockwise (``sideways''), the trend is the same. Errors shown above are computed for the Correspondence Tracker at nominal speed and in diffuse lighting. }
    \label{fig:dtu_match_cov_sideways}
\end{figure}

\begin{figure}[H]
    \centering
    \subfigure[Lucas-Kanade]{
        \includegraphics[width=0.48\textwidth]{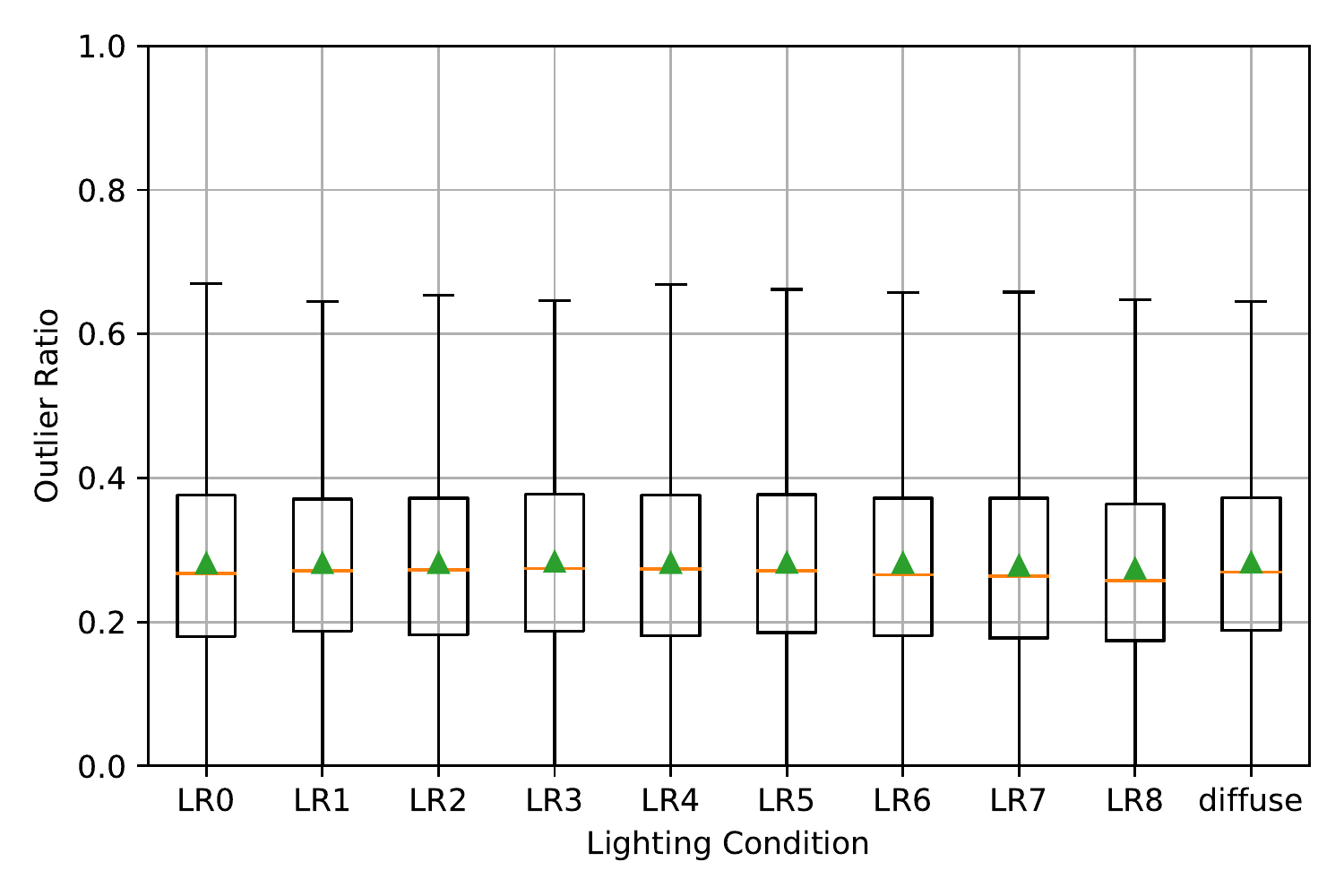}
        \includegraphics[width=0.48\textwidth]{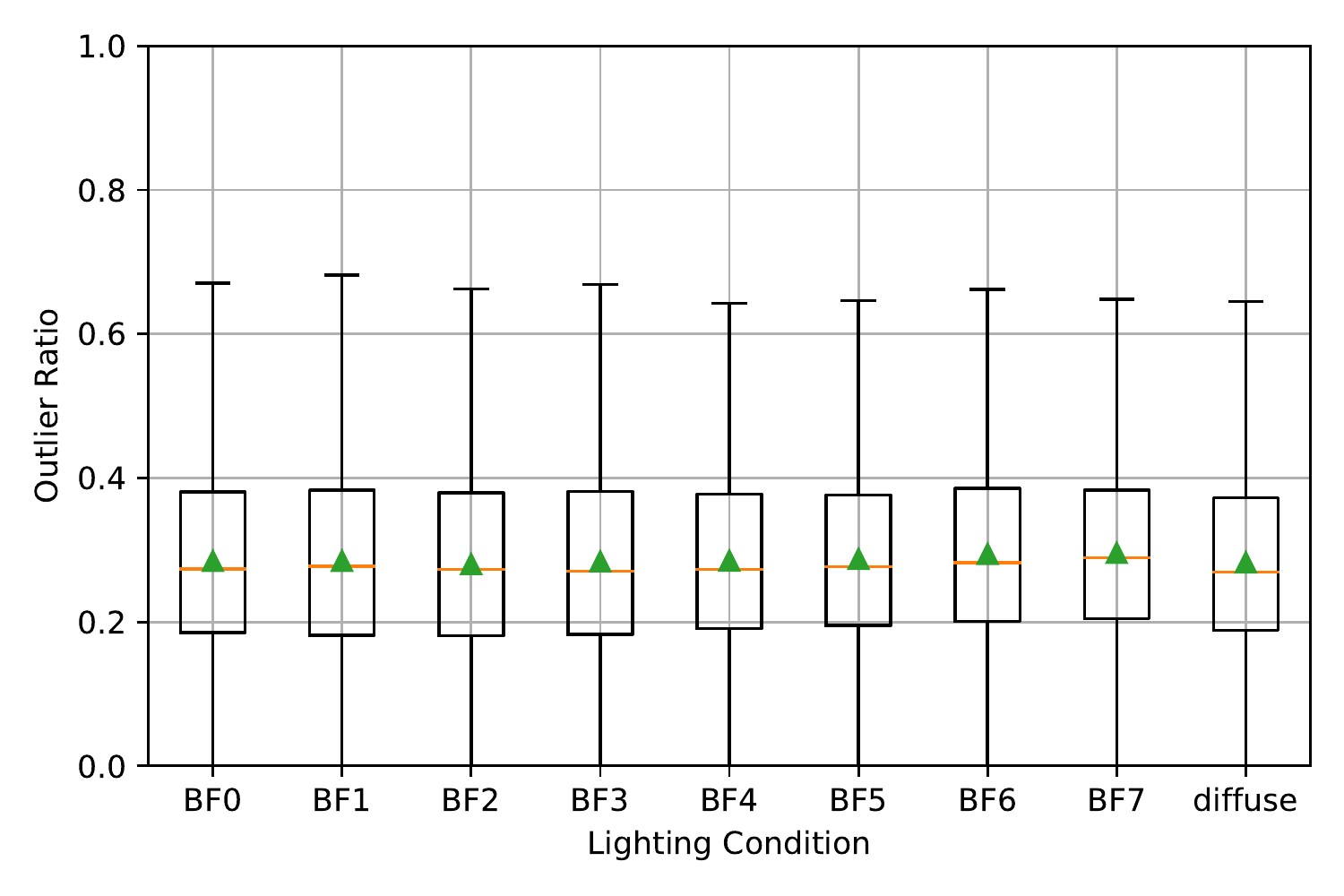}
    }
    \subfigure[Correspondence]{
        \includegraphics[width=0.48\textwidth]{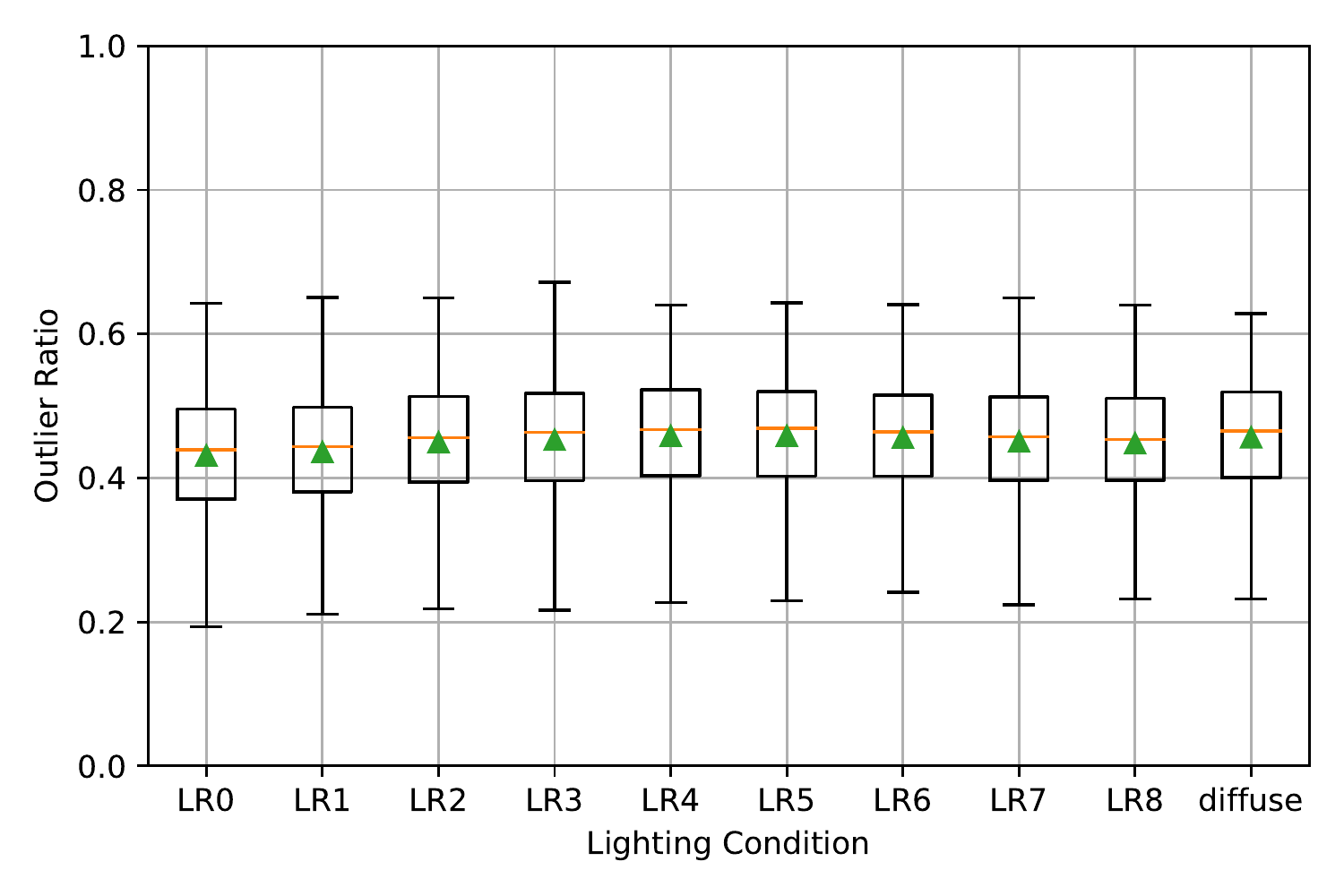}
        \includegraphics[width=0.48\textwidth]{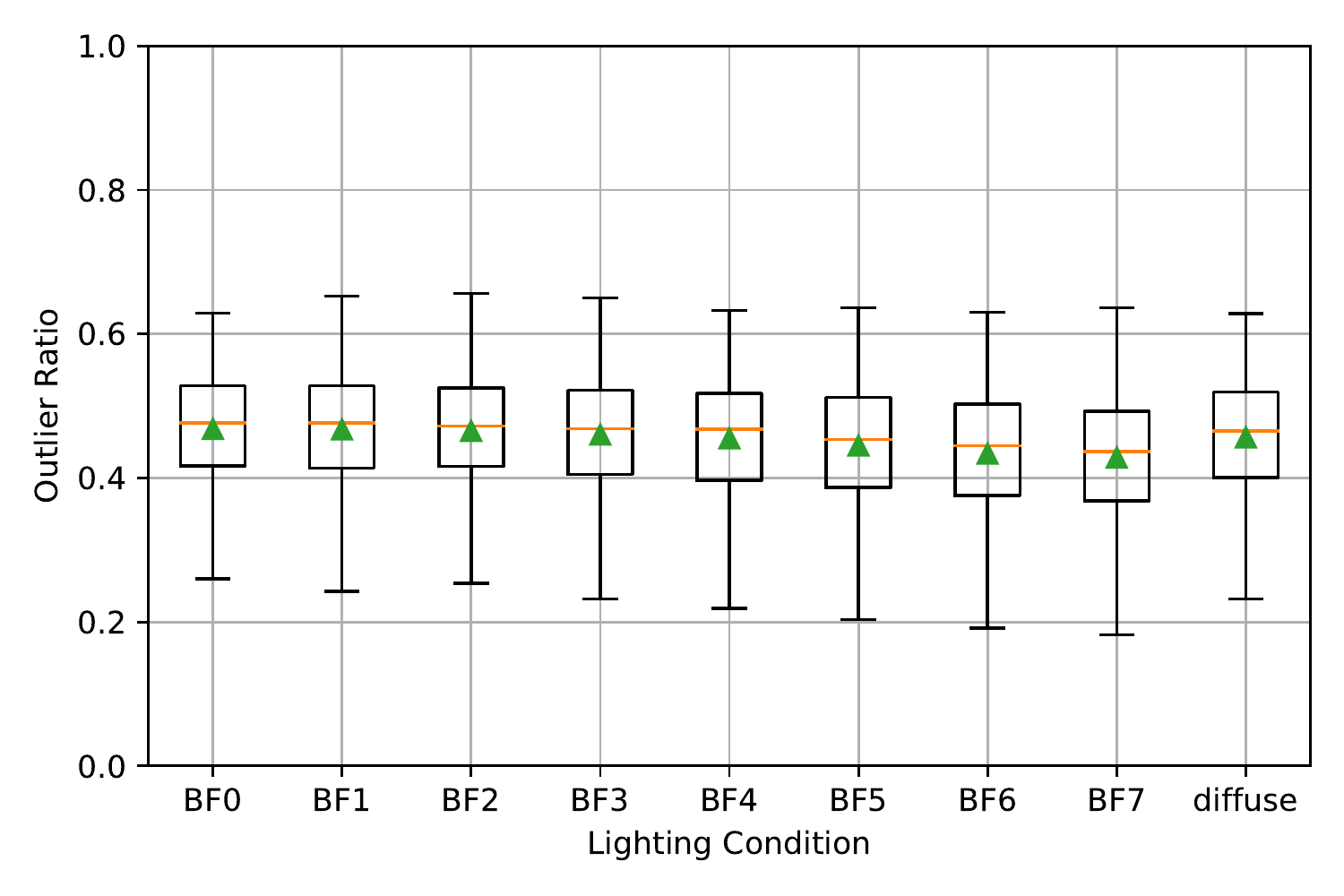}
    }
    \caption{\textbf{At twice nominal speed, the existence of directional lighting does not affect outlier ratio.} In the box-and-whisker plots above, the orange line is the median, the green triangle is the mean, and the box extends from the first to the third quartiles. The whiskers extend up to 1.5x the length of the boxes. Each box-and-whisker plot is computed using features from all 60 scenes, one tracker, speed=2.00, and one of the lighting conditions in Figure \ref{fig:dtu_light_stage}. The distribution of outlier ratio is approximately the same for all lighting conditions.}
    \label{fig:dtu_speed2.00_percent_outlier}
\end{figure}

\begin{figure}[H]
    \centering
    \subfigure[Lucas-Kanade]{
        \includegraphics[width=0.48\textwidth]{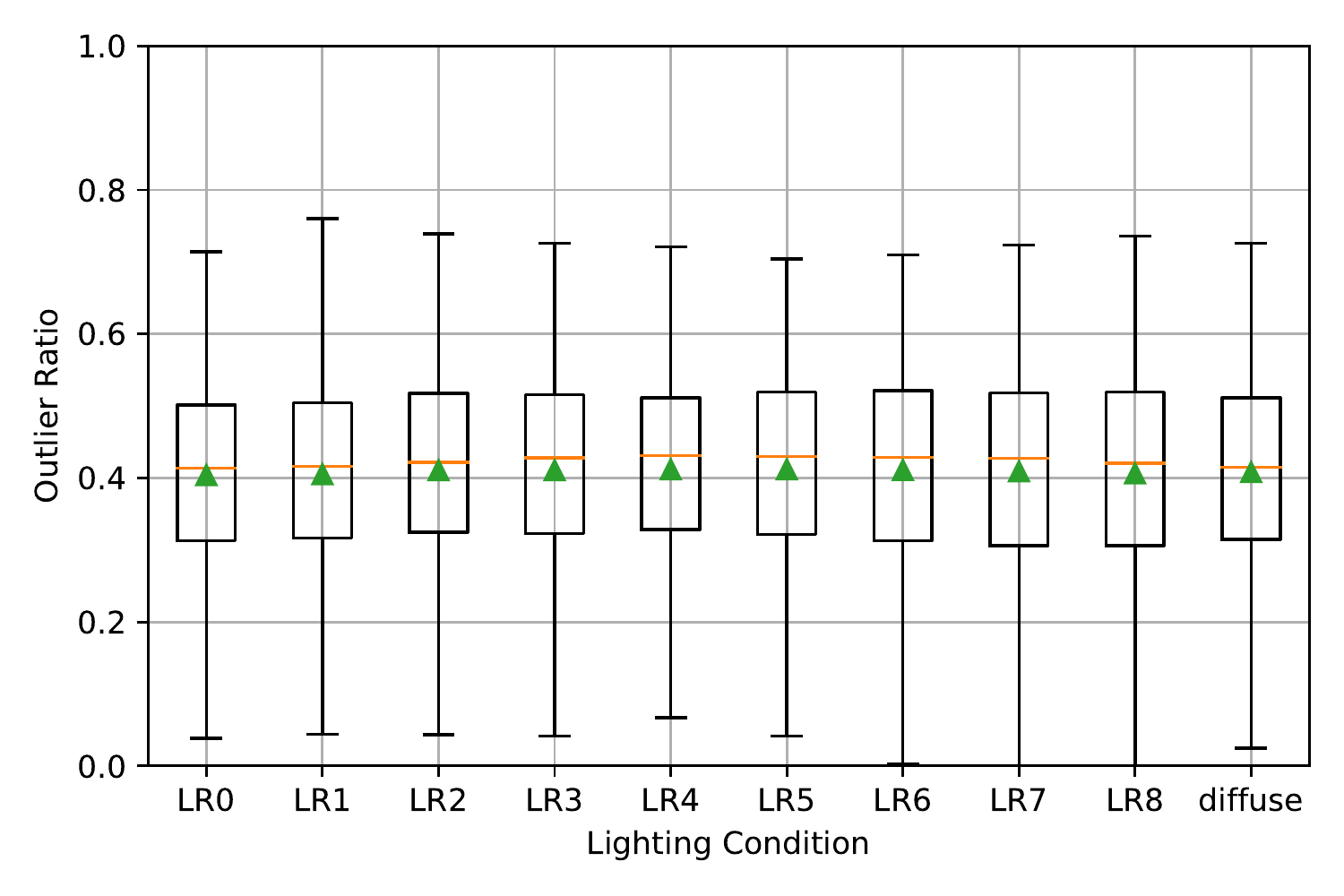}
        \includegraphics[width=0.48\textwidth]{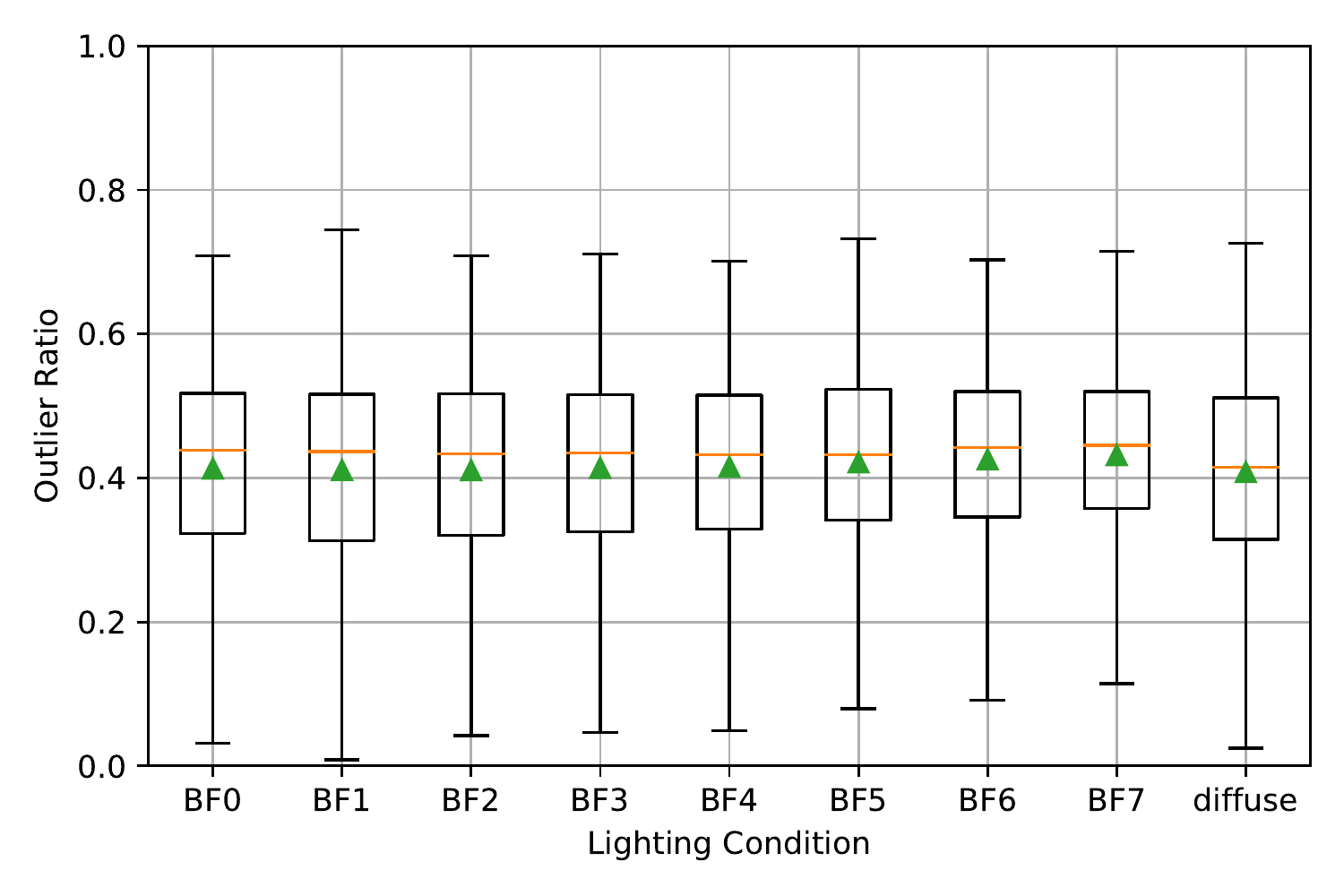}
    }
    \subfigure[Correspondence]{
        \includegraphics[width=0.48\textwidth]{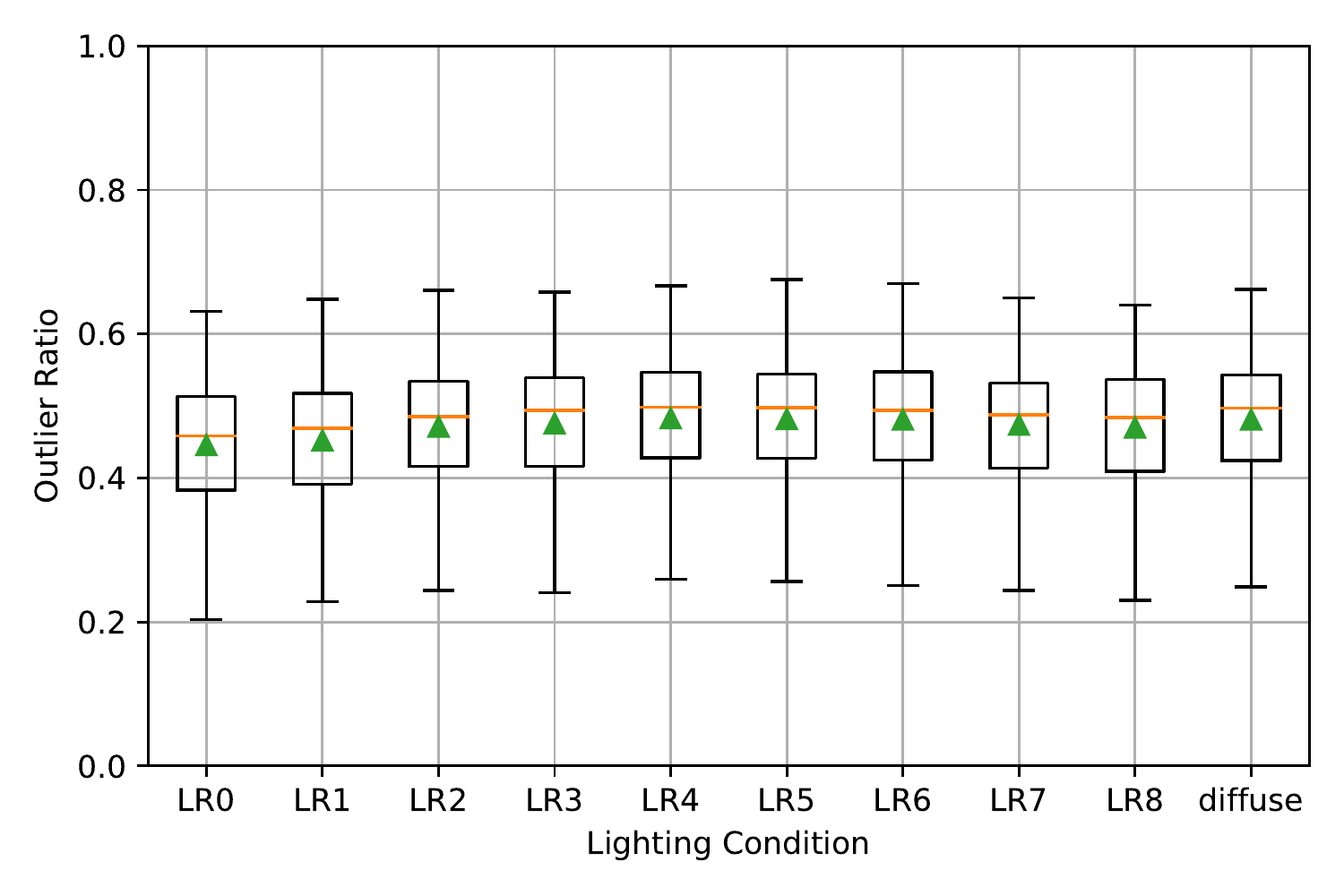}
        \includegraphics[width=0.48\textwidth]{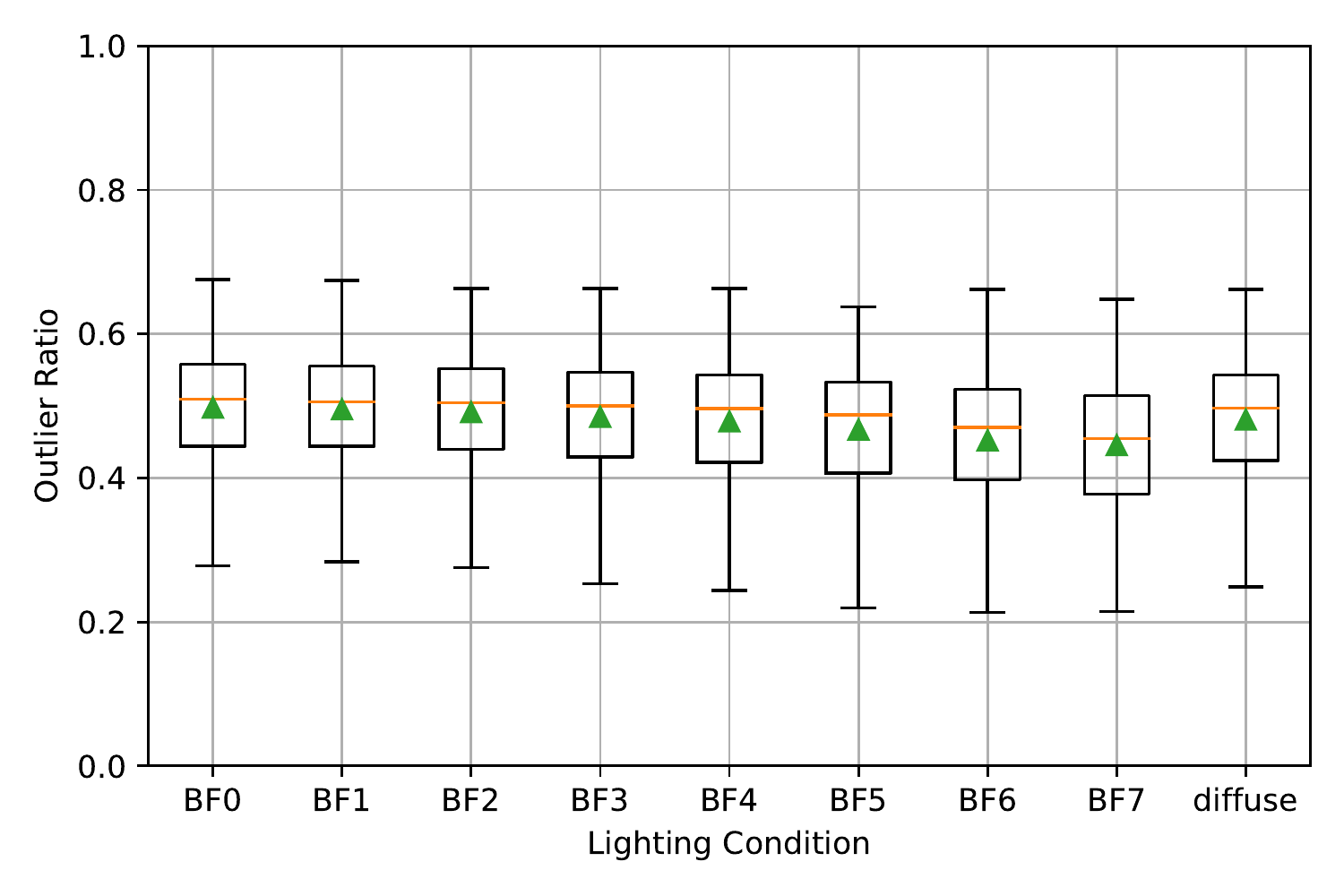}
    }
    \caption{\textbf{At four times nominal speed, the existence of directional lighting does not affect outlier ratio.} In the box-and-whisker plots above, the orange line is the median, the green triangle is the mean, and the box extends from the first to the third quartiles. The whiskers extend up to 1.5x the length of the boxes. Each box-and-whisker plot is computed using features from all 60 scenes, one tracker, speed=4.00, and one of the lighting conditions in Figure \ref{fig:dtu_light_stage}. The distribution of outlier ratio is approximately the same for all lighting conditions.}
    \label{fig:dtu_speed4.00_percent_outlier}
\end{figure}

\begin{figure}[H]
    \centering
    \subfigure[Lucas-Kanade]{
        \includegraphics[width=0.48\textwidth]{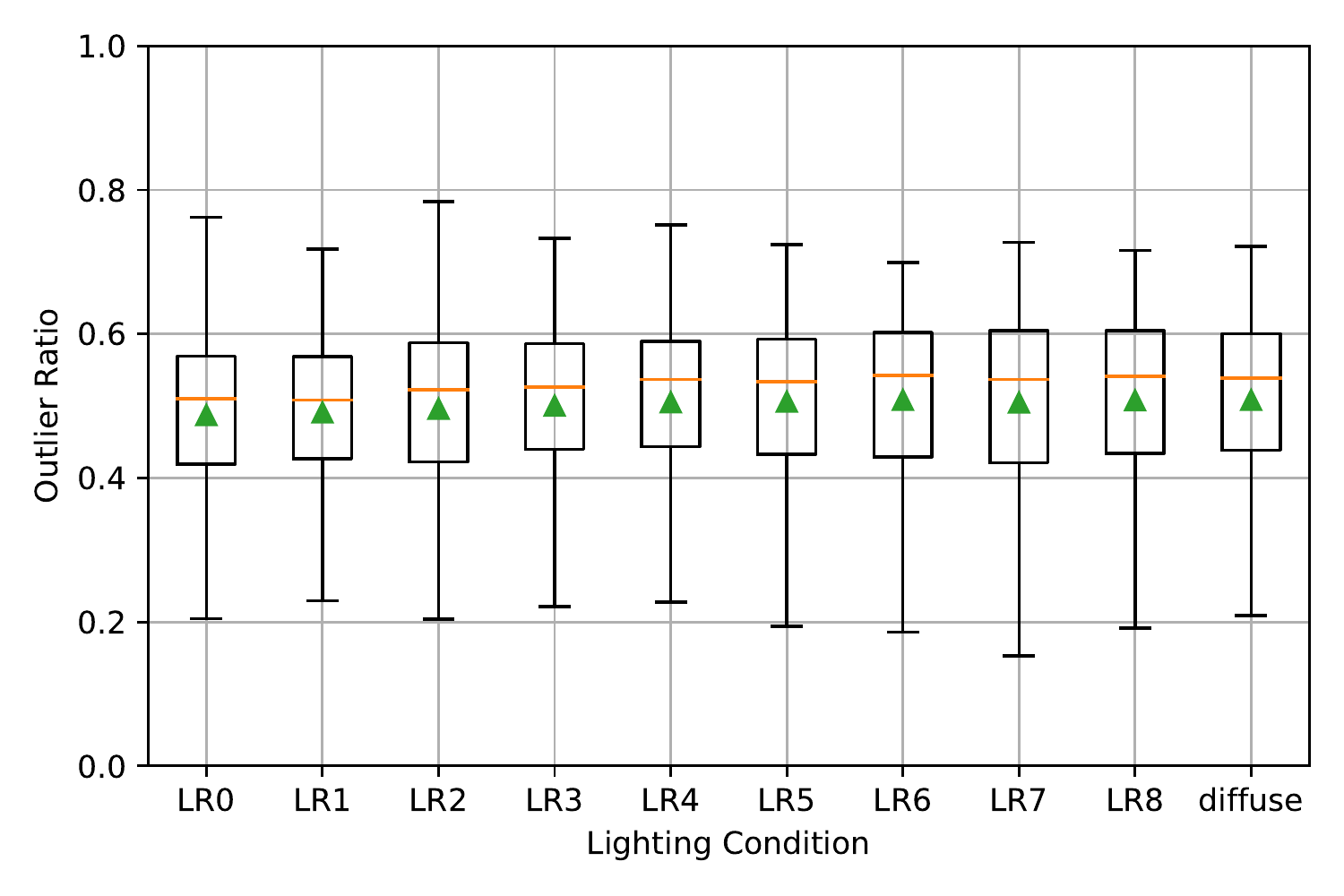}
        \includegraphics[width=0.48\textwidth]{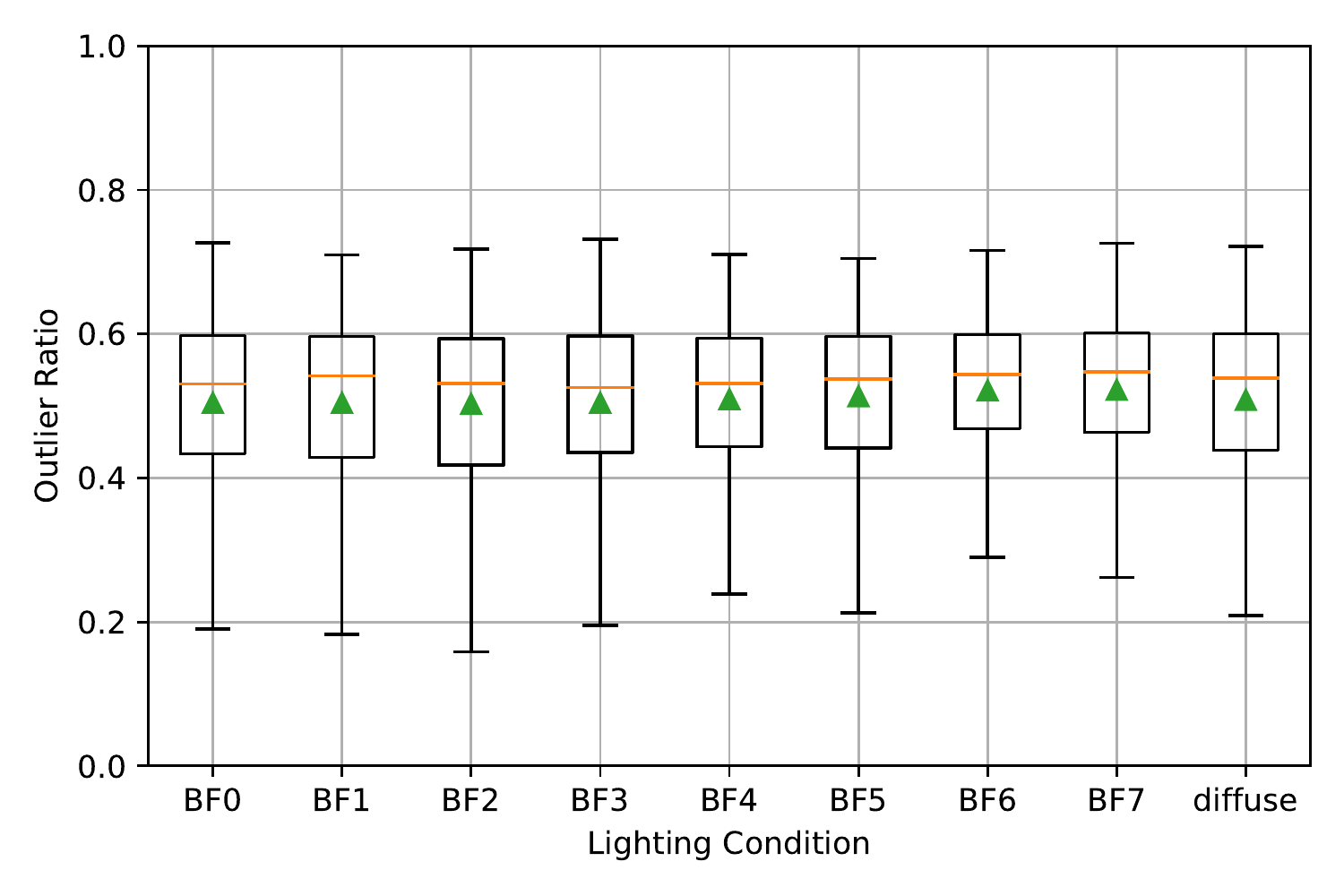}
    }
    \subfigure[Correspondence]{
        \includegraphics[width=0.48\textwidth]{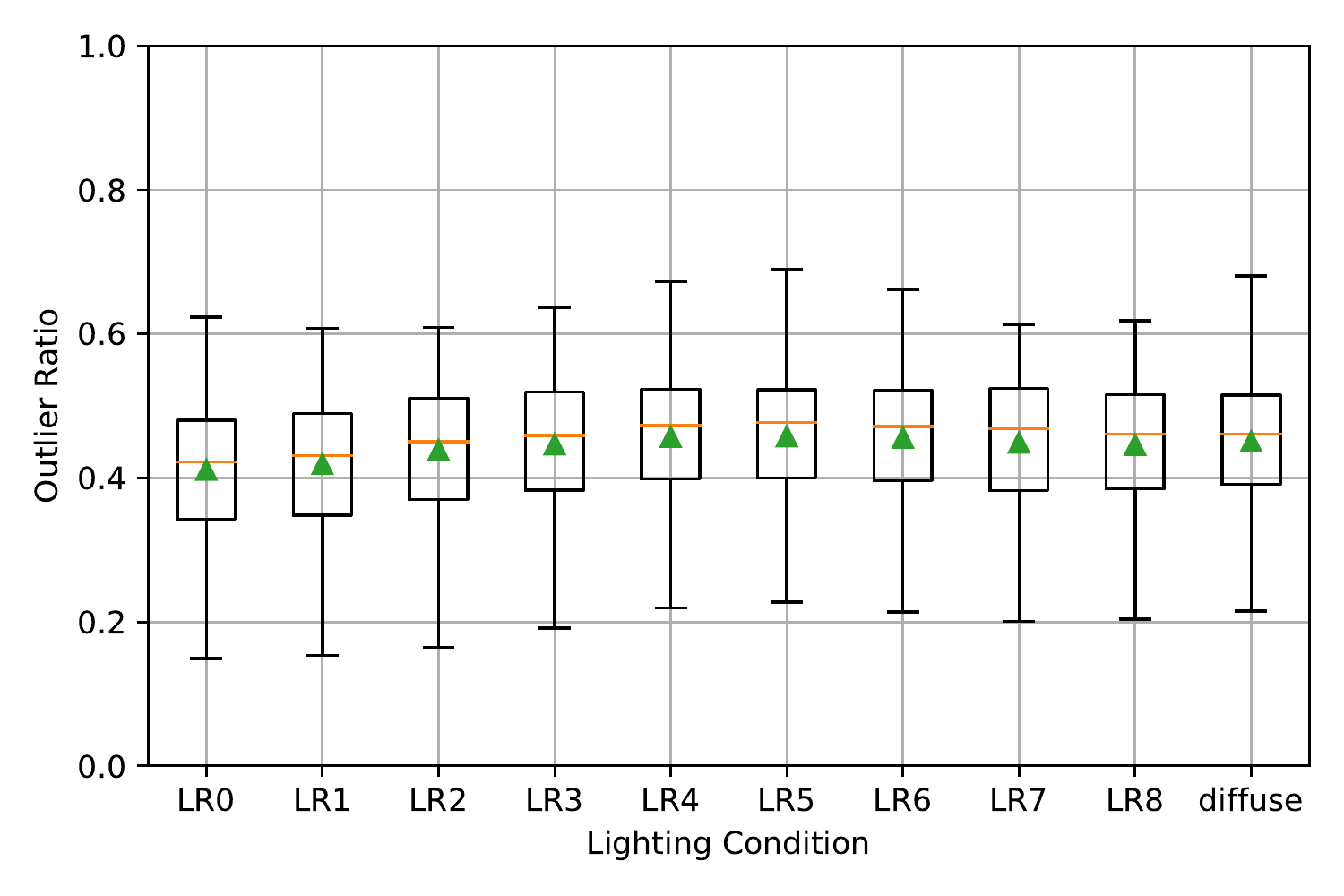}
        \includegraphics[width=0.48\textwidth]{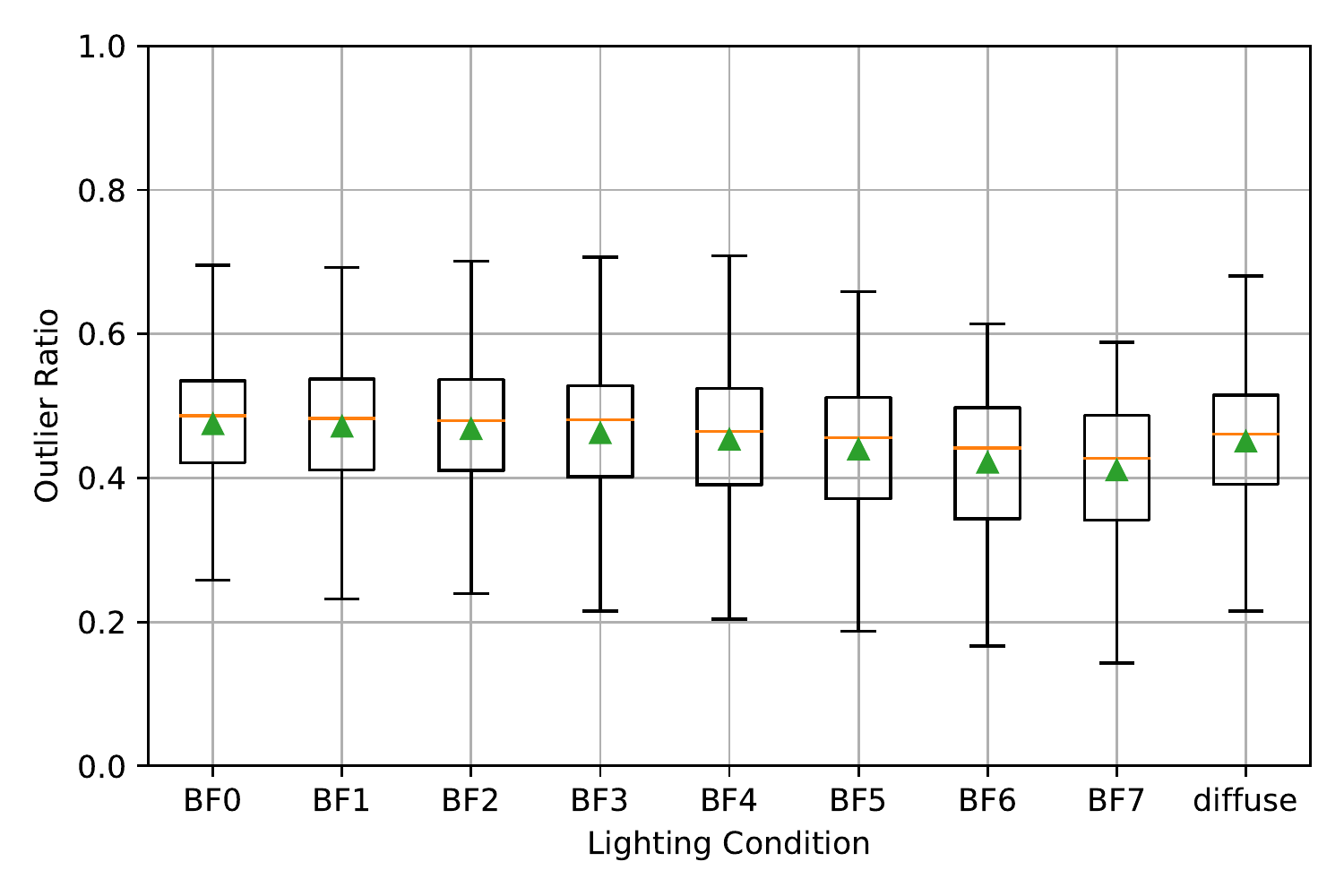}
    }
    \caption{\textbf{At eight times nominal speed, the existence of directional lighting does not affect outlier ratio.} In the box-and-whisker plots above, the orange line is the median, the green triangle is the mean, and the box extends from the first to the third quartiles. The whiskers extend up to 1.5x the length of the boxes. Each box-and-whisker plot is computed using features from all 60 scenes, one tracker, speed=8.00, and one of the lighting conditions in Figure \ref{fig:dtu_light_stage}. The distribution of outlier ratio is approximately the same for all lighting conditions.}
    \label{fig:dtu_speed8.00_percent_outlier}
\end{figure}

\begin{figure}[H]
    \centering
    \subfigure[Lucas-Kanade]{
        \includegraphics[width=0.48\textwidth]{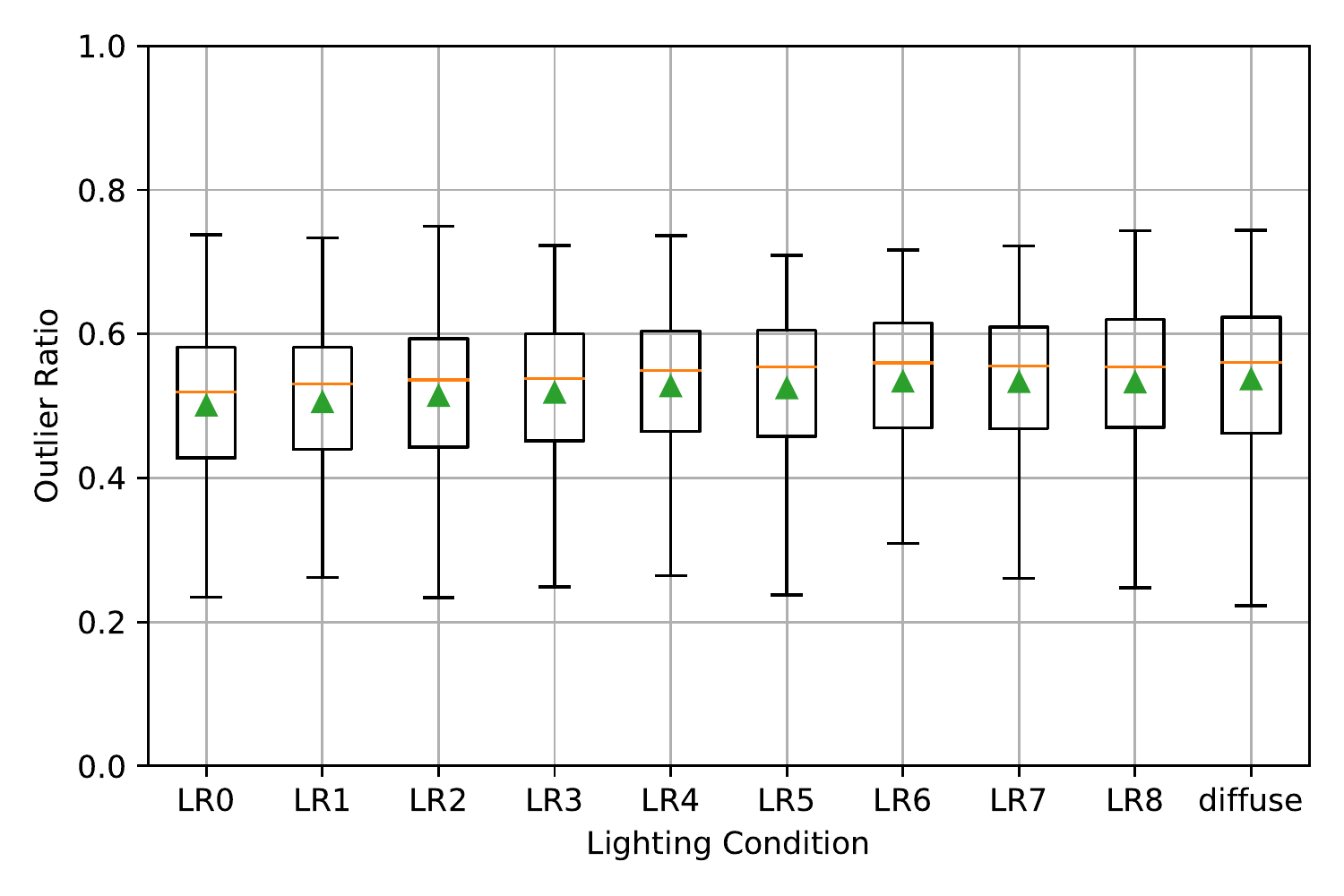}
        \includegraphics[width=0.48\textwidth]{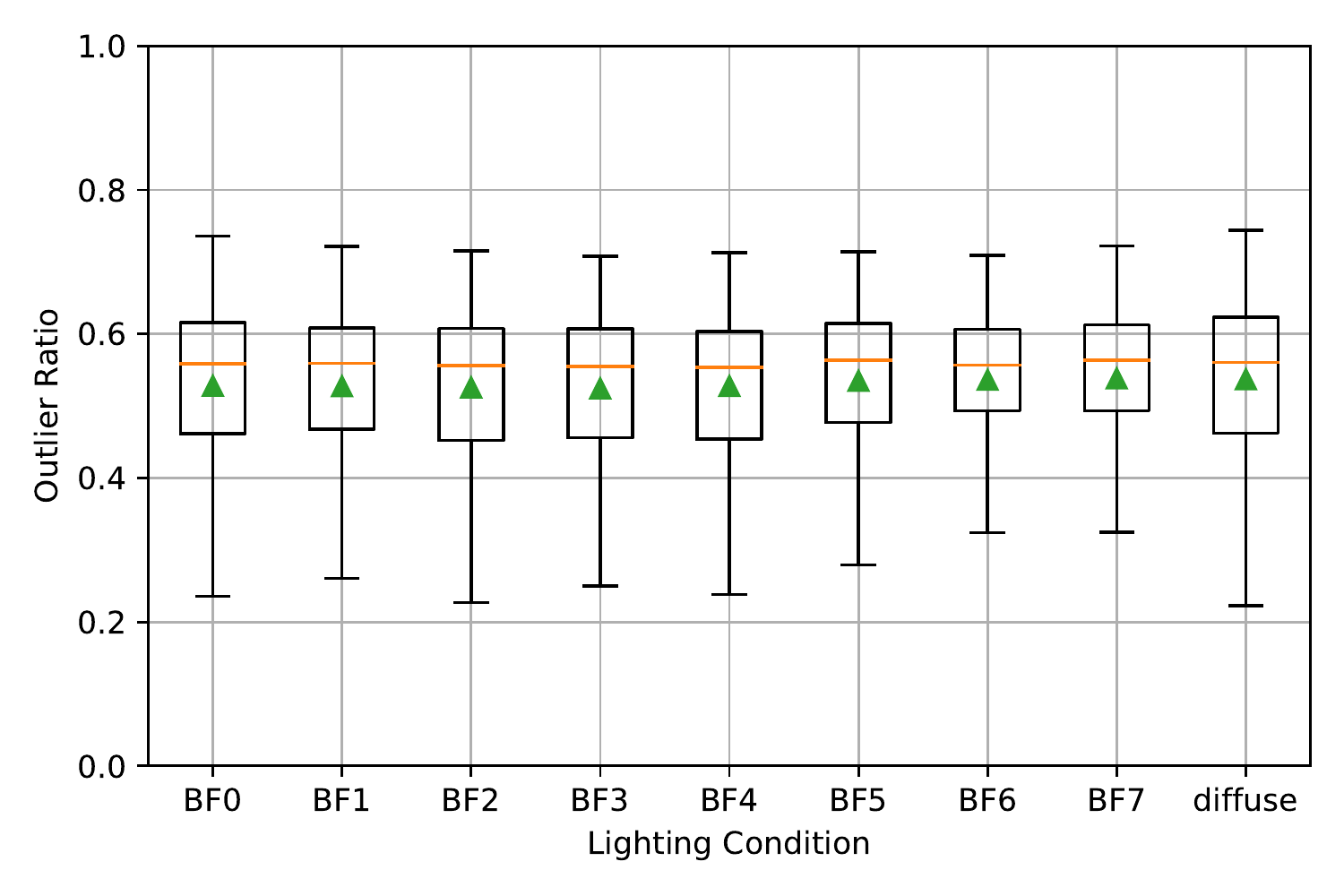}
    }
    \subfigure[Correspondence]{
        \includegraphics[width=0.48\textwidth]{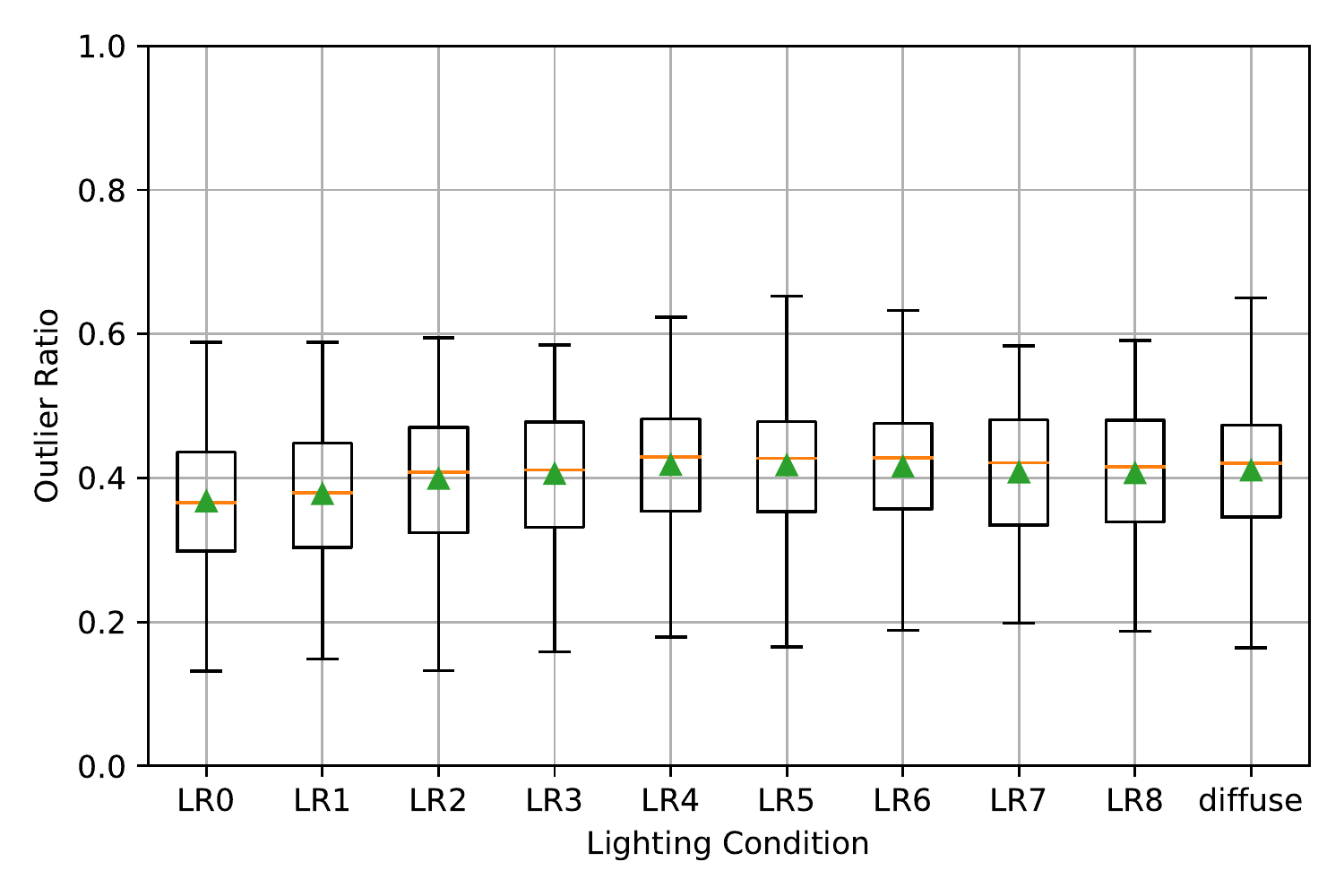}
        \includegraphics[width=0.48\textwidth]{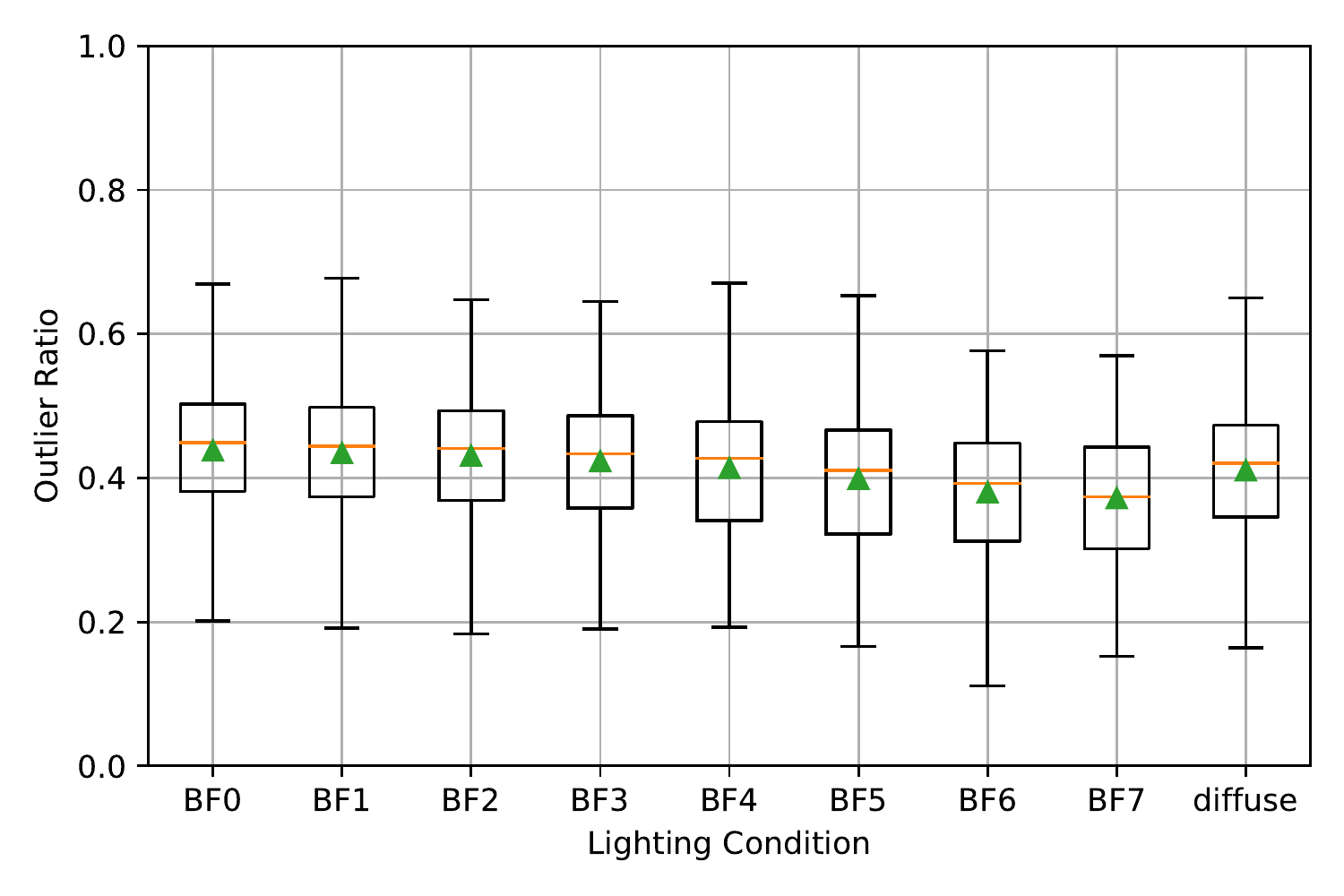}
    }
    \caption{\textbf{At twelve times nominal speed, the existence of directional lighting does not affect outlier ratio.} In the box-and-whisker plots above, the orange line is the median, the green triangle is the mean, and the box extends from the first to the third quartiles. The whiskers extend up to 1.5x the length of the boxes. Each box-and-whisker plot is computed using features from all 60 scenes, one tracker, speed=12.00, and one of the lighting conditions in Figure \ref{fig:dtu_light_stage}. The distribution of outlier ratio is approximately the same for all lighting conditions.}
    \label{fig:dtu_speed12.00_percent_outlier}
\end{figure}

\begin{figure}[H]
    \centering
    \subfigure[Horizontal Coordinate]{
        \includegraphics[width=0.48\textwidth]{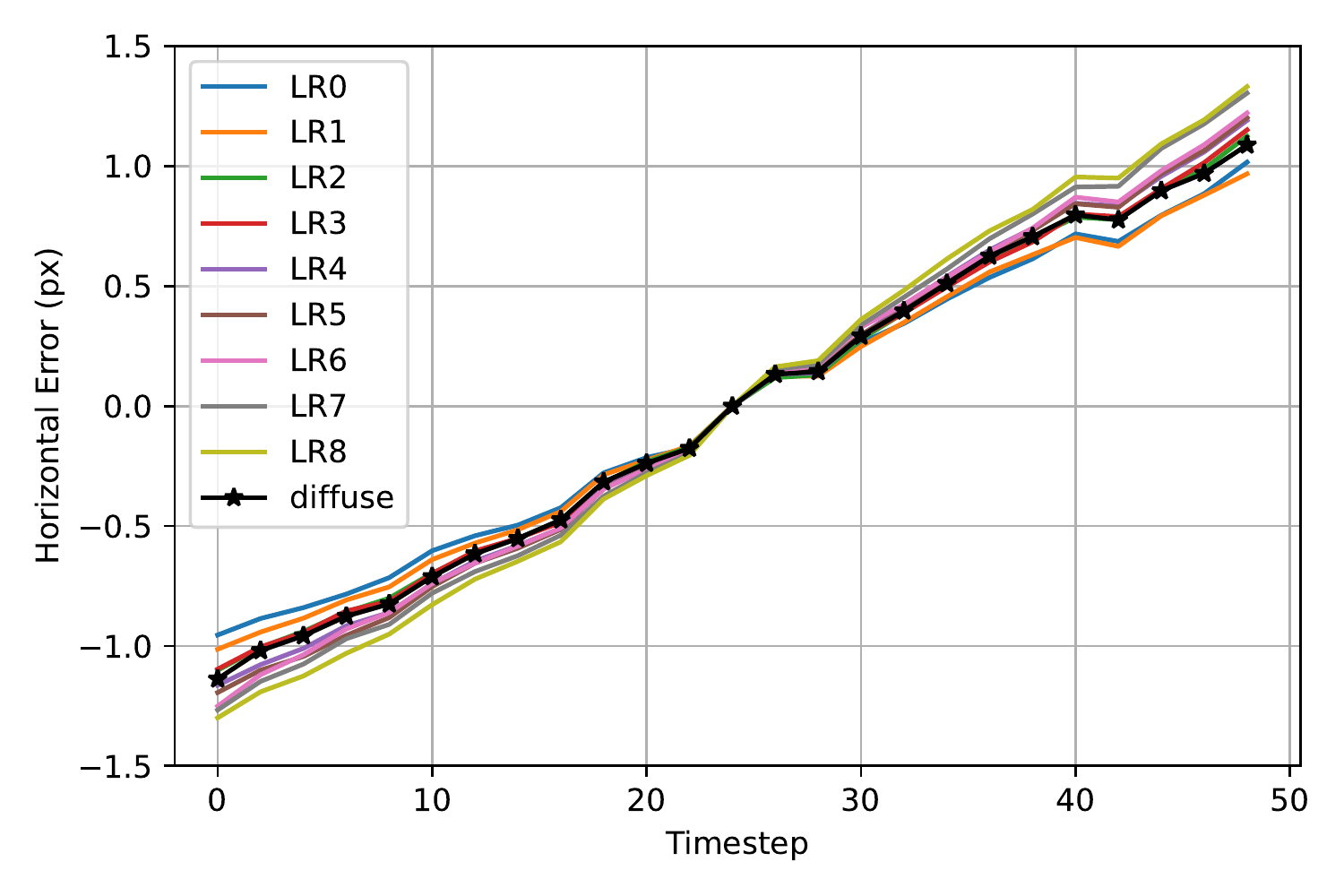}
        \includegraphics[width=0.48\textwidth]{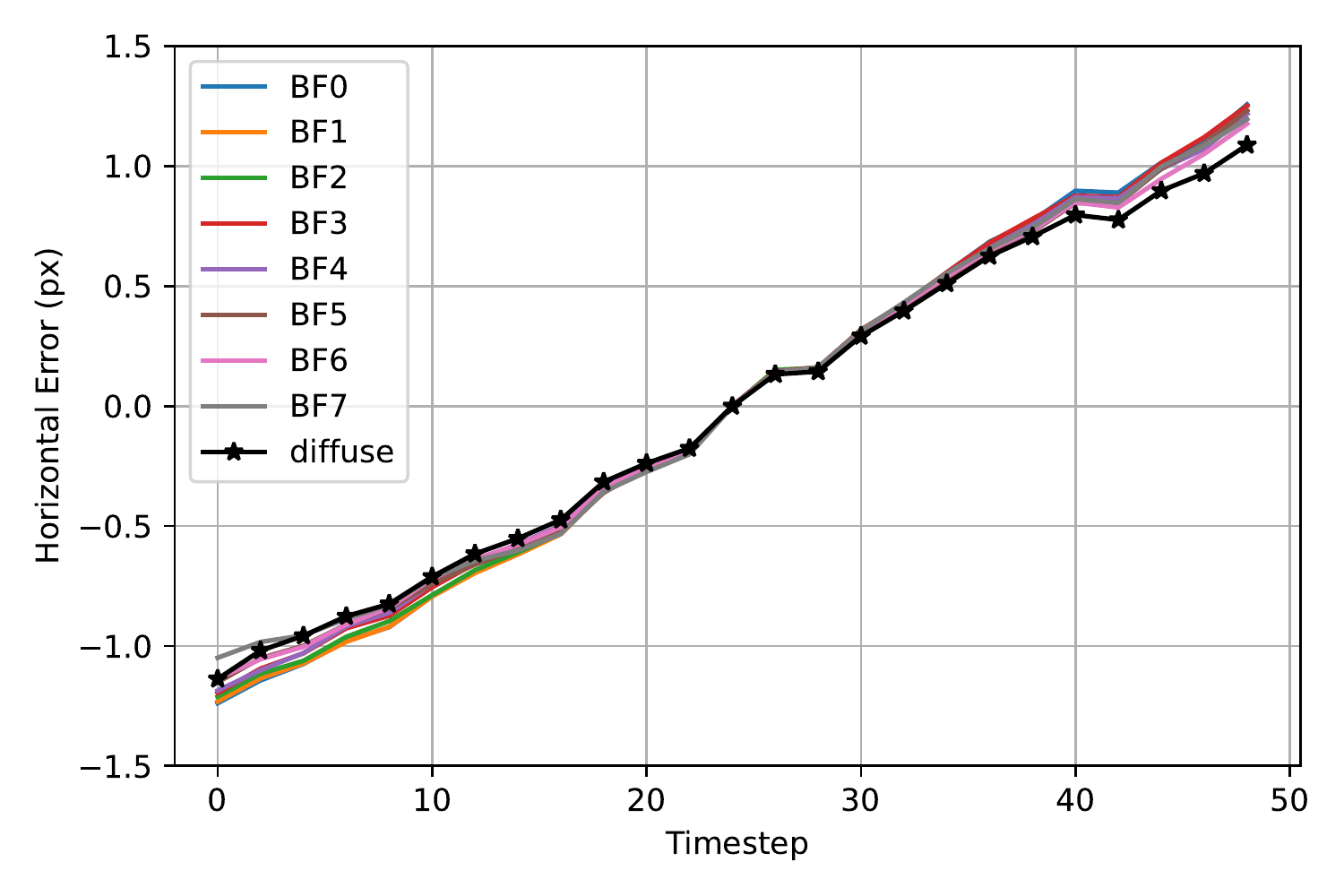}
    }
    \subfigure[Vertical Coordinate]{
        \includegraphics[width=0.48\textwidth]{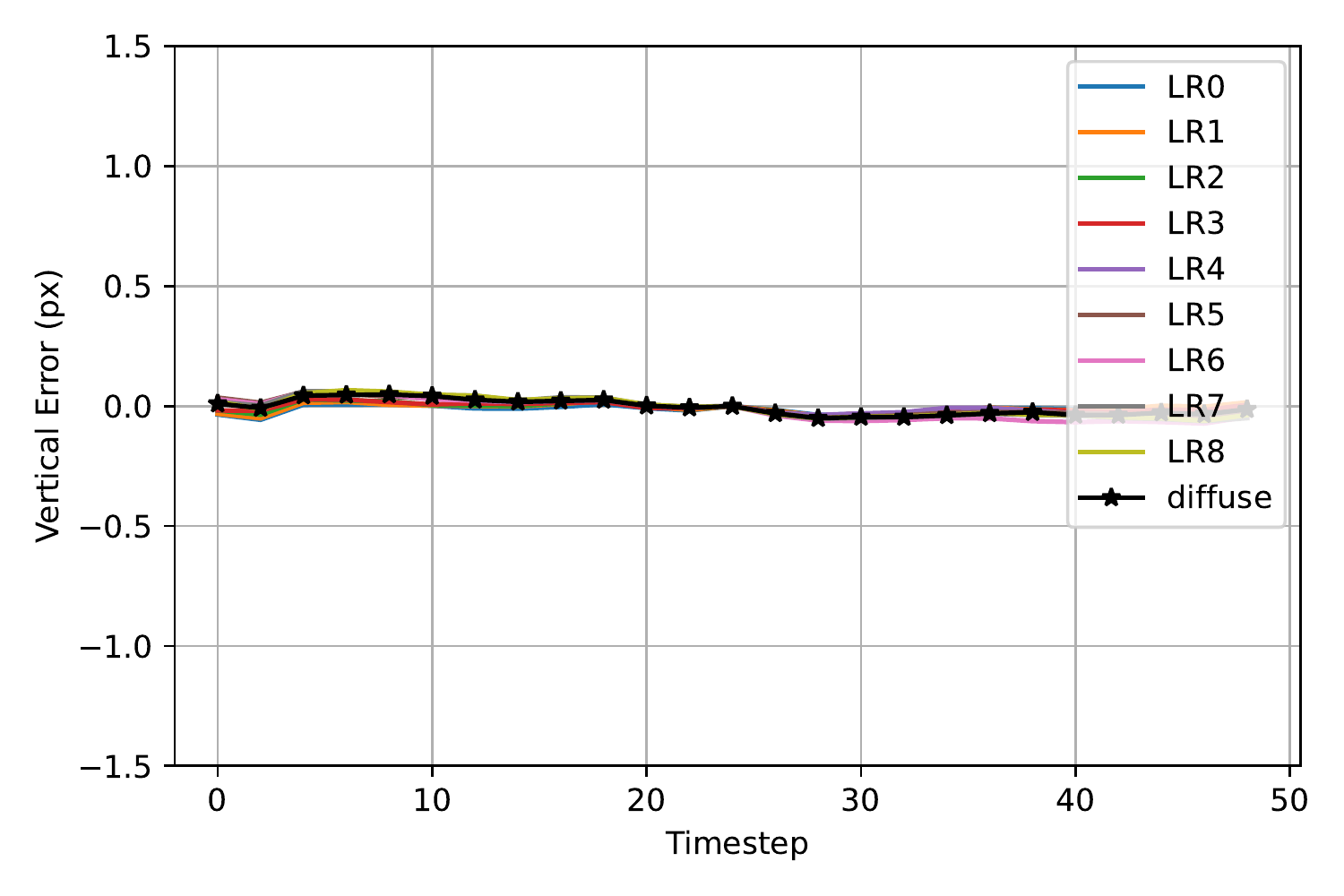}
        \includegraphics[width=0.48\textwidth]{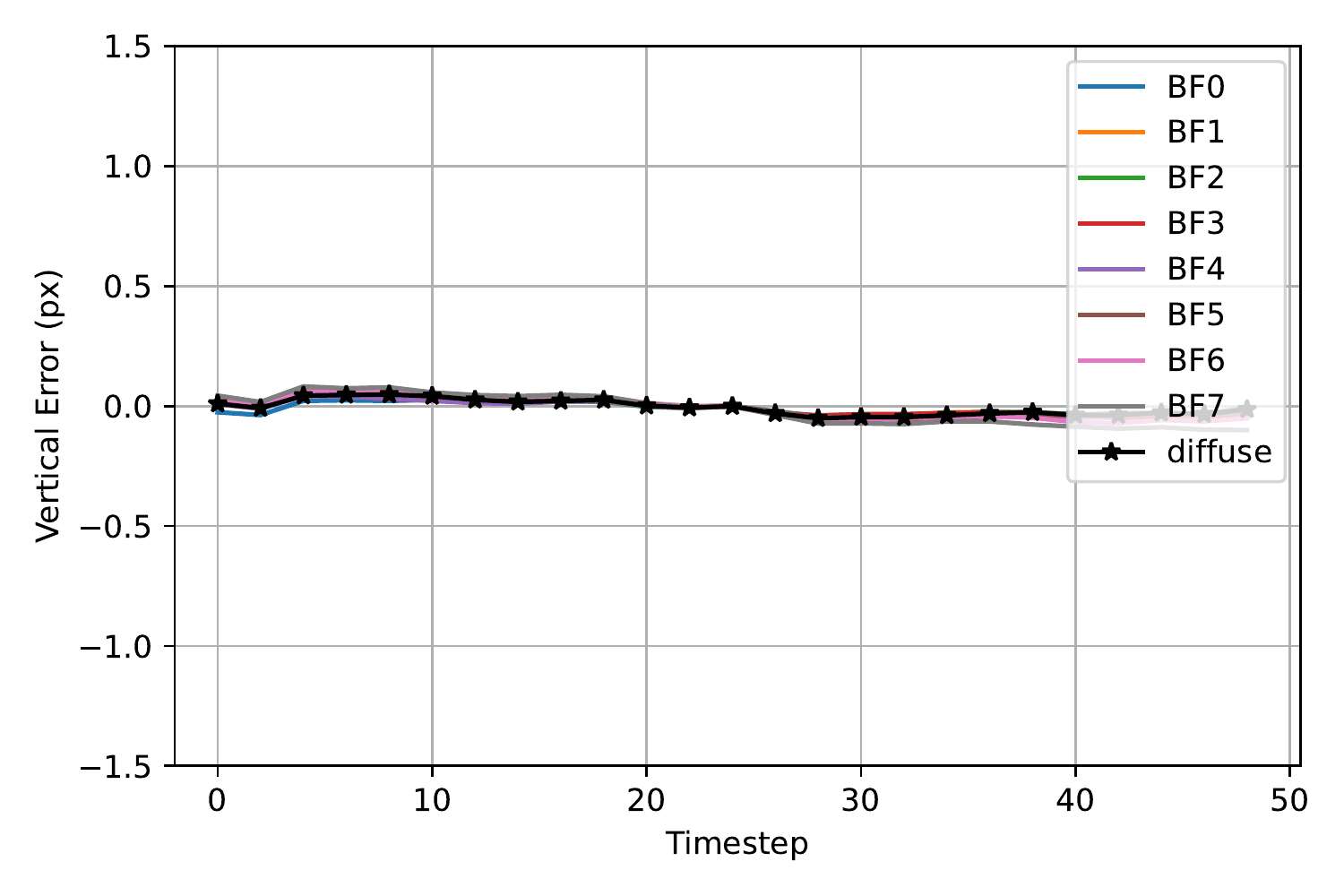}
    }
    \caption{\textbf{DTU Point Features Dataset: At twice nominal speed, lighting condition does not change trends in mean error $\mu(t)$  when using the Lucas-Kanade Tracker.} We compute $\mu(t)$ at each timestep using diffuse lighting (black lines) and each of the directional lighting conditions listed in Figure \ref{fig:dtu_light_stage} using all tracks from all 60 scenes.  The variation of $\mu(t)$ due to the existence of directional lighting is at most 10 percent of the variation common to all plotted lines. The effect of directional lighting is relatively small because changes between adjacent frames are small whether or not the scene contains directional lighting.}
    \label{dtu_LK_mu_speed2.00}
\end{figure}

\begin{figure}[H]
    \centering
    \subfigure[Horizontal Coordinate]{
        \includegraphics[width=0.48\textwidth]{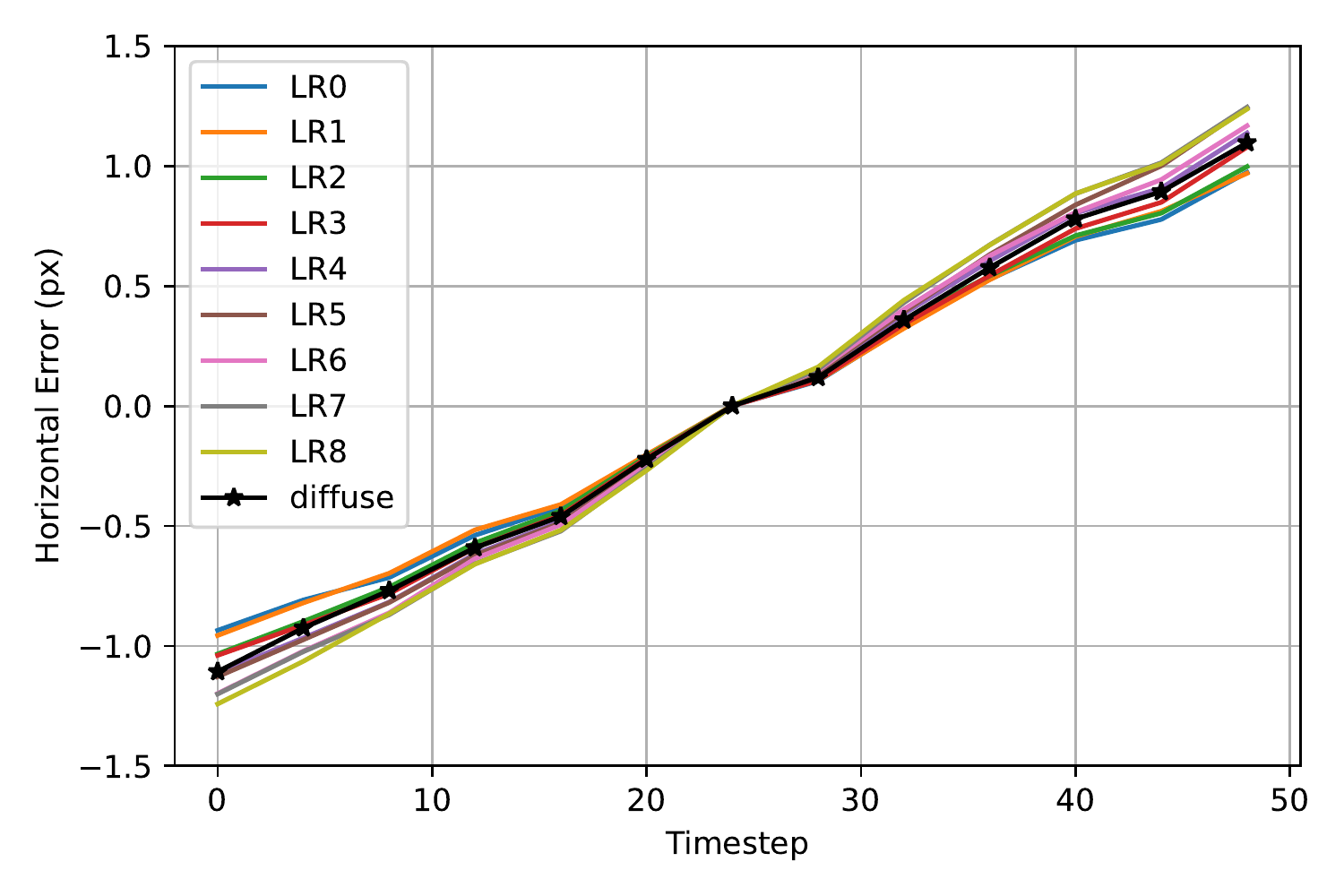}
        \includegraphics[width=0.48\textwidth]{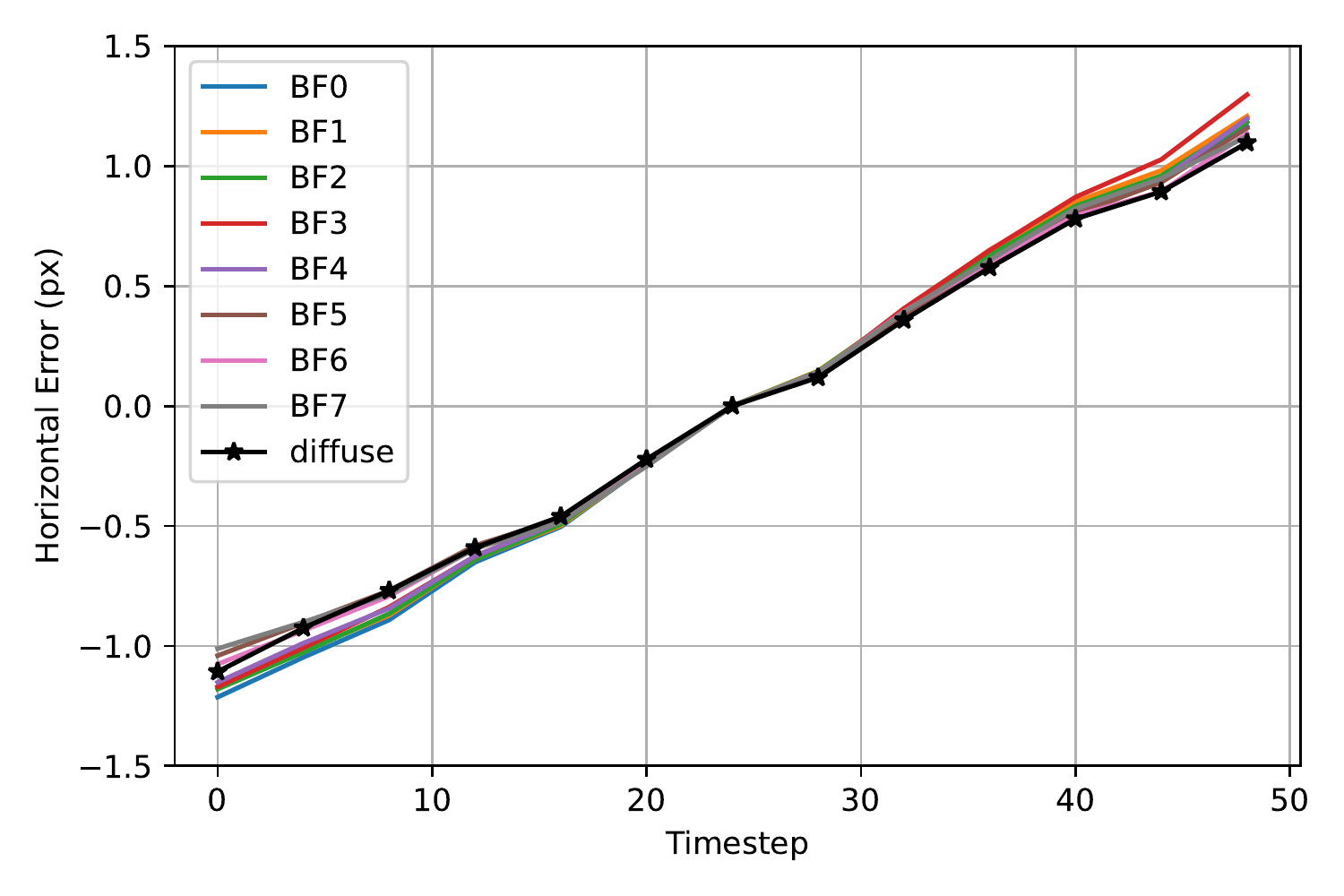}
    }
    \subfigure[Vertical Coordinate]{
        \includegraphics[width=0.48\textwidth]{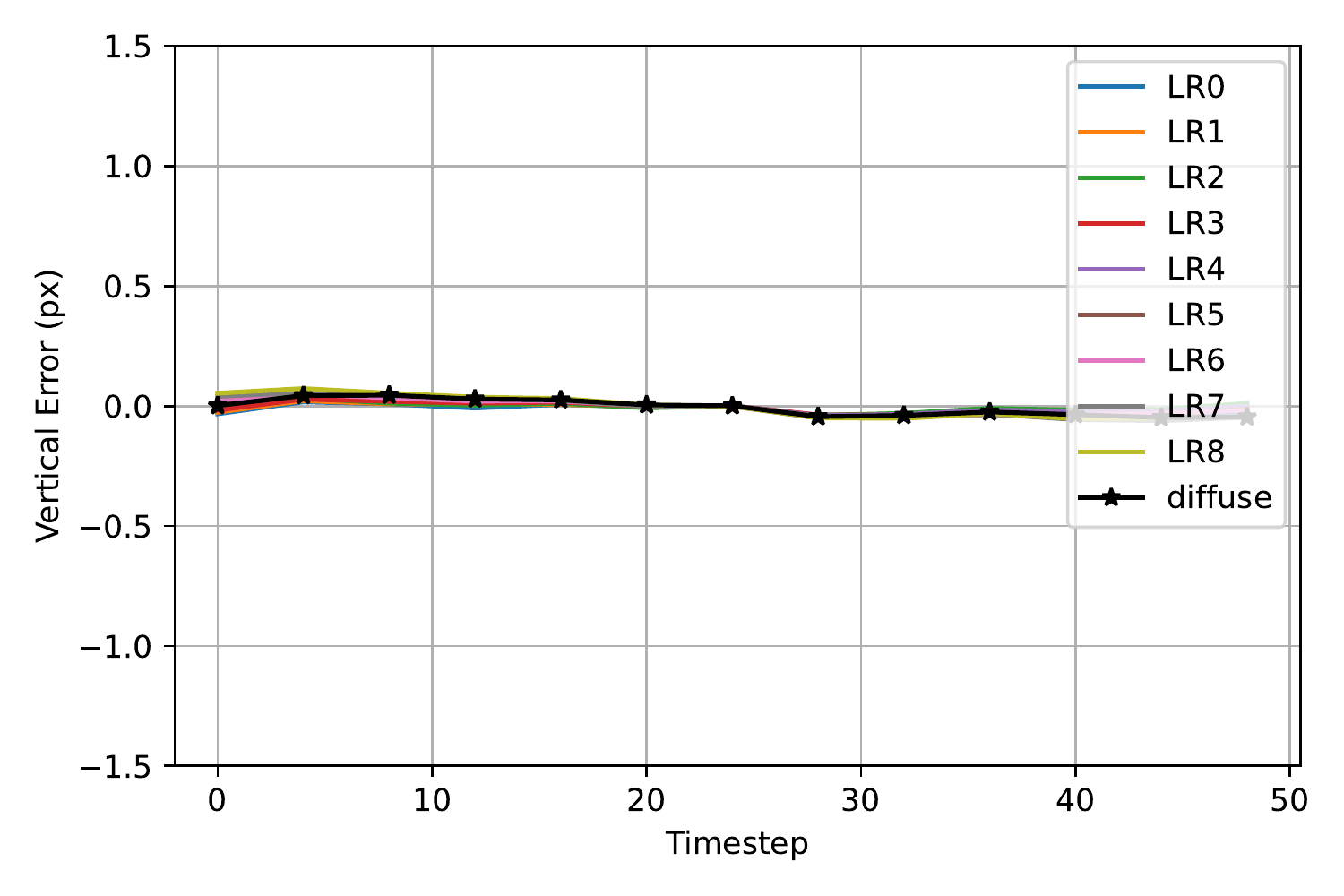}
        \includegraphics[width=0.48\textwidth]{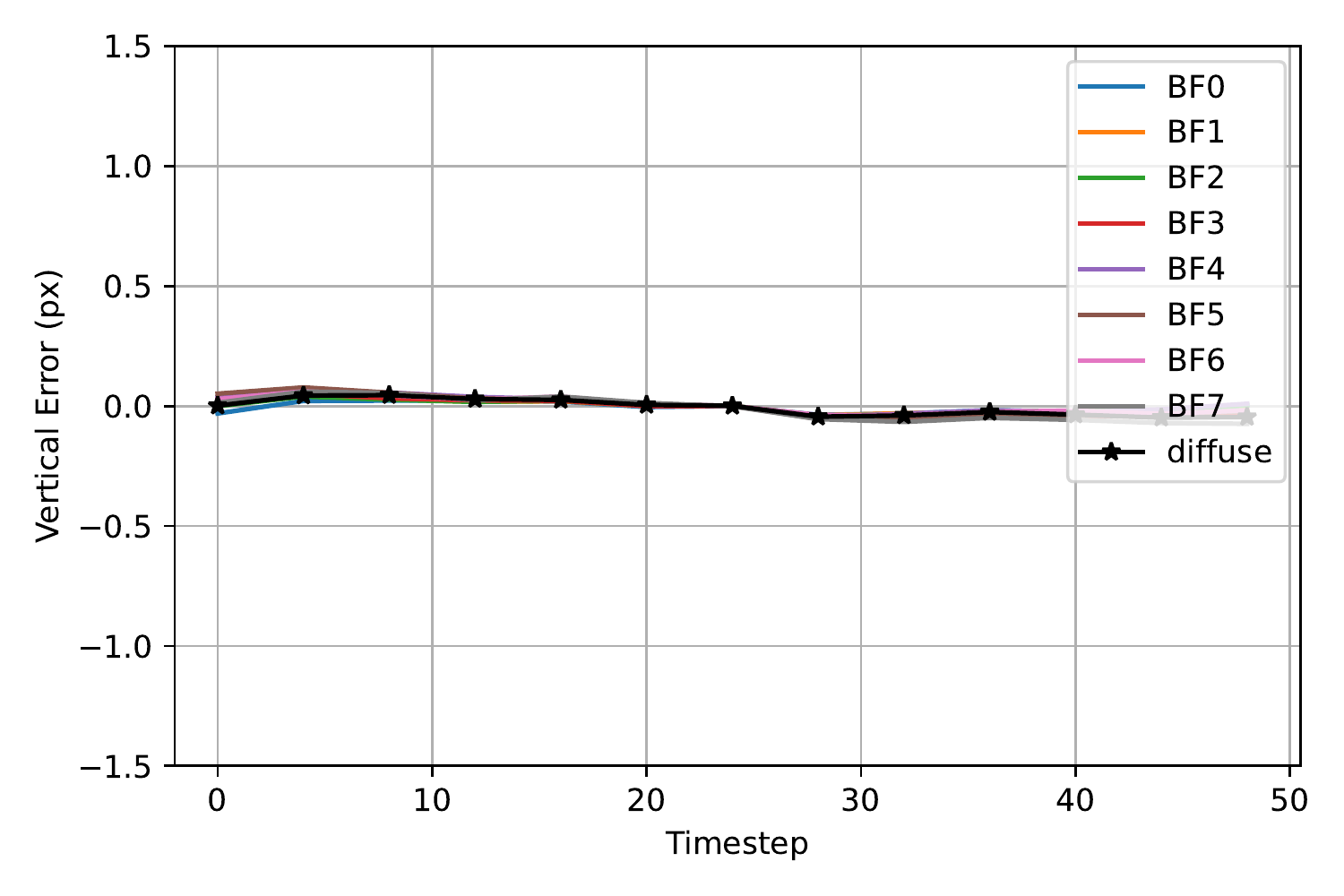}
    }
    \caption{\textbf{DTU Point Features Dataset: At four times nominal speed, lighting condition does not change trends in mean error $\mu(t)$ when using the Lucas-Kanade Tracker.} We compute $\mu(t)$ at each timestep using diffuse lighting (black lines) and each of the directional lighting conditions listed in Figure \ref{fig:dtu_light_stage} using all tracks from all 60 scenes. The variation of $\mu(t)$ due to the existence of directional lighting is at most 10 percent of the variation common to all plotted lines. The effect of directional lighting is relatively small because changes between adjacent frames are small whether or not the scene contains directional lighting.}
    \label{dtu_LK_mu_speed4.00}
\end{figure}

\begin{figure}[H]
    \centering
    \subfigure[Horizontal Coordinate]{
        \includegraphics[width=0.48\textwidth]{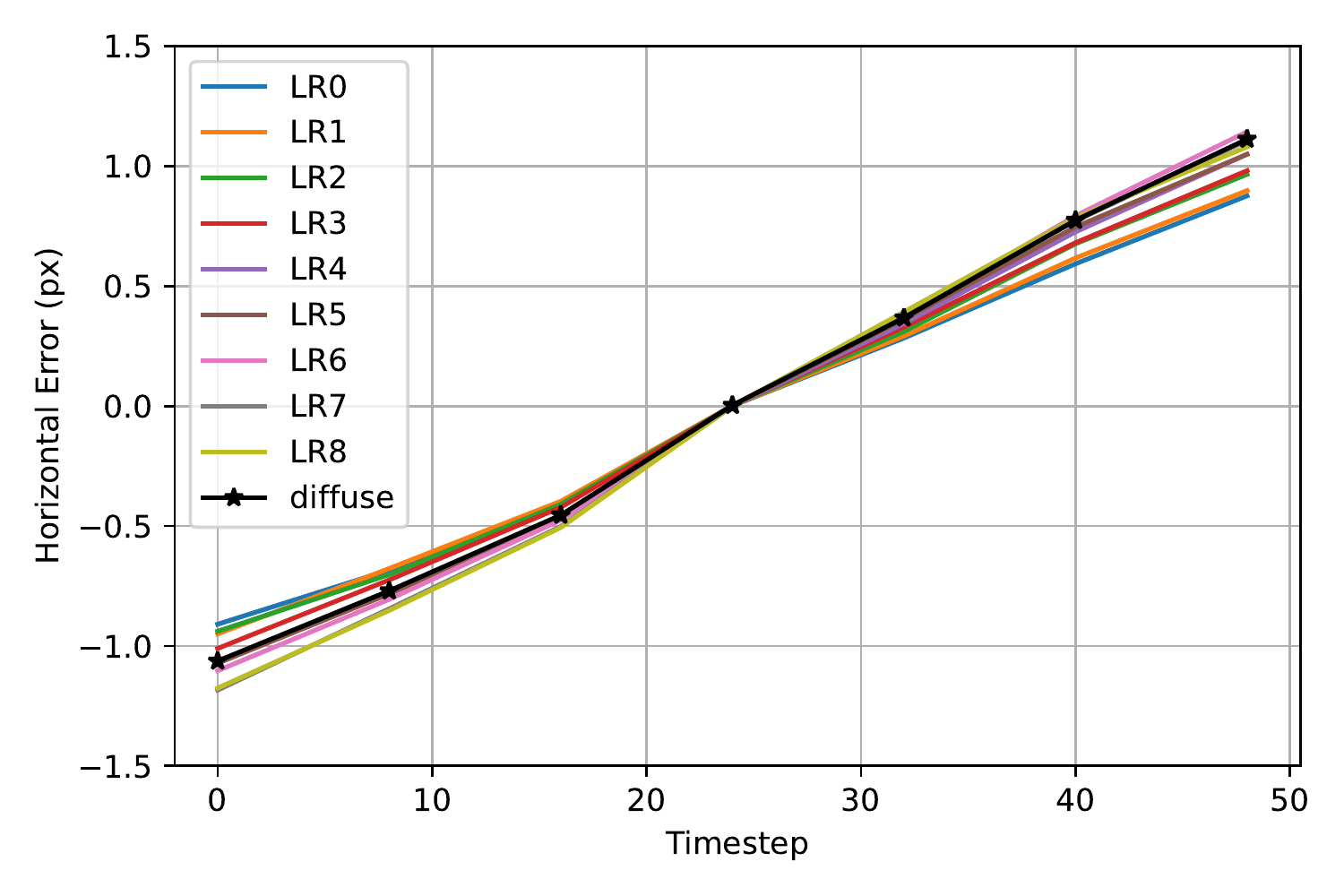}
        \includegraphics[width=0.48\textwidth]{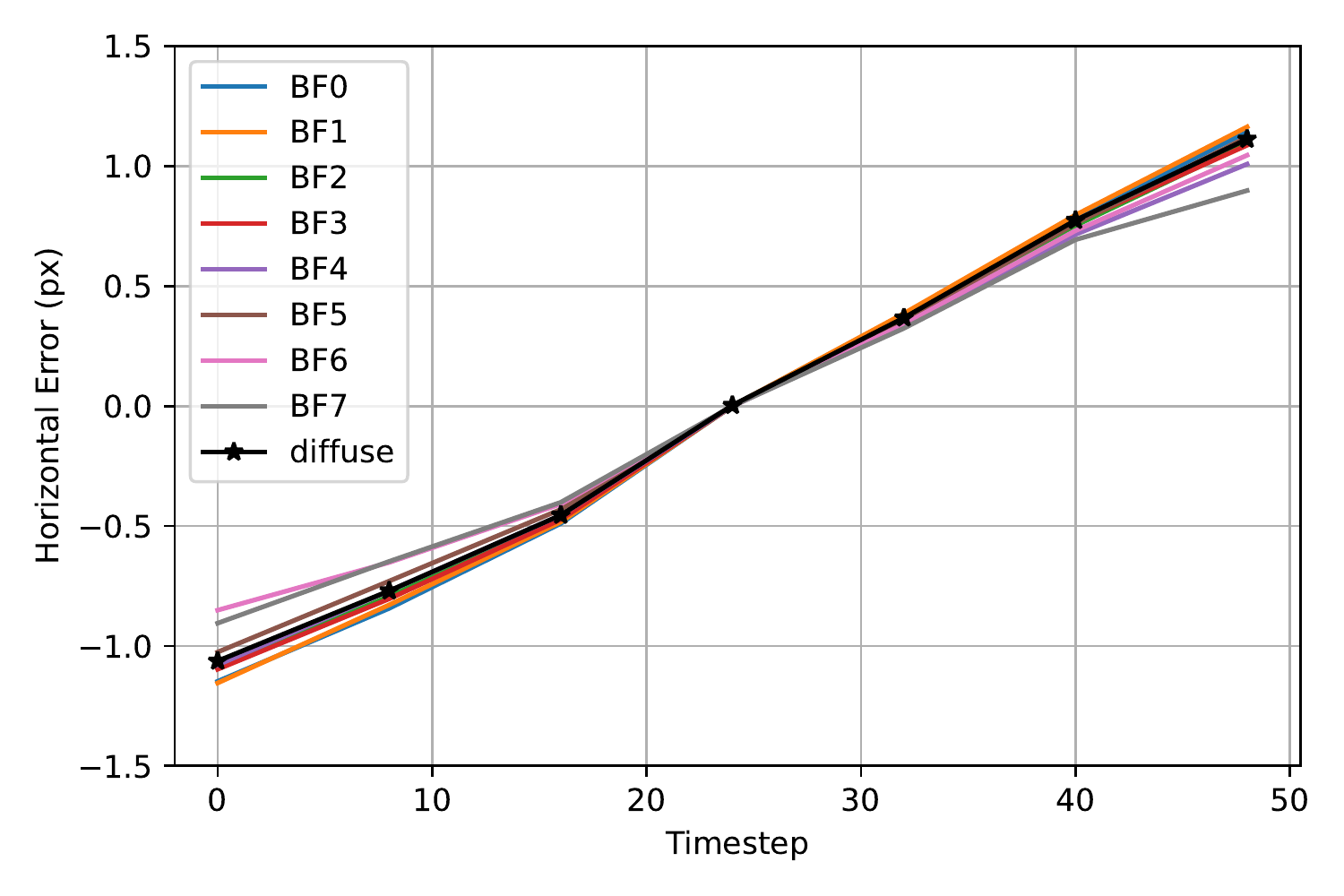}
    }
    \subfigure[Vertical Coordinate]{
        \includegraphics[width=0.48\textwidth]{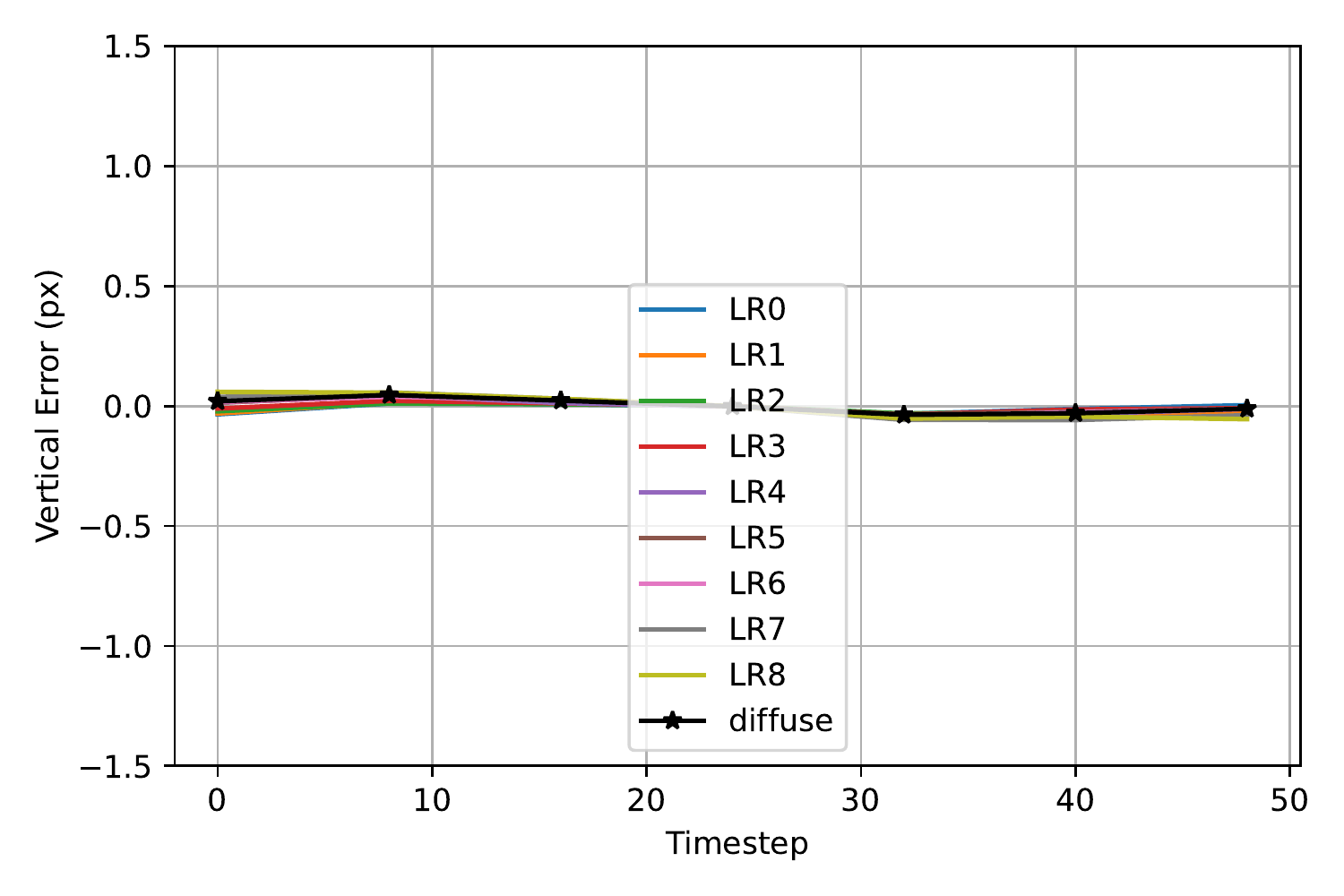}
        \includegraphics[width=0.48\textwidth]{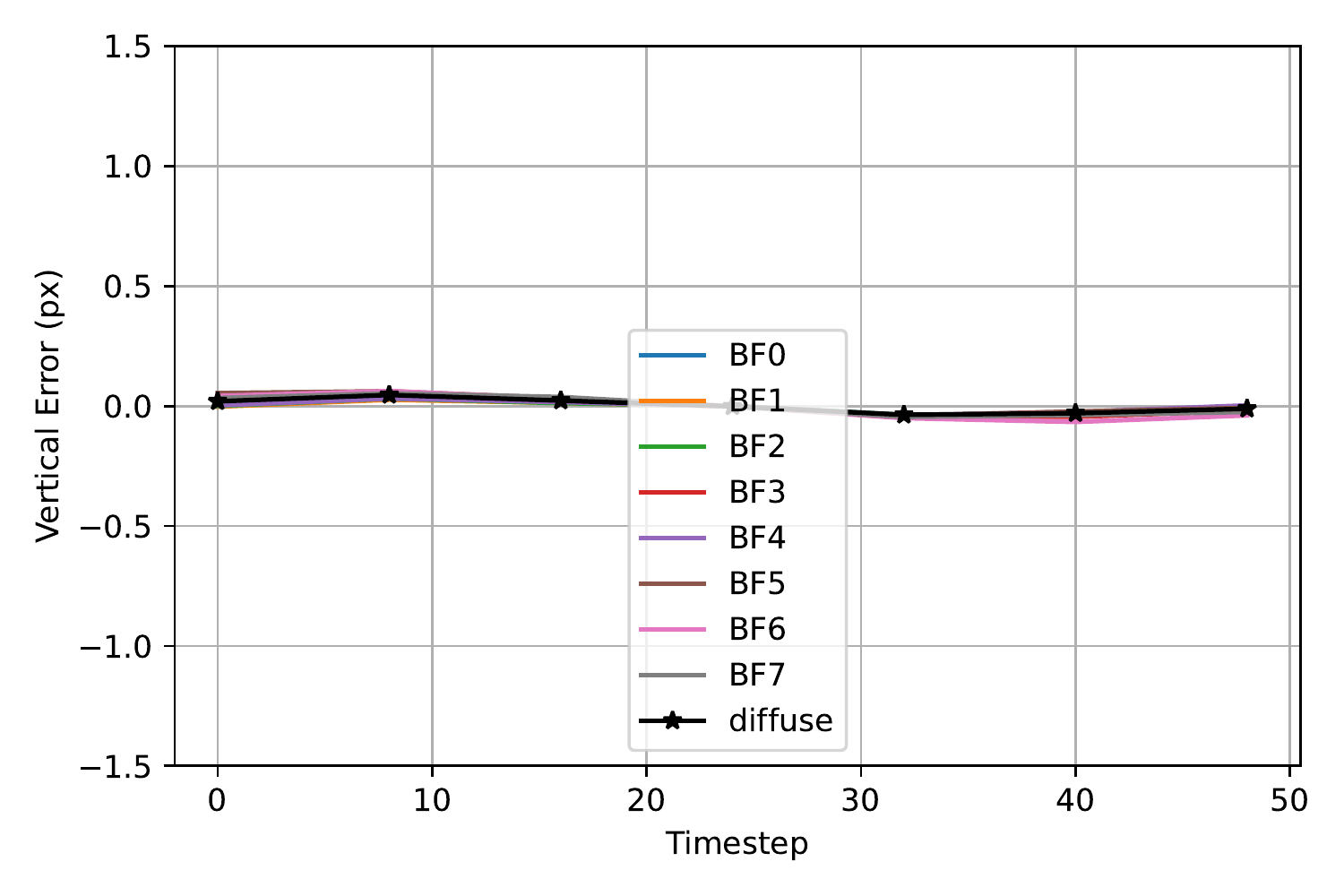}
    }
    \caption{\textbf{DTU Point Features Dataset: At eight times nominal speed, lighting condition does not change trends in mean error $\mu(t)$ when using the Lucas-Kanade Tracker.} We compute $\mu(t)$ at each timestep using diffuse lighting (black lines) and each of the directional lighting conditions listed in Figure \ref{fig:dtu_light_stage} using all tracks from all 60 scenes. The variation of $\mu(t)$ due to the existence of directional lighting is at most 10 percent of the variation common to all plotted lines. The effect of directional lighting is relatively small because changes between adjacent frames are small whether or not the scene contains directional lighting.}
    \label{dtu_LK_mu_speed8.00}
\end{figure}

\begin{figure}[H]
    \centering
    \subfigure[Horizontal Coordinate]{
        \includegraphics[width=0.48\textwidth]{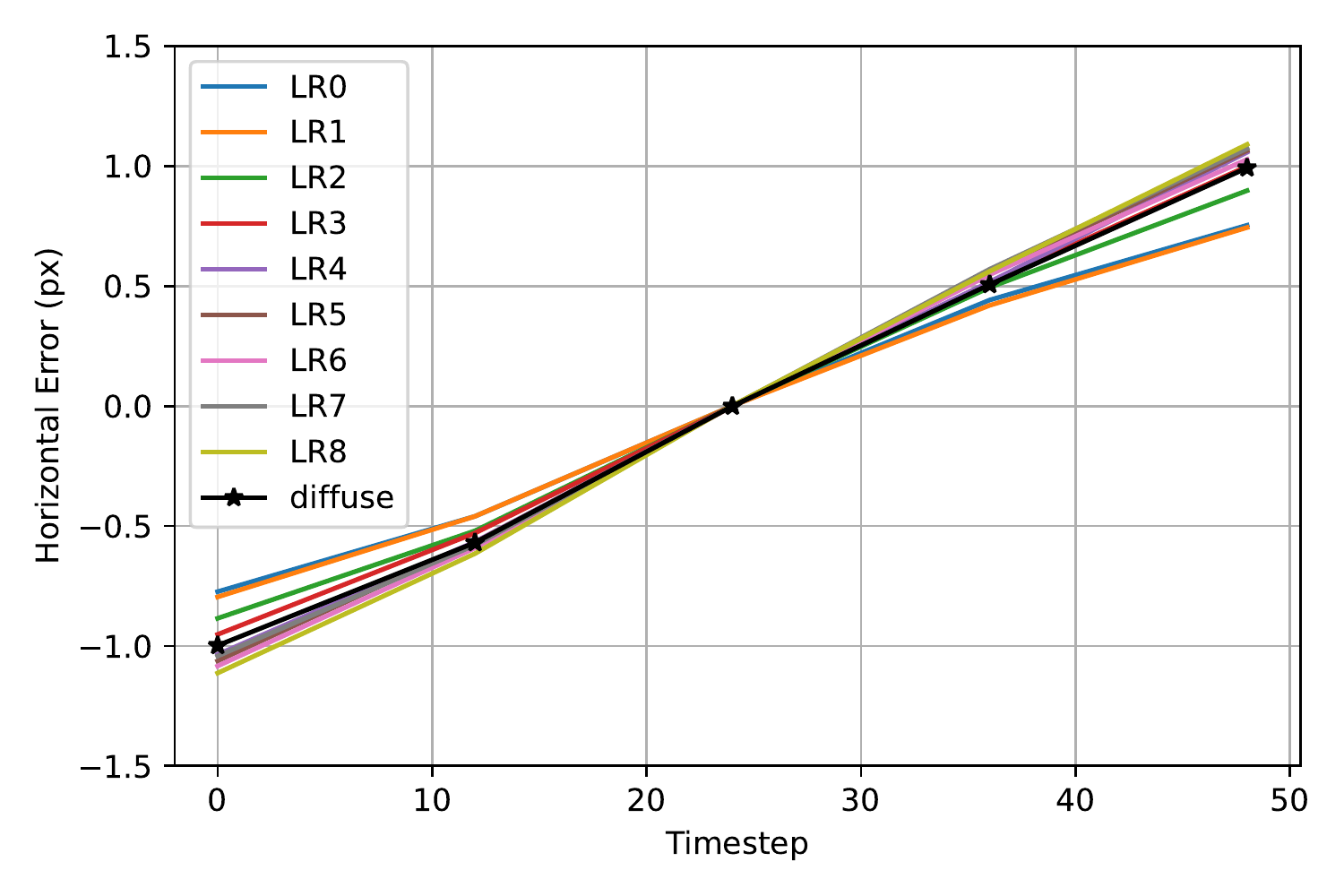}
        \includegraphics[width=0.48\textwidth]{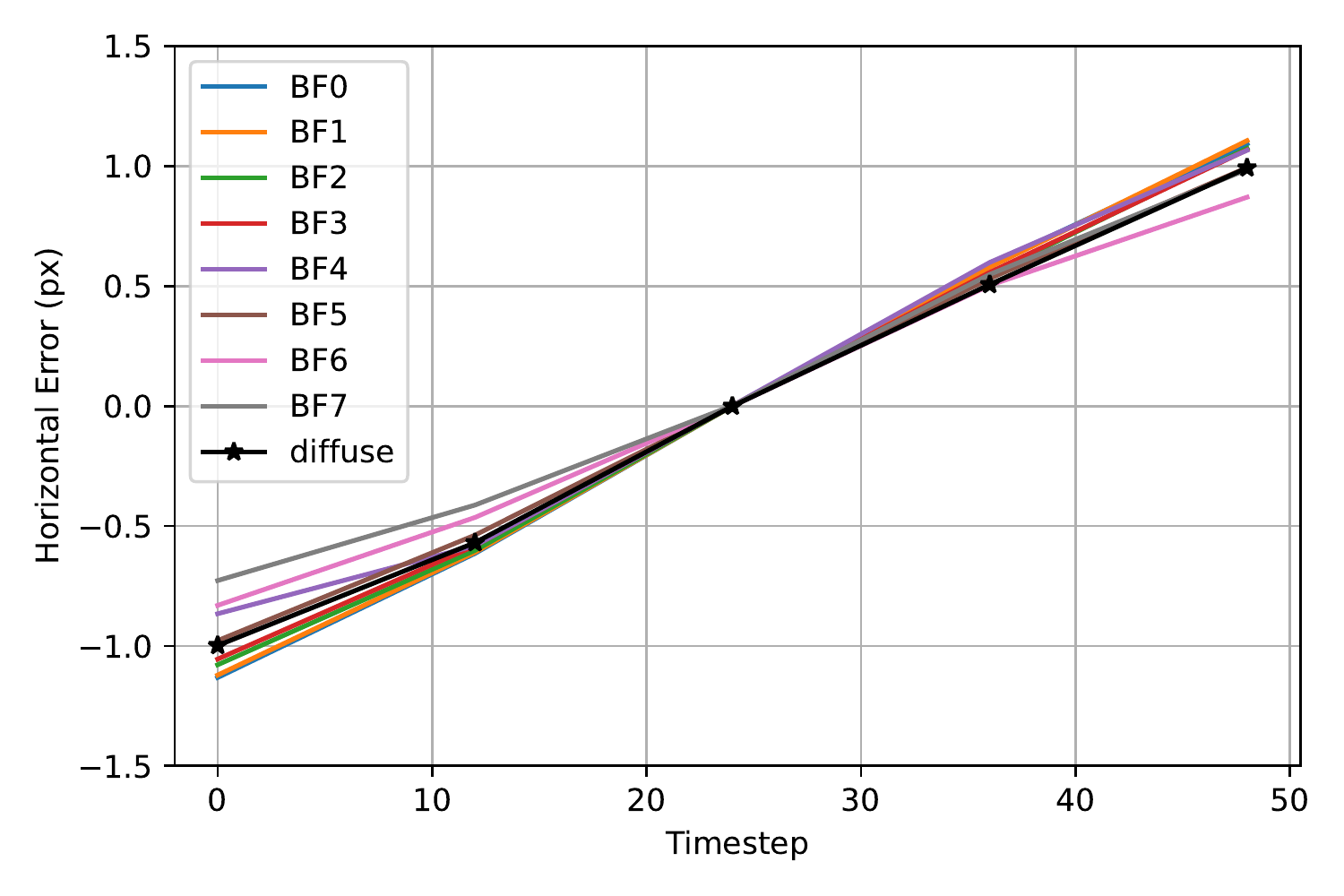}
    }
    \subfigure[Vertical Coordinate]{
        \includegraphics[width=0.48\textwidth]{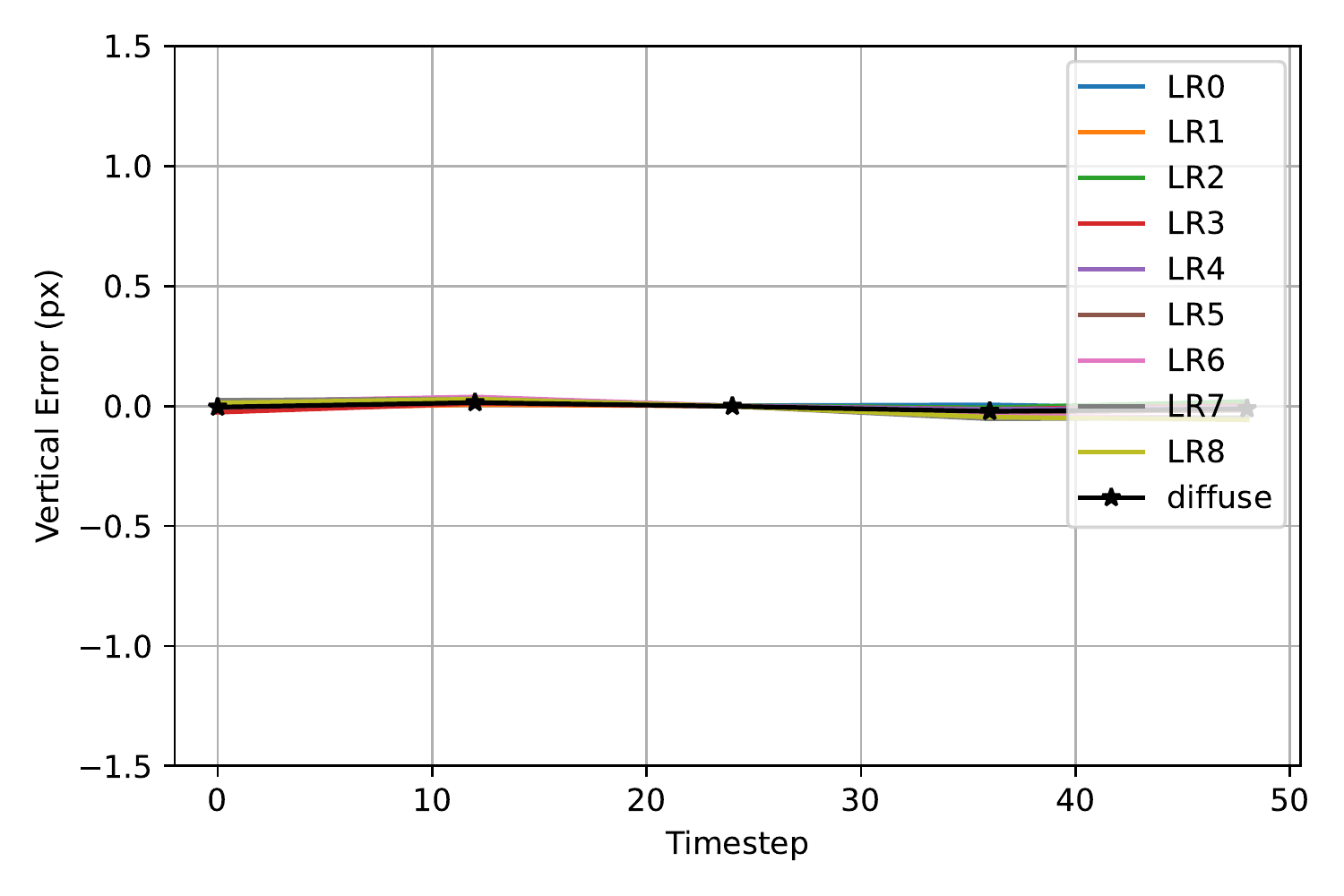}
        \includegraphics[width=0.48\textwidth]{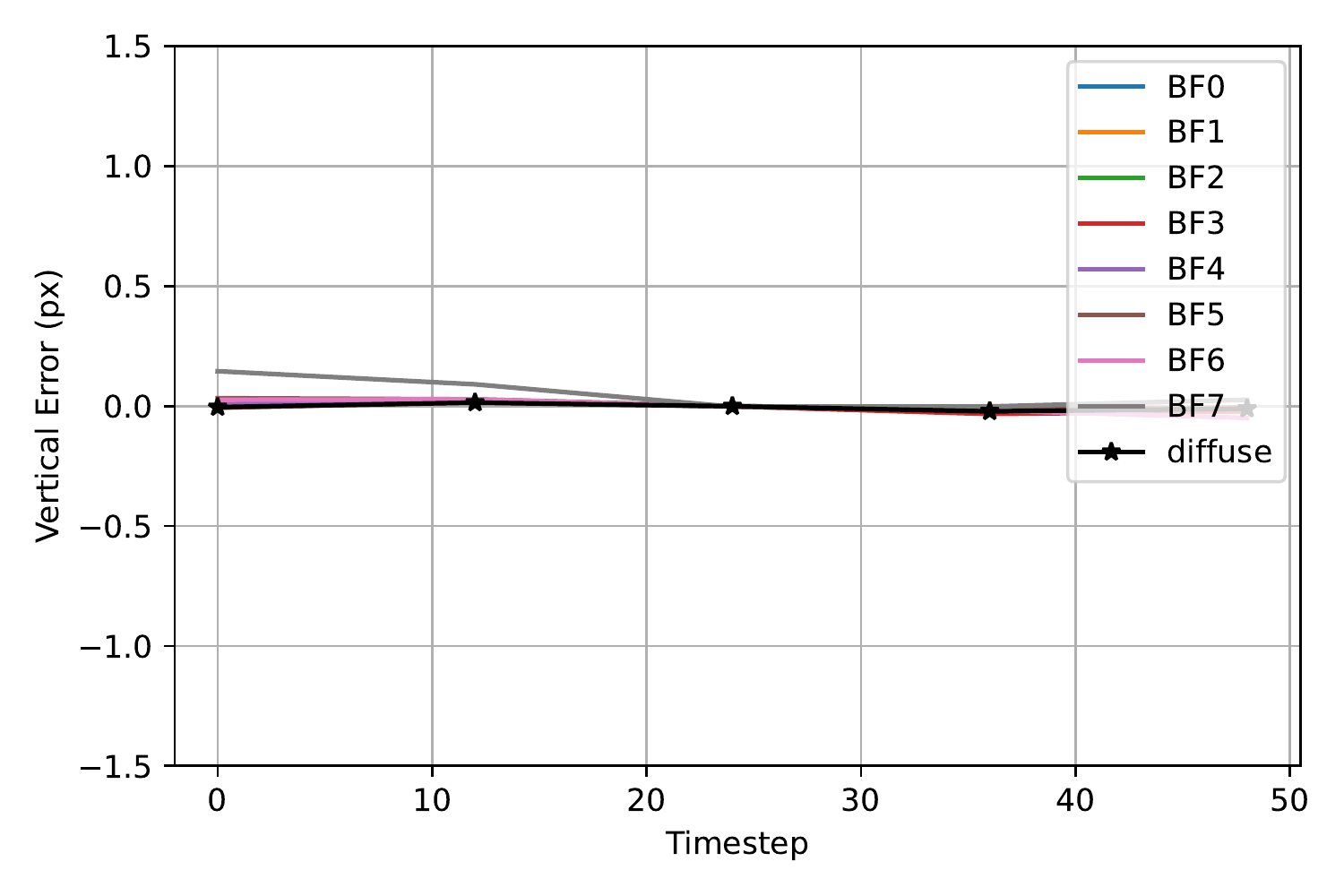}
    } 
    \caption{\textbf{DTU Point Features Dataset: At twelve times nominal speed, lighting condition does not change trends in mean error $\mu(t)$  when using the Lucas-Kanade Tracker.} We compute $\mu(t)$ at each timestep using diffuse lighting (black lines) and each of the directional lighting conditions listed in Figure \ref{fig:dtu_light_stage} using all tracks from all 60 scenes. The variation of $\mu(t)$ due to the existence of directional lighting is at most 10 percent of the variation common to all plotted lines. The effect of directional lighting is relatively small because changes between adjacent frames are small whether or not the scene contains directional lighting.}
    \label{fig:dtu_LK_mu_speed12.00}
\end{figure}

\begin{figure}[H]
    \centering
    \subfigure[Horizontal Coordinate]{
        \includegraphics[width=0.48\textwidth]{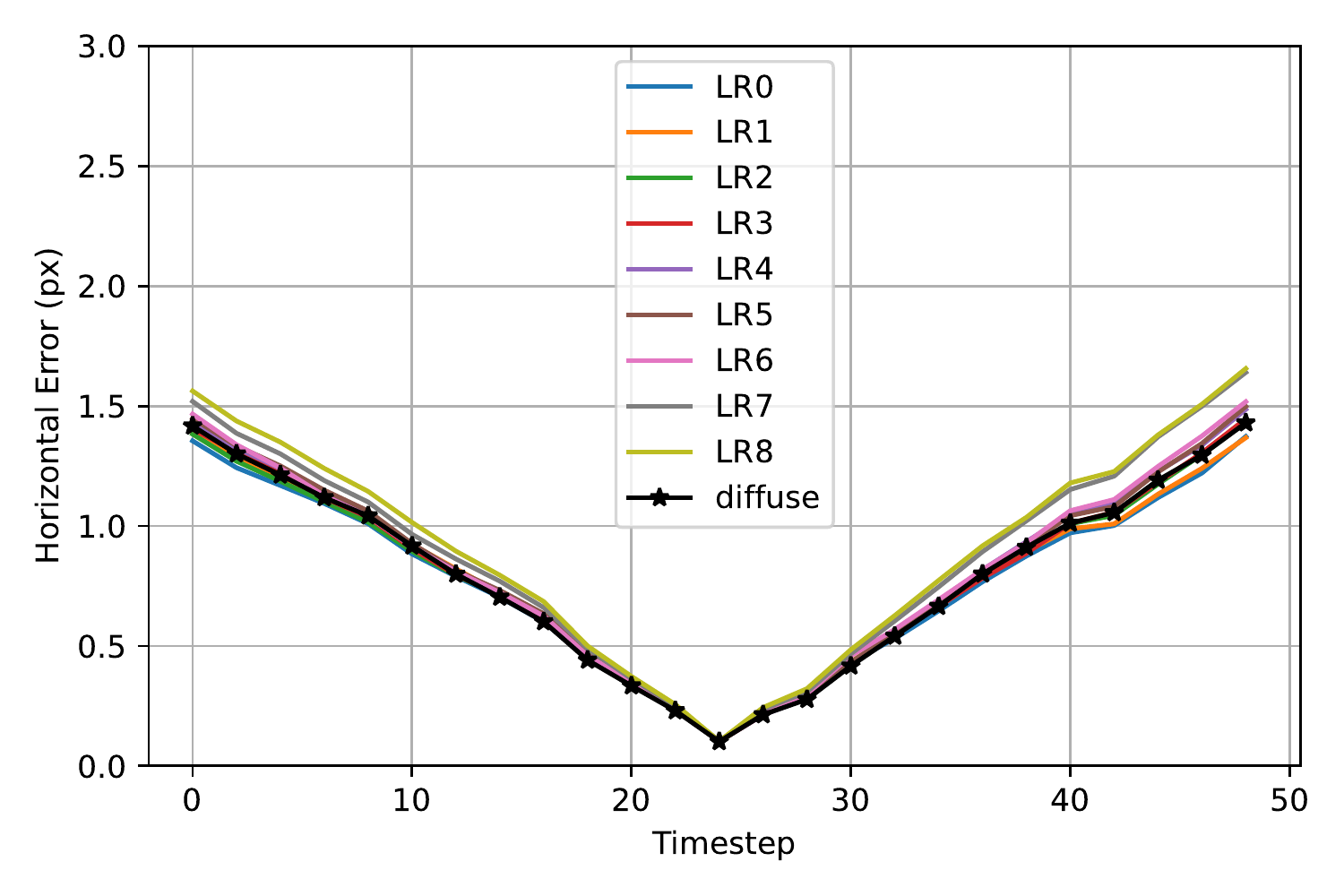}
        \includegraphics[width=0.48\textwidth]{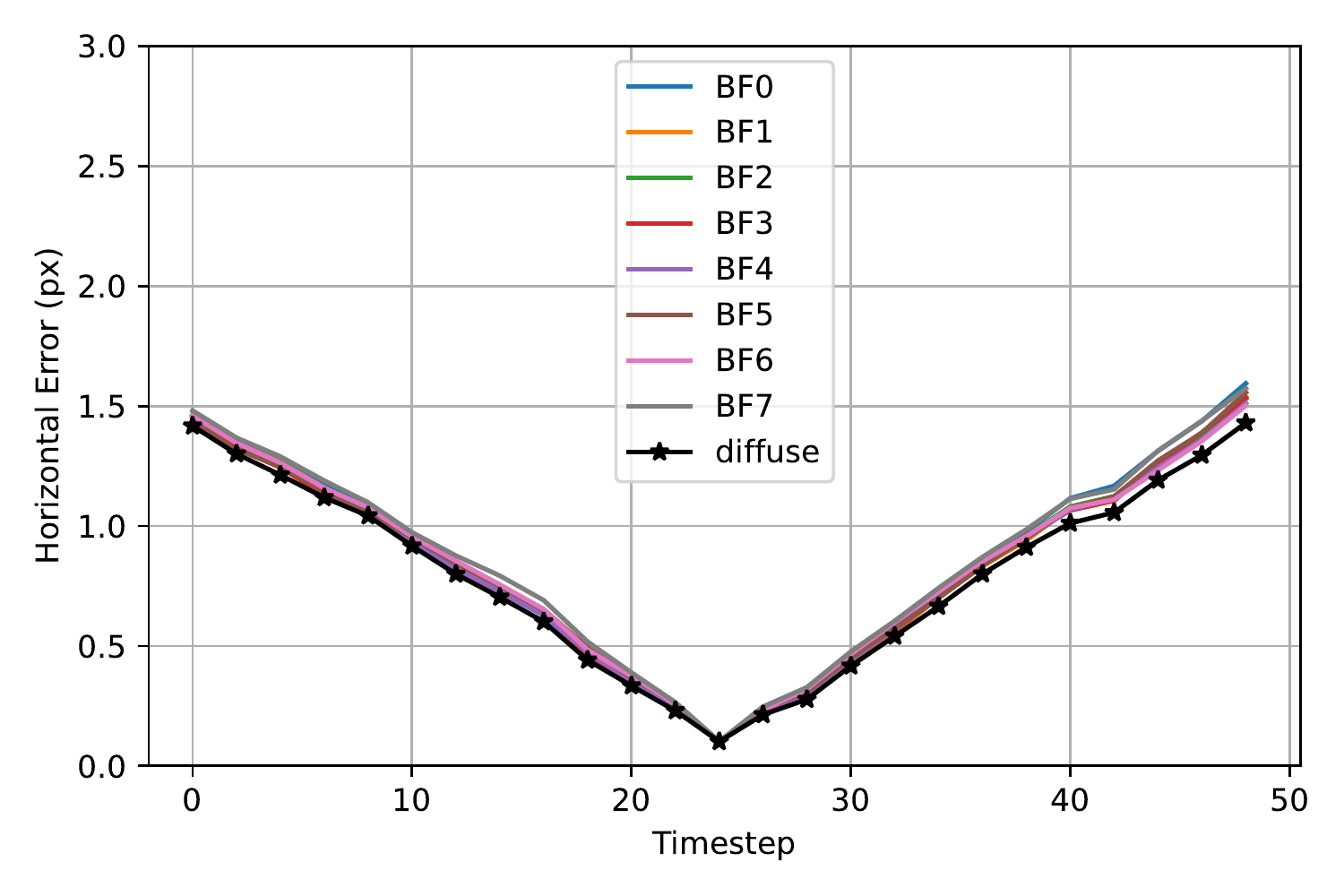}
    }
    \subfigure[Vertical Coordinate]{
        \includegraphics[width=0.48\textwidth]{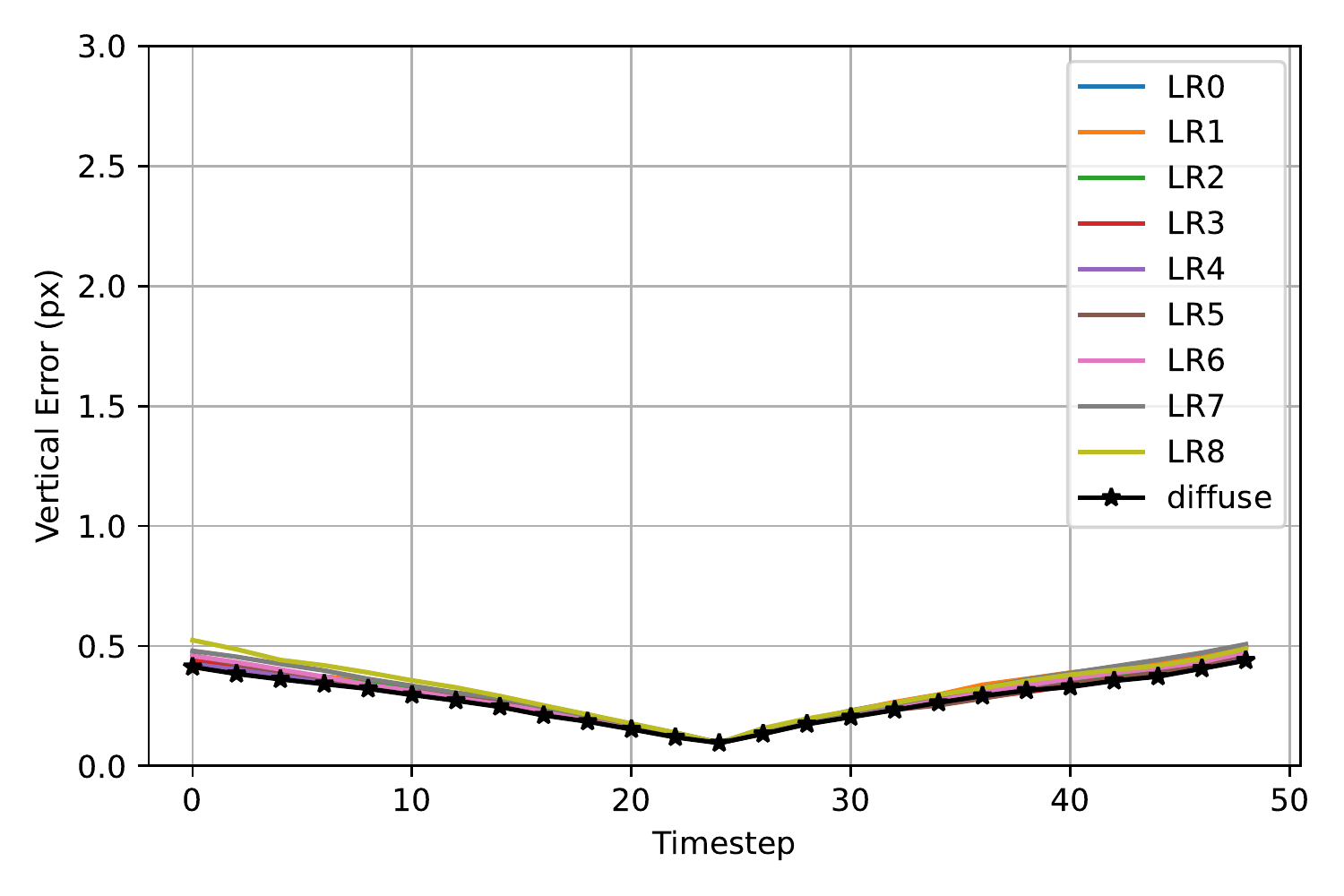}
        \includegraphics[width=0.48\textwidth]{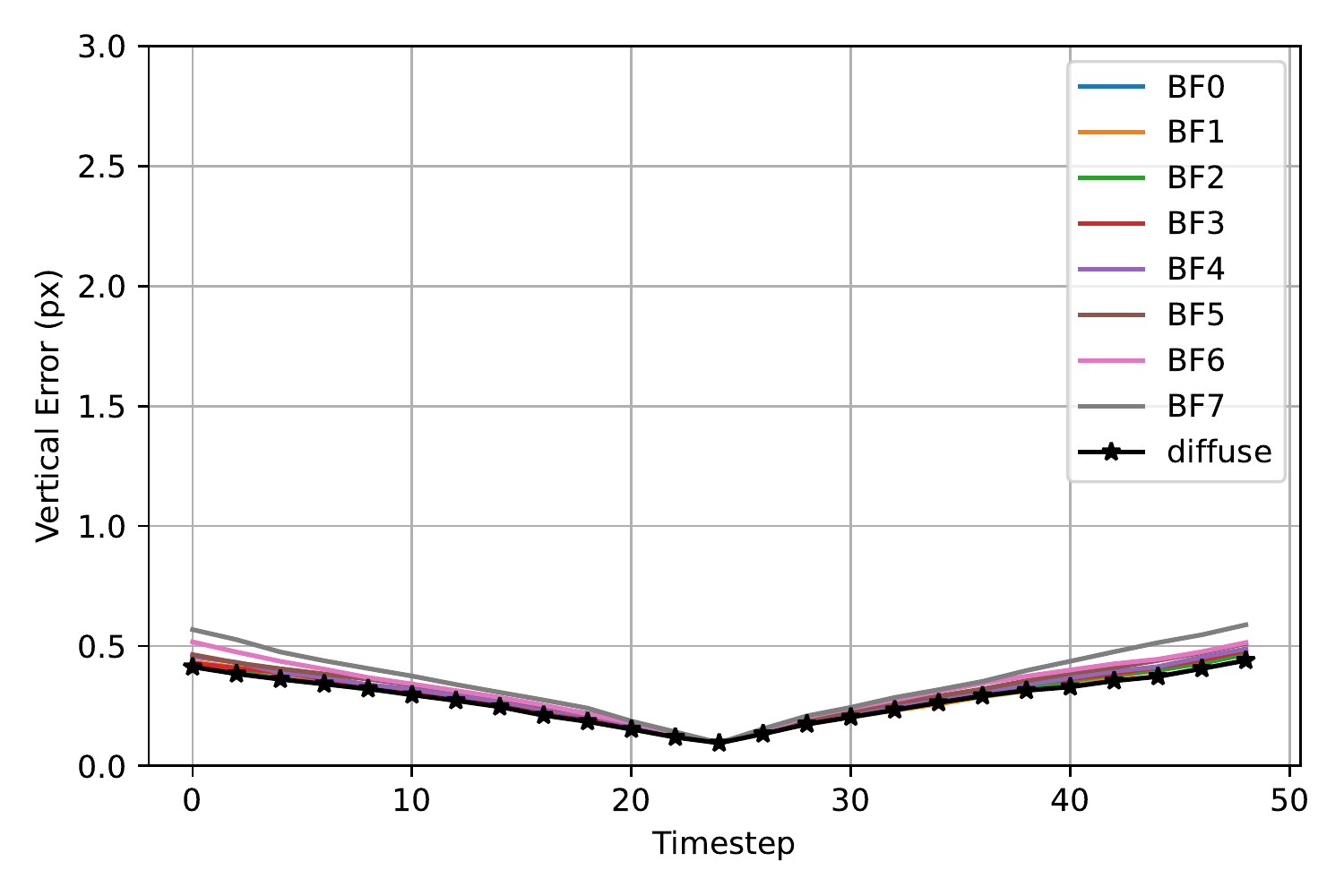}
    }
    \caption{\textbf{DTU Point Features Dataset: At twice nominal speed, lighting condition does not change trends in mean absolute error $\kappa(t)$ when using the Lucas-Kanade Tracker.} We compute $\kappa(t)$ using diffuse lighting (black lines) and each of the directional lighting conditions listed in Figure \ref{fig:dtu_light_stage} using all tracks from all 60 scenes.  The variation of $\kappa(t)$ due to the existence of directional lighting is at most 10 percent of the variation common to all plotted lines. The effect of directional lighting is relatively small because changes between adjacent frames are small whether or not the scene contains directional lighting.}
    \label{dtu_LK_kappa_speed2.00}
\end{figure}

\begin{figure}[H]
    \centering
    \subfigure[Horizontal Coordinate]{
        \includegraphics[width=0.48\textwidth]{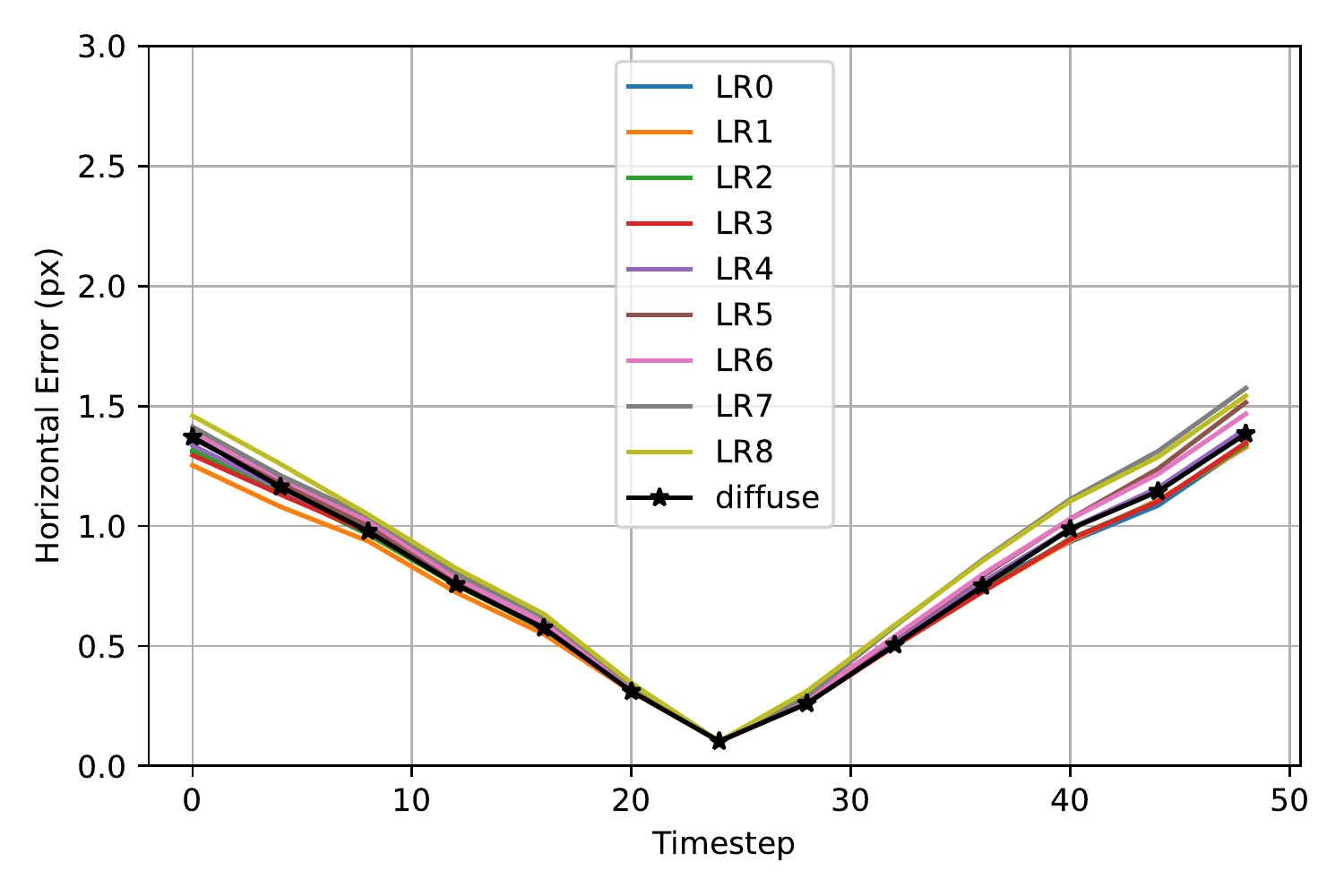}
        \includegraphics[width=0.48\textwidth]{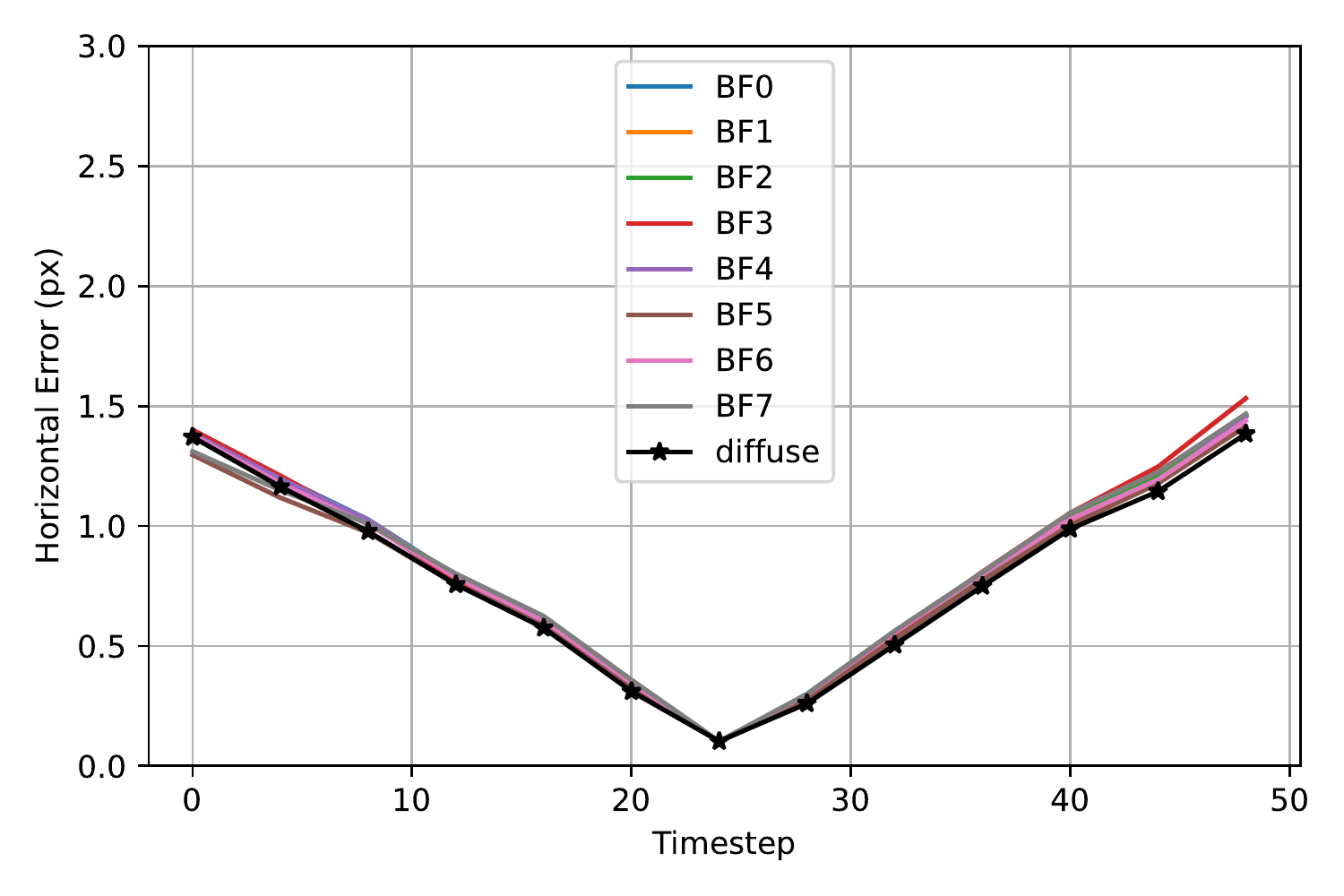}
    }
    \subfigure[Vertical Coordinate]{
        \includegraphics[width=0.48\textwidth]{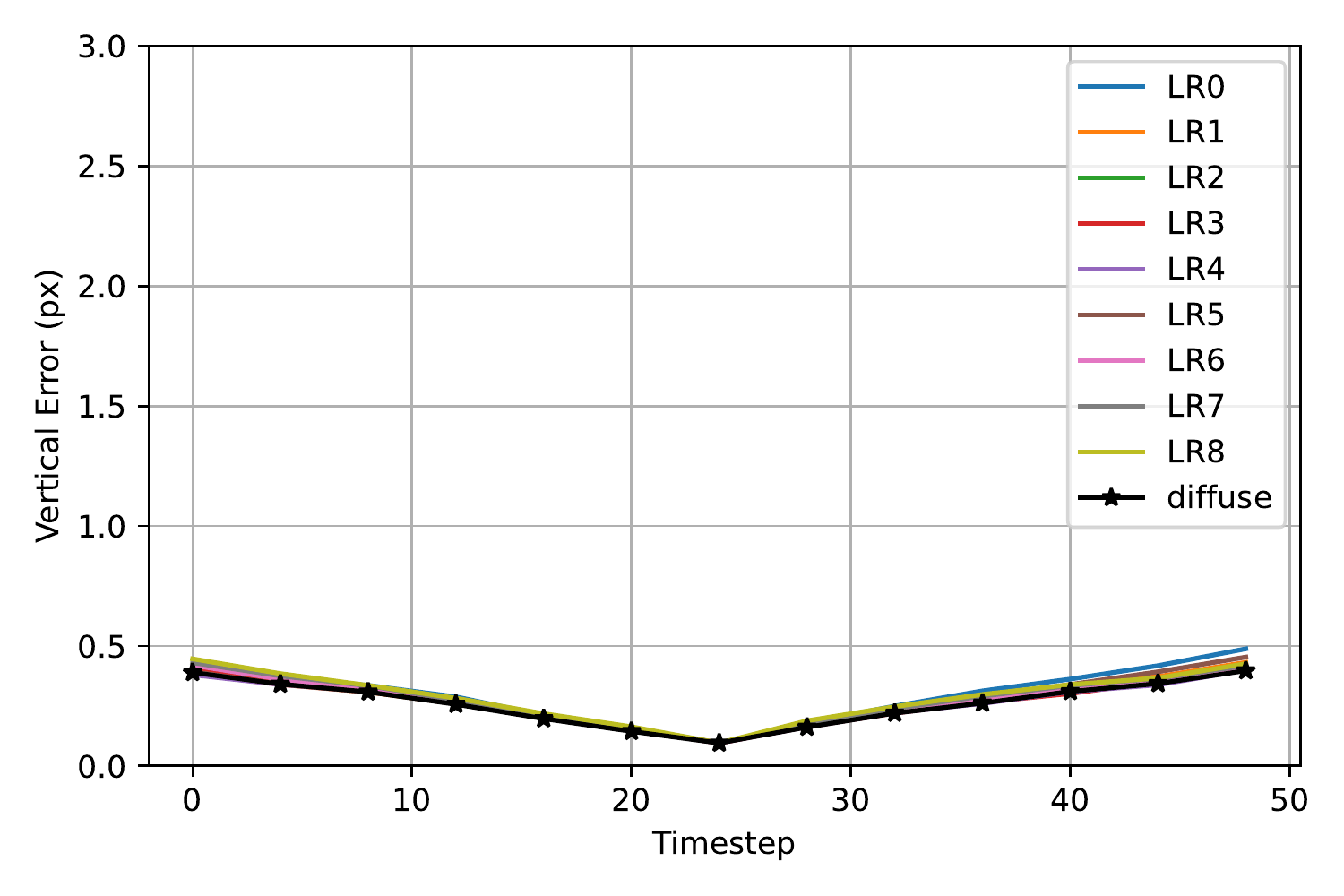}
        \includegraphics[width=0.48\textwidth]{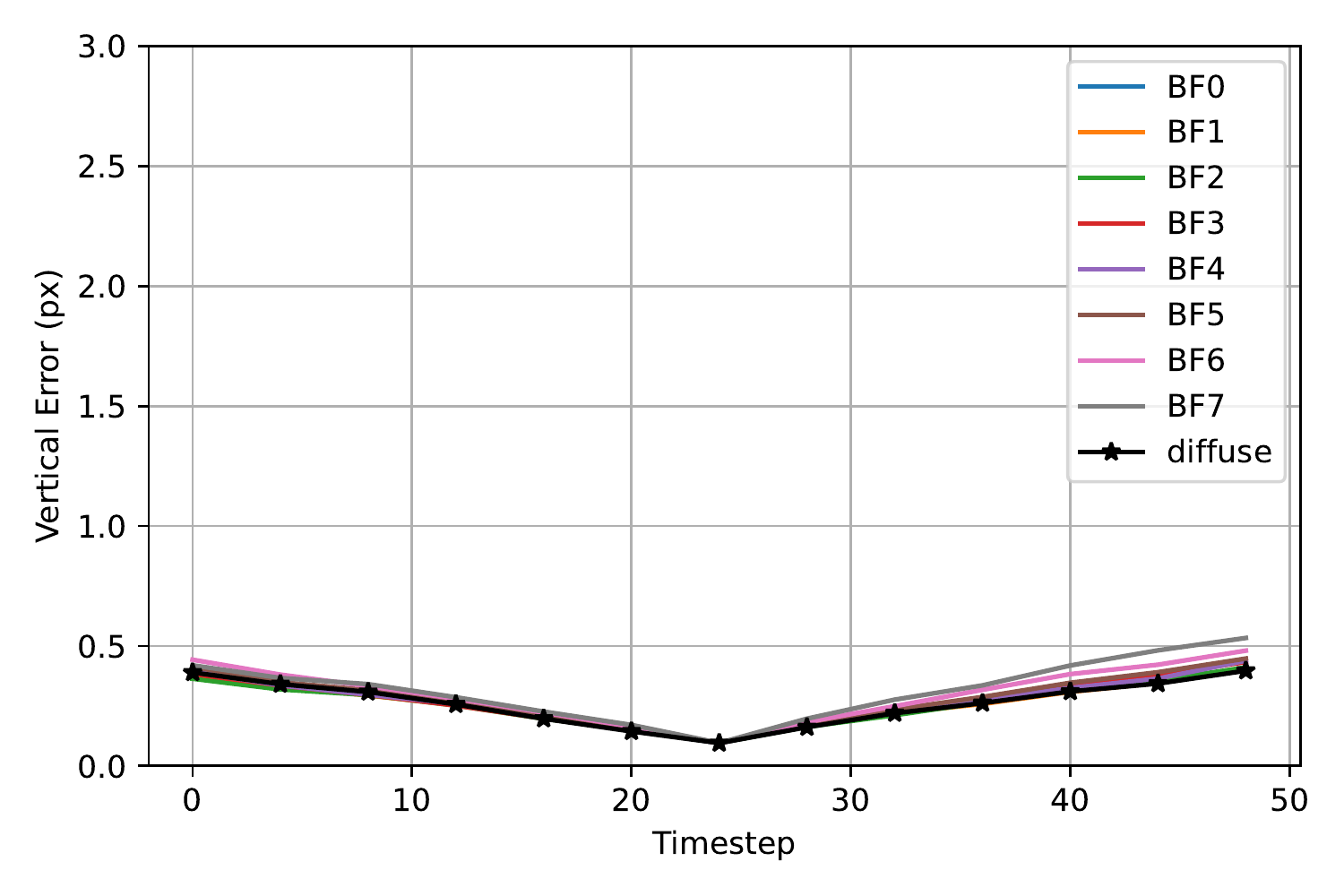}
    }
    \caption{\textbf{DTU Point Features Dataset: At four times nominal speed, lighting condition does not change trends in mean absolute error $\kappa(t)$ when using the Lucas-Kanade Tracker.} We compute $\kappa(t)$ using diffuse lighting (black lines) and each of the directional lighting conditions listed in Figure \ref{fig:dtu_light_stage} using all tracks from all 60 scenes.   The variation of $\kappa(t)$ due to the existence of directional lighting is at most 10 percent of the variation common to all plotted lines. The effect of directional lighting is relatively small because changes between adjacent frames are small whether or not the scene contains directional lighting.}
    \label{dtu_LK_kappa_speed4.00}
\end{figure}

\begin{figure}[H]
    \centering
    \subfigure[Horizontal Coordinate]{
        \includegraphics[width=0.48\textwidth]{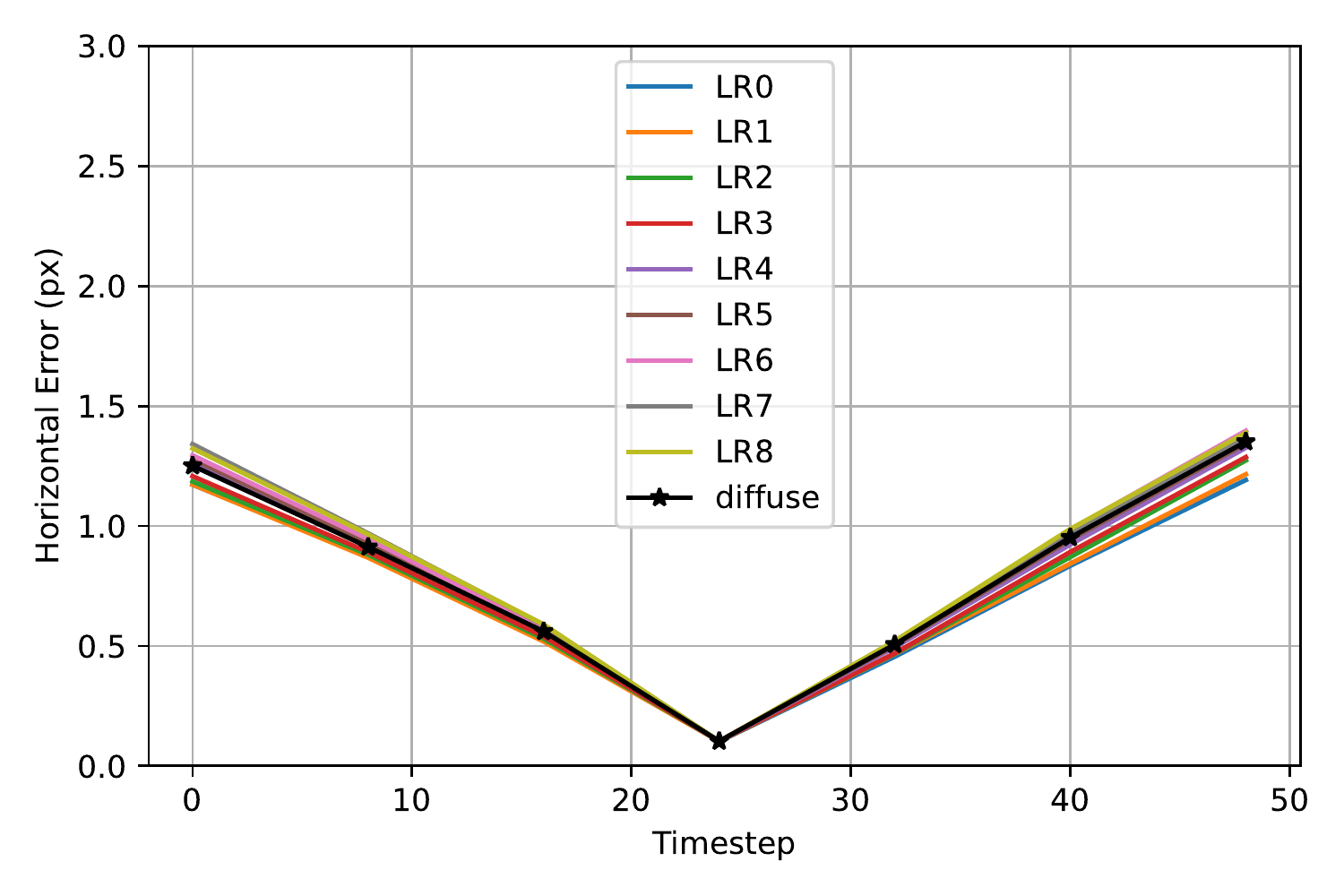}
        \includegraphics[width=0.48\textwidth]{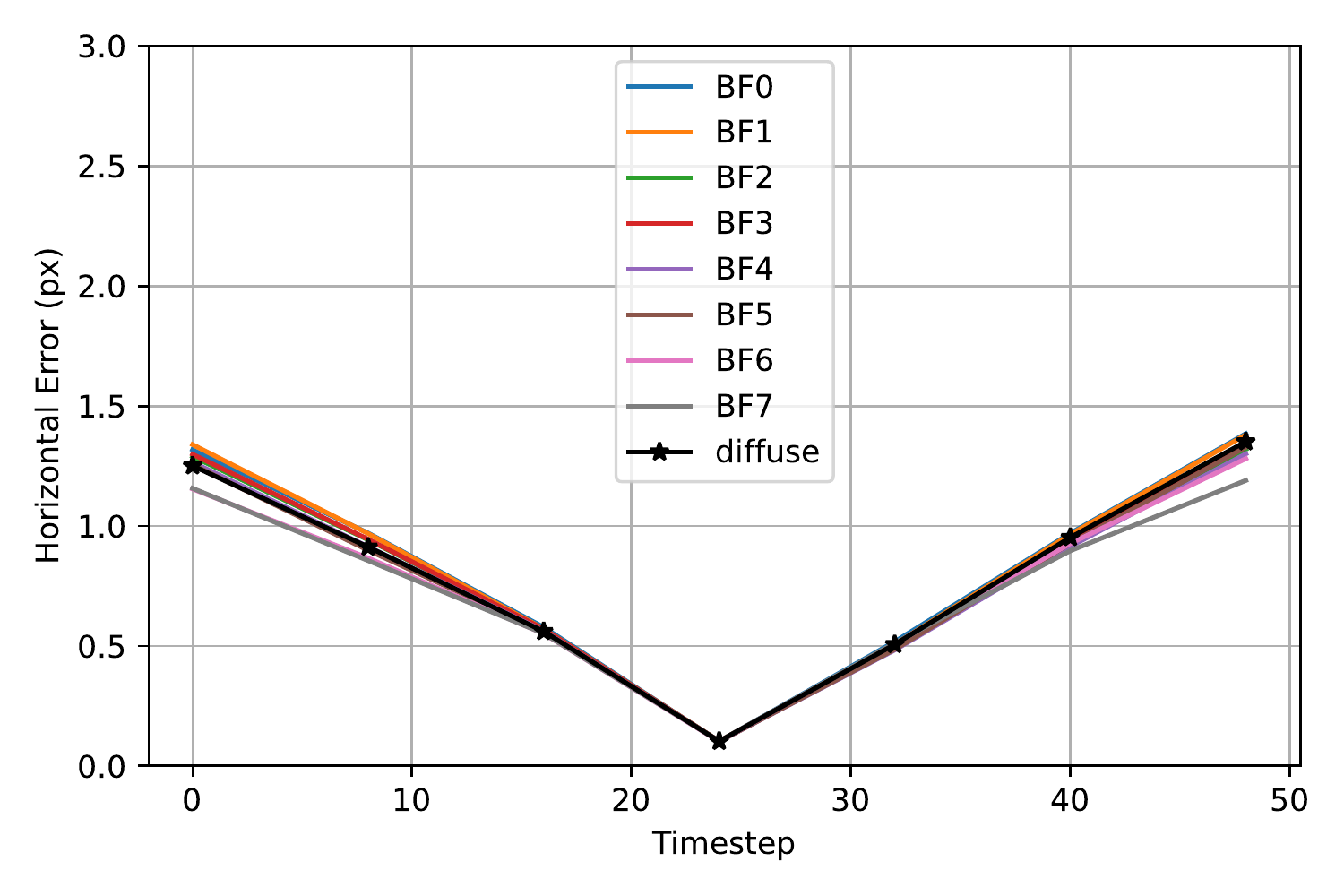}
    }
    \subfigure[Vertical Coordinate]{
        \includegraphics[width=0.48\textwidth]{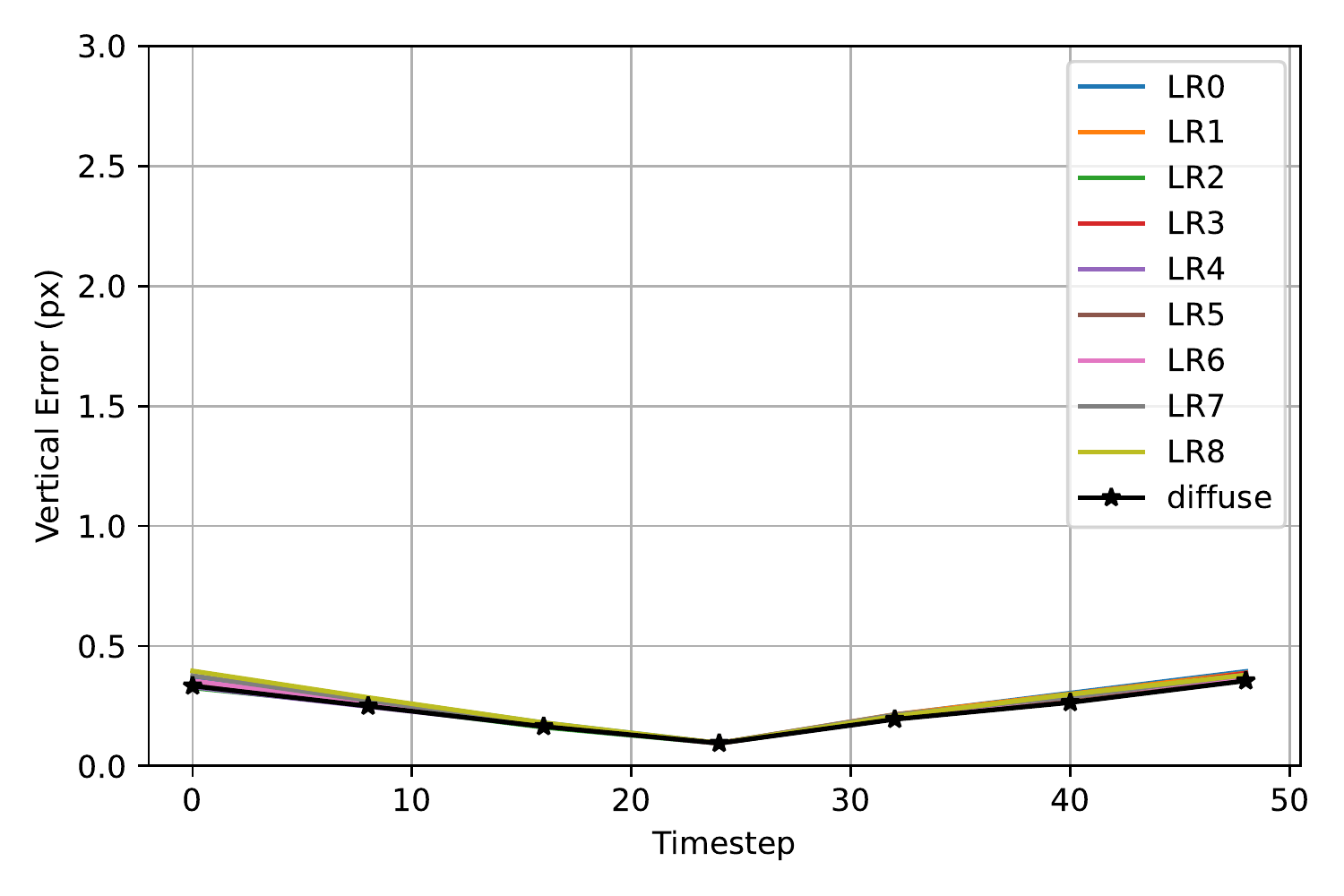}
        \includegraphics[width=0.48\textwidth]{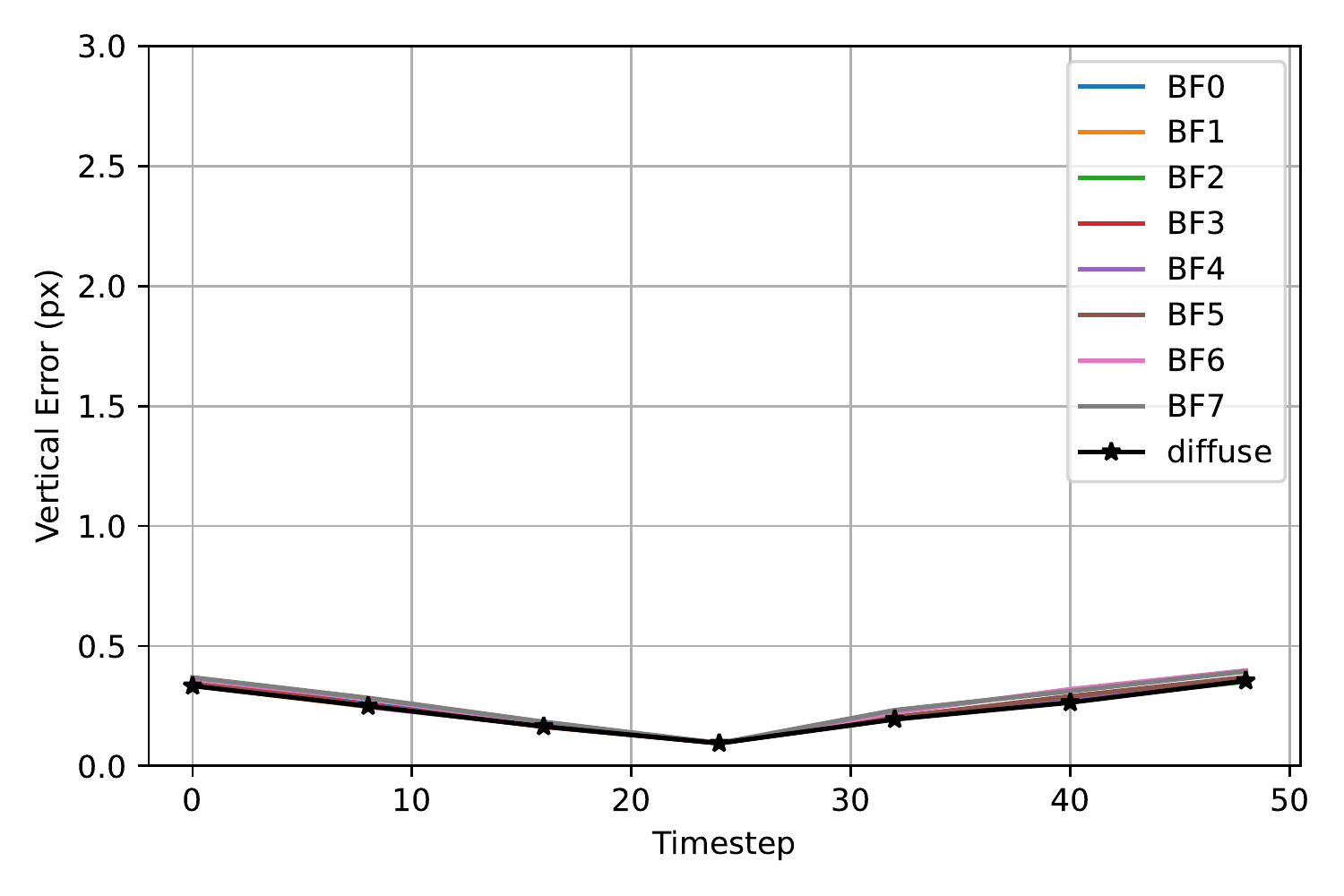}
    }
    \caption{\textbf{DTU Point Features Dataset: At eight times nominal speed, lighting condition does not change trends in $\kappa(t)$ when using the Lucas-Kanade Tracker.} We compute $\kappa(t)$ using diffuse lighting (black lines) and each of the directional lighting conditions listed in Figure \ref{fig:dtu_light_stage} using all tracks from all 60 scenes.  The variation of $\kappa(t)$ due to the existence of directional lighting is at most 10 percent of the variation common to all plotted lines. The effect of directional lighting is relatively small because changes between adjacent frames are small whether or not the scene contains directional lighting.}
    \label{dtu_LK_kappa_speed8.00}
\end{figure}

\begin{figure}[H]
    \centering
    \subfigure[Horizontal Coordinate]{
        \includegraphics[width=0.48\textwidth]{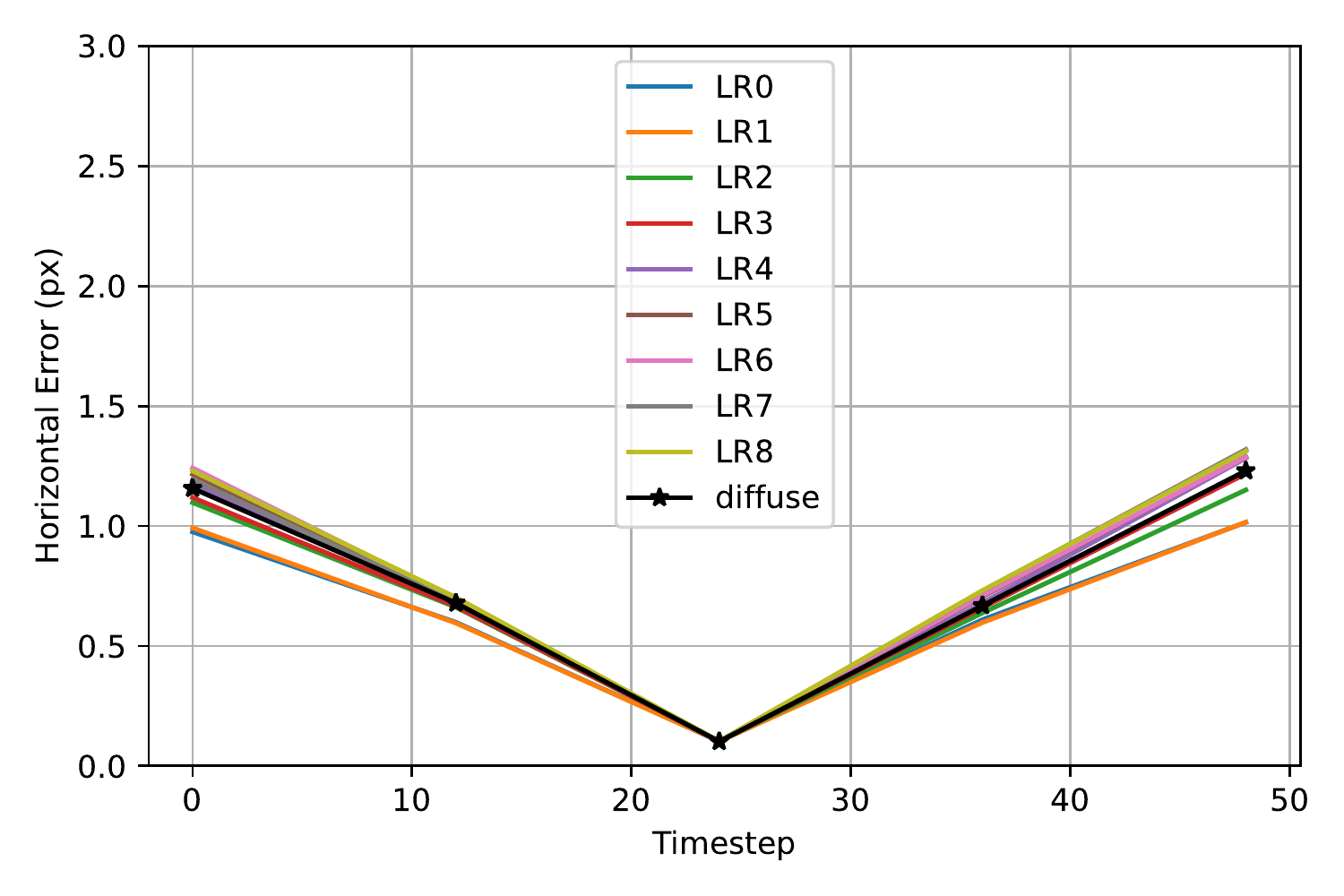}
        \includegraphics[width=0.48\textwidth]{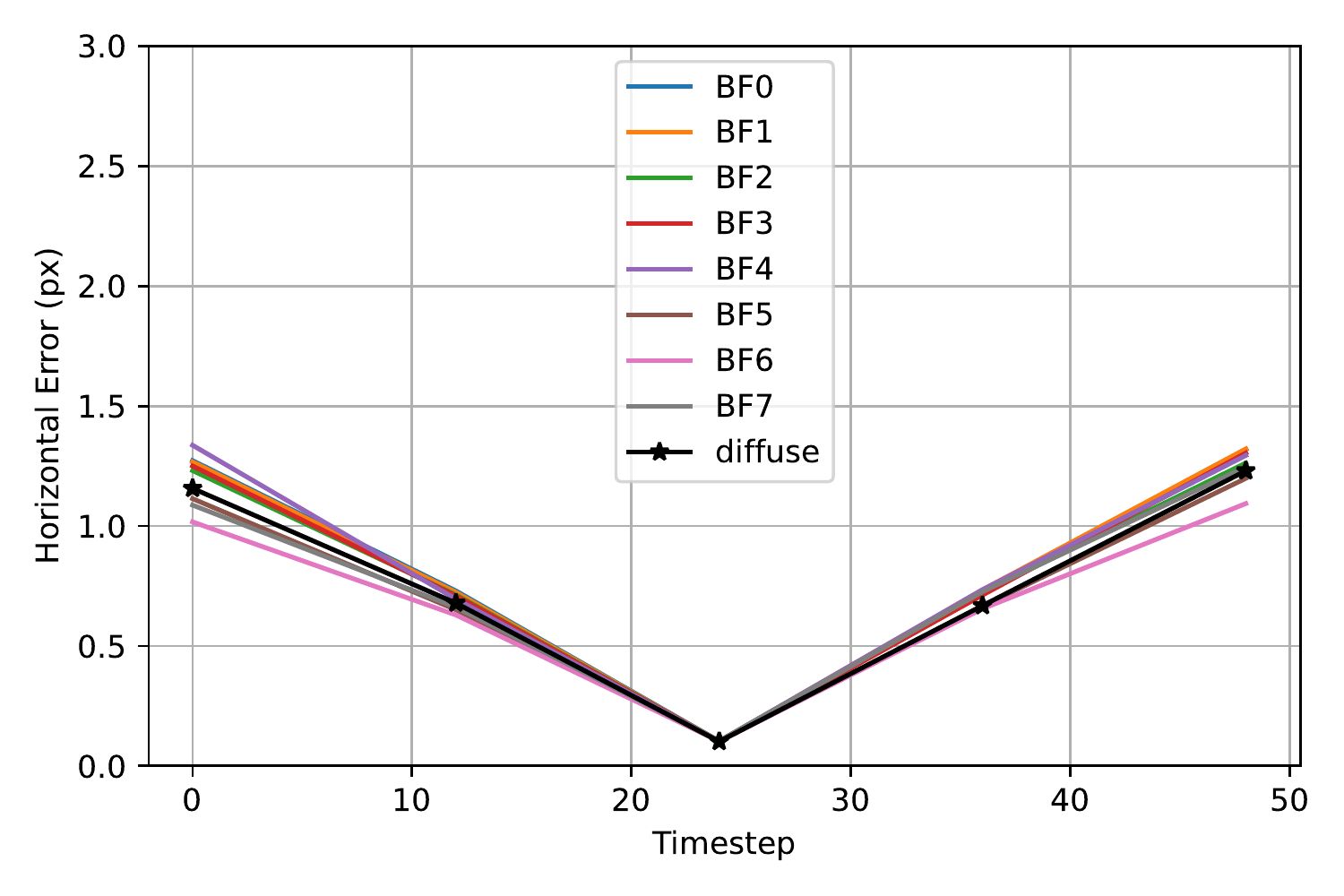}
    }
    \subfigure[Vertical Coordinate]{
        \includegraphics[width=0.48\textwidth]{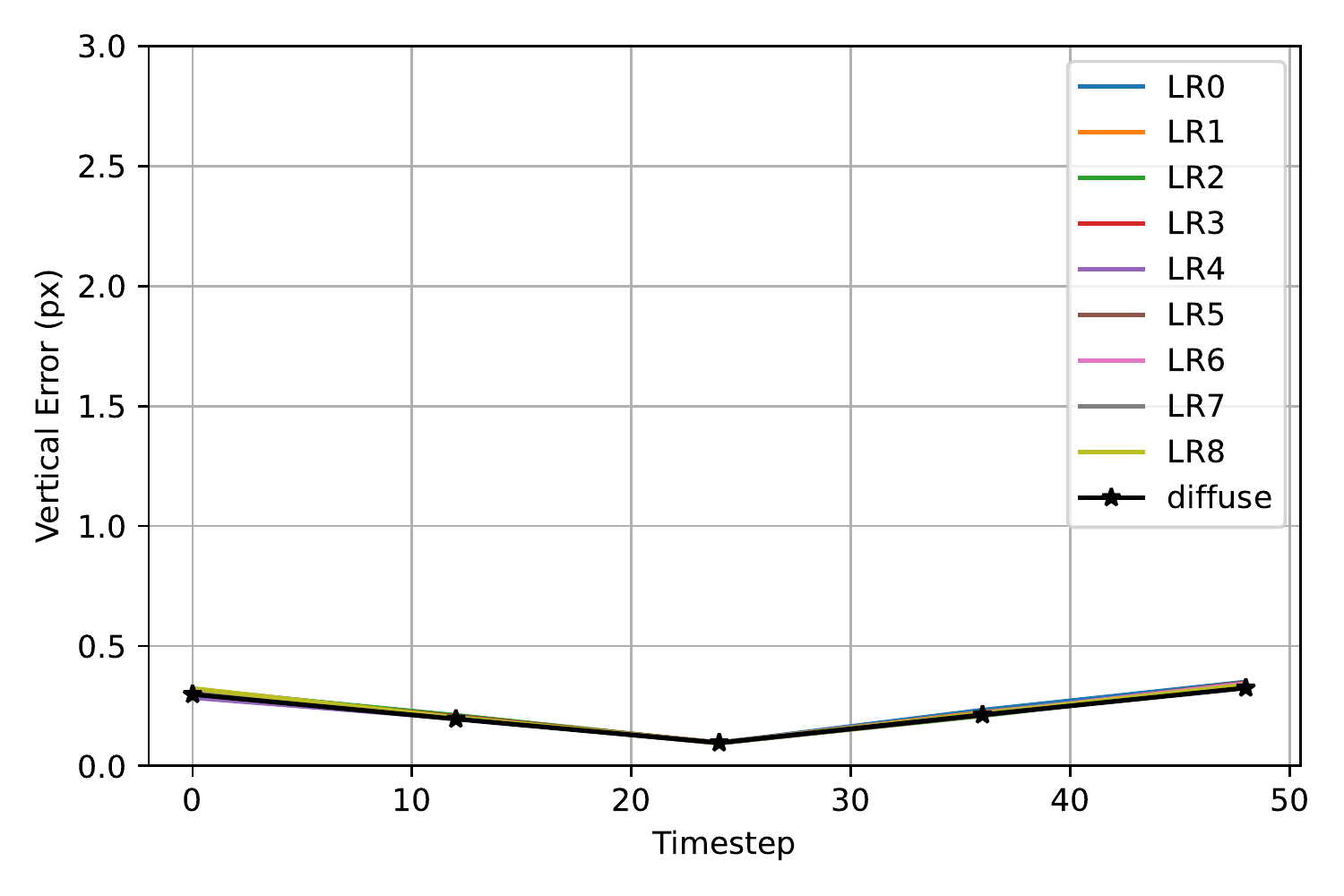}
        \includegraphics[width=0.48\textwidth]{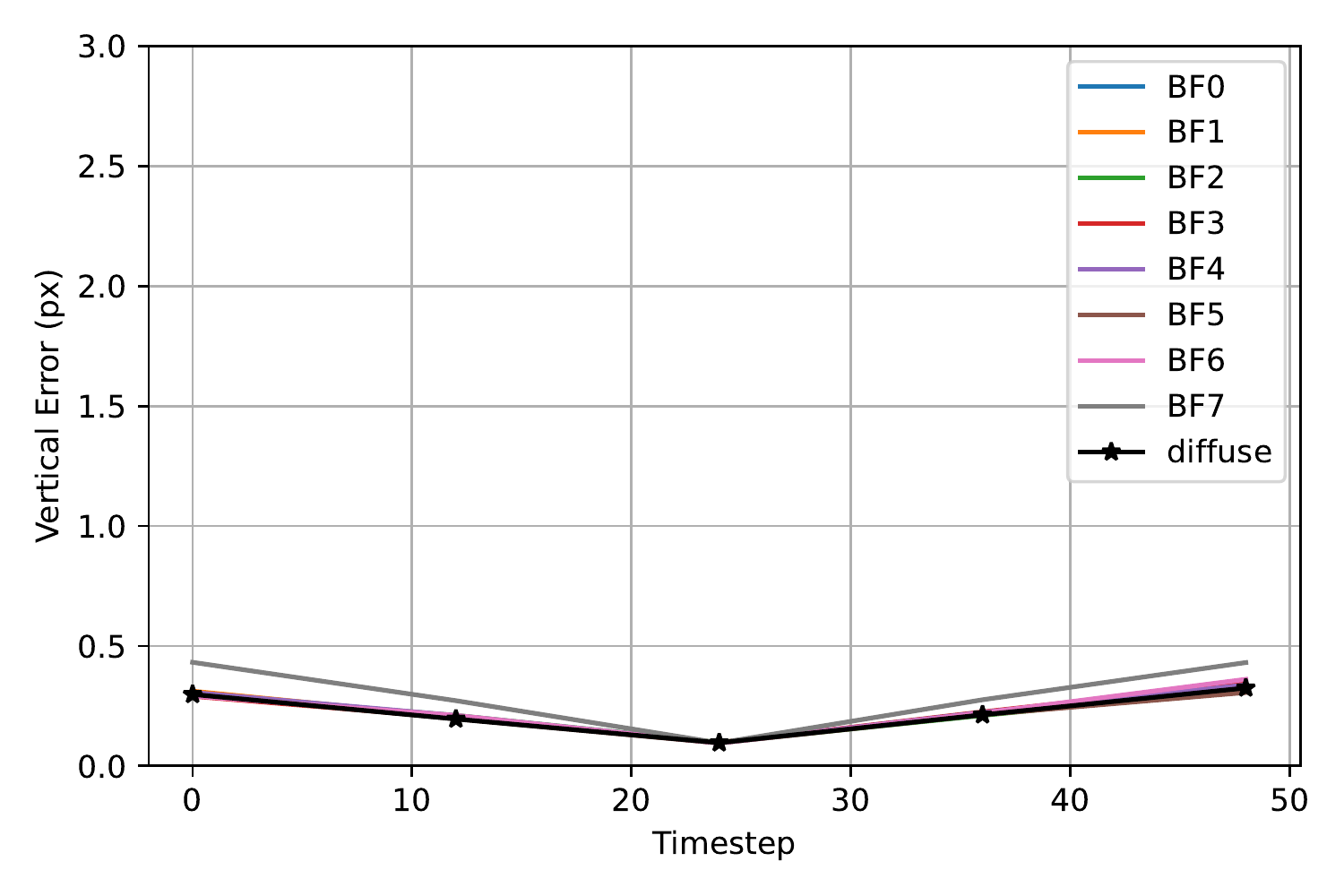}
    } 
    \caption{\textbf{DTU Point Features Dataset: At twelve times nominal speed, lighting condition does not change trends in mean absolute error $\kappa(t)$ when using the Lucas-Kanade Tracker.} We compute $\kappa(t)$ using diffuse lighting (black lines) and each of the directional lighting conditions listed in Figure \ref{fig:dtu_light_stage} using all tracks from all 60 scenes. The variation of $\kappa(t)$ due to the existence of directional lighting is at most 10 percent of the variation common to all plotted lines. The effect of directional lighting is relatively small because changes between adjacent frames are small whether or not the scene contains directional lighting.
}
    \label{fig:dtu_LK_kappa_speed12.00}
\end{figure}

\begin{figure}[H]
    \centering
    \subfigure[Horizontal Coordinate]{
        \includegraphics[width=0.48\textwidth]{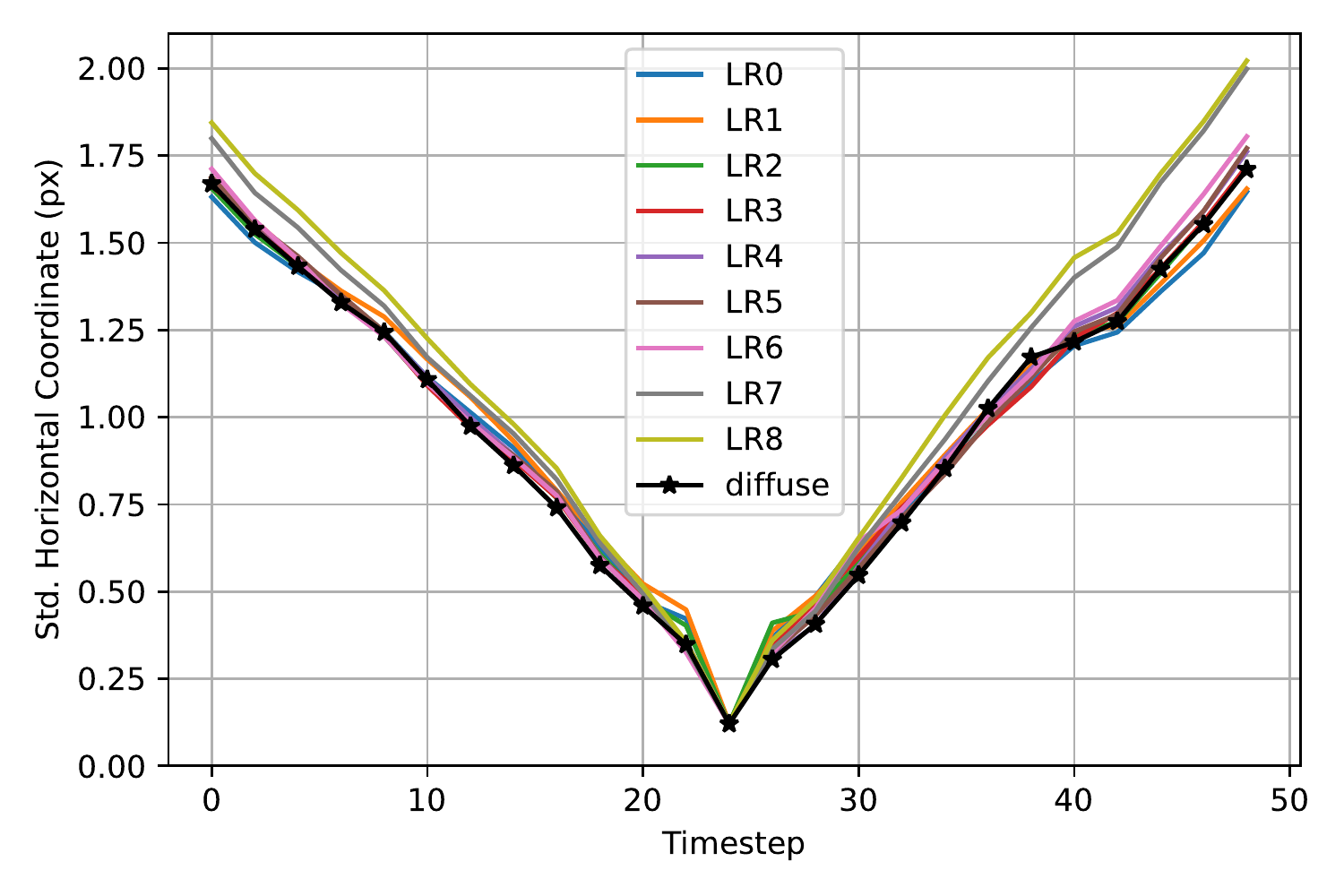}
        \includegraphics[width=0.48\textwidth]{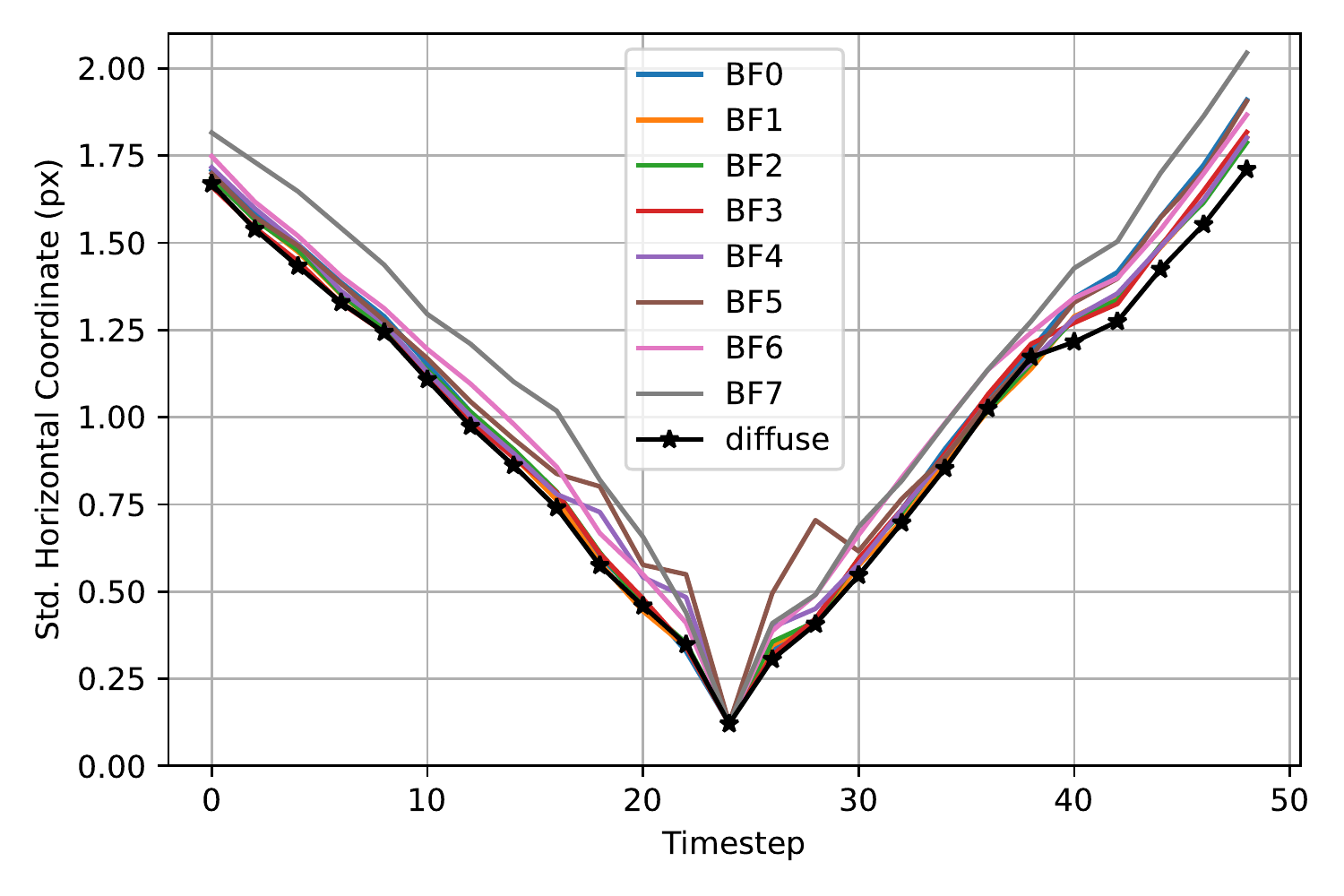}
    }
    \subfigure[Vertical Coordinate]{
        \includegraphics[width=0.48\textwidth]{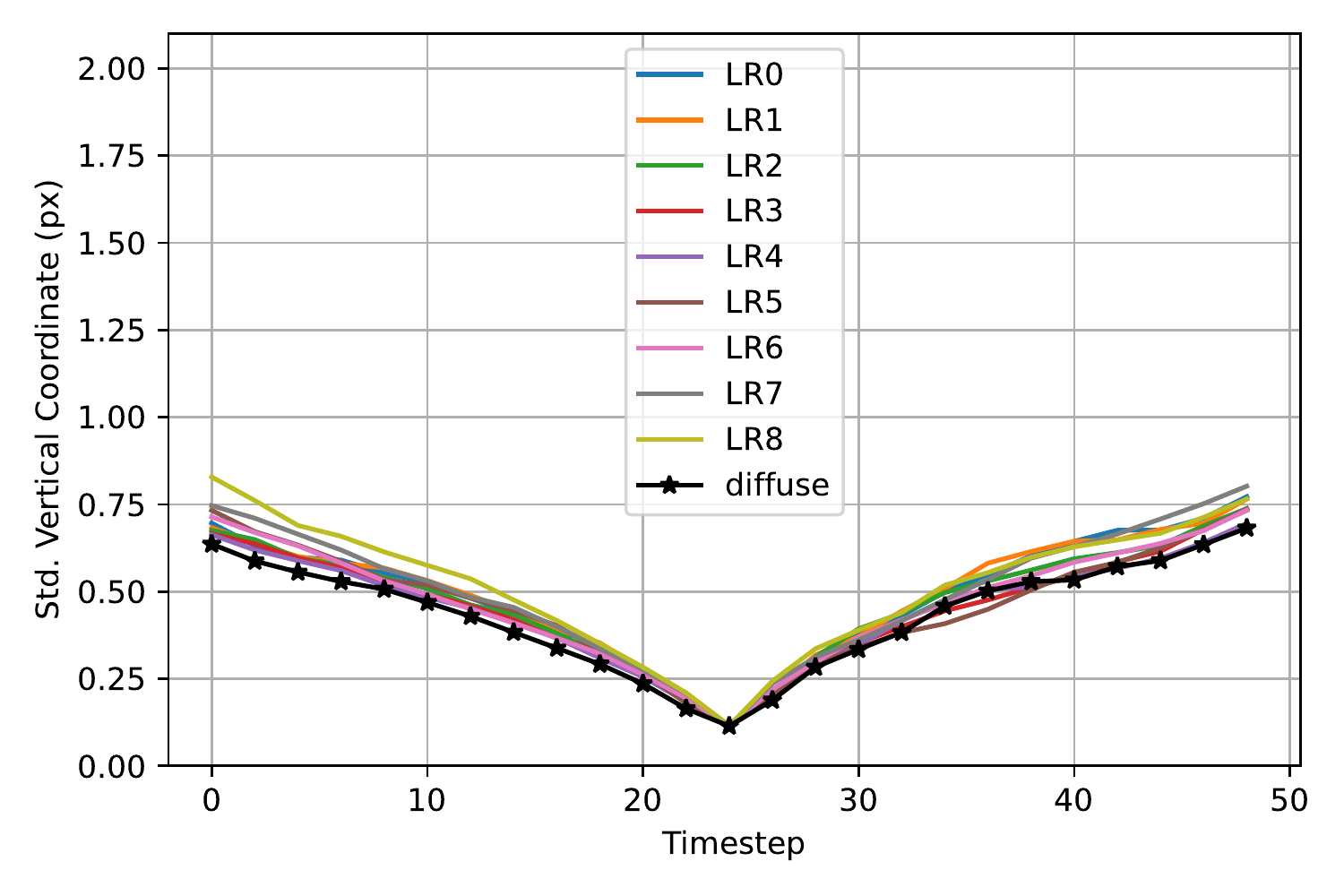}
        \includegraphics[width=0.48\textwidth]{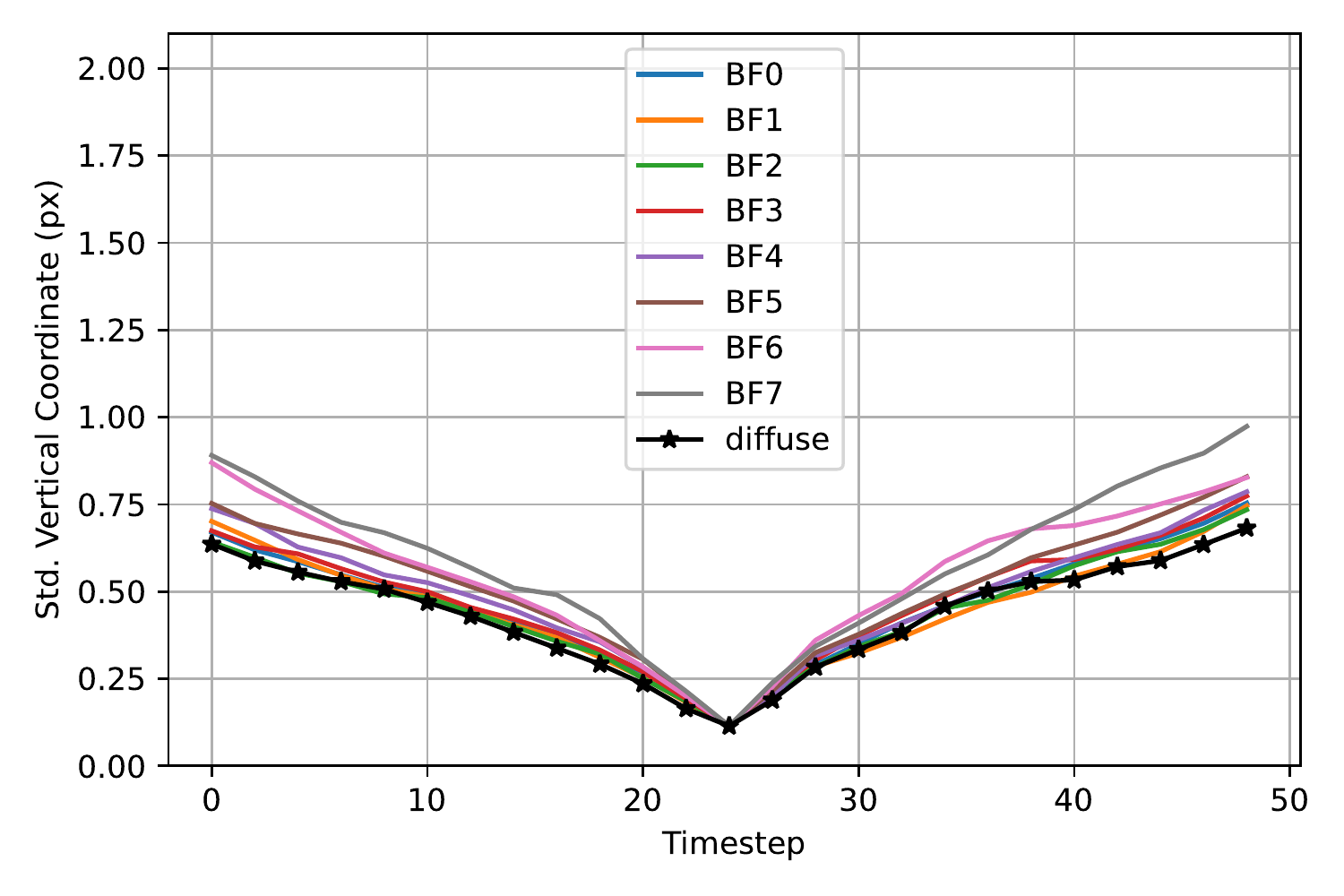}
    }
    \caption{\textbf{DTU Point Features Dataset: At twice nominal speed, lighting condition does not change trends in covariance $\Sigma(t)$  when using the Lucas-Kanade Tracker.} We compute $\Sigma(t)$ using diffuse lighting (black lines) and each of the directional lighting conditions listed in Figure \ref{fig:dtu_light_stage} using all tracks from all 60 scenes. 
    The variation of $\Sigma(t)$ due to the existence of directional lighting is at most 10 percent of the variation common to all plotted lines. The effect of directional lighting is relatively small because changes between adjacent frames are small whether or not the scene contains directional lighting.}
    \label{fig:dtu_LK_cov_speed2.00}
\end{figure}

\begin{figure}[H]
    \centering
    \subfigure[Horizontal Coordinate]{
        \includegraphics[width=0.48\textwidth]{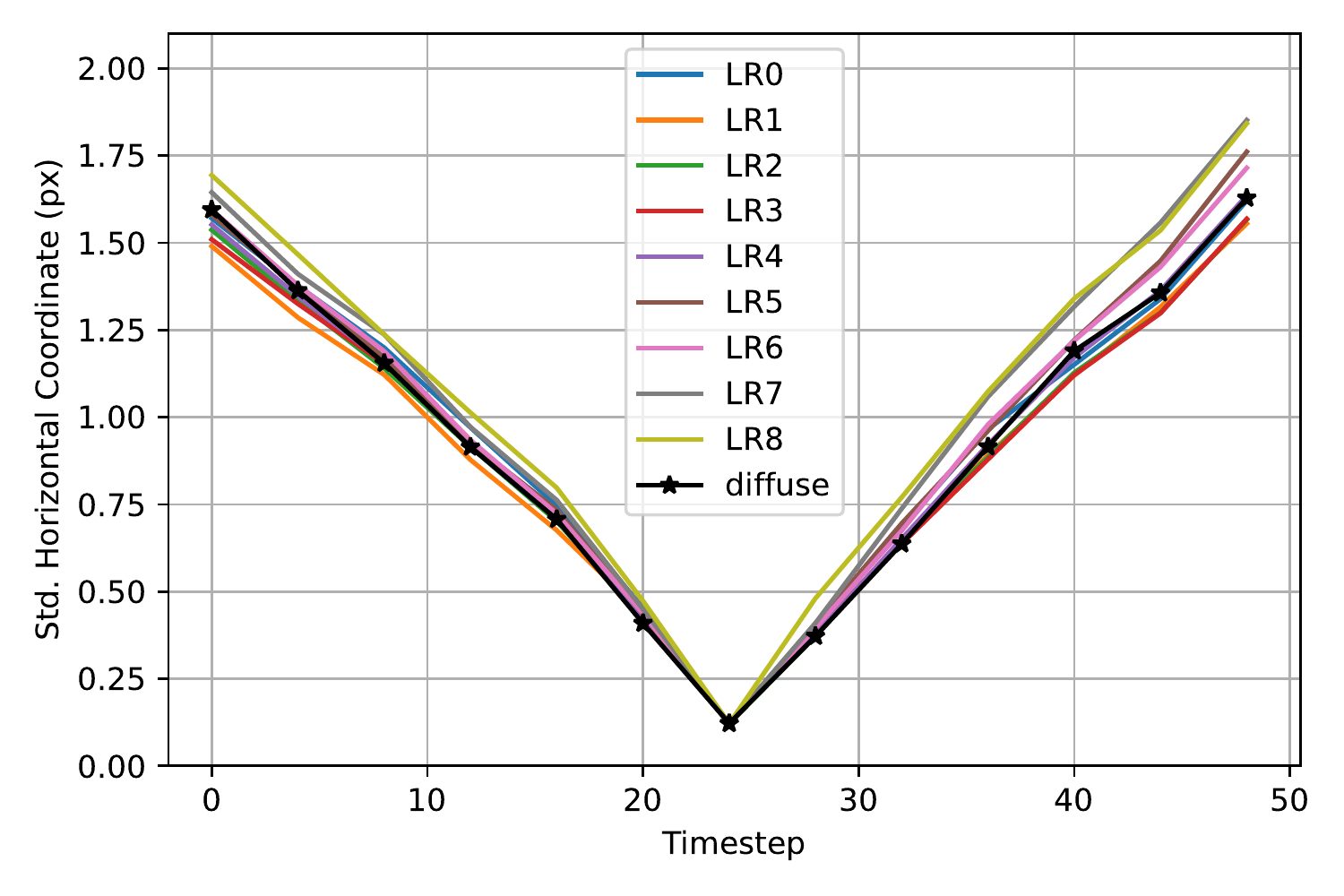}
        \includegraphics[width=0.48\textwidth]{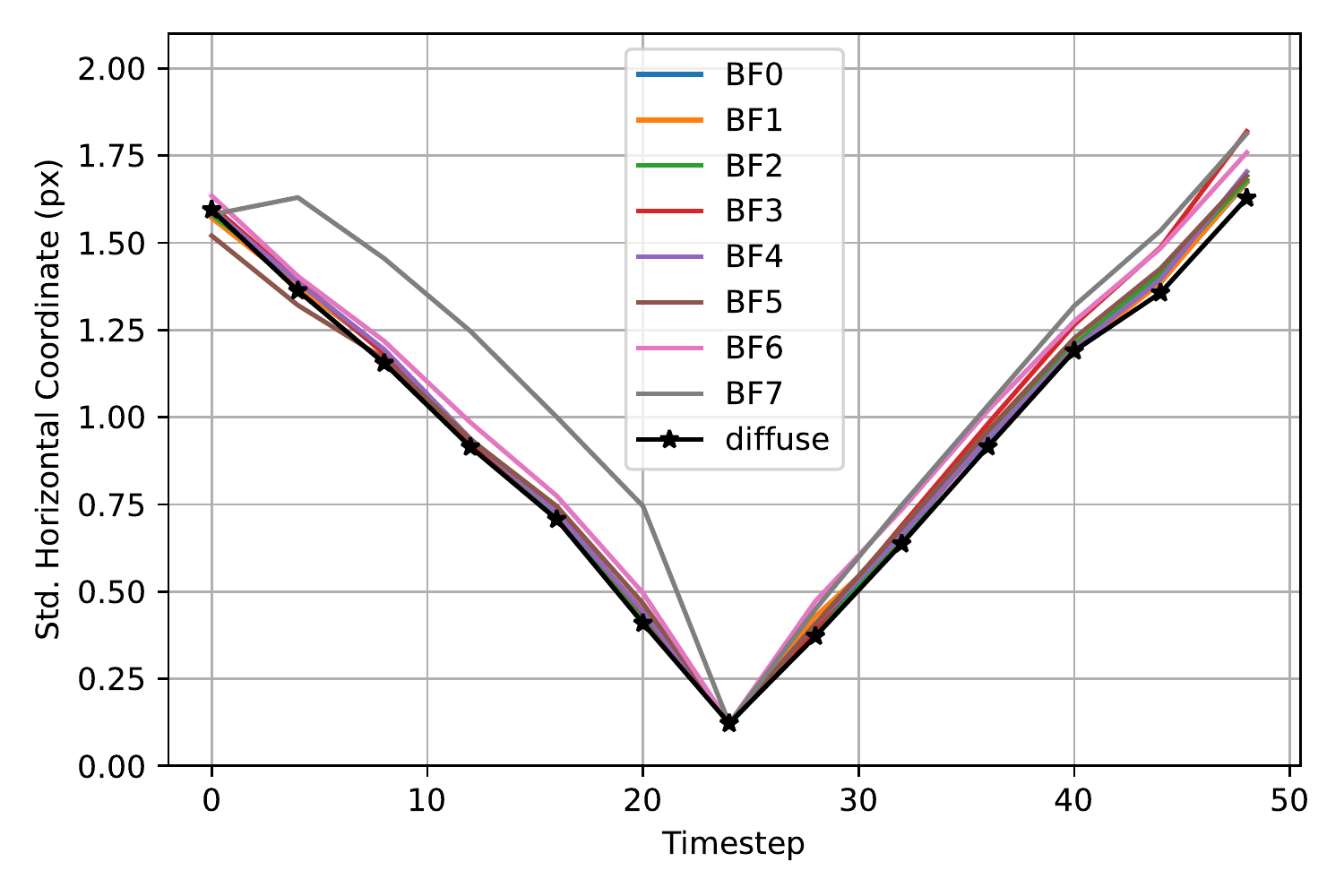}
    }
    \subfigure[Vertical Coordinate]{
        \includegraphics[width=0.48\textwidth]{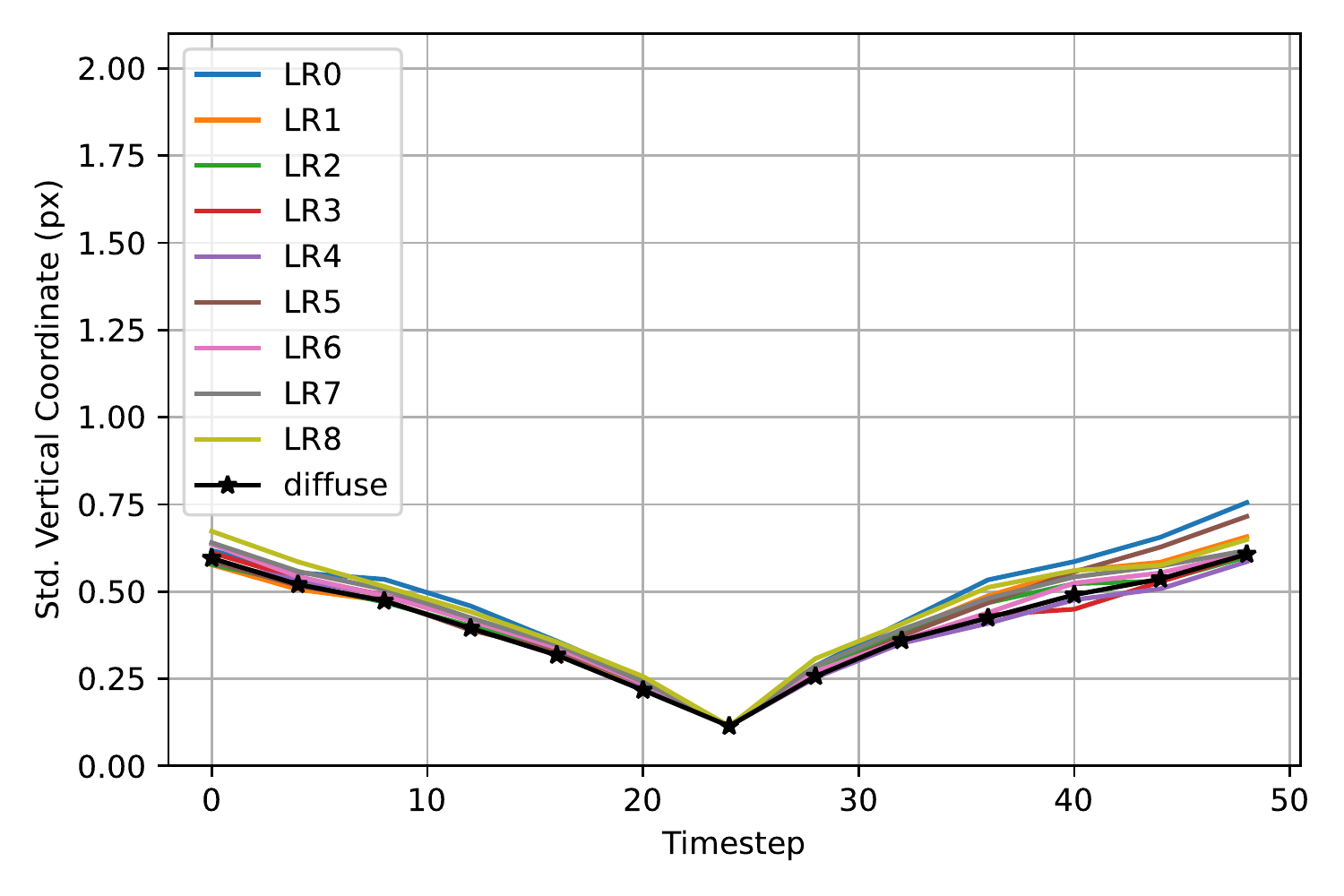}
        \includegraphics[width=0.48\textwidth]{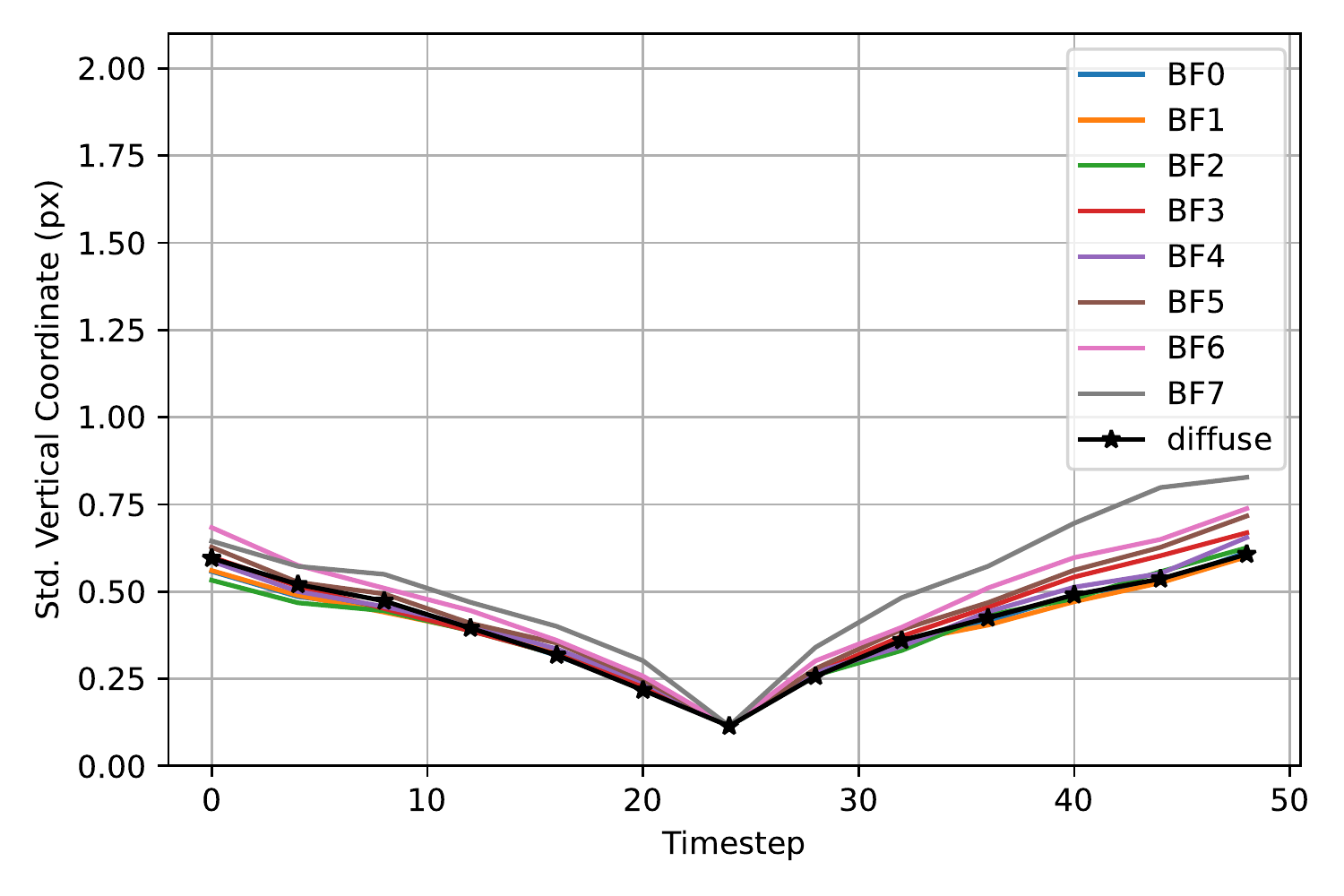}
    }
    \caption{\textbf{DTU Point Features Dataset: At four times nominal speed, lighting condition does not change trends in covariance $\Sigma(t)$  when using the Lucas-Kanade Tracker.} We compute $\Sigma(t)$ using diffuse lighting (black lines) and each of the directional lighting conditions listed in Figure \ref{fig:dtu_light_stage} using all tracks from all 60 scenes. The variation of $\Sigma(t)$ due to the existence of directional lighting is less than 10 percent of the variation common to all plotted lines for all but one lighting condition. The effect of directional lighting is relatively small because changes between adjacent frames are small whether or not the scene contains directional lighting.
    The larger-than average covariance for lighting condition BF7 is caused by a single scene where feature tracking fails.}
    \label{fig:dtu_LK_cov_speed4.00}
\end{figure}

\begin{figure}[H]
    \centering
    \subfigure[Horizontal Coordinate]{
        \includegraphics[width=0.48\textwidth]{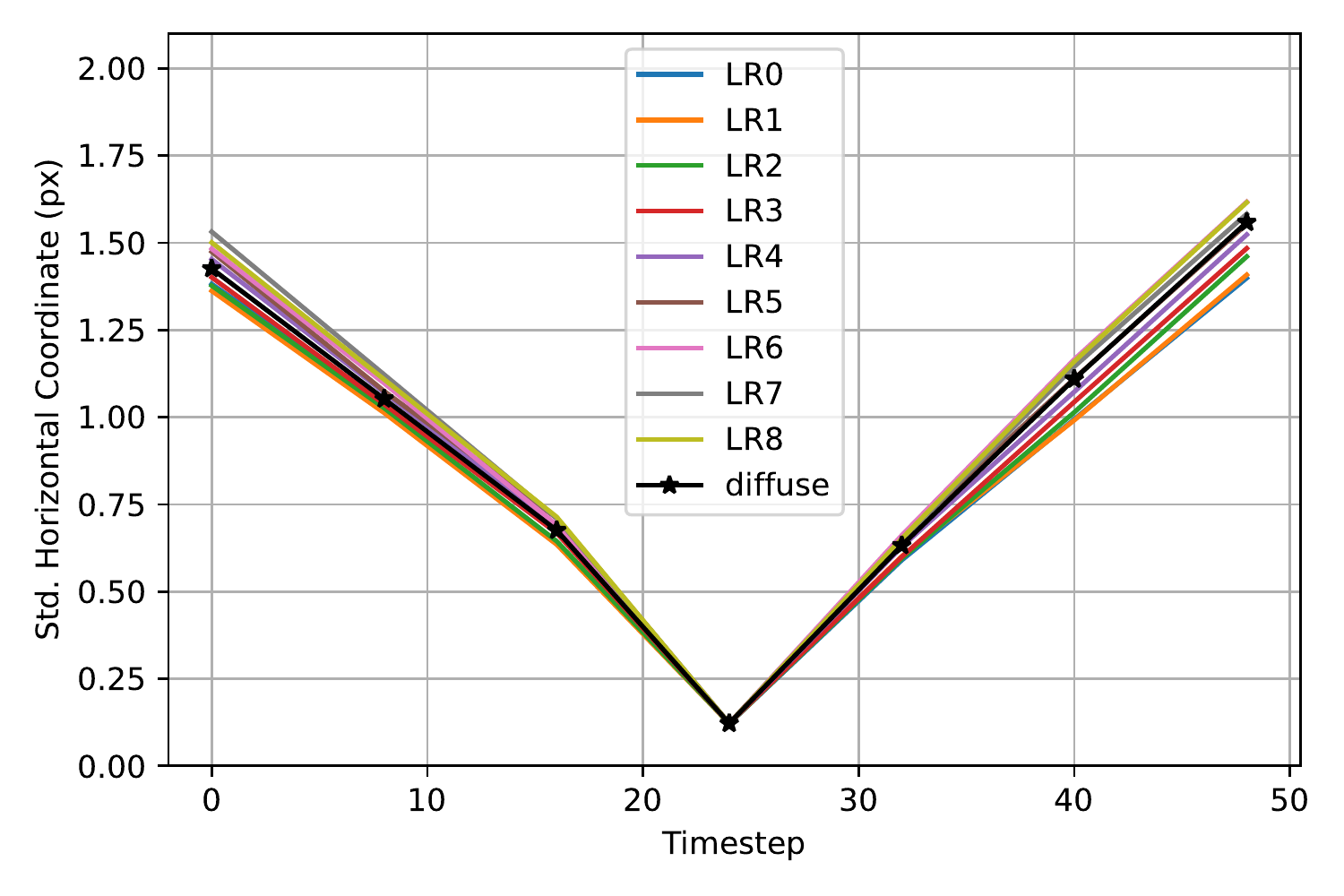}
        \includegraphics[width=0.48\textwidth]{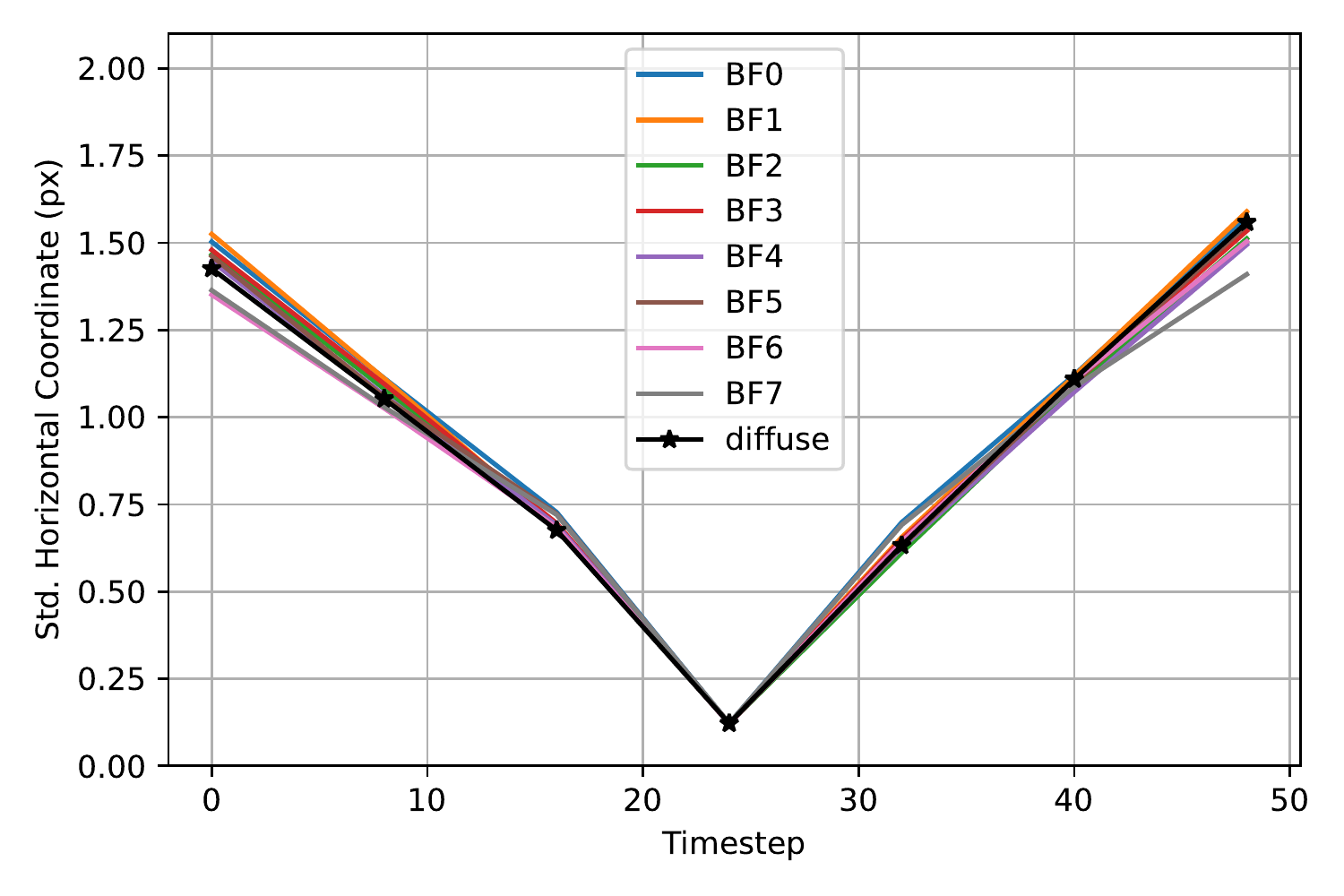}
    }
    \subfigure[Vertical Coordinate]{
        \includegraphics[width=0.48\textwidth]{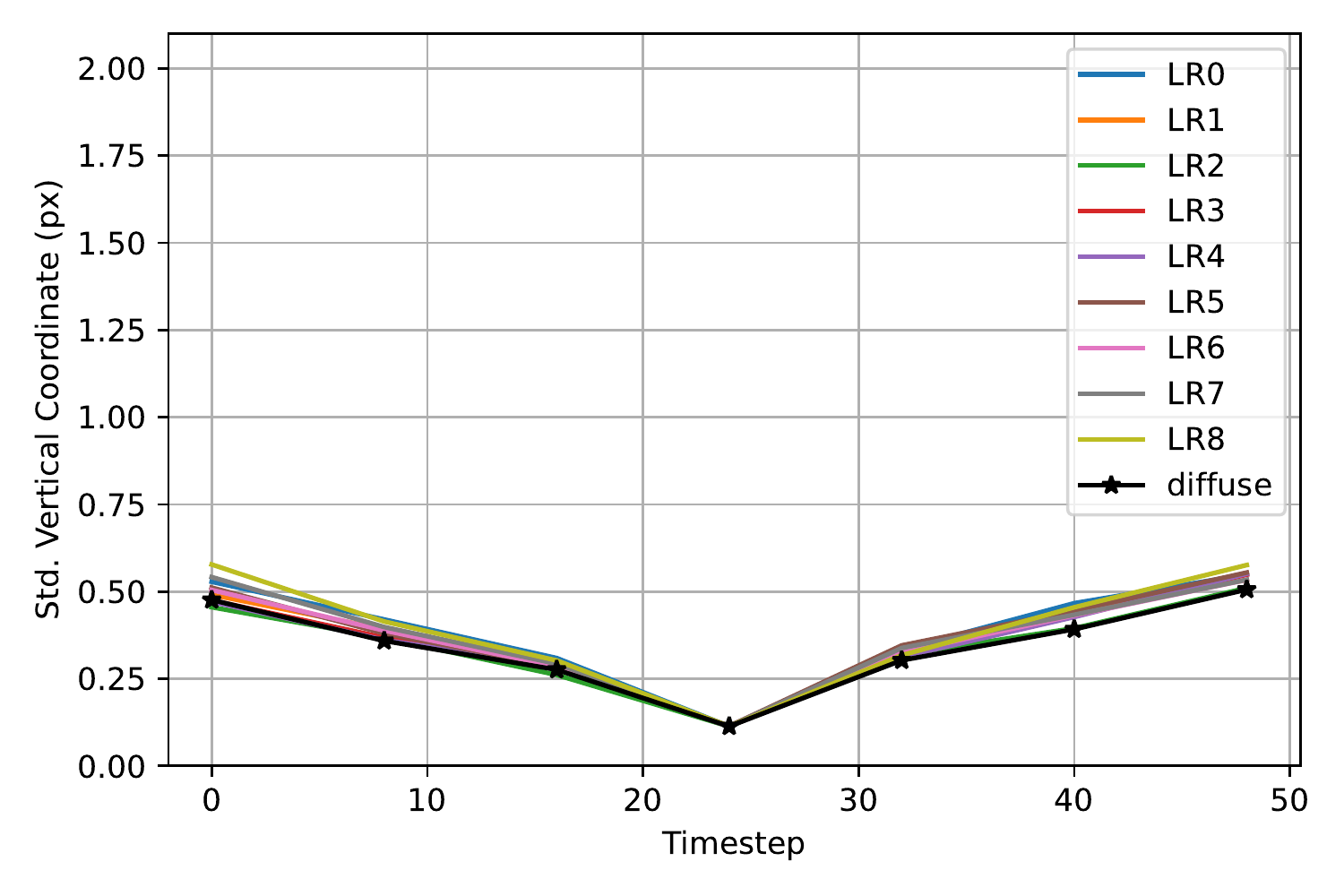}
        \includegraphics[width=0.48\textwidth]{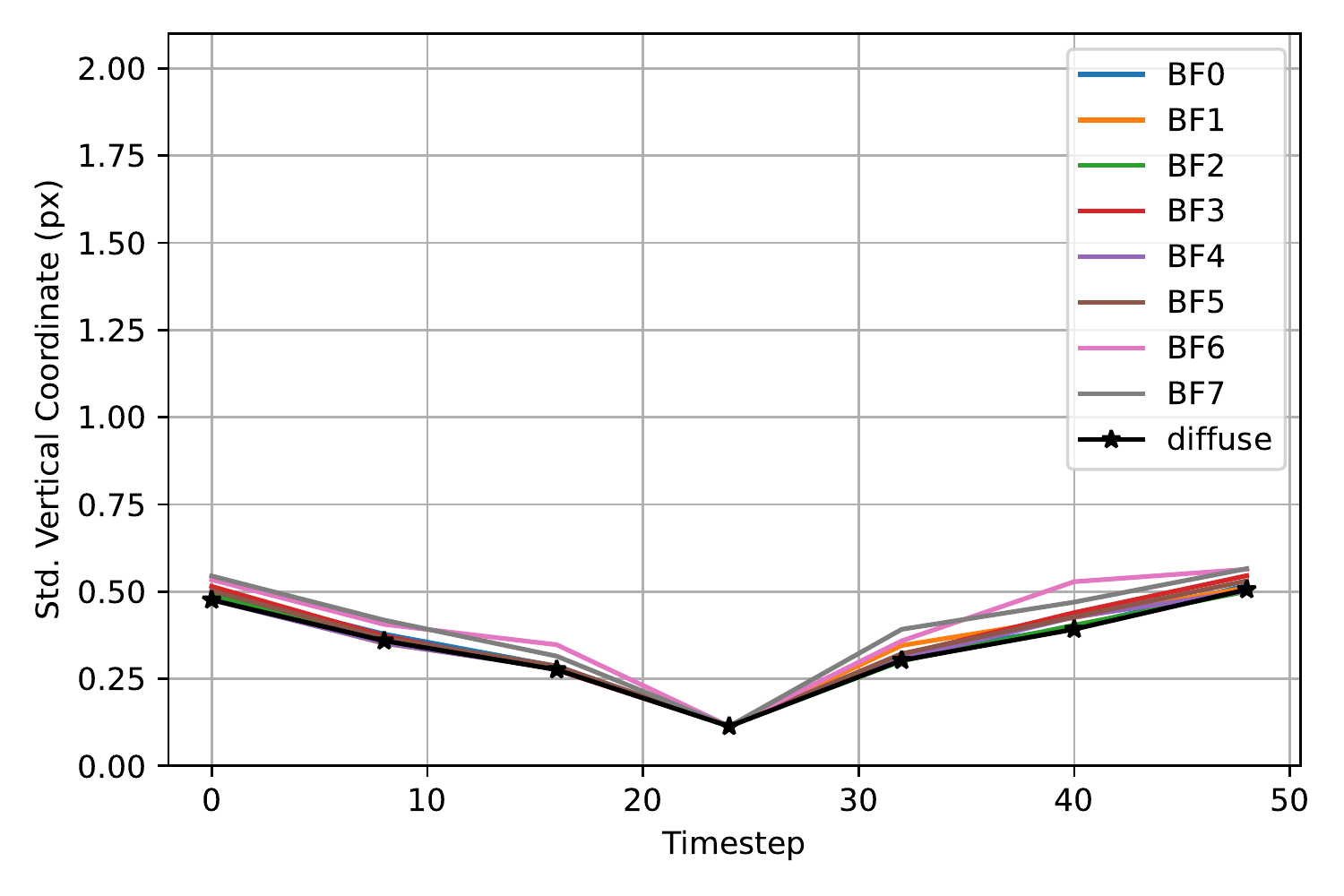}
    }
    \caption{\textbf{DTU Point Features Dataset: At eight times nominal speed, lighting condition does not change trends in covariance $\Sigma(t)$ when using the Lucas-Kanade Tracker.} We compute $\Sigma(t)$ using diffuse lighting (black lines) and each of the directional lighting conditions listed in Figure \ref{fig:dtu_light_stage} using all tracks from all 60 scenes. The variation of $\Sigma(t)$ due to the existence of directional lighting is at most 10 percent of the variation common to all plotted lines. The effect of directional lighting is relatively small because changes between adjacent frames are small whether or not the scene contains directional lighting.}
    \label{fig:dtu_LK_cov_speed8.00}
\end{figure}

\begin{figure}[H]
    \centering
    \subfigure[Horizontal Coordinate]{
        \includegraphics[width=0.48\textwidth]{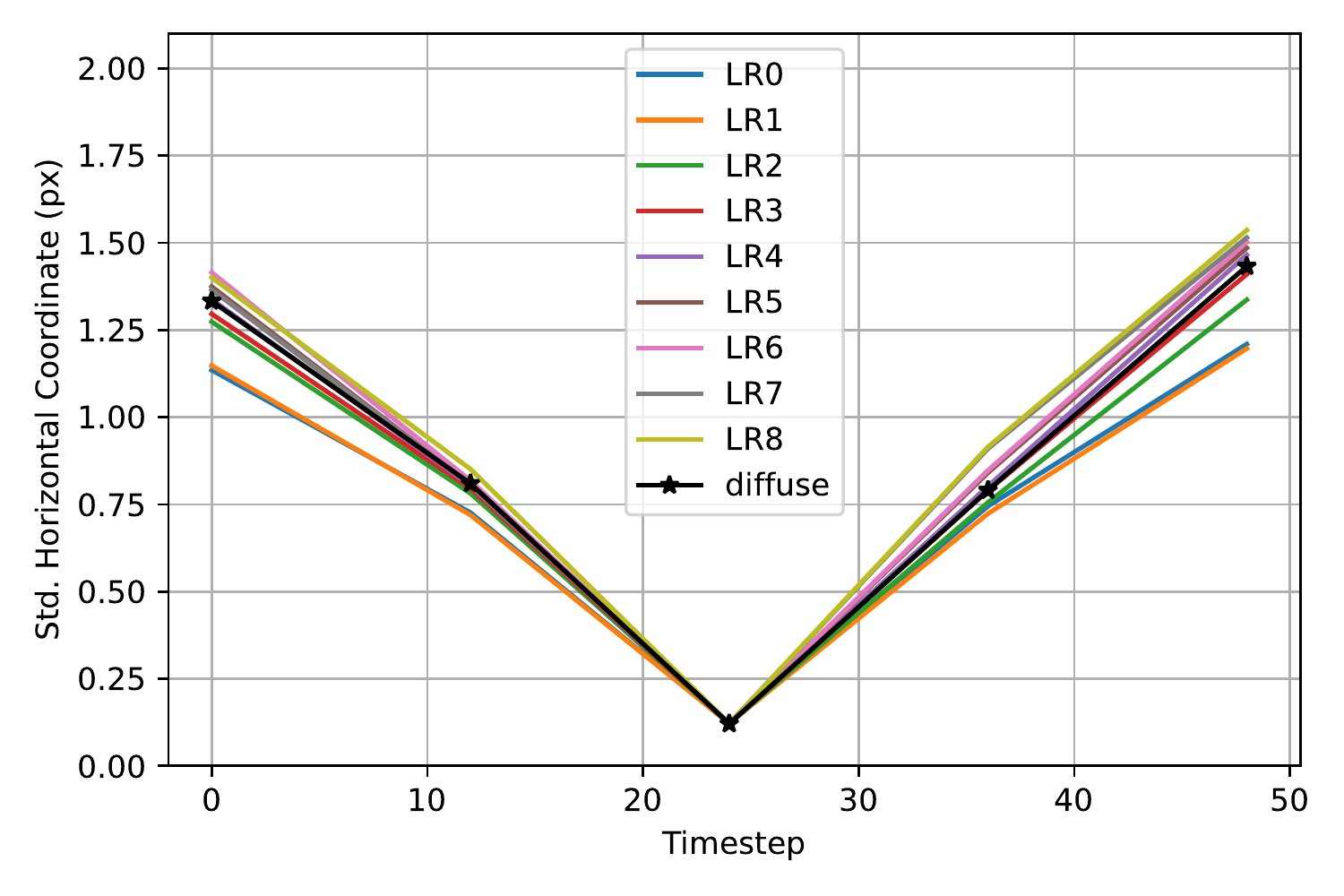}
        \includegraphics[width=0.48\textwidth]{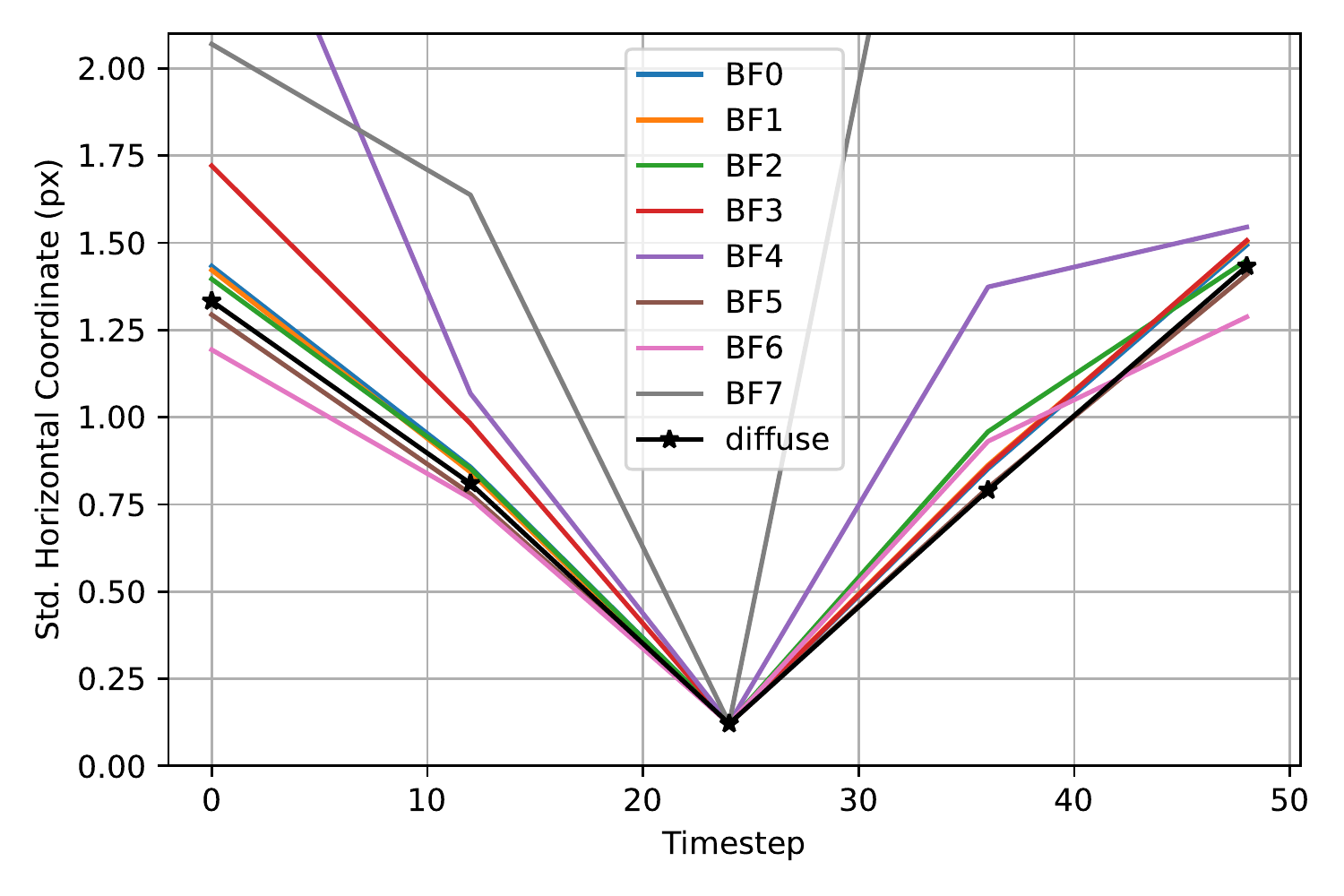}
    }
    \subfigure[Vertical Coordinate]{
        \includegraphics[width=0.48\textwidth]{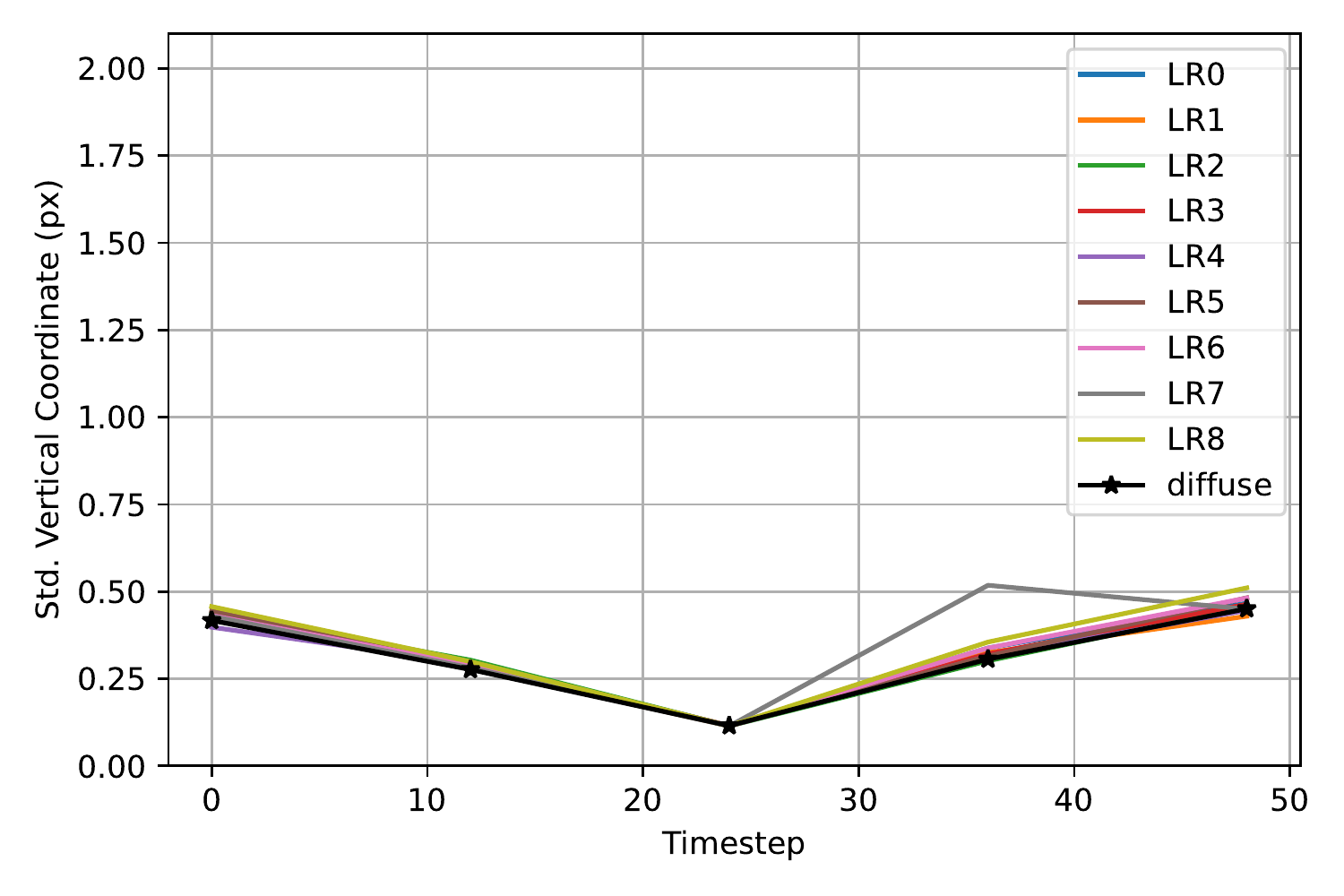}
        \includegraphics[width=0.48\textwidth]{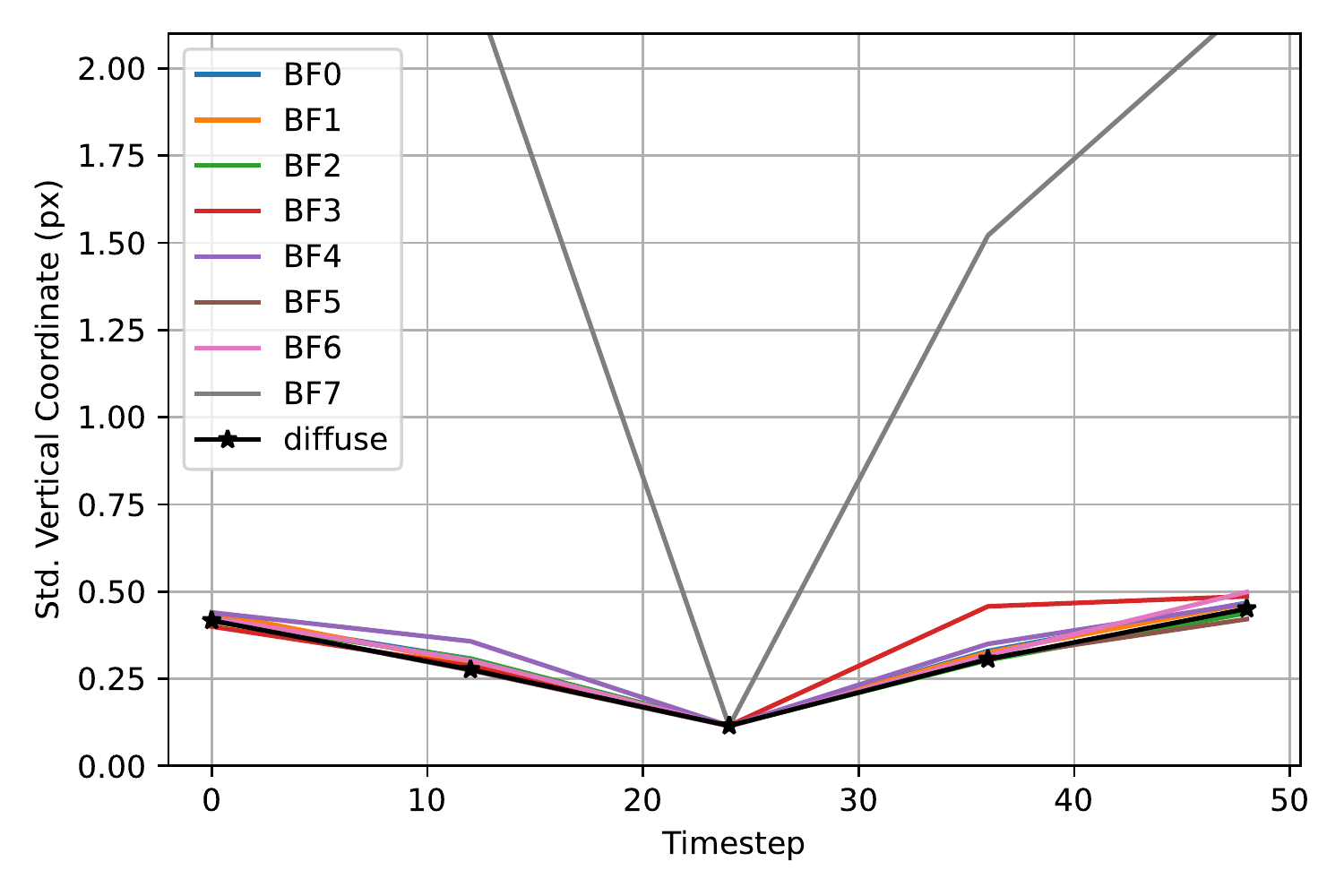}
    }
    \caption{\textbf{DTU Point Features Dataset: At twelve times nominal speed, lighting condition does not change trends in covariance $\Sigma(t)$ when using the Lucas-Kanade Tracker.} We compute $\Sigma(t)$ using diffuse lighting (black lines) and each of the directional lighting conditions listed in Figure \ref{fig:dtu_light_stage} using all tracks from all 60 scenes. 
    At twelve times nominal speed, tracking failures cause large covariances to appear for some lighting conditions. Otherwise, the variation of $\Sigma(t)$ due to the existence of directional lighting is at most 10 percent of the variation common to all plotted lines. The effect of directional lighting is relatively small because changes between adjacent frames are small whether or not the scene contains directional lighting.
    }
    \label{fig:dtu_LK_cov_speed12.00}
\end{figure}

\begin{figure}[H]
    \centering
    \subfigure[Horizontal Coordinate]{
        \includegraphics[width=0.48\textwidth]{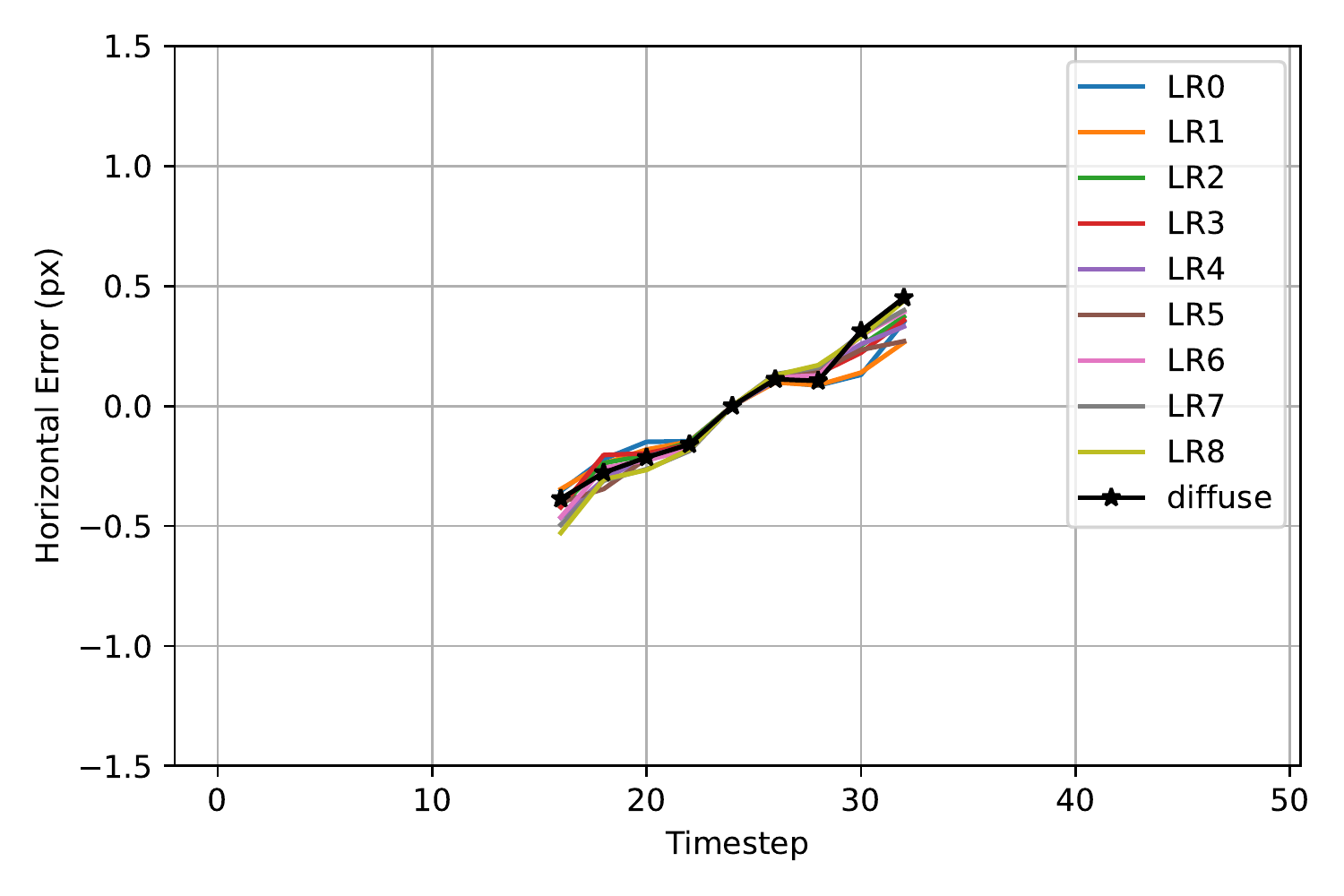}
        \includegraphics[width=0.48\textwidth]{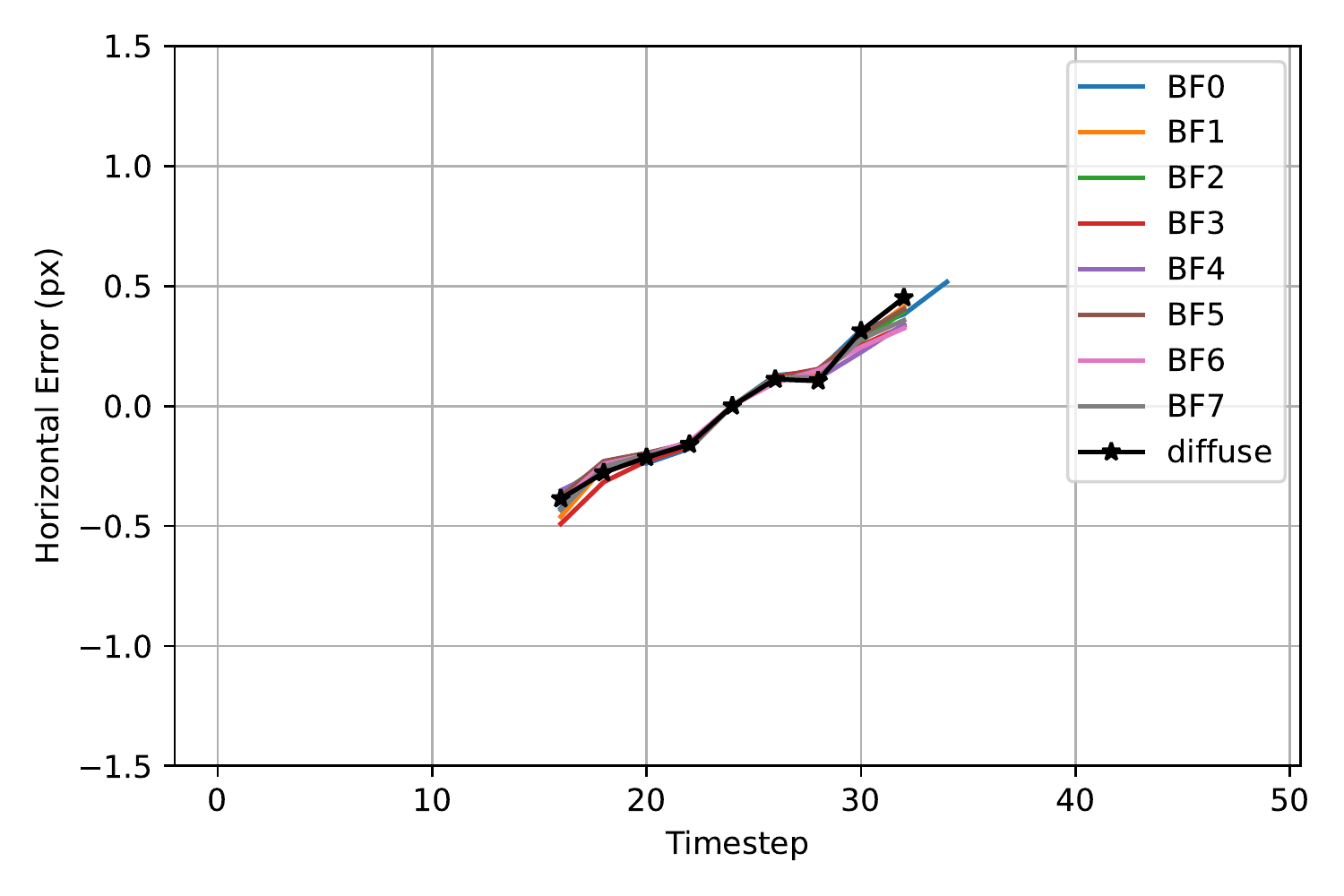}
    }
    \subfigure[Vertical Coordinate]{
        \includegraphics[width=0.48\textwidth]{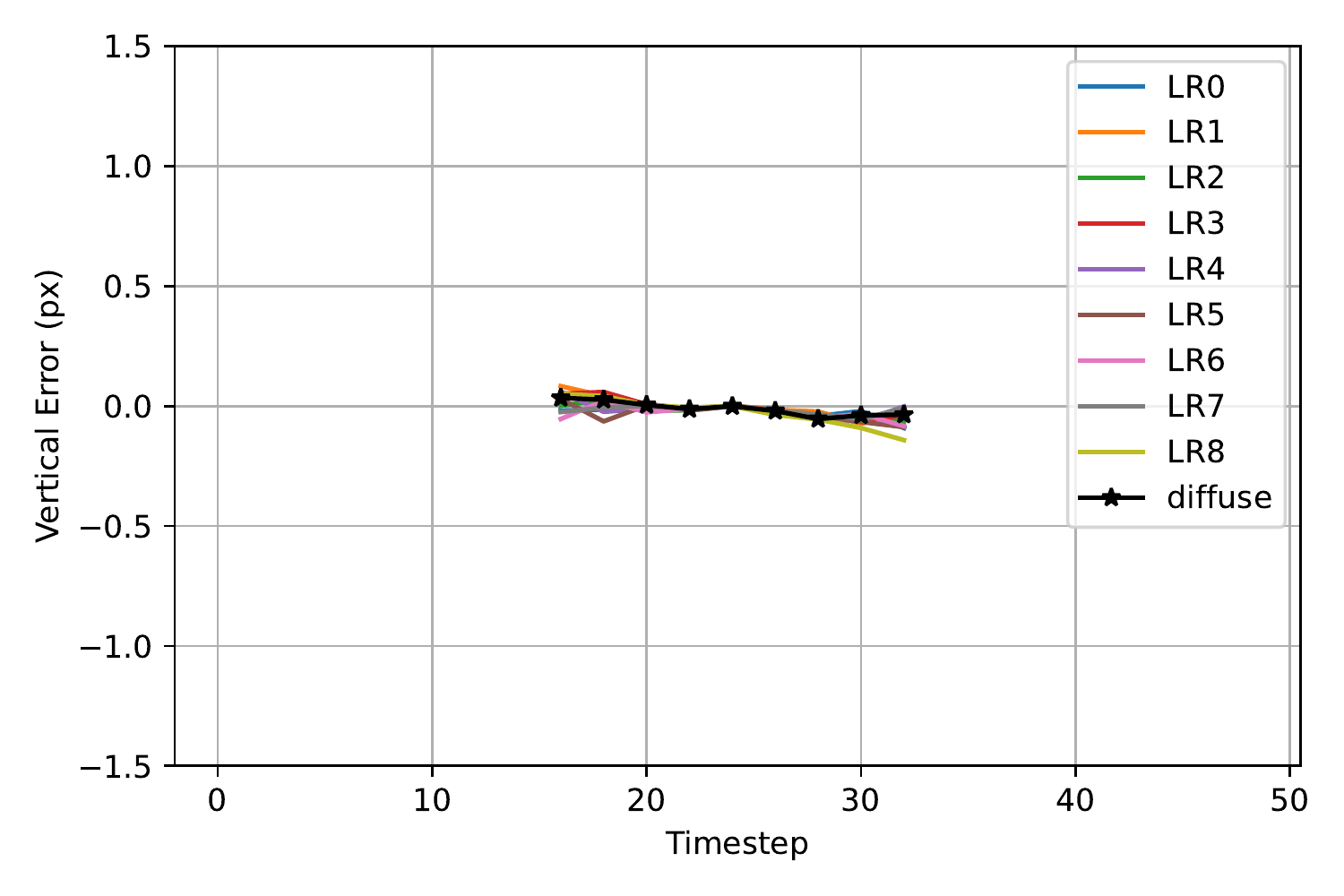}
        \includegraphics[width=0.48\textwidth]{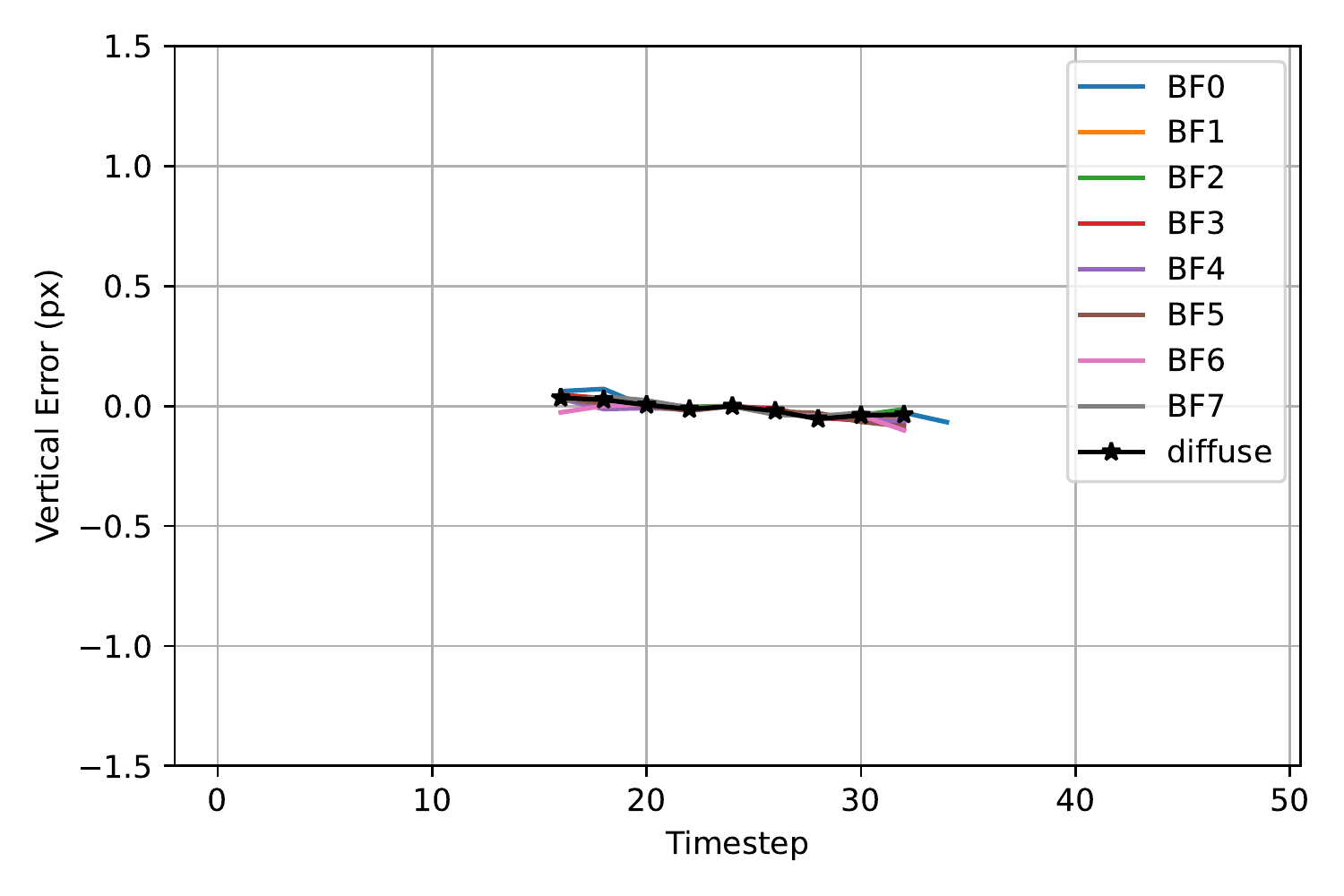}
    }
    \caption{\textbf{DTU Point Features Dataset: At twice nominal speed, lighting condition does not change trends in mean error $\mu(t)$ when using the Correspondence Tracker.} We compute $\mu(t)$ at each timestep using diffuse lighting (black lines) and each of the directional lighting conditions listed in Figure \ref{fig:dtu_light_stage} using all tracks from all 60 scenes. Lines are limited to timesteps containing at least 100 features. The variation of $\mu(t)$ due to the existence of directional lighting is at most 10 percent of the variation common to all plotted lines. The effect of directional lighting is relatively small because changes between adjacent frames are small whether or not the scene contains directional lighting.}
    \label{dtu_match_mu_speed2.00}
\end{figure}

\begin{figure}[H]
    \centering
    \subfigure[Horizontal Coordinate]{
        \includegraphics[width=0.48\textwidth]{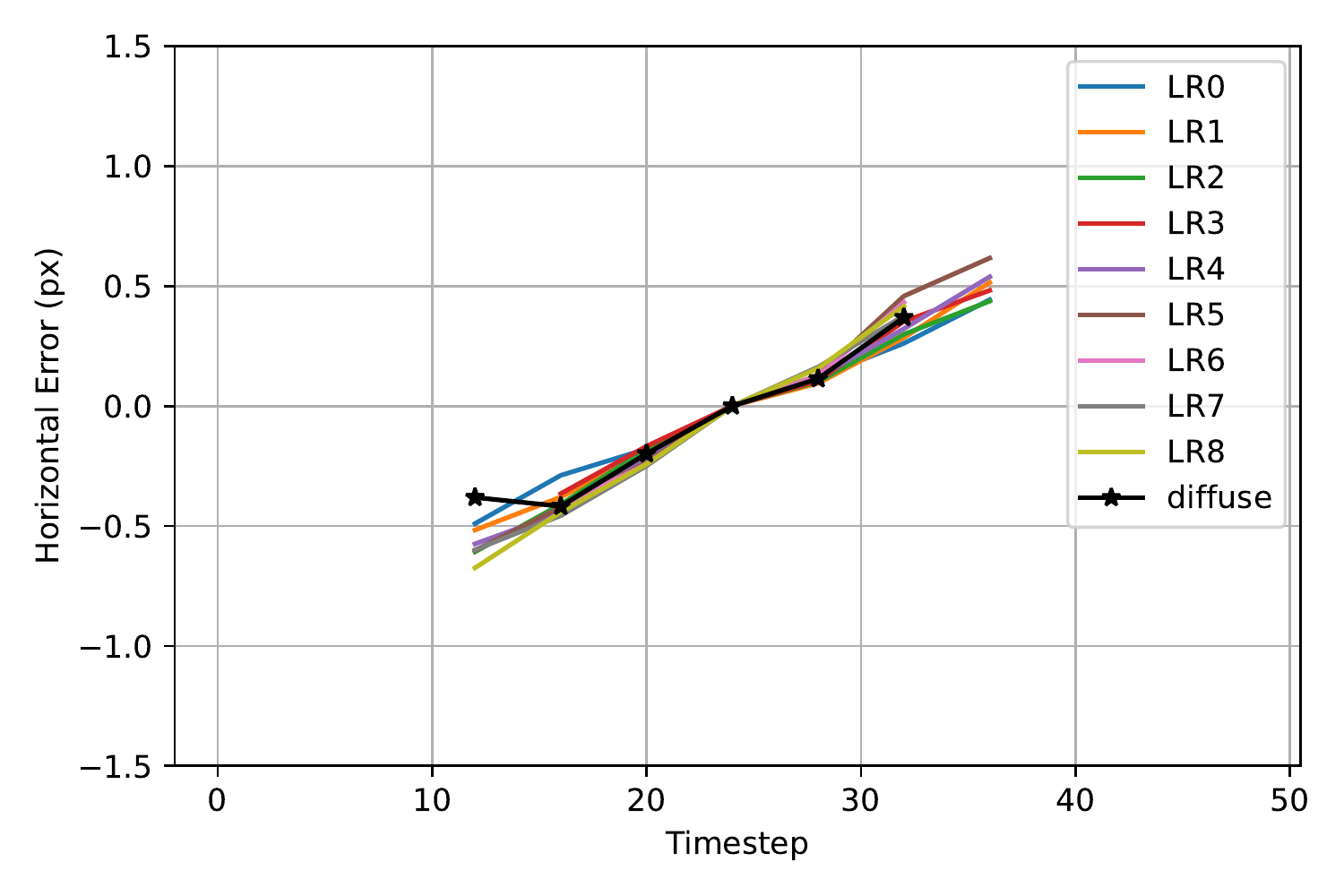}
        \includegraphics[width=0.48\textwidth]{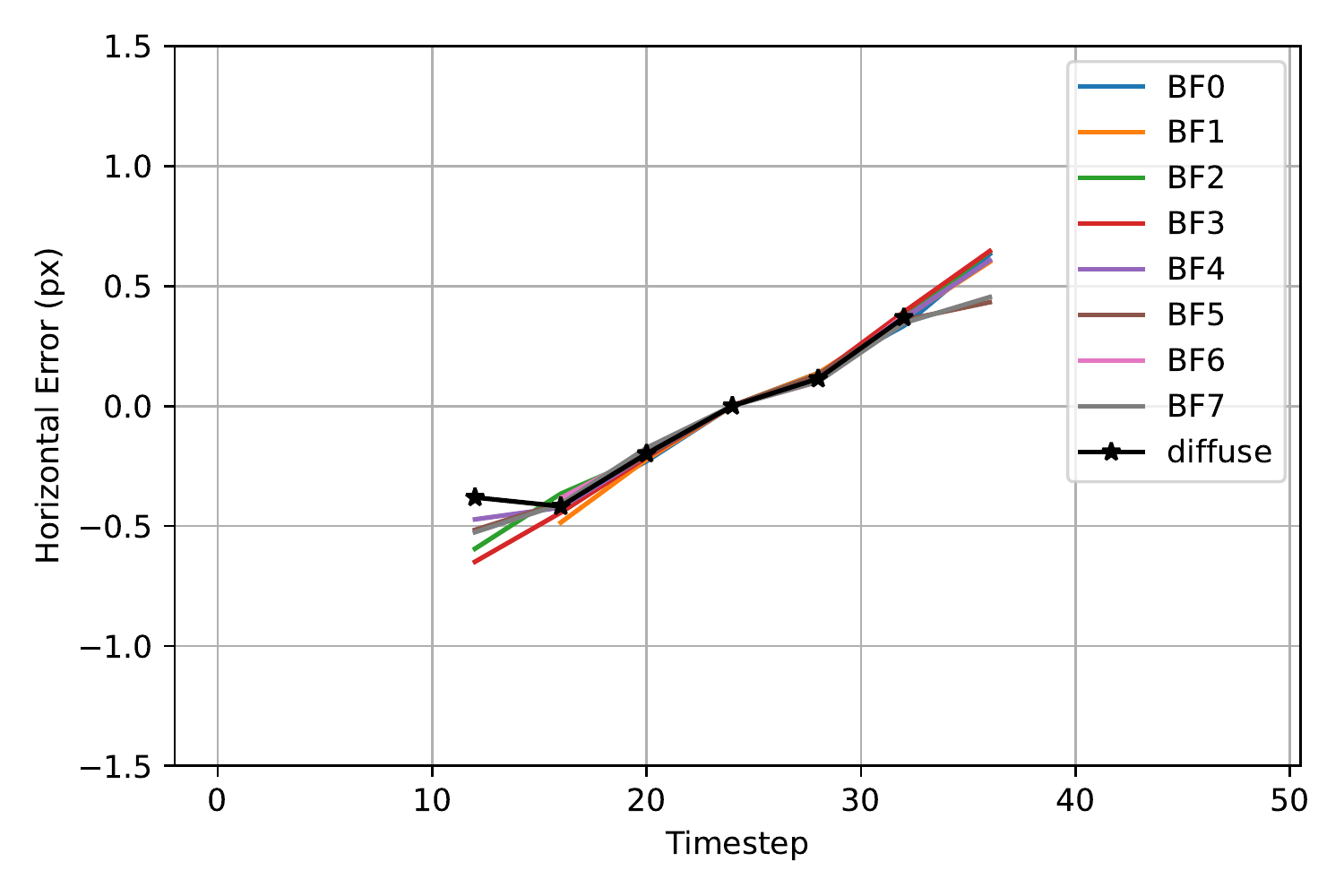}
    }
    \subfigure[Vertical Coordinate]{
        \includegraphics[width=0.48\textwidth]{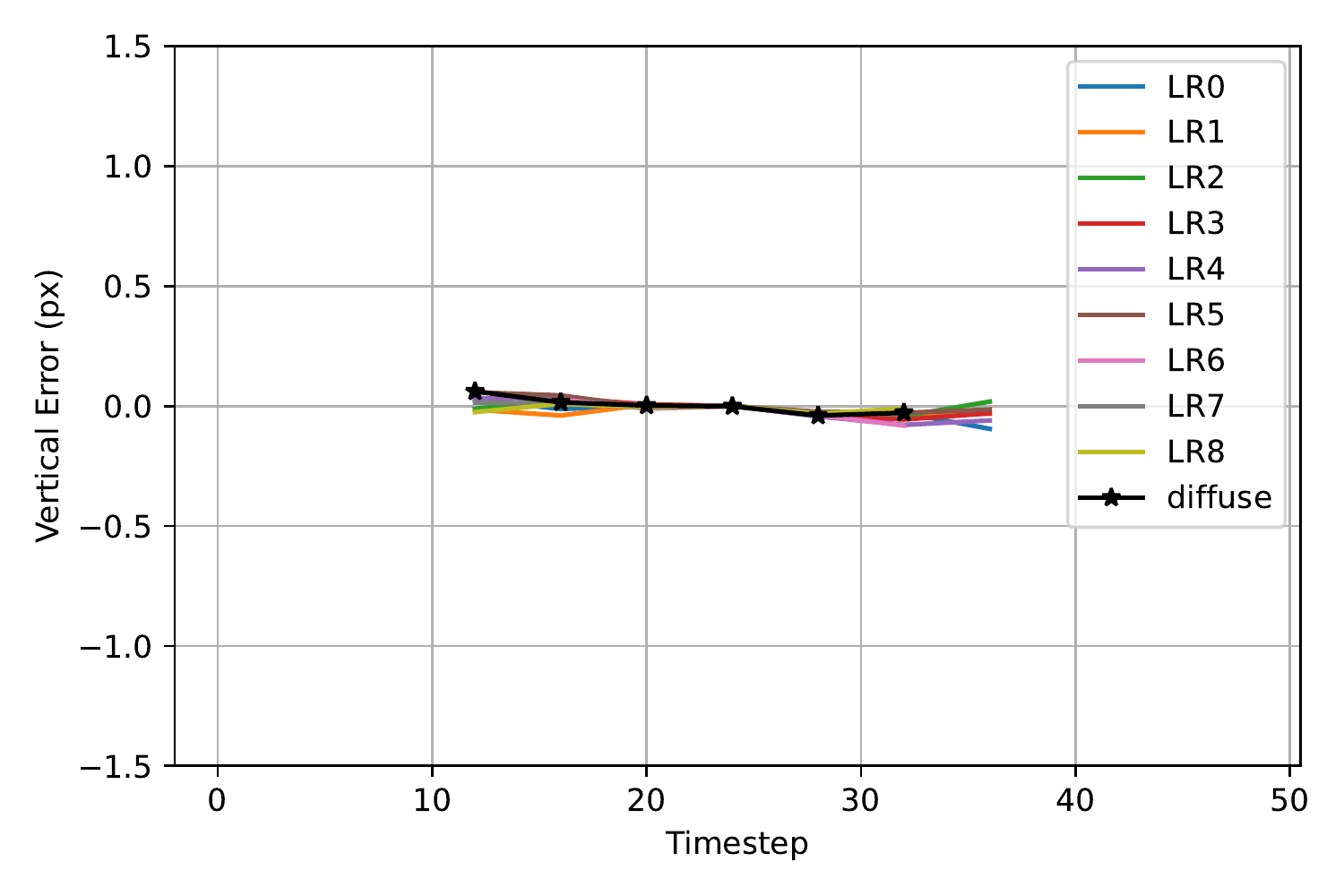}
        \includegraphics[width=0.48\textwidth]{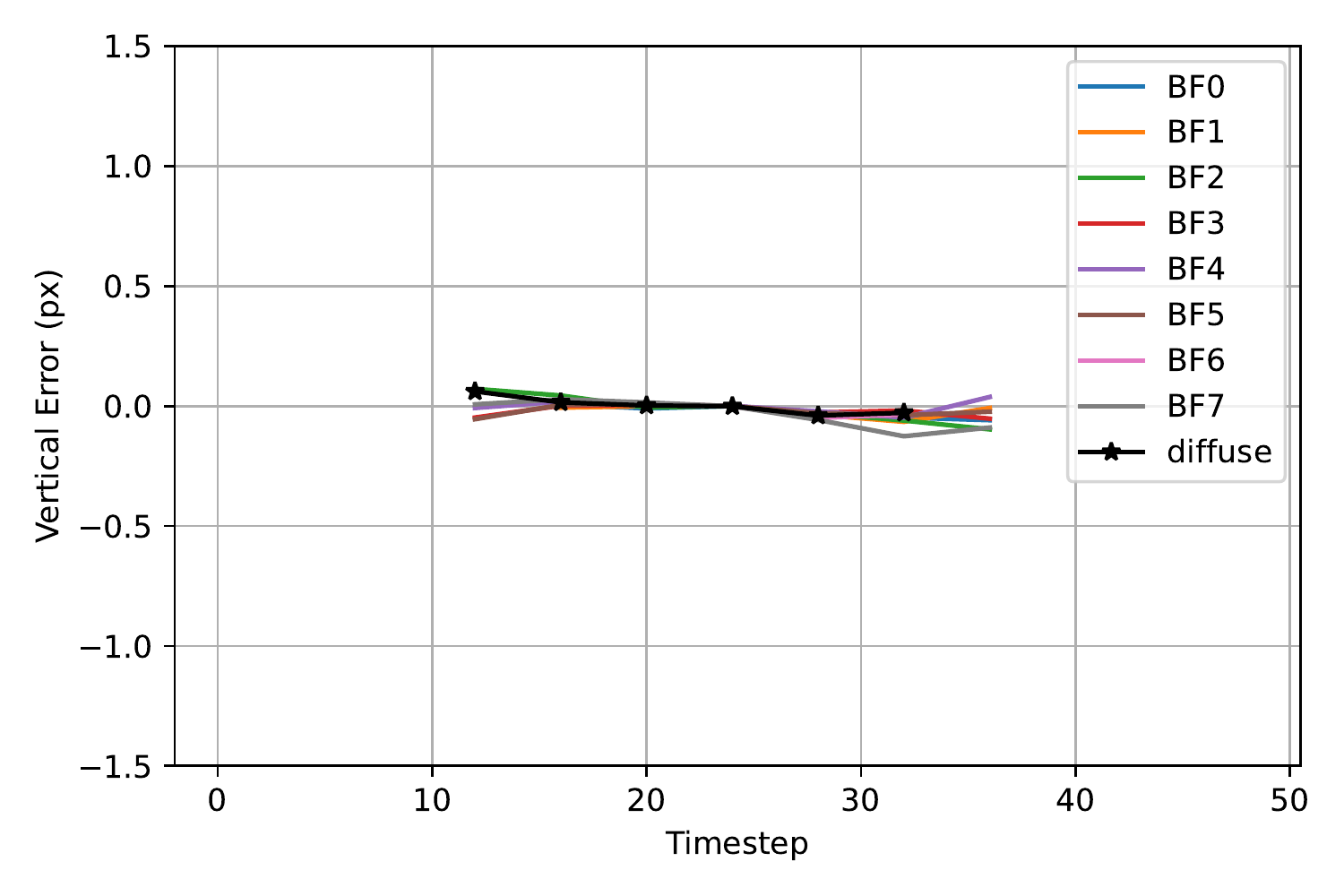}
    }
    \caption{\textbf{DTU Point Features Dataset: At four times nominal speed, lighting condition does not change trends in mean error $\mu(t)$ when using the Correspondence Tracker.} We compute $\mu(t)$ at each timestep using diffuse lighting (black lines) and each of the directional lighting conditions listed in Figure \ref{fig:dtu_light_stage} using all tracks from all 60 scenes. Lines are limited to timesteps containing at least 100 features. The variation of $\mu(t)$ due to the existence of directional lighting is at most 10 percent of the variation common to all plotted lines. The effect of directional lighting is relatively small because changes between adjacent frames are small whether or not the scene contains directional lighting.}
    \label{dtu_match_mu_speed4.00}
\end{figure}

\begin{figure}[H]
    \centering
    \subfigure[Horizontal Coordinate]{
        \includegraphics[width=0.48\textwidth]{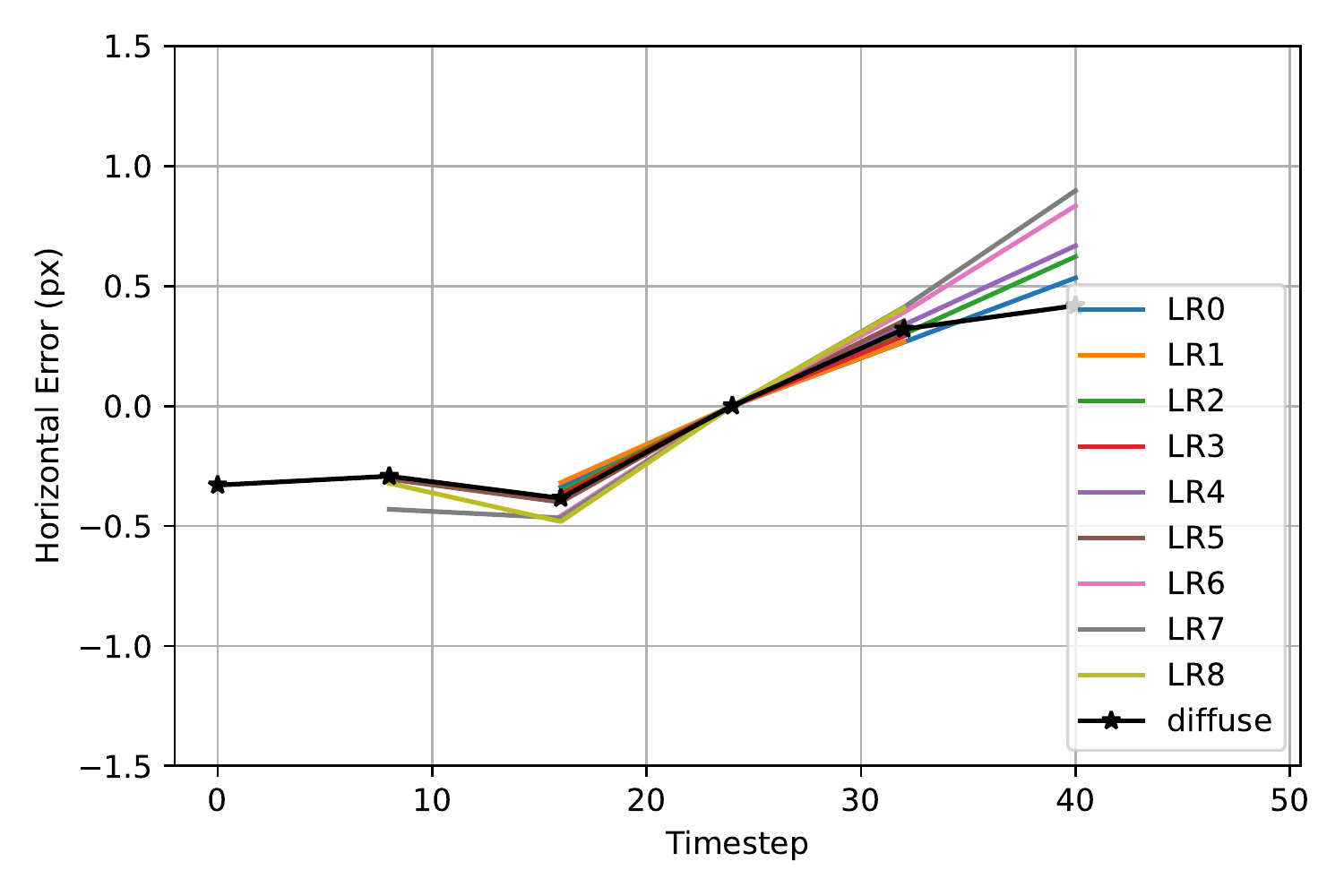}
        \includegraphics[width=0.48\textwidth]{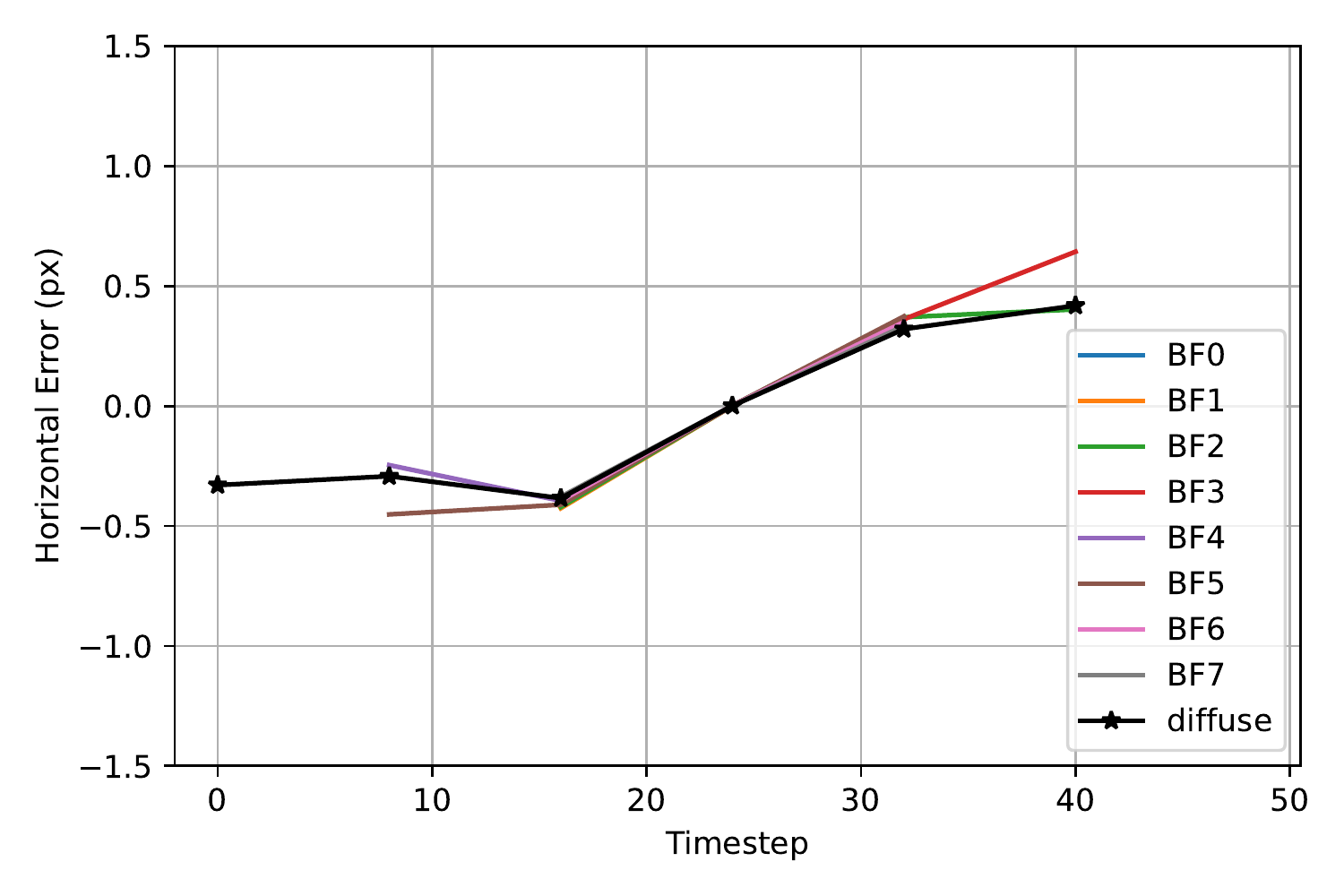}
    }
    \subfigure[Vertical Coordinate]{
        \includegraphics[width=0.48\textwidth]{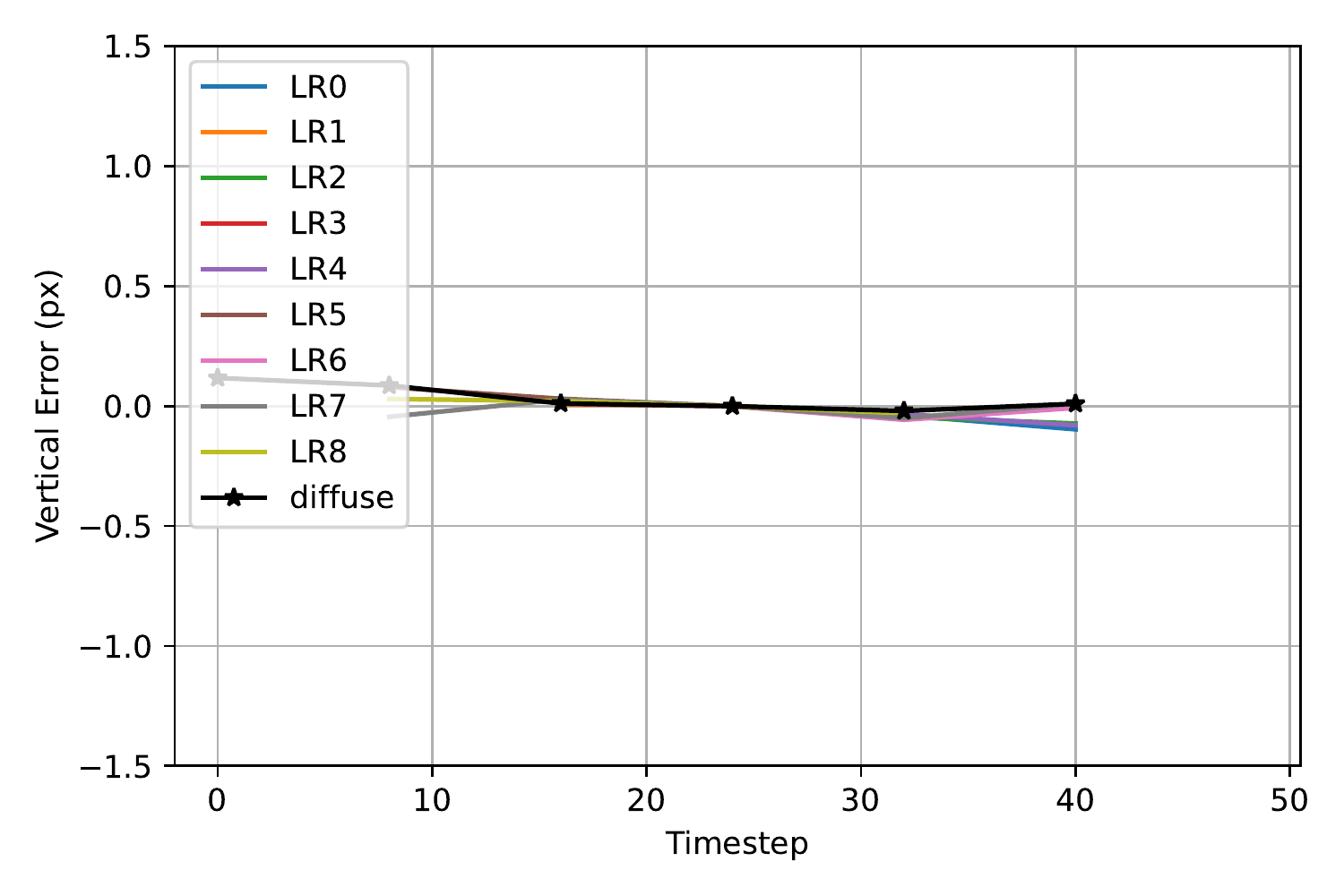}
        \includegraphics[width=0.48\textwidth]{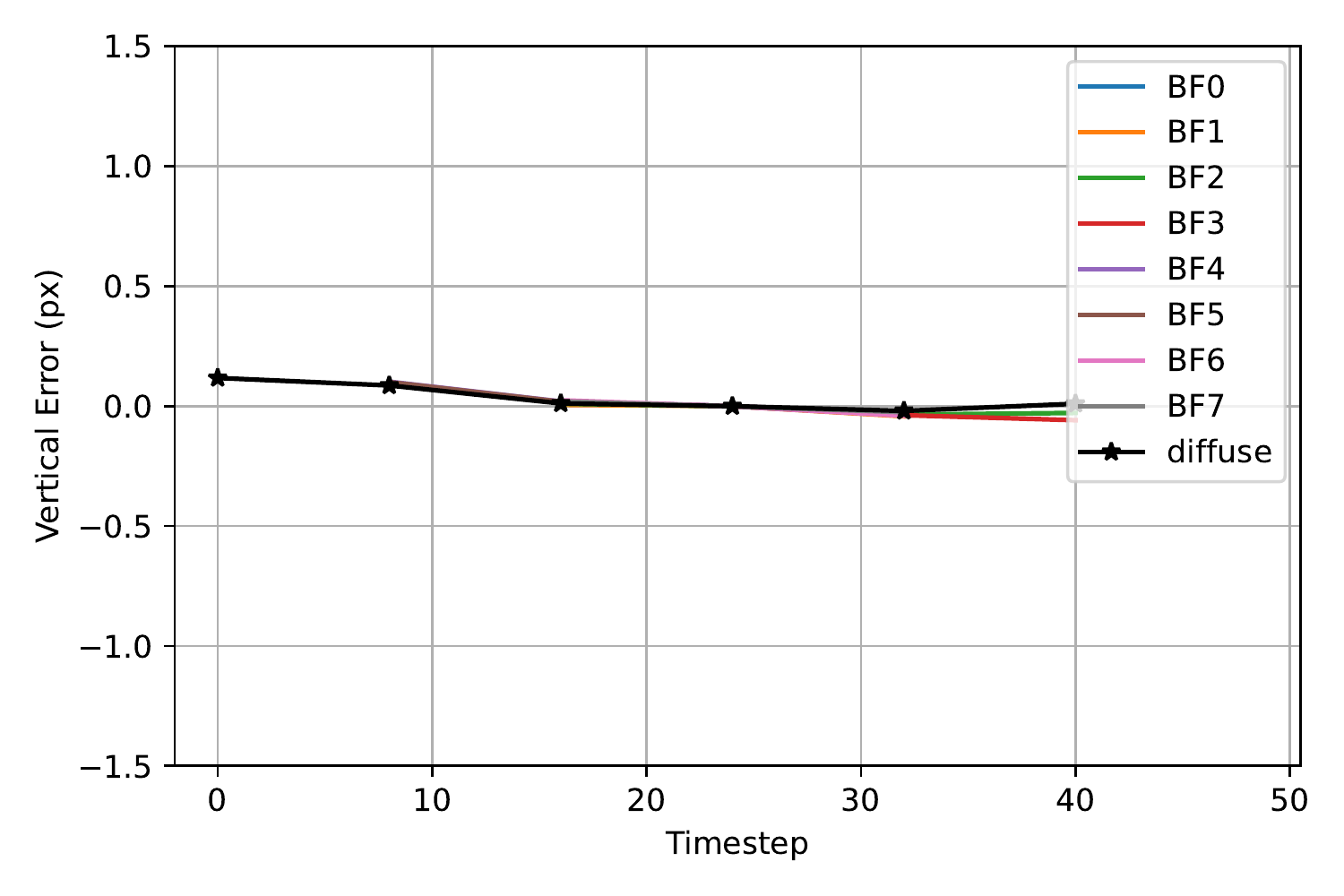}
    }
    \caption{\textbf{DTU Point Features Dataset: At eight times nominal speed, lighting condition does not change trends in mean error $\mu(t)$  when using the Correspondence Tracker.} We compute $\mu(t)$ at each timestep using diffuse lighting (black lines) and each of the directional lighting conditions listed in Figure \ref{fig:dtu_light_stage} using all tracks from all 60 scenes. Lines are limited to timesteps containing at least 100 features. The variation of $\mu(t)$ due to the existence of directional lighting is smaller than the variation common to all plotted lines. The effect of directional lighting is relatively small because changes between adjacent frames are small whether or not the scene contains directional lighting.}
    \label{dtu_match_mu_speed8.00}
\end{figure}

\begin{figure}[H]
    \centering
    \subfigure[Horizontal Coordinate]{
        \includegraphics[width=0.48\textwidth]{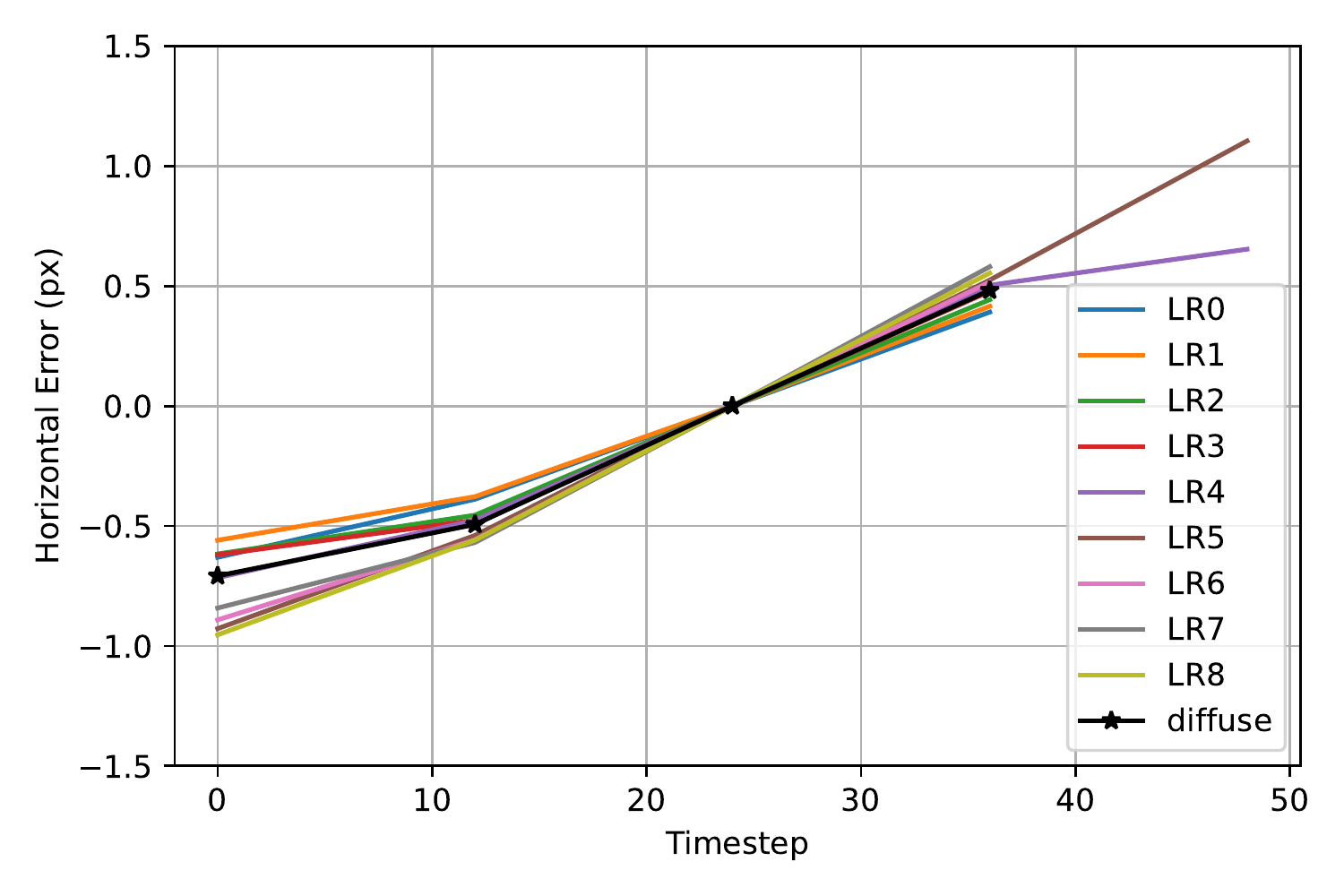}
        \includegraphics[width=0.48\textwidth]{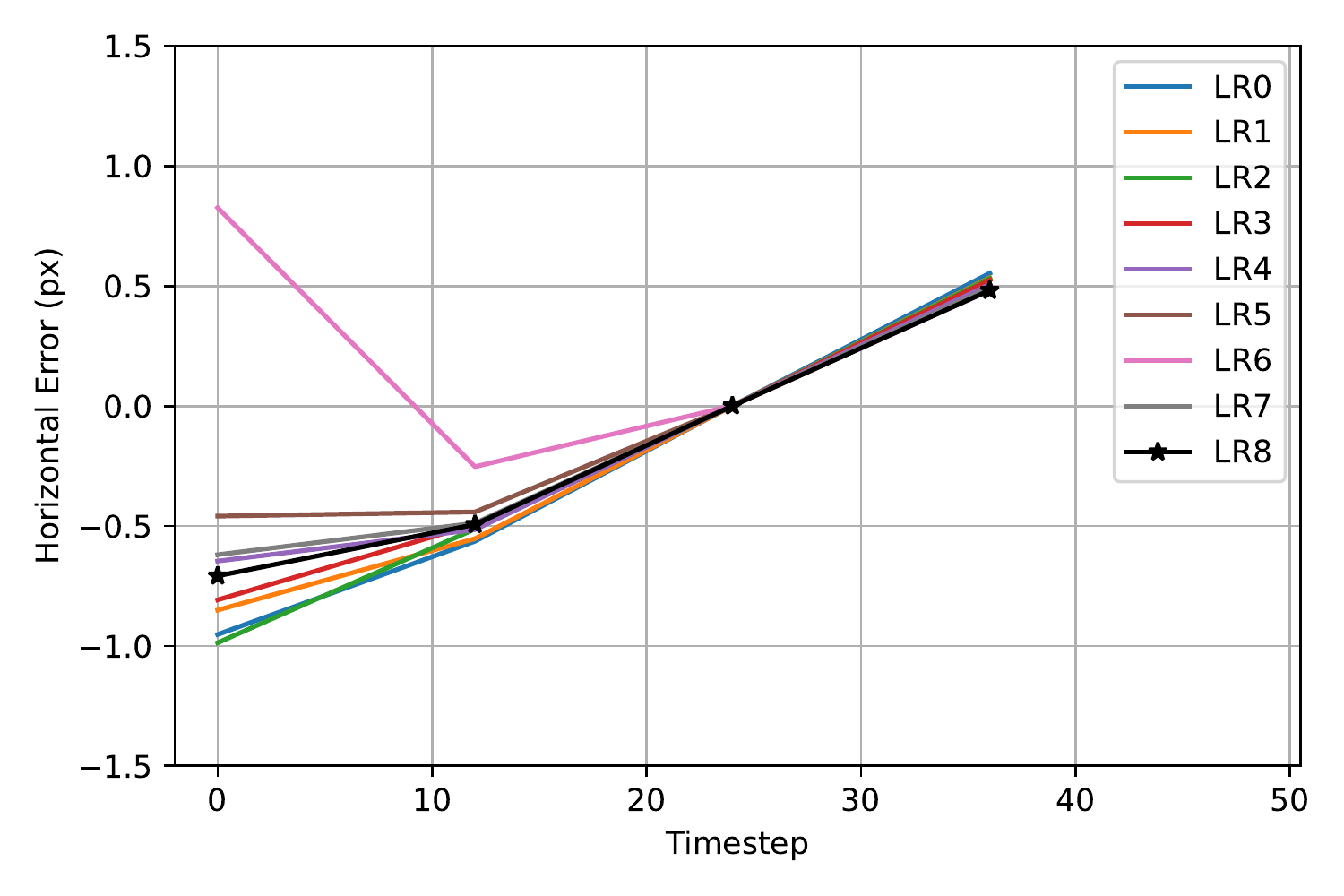}
    }
    \subfigure[Vertical Coordinate]{
        \includegraphics[width=0.48\textwidth]{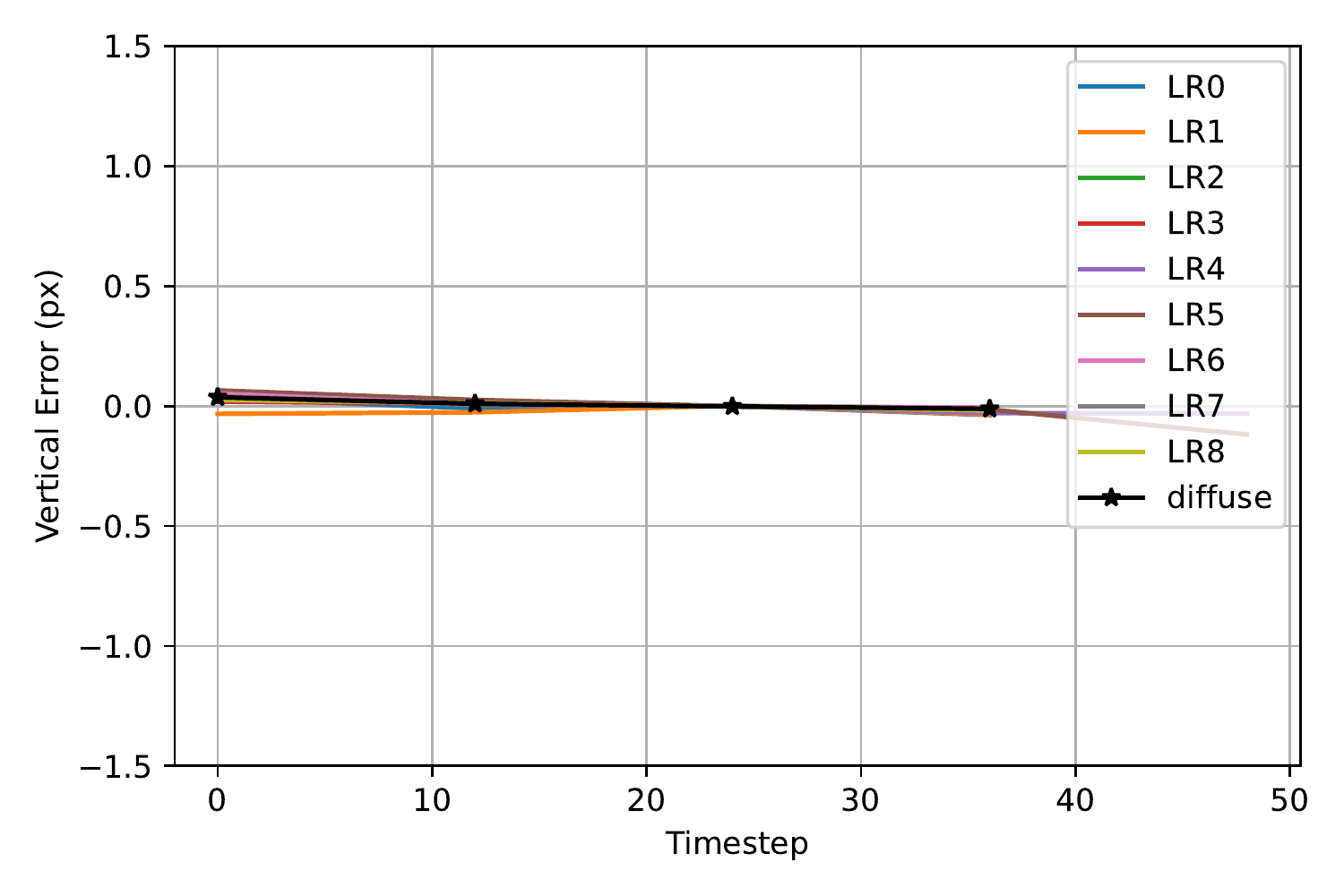}
        \includegraphics[width=0.48\textwidth]{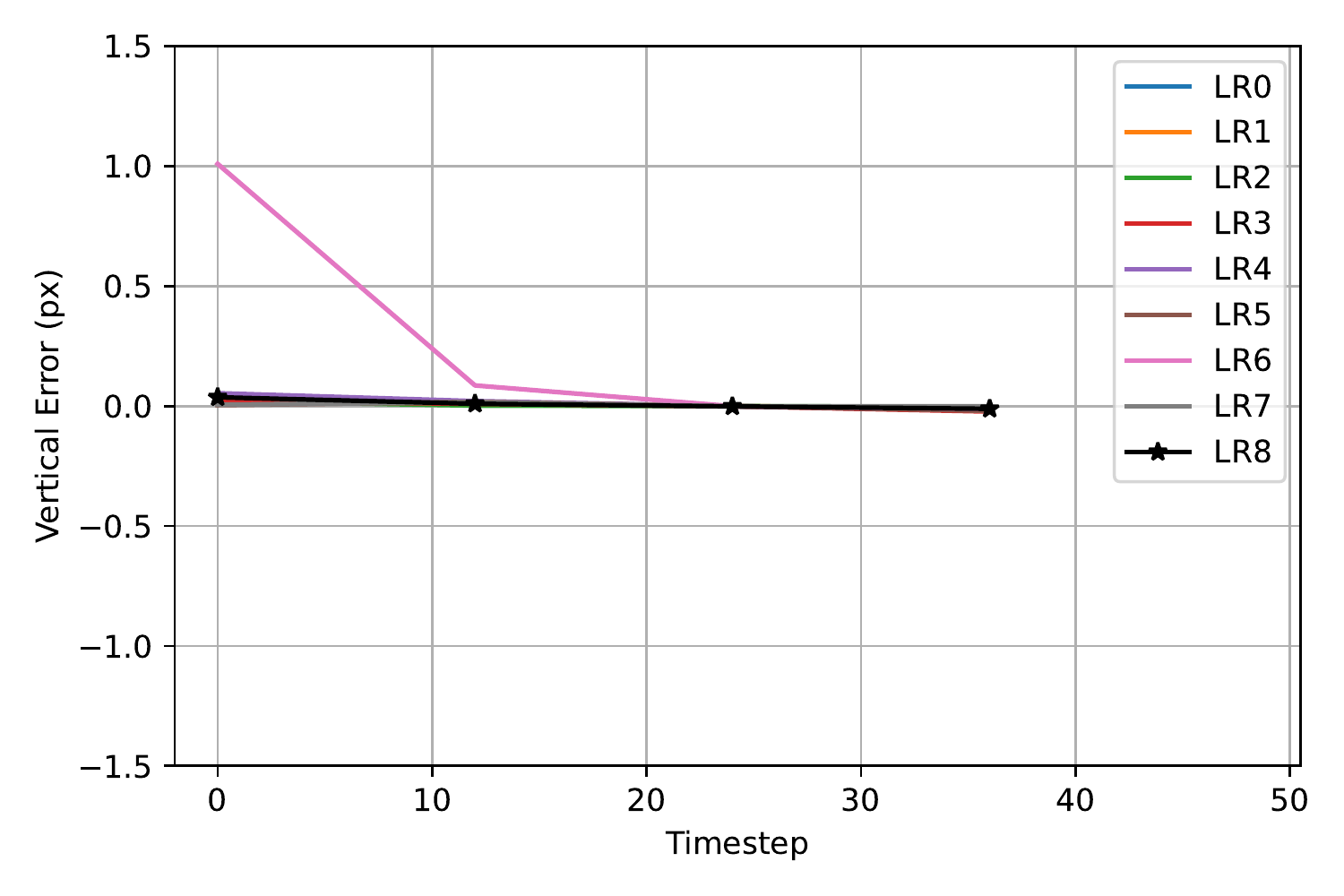}
    } 
    \caption{\textbf{DTU Point Features Dataset: At twelve times nominal speed, lighting condition does not change trends in mean error $\mu(t)$  when using the Correspondence Tracker.} We compute $\mu(t)$ at each timestep using diffuse lighting (black lines) and each of the directional lighting conditions listed in Figure \ref{fig:dtu_light_stage} using all tracks from all 60 scenes. Lines are limited to timesteps containing at least 100 features. 
    With the exception of one lighting condition, the variation of $\mu(t)$ due to the existence of directional lighting is at most 10 percent of the variation common to all plotted lines. The effect of directional lighting is relatively small because changes between adjacent frames are small whether or not the scene contains directional lighting. 
    The large variation in lighting condition LR6 is caused by tracking failures. }
    \label{fig:dtu_match_mu_speed12.00}
\end{figure}

\begin{figure}[H]
    \centering
    \subfigure[Horizontal Coordinate]{
        \includegraphics[width=0.48\textwidth]{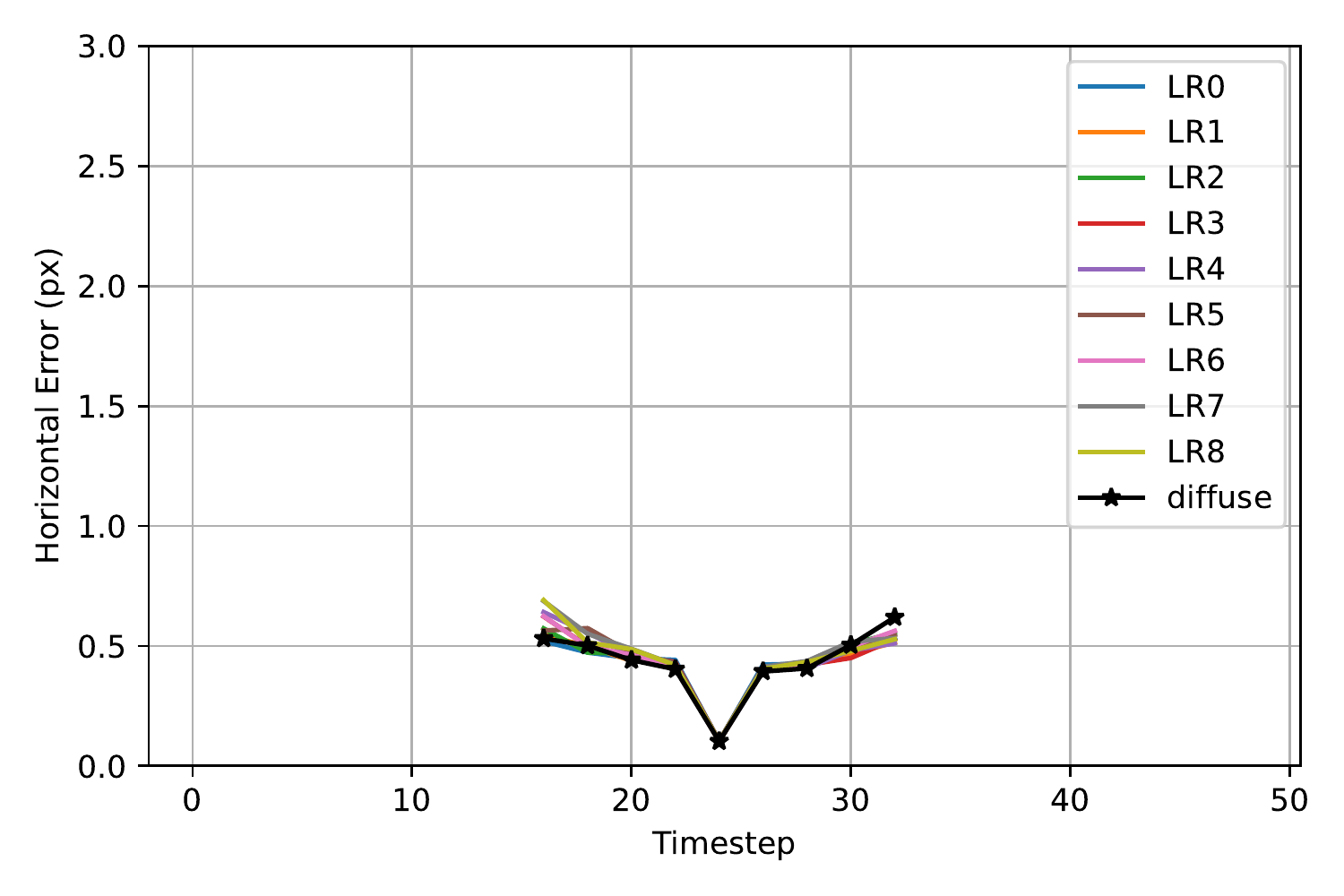}
        \includegraphics[width=0.48\textwidth]{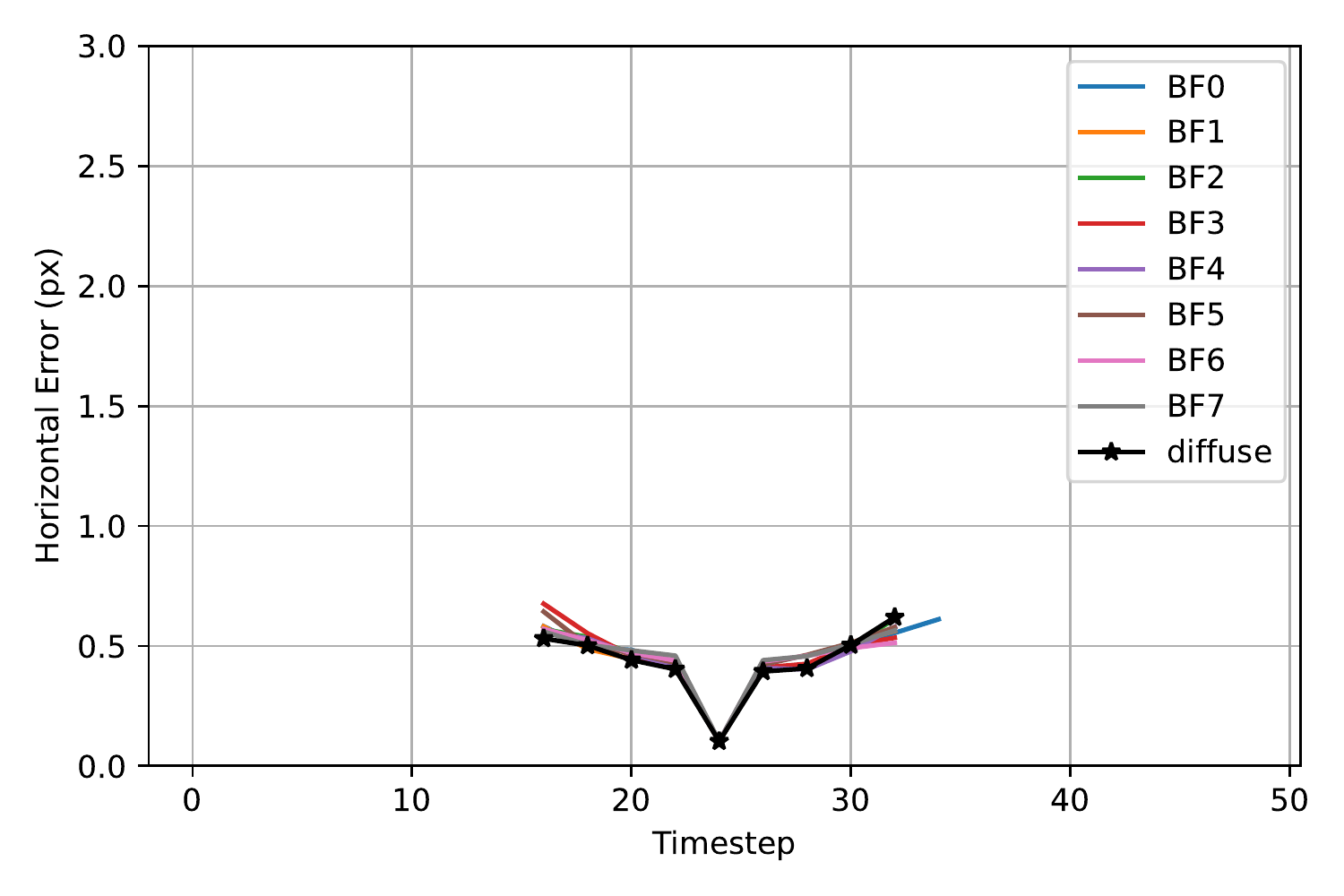}
    }
    \subfigure[Vertical Coordinate]{
        \includegraphics[width=0.48\textwidth]{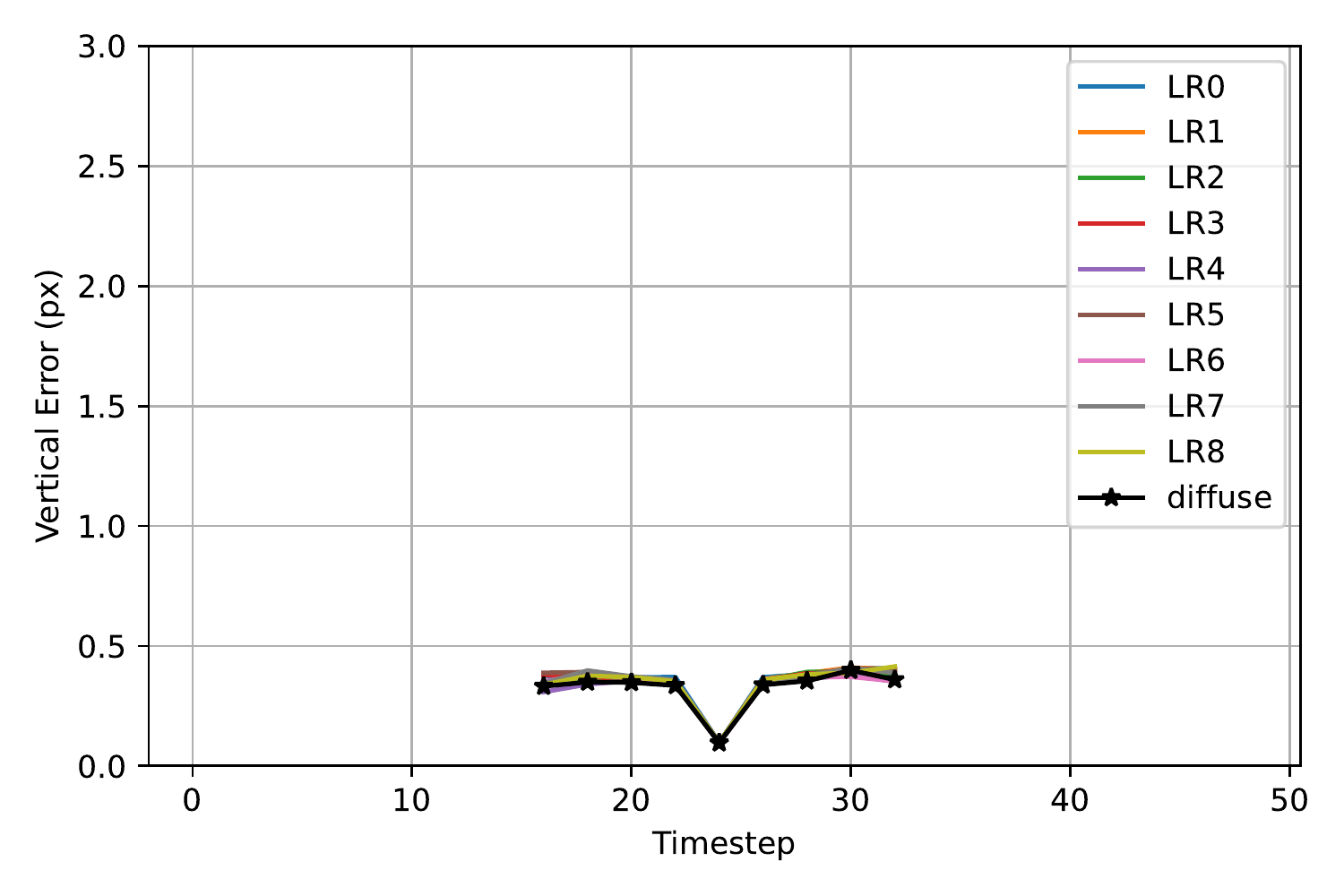}
        \includegraphics[width=0.48\textwidth]{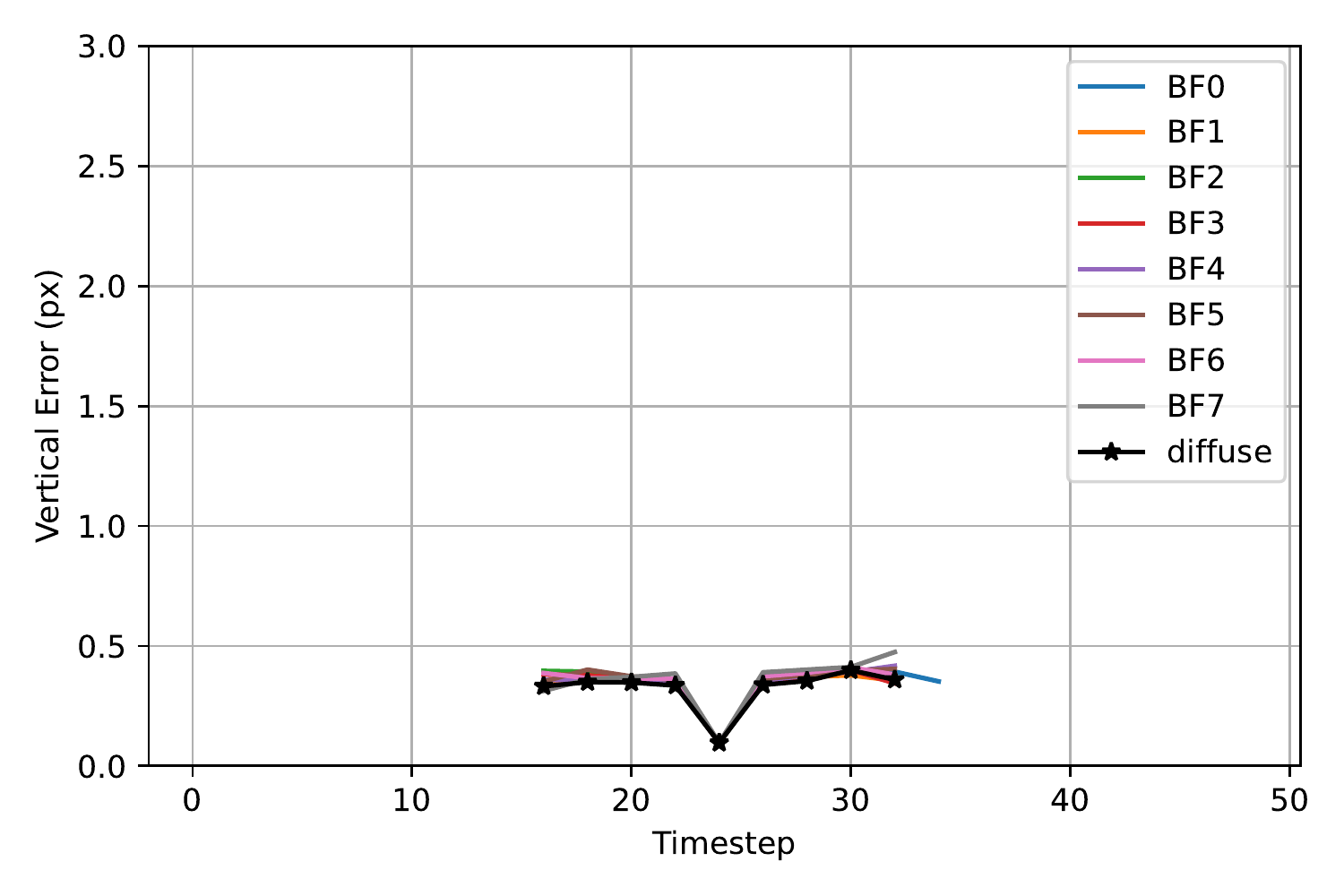}
    }
    \caption{\textbf{DTU Point Features Dataset: At twice nominal speed, lighting condition does not change trends in mean absolute error $\kappa(t)$  when using the Correspondence Tracker.} We compute $\kappa(t)$ using diffuse lighting (black lines) and each of the directional lighting conditions listed in Figure \ref{fig:dtu_light_stage} using all tracks from all 60 scenes. Lines are limited to timesteps containing at least 100 features.  The variation of $\kappa(t)$ due to the existence of directional lighting is at most 10 percent of the variation common to all plotted lines. The effect of directional lighting is relatively small because changes between adjacent frames are small whether or not the scene contains directional lighting. }
    \label{dtu_match_kappa_speed2.00}
\end{figure}

\begin{figure}[H]
    \centering
    \subfigure[Horizontal Coordinate]{
        \includegraphics[width=0.48\textwidth]{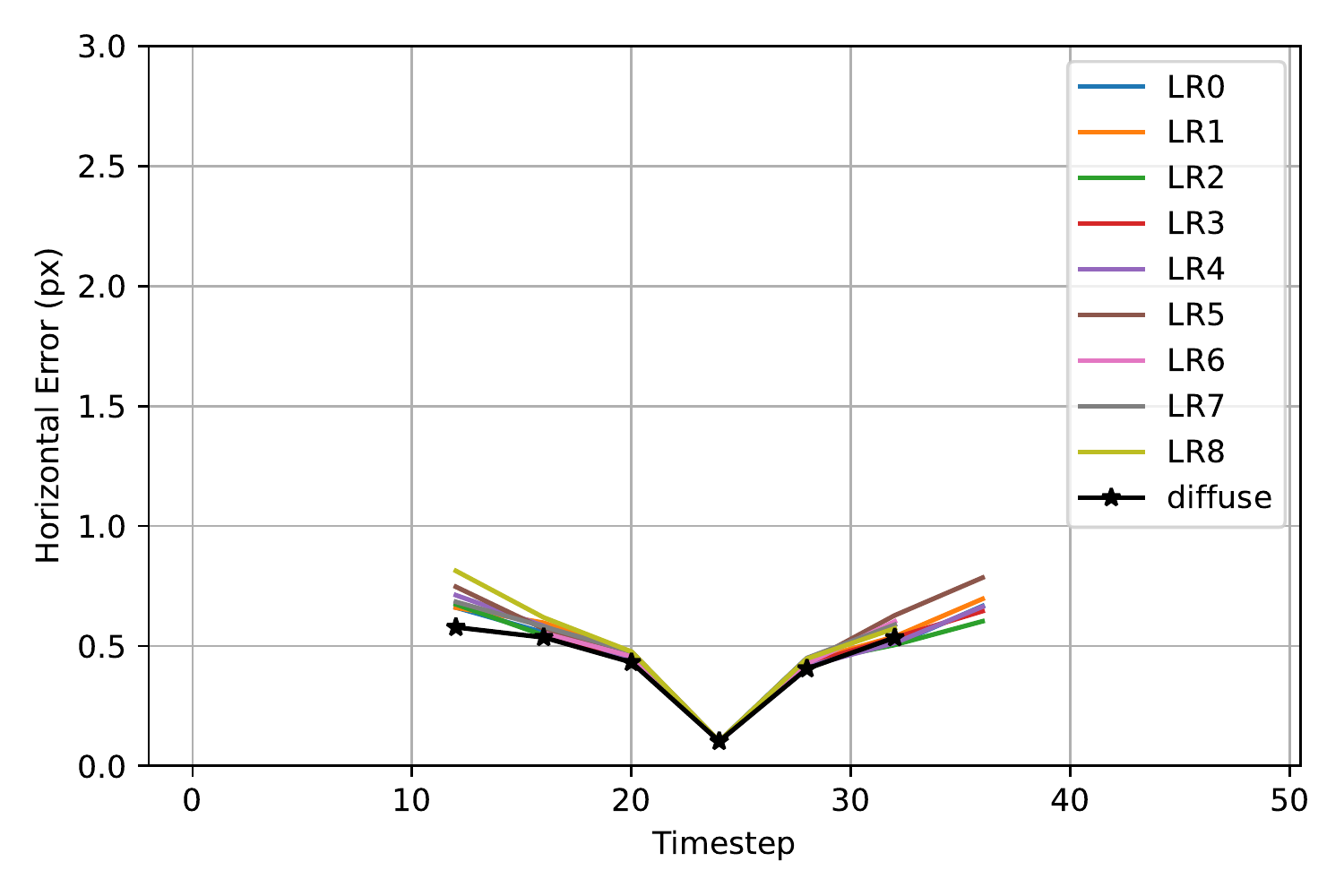}
        \includegraphics[width=0.48\textwidth]{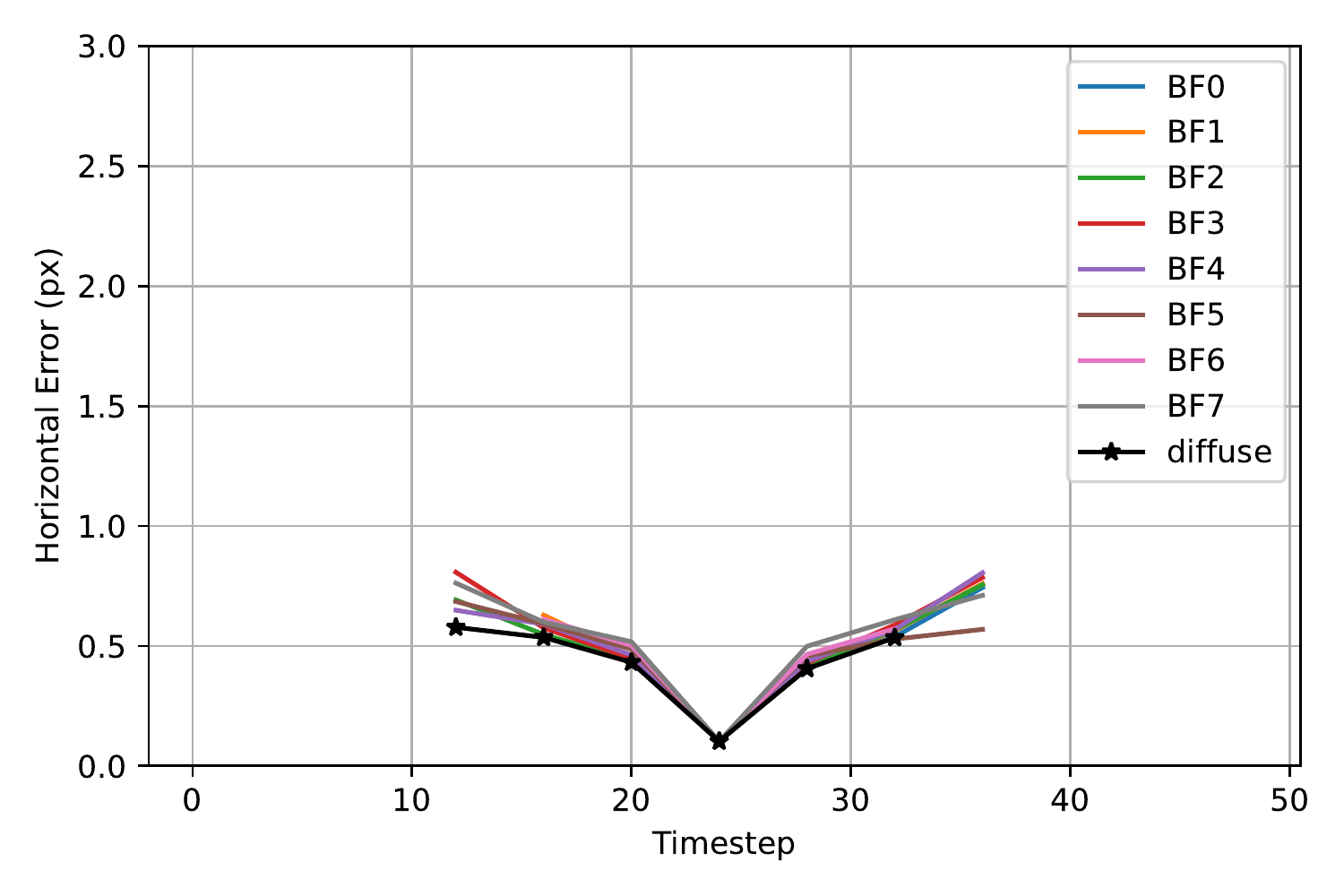}
    }
    \subfigure[Vertical Coordinate]{
        \includegraphics[width=0.48\textwidth]{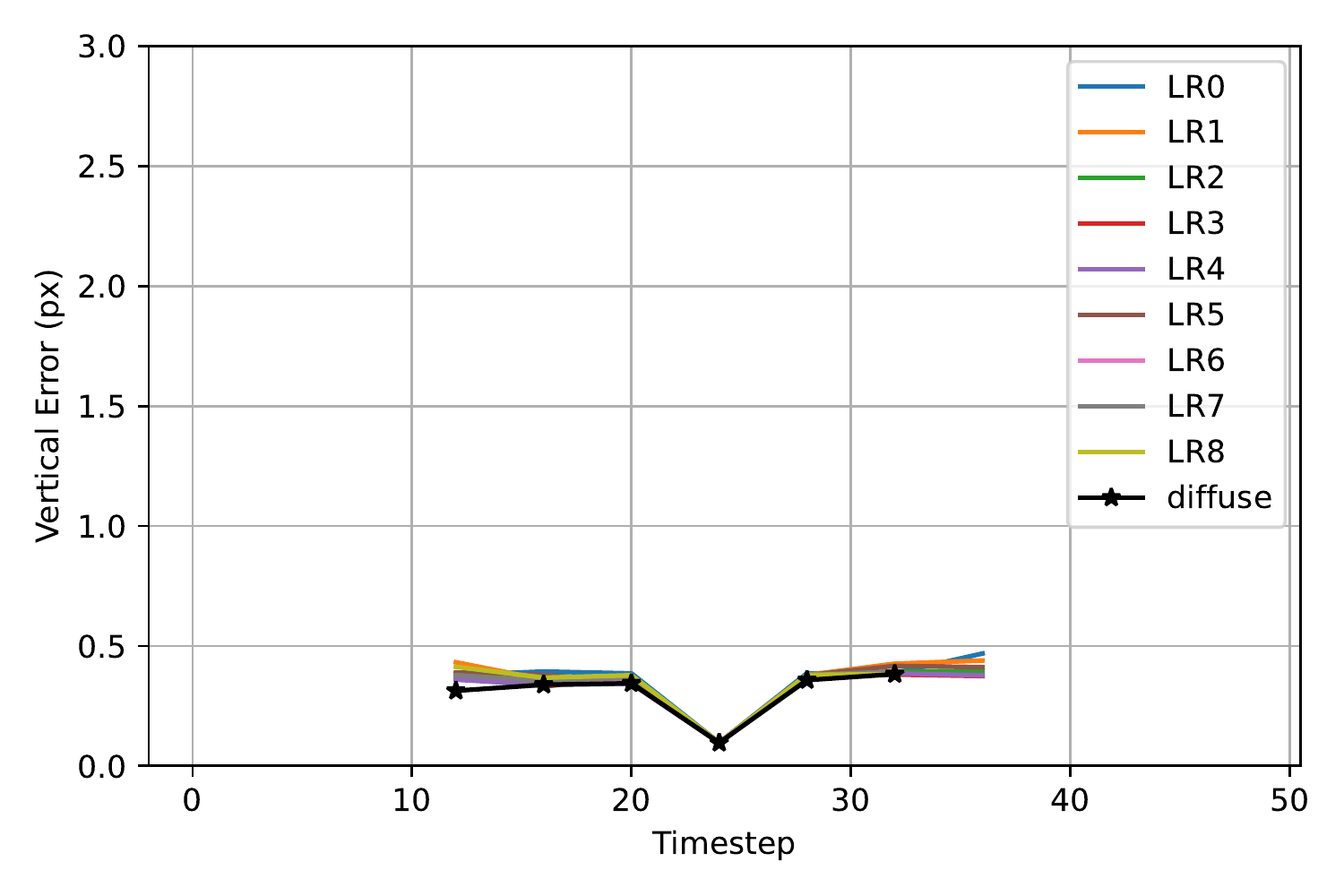}
        \includegraphics[width=0.48\textwidth]{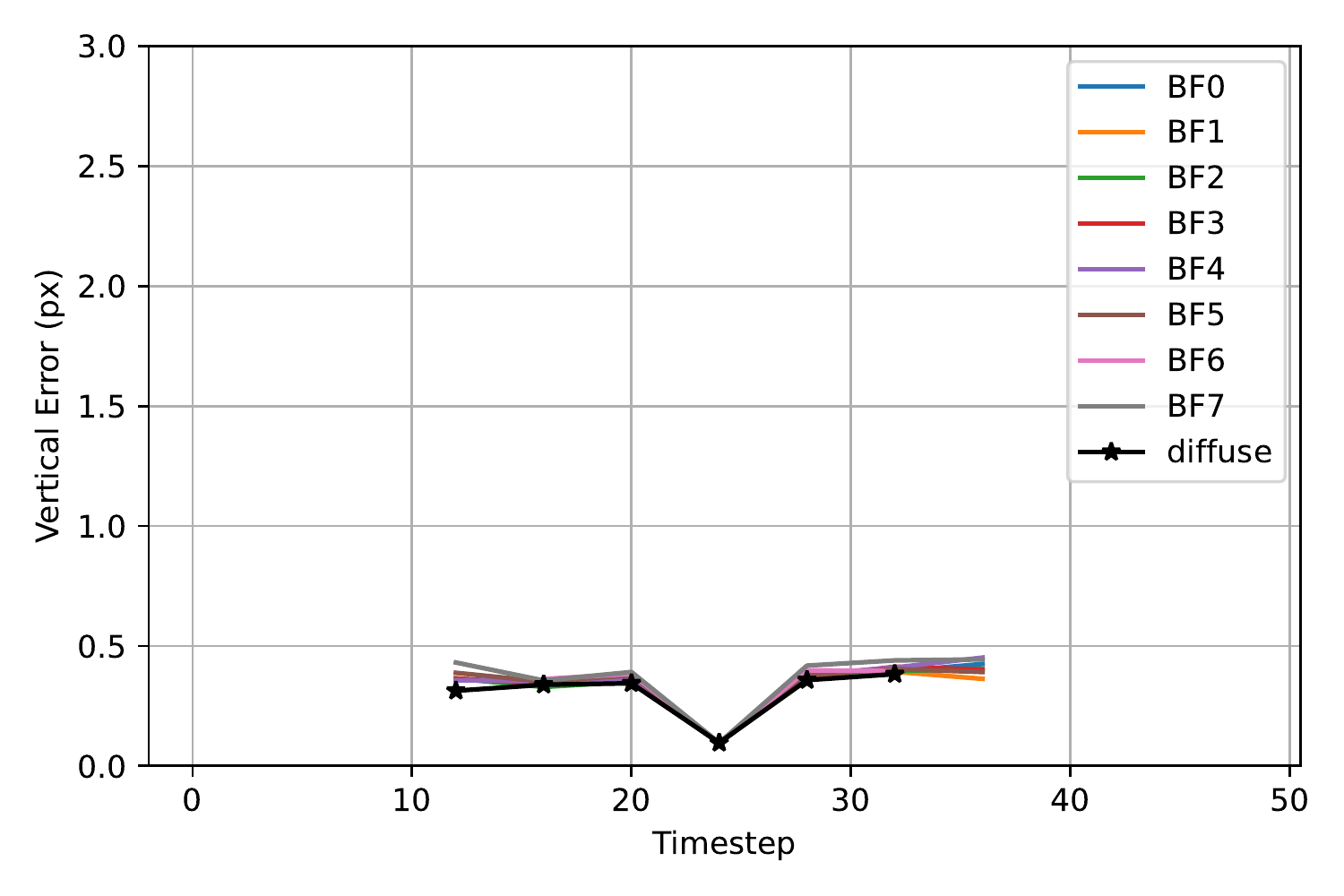}
    }
    \caption{\textbf{DTU Point Features Dataset: At four times nominal speed, lighting condition does not change trends in mean absolute error $\kappa(t)$ remains independent of lighting condition when using the Correspondence Tracker.} We compute $\kappa(t)$ using diffuse lighting (black lines) and each of the directional lighting conditions listed in Figure \ref{fig:dtu_light_stage} using all tracks from all 60 scenes. Lines are limited to timesteps containing at least 100 features. There are no significant differences between lines. The effect of directional lighting is small because changes from frame-to-frame are small.}
    \label{dtu_match_kappa_speed4.00}
\end{figure}

\begin{figure}[H]
    \centering
    \subfigure[Horizontal Coordinate]{
        \includegraphics[width=0.48\textwidth]{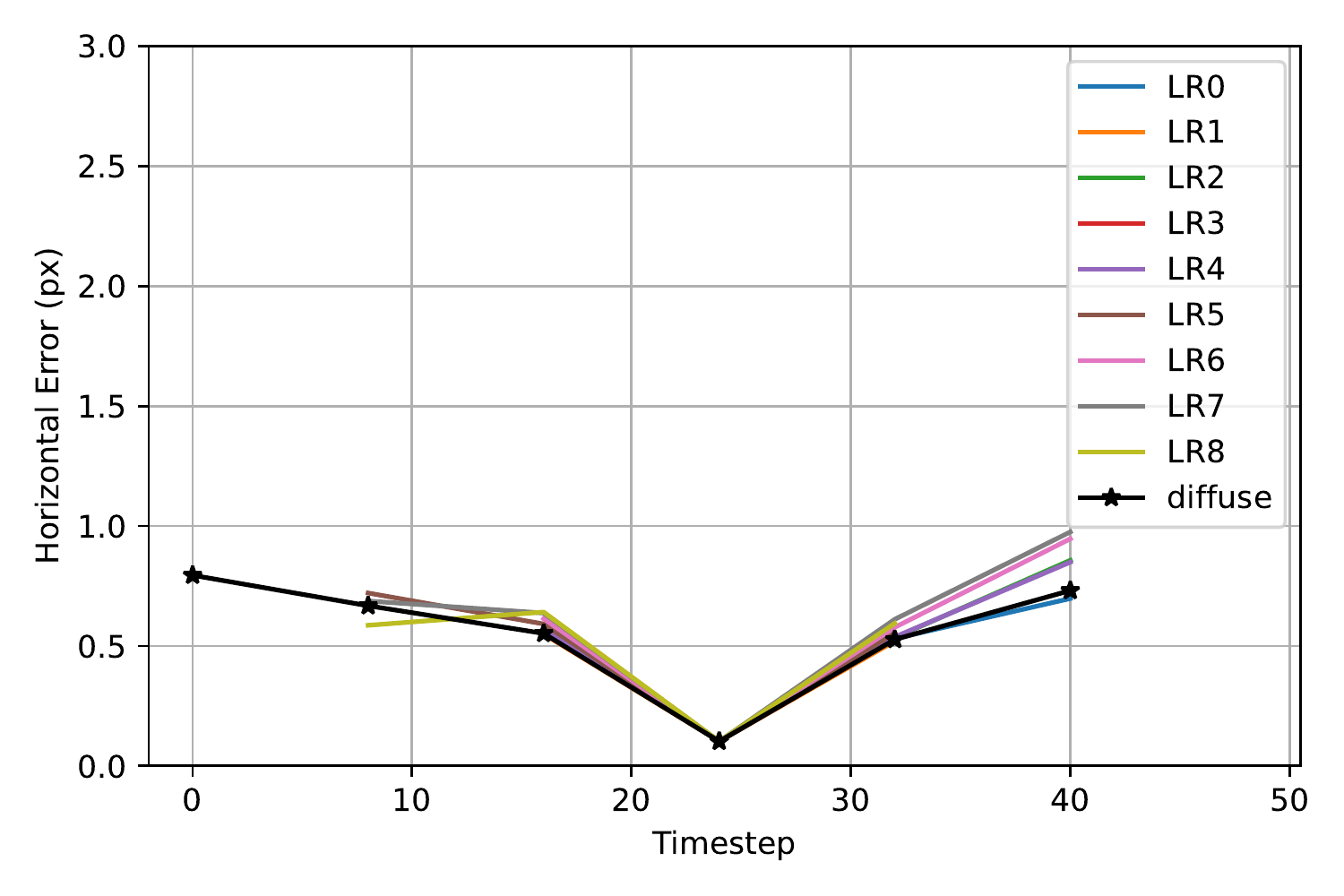}
        \includegraphics[width=0.48\textwidth]{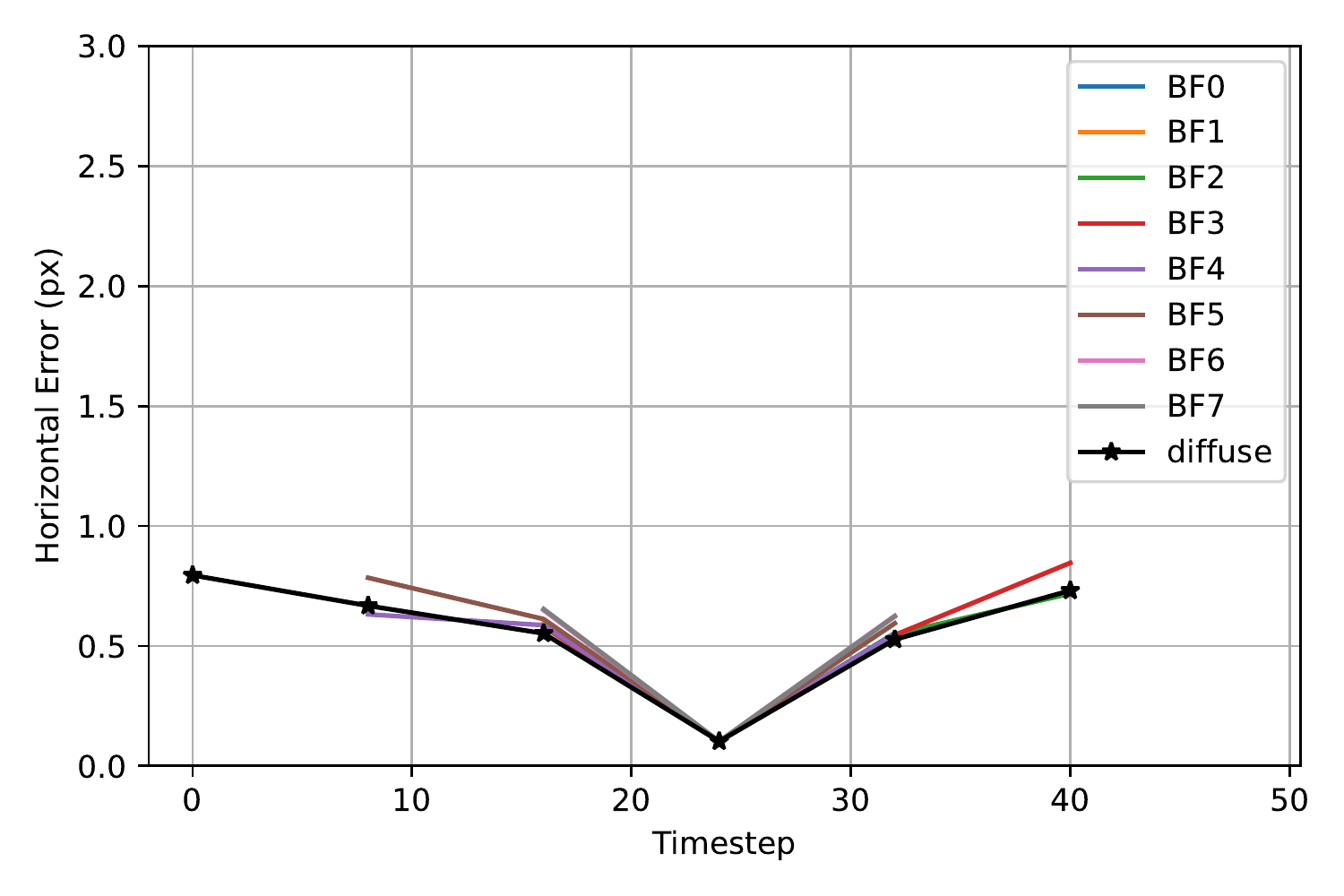}
    }
    \subfigure[Vertical Coordinate]{
        \includegraphics[width=0.48\textwidth]{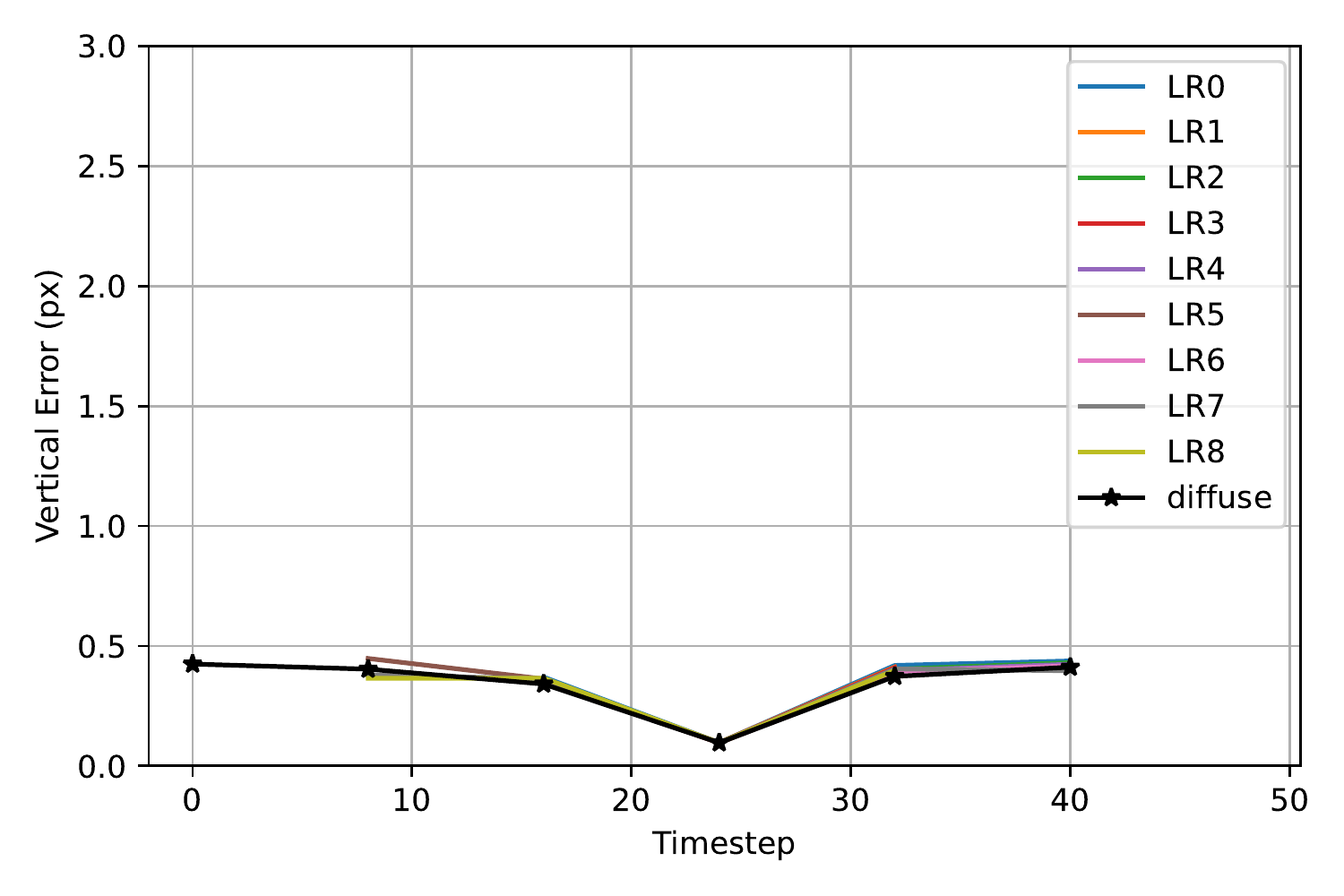}
        \includegraphics[width=0.48\textwidth]{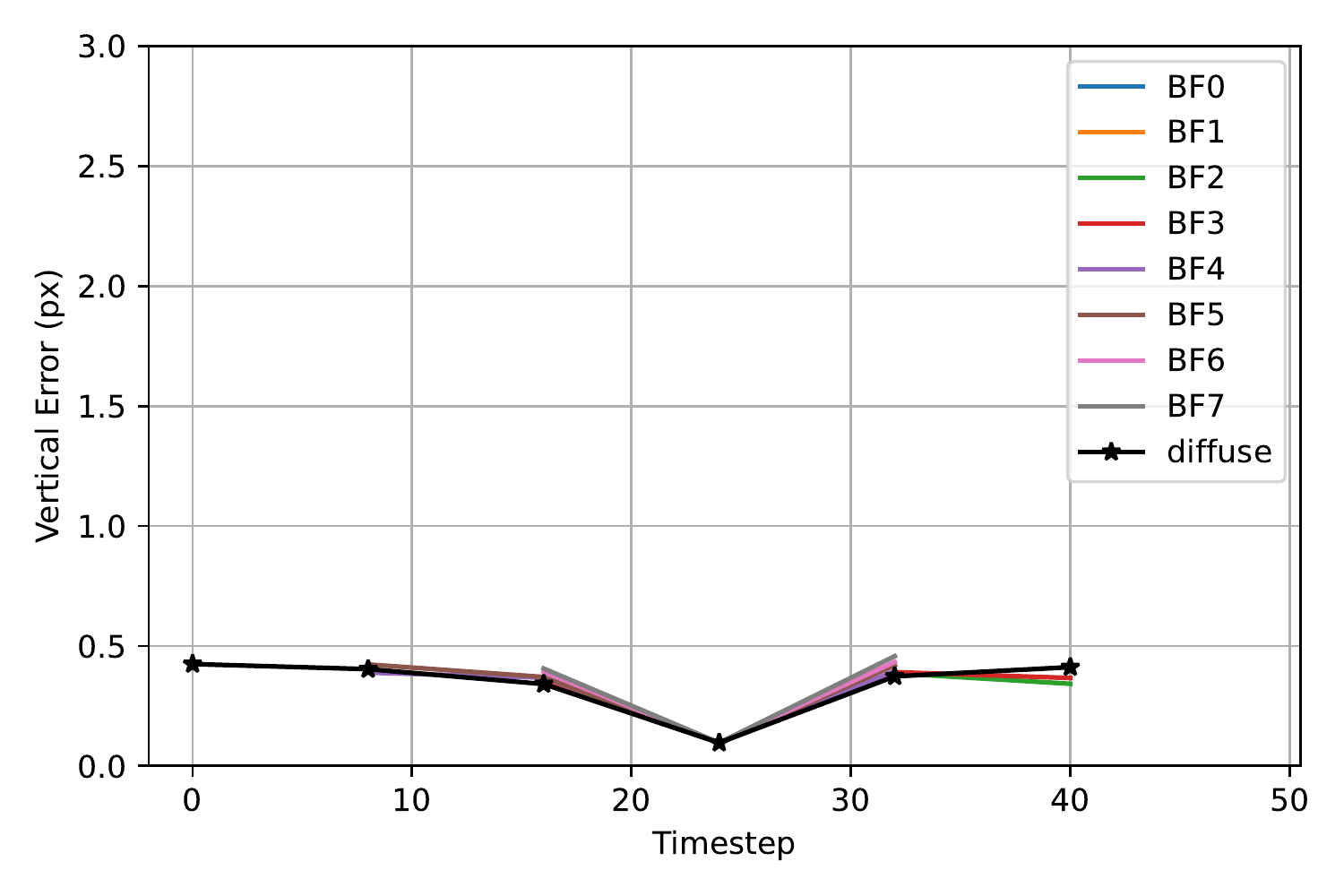}
    }
    \caption{\textbf{DTU Point Features Dataset: At eight times nominal speed, lighting condition does not change trends in mean absolute error $\kappa(t)$ when using the Correspondence Tracker.} We compute $\kappa(t)$ using diffuse lighting (black lines) and each of the directional lighting conditions listed in Figure \ref{fig:dtu_light_stage} using all tracks from all 60 scenes. Lines are limited to timesteps containing at least 100 features. The variation of $\kappa(t)$ due to the existence of directional lighting is at most 10 percent of the variation common to all plotted lines. The effect of directional lighting is relatively small because changes between adjacent frames are small whether or not the scene contains directional lighting. }
    \label{dtu_match_kappa_speed8.00}
\end{figure}

\begin{figure}[H]
    \centering
    \subfigure[Horizontal Coordinate]{
        \includegraphics[width=0.48\textwidth]{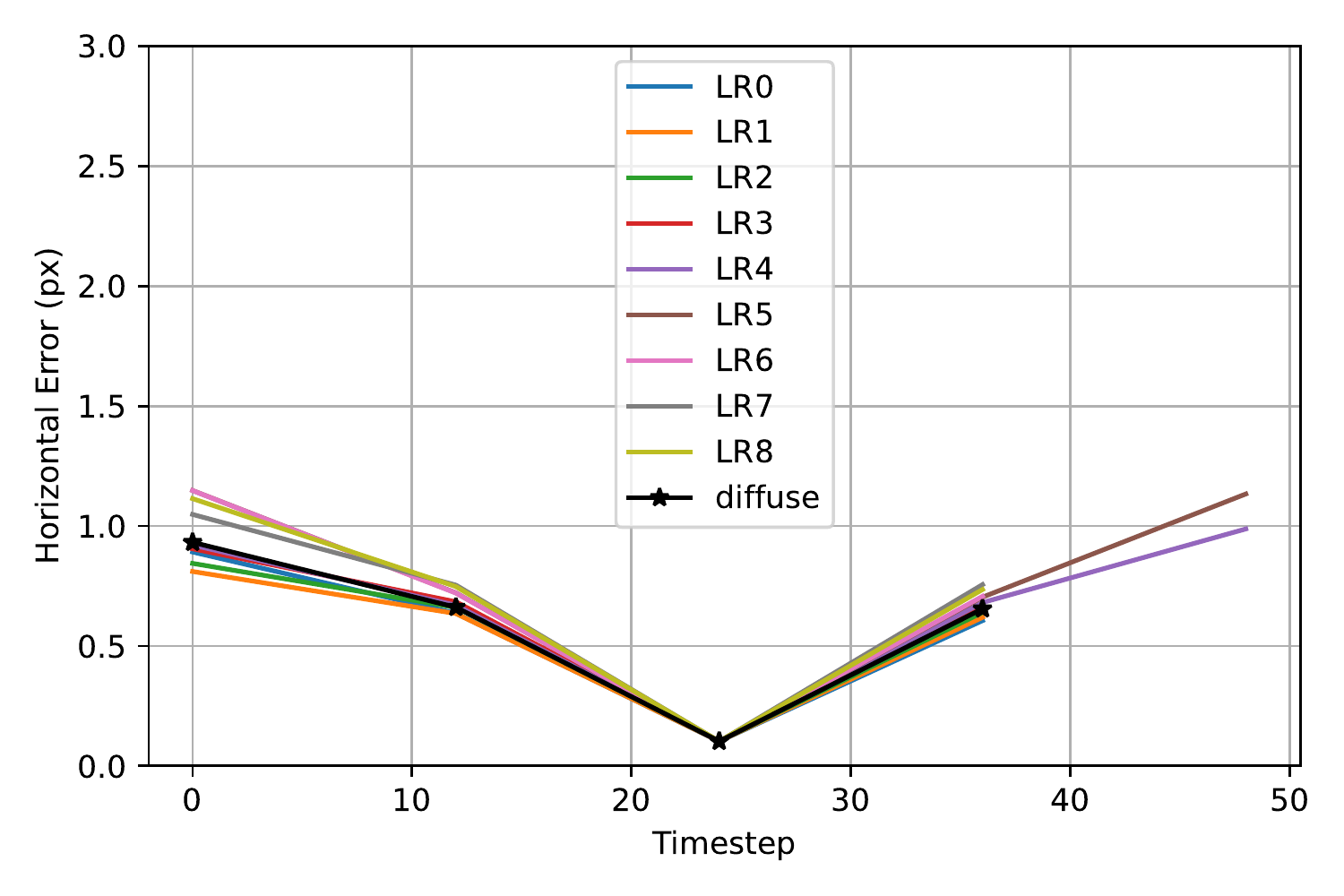}
        \includegraphics[width=0.48\textwidth]{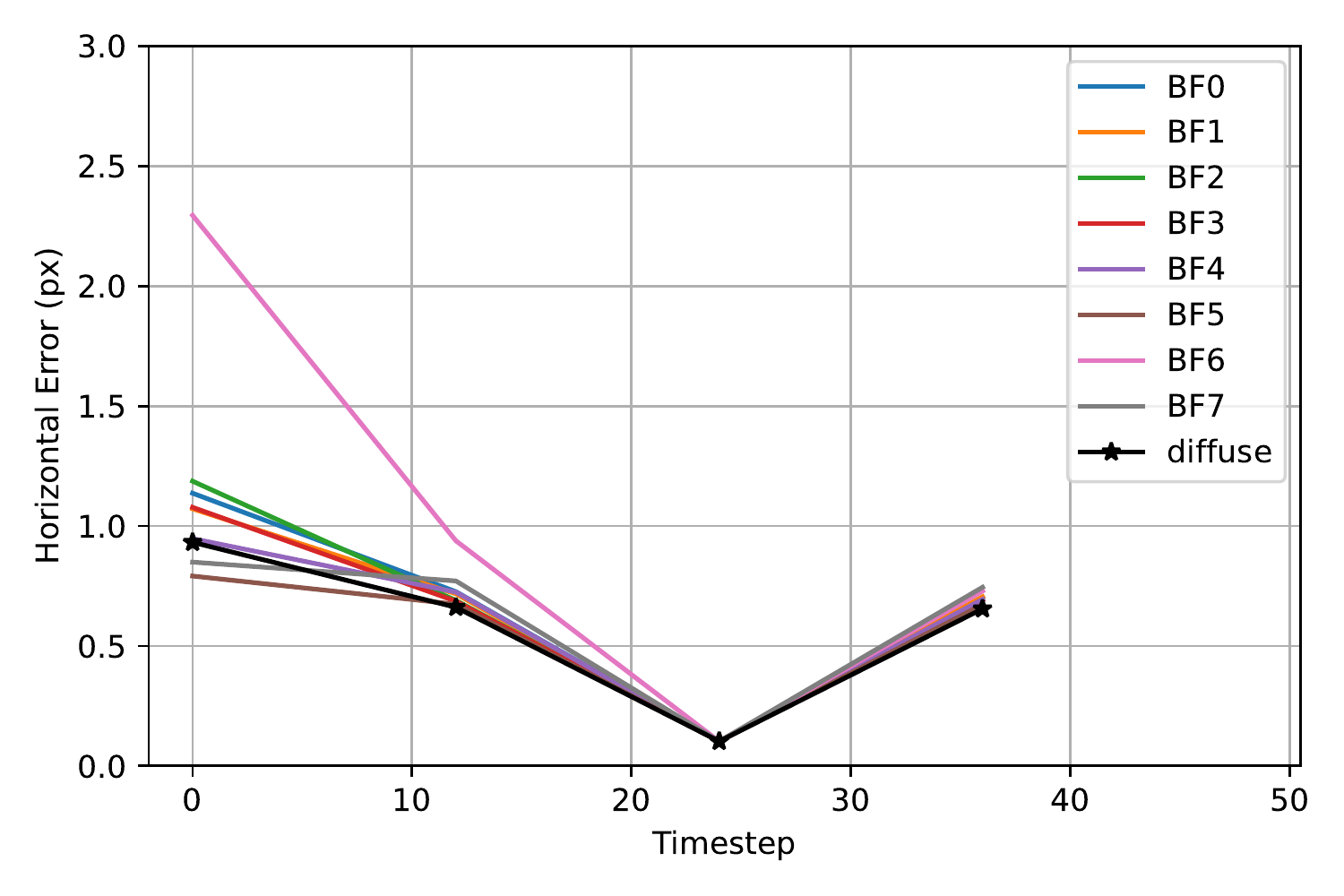}
    }
    \subfigure[Vertical Coordinate]{
        \includegraphics[width=0.48\textwidth]{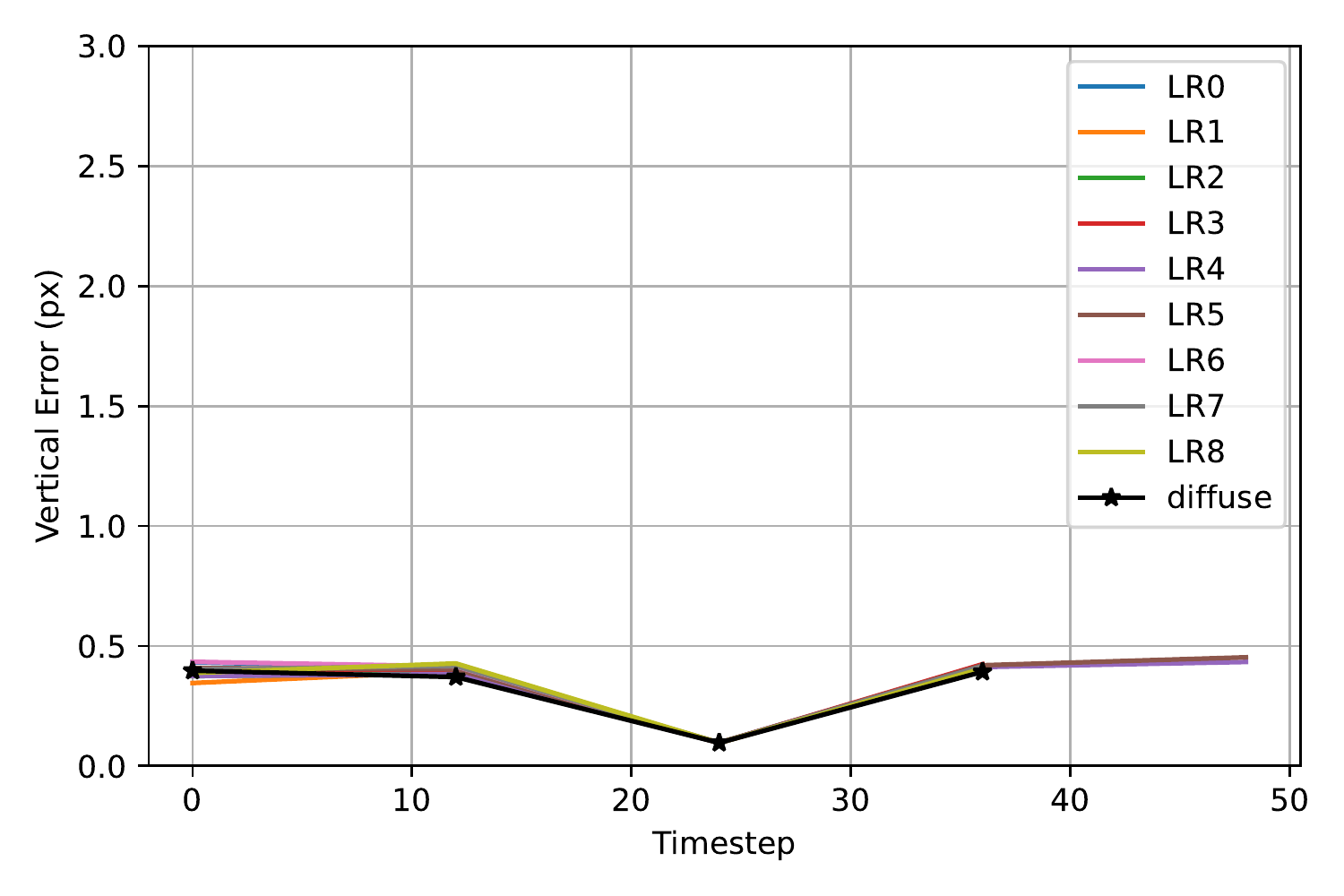}
        \includegraphics[width=0.48\textwidth]{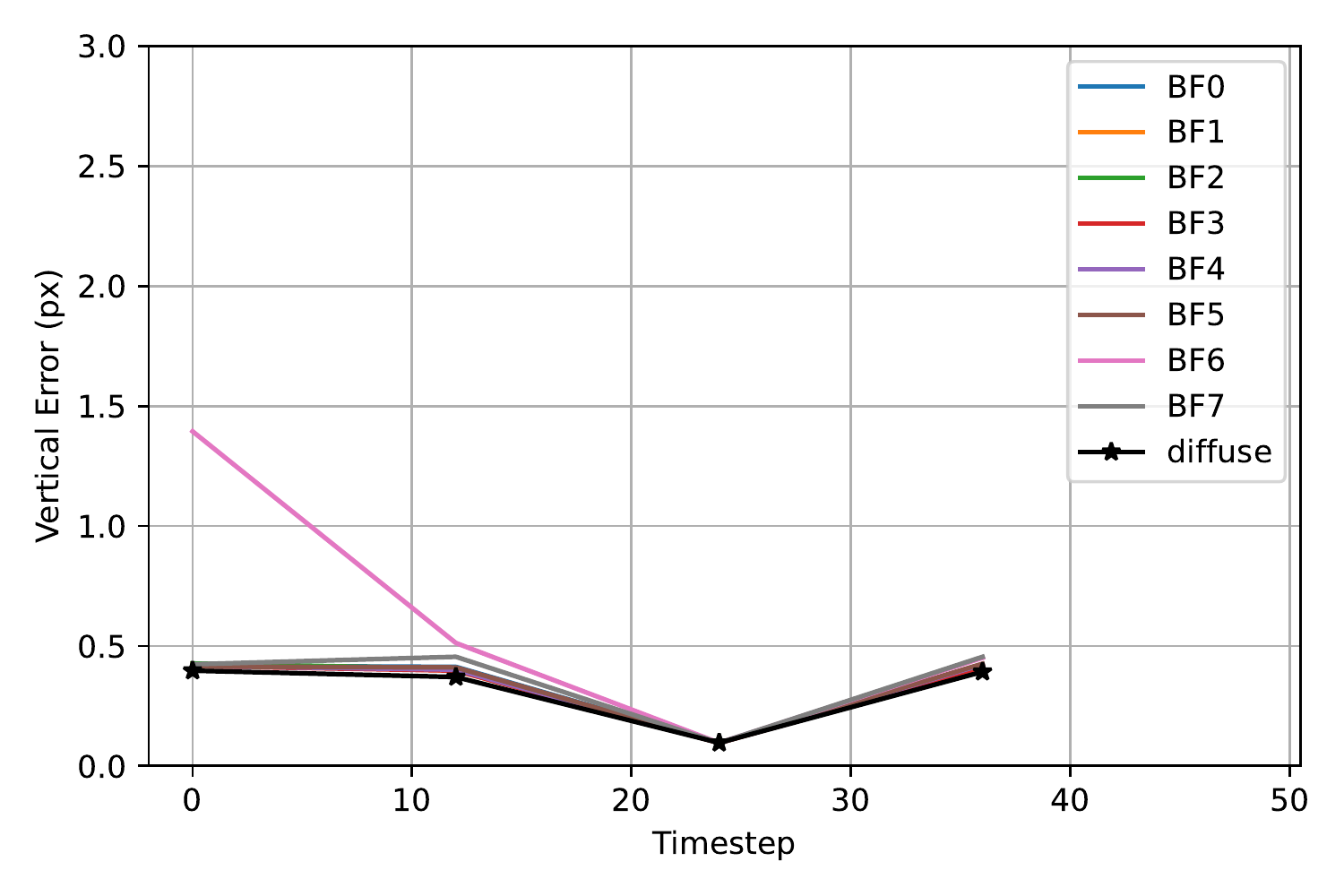}
    } 
    \caption{\textbf{DTU Point Features Dataset: At twelve times nominal speed, lighting condition does not change trends in mean absolute error $\kappa(t)$  when using the Correspondence Tracker.} We compute $\kappa(t)$ using diffuse lighting (black lines) and each of the directional lighting conditions listed in Figure \ref{fig:dtu_light_stage} using all tracks from all 60 scenes. Lines are limited to timesteps containing at least 100 features. With the exception of lighting condition BF6, the variation of $\kappa(t)$ due to the existence of directional lighting is at most 10 percent of the variation common to all plotted lines. The effect of directional lighting is relatively small because changes between adjacent frames are small whether or not the scene contains directional lighting.
    }
    \label{fig:dtu_match_kappa_speed12.00}
\end{figure}

\begin{figure}[H]
    \centering
    \subfigure[Horizontal Coordinate]{
        \includegraphics[width=0.48\textwidth]{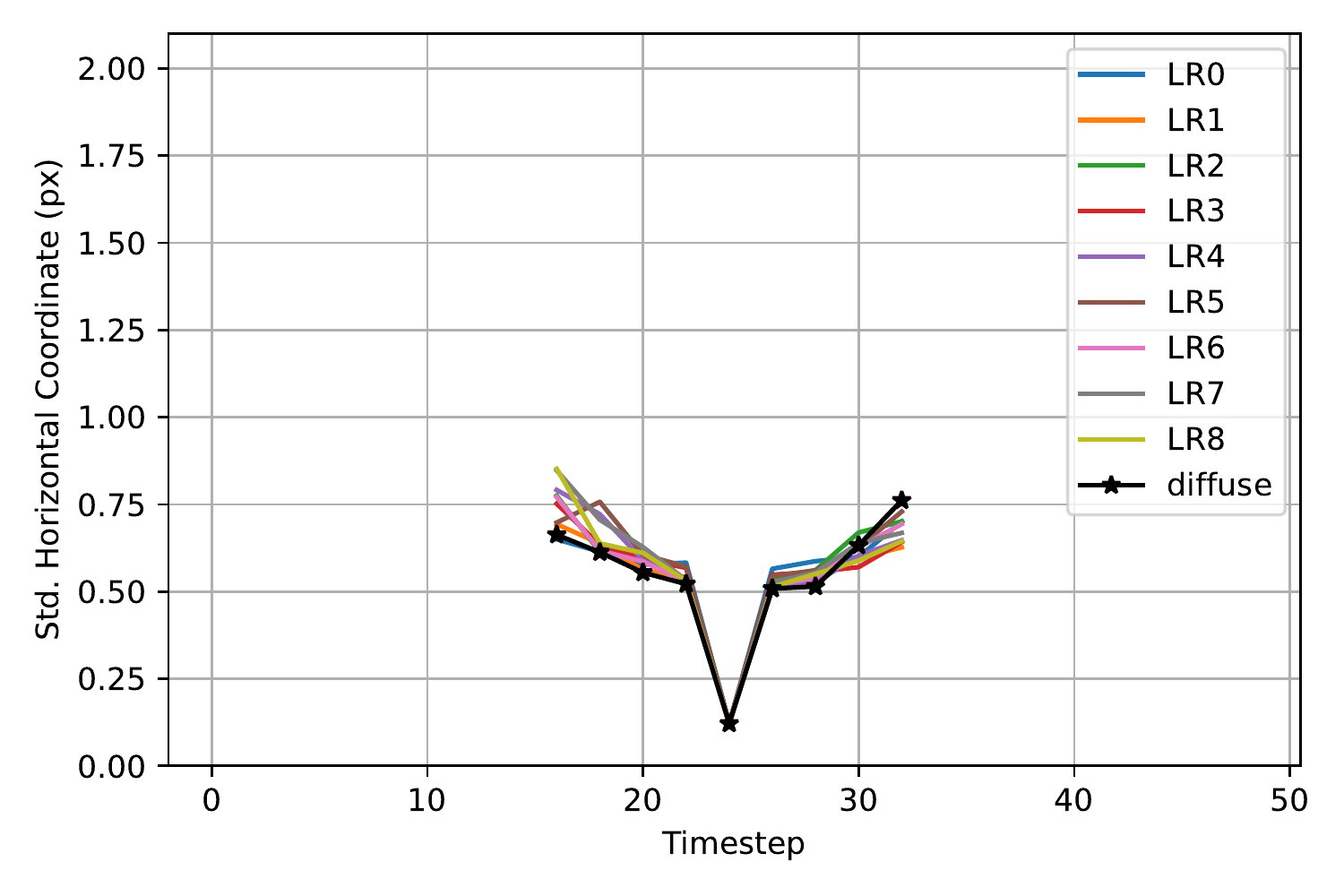}
        \includegraphics[width=0.48\textwidth]{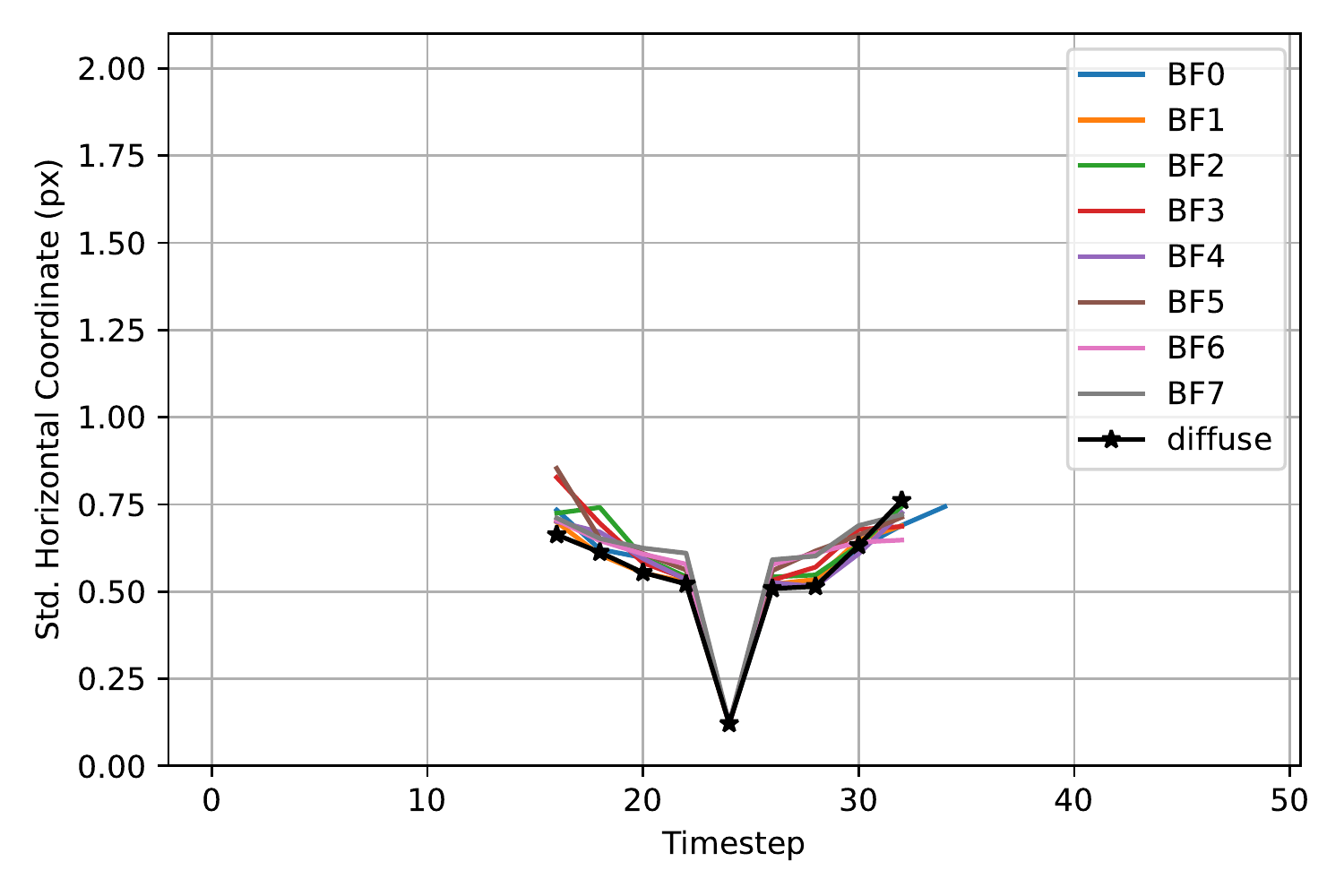}
    }
    \subfigure[Vertical Coordinate]{
        \includegraphics[width=0.48\textwidth]{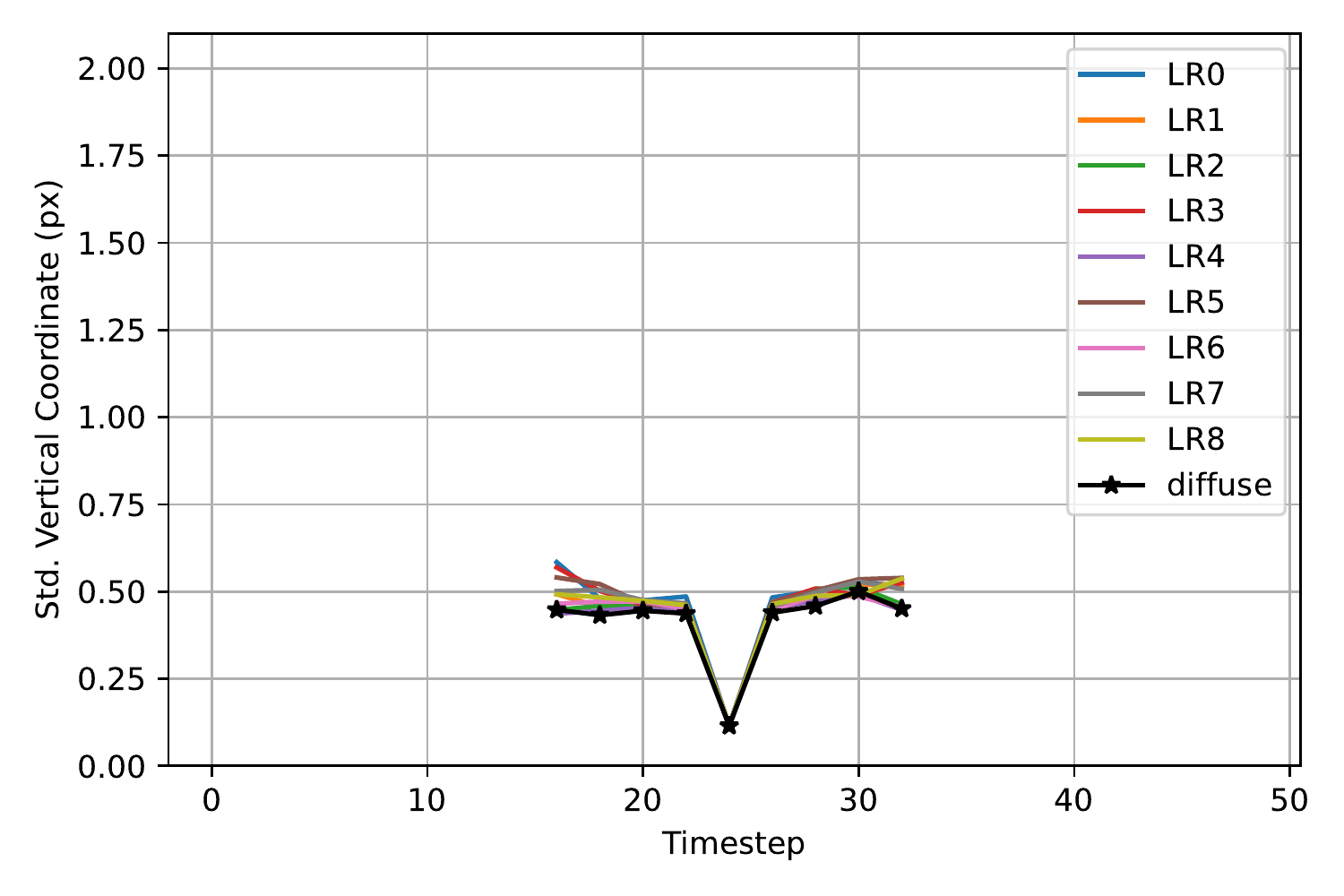}
        \includegraphics[width=0.48\textwidth]{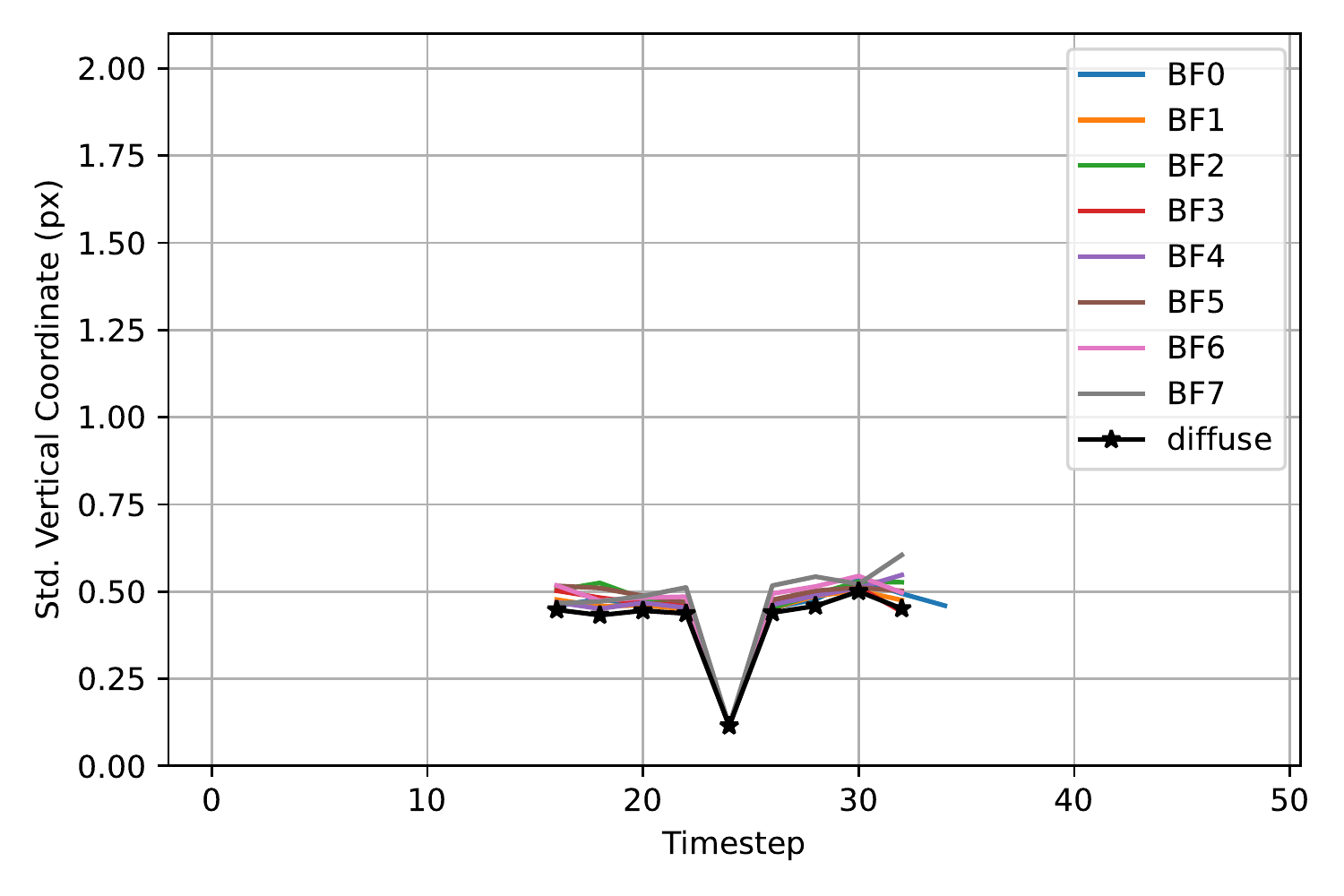}
    }
    \caption{\textbf{DTU Point Features Dataset: At twice nominal speed, lighting condition does not change trends in covariance $\Sigma(t)$  when using the Correspondence Tracker.} We compute $\Sigma(t)$ using diffuse lighting (black lines) and each of the directional lighting conditions listed in Figure \ref{fig:dtu_light_stage} using all tracks from all 60 scenes. Timesteps are limited to those with at least 100 features. The variation of $\Sigma(t)$ due to the existence of directional lighting is at most 10 percent of the variation common to all plotted lines. The effect of directional lighting is relatively small because changes between adjacent frames are small whether or not the scene contains directional lighting.}
    \label{fig:dtu_match_cov_speed2.00}
\end{figure}

\begin{figure}[H]
    \centering
    \subfigure[Horizontal Coordinate]{
        \includegraphics[width=0.48\textwidth]{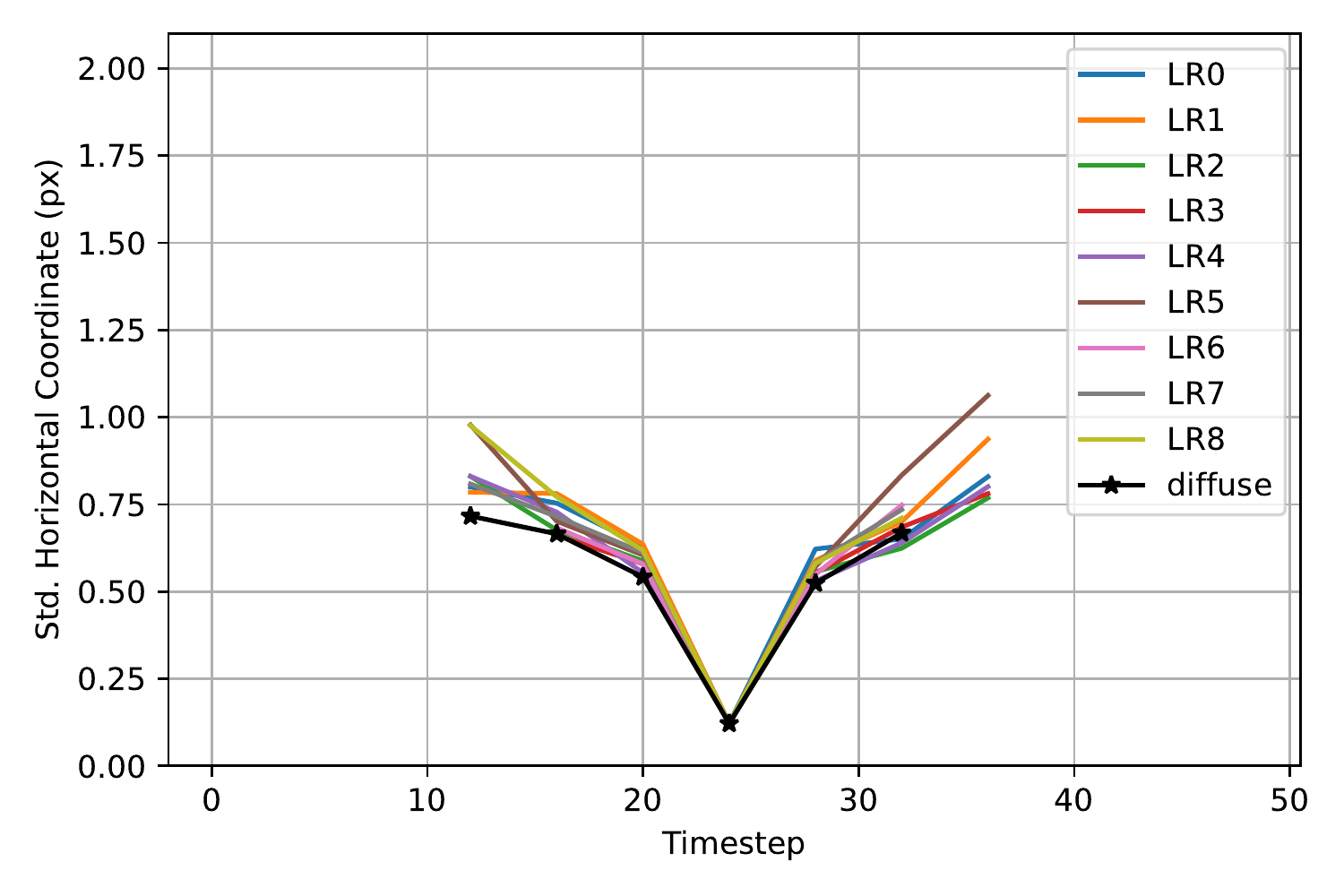}
        \includegraphics[width=0.48\textwidth]{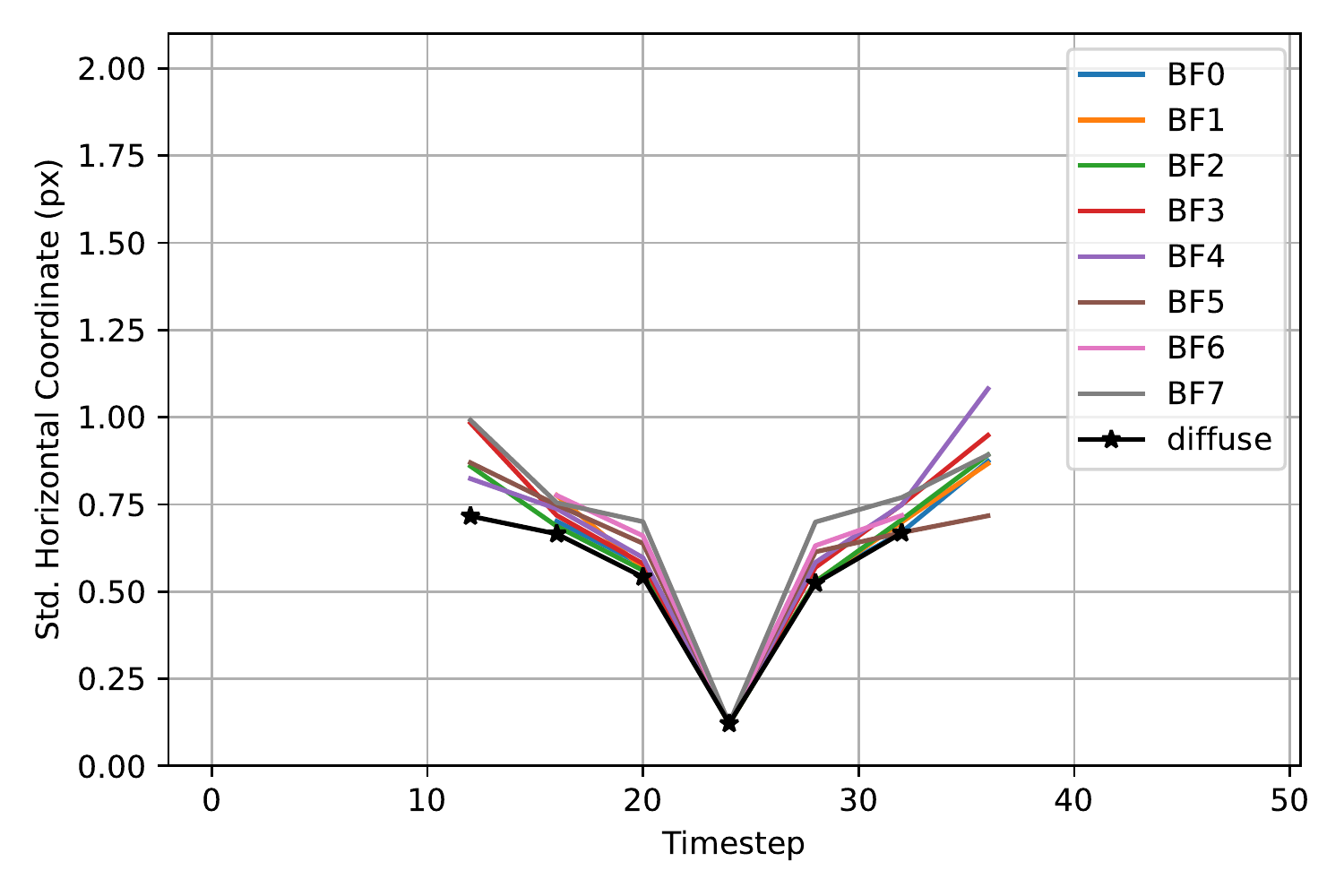}
    }
    \subfigure[Vertical Coordinate]{
        \includegraphics[width=0.48\textwidth]{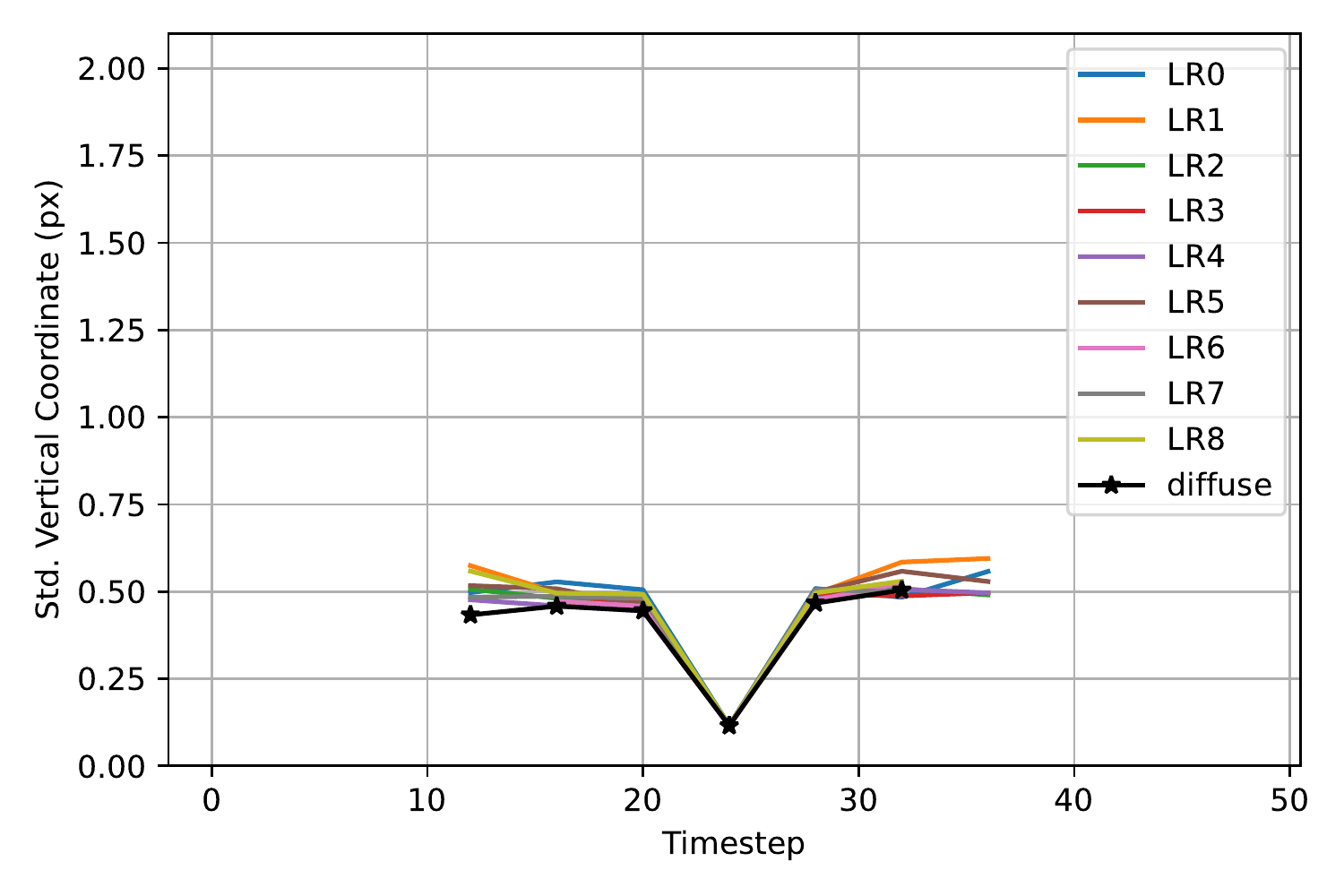}
        \includegraphics[width=0.48\textwidth]{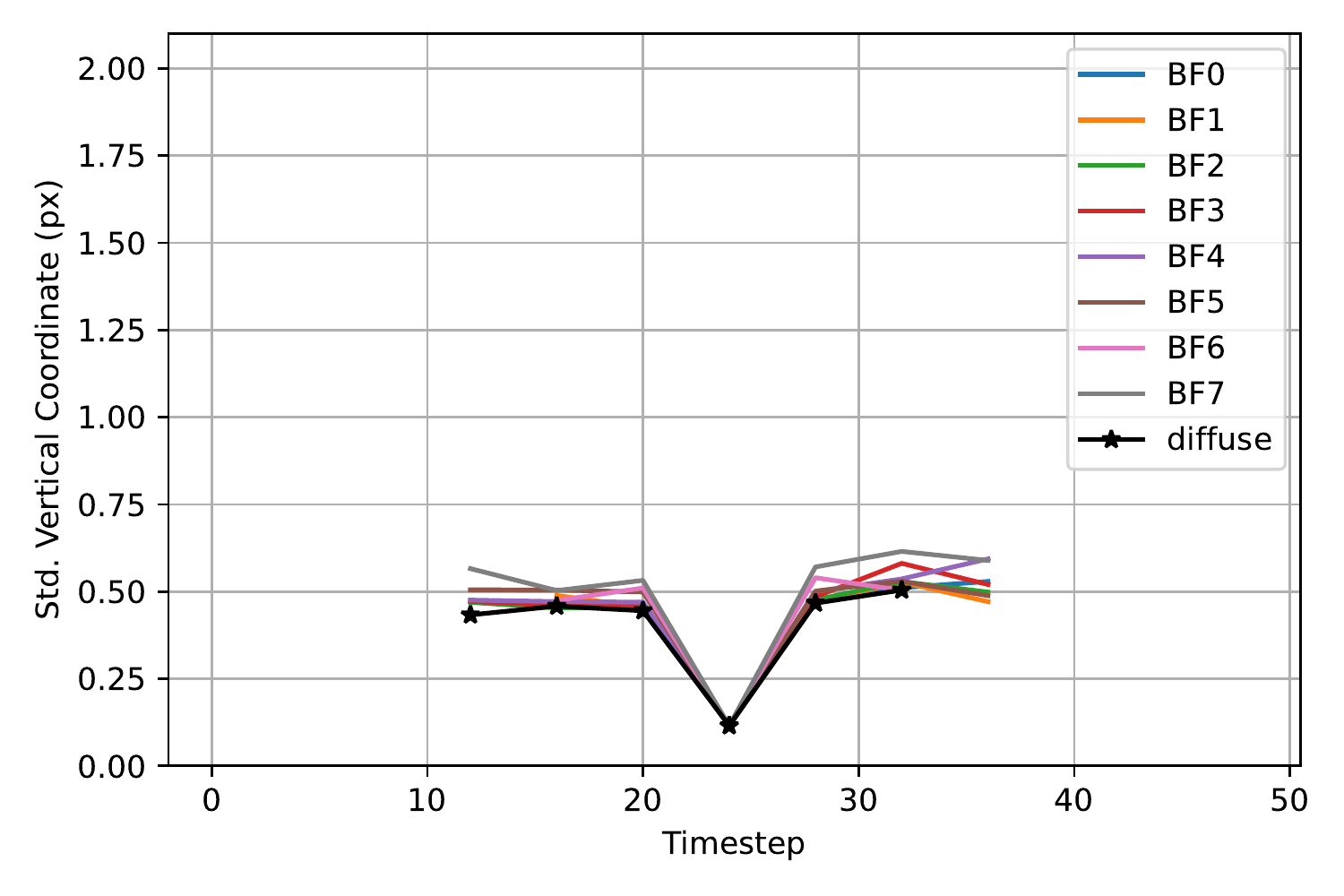}
    }
    \caption{\textbf{DTU Point Features Dataset: At four times nominal speed, lighting condition does not change trends in covariance $\Sigma(t)$  when using the Correspondence Tracker.} We compute $\Sigma(t)$ using diffuse lighting (black lines) and each of the directional lighting conditions listed in Figure \ref{fig:dtu_light_stage} using all tracks from all 60 scenes. Timesteps are limited to those with at least 100 features.  The variation of $\Sigma(t)$ due to the existence of directional lighting is at most 10 percent of the variation common to all plotted lines. The effect of directional lighting is relatively small because changes between adjacent frames are small whether or not the scene contains directional lighting.}
    \label{fig:dtu_match_cov_speed4.00}
\end{figure}

\begin{figure}[H]
    \centering
    \subfigure[Horizontal Coordinate]{
        \includegraphics[width=0.48\textwidth]{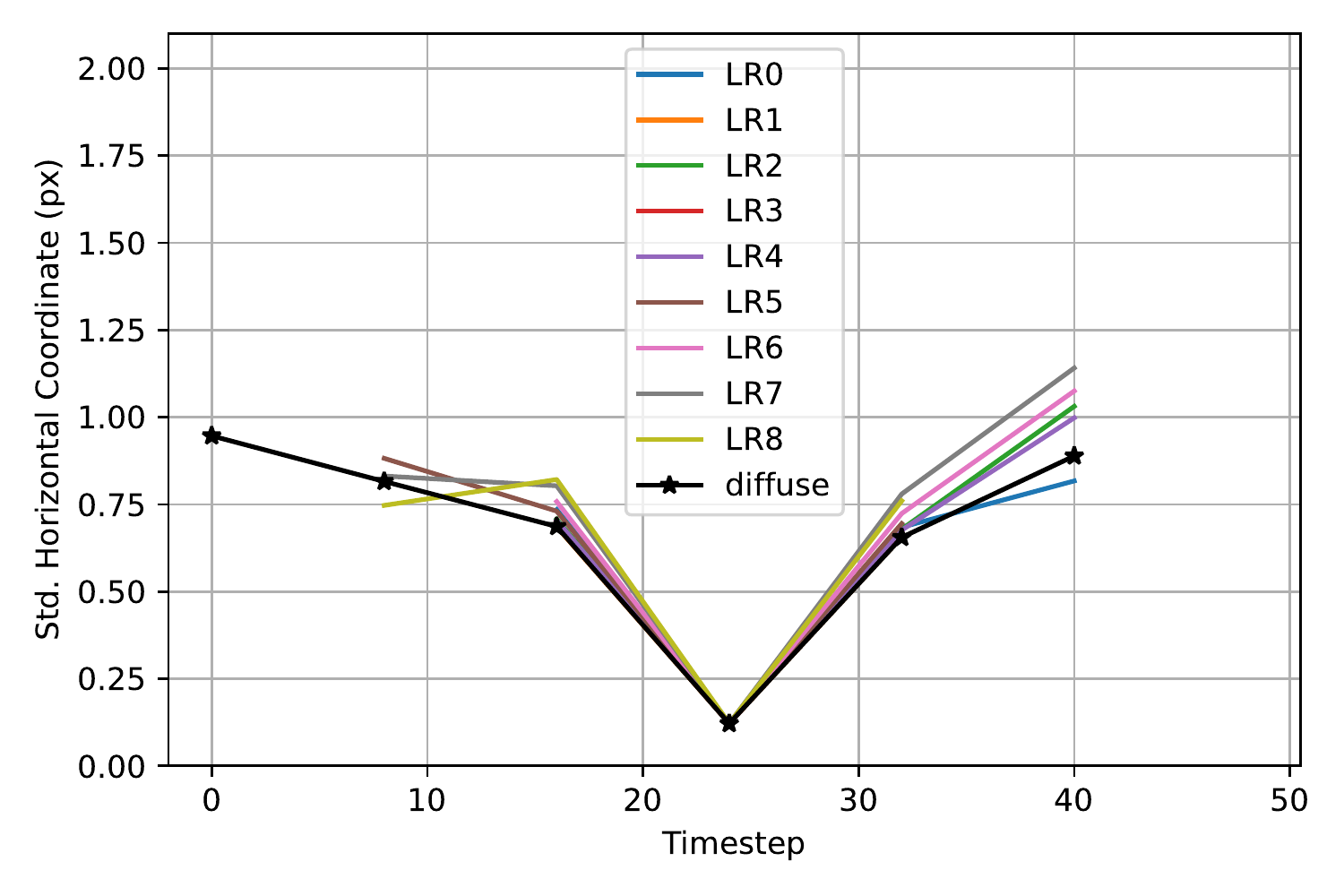}
        \includegraphics[width=0.48\textwidth]{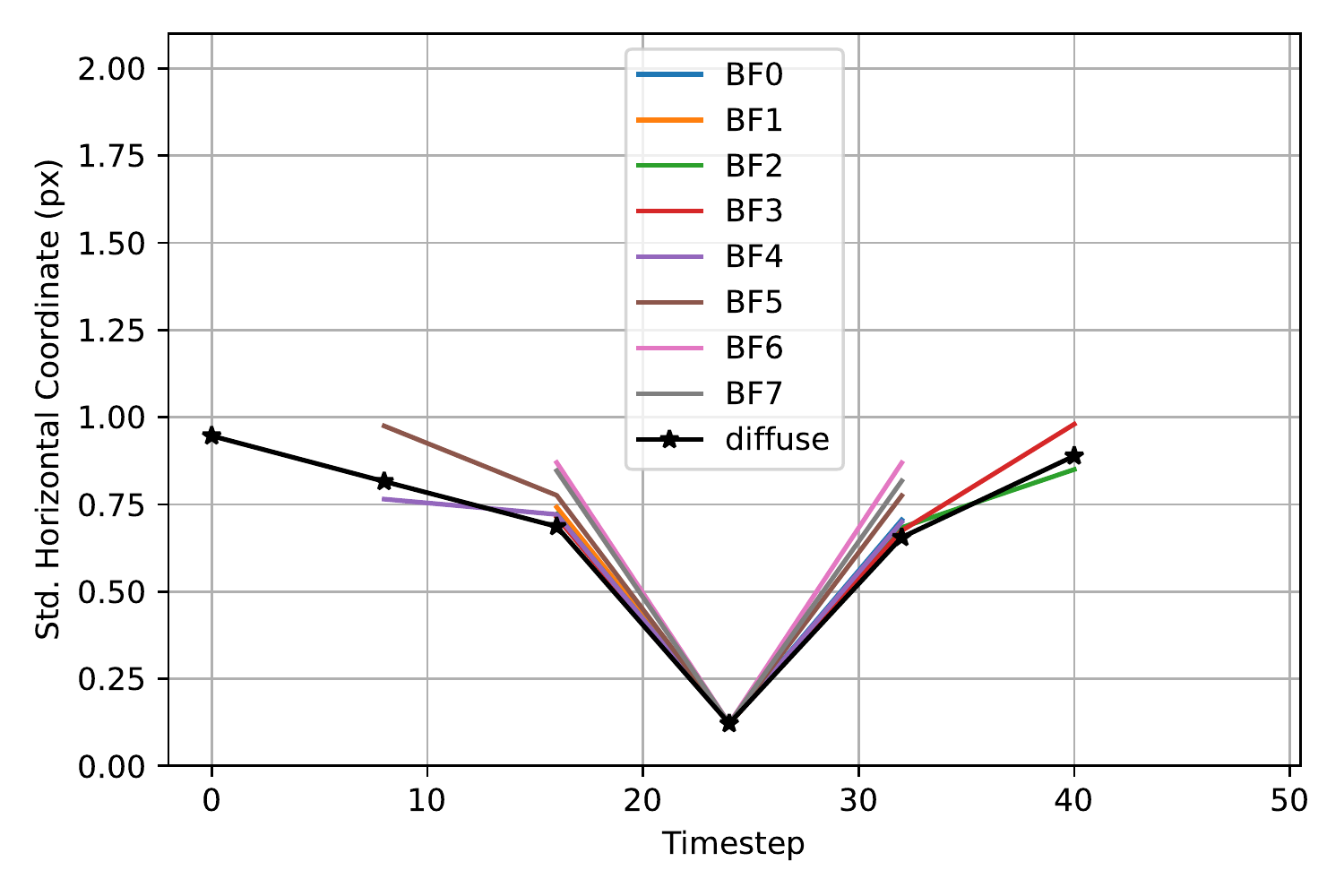}
    }
    \subfigure[Vertical Coordinate]{
        \includegraphics[width=0.48\textwidth]{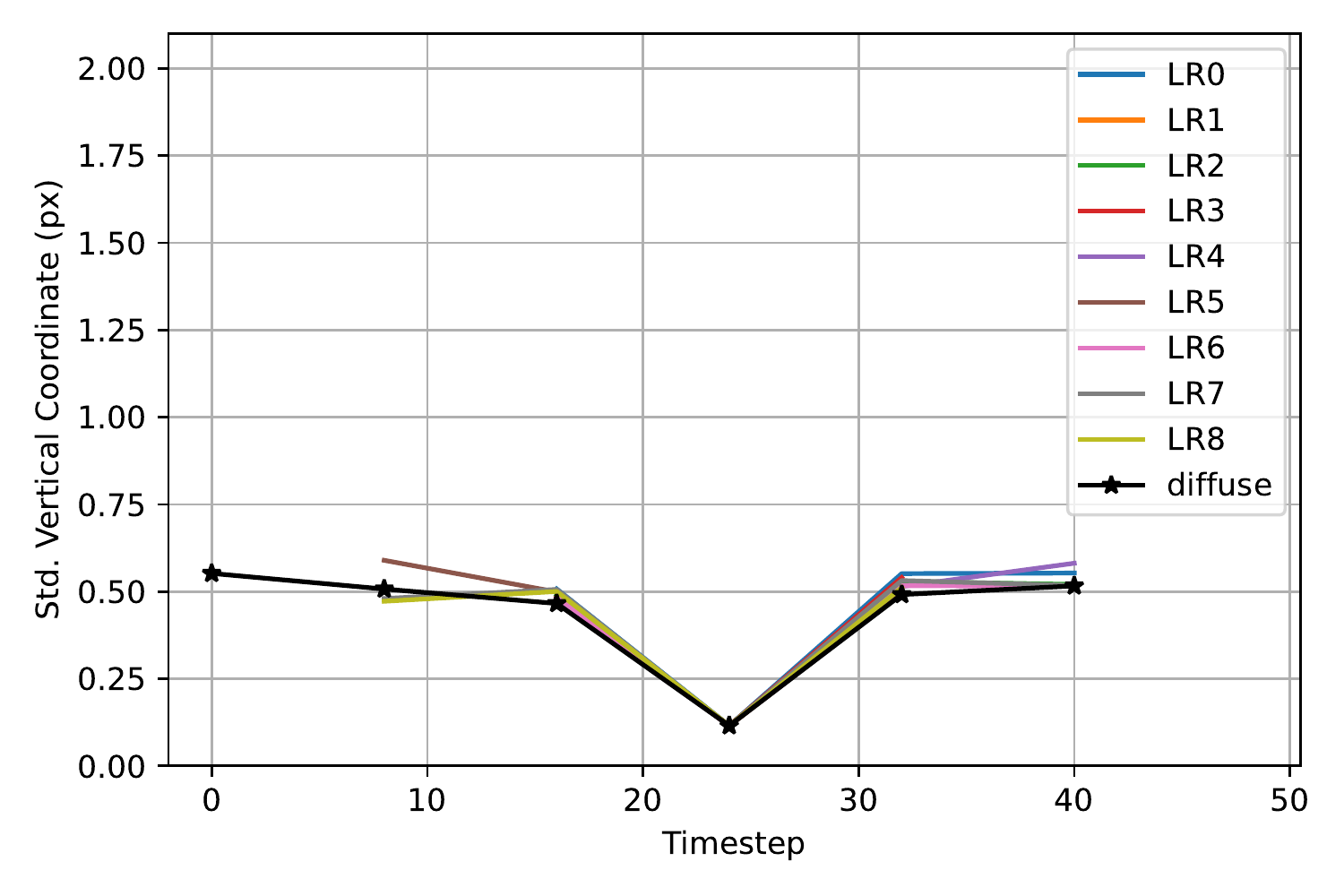}
        \includegraphics[width=0.48\textwidth]{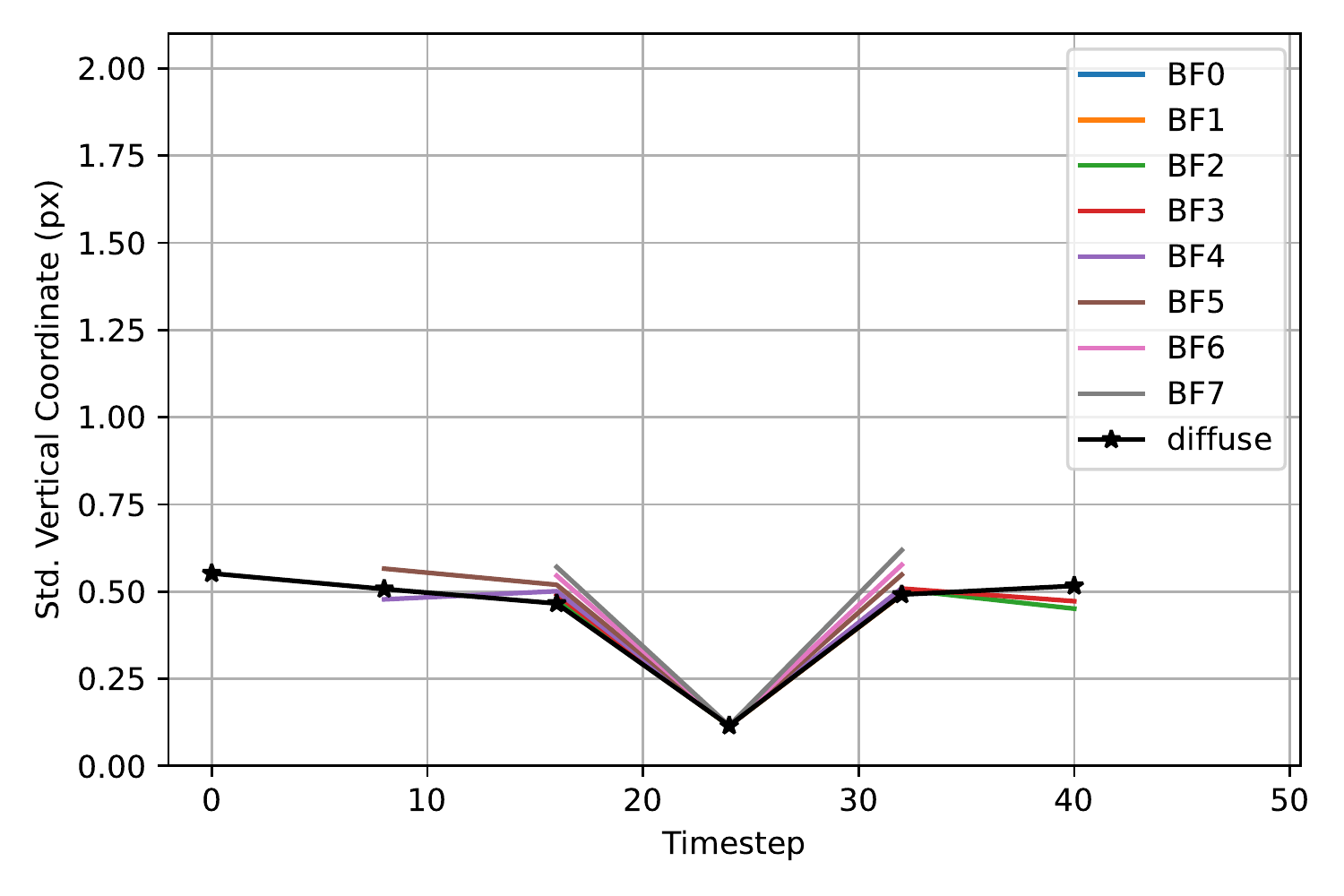}
    }
    \caption{\textbf{DTU Point Features Dataset: At eight times nominal speed, lighting condition does not change trends in covariance $\Sigma(t)$ when using the Correspondence Tracker.} We compute $\Sigma(t)$ using diffuse lighting (black lines) and each of the directional lighting conditions listed in Figure \ref{fig:dtu_light_stage} using all tracks from all 60 scenes. Timesteps are limited to those with at least 100 features. The variation of $\Sigma(t)$ due to the existence of directional lighting is at most 10 percent of the variation common to all plotted lines. The effect of directional lighting is relatively small because changes between adjacent frames are small whether or not the scene contains directional lighting.}
    \label{fig:dtu_match_cov_speed8.00}
\end{figure}

\begin{figure}[H]
    \centering
    \subfigure[Horizontal Coordinate]{
        \includegraphics[width=0.48\textwidth]{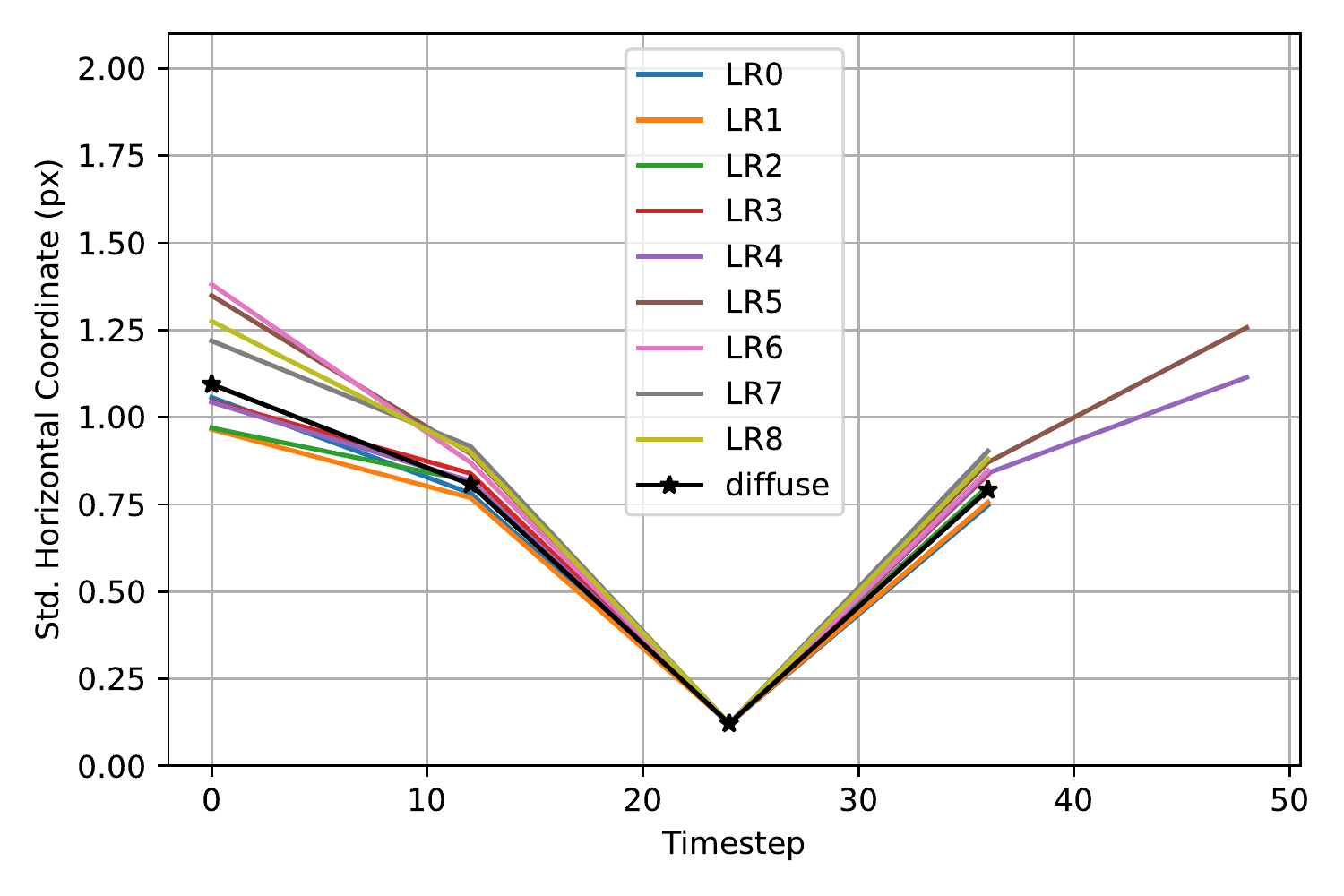}
        \includegraphics[width=0.48\textwidth]{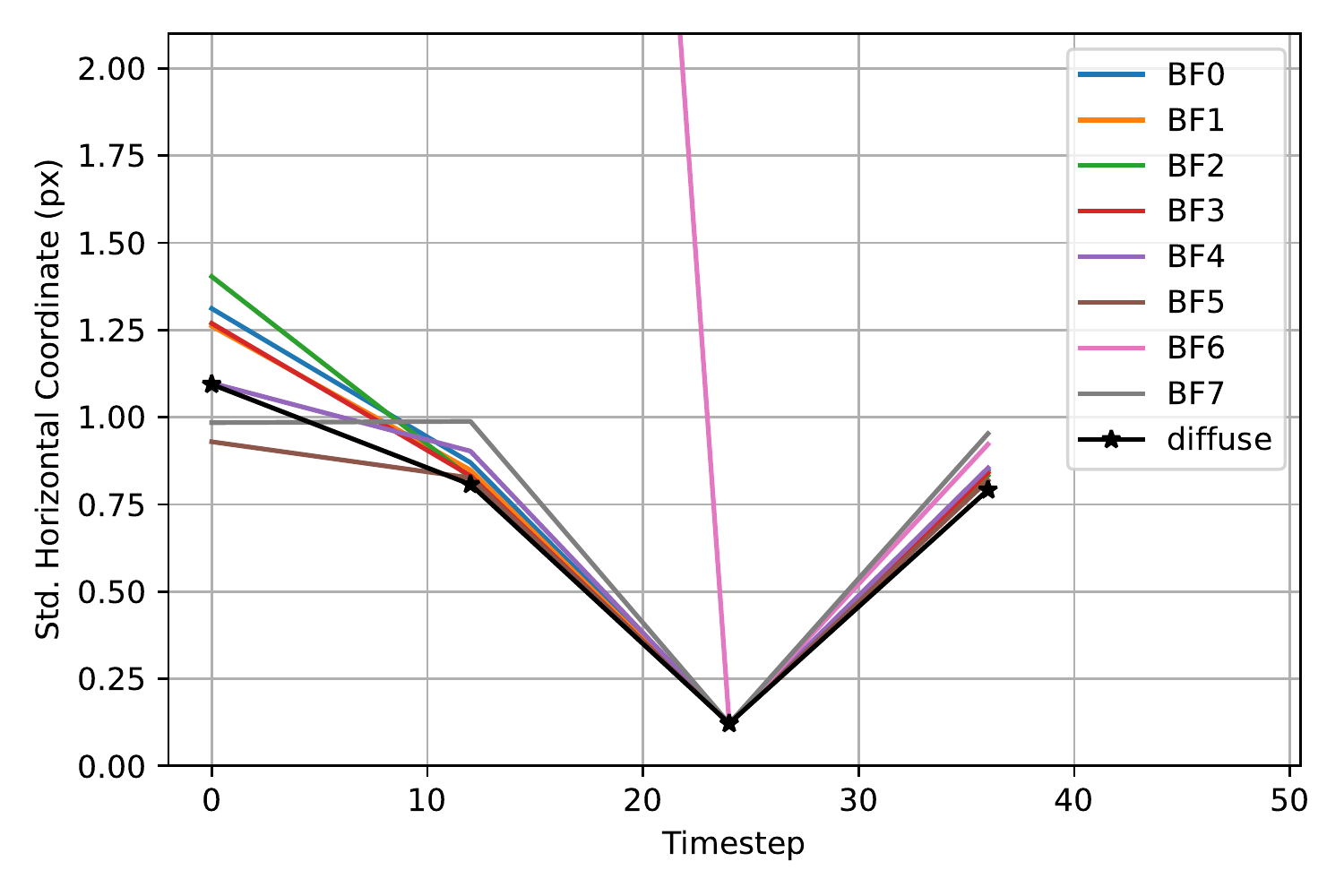}
    }
    \subfigure[Vertical Coordinate]{
        \includegraphics[width=0.48\textwidth]{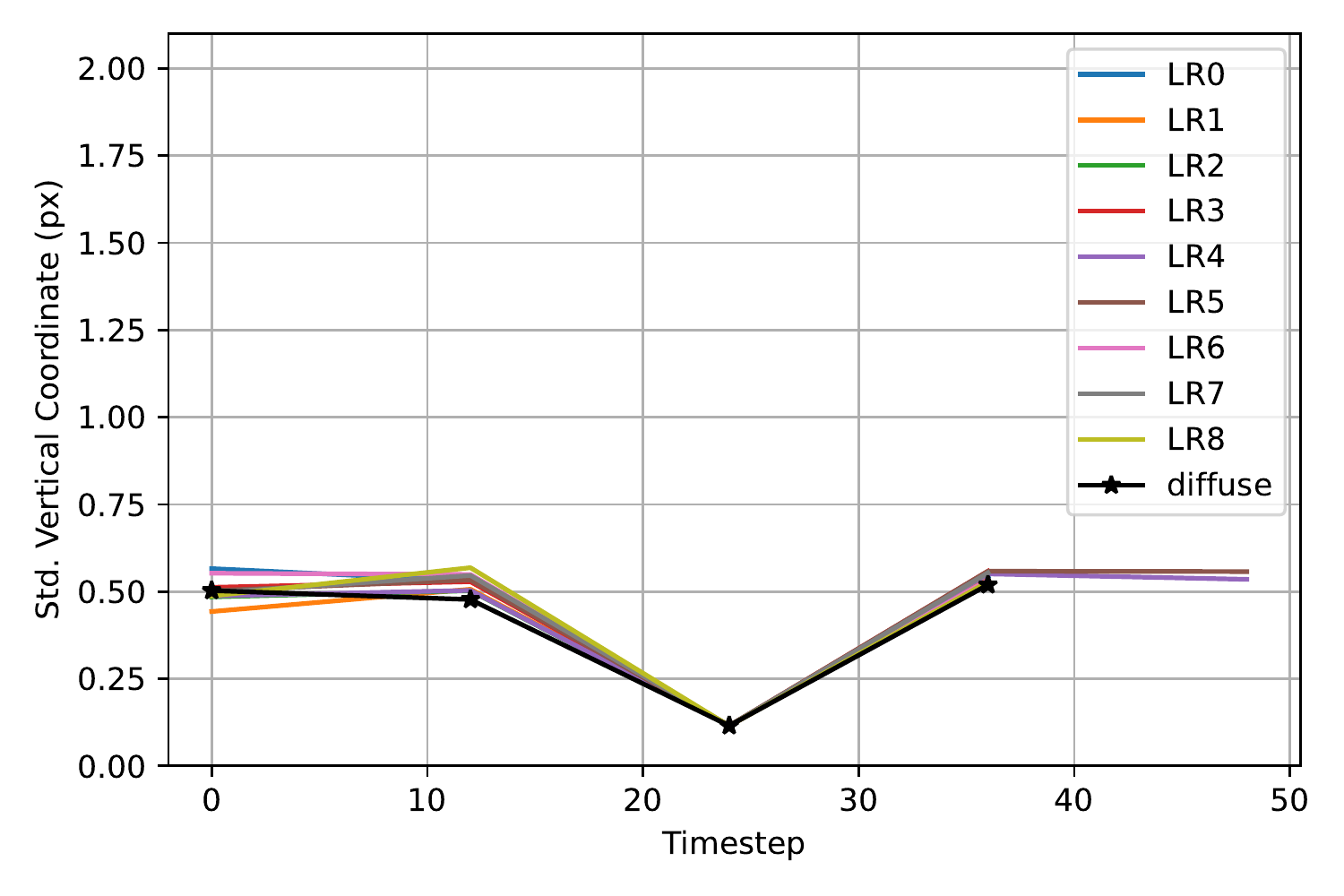}
        \includegraphics[width=0.48\textwidth]{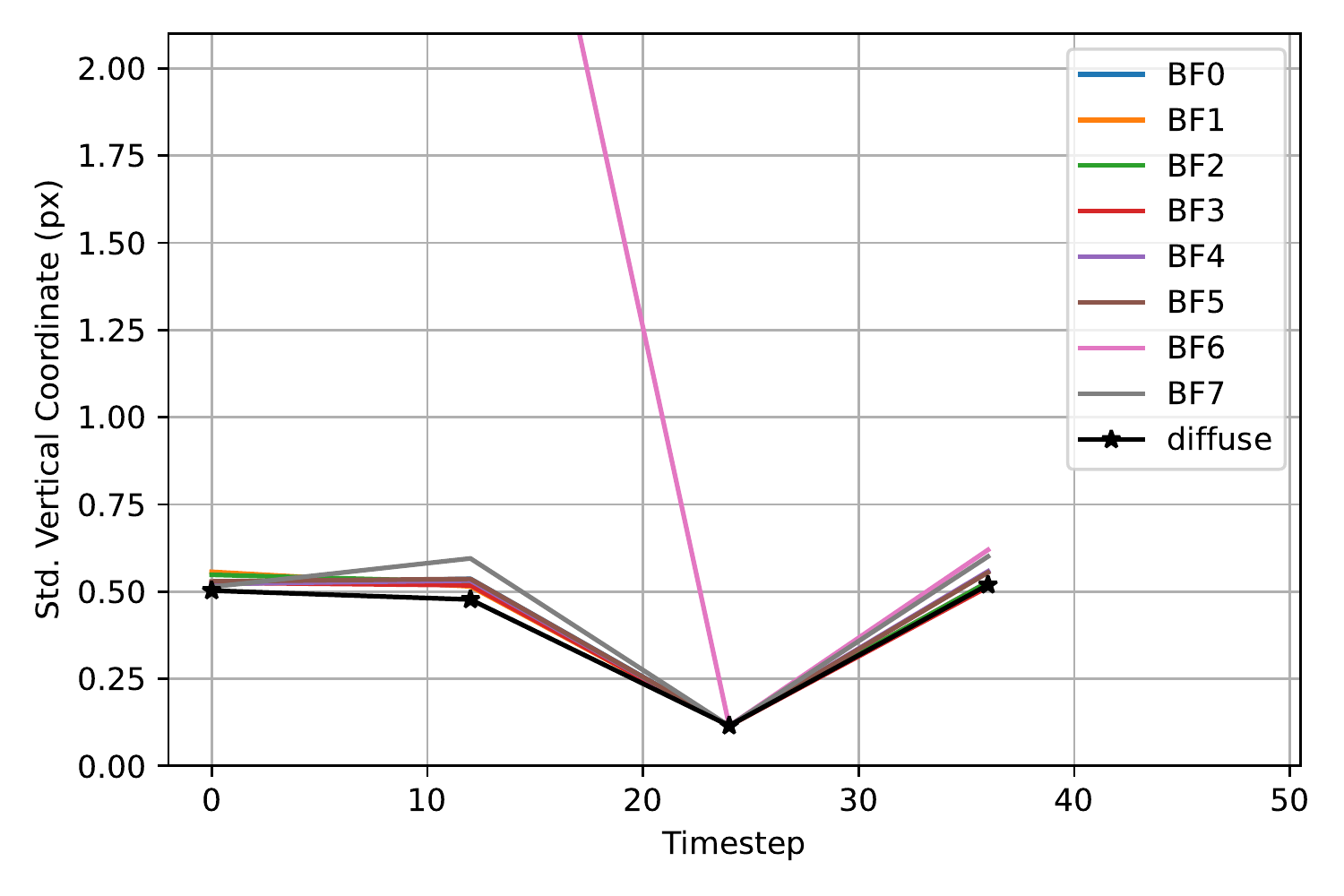}
    }
    \caption{\textbf{DTU Point Features Dataset: At twelve times nominal speed, lighting condition does not change trends in covariance $\Sigma(t)$  when using the Correspondence Tracker.} We compute $\Sigma(t)$ using diffuse lighting (black lines) and each of the directional lighting conditions listed in Figure \ref{fig:dtu_light_stage} using all tracks from all 60 scenes. Timesteps are limited to those with at least 100 features. With the exception of feature track failures in lighting condition BF6, the variation of $\Sigma(t)$ due to the existence of directional lighting is a fraction of the variation common to all plotted lines. The effect of directional lighting is relatively small because changes between adjacent frames are small whether or not the scene contains directional lighting. }
    \label{fig:dtu_match_cov_speed12.00}
\end{figure}

\subsection{Supporting Figures for KITTI Vision Suite}
\label{sec:all_kitti_figs}

\begin{figure}[H]
\centering
\includegraphics[width=\textwidth]{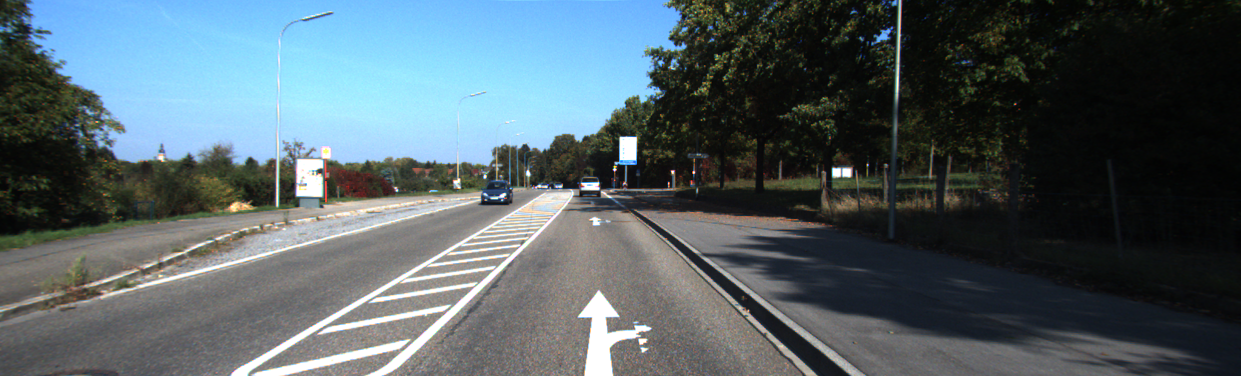}
\includegraphics[width=0.48\textwidth]{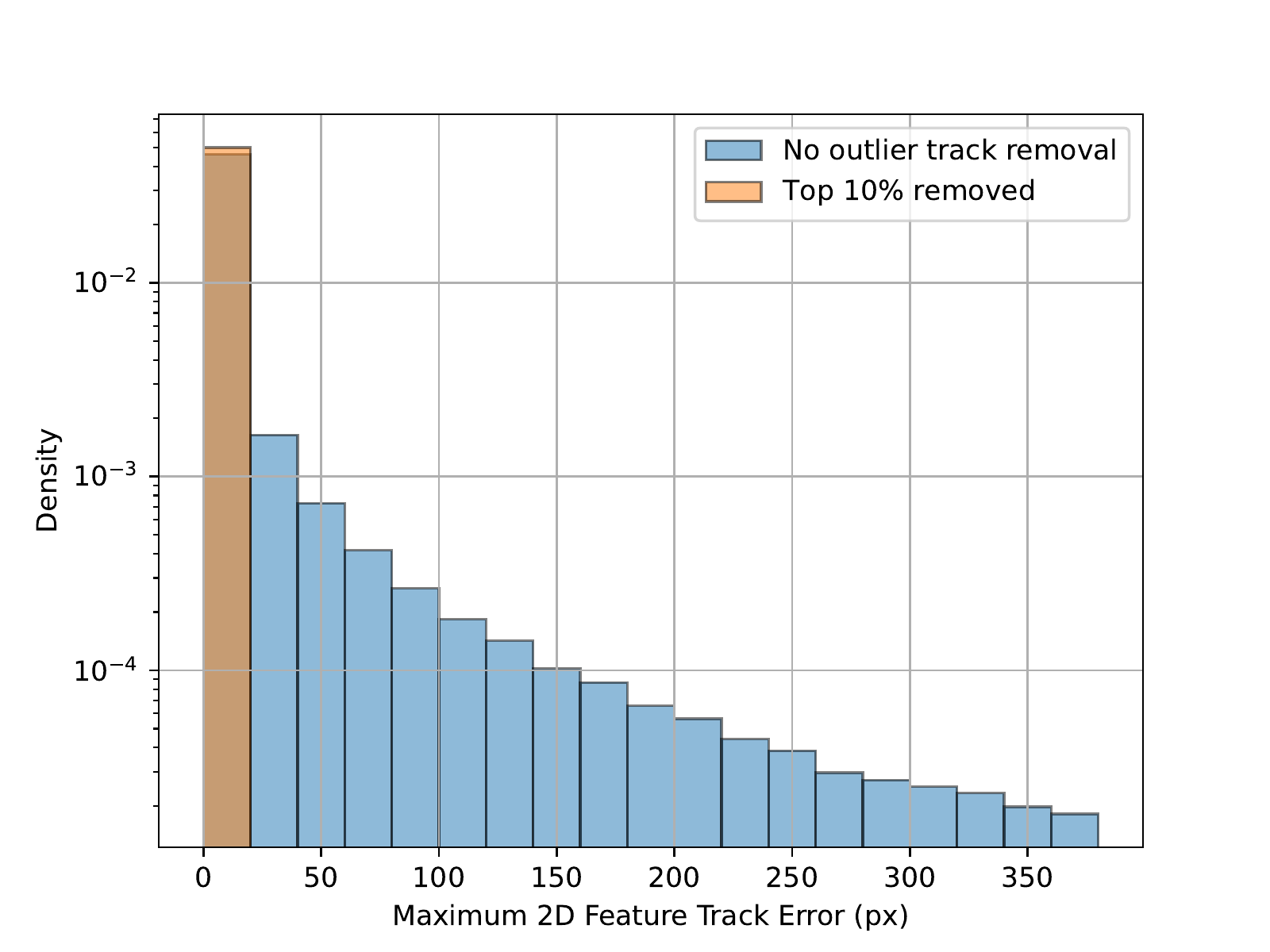}
\caption{\textbf{KITTI Dataset: We will throw out the 10\% of tracks from each scene with the most error.} The bottom figure plots the histogram density of the maximum L2 error of all feature tracks of a single scene in log scale. The corresponding scene is pictured on top. The outlier errors are caused by noisy data in the depth image collection process.}
\label{fig:kitti_error_throwout}
\end{figure}

\begin{figure}[H]
    \centering
    \subfigure[Lucas-Kanade]{\includegraphics[width=0.48\textwidth]{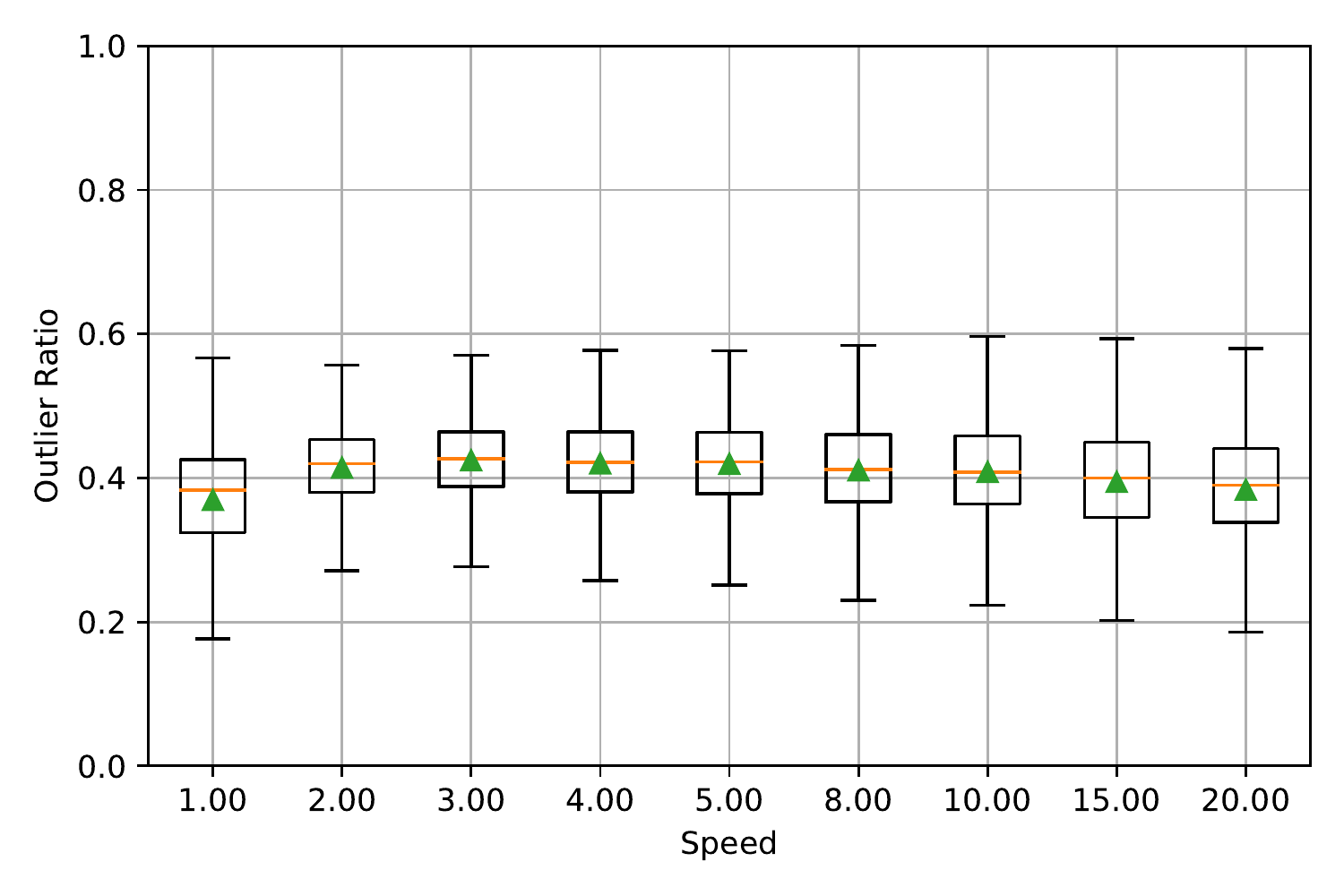}}
    \subfigure[Correspondence]{\includegraphics[width=0.48\textwidth]{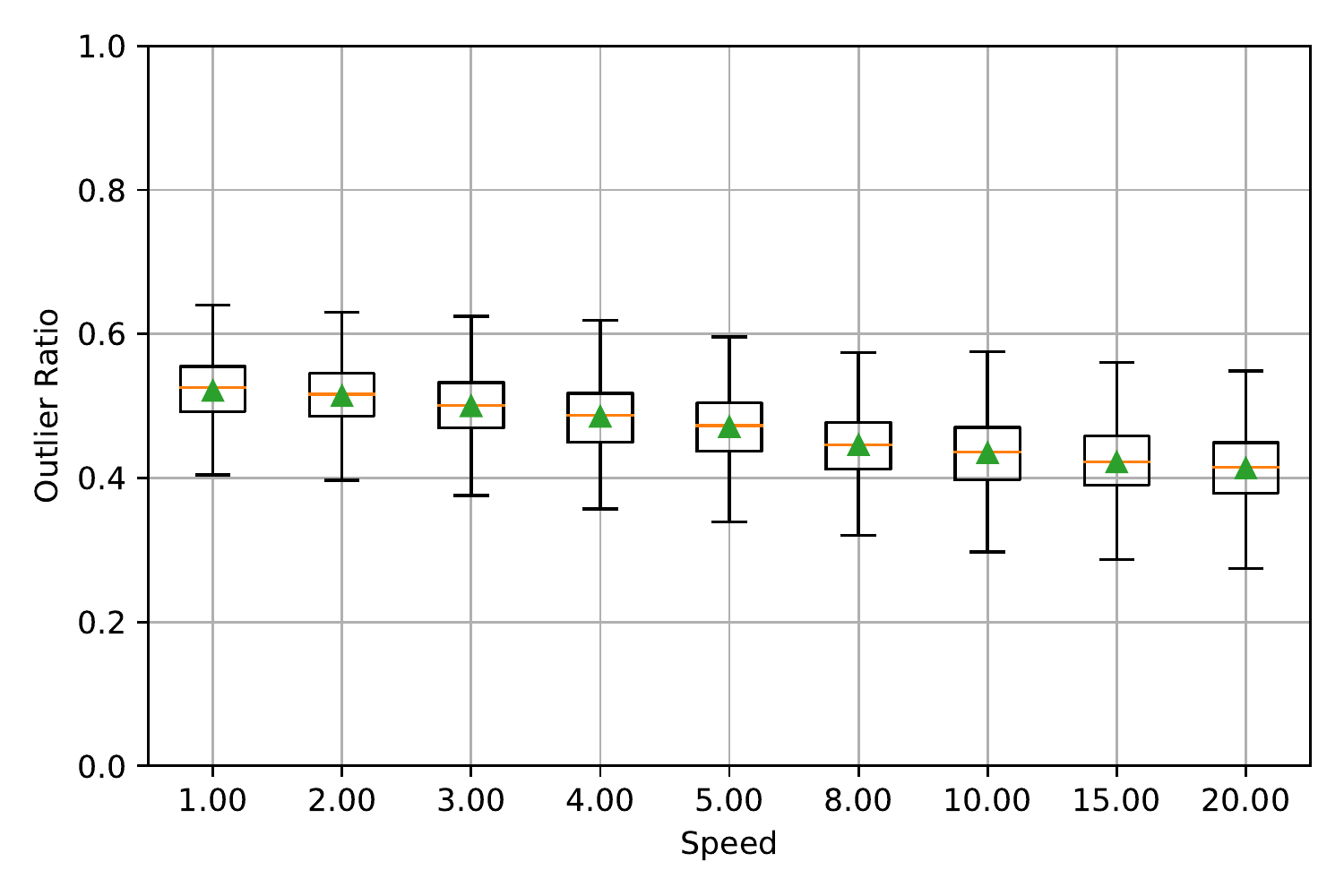}}
    \caption{\textbf{KITTI Dataset: Outlier ratios are above 40 percent.} Outlier ratios per frame are shown as box-and-whisker plots for the Lucas-Kanade tracker on the left and the Correspondence tracker on the right. For the Lucas-Kanade tracker, outlier ratios remain a constant 40 percent. For the correspondence tracker, outlier ratios are higher, around 50 percent, for lower speeds and then decrease. The decreases exists not because of improvements in feature matching with higher speeds, but because fewer features are matched at all.}
    \label{fig:kitti_outlier_ratio}
\end{figure}

\begin{figure}[H]
    \centering
    \subfigure[$\nu(t)$, Horizontal Coordinate]{\includegraphics[width=0.48\textwidth]{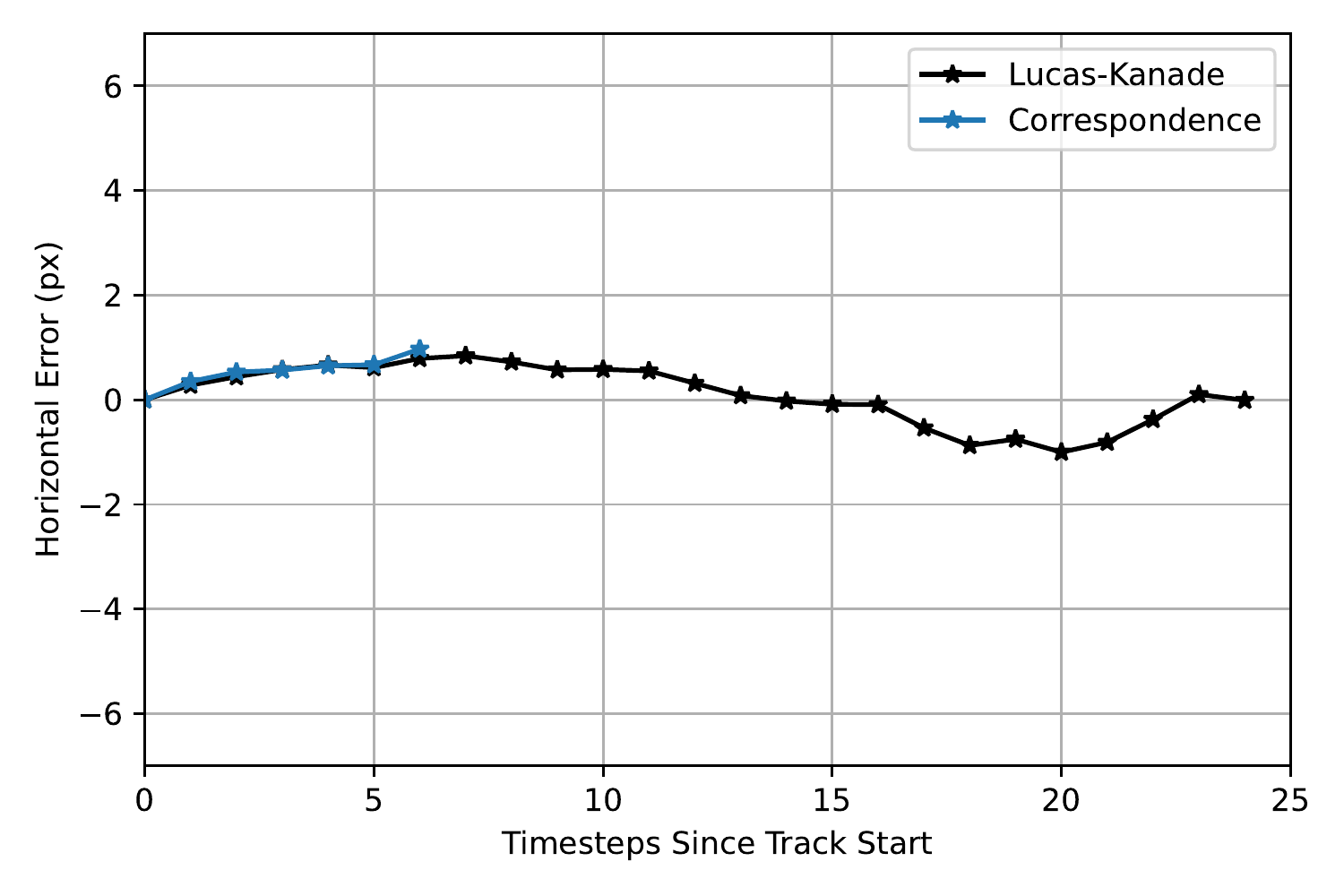}}
    \subfigure[$\nu(t)$, Vertical Coordinate]{\includegraphics[width=0.48\textwidth]{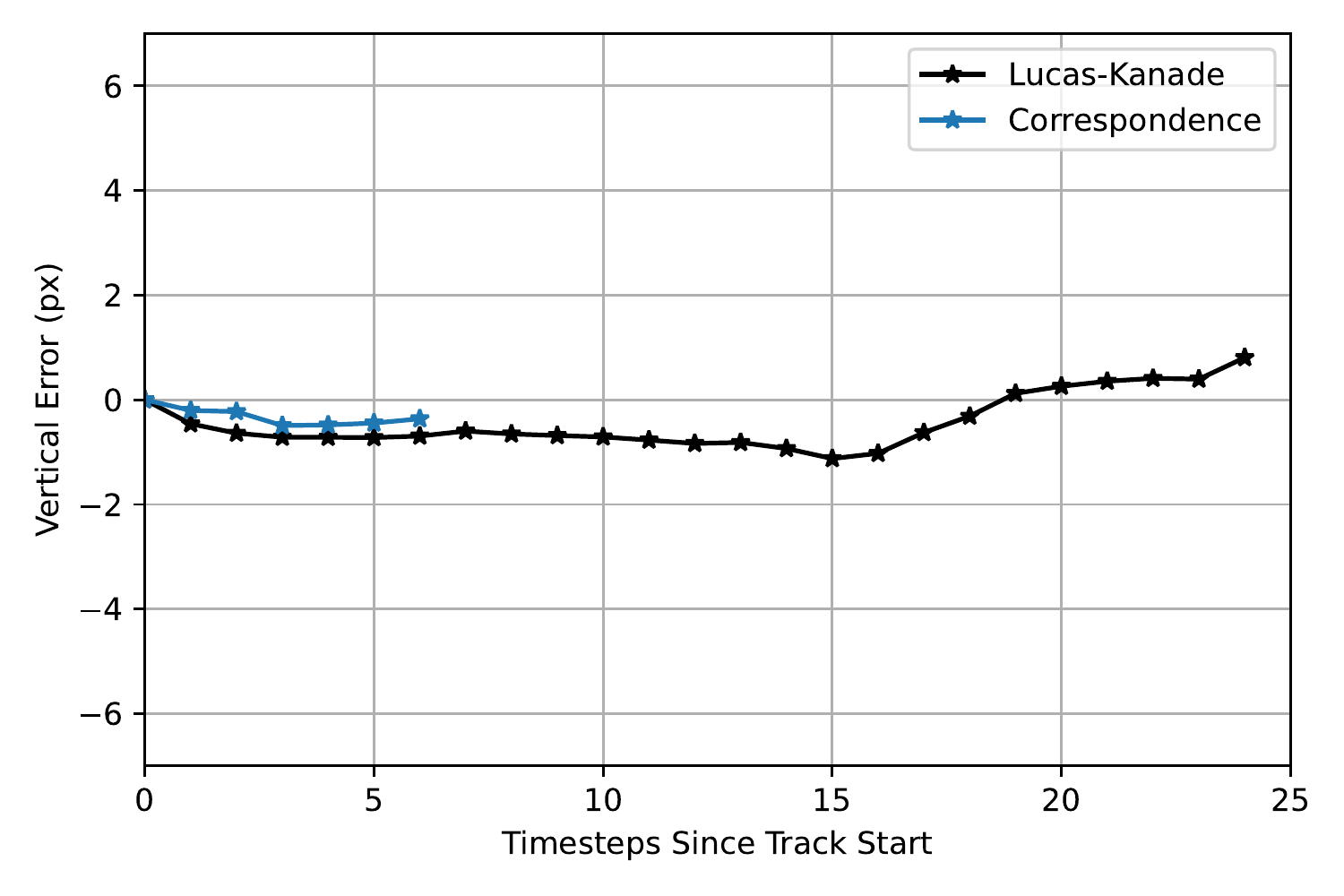}}
    \caption{\textbf{KITTI Dataset: The zero-mean assumption approximately holds for both the Lucas-Kanade Tracker and the Correspondence Tracker at nominal speed.} Lines shown are horizontal (left) and vertical (right) coordinates of mean error $\nu(t)$ calculated using tracks averaged over all scenes; calculation is cutoff at 24 frames for the Lucas-Kanade Tracker and 6 frames for the Correspondence Tracker so that averages can be computed with at least 100 features. Mean errors remain at roughly zero.}
    \label{fig:kitti_1.00_meanerror}
\end{figure}

\begin{figure}[H]
    \centering
    \subfigure[$\eta(t)$, Horizontal Coordinate]{\includegraphics[width=0.48\textwidth]{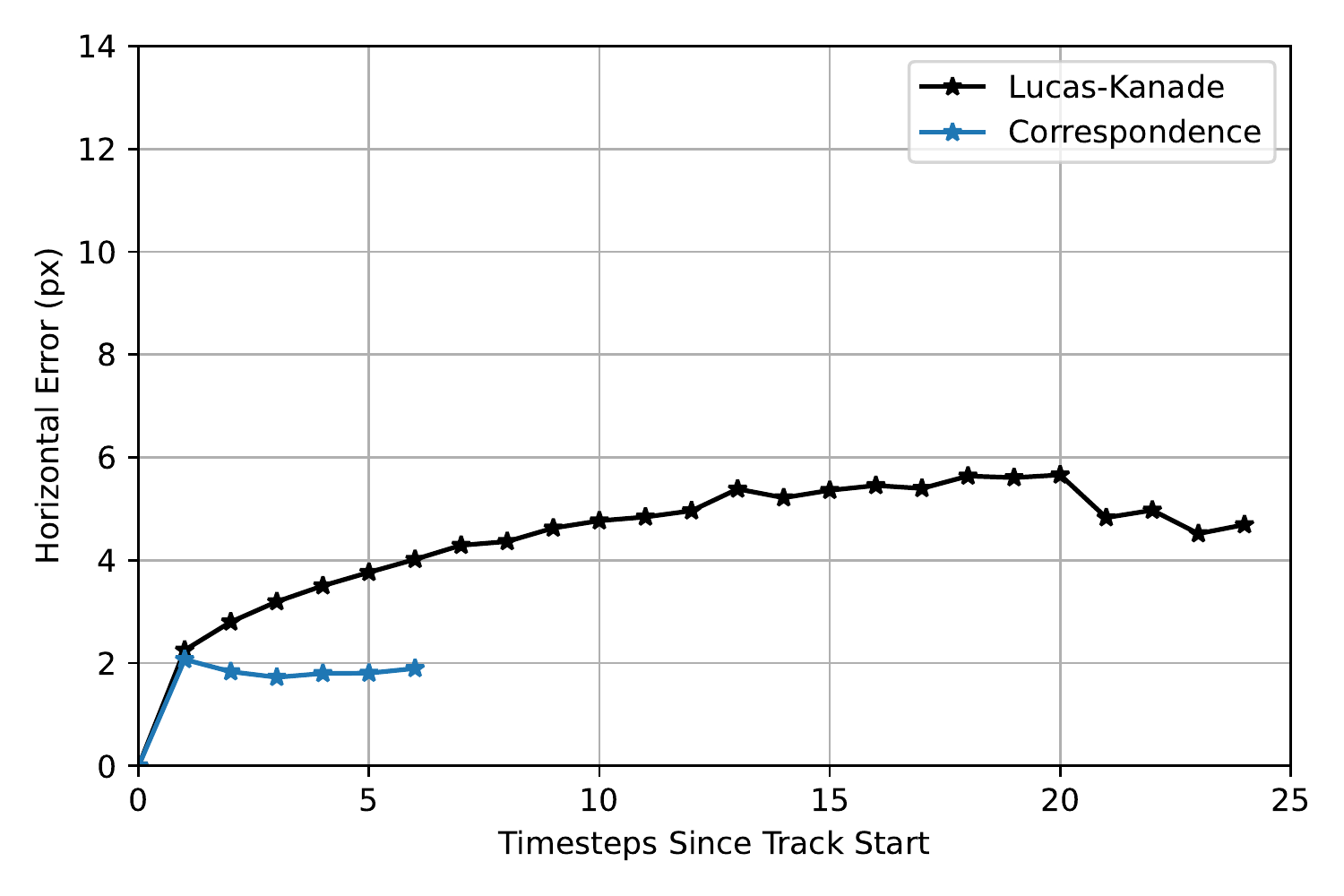}}
    \subfigure[$\eta(t)$, Vertical Coordinate]{\includegraphics[width=0.48\textwidth]{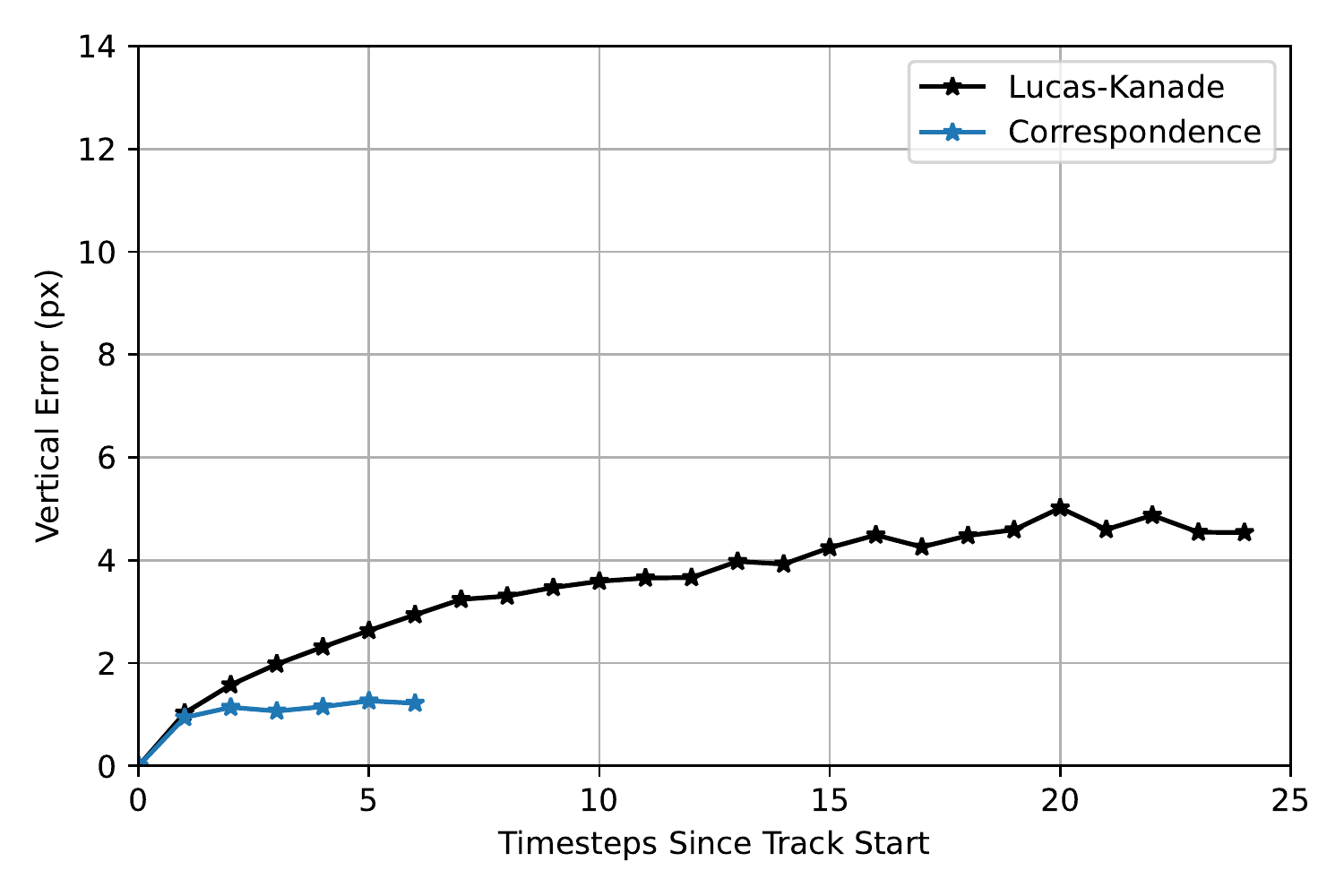}}
    \subfigure[$\Phi(t)$, Horizontal Coordinate]{\includegraphics[width=0.48\textwidth]{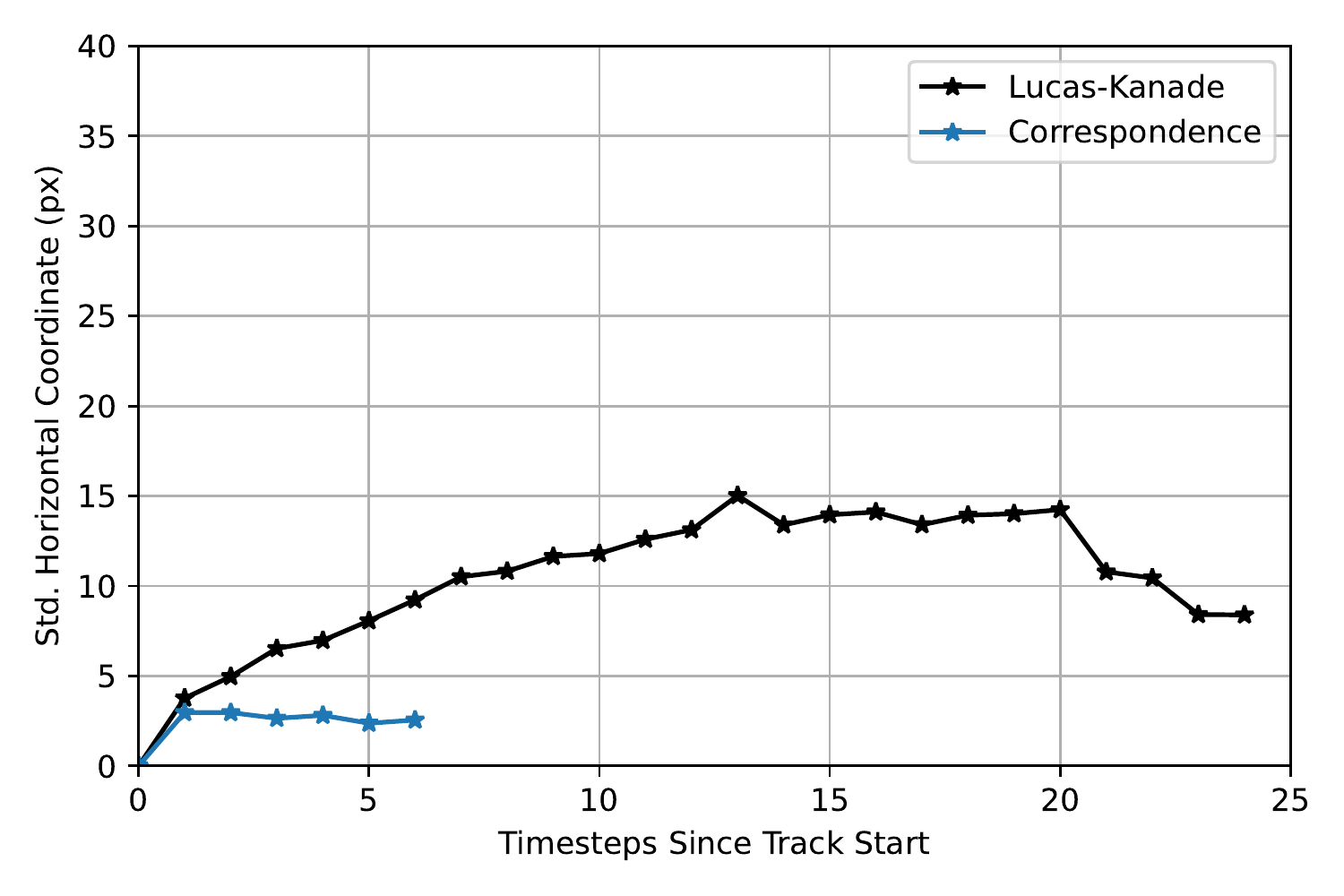}}
    \subfigure[$\Phi(t)$, Vertical Coordinate]{\includegraphics[width=0.48\textwidth]{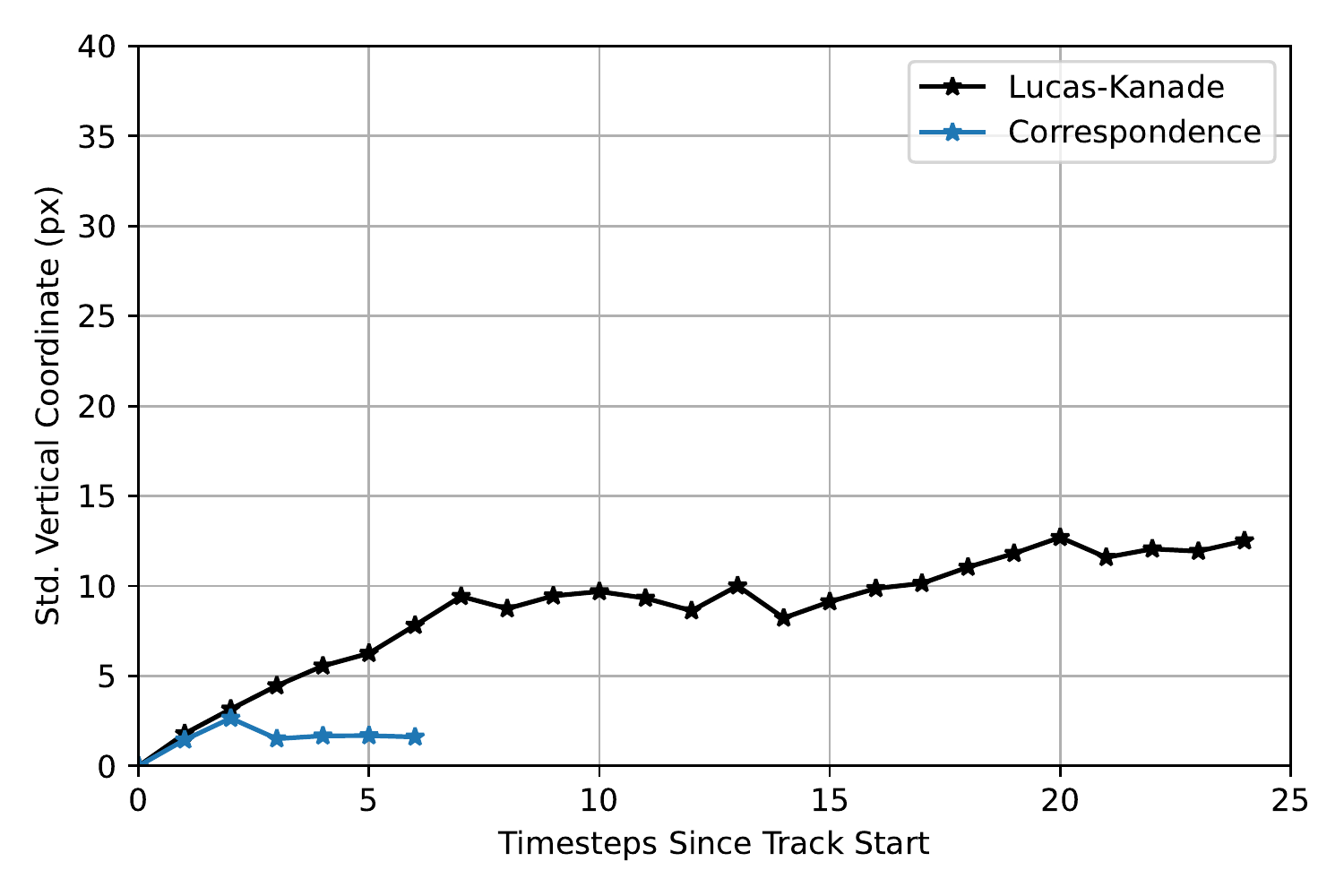}}  
    \caption{\textbf{KITTI Dataset: The Lucas-Kanade Tracker drifts considerably more than the Correspondence Tracker in all directions.} Lines shown are horizontal (left column) and vertical (right column) coordinates of mean absolute error $\eta(t)$ (top row) and covariance $\Phi(t)$ (bottom row) calculated using tracks averaged over all scenes; calculation is cutoff at 24 frames for Lucas-Kanade Tracker and 6 frames for the Correspondence Tracker so that averages can be computed with at least 100 features. Both mean absolute error and covariance are roughly constant when using the Correspondence Tracker. On the other hand, both drift slightly upwards and then level off when using the Lucas-Kanade Tracker.}
    \label{fig:kitti_1.00_error_cov}
\end{figure}

\begin{figure}[H]
    \centering
    \subfigure[$\nu(t)$, Horizontal Coordinate]{
        \includegraphics[width=0.48\textwidth]{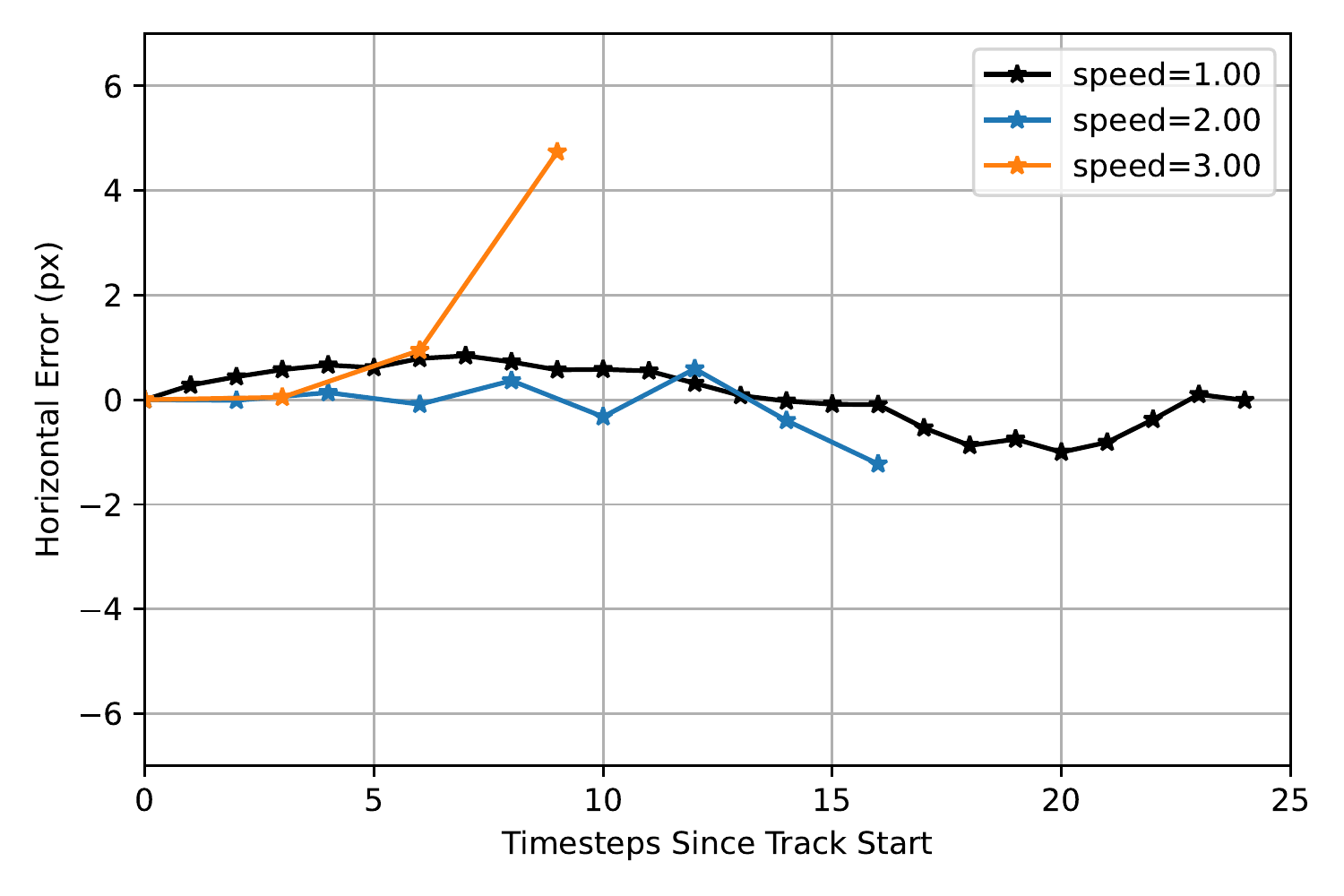}
        \includegraphics[width=0.48\textwidth]{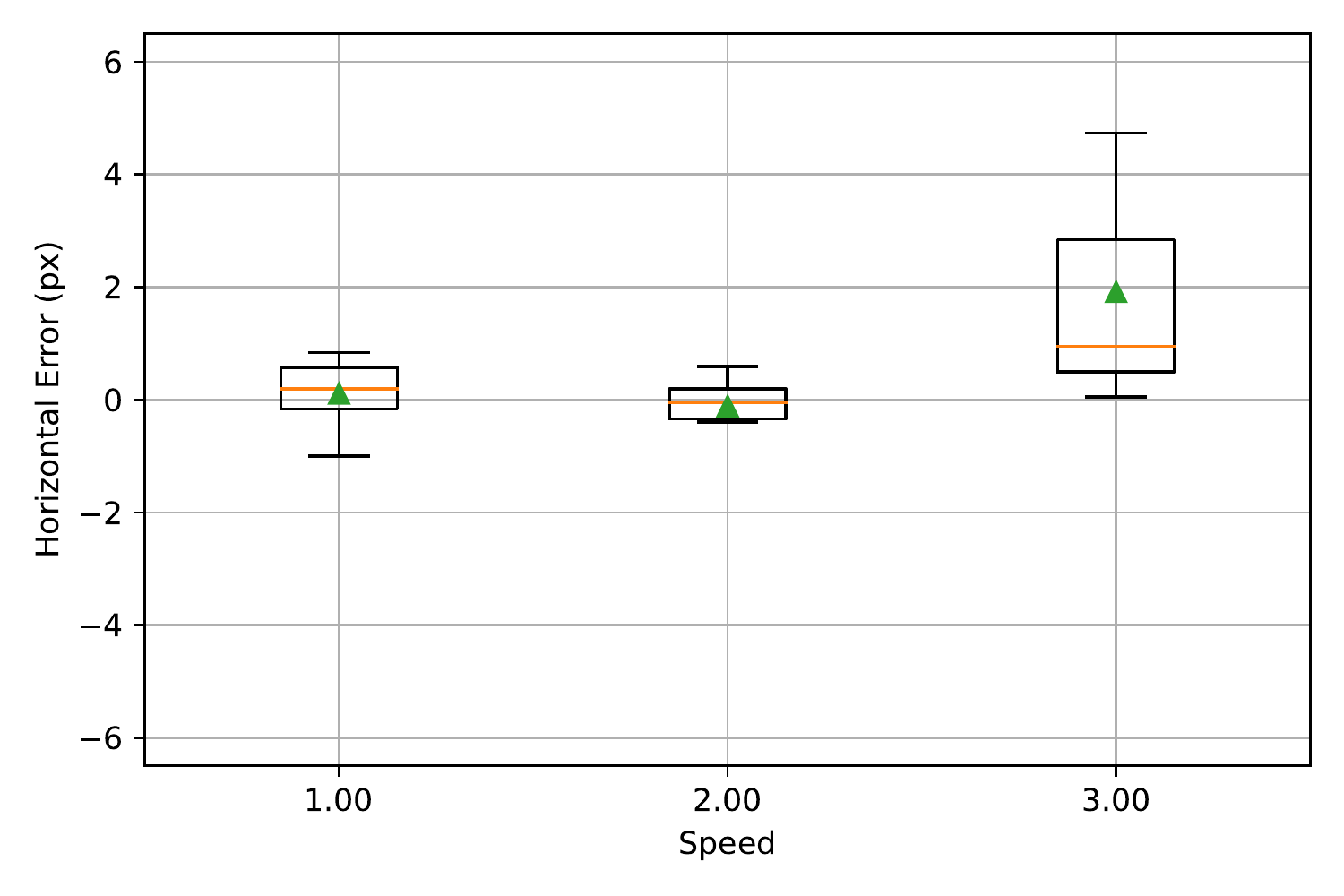}
    }
    \subfigure[$\nu(t)$, Vertical Coordinate]{
        \includegraphics[width=0.48\textwidth]{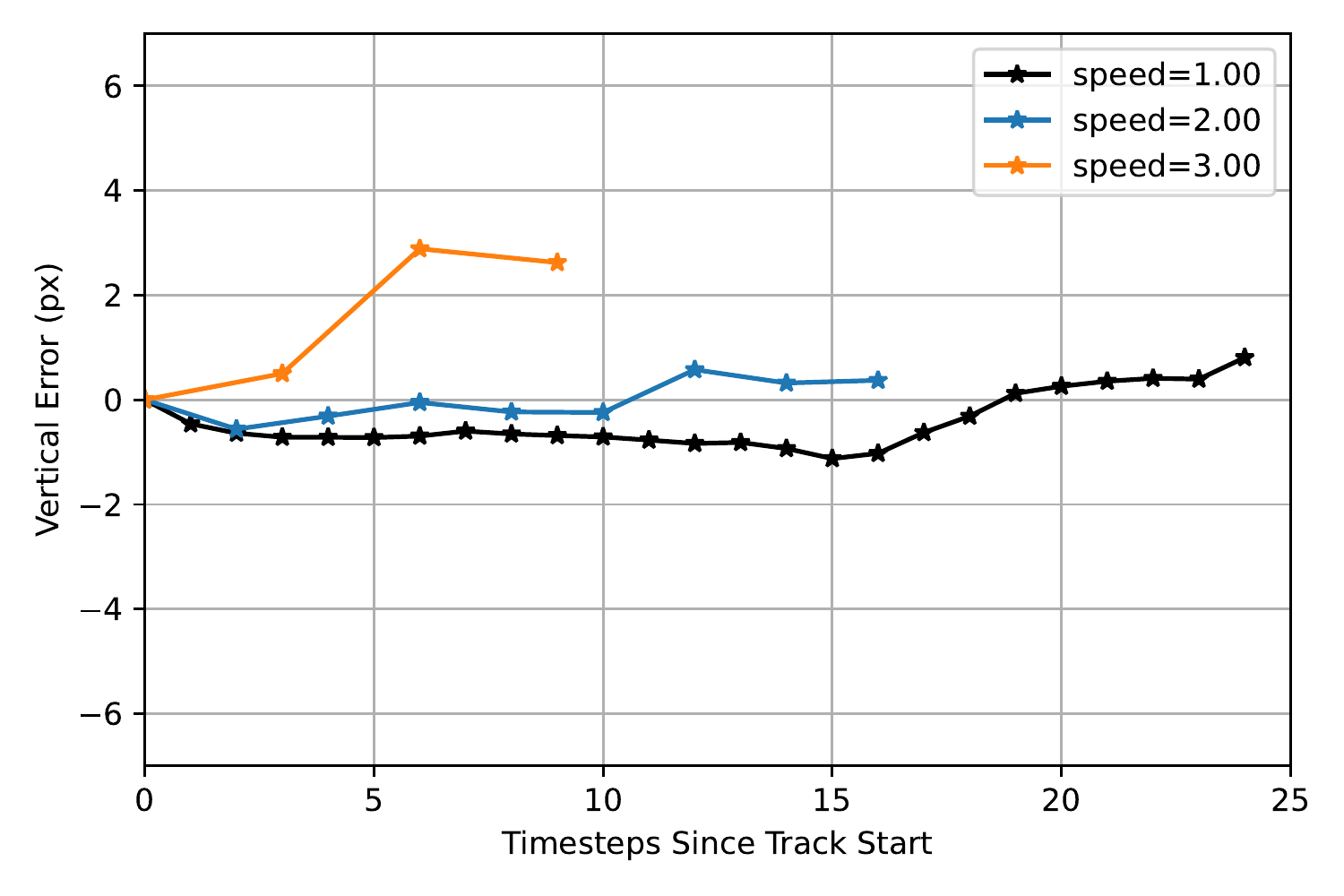}
        \includegraphics[width=0.48\textwidth]{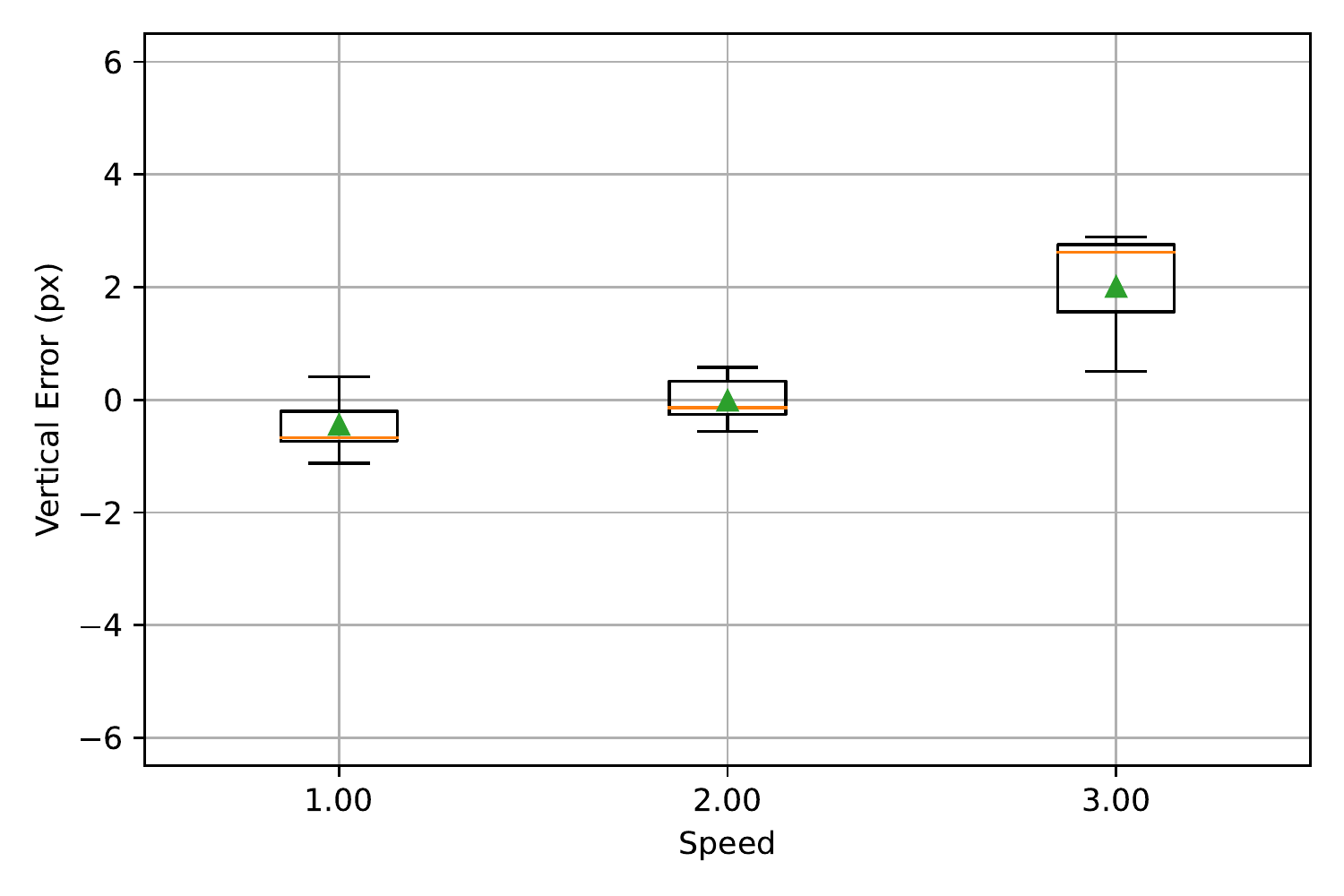}
    }
    \caption{\textbf{KITTI Dataset: Mean tracking errors increase with speed when using the Lucas-Kanade Tracker.}
    The left column contains plots of the horizontal (top row) and vertical (bottom row) components of the mean tracking error $\nu(t)$ at each timestep $t$ after initial feature detection at multiple speeds. Each dot corresponds to a processed frame; lines for higher speeds contain data from fewer frames and therefore show fewer dots. The right column plots the ordinate values of each line for $t>0$ in the left figures as a box plot: means are shown as green triangles and medians are shown as orange lines.
    The mean and median values of the horizontal and vertical coordinates of $\nu(t)$ increases by about two pixels when speed is increased from 2.00 to 3.00. There is no such increase in $\nu(t)$ when speed is increased from 1.00 to 2.00.} \label{fig:kitti_LK_meanerror}
\end{figure}

\begin{figure}[H]
    \centering
    \subfigure[$\eta(t)$, Horizontal Coordinate]{
        \includegraphics[width=0.48\textwidth]{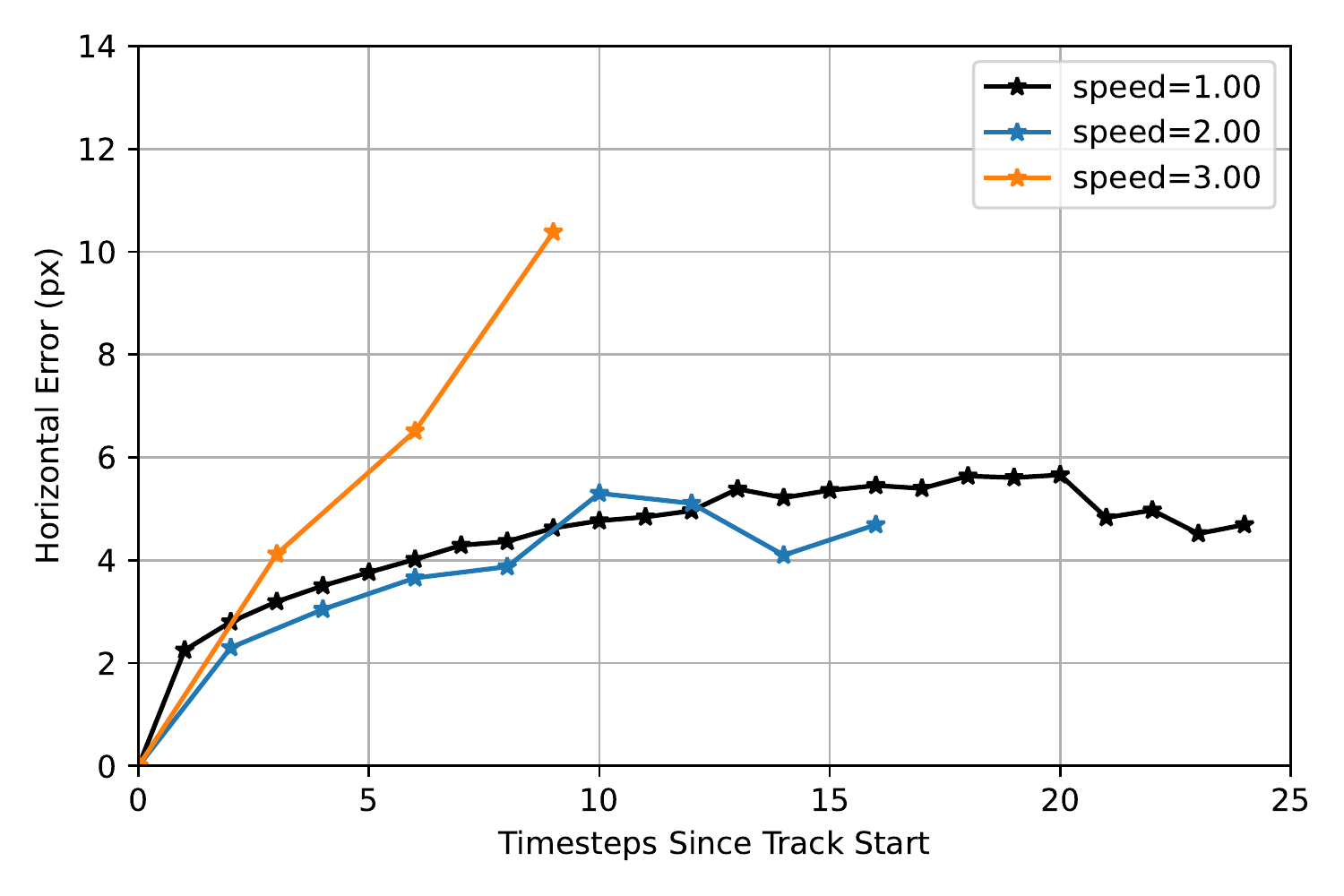}
        \includegraphics[width=0.48\textwidth]{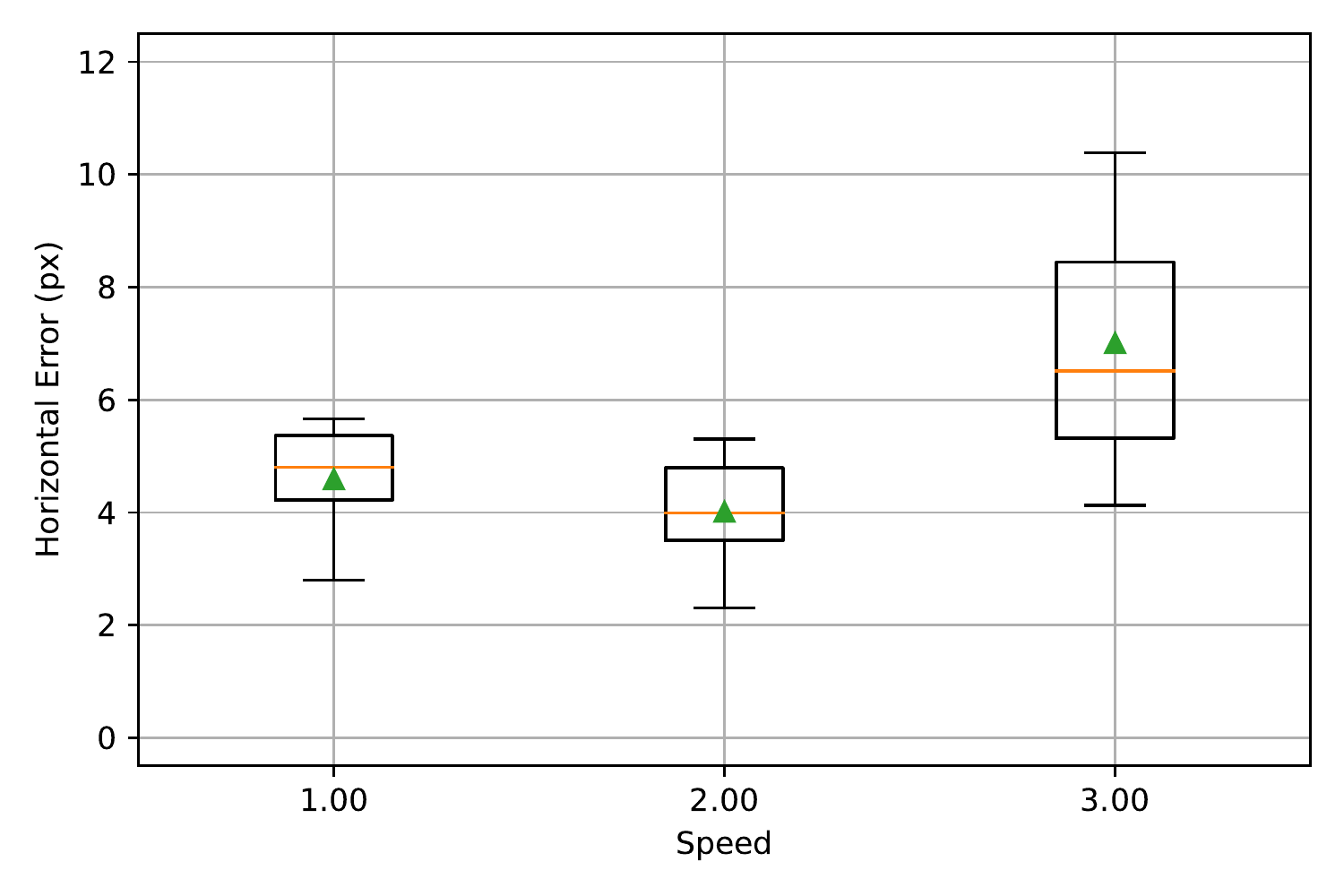}
    }
    \subfigure[$\eta(t)$, Vertical Coordinate]{
        \includegraphics[width=0.48\textwidth]{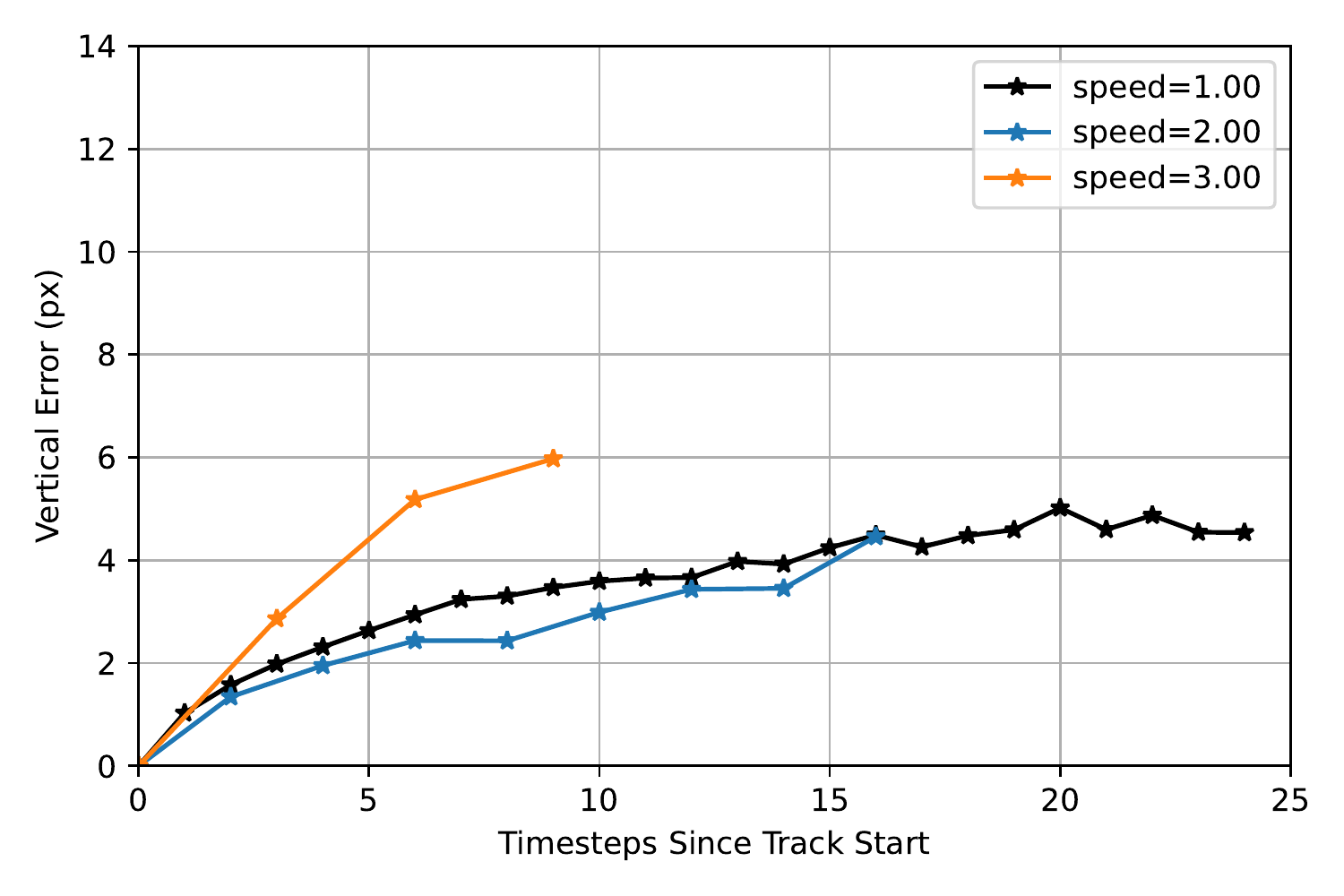}
        \includegraphics[width=0.48\textwidth]{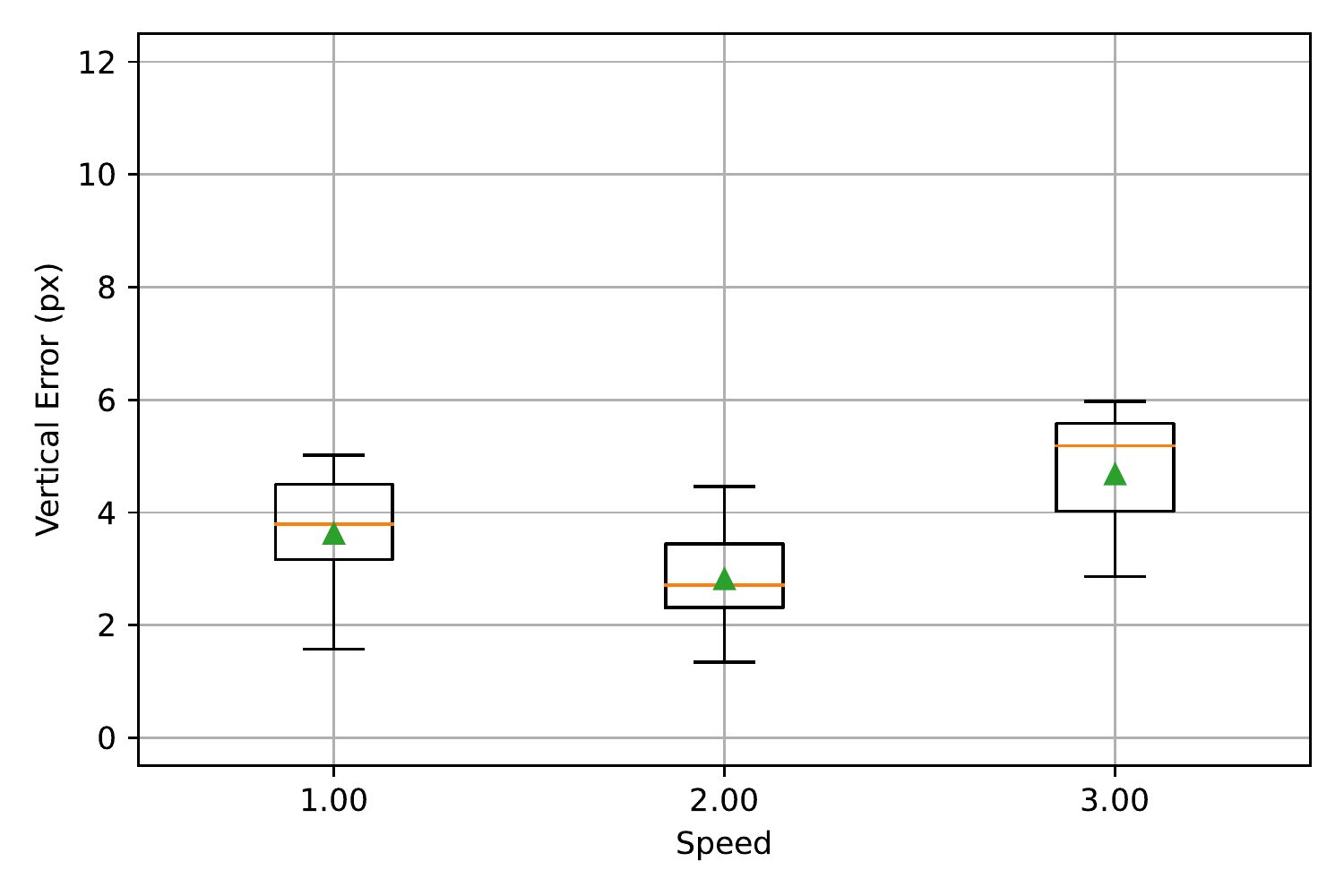}
    } 
    \caption{\textbf{KITTI Dataset: Mean absolute errors increase with speed when using the Lucas-Kanade Tracker.}
    The left column contains plots of the horizontal (top row) and vertical (bottom row) components of the mean absolute error $\eta(t)$ at each timestep $t$ after initial feature detection at multiple speeds. Each dot corresponds to a processed frame; lines for higher speeds contain data from fewer frames and therefore show fewer dots. The right column plots the ordinate values of each line for $t>0$ in the left figures as a box plot: means are shown as green triangles and medians are shown as orange lines.
    The mean and median values of $\eta(t)$ jump when speed is increased from 2.00 to 3.00. Left column plots show that $\eta(t)$ is approximately unchanged when speed is increased from 1.00 to 2.00. Since the box plot for speed=1.00 contains more points at larger values of $t$ than the box plot for speed=2.00, the mean and median values in the box plot decrease.
    }
    \label{fig:kitti_LK_MAE}
\end{figure}

\begin{figure}[H]
    \centering
    \subfigure[$\Phi(t)$, Horizontal Coordinates]{
        \includegraphics[width=0.48\textwidth]{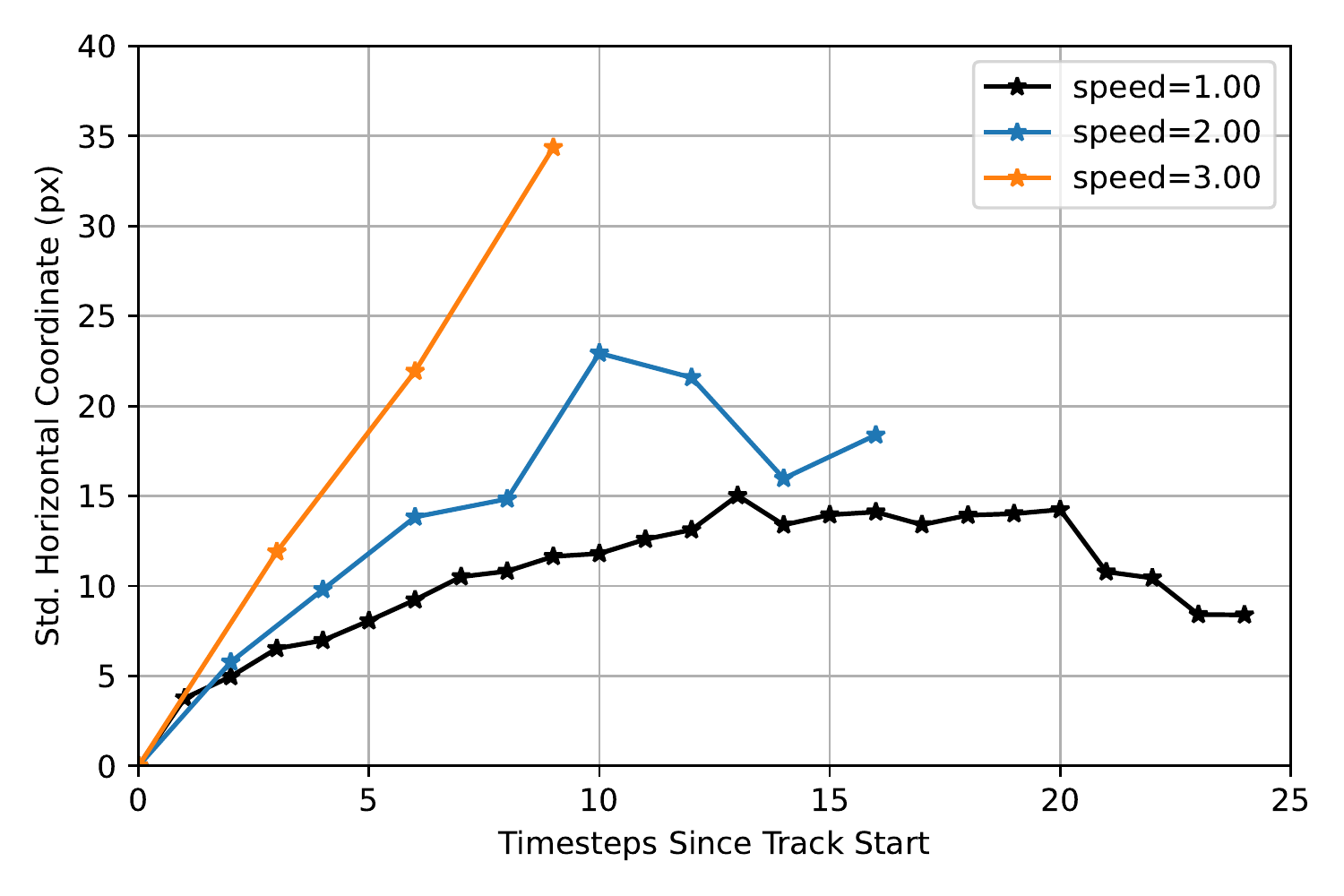}
        \includegraphics[width=0.48\textwidth]{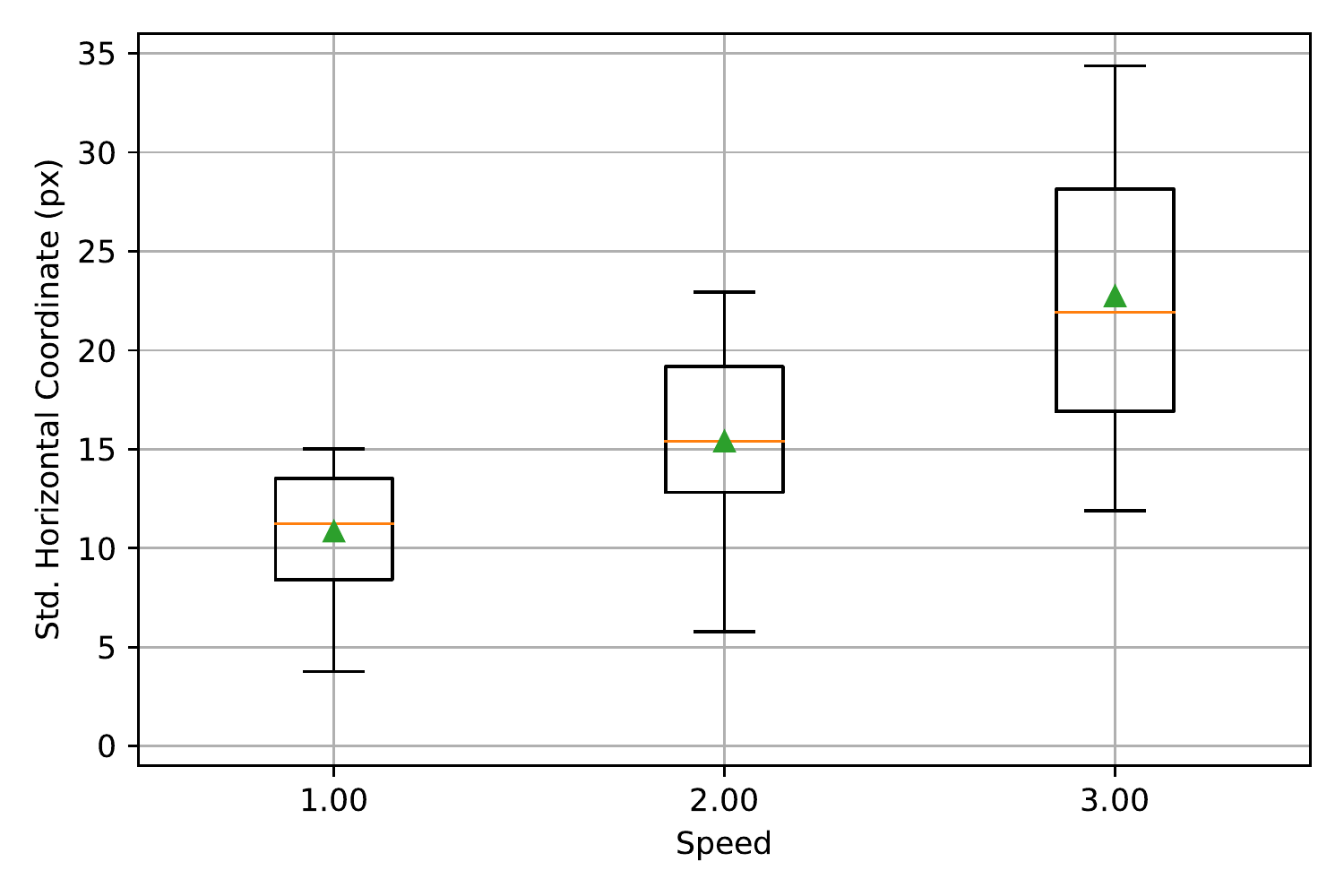}
    }
    \subfigure[$\Phi(t)$, Vertical Coordinates]{
        \includegraphics[width=0.48\textwidth]{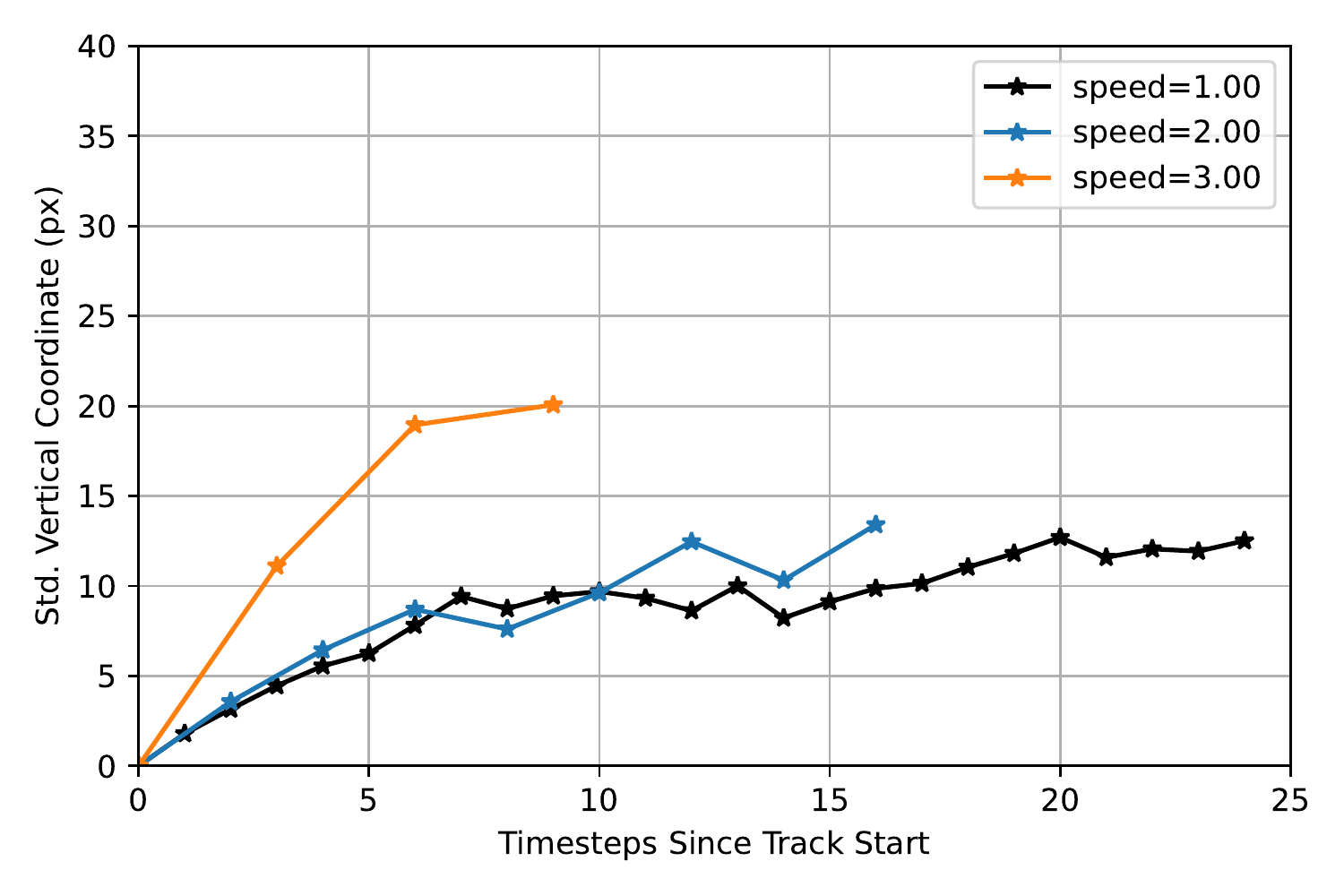}
        \includegraphics[width=0.48\textwidth]{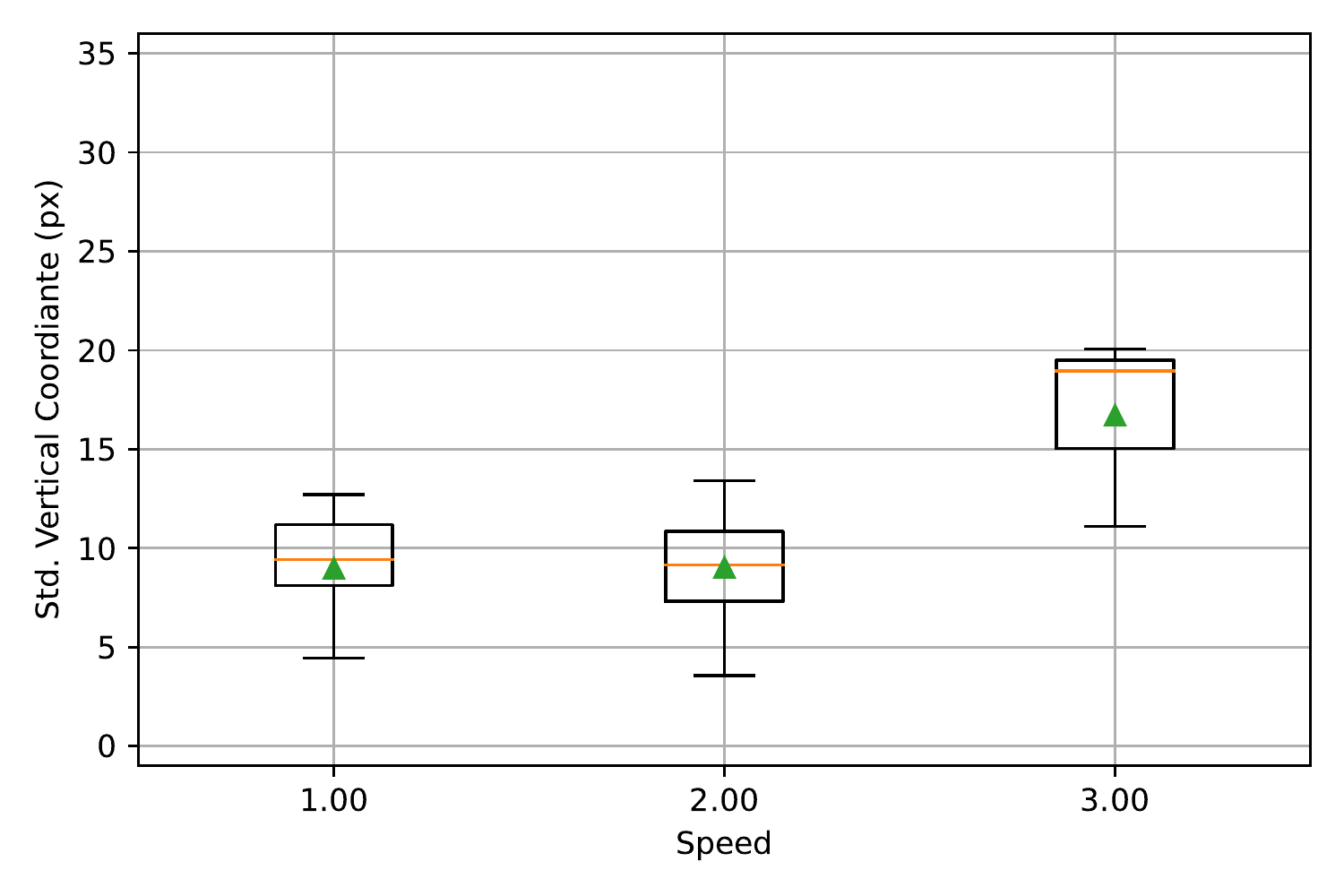}
    }
    \caption{\textbf{KITTI Dataset: Covariances increase with speed when using the Lucas-Kanade Tracker.}
    The left column contains plots of the horizontal (top row) and vertical (bottom row) components of the covariance $\Phi(t)$ at each timestep $t$ after initial feature detection at multiple speeds. Each dot corresponds to a processed frame; lines for higher speeds contain data from fewer frames and therefore show fewer dots. The right column plots the ordinate values of each line for $t>0$ in the left figures as a box plot: means are shown as green triangles and medians are shown as orange lines.
    We see a linear increase in covariance in the horizontal coordinate with speed. The increase in the vertical coordinate follows the same trend noted in Figures \ref{fig:kitti_LK_meanerror} and \ref{fig:kitti_LK_MAE}.
    }
    \label{fig:kitti_LK_cov}
\end{figure}

\begin{figure}[H]
    \centering
    \subfigure[$\nu(t)$, Horizontal Coordinate]{
        \includegraphics[width=0.48\textwidth]{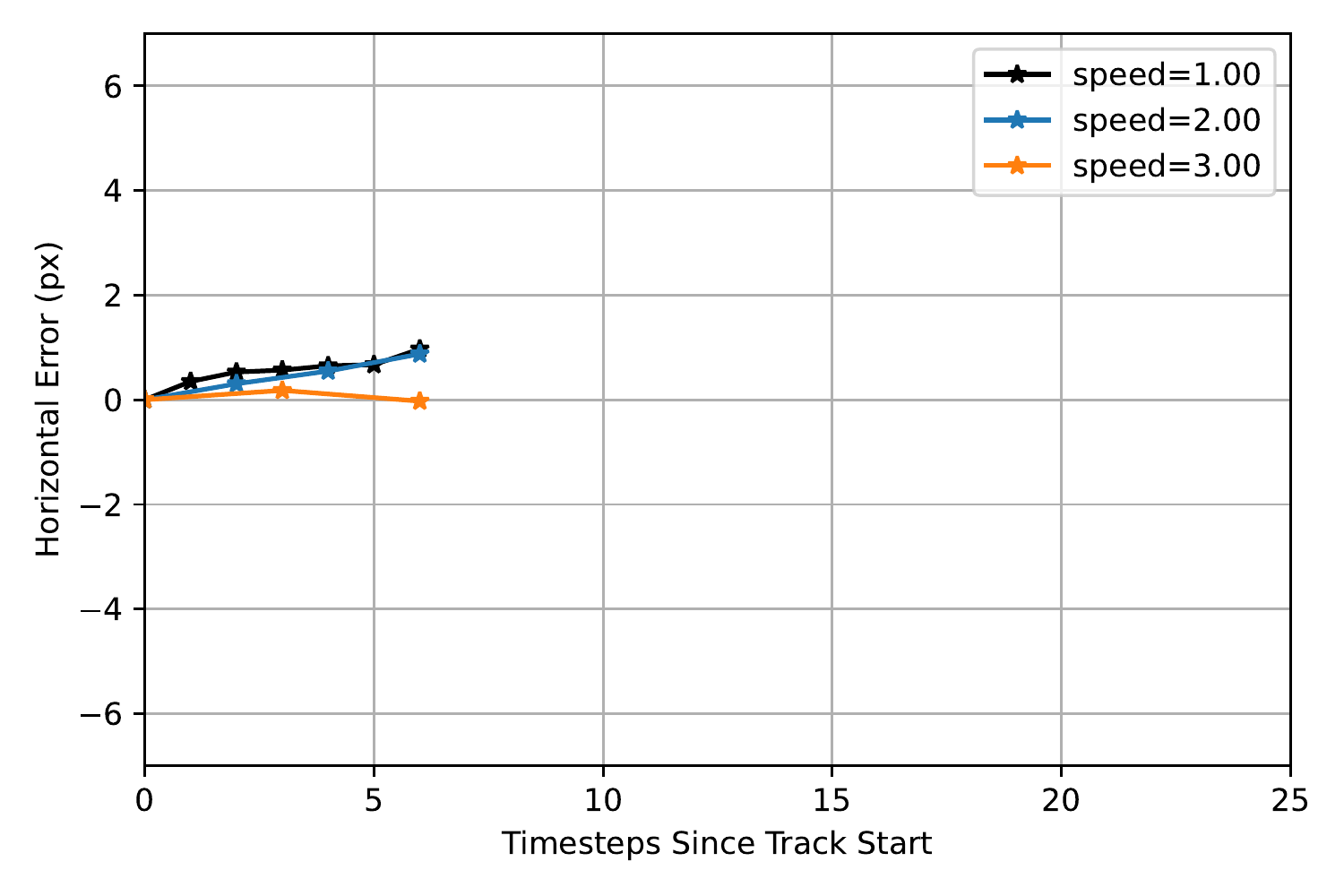}
        \includegraphics[width=0.48\textwidth]{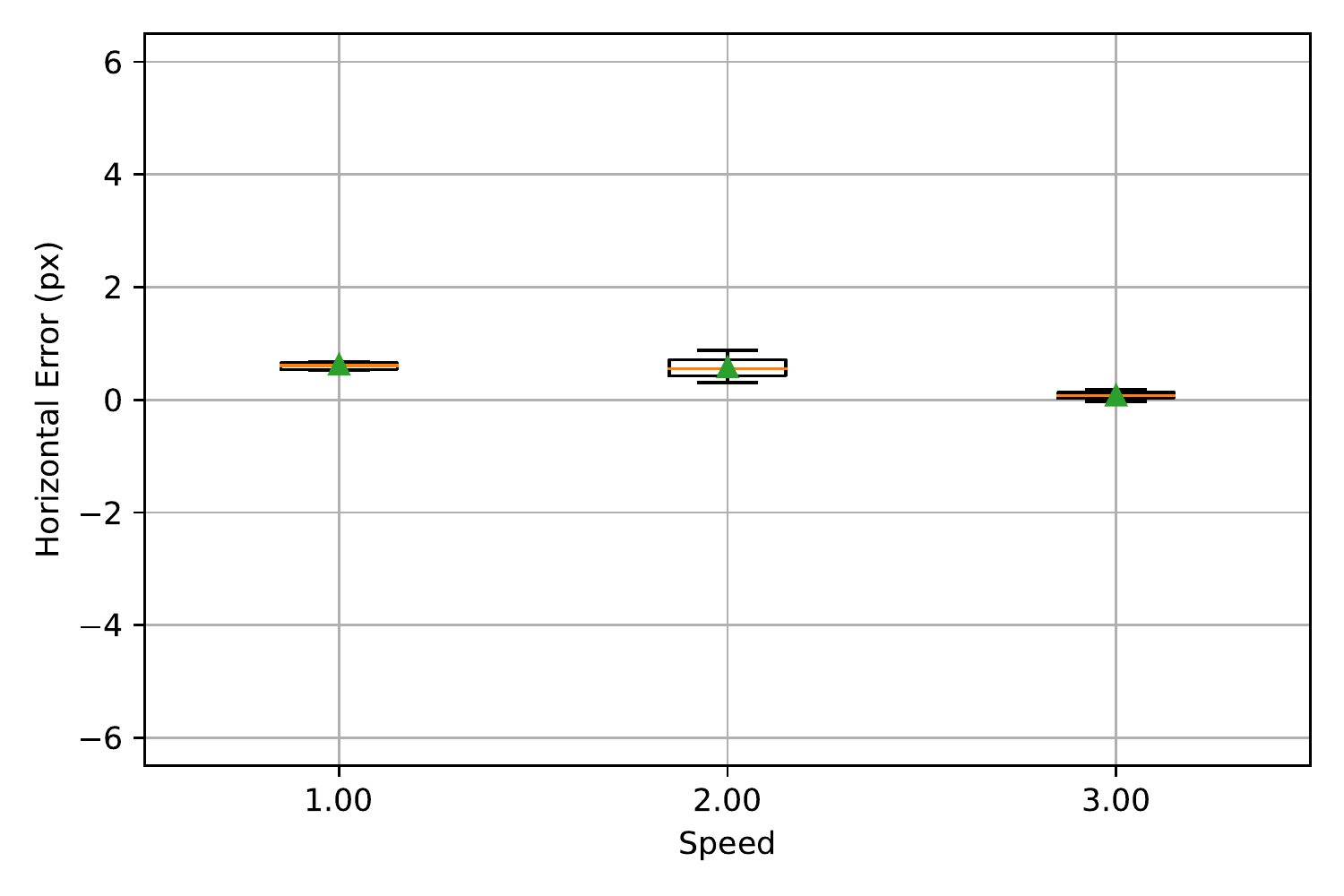}
    }
    \subfigure[$\nu(t)$, Vertical Coordinate]{
        \includegraphics[width=0.48\textwidth]{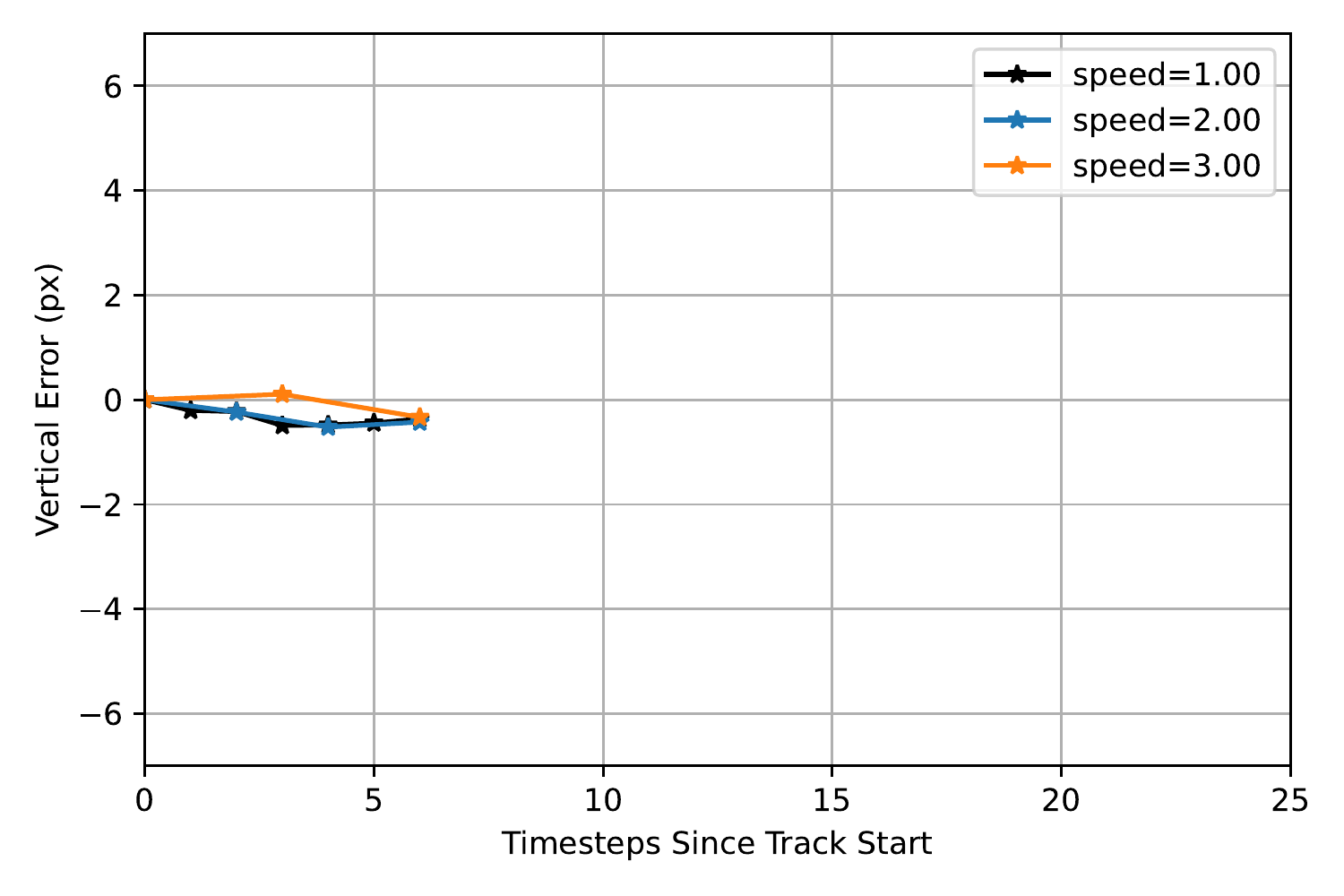}
        \includegraphics[width=0.48\textwidth]{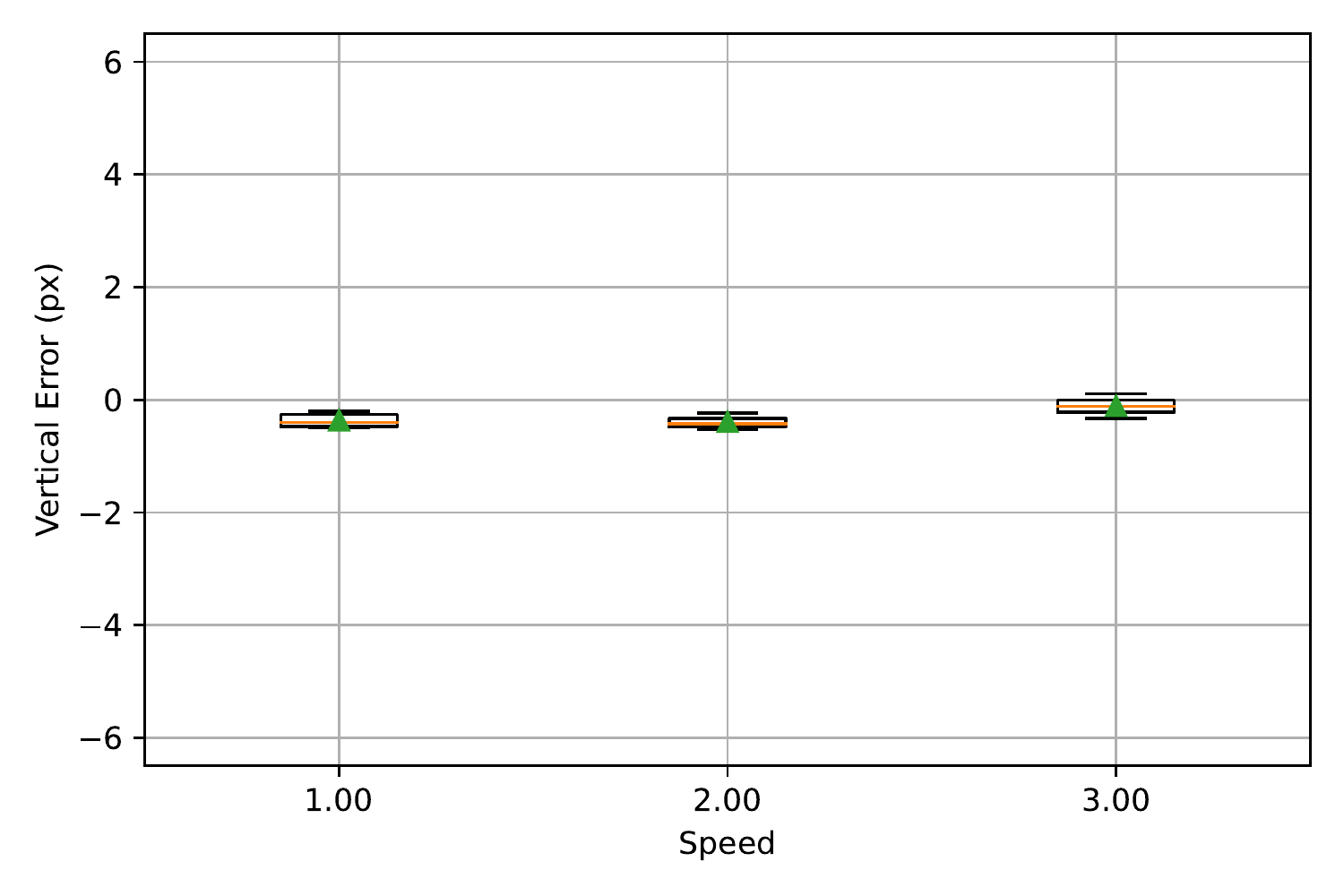}
    }
    \caption{\textbf{KITTI Dataset: Mean errors are unaffected by speed when using the Correspondence Tracker.}
    The left column contains plots of the horizontal (top row) and vertical (bottom row) components of the mean tracking error $\nu(t)$ at each timestep $t$ after initial feature detection at multiple speeds. Each dot corresponds to a processed frame; lines for higher speeds contain data from fewer frames and therefore show fewer dots. The right column plots the ordinate values of each line for $t>0$ in the left figures as a box plot: means are shown as green triangles and medians are shown as orange lines. Compared to the results for the Lucas-Kanade Tracker in Figure \ref{fig:kitti_LK_meanerror}, mean errors do not change when speed is increased from 1.00 to 3.00.
    }
    \label{fig:kitti_match_meanerror}
\end{figure}

\begin{figure}[H]
    \centering
    \subfigure[$\eta(t)$, Horizontal Coordinate]{
        \includegraphics[width=0.48\textwidth]{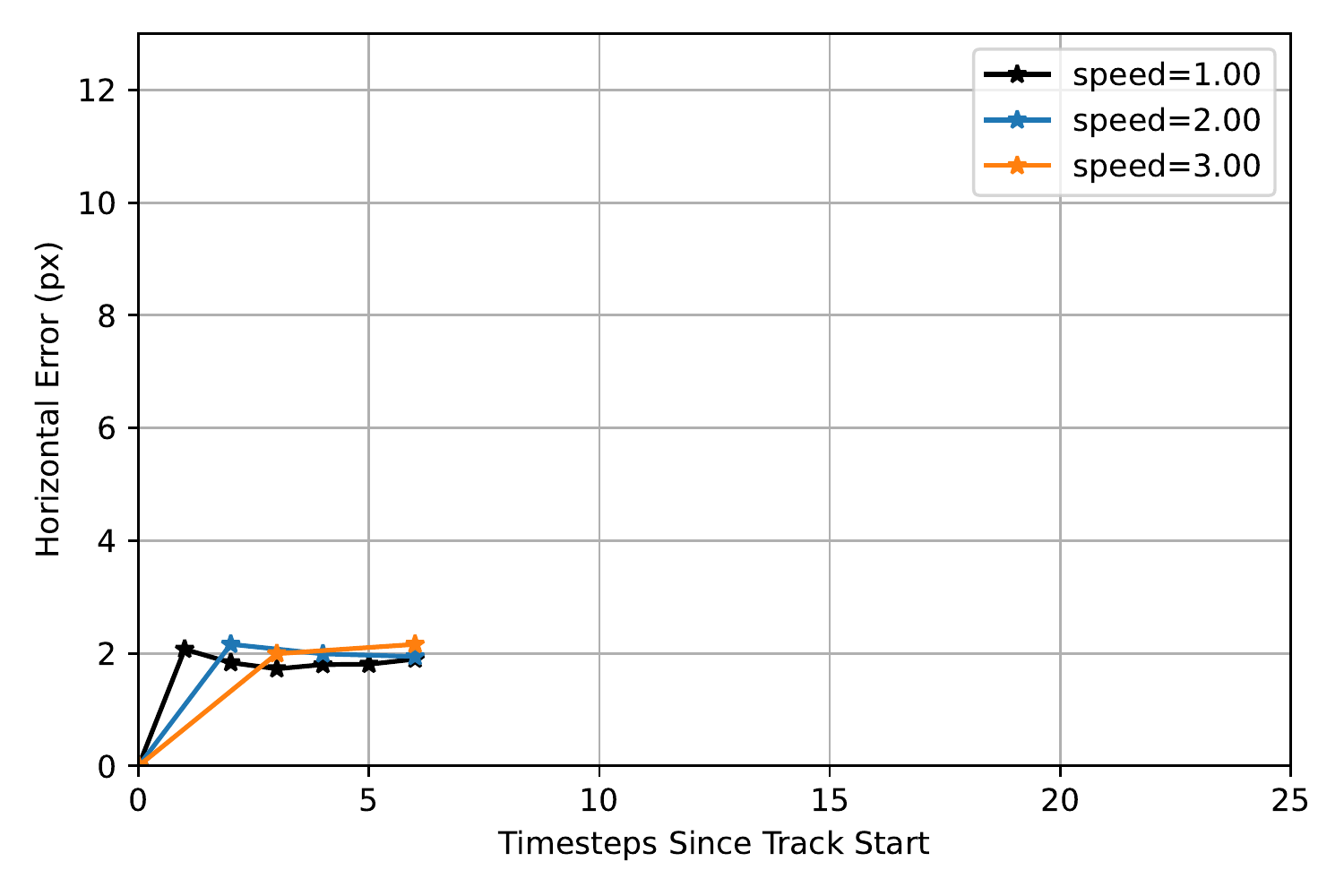}
        \includegraphics[width=0.48\textwidth]{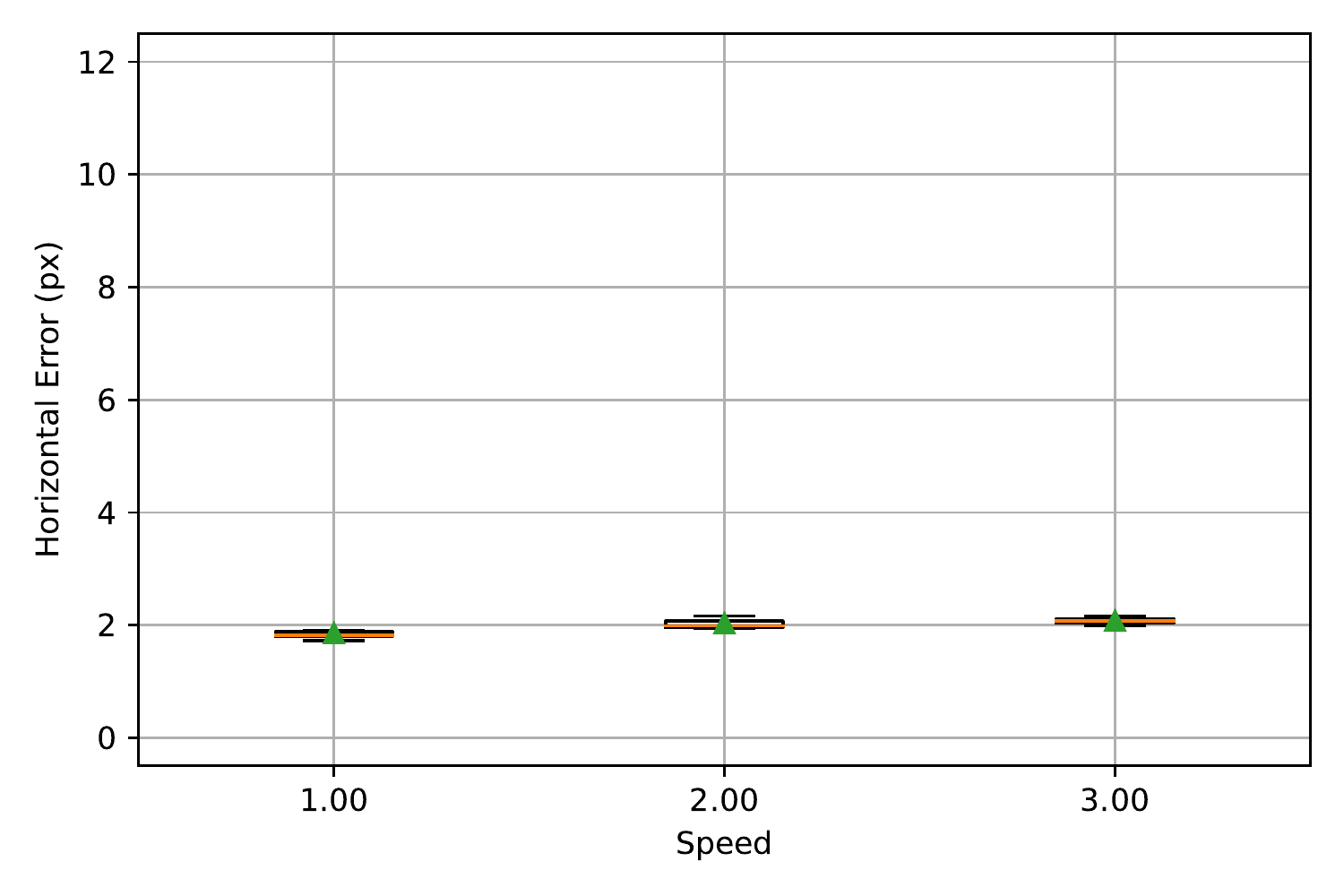}
    }
    \subfigure[$\eta(t)$, Vertical Coordinate]{
        \includegraphics[width=0.48\textwidth]{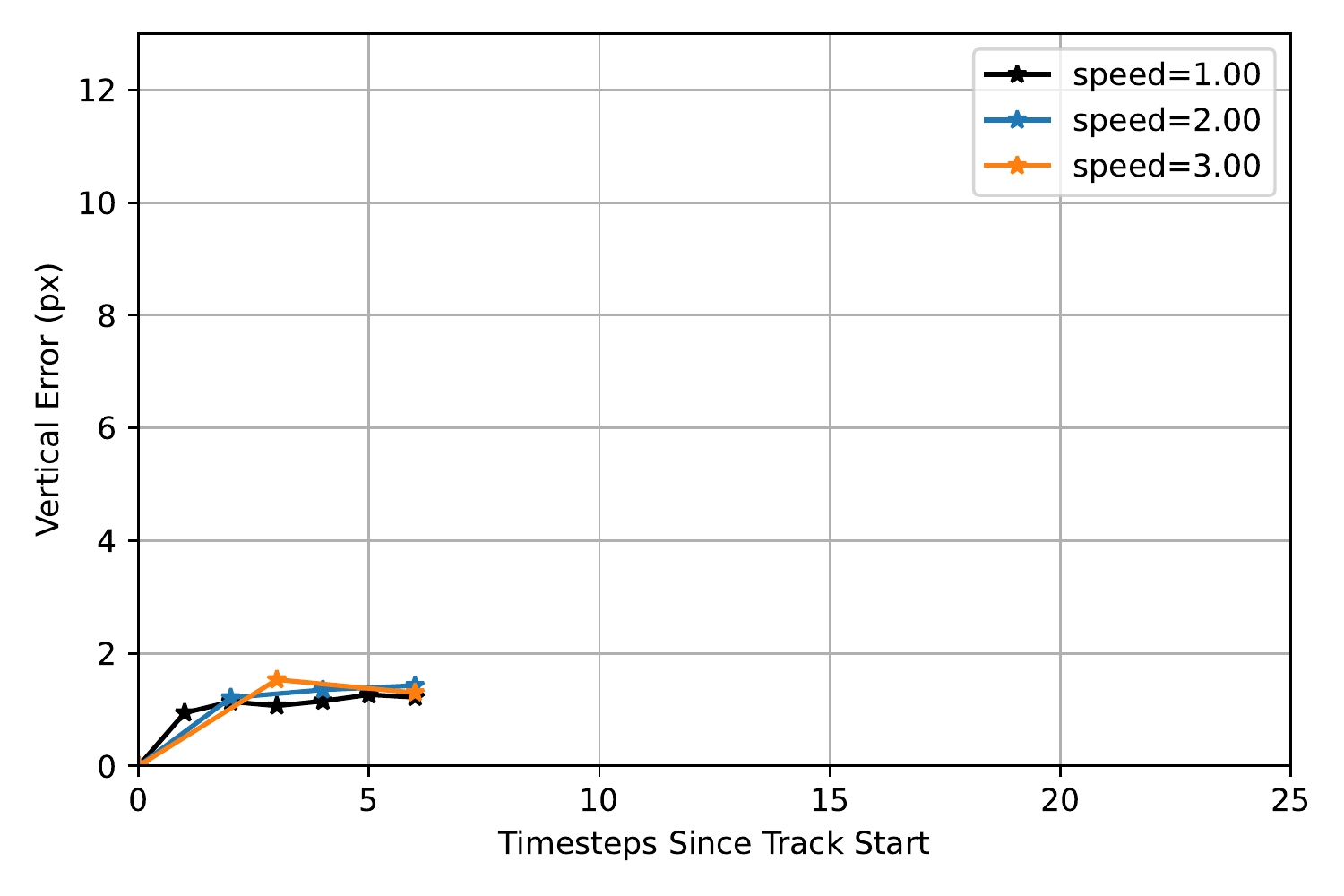}
        \includegraphics[width=0.48\textwidth]{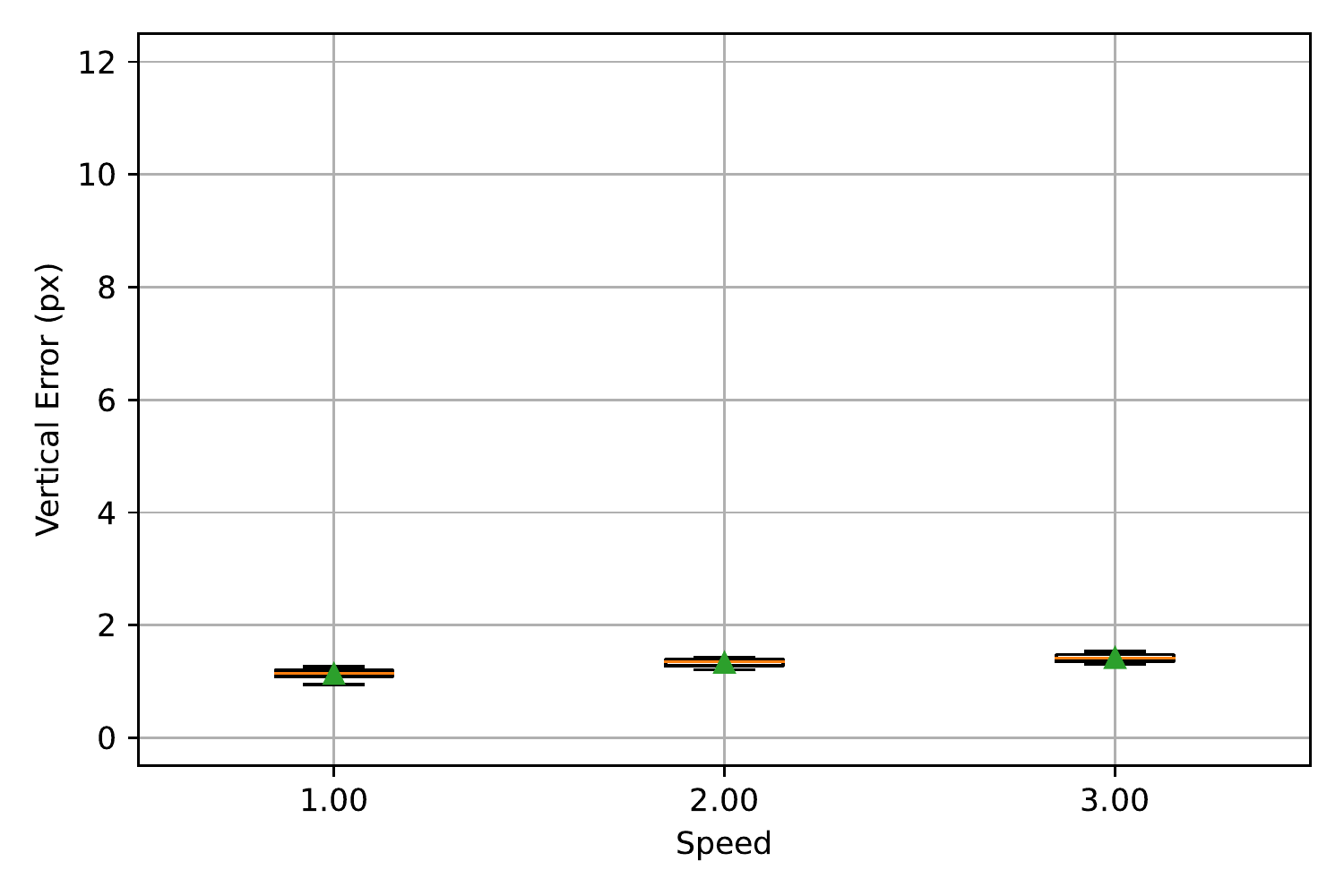}
    }
    \caption{\textbf{KITTI Dataset: Mean absolute errors are unaffected by speed when using the Correspondence Tracker.}
    The left column contains plots of the horizontal (top row) and vertical (bottom row) components of the mean absolute error $\eta(t)$ at each timestep $t$ after initial feature detection at multiple speeds. Each dot corresponds to a processed frame; lines for higher speeds contain data from fewer frames and therefore show fewer dots. The right column plots the ordinate values of each line for $t>0$ in the left figures as a box plot: means are shown as green triangles and medians are shown as orange lines. Compared to the results for the Lucas-Kanade Tracker in Figure \ref{fig:kitti_LK_MAE}, mean errors do not change when speed is increased from 1.00 to 3.00.}
    \label{fig:kitti_match_MAE}
\end{figure}

\begin{figure}[H]
    \centering
    \subfigure[$\Phi(t)$, Horizontal Coordinates]{
        \includegraphics[width=0.48\textwidth]{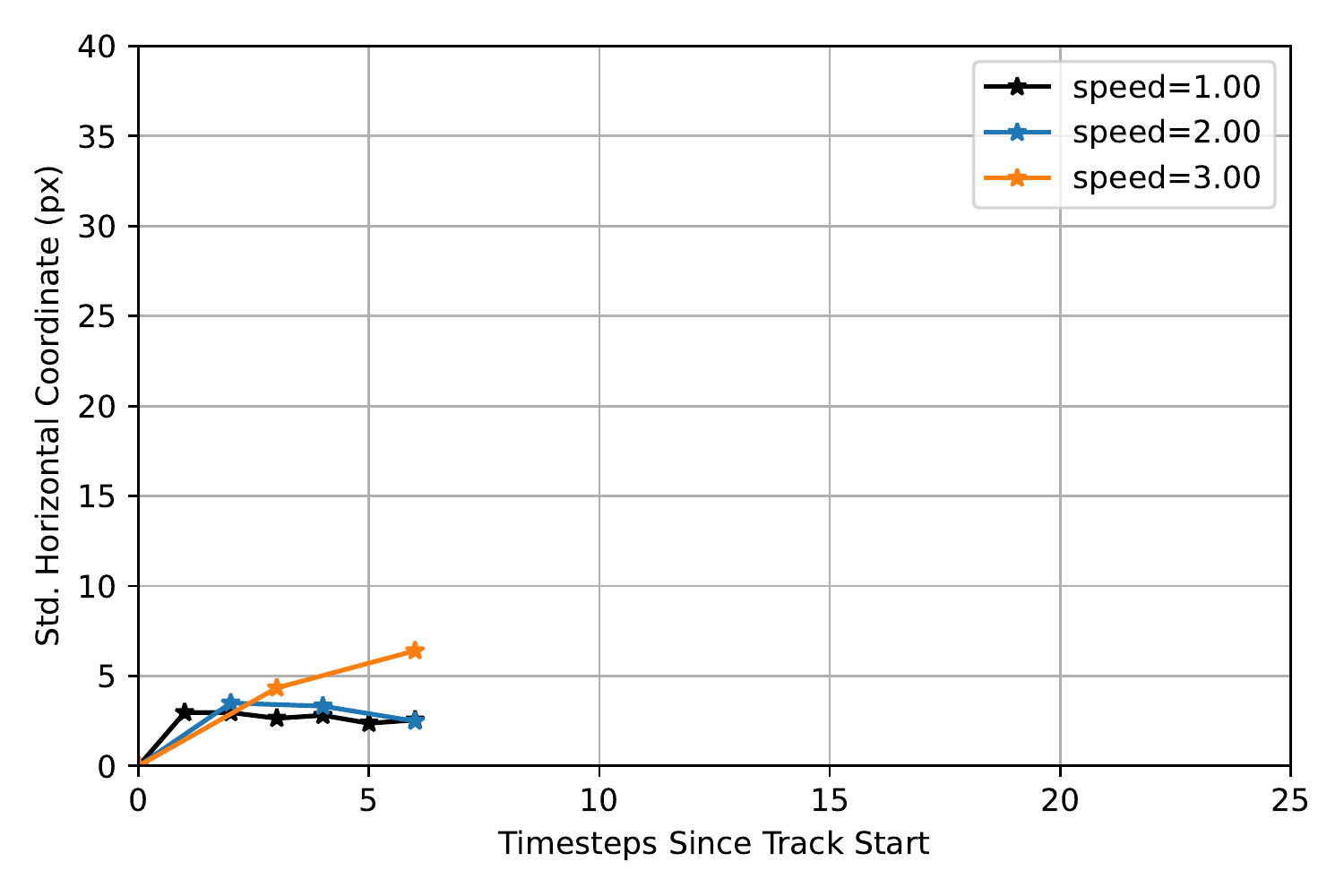}
        \includegraphics[width=0.48\textwidth]{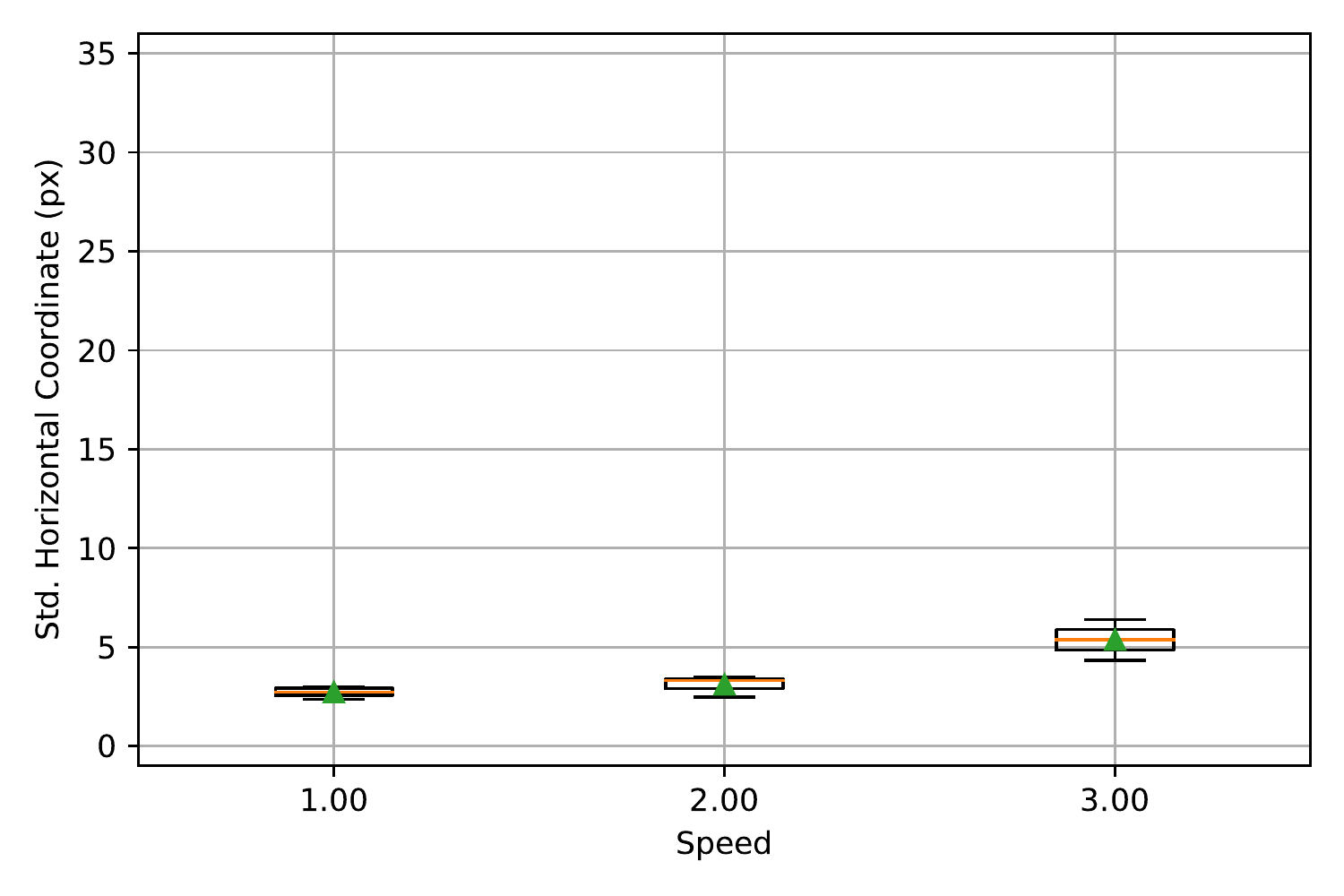}
    }
    \subfigure[$\Phi(t)$, Vertical Coordinates]{
        \includegraphics[width=0.48\textwidth]{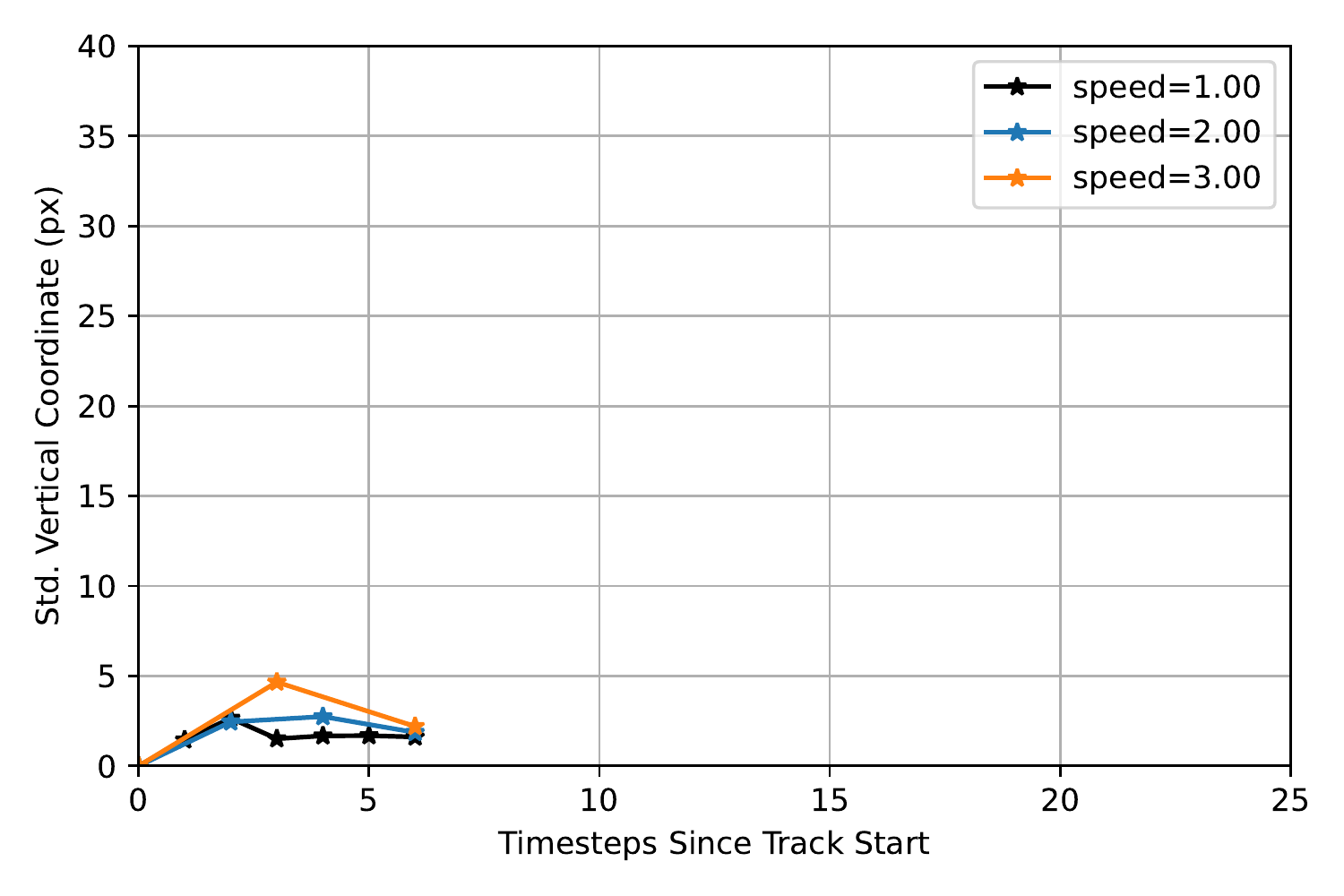}
        \includegraphics[width=0.48\textwidth]{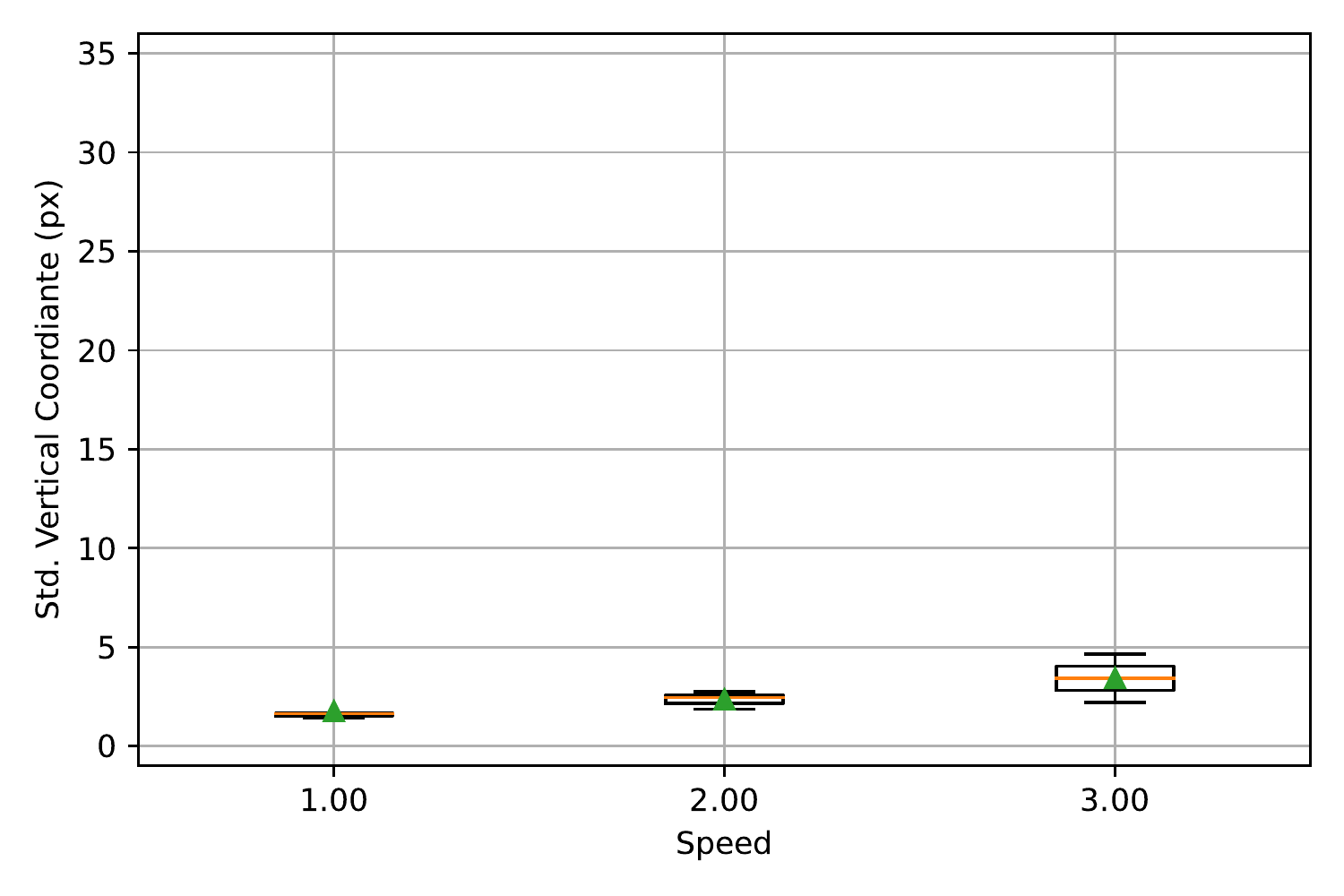}
    }
    \caption{\textbf{KITTI Dataset: Covariance is unaffected by speed when using the Correspondence Tracker.}  
    The left column contains plots of the horizontal (top row) and vertical (bottom row) components of the mean absolute error $\eta(t)$ at each timestep $t$ after initial feature detection at multiple speeds. Each dot corresponds to a processed frame; lines for higher speeds contain data from fewer frames and therefore show fewer dots. The right column plots the ordinate values of each line for $t>0$ in the left figures as a box plot: means are shown as green triangles and medians are shown as orange lines. 
    Compared to the results for the Lucas-Kanade Tracker in Figure \ref{fig:kitti_LK_cov}, covariances do not change when speed is increased from 1.00 to 2.00. Covariances show an increase of about 2 pixels when speed is increased from 2.00 to 3.00, however.
    }
    \label{fig:kitti_match_cov}
\end{figure}

\subsection{Supporting Figures for Gazebo Dataset with Linear Motion}
\label{sec:all_gazebo_linear_figs}

\begin{figure}[H]
    \subfigure[Lucas-Kanade]{\includegraphics[width=0.48\textwidth]{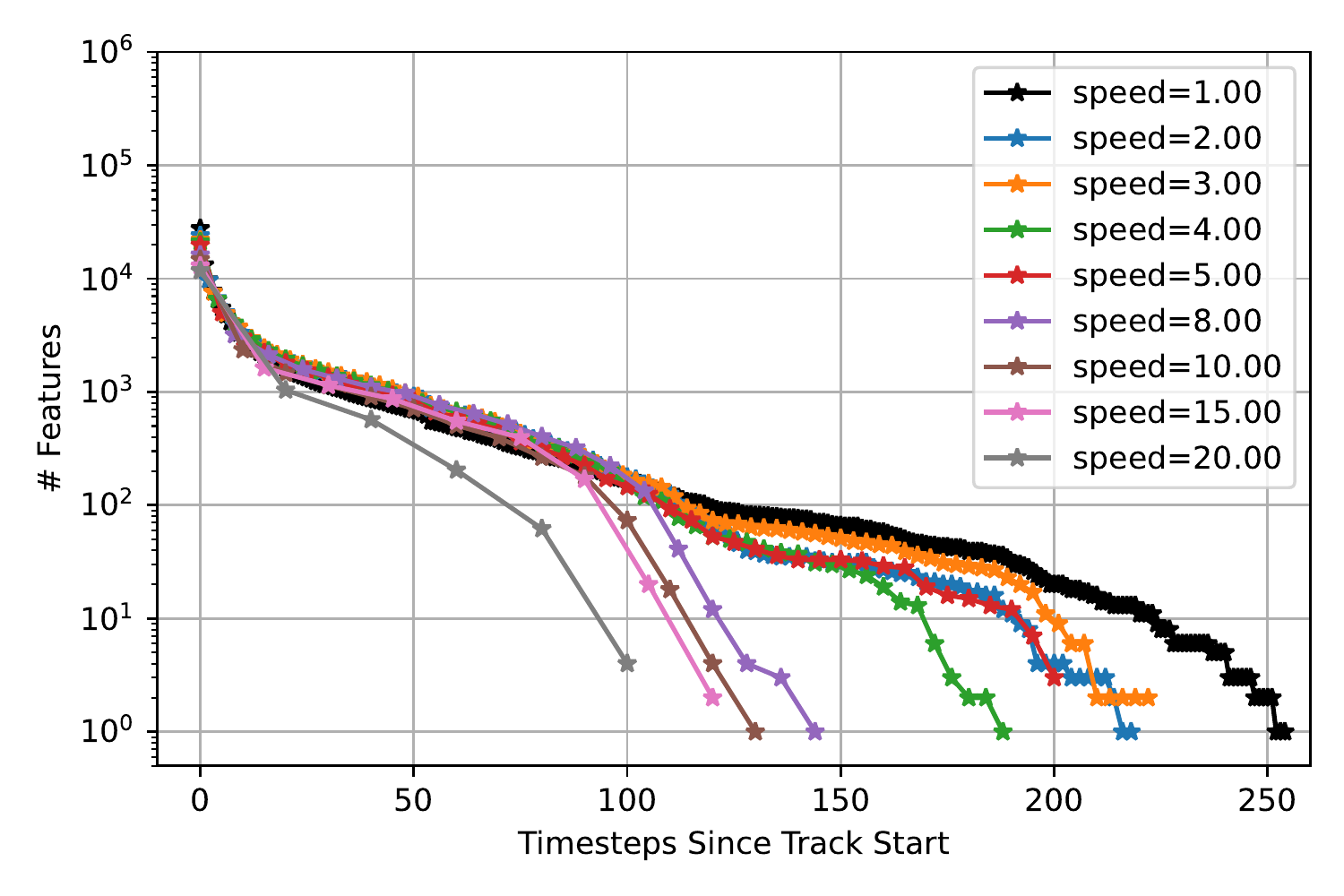}}
    \subfigure[Correspondence]{\includegraphics[width=0.48\textwidth]{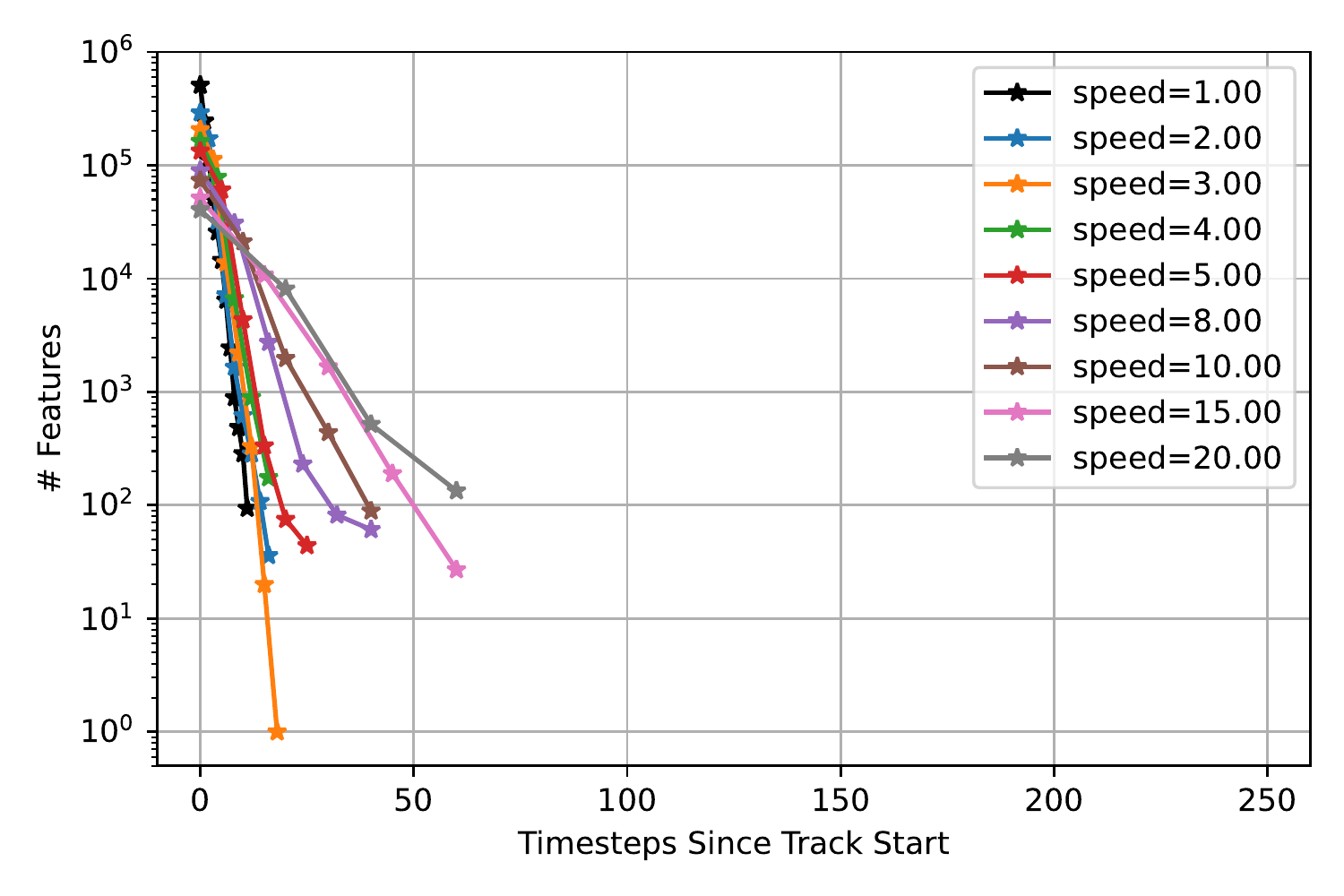}}
    \caption{Feature Lifetime is plotted on the horizontal axis. The vertical axis, in log scale, shows the number of features in all 11 scenes that lived at least that long for every tested speed. The number of features drops very fast, especially when the Correspondence Tracker is used. \textbf{In subsequent analyses, we only compute means errors and covariances at timesteps with at least 500 features on the Gazebo Linear Dataset.}}
    \label{fig:gazebo_linear_avg_feats}
\end{figure}

\begin{figure}[H]
    \centering
    \subfigure[Lucas-Kanade]{\includegraphics[width=0.48\textwidth]{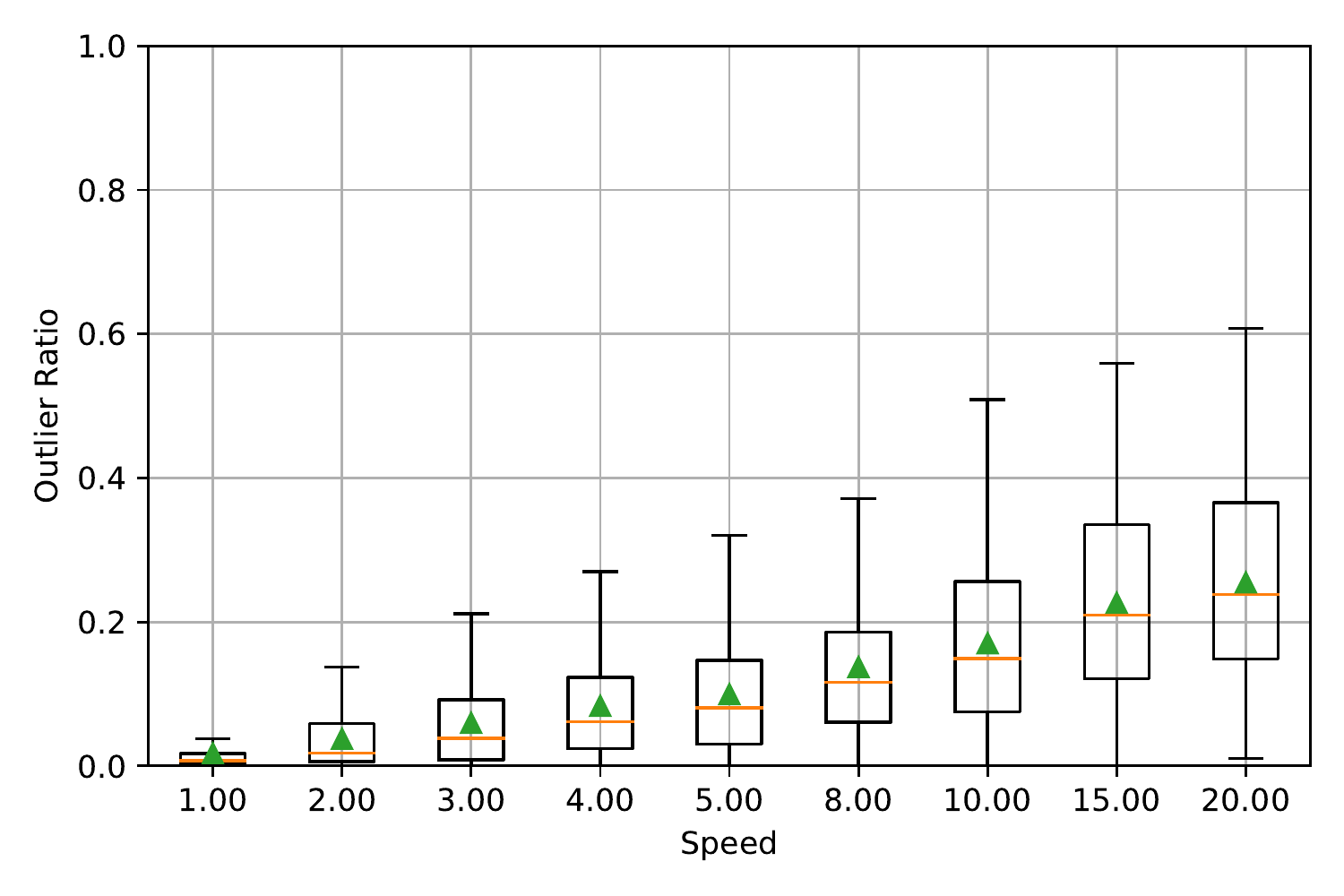}}
    \subfigure[Correspondence]{\includegraphics[width=0.48\textwidth]{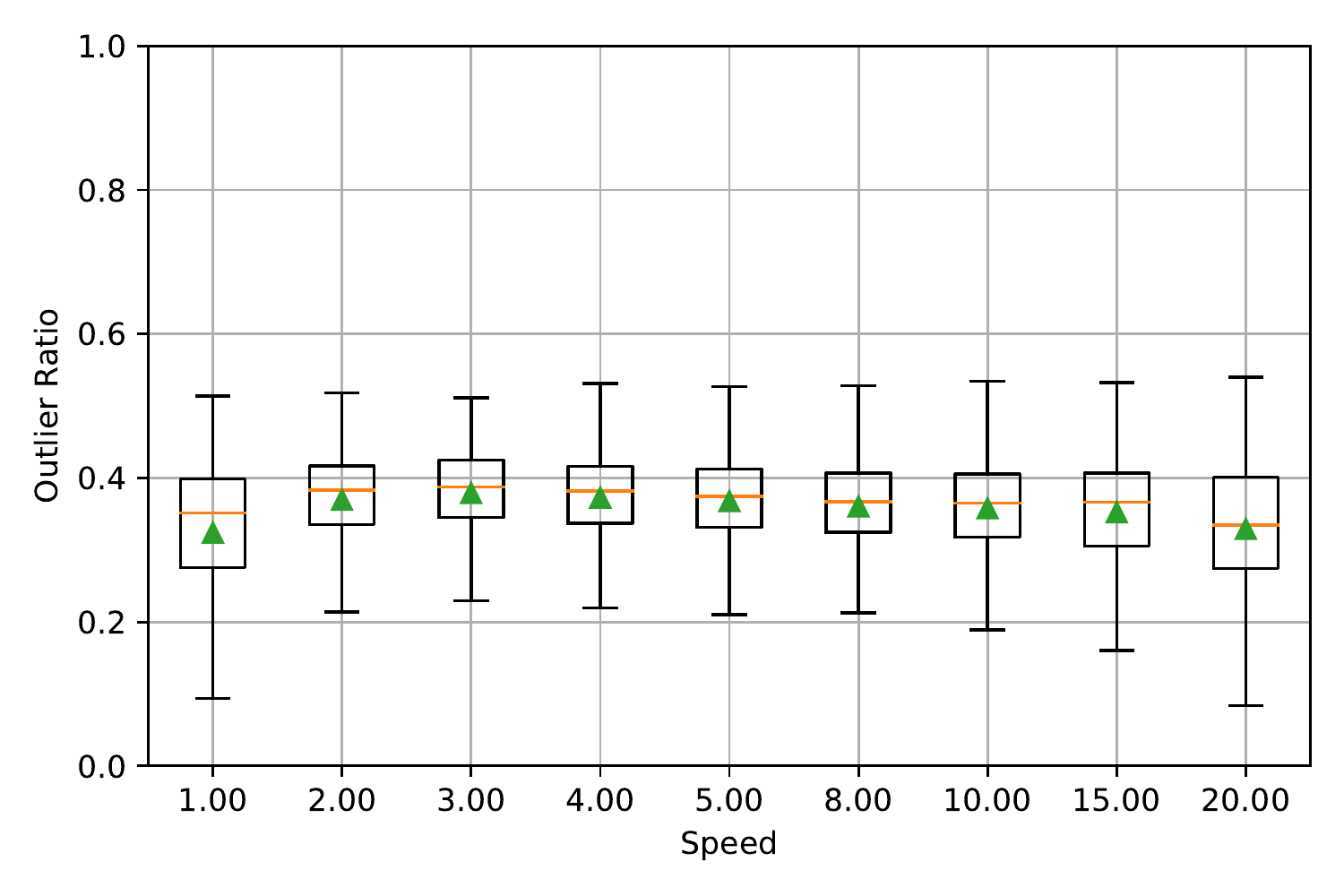}}
    \caption{\textbf{Gazebo Linear Dataset: Outlier Ratios are a function of speed when using the Lucas-Kanade Tracker and constant for the Correspondence Tracker.} Outlier ratios per frame are shown as box-and-whisker plots for tested speeds for the Lucas-Kanade tracker on the left and the Correspondence Tracker on the right. Mean values are shown as green triangles and median values are shown as orange lines. For lower speeds, the Lucas-Kanade tracker produces fewer outliers. Outlier ratios then increase with speed. On the other hand, the outlier ratio for the Correspondence Tracker remains constant, at around 40 percent.}
    \label{fig:gazebo_linear_outlier_ratio}
\end{figure}

\begin{figure}[H]
    \centering
    \subfigure[$\nu(t)$, Horizontal Coordinate]{\includegraphics[width=0.48\textwidth]{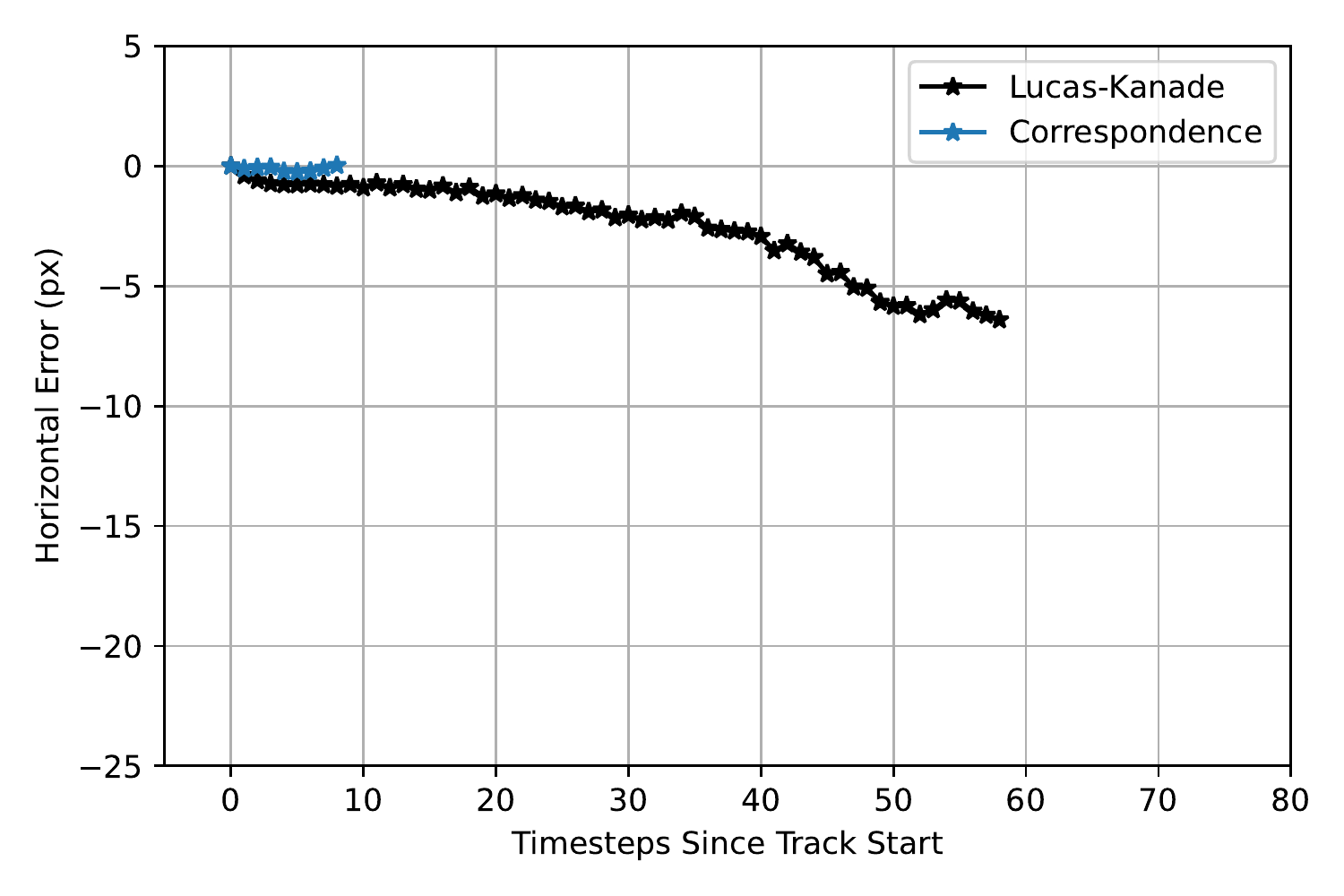}}
    \subfigure[$\nu(t)$, Vertical Coordinate]{\includegraphics[width=0.48\textwidth]{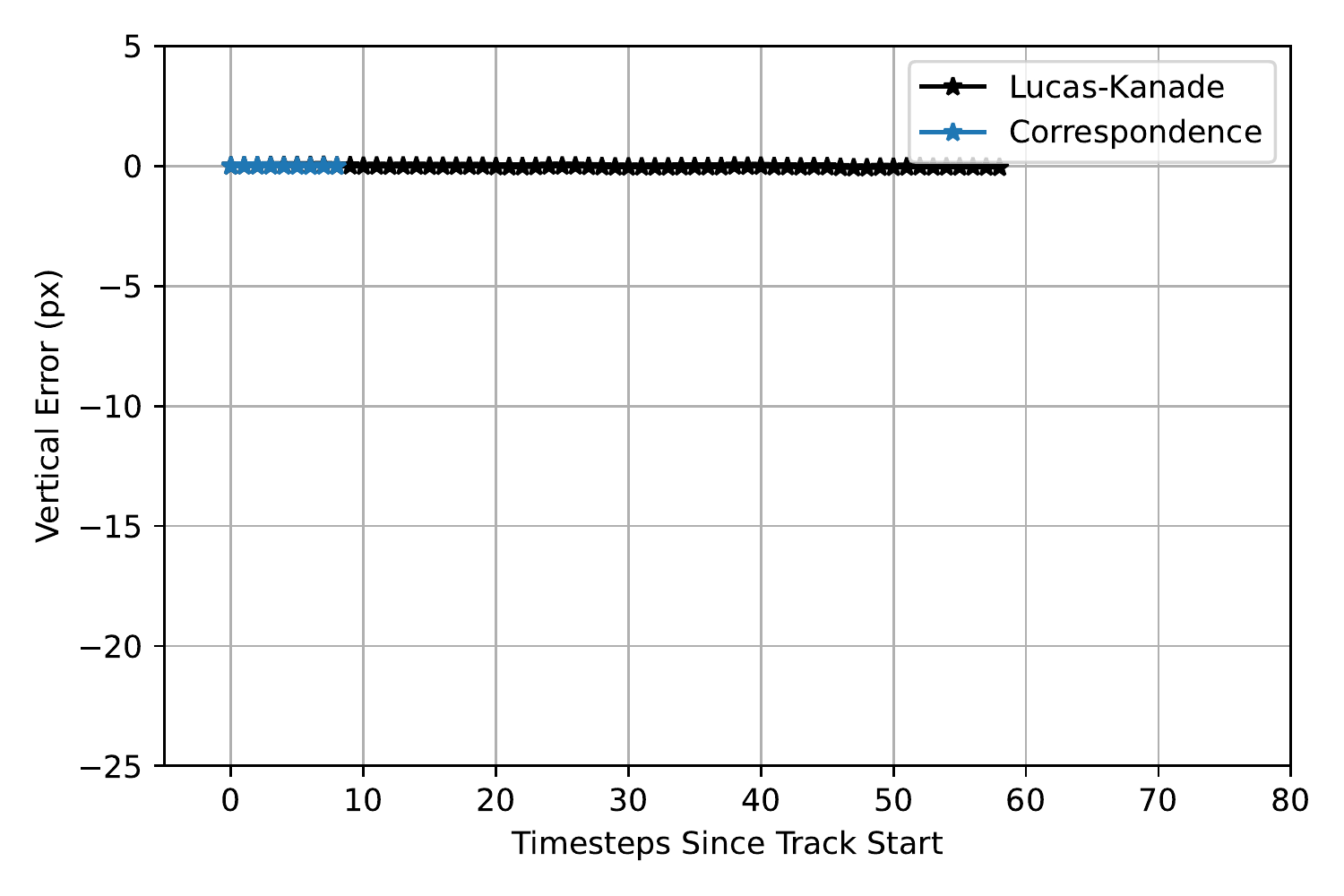}}
    \caption{\textbf{Gazebo Linear Dataset: At nominal speed, the Lucas-Kanade Tracker slowly accumulates negative error in the horizontal direction. The Correspondence Tracker has zero mean error.} Lines shown are horizontal (left) and vertical (right) coordinates of mean error $\nu(t)$ calculated using tracks averaged over all scenes; calculation is cut off at 58 frames for the Lucas-Kanade Tracker and 9 frames for the Correspondence Tracker so that averages can be computed with at least 500 features.}
    \label{fig:gazebo_linear_1.00_meanerror}
\end{figure}

\begin{figure}[H]
    \centering
    \subfigure[$\eta(t)$, Horizontal Coordinate]{\includegraphics[width=0.48\textwidth]{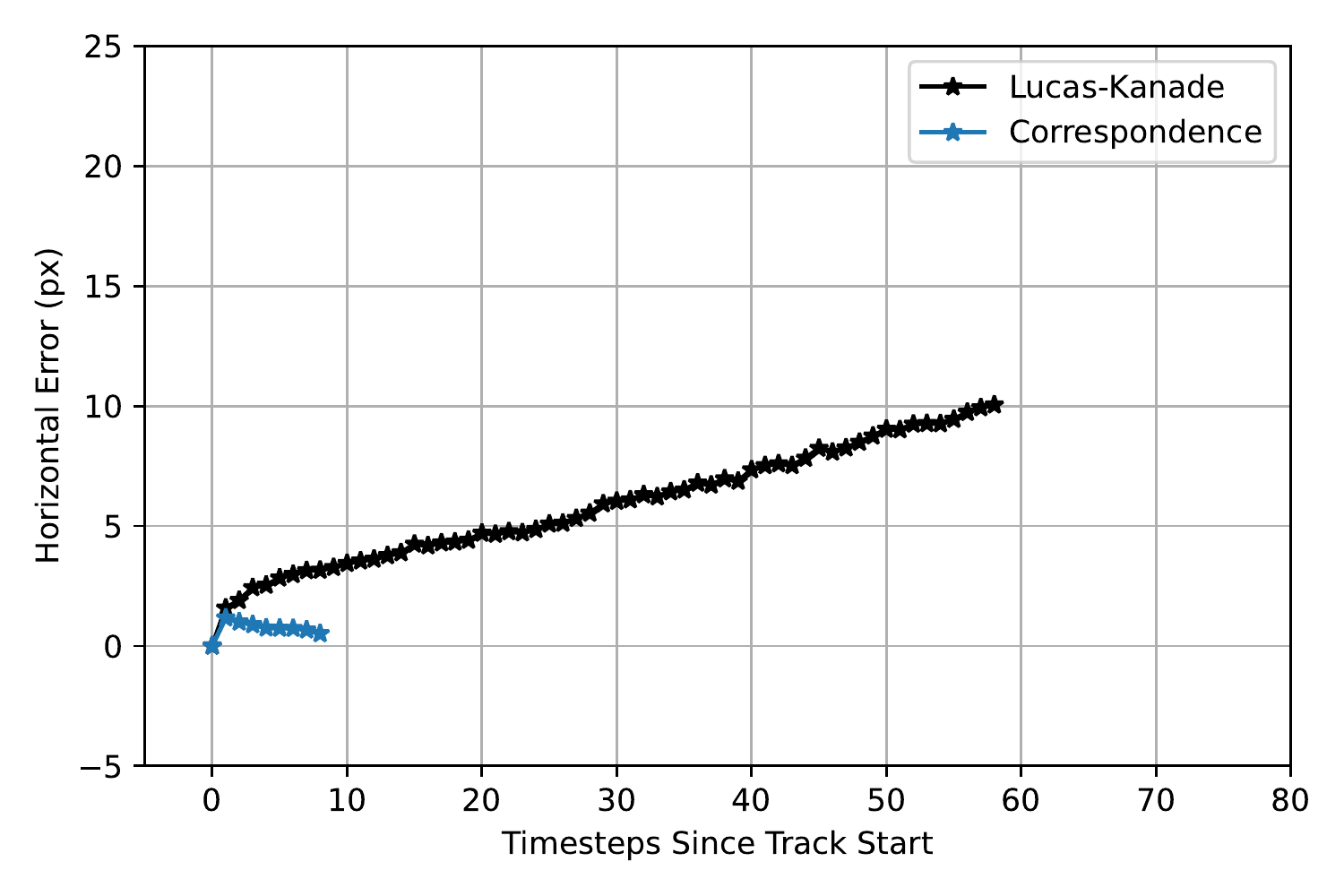}}
    \subfigure[$\eta(t)$, Vertical Coordinate]{\includegraphics[width=0.48\textwidth]{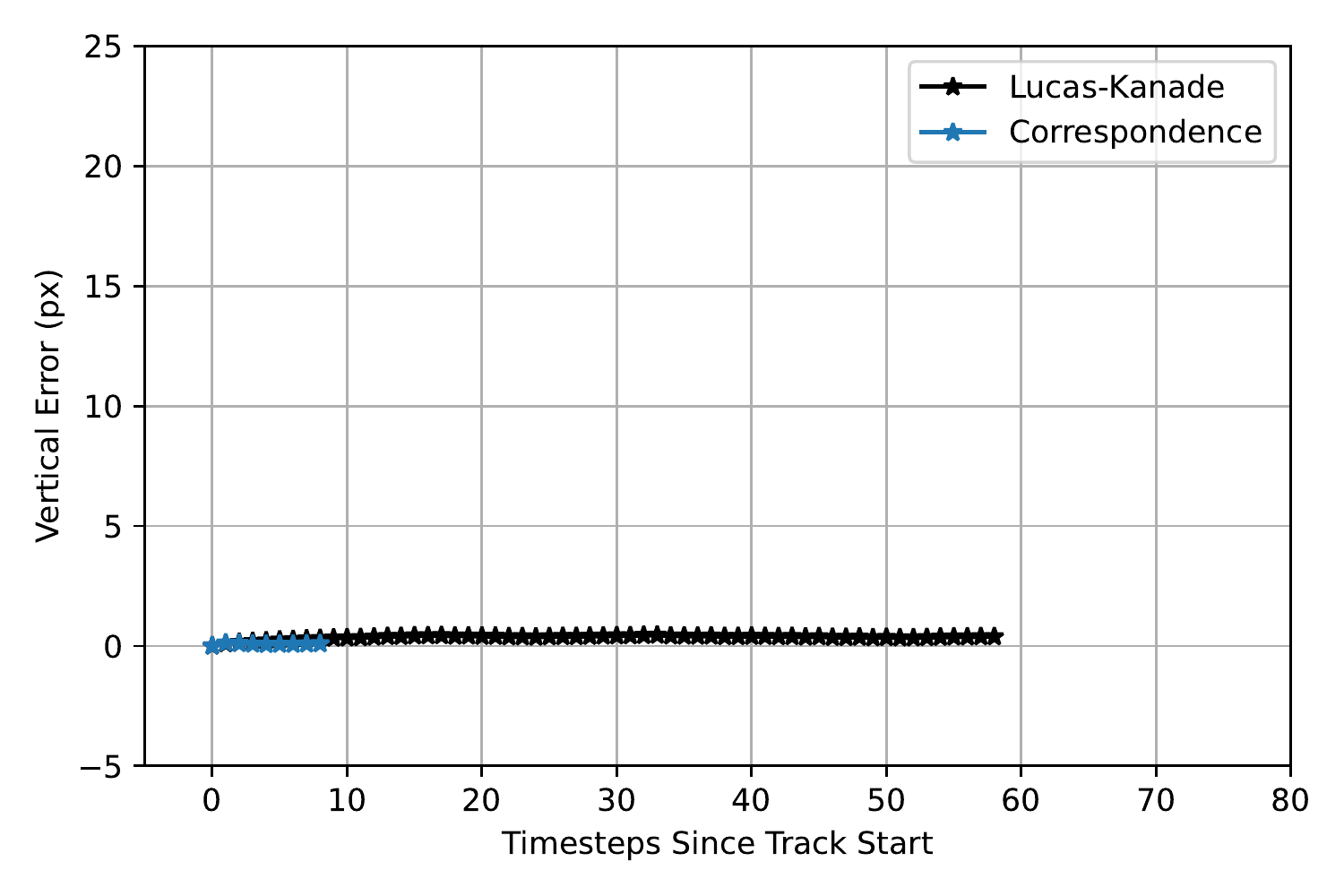}}   
    \subfigure[$\Phi(t)$, Horizontal Coordinate]{\includegraphics[width=0.48\textwidth]{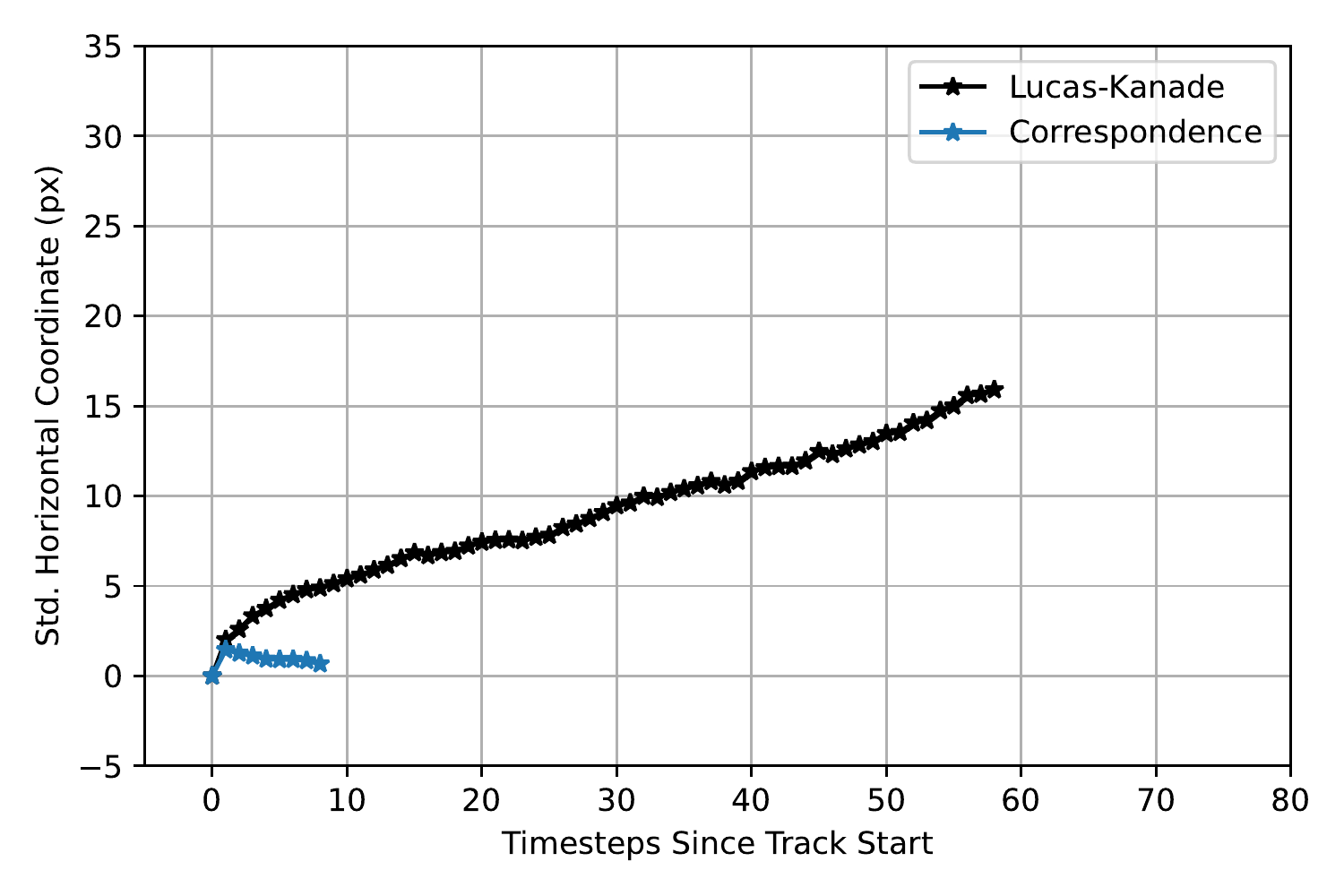}}
    \subfigure[$\Phi(t)$, Vertical Coordinate]{\includegraphics[width=0.48\textwidth]{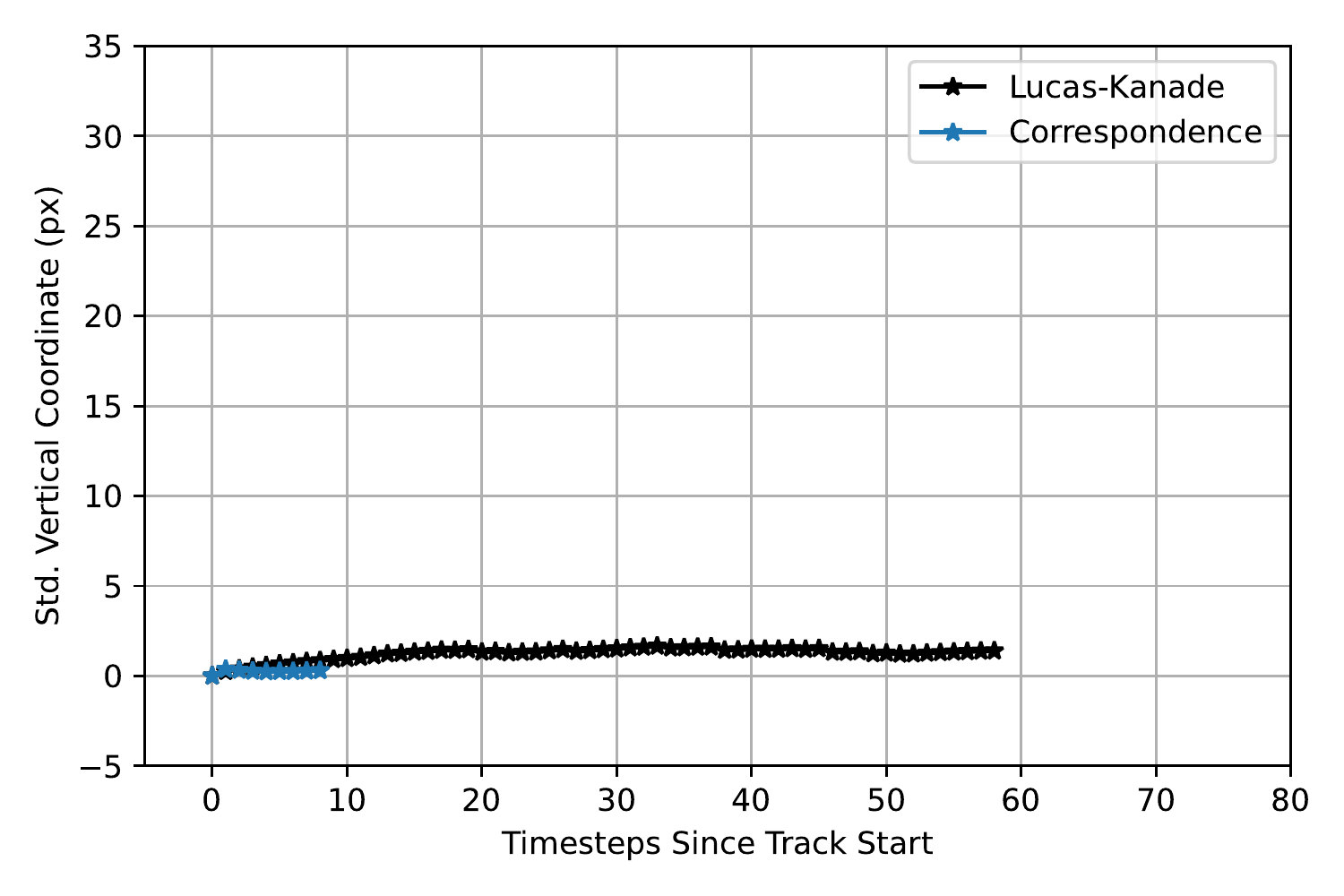}}      
    \caption{\textbf{Gazebo Linear Dataset: The Lucas-Kanade tracker drifts considerably more than the Correspondence Tracker, but only in the horizontal direction.} Lines shown are horizontal (left column) and vertical (right column) coordinates of mean absolute error $\eta(t)$ (top row) and covariance $\Phi(t)$ (bottom row) calculated using tracks averaged over all scenes; calculation is cut off at 58 frames for Lucas-Kanade Tracker and 9 frames for the Correspondence Tracker so that averages can be computed with at least 500 features. Both mean absolute error and covariance are constant when using the Correspondence Tracker. On the other hand, the horizontal coordinate of $\eta(t)$ and $\Phi(t)$ drifts upwards when using the Lucas-Kanade Tracker.}
    \label{fig:gazebo_linear_1.00_error_cov}
\end{figure}

\begin{figure}[H]
    \centering
    \subfigure[$\eta(t)$, Horizontal Coordinate]{
        \includegraphics[width=0.48\textwidth]{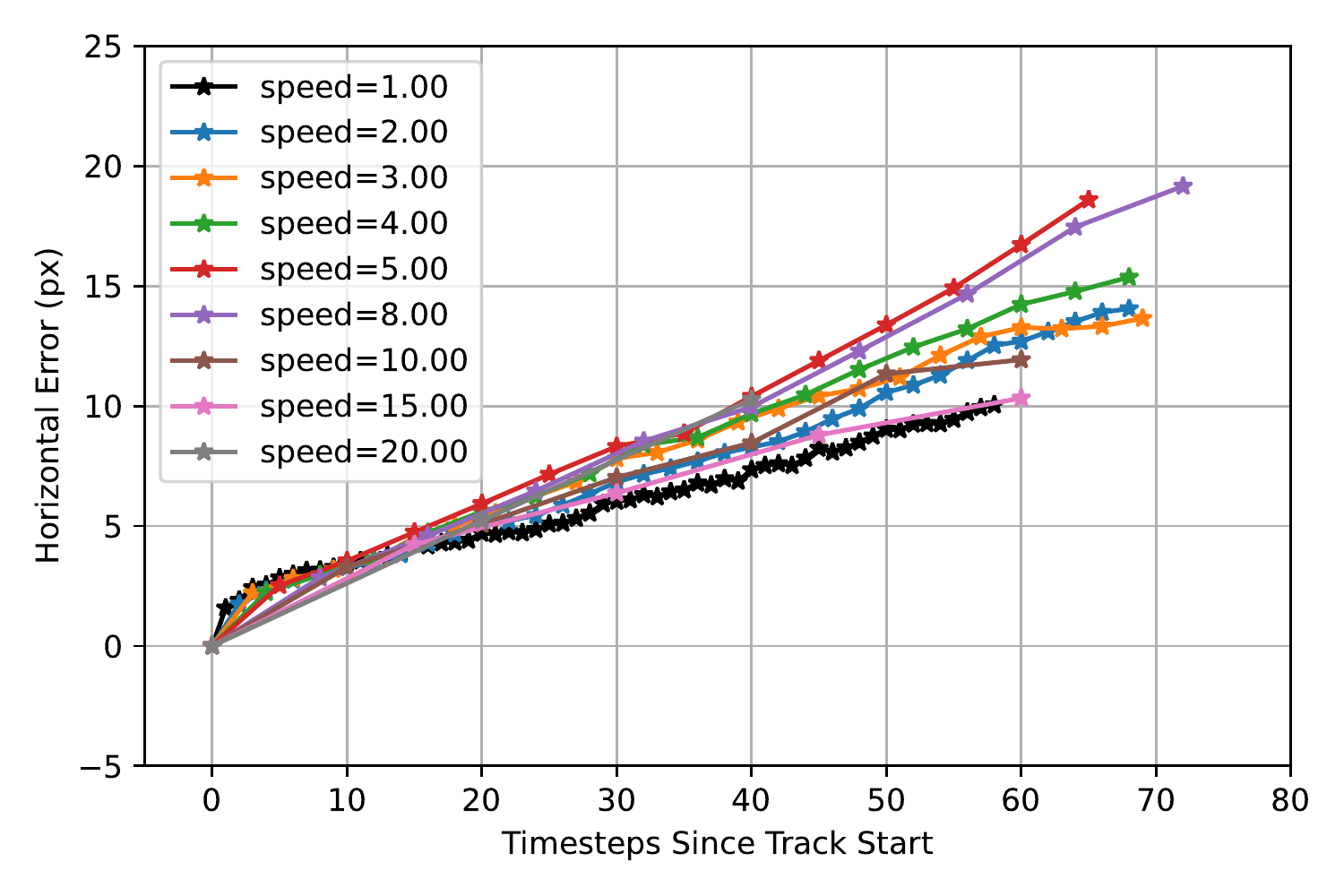}
        \includegraphics[width=0.48\textwidth]{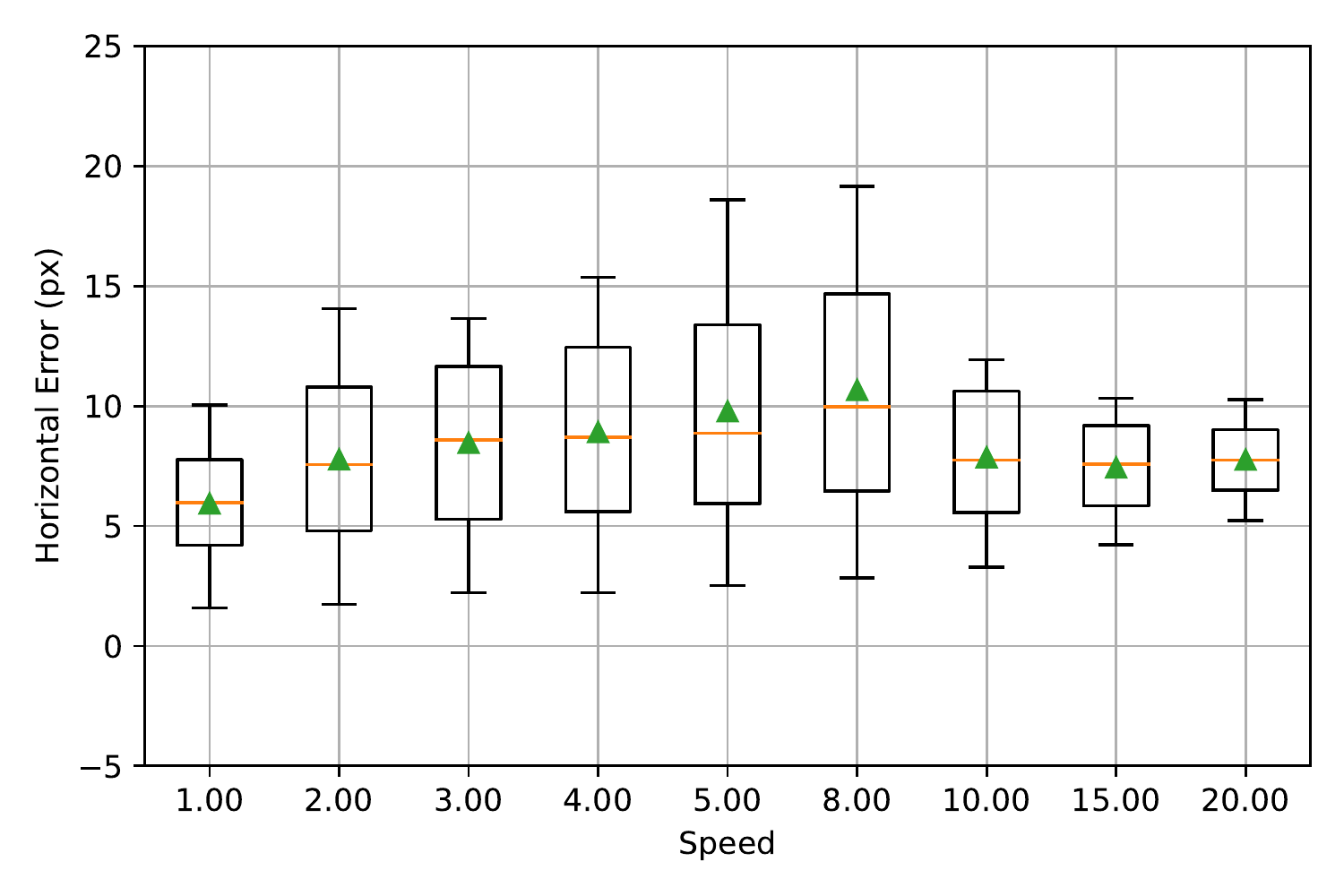}
    }
    \subfigure[$\eta(t)$, Vertical Coordinate]{
        \includegraphics[width=0.48\textwidth]{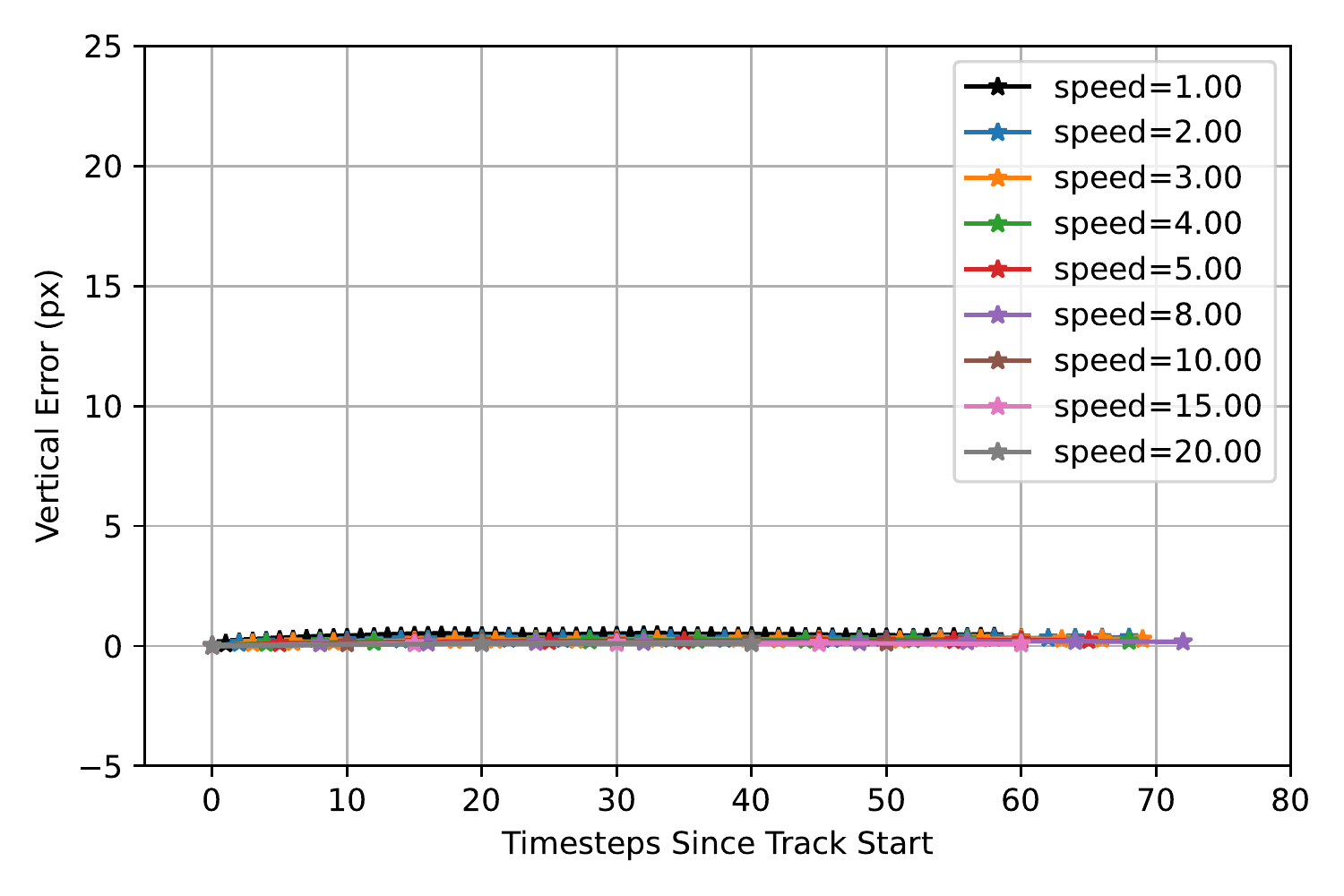}
        \includegraphics[width=0.48\textwidth]{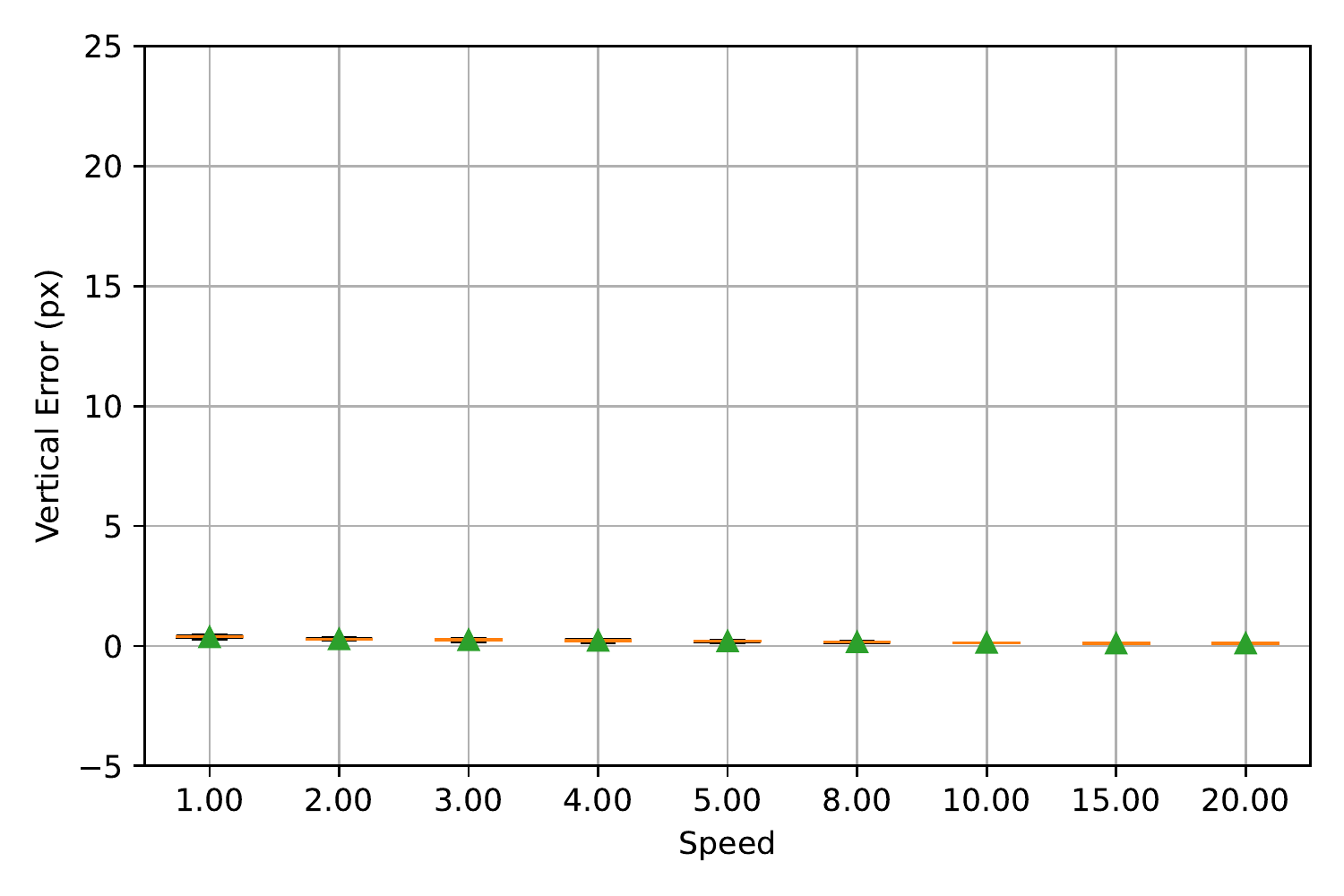}
    }
    \caption{\textbf{Gazebo Linear Dataset: Mean absolute errors increase with speed when using the Lucas-Kanade Tracker.}
    The left column contains plots of the horizontal (top row) and vertical (bottom row) components of the mean absolute error $\eta(t)$ at each timestep $t$ after initial feature detection at multiple speeds. Each dot corresponds to a processed frame; lines for higher speeds contain data from fewer frames and therefore show fewer dots. The right column plots the ordinate values of each line for $t>0$ in the left figures as a box plot: means are shown as green triangles and medians are shown as orange lines.
    Mean absolute errors in the horizontal coordinate increase as speed is increased from 1.00 to 8.00. $\eta(t)$ then decreases for speeds=10.00 (brown line), 15.00 (pink line), and 20.00 (gray line), showing that both the number of elapsed frames, and the speed are both factors that affect $\nu(t)$. For all speeds, mean absolute error is close to zero in the vertical coordinate.    
    }
    \label{fig:gazebo_linear_LK_MAE}
\end{figure}

\begin{figure}[H]
    \centering
    \subfigure[$\Phi(t)$, Horizontal Coordinate]{
        \includegraphics[width=0.48\textwidth]{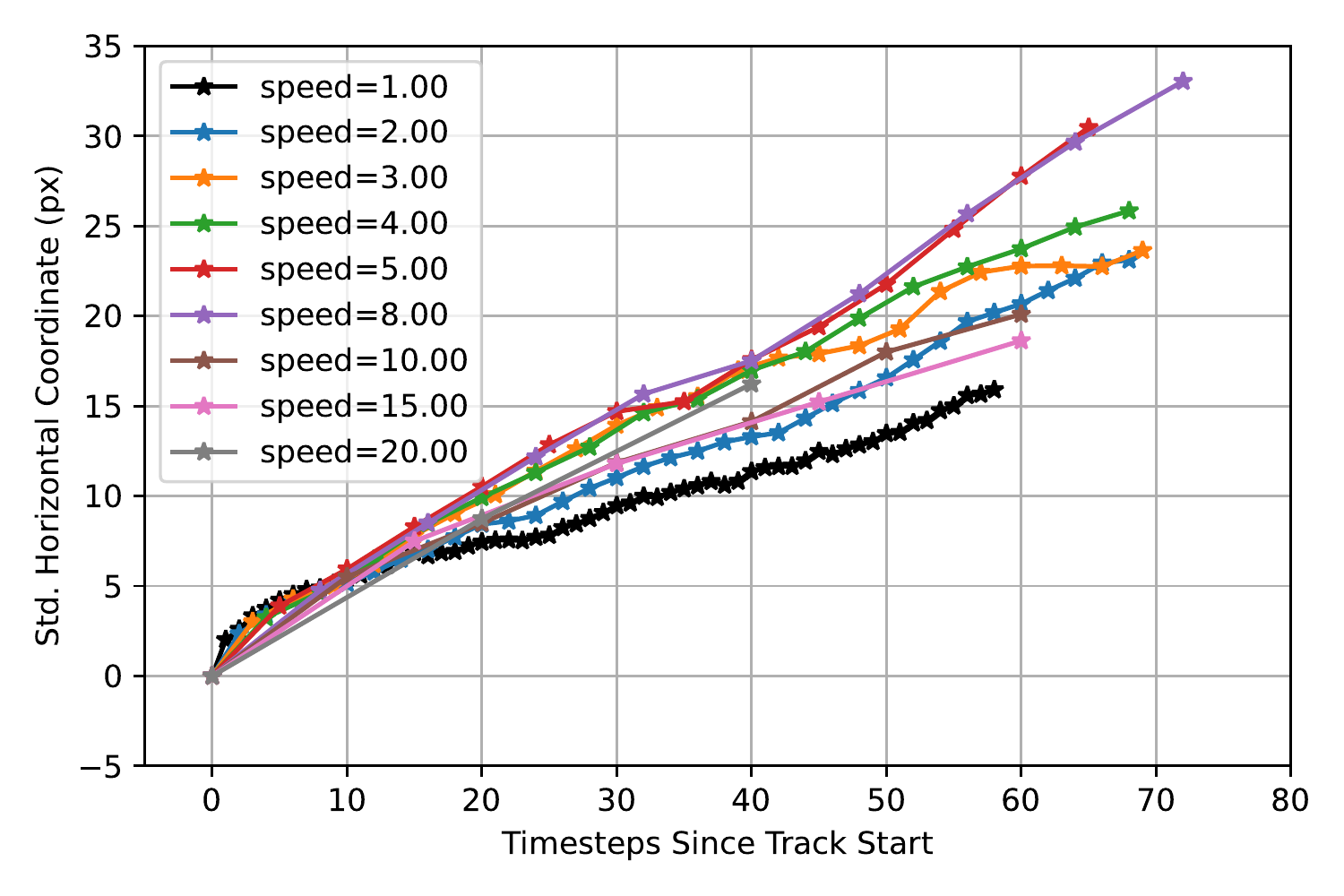}
        \includegraphics[width=0.48\textwidth]{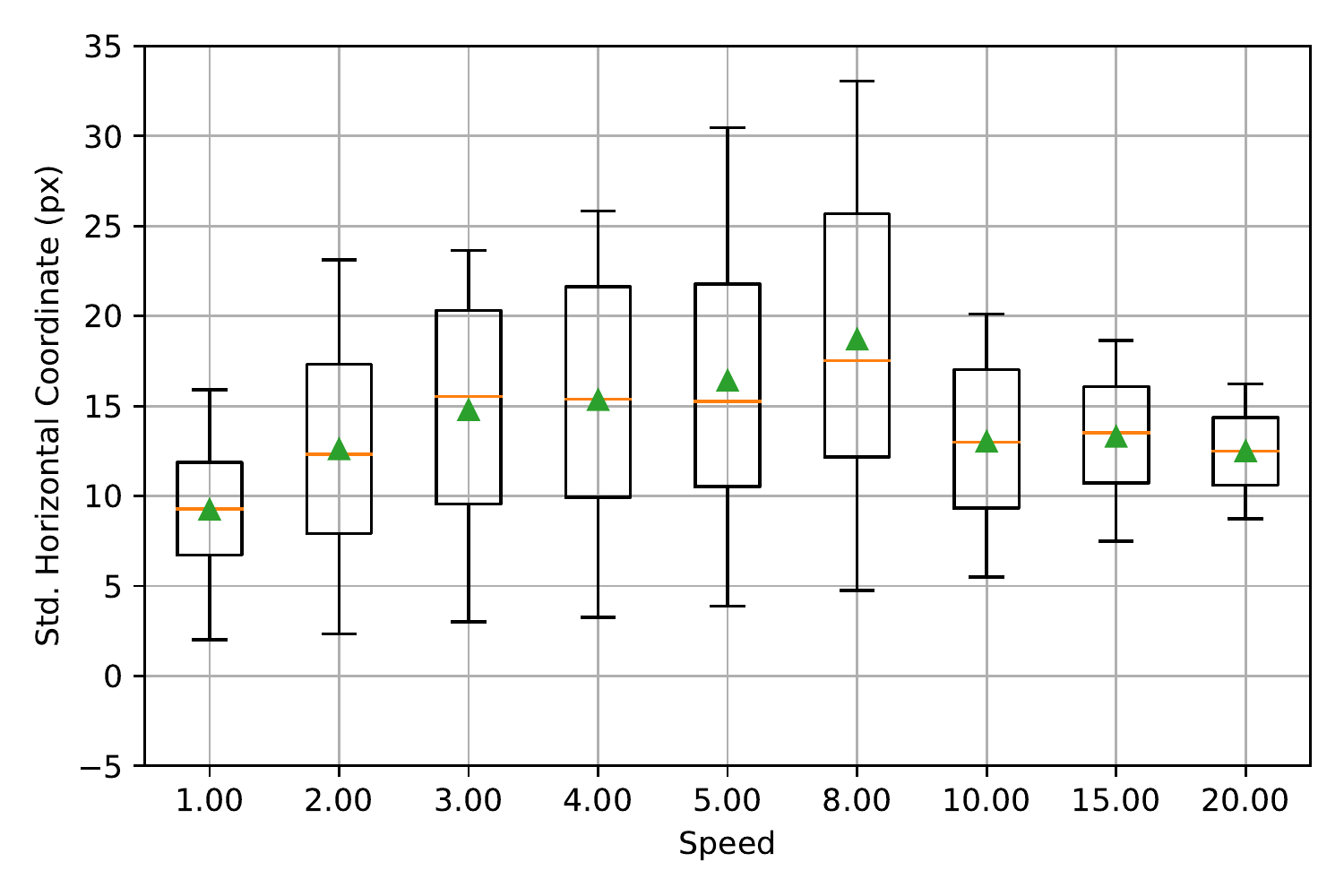}
    }
    \subfigure[$\Phi(t)$, Vertical Coordinate]{
        \includegraphics[width=0.48\textwidth]{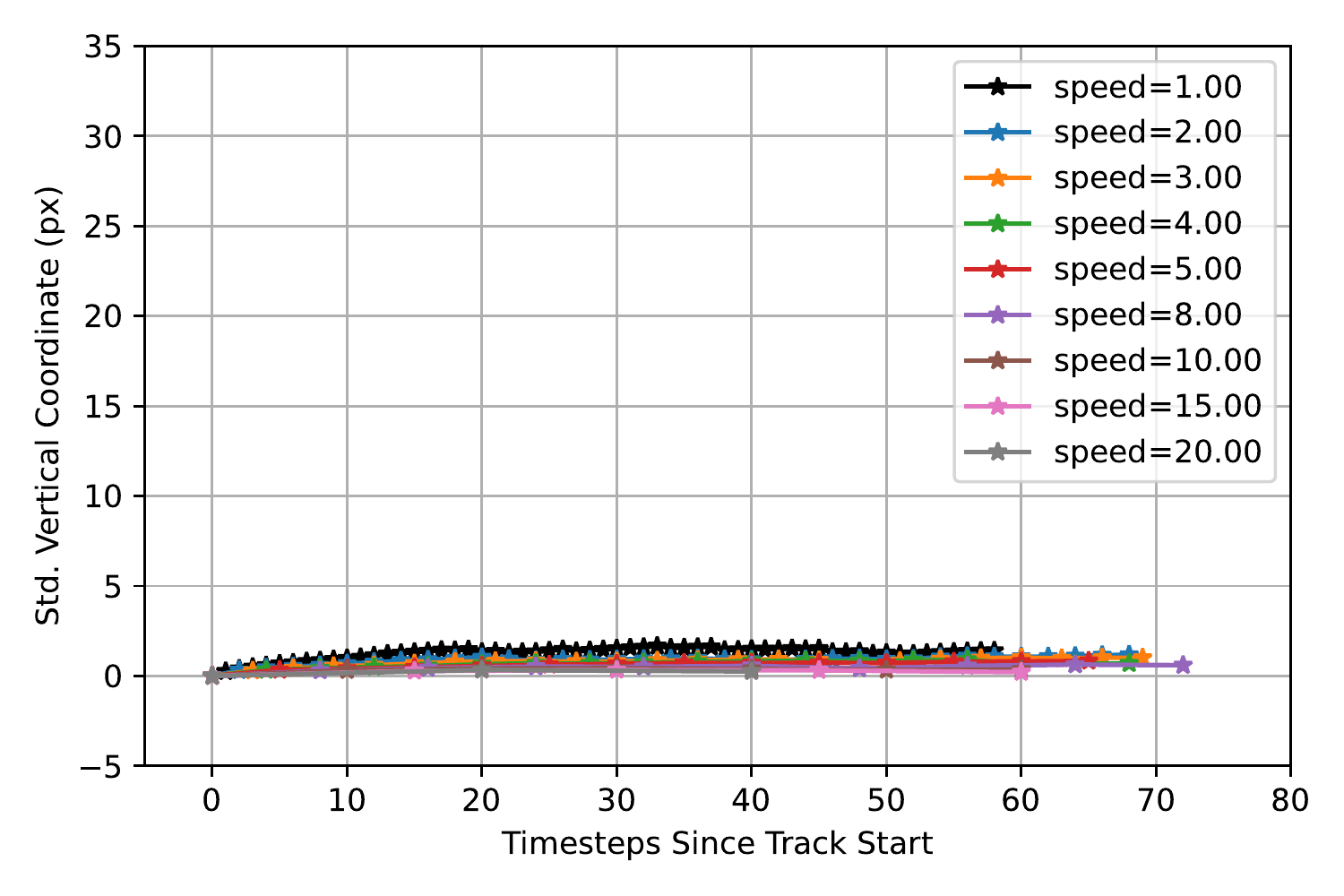}
        \includegraphics[width=0.48\textwidth]{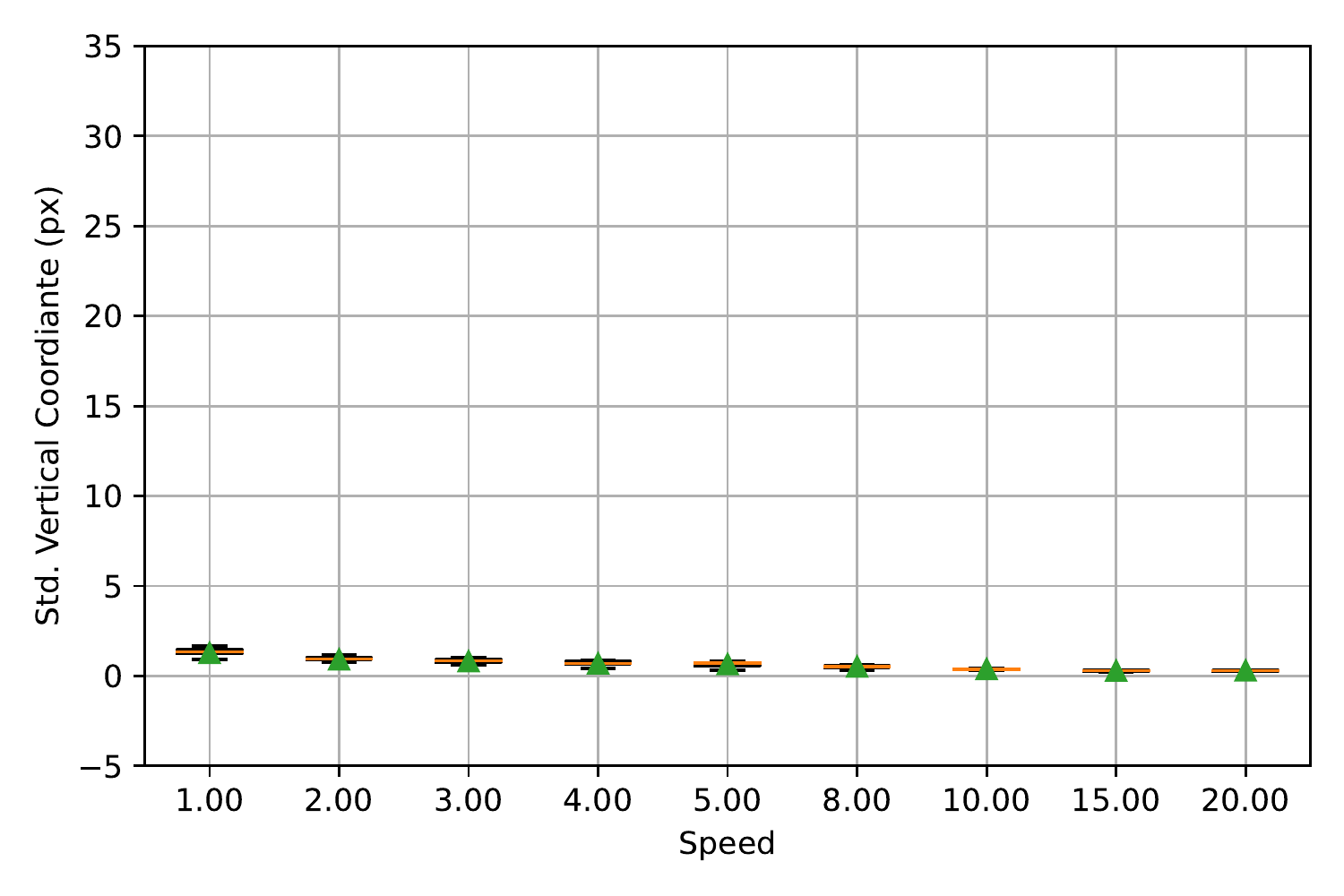}
    }
    \caption{\textbf{Gazebo Linear Dataset: Covariance increases with speed when using the Lucas-Kanade Tracker.}
    The left column contains plots of the horizontal (top row) and vertical (bottom row) components of the covariance $\Phi(t)$ at each timestep $t$ after initial feature detection at multiple speeds. Each dot corresponds to a processed frame; lines for higher speeds contain data from fewer frames and therefore show fewer dots. The right column plots the ordinate values of each line for $t>0$ in the left figures as a box plot: means are shown as green triangles and medians are shown as orange lines.
    Covariance increases in the horizontal coordinate increase as speed is increased from 1.00 to 8.00. The covariance then decreases for speeds=10.00 (brown line), 15.00 (pink line), and 20.00 (gray line), showing that both the number of elapsed frames, and the speed are both factors that affect $\Phi(t)$. For all speeds, covariance is close to zero in the vertical coordinate.
    }
    \label{fig:gazebo_linear_LK_cov}
\end{figure}

\begin{figure}[H]
    \centering
    \subfigure[$\nu(t)$, Horizontal Coordinate]{
        \includegraphics[width=0.48\textwidth]{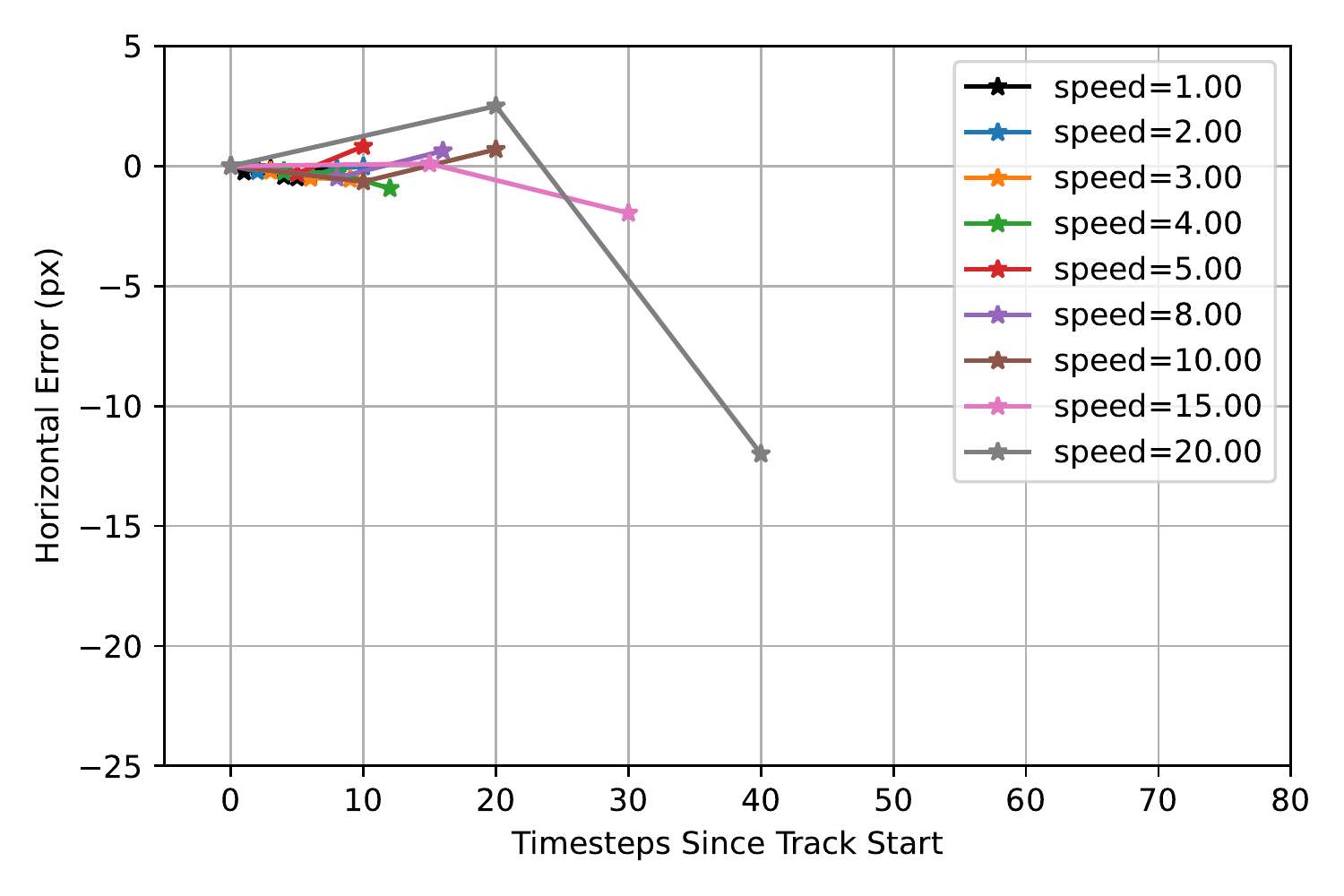}
        \includegraphics[width=0.48\textwidth]{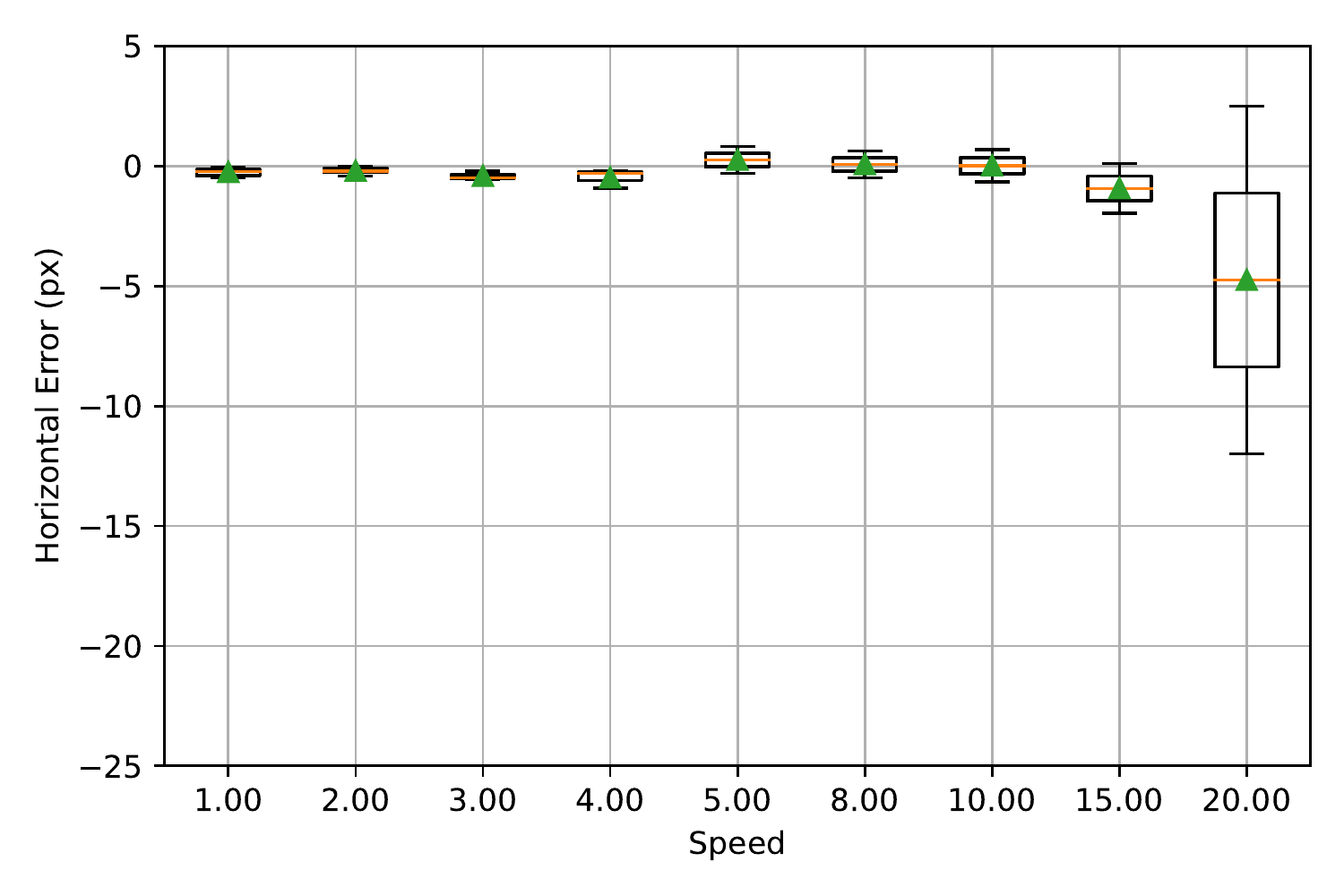}
    }
    \subfigure[$\nu(t)$, Vertical Coordinate]{
        \includegraphics[width=0.48\textwidth]{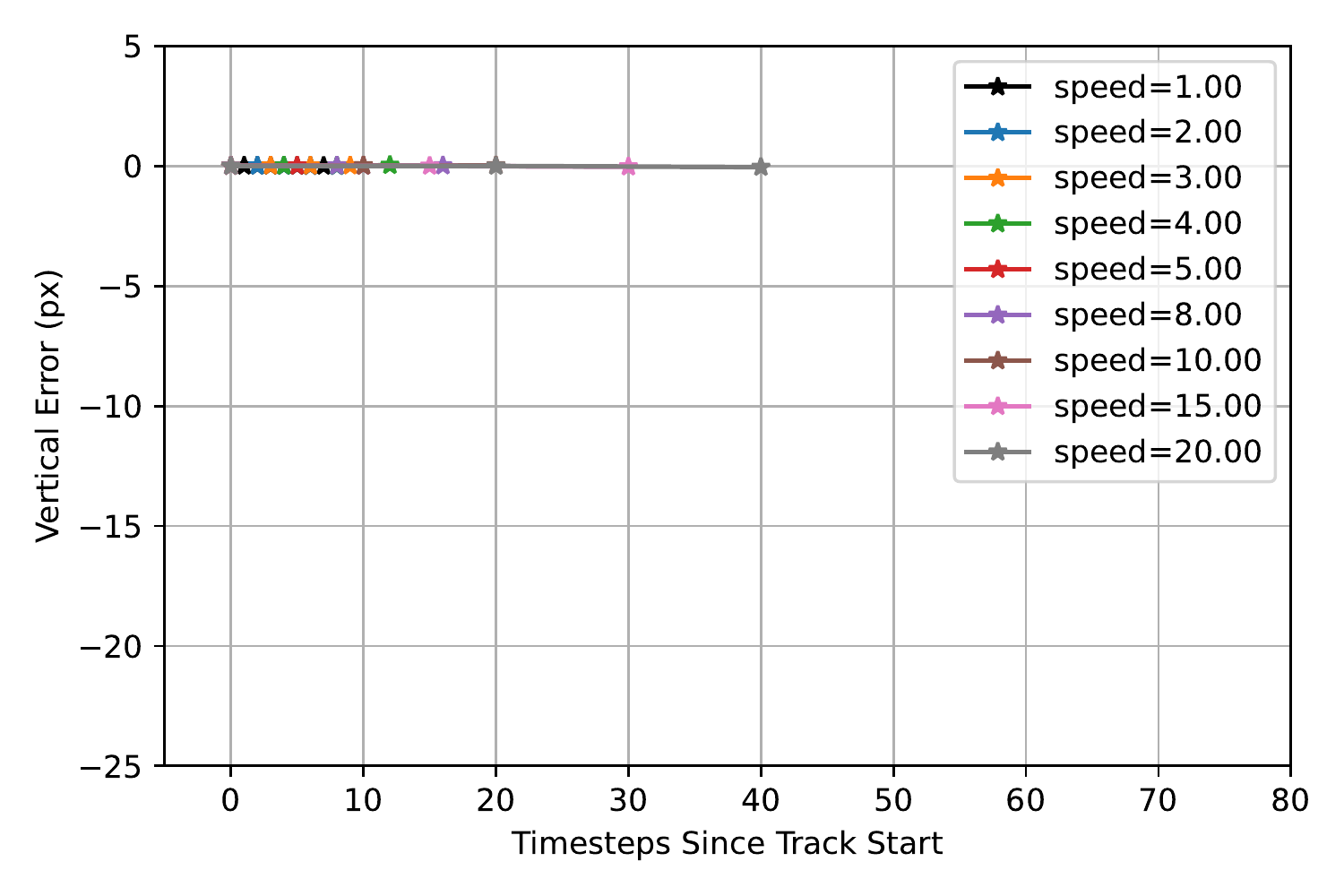}
        \includegraphics[width=0.48\textwidth]{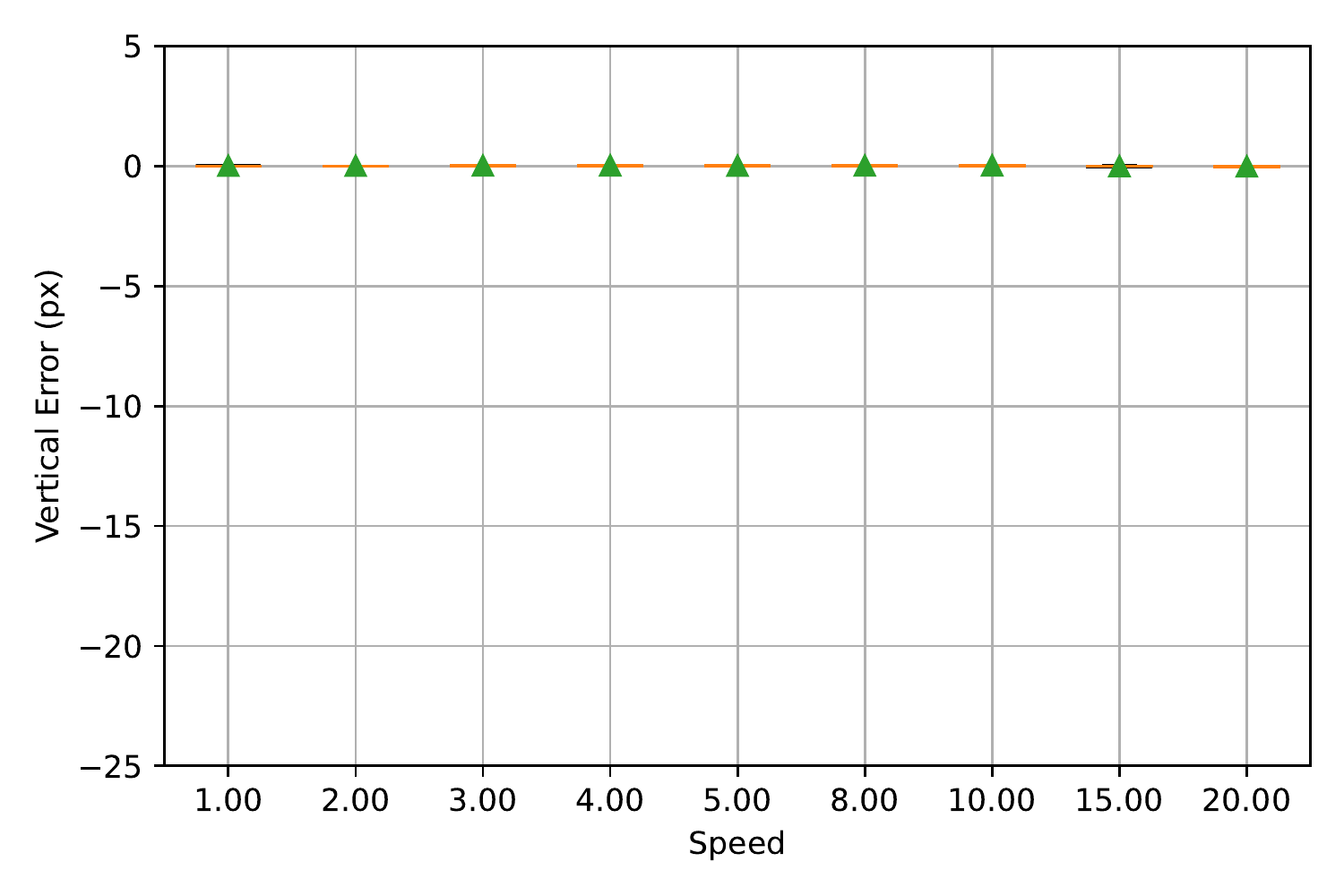}
    }
    \caption{\textbf{Gazebo Linear Dataset: Mean errors are unaffected by speed when using the Correspondence Tracker until tracking failure occurs.}
    The left column contains plots of the horizontal (top row) and vertical (bottom row) components of the mean tracking error $\nu(t)$ at each timestep $t$ after initial feature detection at multiple speeds. Each dot corresponds to a processed frame; lines for higher speeds contain data from fewer frames and therefore show fewer dots. The right column plots the ordinate values of each line for $t>0$ in the left figures as a box plot: means are shown as green triangles and medians are shown as orange lines. In the horizontal coordinate, mean errors remain near zero as speed is increased from 1.00 to 15.00. Mean errors are larger when speed=20.00. The mean error is close to zero in the vertical coordinate.
    }
    \label{fig:gazebo_linear_match_meanerror}
\end{figure}

\begin{figure}[H]
    \centering
    \subfigure[$\eta(t)$, Horizontal Coordinate]{
        \includegraphics[width=0.48\textwidth]{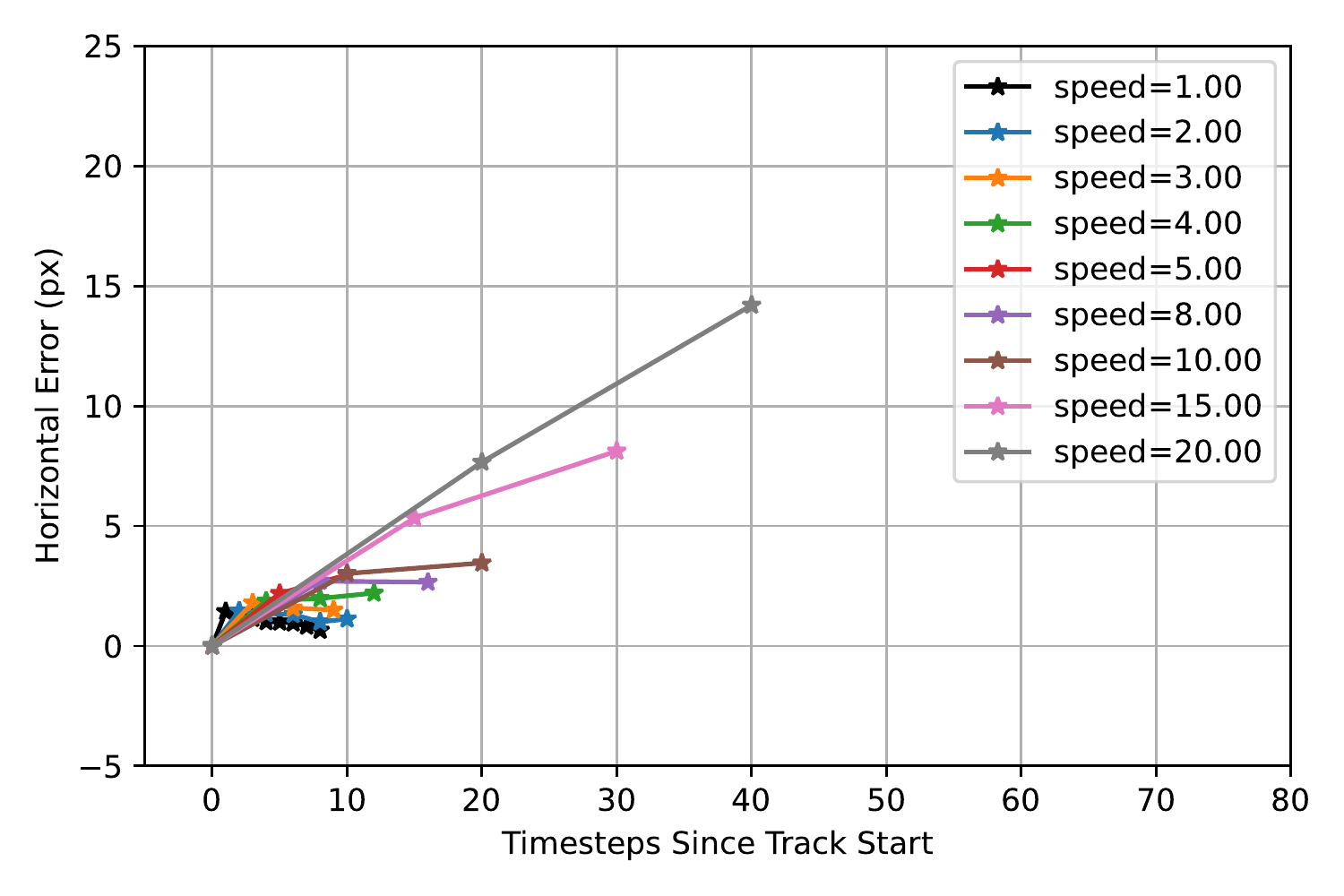}
        \includegraphics[width=0.48\textwidth]{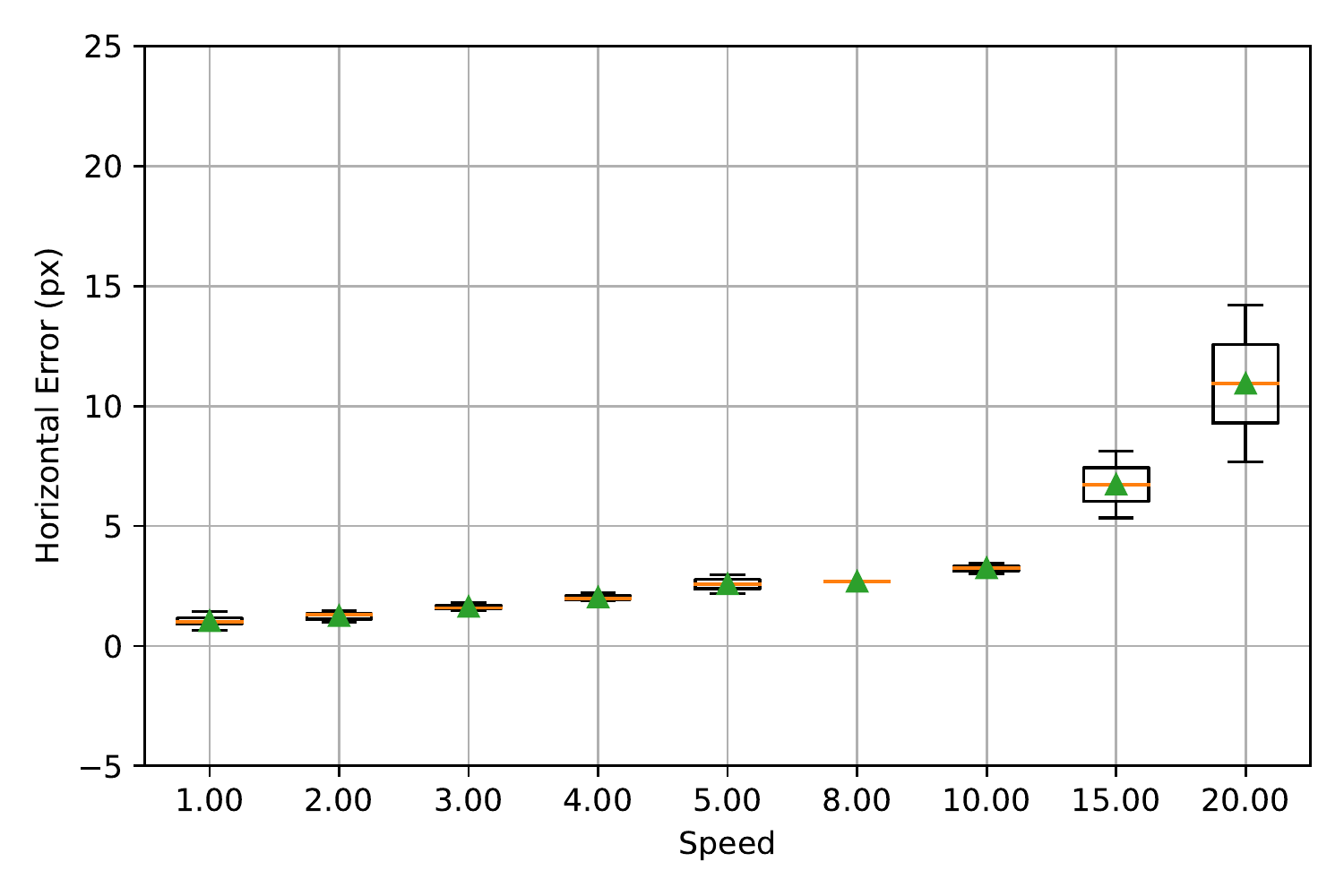}
    }
    \subfigure[$\eta(t)$, Vertical Coordinate]{
        \includegraphics[width=0.48\textwidth]{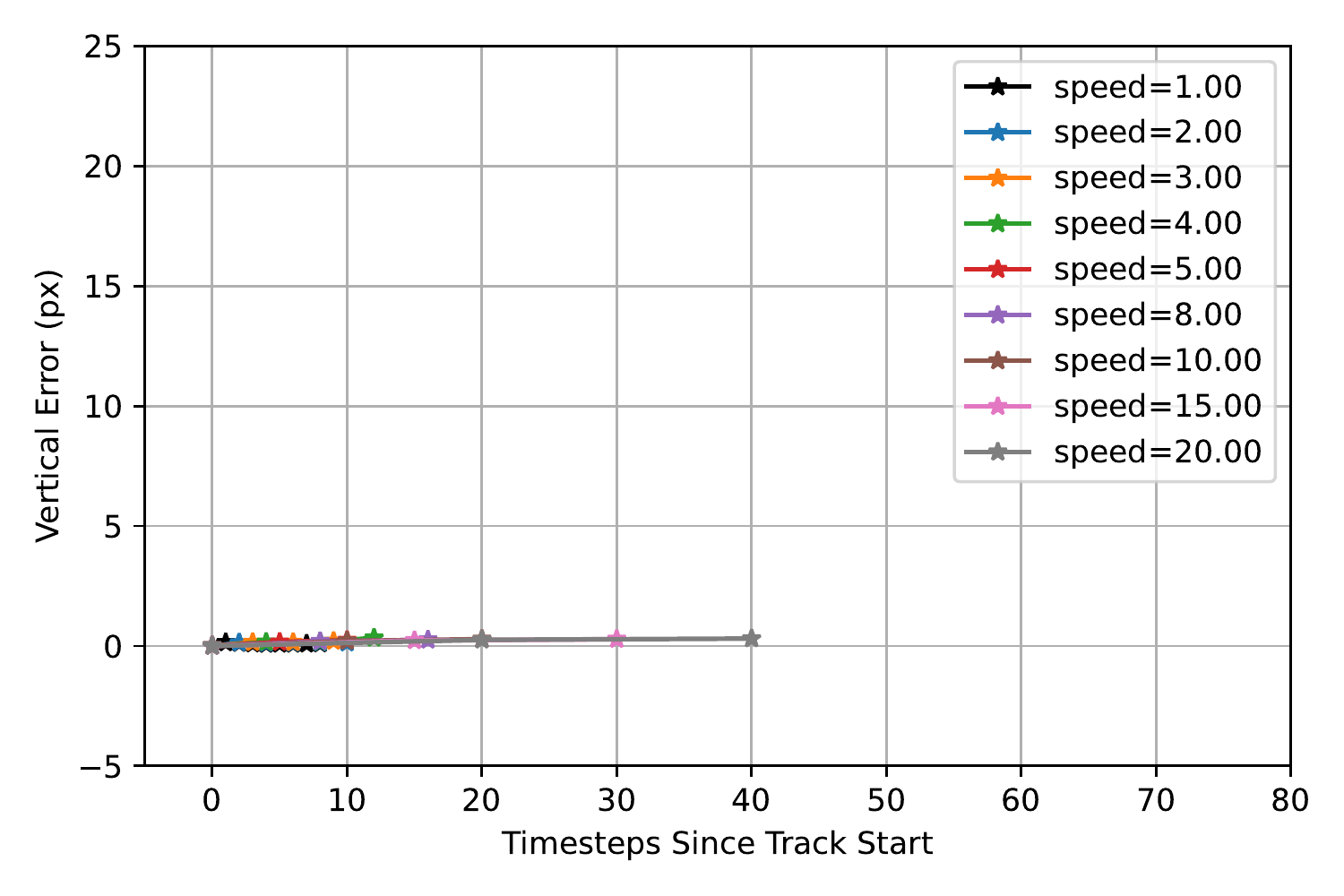}
        \includegraphics[width=0.48\textwidth]{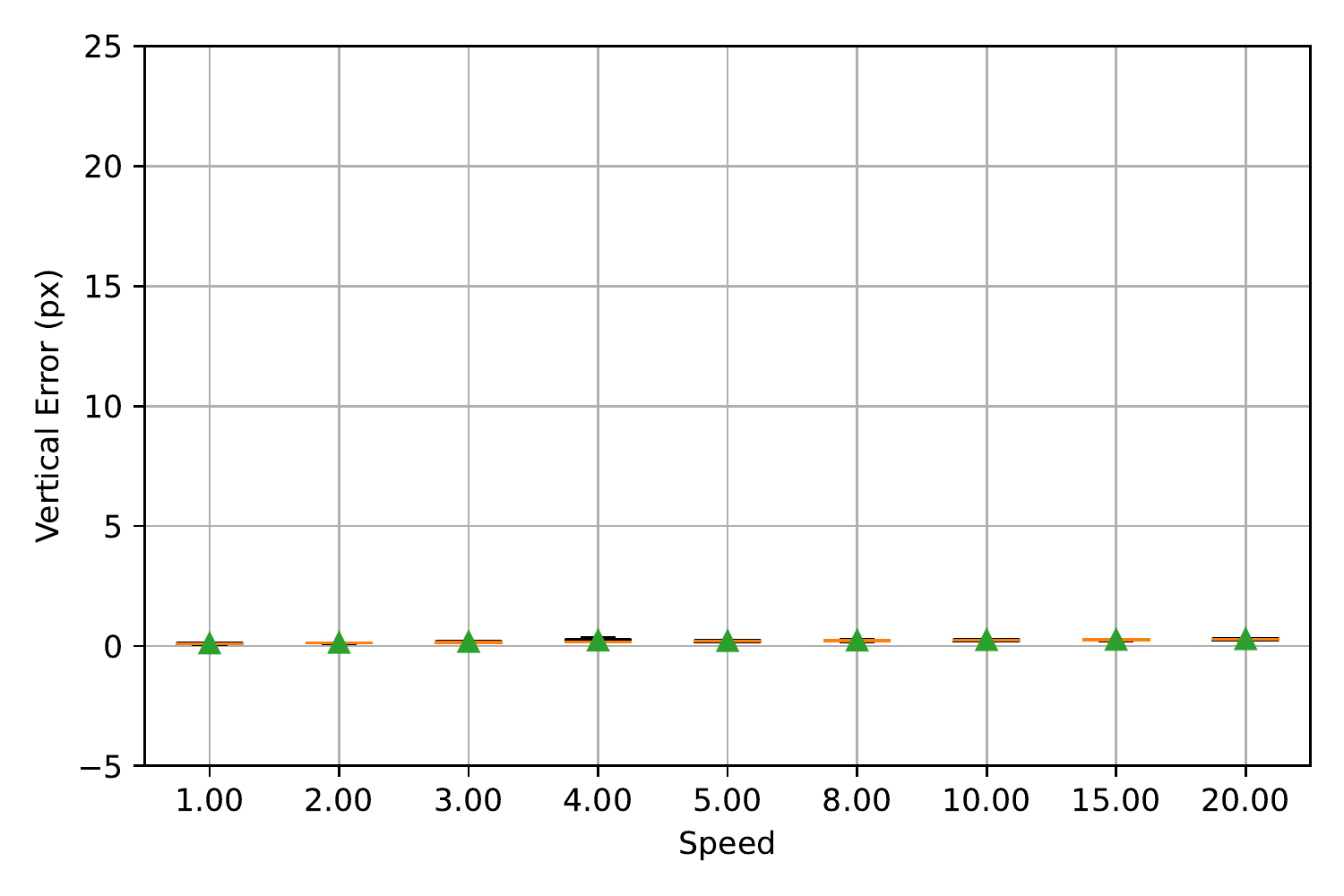}
    }
    \caption{\textbf{Gazebo Linear Dataset: Mean absolute errors increase with speed when using the Correspondence Tracker.}
    The left column contains plots of the horizontal (top row) and vertical (bottom row) components of the mean absolute error $\eta(t)$ at each timestep $t$ after initial feature detection at multiple speeds. Each dot corresponds to a processed frame; lines for higher speeds contain data from fewer frames and therefore show fewer dots. The right column plots the ordinate values of each line for $t>0$ in the left figures as a box plot: means are shown as green triangles and medians are shown as orange lines. In the horizontal coordinate, mean absolute errors increase slowly with speed at first; increases are larger from speed=10.00 to speed=15.00 and speed=15.00 to speed=20.00. The mean absolute error is approximately 0 in the vertical coordinate.
    }
    \label{fig:gazebo_linear_match_MAE}
\end{figure}

\begin{figure}[H]
    \centering
    \subfigure[$\Phi(t)$, Horizontal Coordinate]{
        \includegraphics[width=0.48\textwidth]{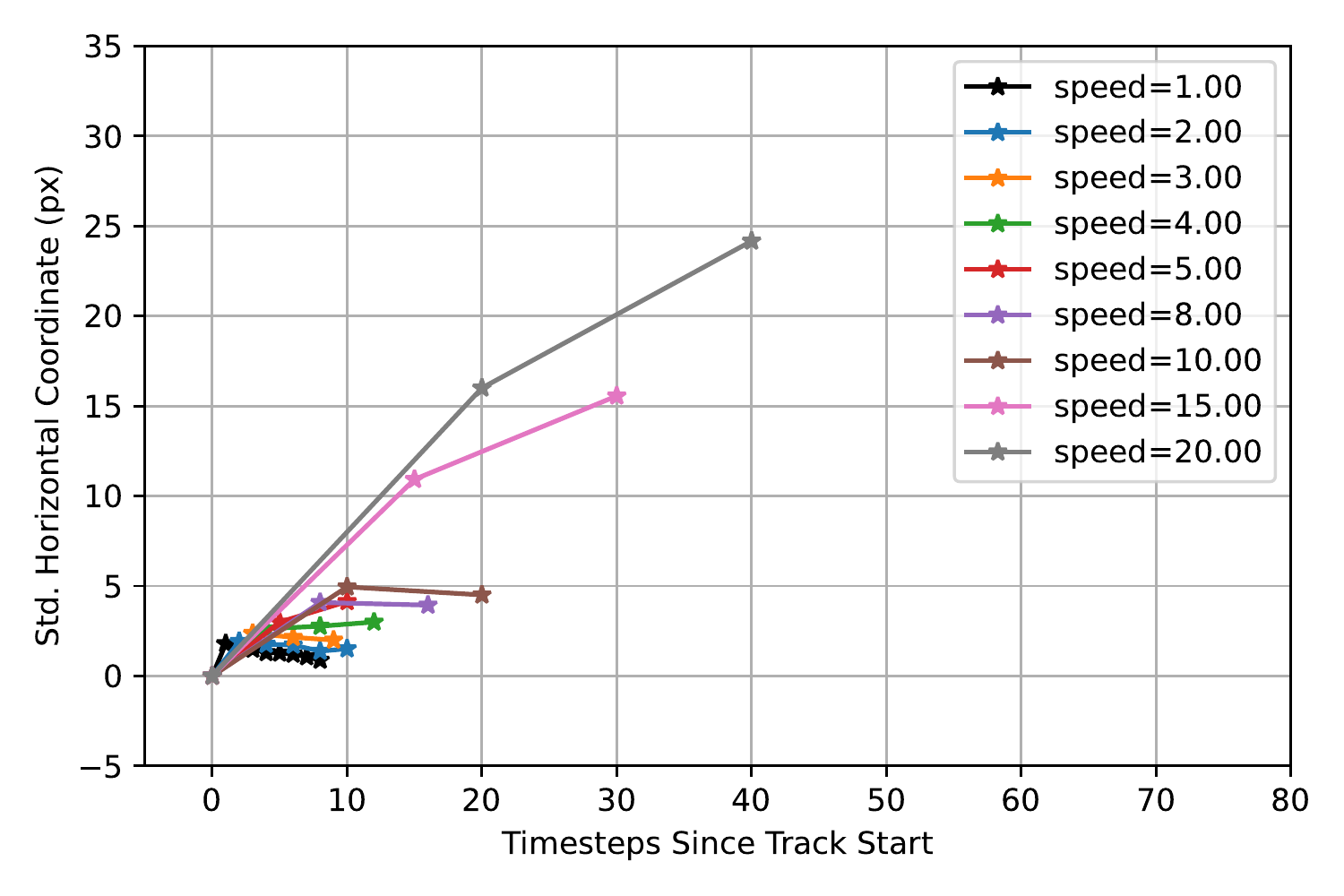}
        \includegraphics[width=0.48\textwidth]{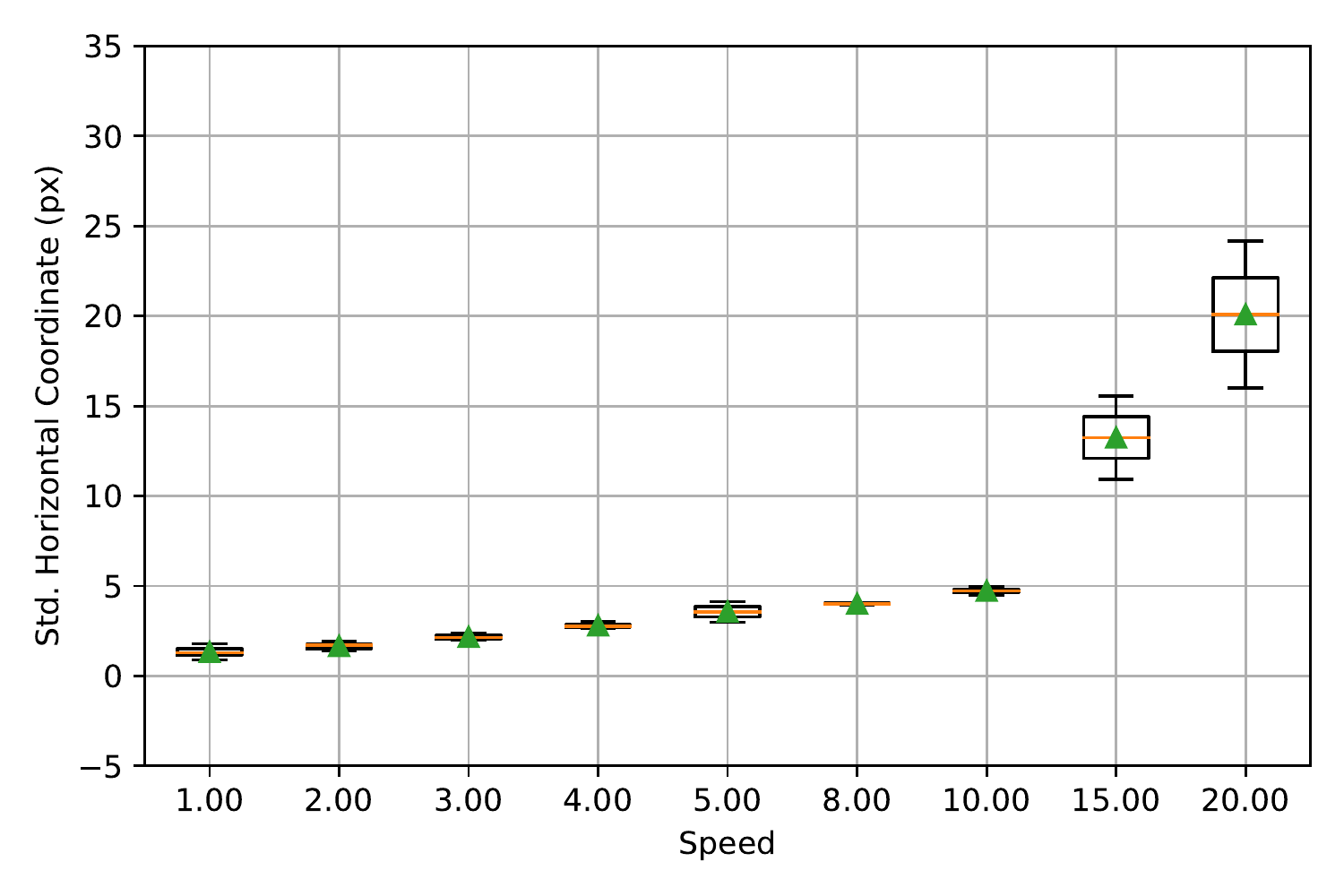}
    }
    \subfigure[$\Phi(t)$, Vertical Coordinate]{
        \includegraphics[width=0.48\textwidth]{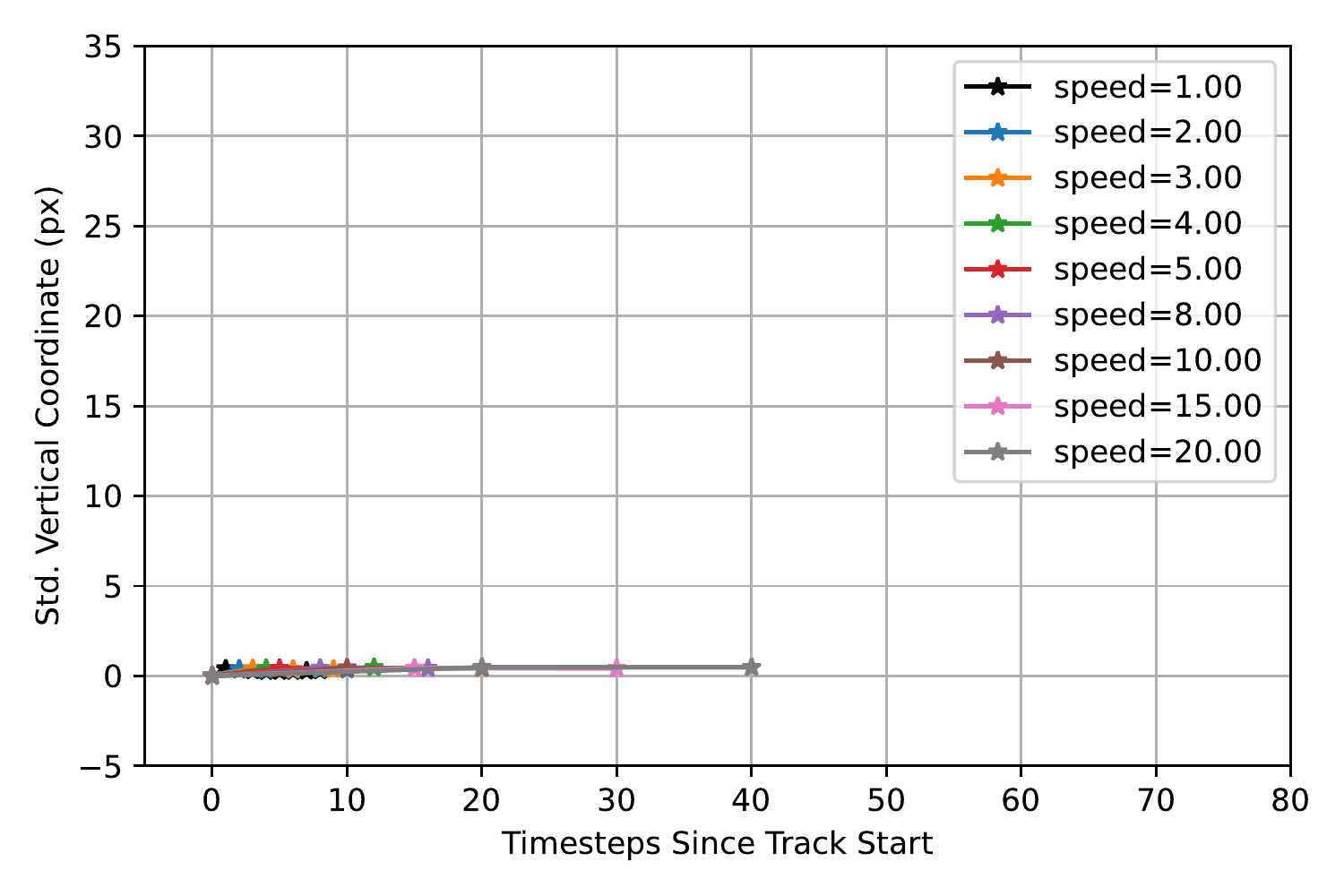}
        \includegraphics[width=0.48\textwidth]{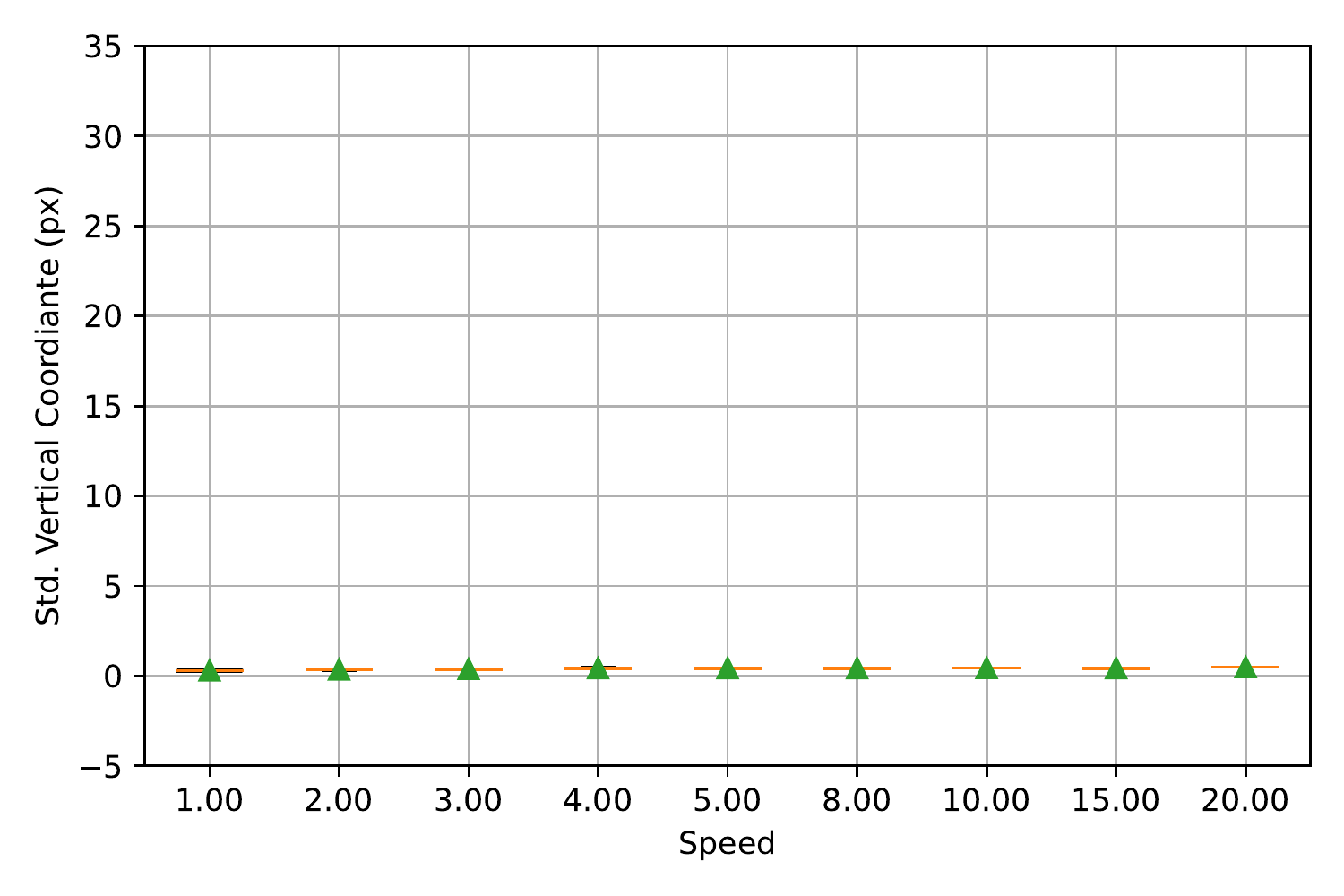}
    }
    \caption{\textbf{Gazebo Linear Dataset: Covariance increases speed when using the Correspondence Tracker.}
    The left column contains plots of the horizontal (top row) and vertical (bottom row) components of the mean tracking error $\Phi(t)$ at each timestep $t$ after initial feature detection at multiple speeds. Each dot corresponds to a processed frame; lines for higher speeds contain data from fewer frames and therefore show fewer dots. The right column plots the ordinate values of each line for $t>0$ in the left figures as a box plot: means are shown as green triangles and medians are shown as orange lines. In the horizontal coordinate, covariance increases slowly with speed at first; increases are larger from speed=10.00 to speed=15.00 and speed=15.00 to speed=20.00. The covariance is close to zero in the vertical coordinate.
    }
    \label{fig:gazebo_linear_match_cov}
\end{figure}

\begin{figure}
    \centering
    \subfigure[$\nu(t)$, Horizontal Coordinate]{\includegraphics[width=0.48\textwidth]{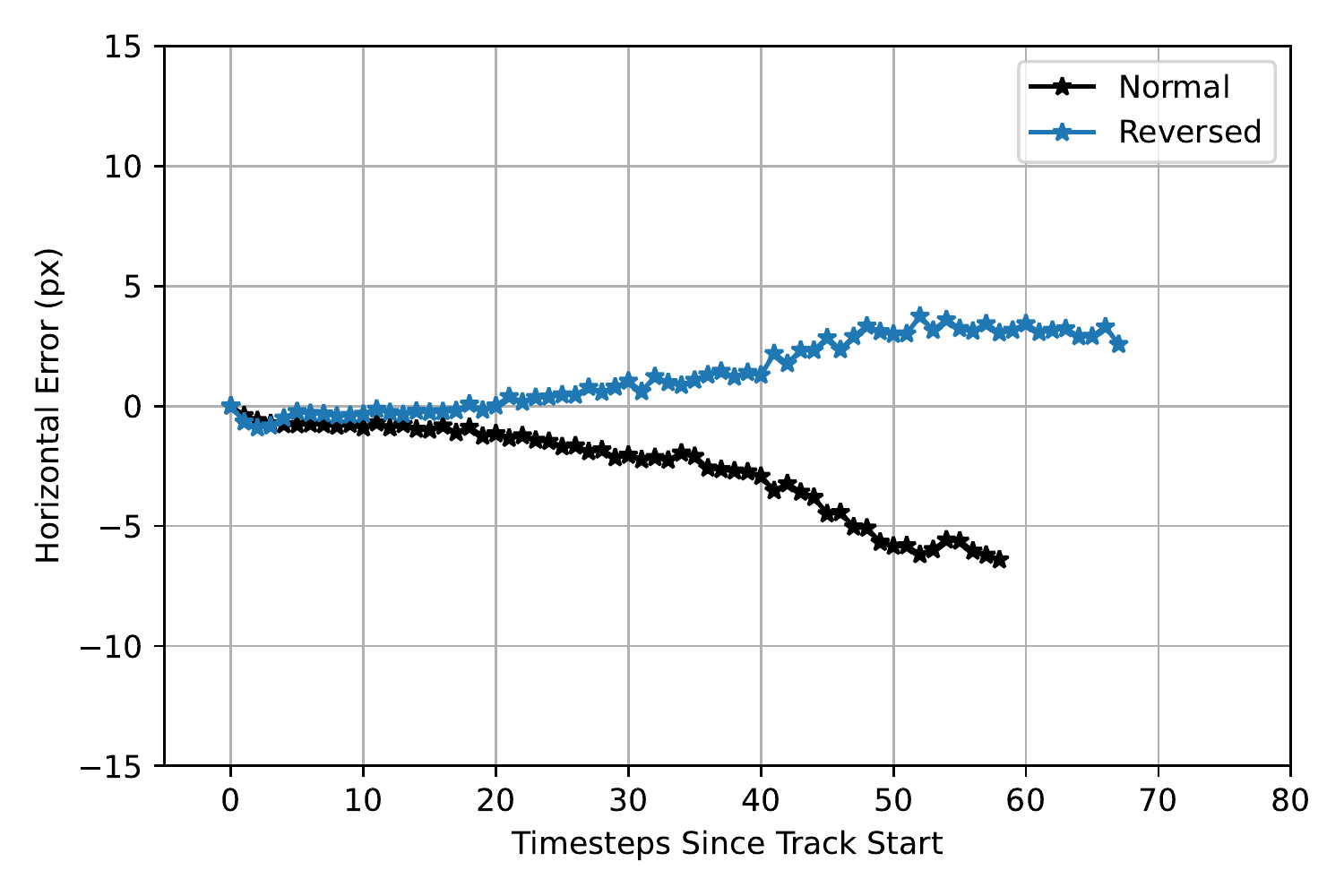}}
    \subfigure[$\nu(t)$, Vertical Coordinate]{\includegraphics[width=0.48\textwidth]{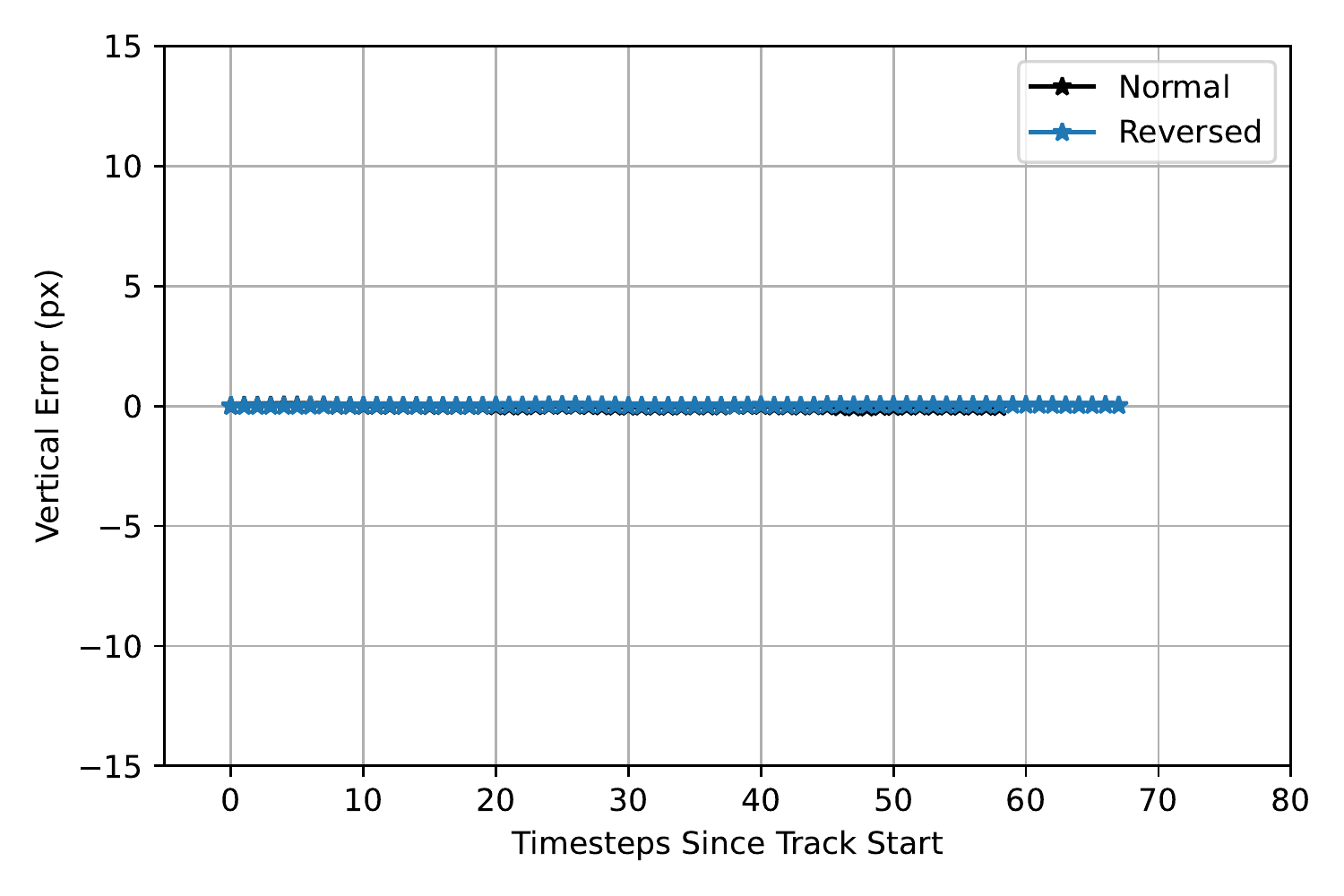}}
    \caption{\textbf{Gazebo Linear Dataset: The Lucas-Kanade Tracker drifts opposite the direction of motion.} Lines above contain $\nu(t)$ computed from tracks using the Lucas-Kanade Tracker. In the black lines, the quadrotor is flying horizontally from left to right, as is the case in the rest of the experiments on the Gazebo Linear Dataset. In the blue lines, the quadrotor is flying horizontally from right to left while observing the same scene; the scene is not mirror-imaged, so the features tracked in the two trajectories are not identical. Once again, there is nearly no mean error in the vertical direction. However, mean horizontal error is positive instead of negative.}
    \label{fig:gazebo_linear_backwards}
\end{figure}

\subsection{Supporting Figures for Gazebo Dataset with AR/VR Motion}
\label{sec:all_gazebo_arvr_figs}

\begin{figure}[H]
\centering
\includegraphics[width=0.48\textwidth]{feature_tracker_uq/gazebo_linear_figs/gazebo_wall.png}
\includegraphics[width=0.48\textwidth]{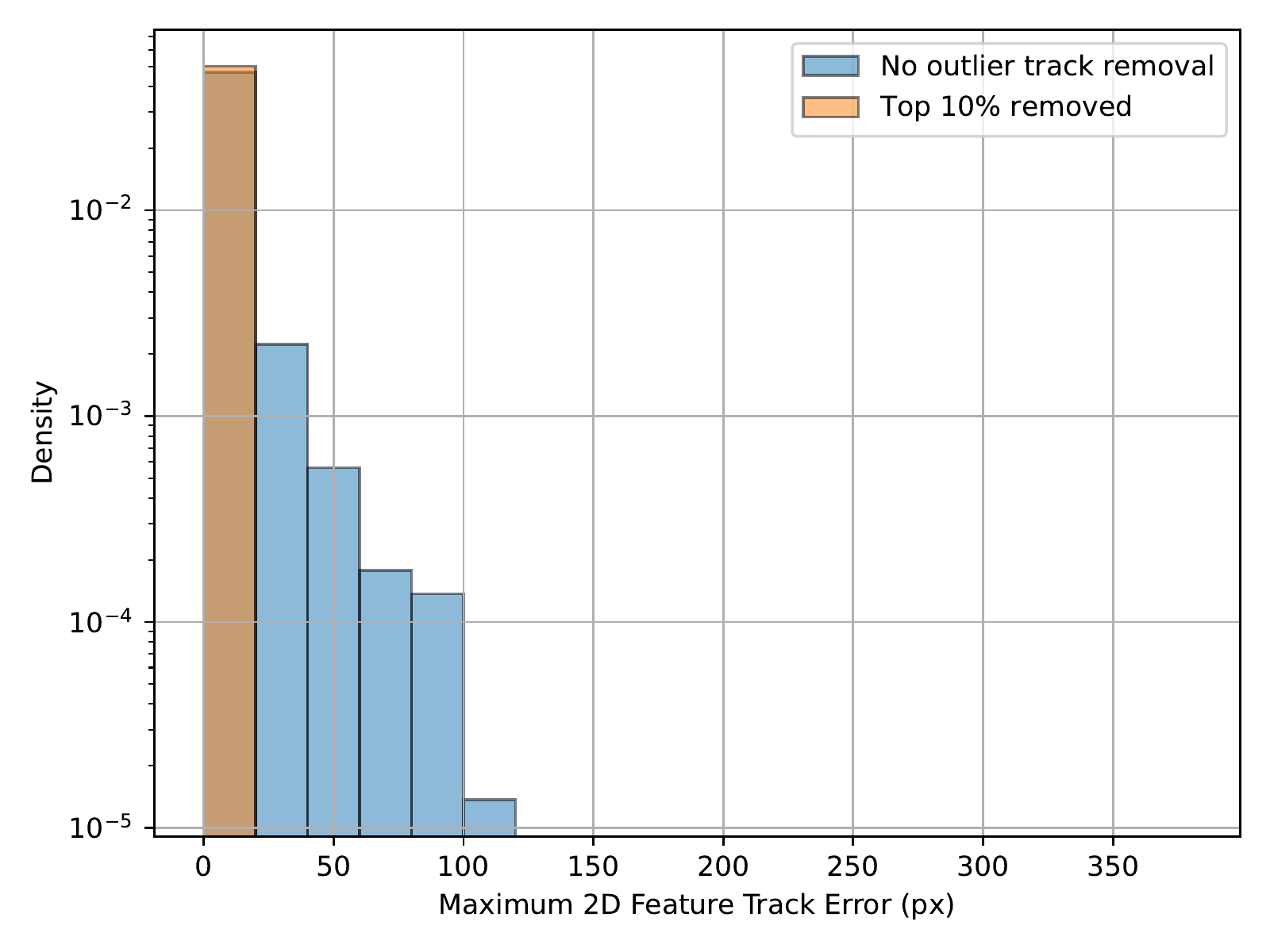}
\caption{\textbf{Gazebo AR/VR Dataset: We will throw out the 10\% of tracks from each scene with the most error.} The right figure plots the histogram density of the maximum L2 error of all feature tracks of a single scene in log scale. The corresponding scene is pictured on the left. The outlier errors are caused by poor depth association that is the result of depth images and AR/VR images not being collected synchronously in the Gazebo simulation environment.}
\label{fig:gazebo_arvr_error_throwout}
\end{figure}

\begin{figure}[H]
    \subfigure[Lucas-Kanade]{\includegraphics[width=0.48\textwidth]{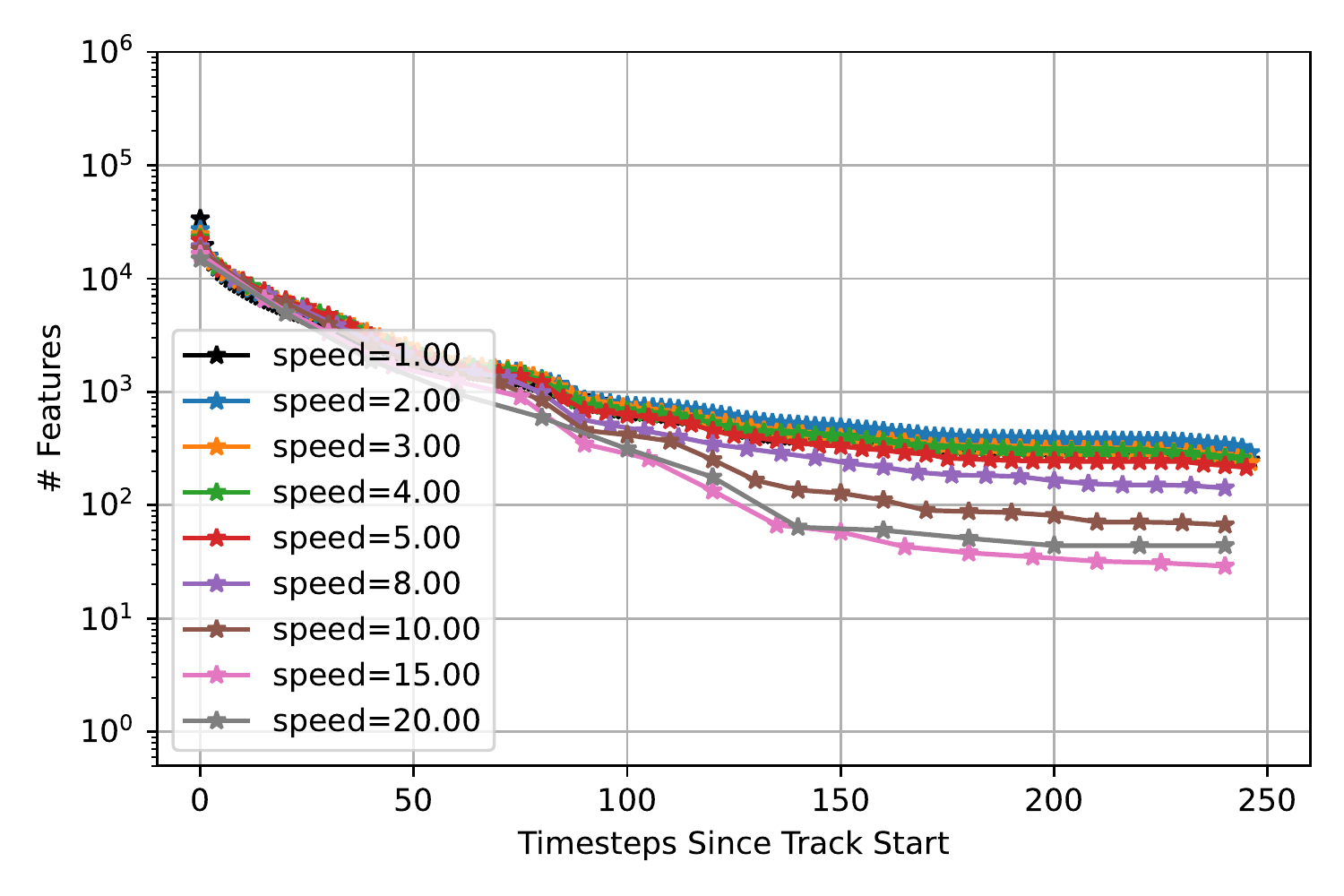}}
    \subfigure[Correspondence]{\includegraphics[width=0.48\textwidth]{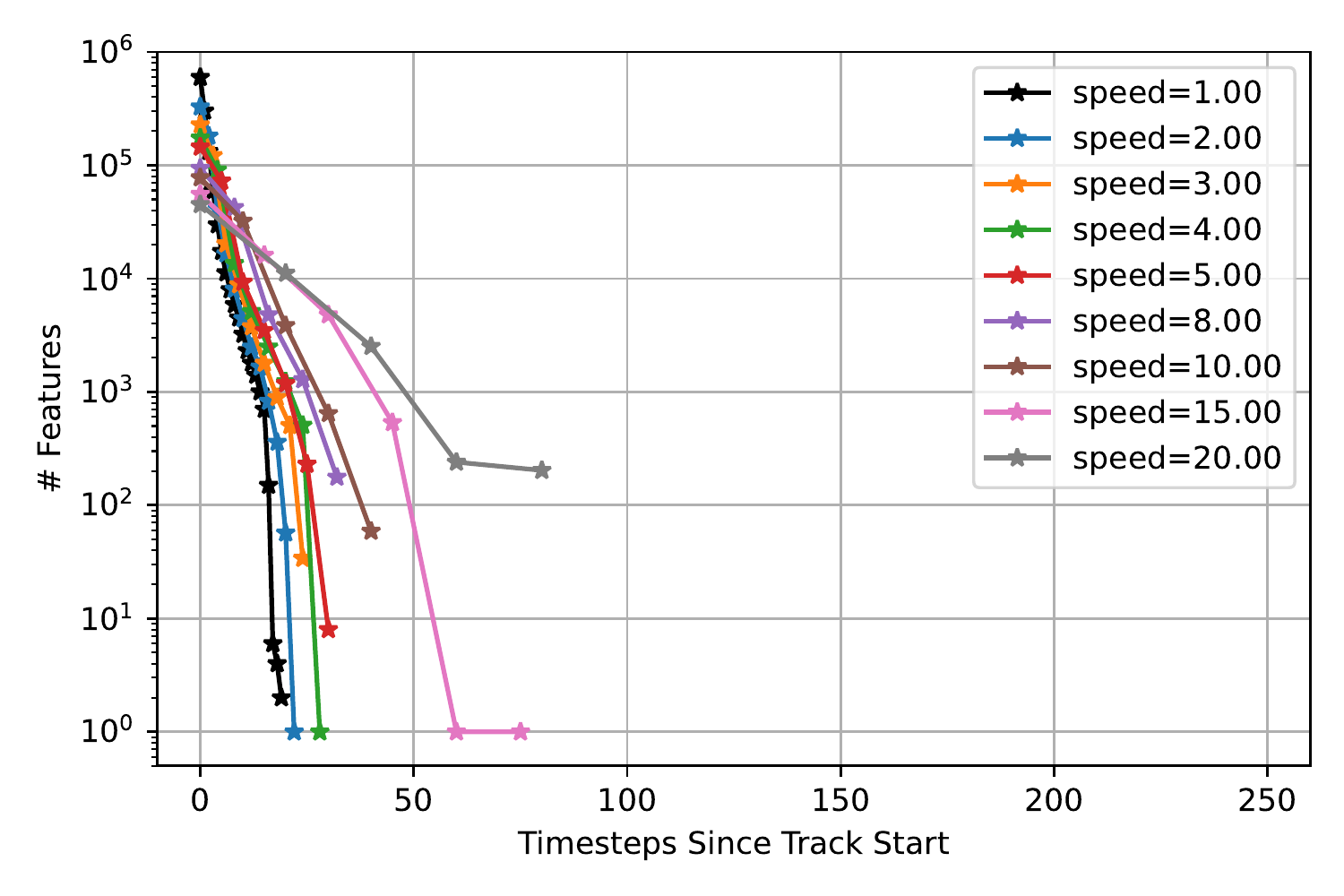}}
    \caption{Feature Lifetime is plotted on the horizontal axis. The vertical axis, in log scale, shows the number of features in all 11 scenes that lived at least that long for every tested speed. The number of features drops very fast when the Correspondence Tracker is used. On the other hand, many features have a long lifetime when the Lucas-Kanade tracker is used, since the scene is persistent. \textbf{In subsequent analyses, we only compute mean errors and covariances at timesteps with at least 100 features on the Gazebo AR/VR Dataset.}}
    \label{fig:gazebo_arvr_avg_feats}
\end{figure}

\begin{figure}[H]
    \centering
    \subfigure[Lucas-Kanade]{\includegraphics[width=0.48\textwidth]{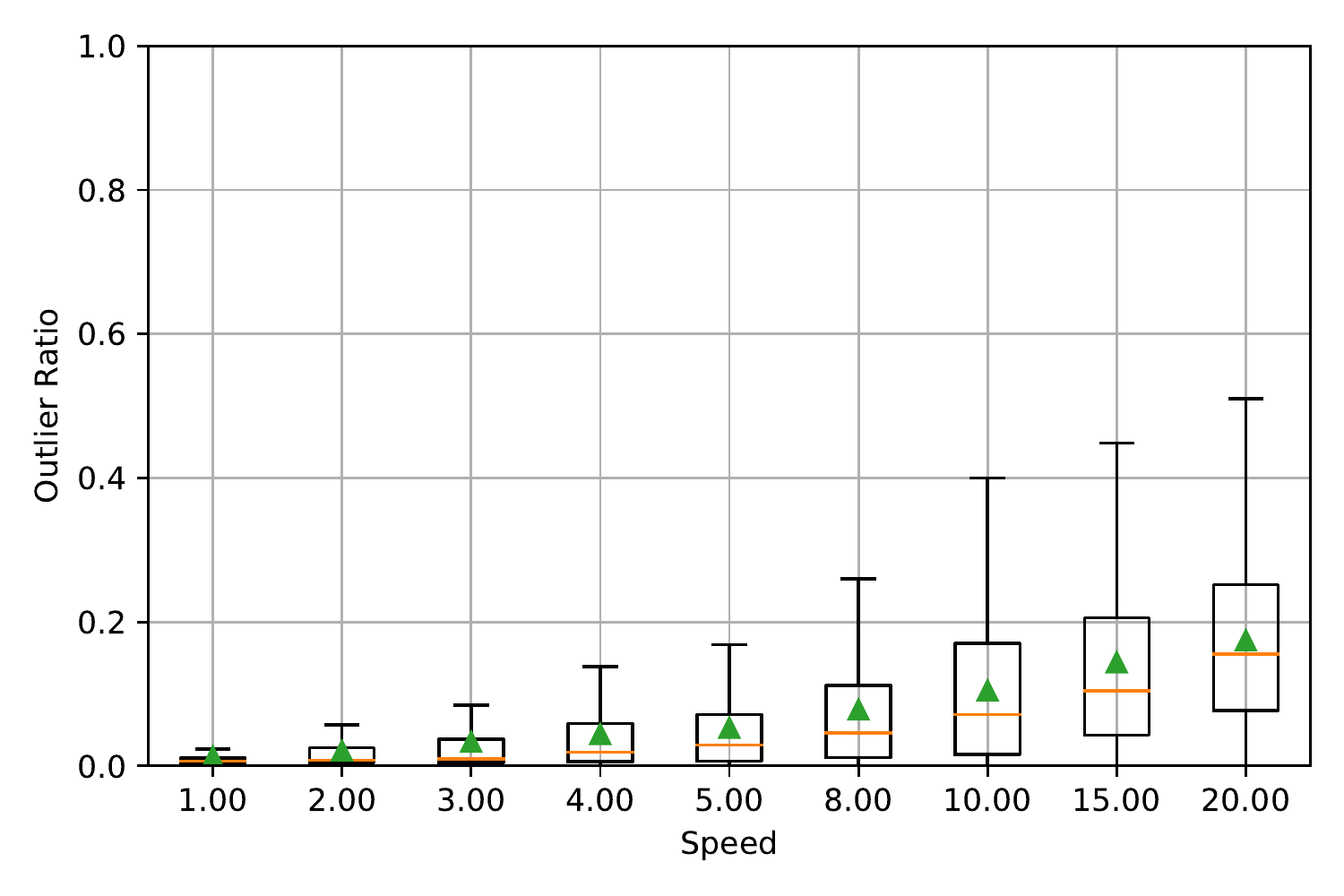}}
    \subfigure[Correspondence]{\includegraphics[width=0.48\textwidth]{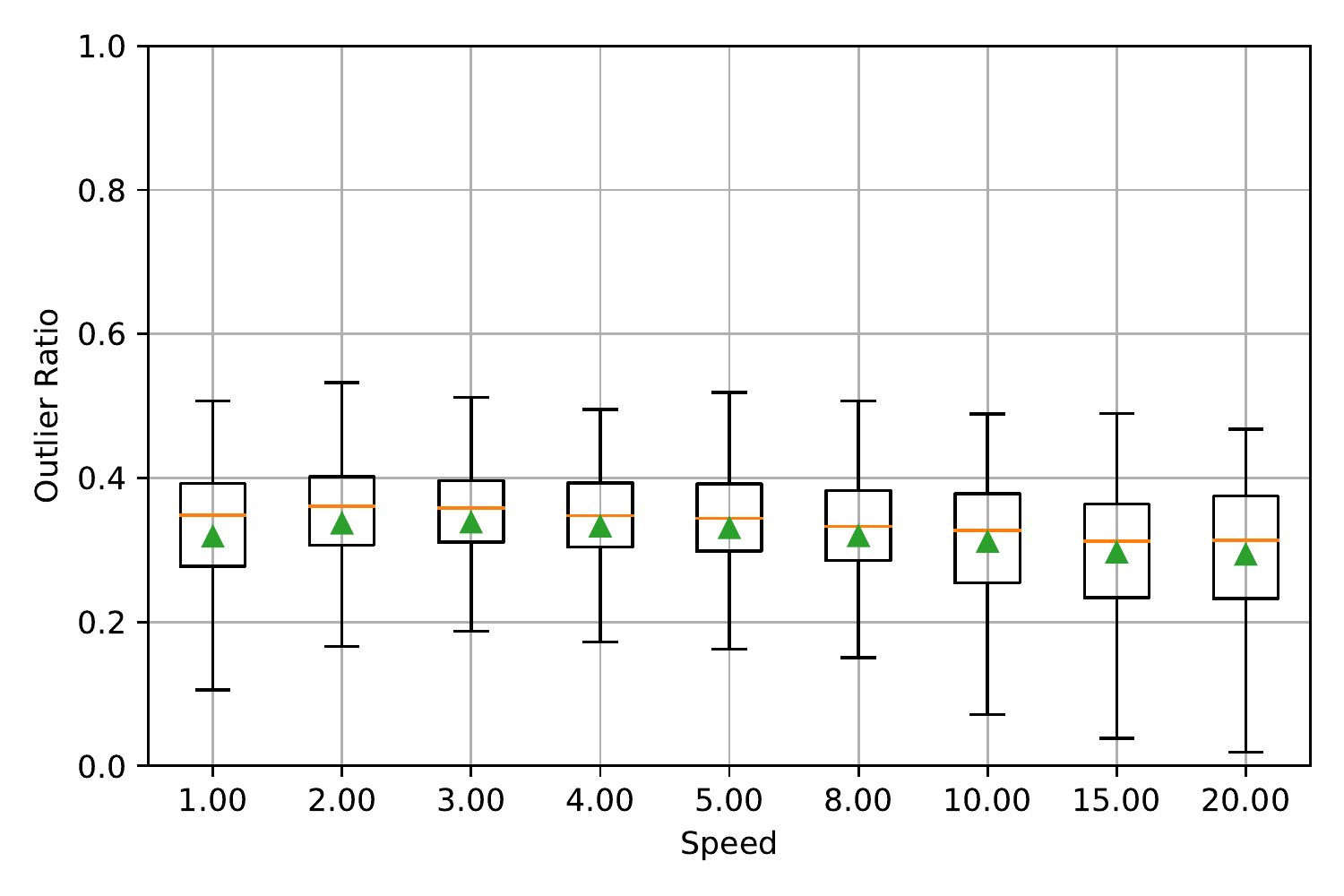}}
    \caption{\textbf{Gazebo AR/VR Dataset: Outlier Ratio increases with speed when using the Lucas-Kanade Tracker and are constant when using the Correspondence Tracker.} Outlier ratios are shown as box-and-whisker plots for tested speeds for the Lucas-Kanade tracker on the left and the Correspondence Tracker on the right. For lower speeds, the Lucas-Kanade tracker produces fewer outliers. Outlier ratios then increase with speed. On the other hand, the outlier ratio for the Correspondence Tracker remains constant, between 30 and 40 percent.}
    \label{fig:gazebo_arvr_outlier_ratio}
\end{figure}

\begin{figure}[H]
    \centering
    \subfigure[$\nu(t)$, Horizontal Coordinate]{\includegraphics[width=0.48\textwidth]{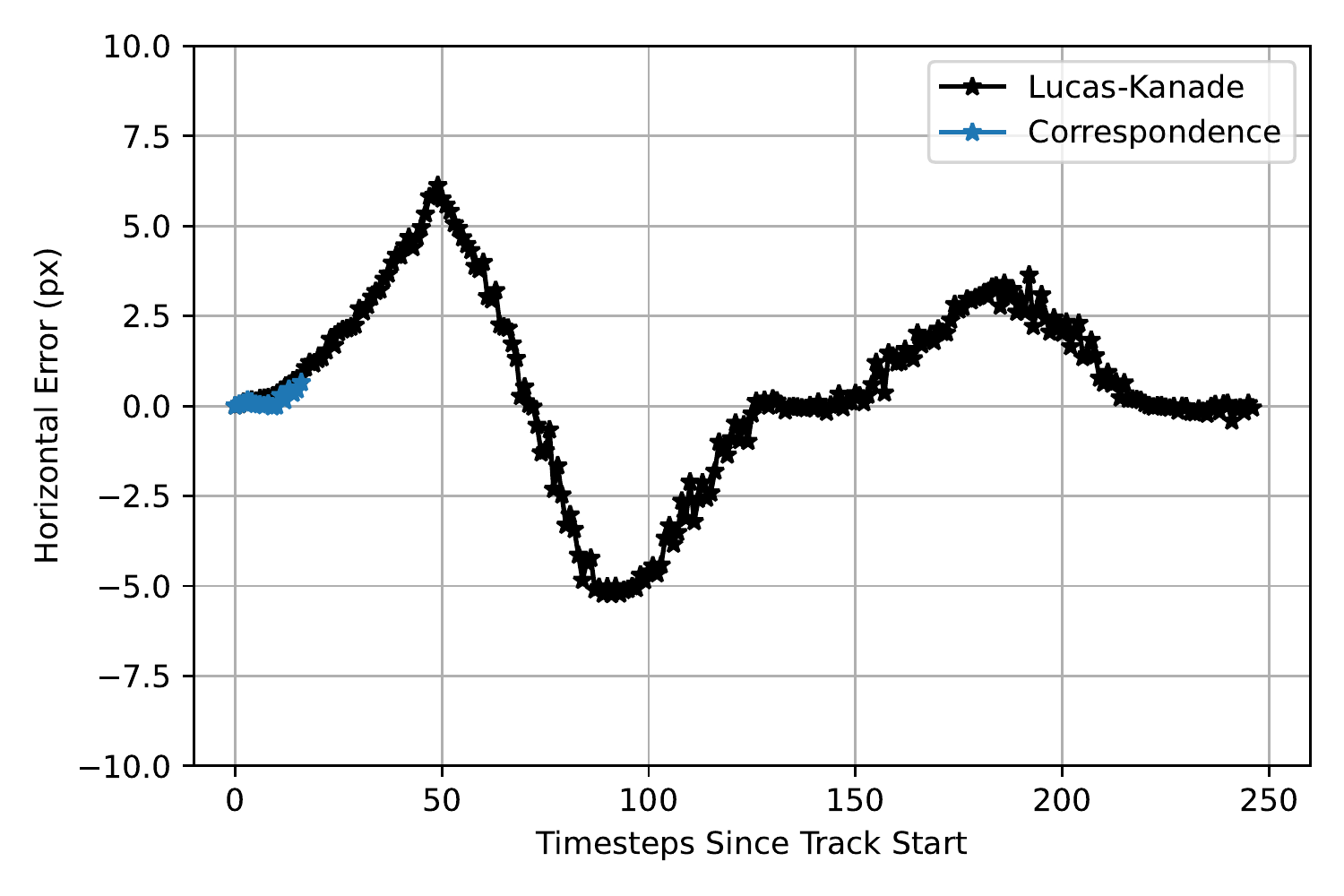}}
    \subfigure[$\nu(t)$, Vertical Coordinate]{\includegraphics[width=0.48\textwidth]{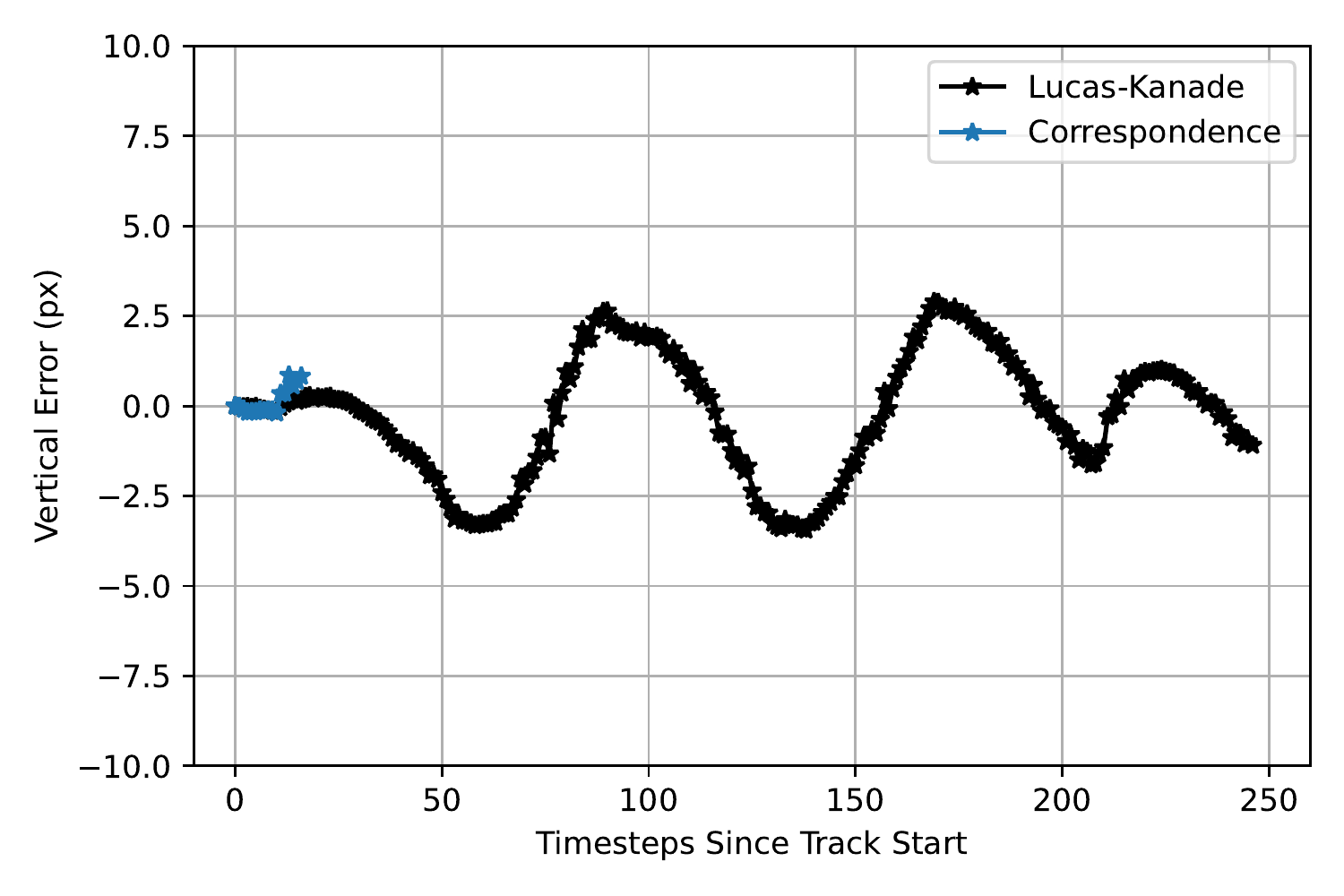}}
    \caption{\textbf{Gazebo AR/VR Dataset: At nominal speed, the Lucas-Kanade Tracker accumulates drift that changes with motion. The Correspondence Tracker's error is not zero-mean, but tracks do not live long enough to accumulate.} Lines shown are horizontal (left) and vertical (right) coordinates of mean error $\nu(t)$ calculated using tracks averaged over all scenes; calculation is cut off at 58 frames for the Lucas-Kanade Tracker and 247 frames for the Correspondence Tracker so that averages can be computed with at least 100 features.}
    \label{fig:gazebo_arvr_1.00_meanerror}
\end{figure}

\begin{figure}[H]
    \centering
    \subfigure[$\eta(t)$, Horizontal Coordinate]{\includegraphics[width=0.48\textwidth]{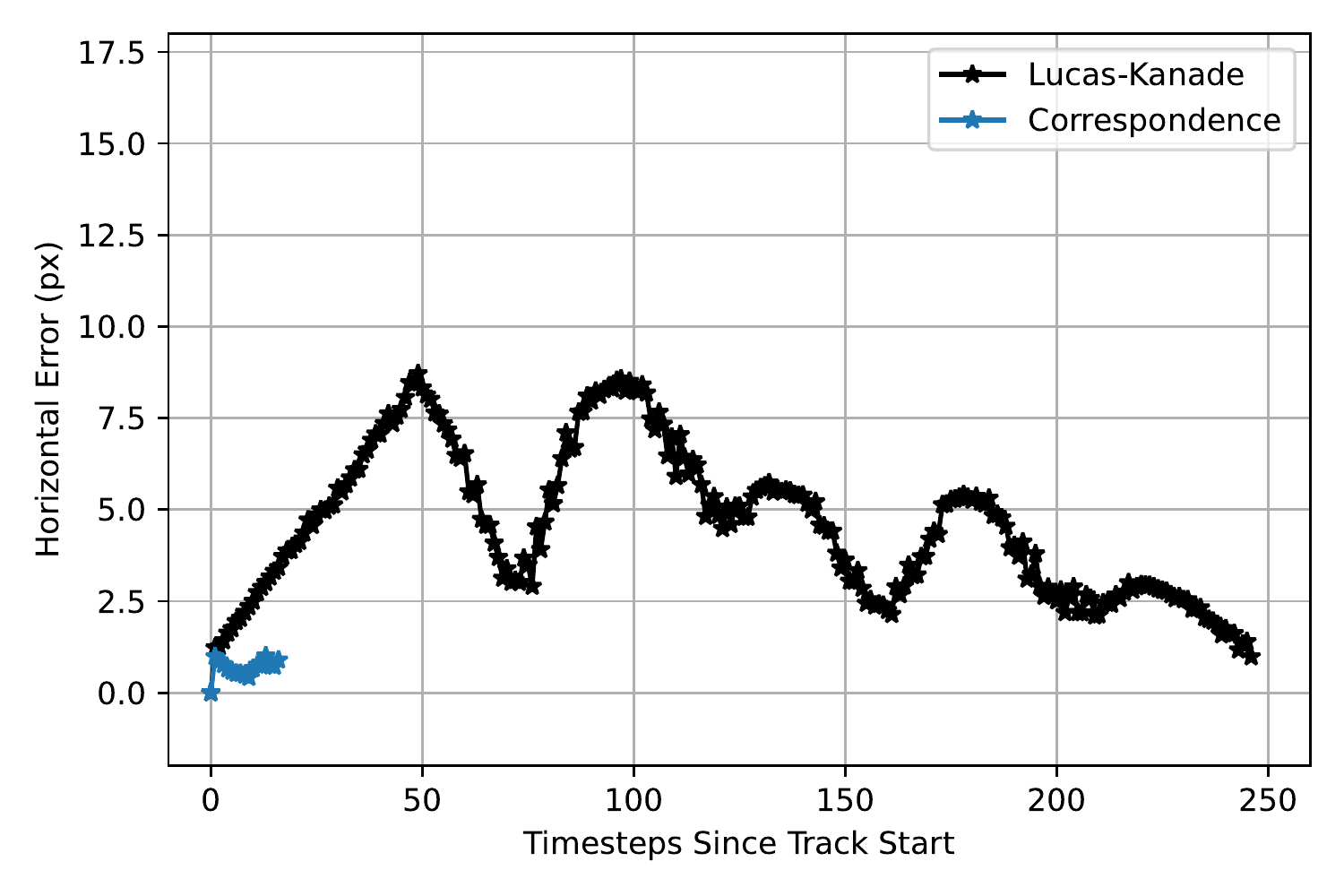}}
    \subfigure[$\eta(t)$, Vertical Coordinate]{\includegraphics[width=0.48\textwidth]{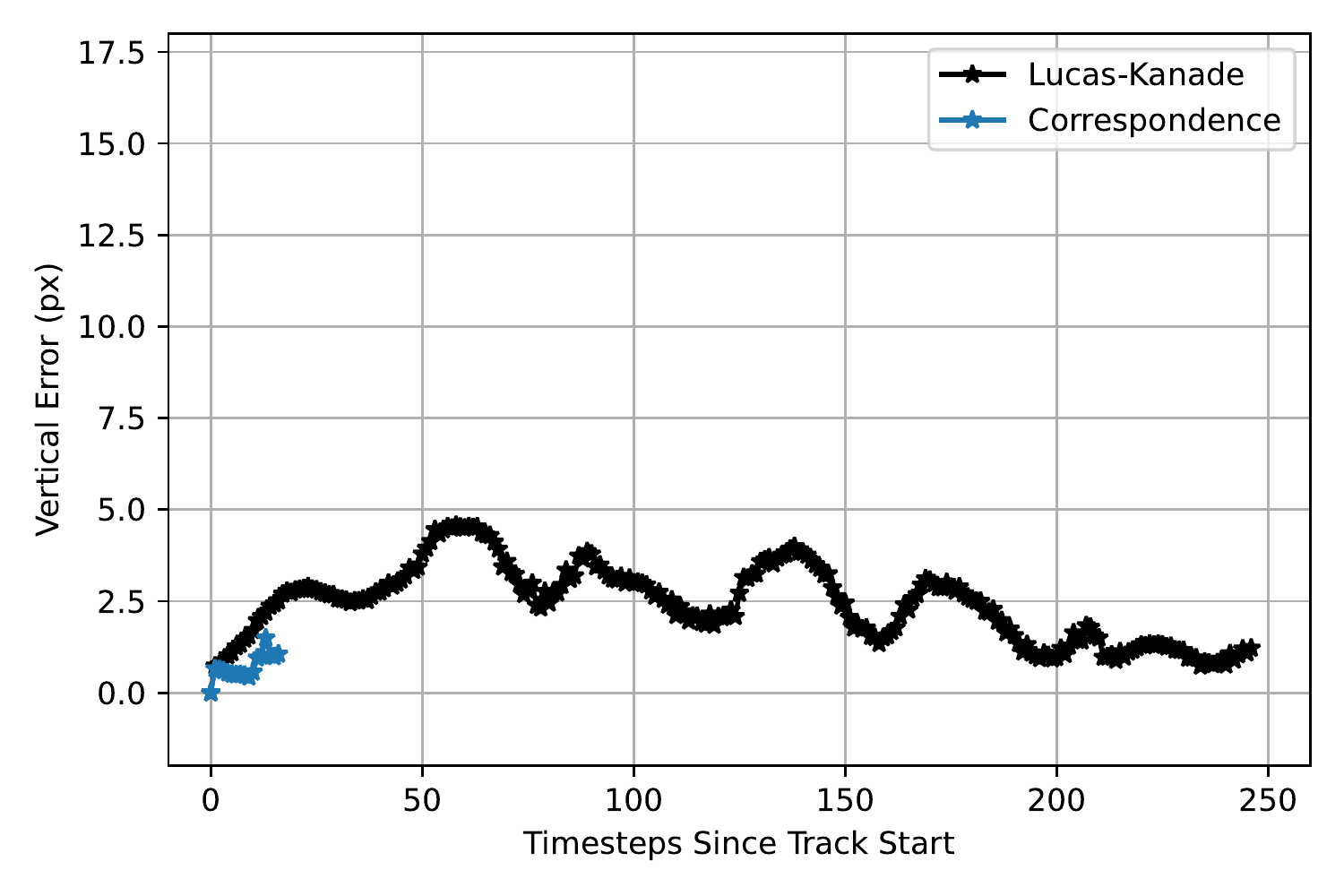}}   
    \subfigure[$\Phi(t)$, Horizontal Coordinate]{\includegraphics[width=0.48\textwidth]{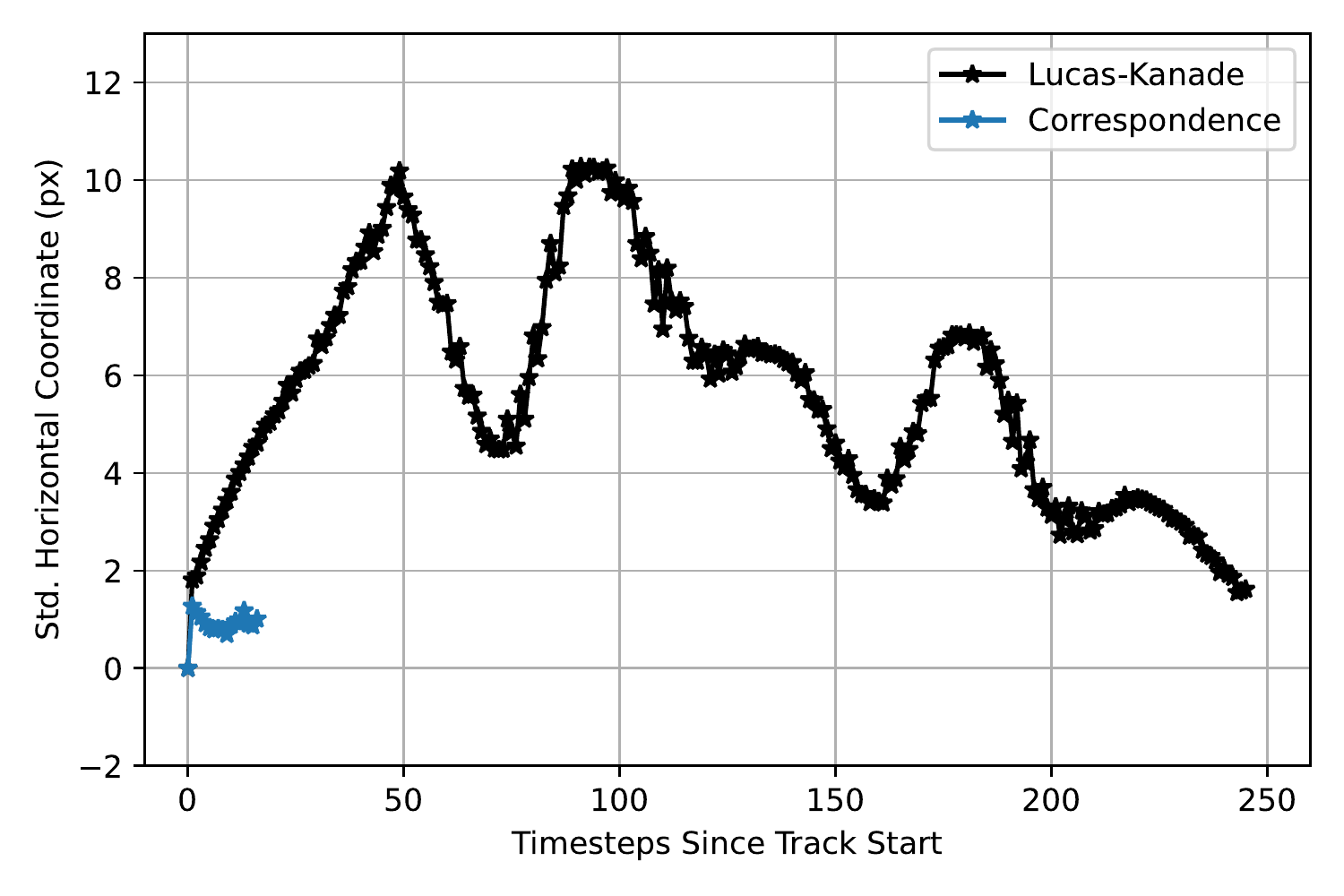}}
    \subfigure[$\Phi(t)$, Vertical Coordinate]{\includegraphics[width=0.48\textwidth]{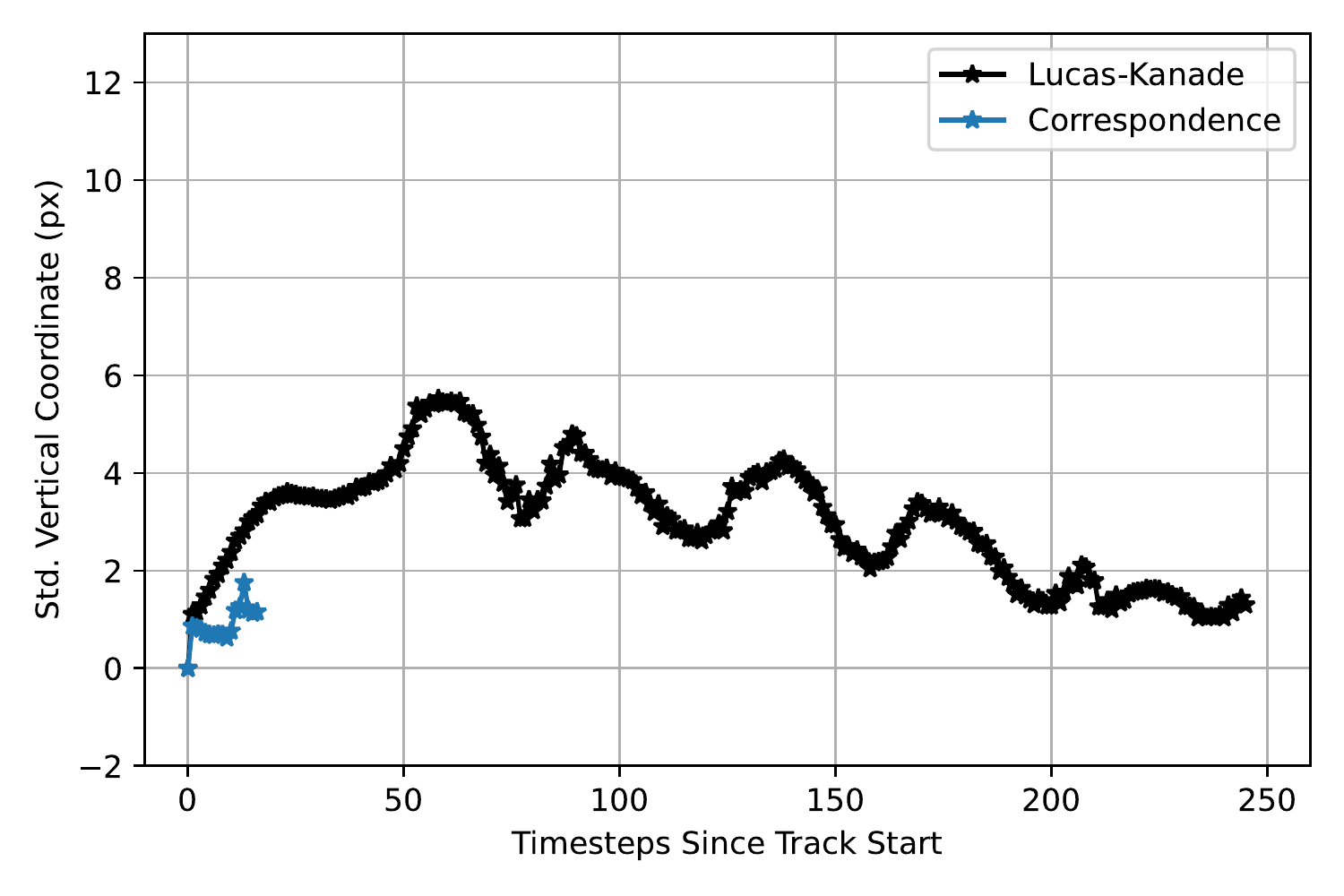}}      
    \caption{\textbf{Gazebo AR/VR Dataset: The Lucas-Kanade tracker drifts considerably more than the Correspondence Tracker.} Lines shown are horizontal (left column) and vertical (right column) coordinates of mean absolute error $\eta(t)$ (top row) and covariance $\Phi(t)$ (bottom row) calculated using tracks averaged over all scenes; calculation is cutoff at 247 frames for Lucas-Kanade Tracker and 17 frames for the Correspondence Tracker so that averages can be computed with at least 100 features. Both mean absolute error and covariance are roughly constant and small when using the Correspondence Tracker. On the other hand, the horizontal coordinate of $\eta(t)$ and $\Phi(t)$ drifts and changes with motion when using the Lucas-Kanade Tracker.}
    \label{fig:gazebo_arvr_1.00_error_cov}
\end{figure}

\begin{figure}[H]
    \centering
    \subfigure[$\eta(t)$, Horizontal Coordinate]{
        \includegraphics[width=0.48\textwidth]{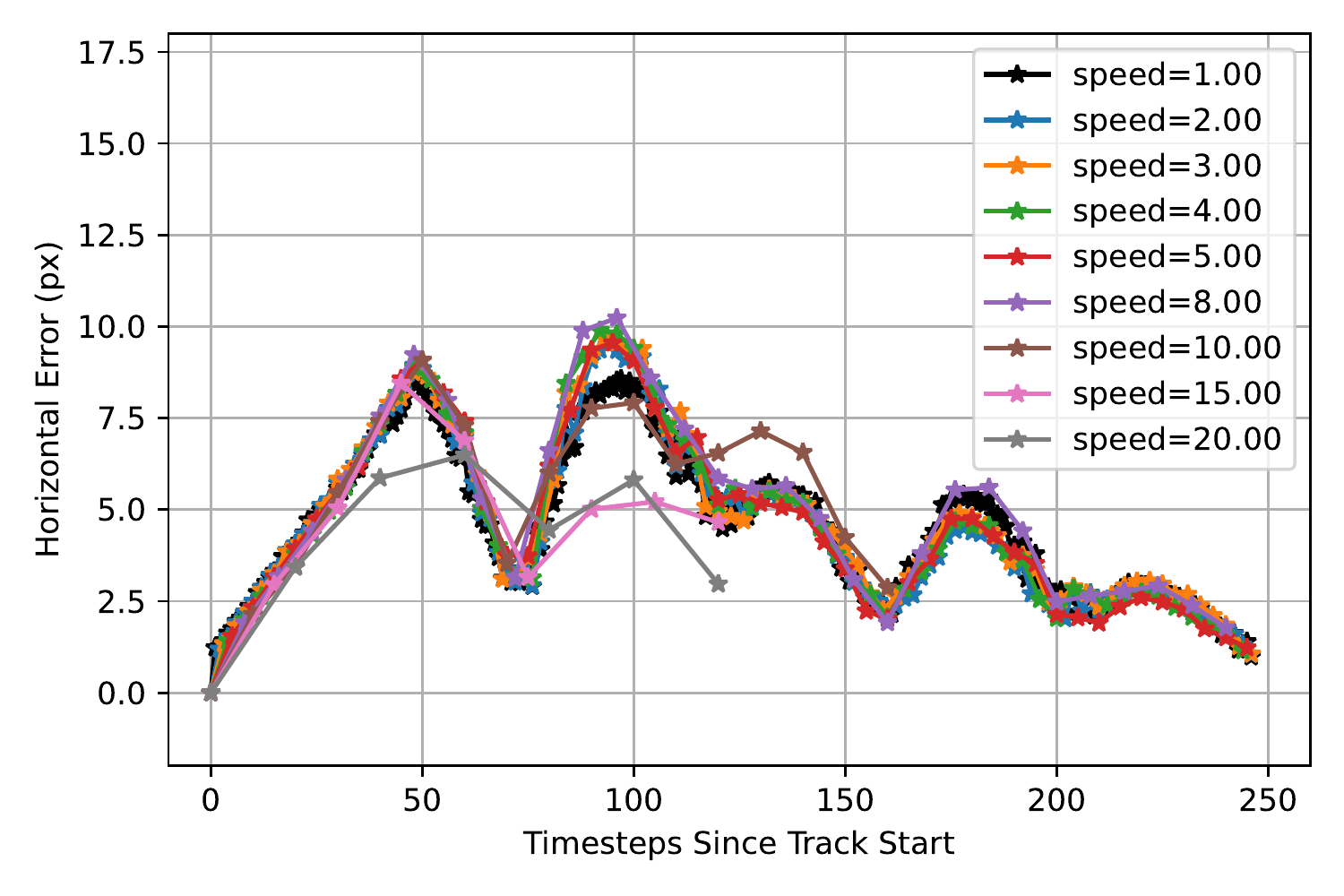}
        \includegraphics[width=0.48\textwidth]{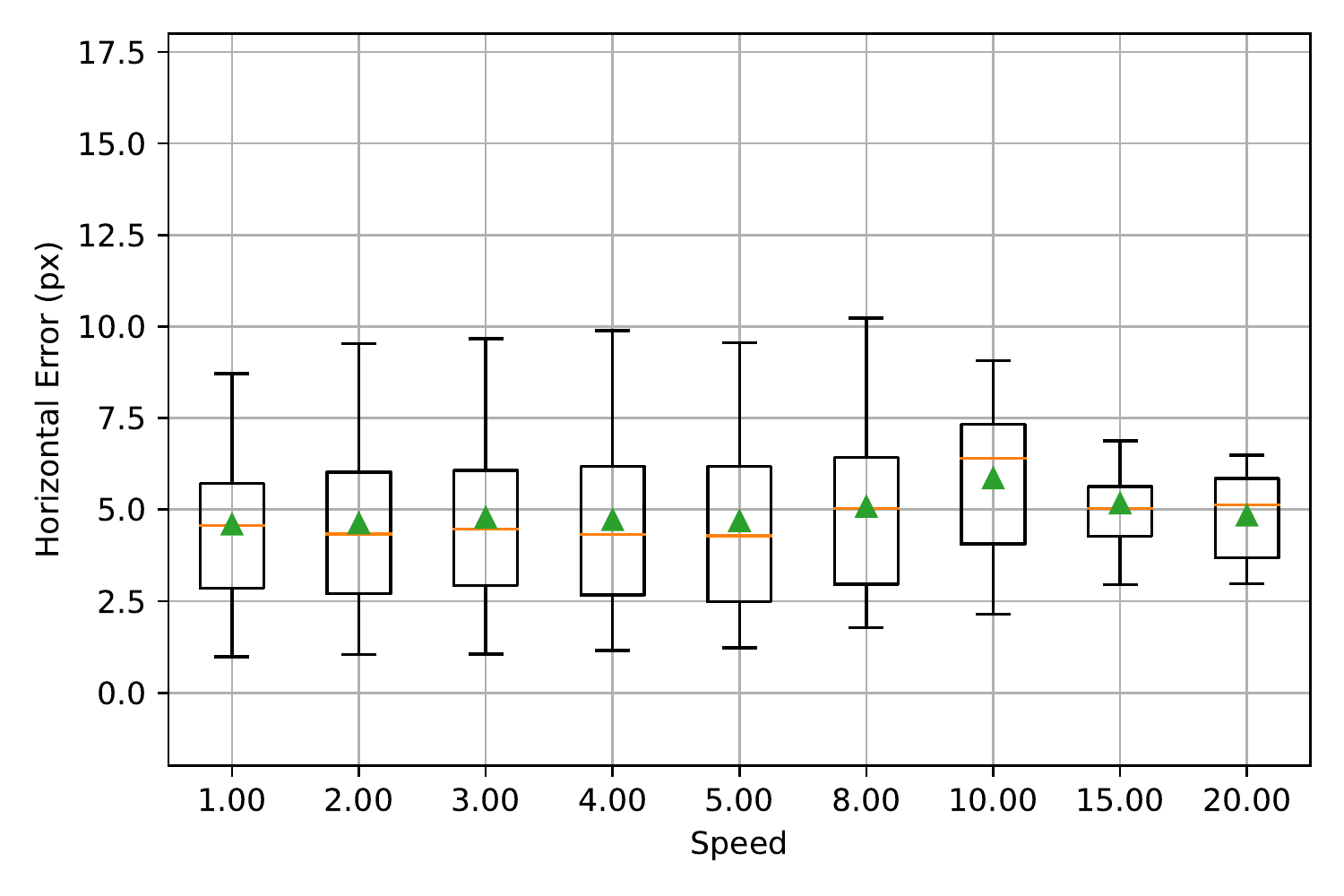}
    }
    \subfigure[$\eta(t)$, Vertical Coordinate]{
        \includegraphics[width=0.48\textwidth]{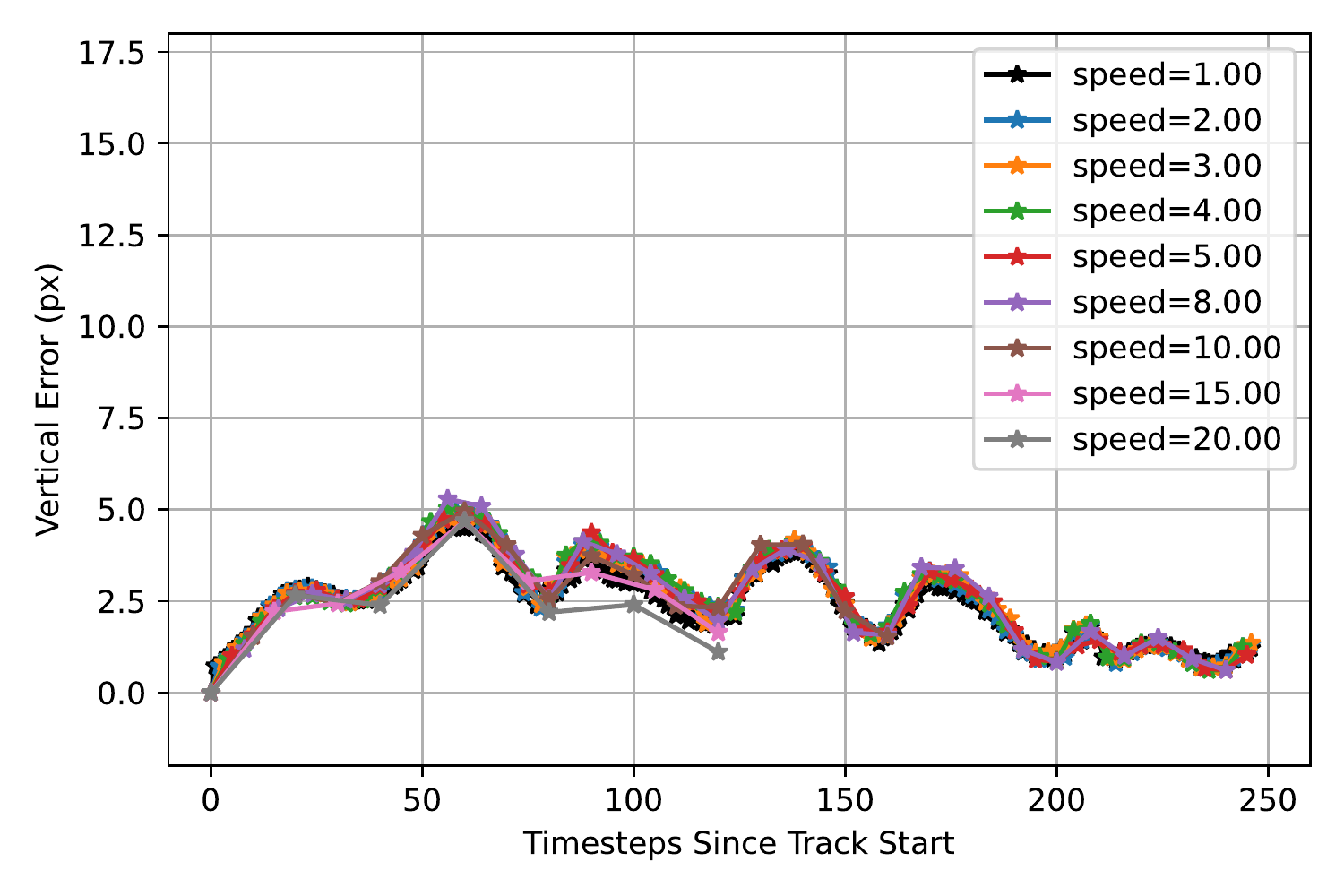}
        \includegraphics[width=0.48\textwidth]{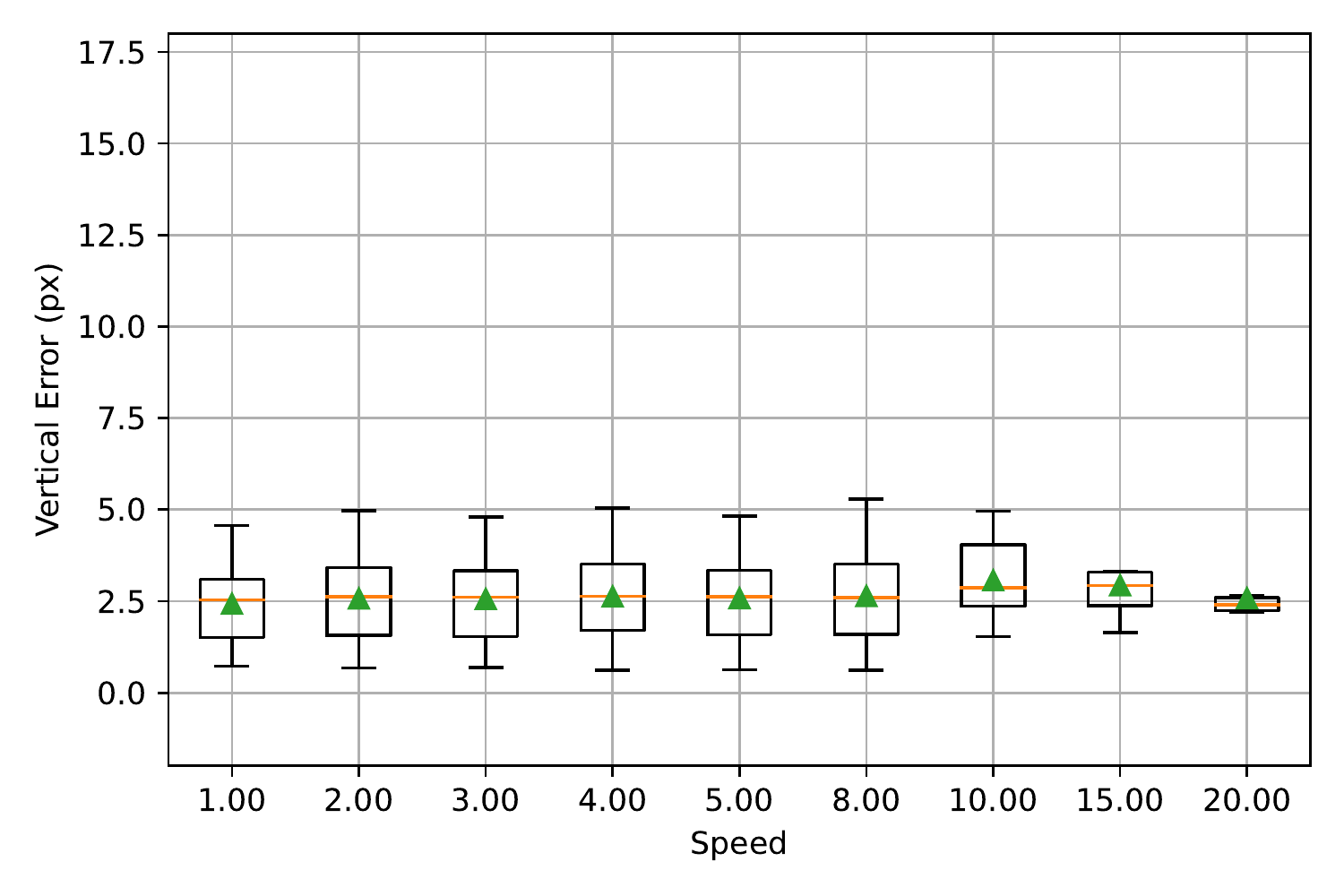}
    }
    \caption{\textbf{Gazebo AR/VR Dataset: Mean absolute errors are not affected by speed when using the Lucas-Kanade Tracker.}
    The left column contains plots of the horizontal (top row) and vertical (bottom row) components of the mean absolute error $\eta(t)$ at each timestep $t$ after initial feature detection at multiple speeds. Each dot corresponds to a processed frame; lines for higher speeds contain data from fewer frames and therefore show fewer dots. The right column plots the ordinate values of each line for $t>0$ in the left figures as a box plot: means are shown as green triangles and medians are shown as orange lines.
    The left-column plots show that that lines of $\eta(t)$ for different speeds are largely on top of one another. Some of the lines at higher speeds (brown, pink, and gray) have lower mean absolute errors than the lines at lower speeds, indicating that the number of frames, and not just absolute distance, also affects the total drift.
    }
    \label{fig:gazebo_arvr_LK_MAE}
\end{figure}

\begin{figure}[H]
    \centering
    \subfigure[$\Phi(t)$, Horizontal Coordinate]{
        \includegraphics[width=0.48\textwidth]{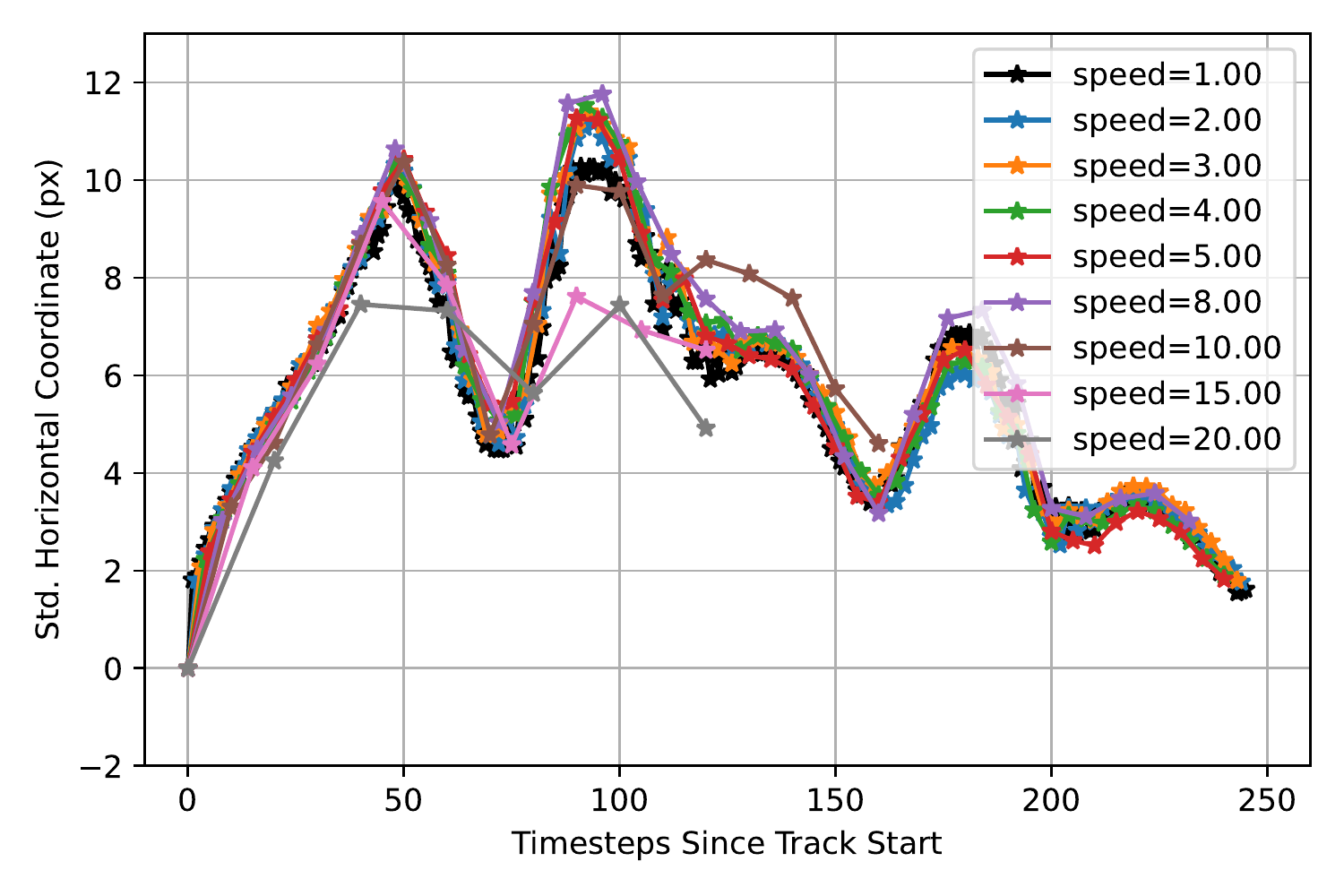}
        \includegraphics[width=0.48\textwidth]{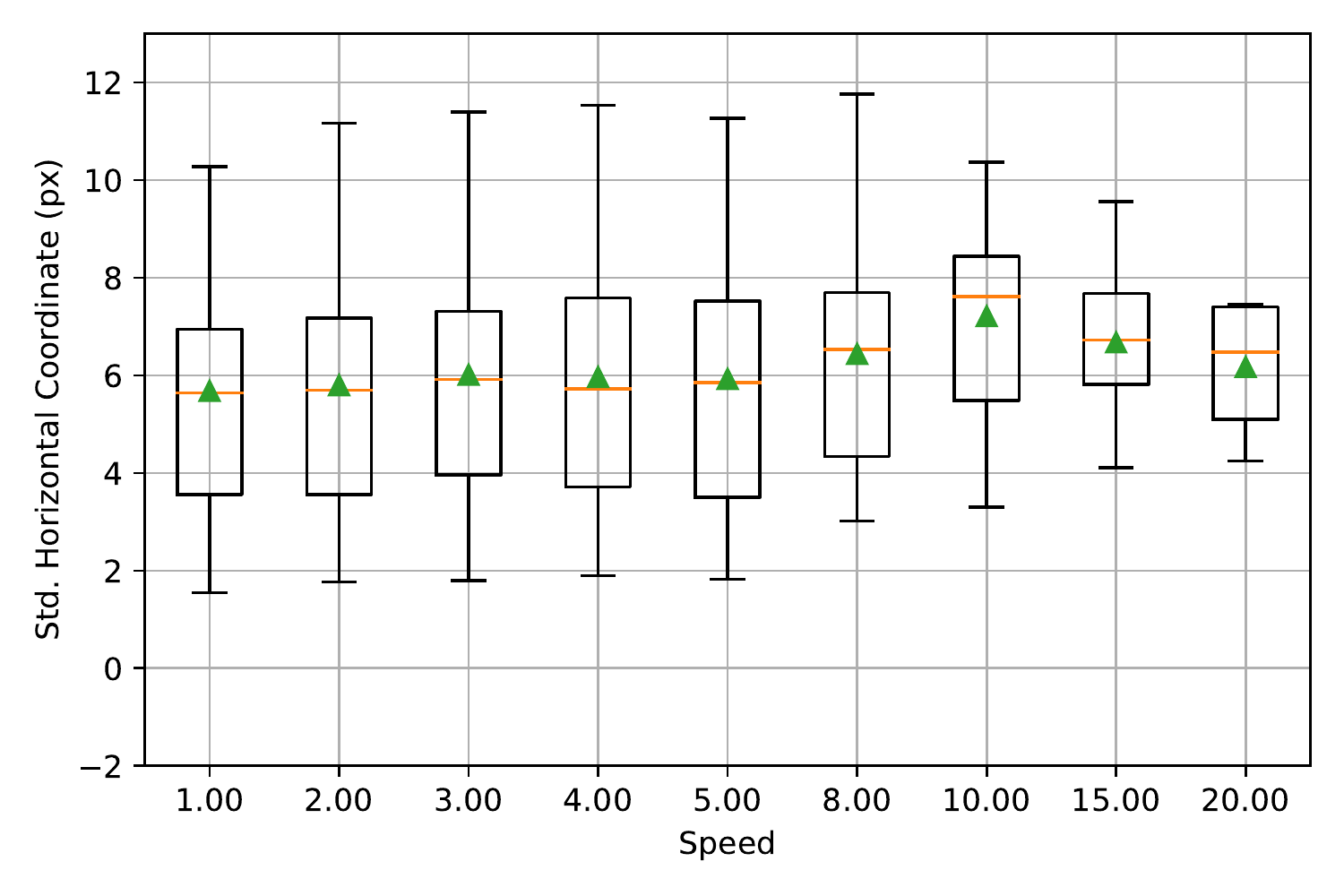}
    }
    \subfigure[$\Phi(t)$, Vertical Coordinate]{
        \includegraphics[width=0.48\textwidth]{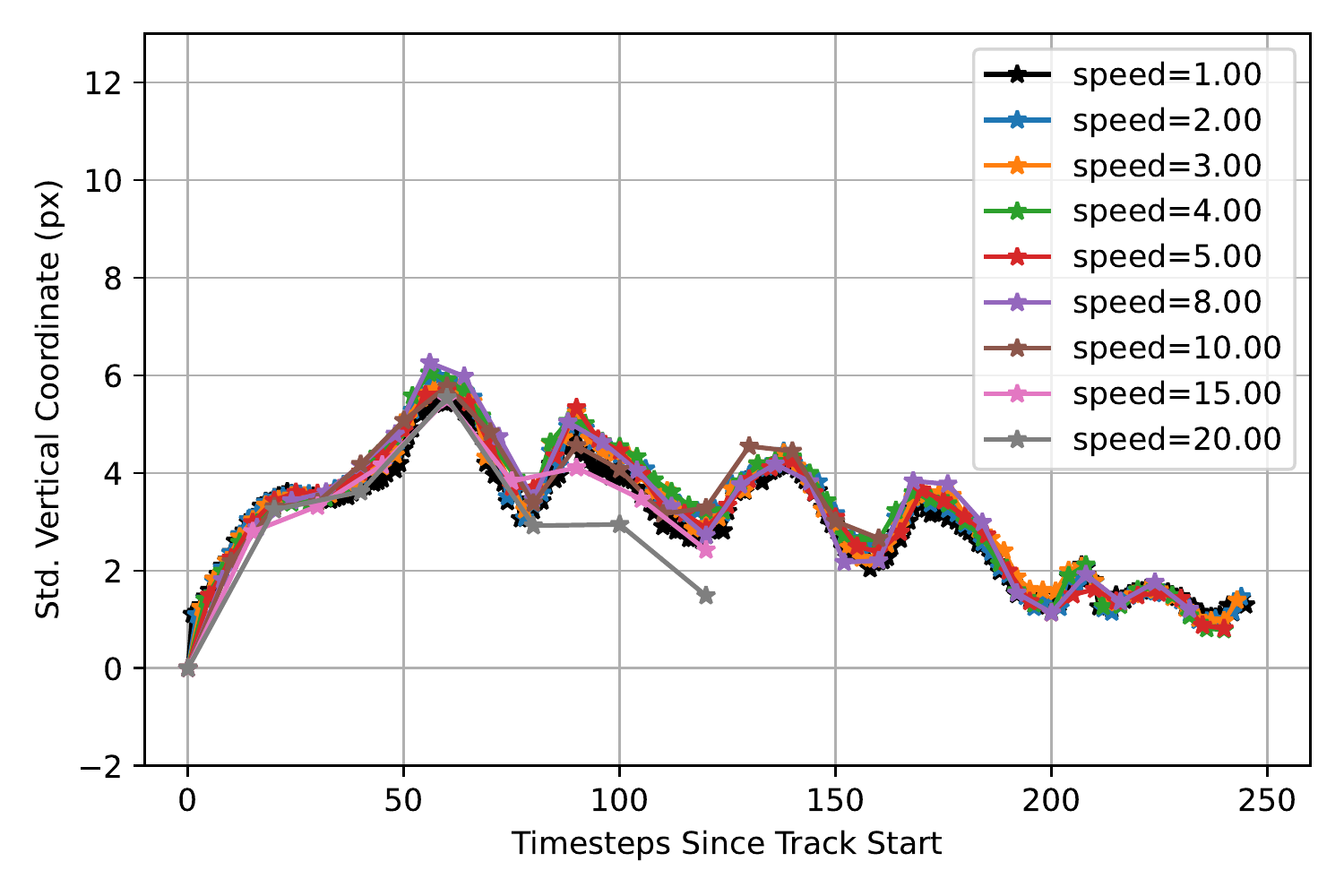}
        \includegraphics[width=0.48\textwidth]{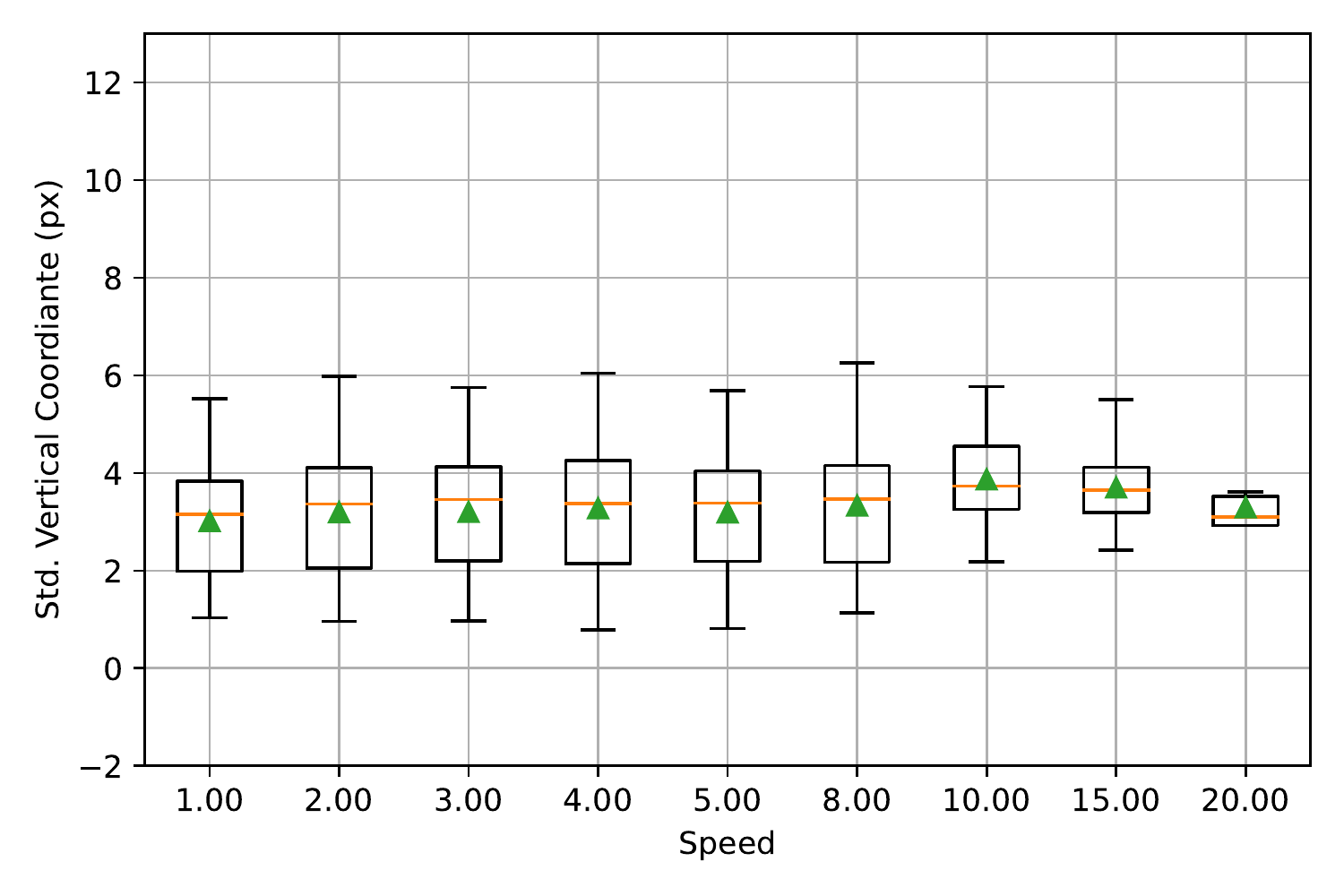}
    }
    \caption{\textbf{Gazebo AR/VR Dataset: Covariance is not affected by speed when using the Lucas-Kanade Tracker.}
    The left column contains plots of the horizontal (top row) and vertical (bottom row) components of the covariance $\Phi(t)$ at each timestep $t$ after initial feature detection at multiple speeds. Each dot corresponds to a processed frame; lines for higher speeds contain data from fewer frames and therefore show fewer dots. The right column plots the ordinate values of each line for $t>0$ in the left figures as a box plot: means are shown as green triangles and medians are shown as orange lines.
    The left-column plots show that that lines of $\eta(t)$ for different speeds are largely on top of one another. The right-column plots show that the corresponding box plots remain similar until speed is increased to 10.00. Then, variation in $\Phi(t)$ (i.e., the height of the boxes) shrinks because there are fewer points over time. The location of the green triangles in the box plots also changes because there are fewer points with low values of $t$ and low-errors contributing to it. Some of the lines at higher speeds (brown, pink, and gray) show less covariance than the lines at lower speeds, indicating that the number of frames, and not just absolute distance, also affects the total drift.
    }
    \label{fig:gazebo_arvr_LK_cov}
\end{figure}

\begin{figure}[H]
    \centering
    \subfigure[$\nu(t)$, Horizontal Coordinate]{
        \includegraphics[width=0.48\textwidth]{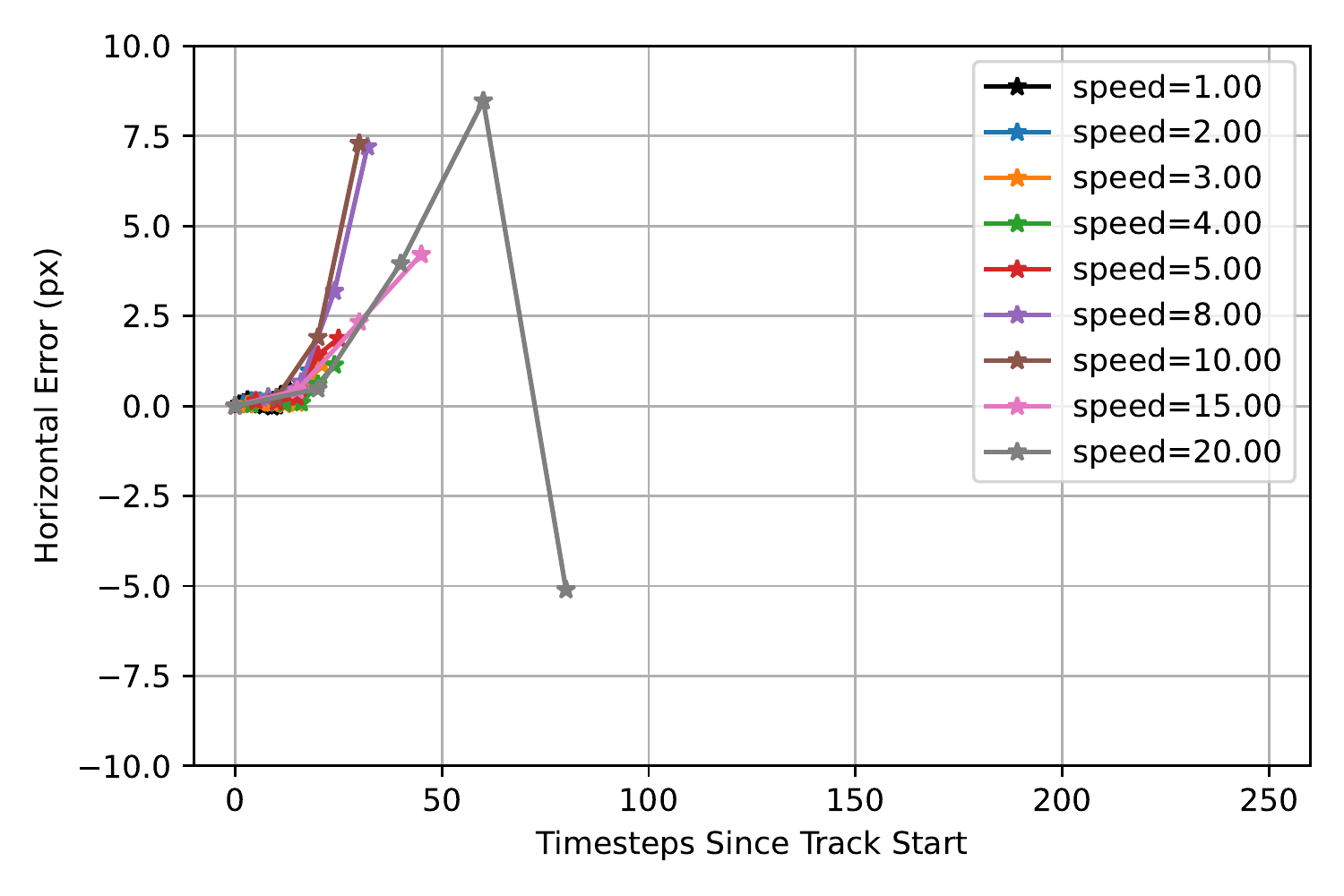}
        \includegraphics[width=0.48\textwidth]{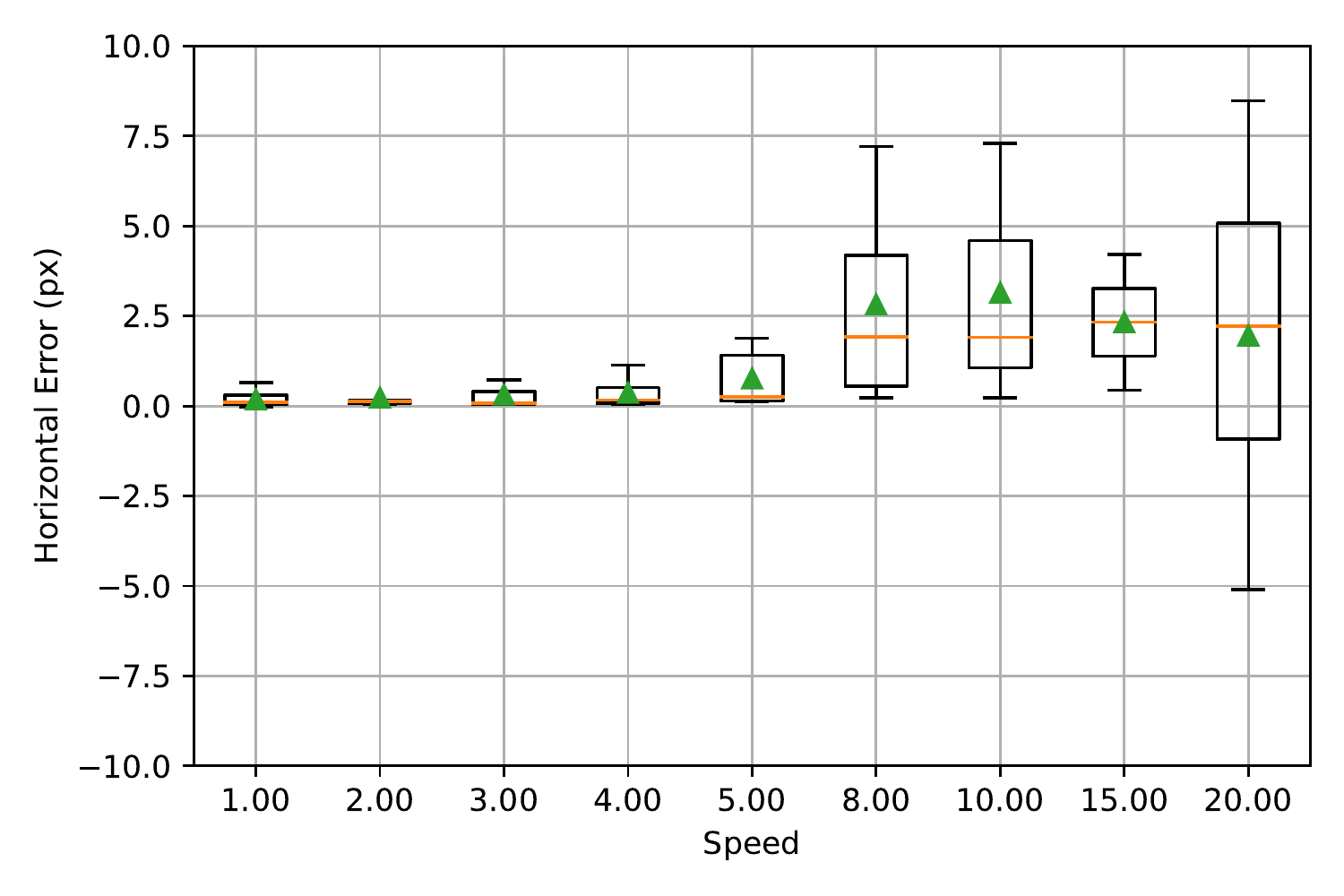}
    }
    \subfigure[$\nu(t)$, Vertical Coordinate]{
        \includegraphics[width=0.48\textwidth]{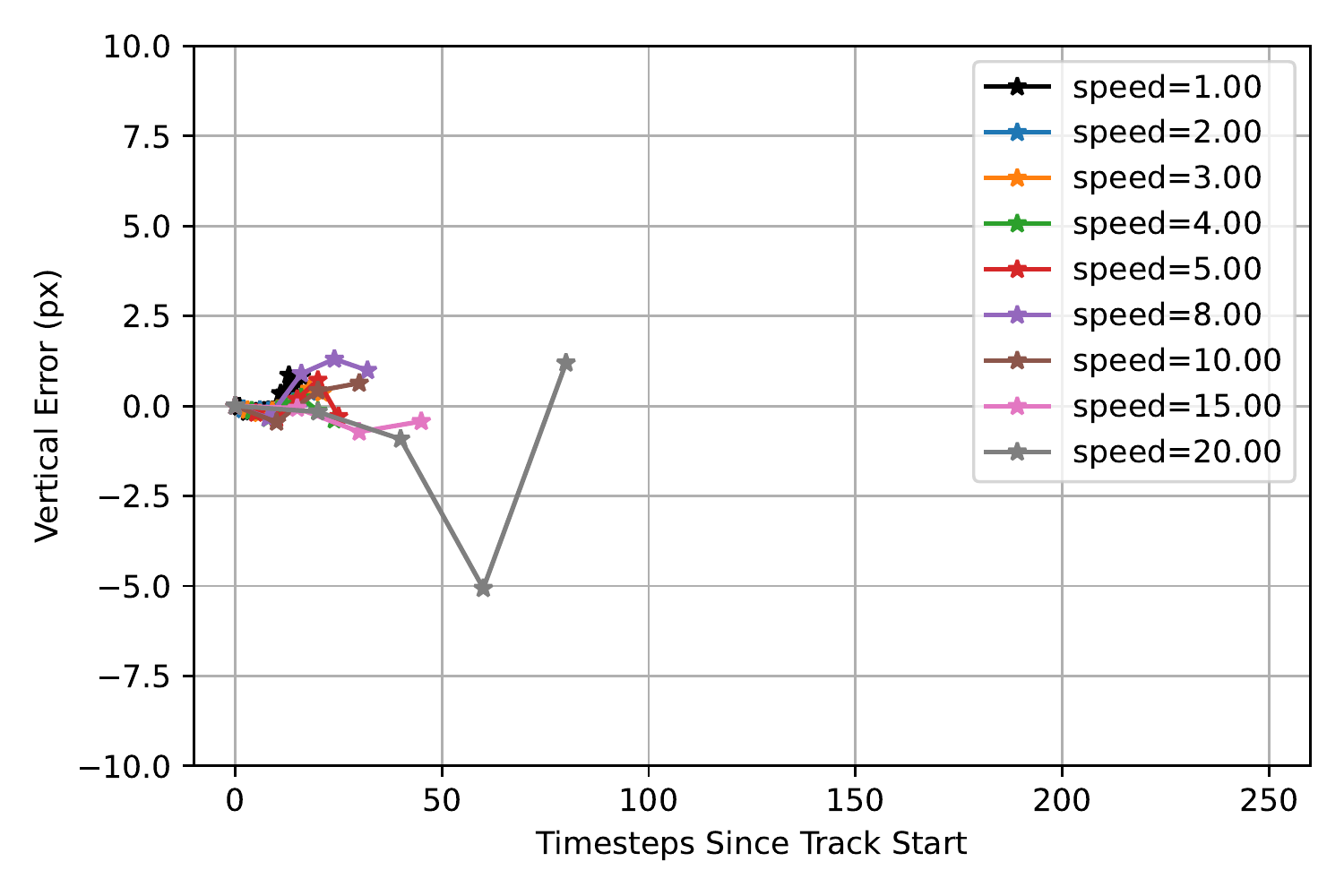}
        \includegraphics[width=0.48\textwidth]{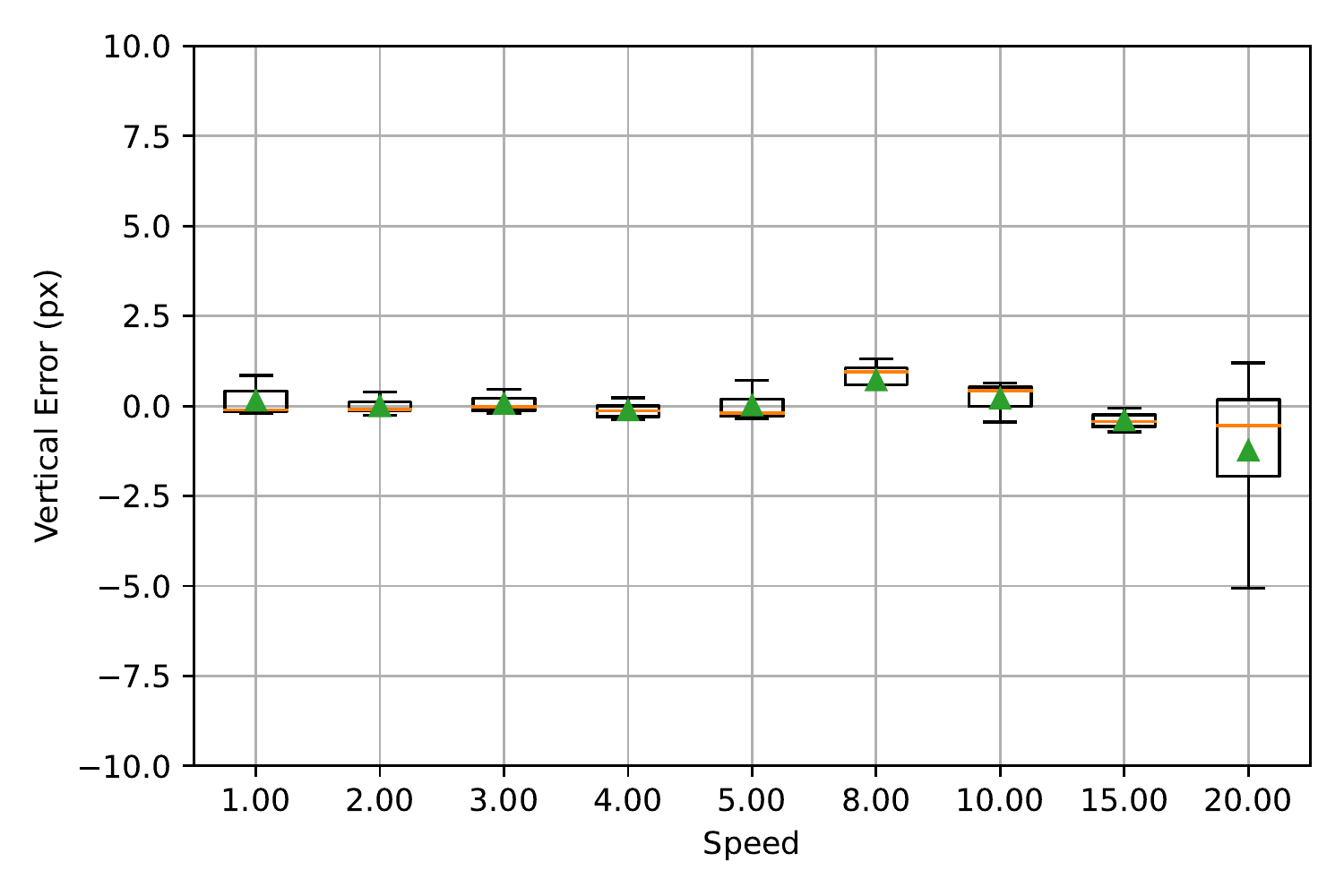}
    }
    \caption{\textbf{Gazebo AR/VR Dataset: Mean errors are unaffected by speed when using the Correspondence Tracker until tracking failure occurs.}
    The left column contains plots of the horizontal (top row) and vertical (bottom row) components of the mean tracking error $\nu(t)$ at each timestep $t$ after initial feature detection at multiple speeds. Each dot corresponds to a processed frame; lines for higher speeds contain data from fewer frames and therefore show fewer dots. The right column plots the ordinate values of each line for $t>0$ in the left figures as a box plot: means are shown as green triangles and medians are shown as orange lines. 
    In both the horizontal and vertical coordinates, mean errors over time are largely the same until speed=5.00. Then, tracking failures cause larger errors.
    }
    \label{fig:gazebo_arvr_match_meanerror}
\end{figure}

\begin{figure}[H]
    \centering
    \subfigure[$\eta(t)$, Horizontal Coordinate]{
        \includegraphics[width=0.48\textwidth]{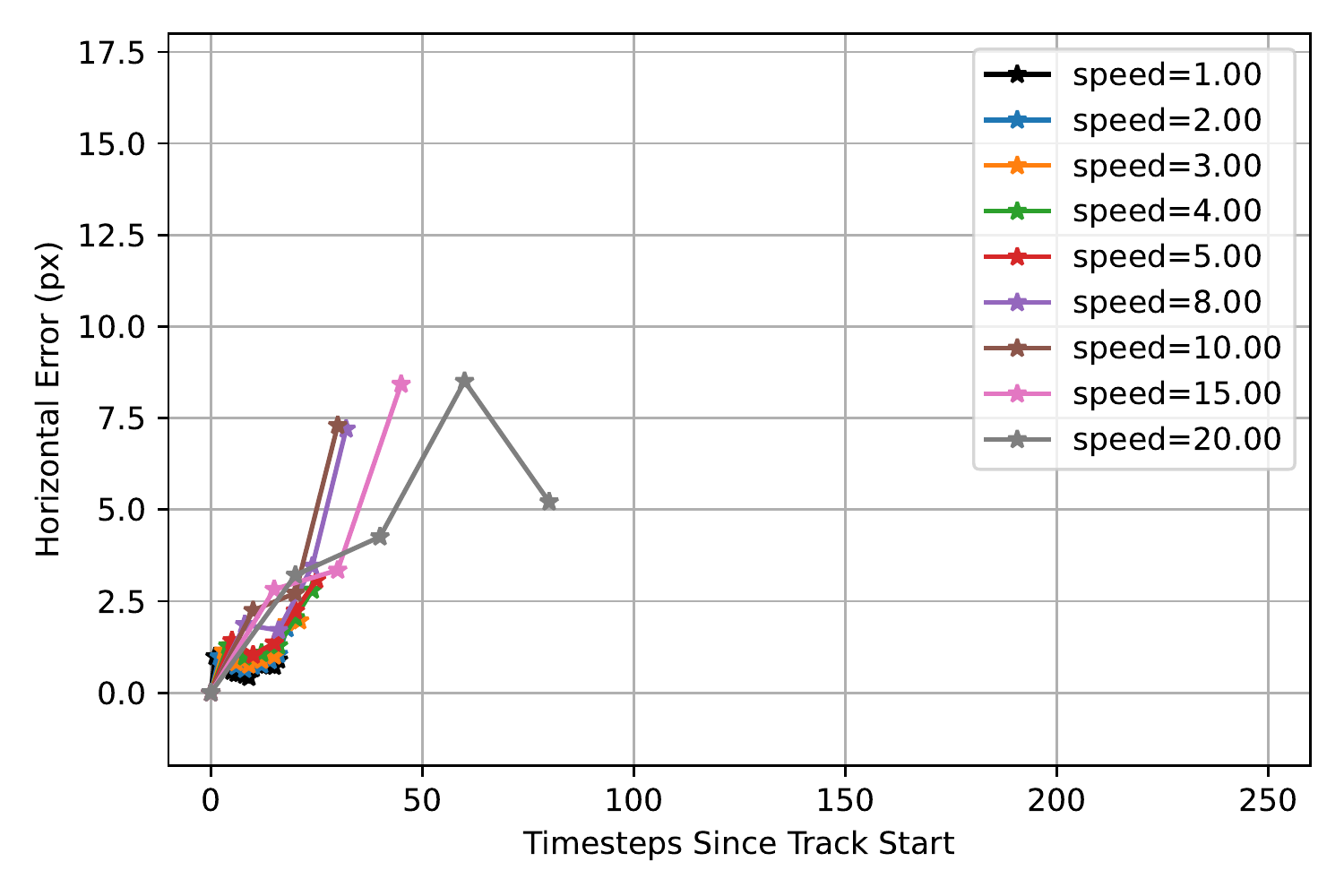}
        \includegraphics[width=0.48\textwidth]{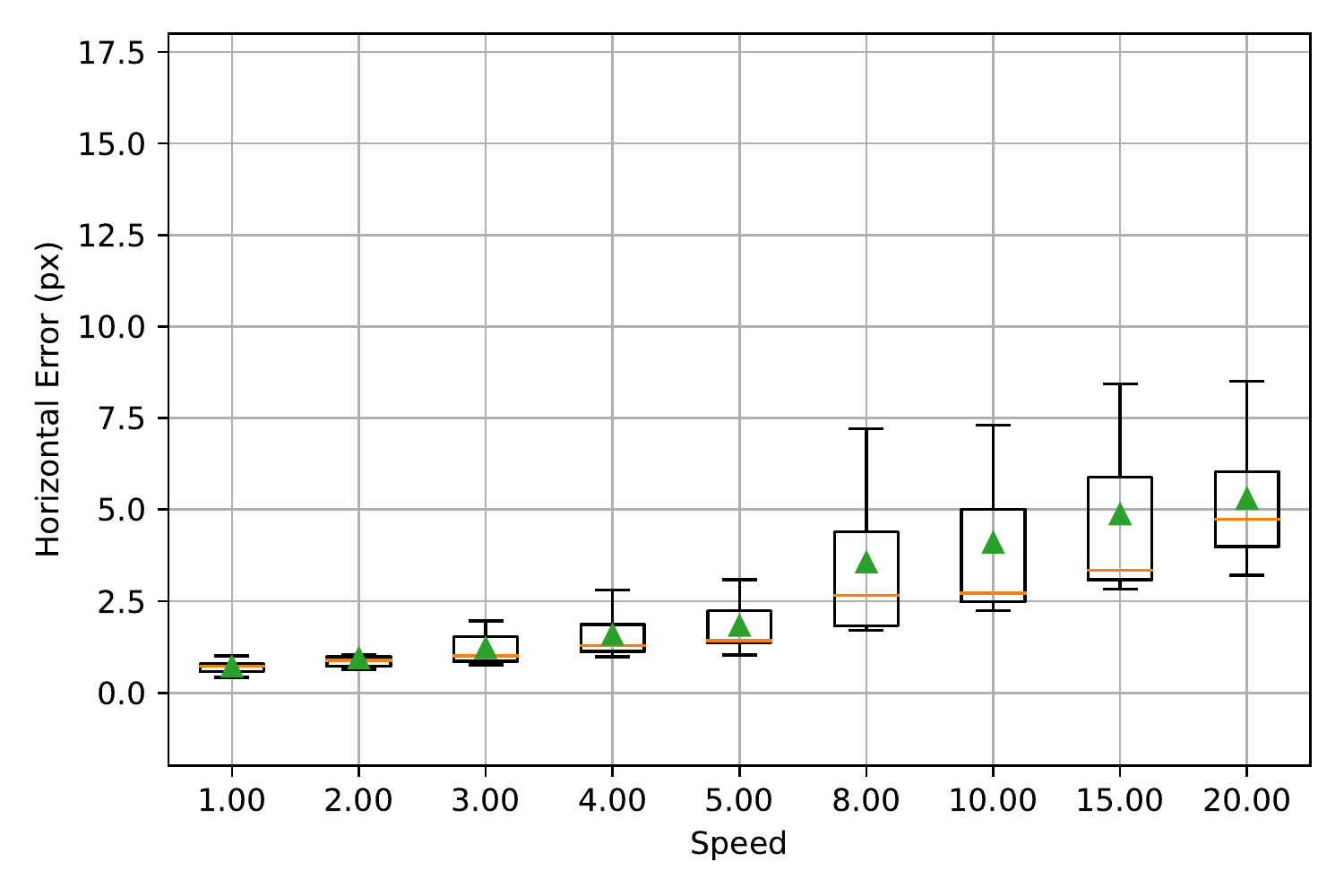}
    }
    \subfigure[$\eta(t)$, Vertical Coordinate]{
        \includegraphics[width=0.48\textwidth]{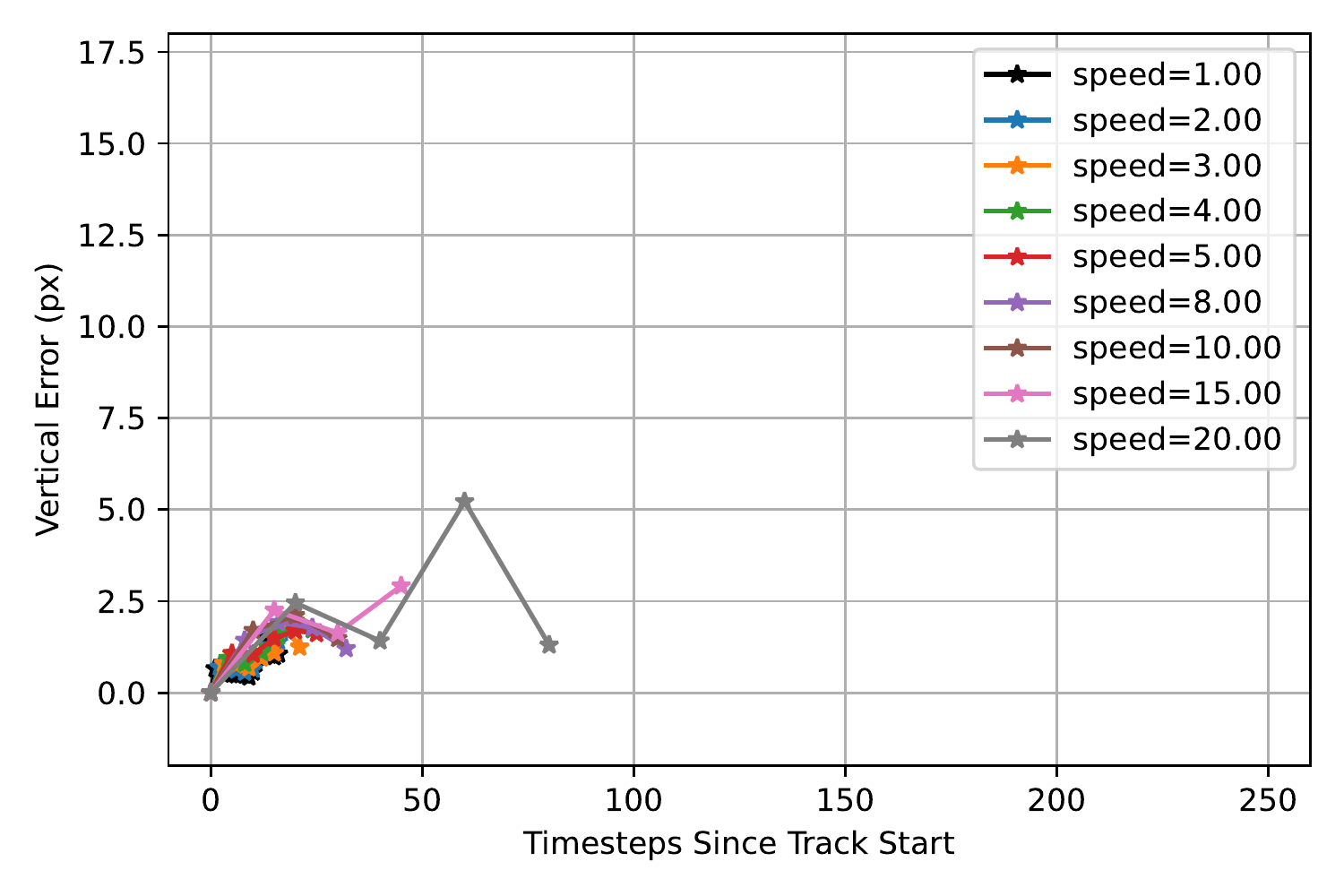}
        \includegraphics[width=0.48\textwidth]{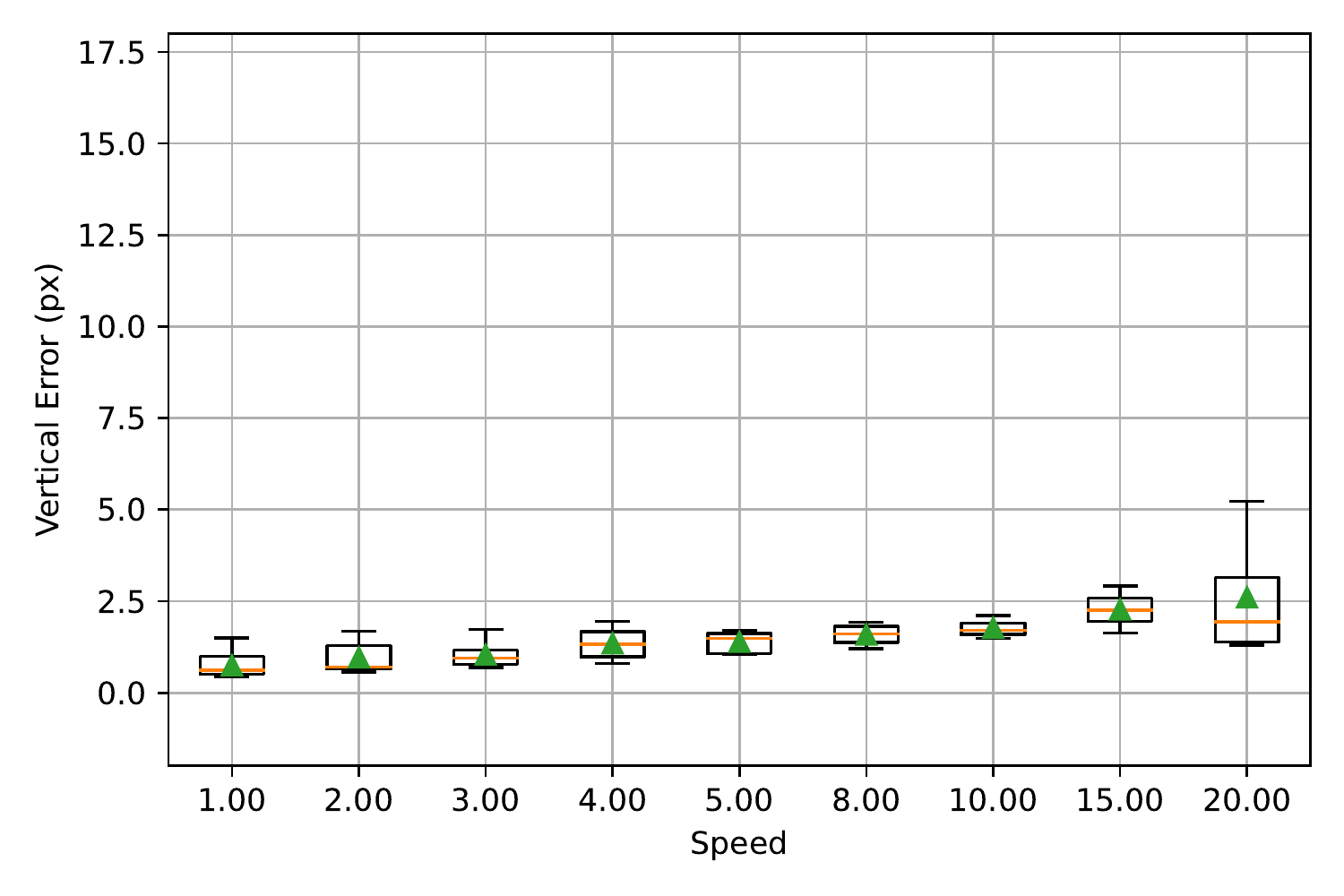}
    }
    \caption{\textbf{Gazebo AR/VR Dataset: Mean absolute errors increase with speed when using the Correspondence Tracker.}
    The left column contains plots of the horizontal (top row) and vertical (bottom row) components of the mean absolute error $\eta(t)$ at each timestep $t$ after initial feature detection at multiple speeds. Each dot corresponds to a processed frame; lines for higher speeds contain data from fewer frames and therefore show fewer dots. The right column plots the ordinate values of each line for $t>0$ in the left figures as a box plot: means are shown as green triangles and medians are shown as orange lines. For both the horizontal and vertical coordinates, there is a steady rise in the mean absolute error in the right column plots.
    }
    \label{fig:gazebo_arvr_match_MAE}
\end{figure}

\begin{figure}[H]
    \centering
    \subfigure[$\Phi(t)$, Horizontal Coordinate]{
        \includegraphics[width=0.48\textwidth]{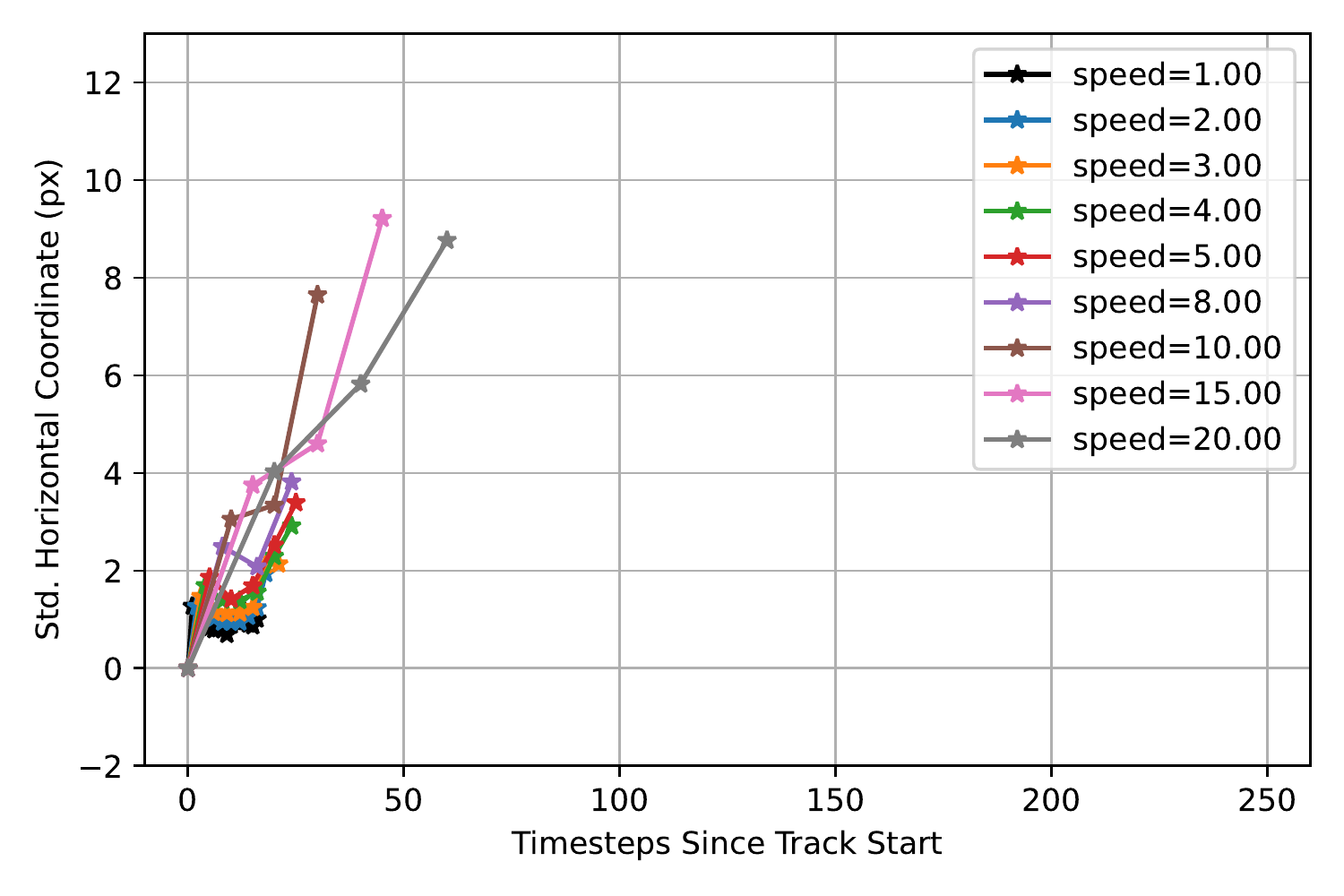}
        \includegraphics[width=0.48\textwidth]{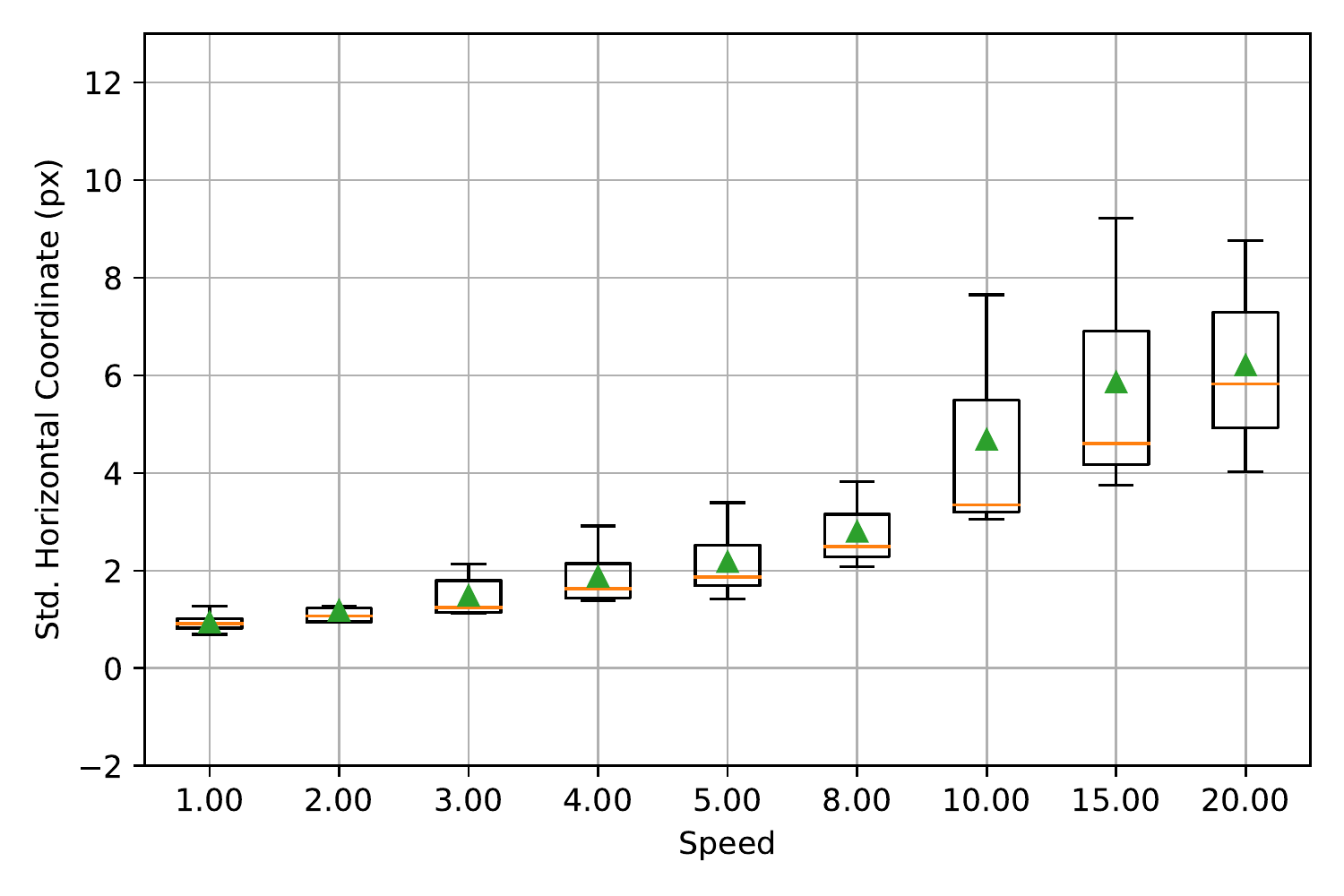}
    }
    \subfigure[$\Phi(t)$, Vertical Coordinate]{
        \includegraphics[width=0.48\textwidth]{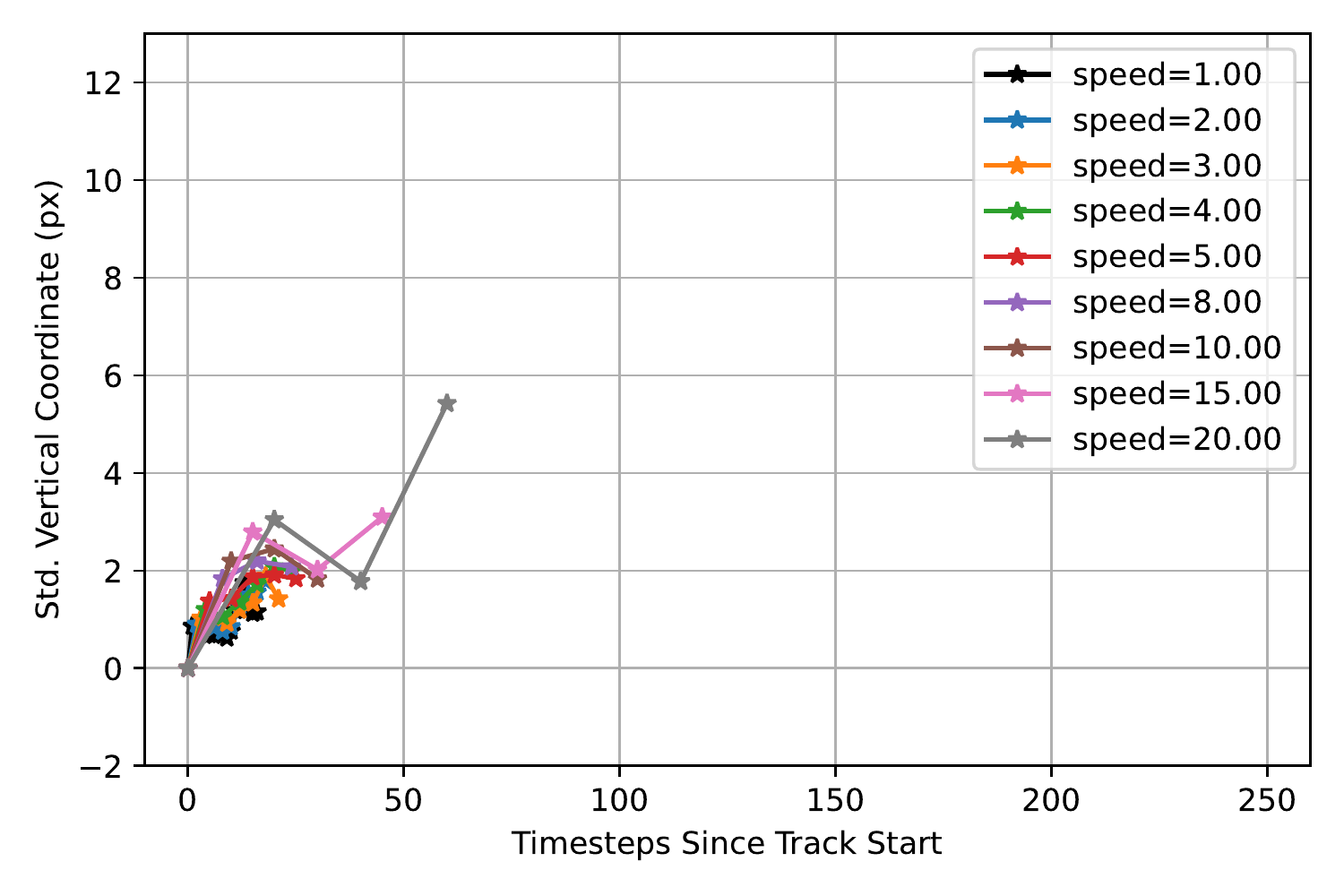}
        \includegraphics[width=0.48\textwidth]{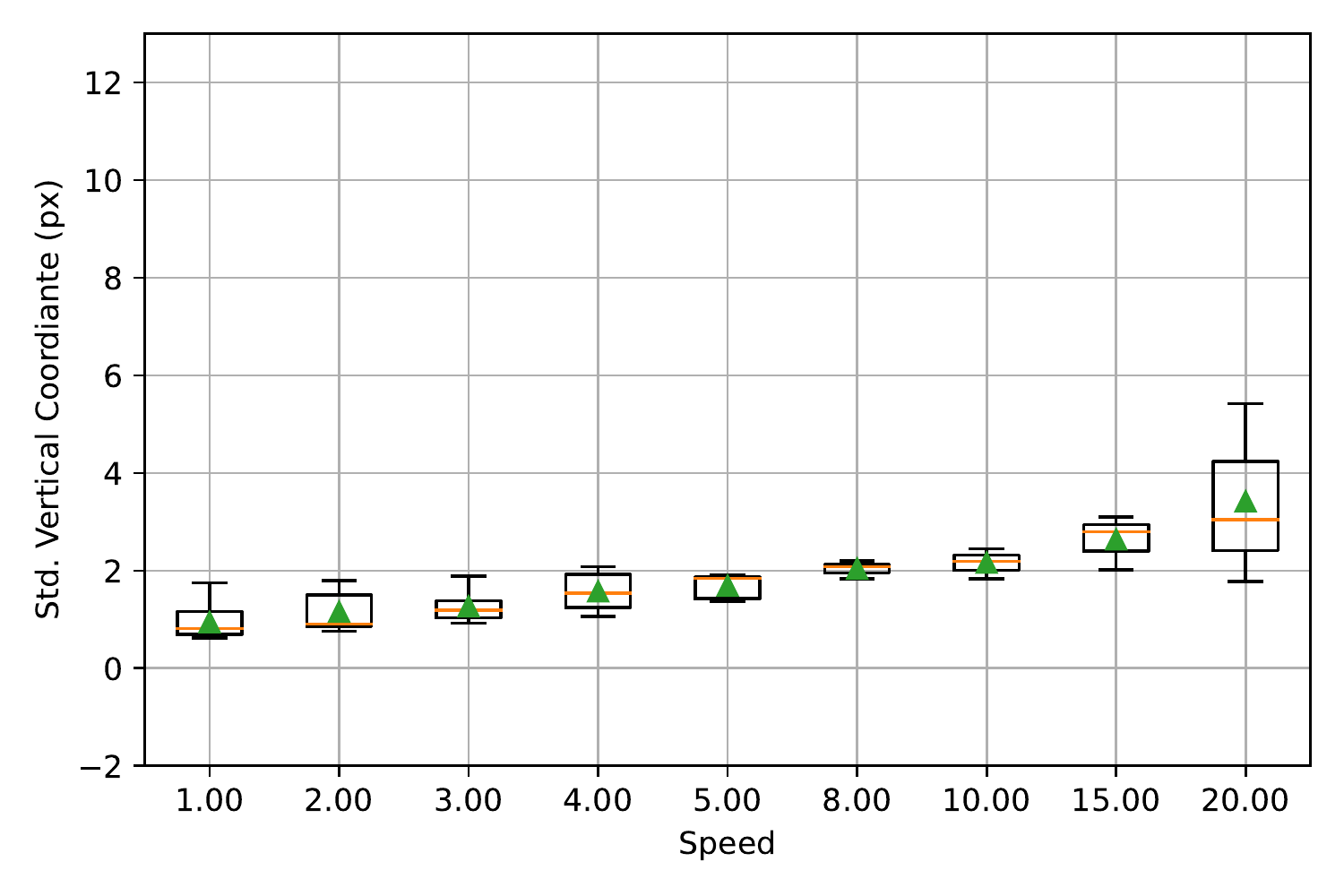}
    }
    \caption{\textbf{Gazebo AR/VR Dataset: Covariance increases with speed when using the Correspondence Tracker.}
    The left column contains plots of the horizontal (top row) and vertical (bottom row) components of the mean tracking error $\Phi(t)$ at each timestep $t$ after initial feature detection at multiple speeds. Each dot corresponds to a processed frame; lines for higher speeds contain data from fewer frames and therefore show fewer dots. The right column plots the ordinate values of each line for $t>0$ in the left figures as a box plot: means are shown as green triangles and medians are shown as orange lines. For both the horizontal and vertical coordinates, there is a steady rise in the covariance in the right column plots.
    }
    \label{fig:gazebo_arvr_match_cov}
\end{figure}

\bibliographystyle{plain}
\bibliography{feature_tracker_uq/references.bib}

\end{document}